\documentclass[12pt]{puthesis}
\newcommand{\proquestmode}{}

\title{Reinforcement Learning: From Algorithms To Foundation Models}

\submitted{May 2026}  
\copyrightyear{2026}  
\author{Zihan Ding}
\adviser{Professor Chi Jin}  
\department{Electrical Computer Engineering}

\usepackage{graphicx}

\usepackage{verbatim}

\usepackage{multirow}
\usepackage{longtable}

\usepackage{booktabs}
\usepackage{tabularray}
\UseTblrLibrary{booktabs}
\usepackage{tikz}
\usetikzlibrary{arrows.meta, positioning}

\usepackage{adjustbox}
\usepackage{algorithm}
\usepackage[noend]{algorithmic}
\usepackage{amsfonts}				
\usepackage{amsmath}
\usepackage{amsthm}
\usepackage{amssymb}
\usepackage[page,toc,titletoc,title]{appendix}
\usepackage[english]{babel} 		
\usepackage{bm}						
\usepackage{booktabs}
\usepackage{caption}
\usepackage{changepage,threeparttable} 
\usepackage{csvsimple}
\usepackage{etoolbox}               
\usepackage{fancyhdr}
\usepackage{float}
\usepackage{graphicx}
\usepackage[hidelinks]{hyperref}
\usepackage{mathabx}				
\usepackage{mathtools}
\usepackage{minted}
\usepackage{multirow}
\usepackage[super]{nth}						                
\usepackage[square,authoryear]{natbib}
\let\cite\citep
\usepackage{pifont}
\usepackage{rotating}
\usepackage{setspace}
\usepackage{siunitx}  				
\usepackage{soul}                   
\usepackage{subcaption}
\usepackage{tabularx}               
\usepackage{textcomp}				
\usepackage[table]{xcolor}
\usepackage{tikz}
\usepackage{titlesec}
\usepackage[nottoc]{tocbibind}					
\usepackage{tocloft}
\usepackage{url}
\usepackage{xfrac}					
\usepackage{xpatch}
\usepackage{xspace}
\usepackage[nameinlink, capitalise]{cleveref}
\usepackage{latexsym}
\usepackage{relsize}
\usepackage{enumerate}
\usepackage{dsfont}
\usepackage{algorithm}
\usepackage[left=1.5in, right=1in, top=1in, bottom=1in]{geometry}
\usepackage{algorithmic}
\usepackage{bbm}
\usepackage{framed}
\usepackage{color}
\usepackage{makecell}
\usepackage{nicefrac}  
\usepackage{diagbox}
\usepackage{wrapfig}
\usepackage{mdframed}
\usepackage{subcaption}
\usepackage{tablefootnote}
\usepackage{scrextend}
\usepackage{bbm}
\usepackage{mathabx}
\usepackage{quoting}
\usepackage[T1]{fontenc}
\usepackage{enumitem}

\newcommand{\nd}{\textit{Nash-DQN}\xspace}
\newcommand{\nde}{\textit{Nash-DQN-Exploiter}\xspace}
\newcommand{\nv}{\textit{Nash-VI}\xspace}
\newcommand{\nve}{\textit{Nash-VI-Exploiter}\xspace}
\newcommand{\golf}{\textit{Golf-with-Exploiter}\xspace}
\newcommand{\minibatch}{{\mathcal{M}}}
\newcommand{\reb}[1]{{\color{black}{#1}}}
\newtheorem{theorem}{Theorem}
\newtheorem{lemma}[theorem]{Lemma}
\newtheorem{corollary}[theorem]{Corollary}
\newtheorem{remark}[theorem]{Remark}

\newtheorem{proposition}[theorem]{Proposition}

\theoremstyle{definition}
\newtheorem{definition}[theorem]{Definition}

\newtheorem{assumption}{Assumption}
\newcommand{\alglinelabel}{%
  \addtocounter{ALC@line}{-1}
  \refstepcounter{ALC@line}
  \label
}
\newcommand\benchname{FightLadder}
\newcommand{\ie}{\textit{i.e.}\ }
\newcommand{\eg}{\textit{e.g.}\ }

\newcommand{\etc}{etc.\ }

  \newcommand{\INPUT}{\item[\algorithmicinput]}

\newcommand{\algorithmicinput}{\textbf{input}}

\ifdefined\printmode

\usepackage{url}

\else

\ifdefined\proquestmode

\usepackage{hyperref}
\hypersetup{bookmarksnumbered}

\makeatletter
\hypersetup{pdftitle=\@title,pdfauthor=\@author}
\makeatother

\else

\usepackage{hyperref}
\hypersetup{colorlinks,bookmarksnumbered}

\makeatletter
\hypersetup{pdftitle=\@title,pdfauthor=\@author}
\makeatother

\fi 
\fi 

\ifodd 0
\renewcommand{\maketitlepage}{}
\renewcommand*{\makecopyrightpage}{}
\renewcommand*{\makeabstract}{}

\else

\newcommand{\brac}[1]{{\left[ #1 \right]}}
\newcommand{\set}[1]{{\left\{ #1 \right\}}}

\newcommand{\defeq}{\mathrel{\mathop:}=}

\newcommand{\mat}[1]{\ensuremath{\mathbf{#1}}}

\newcommand{\argmin}{\mathop{\rm argmin}}

\newcommand{\trans}{^{\top}}

\newcommand{\norm}[1]{\|{#1} \|}

\newcommand{\E}{\mathbb{E}}
\newcommand{\D}{\mathbb{D}}
\renewcommand{\P}{\mathbb{P}}

\newcommand{\R}{\mathbb{R}}

\newcommand{\A}{\mat{A}}

\newcommand{\cM}{\mathcal{M}}

\newcommand{\cD}{\mathcal{D}}

\newcommand{\cS}{\mathcal{S}}
\newcommand{\cA}{\mathcal{A}}
\newcommand{\cB}{\mathcal{B}}

\newcommand{\eps}{\varepsilon}

\newenvironment{proof-sketch}{\noindent{\bf Proof Sketch}
  \hspace*{1em}}{\qed\bigskip\\}
\newenvironment{proof-idea}{\noindent{\bf Proof Idea}
  \hspace*{1em}}{\qed\bigskip\\}
\newenvironment{proof-of-lemma}[1][{}]{\noindent{\bf Proof of Lemma {#1}}
  \hspace*{1em}}{\qed\bigskip\\}
\newenvironment{proof-of-proposition}[1][{}]{\noindent{\bf
    Proof of Proposition {#1}}
  \hspace*{1em}}{\qed\bigskip\\}
\newenvironment{proof-of-theorem}[1][{}]{\noindent{\bf Proof of Theorem {#1}}
  \hspace*{1em}}{\qed\bigskip\\}
\newenvironment{inner-proof}{\noindent{\bf Proof}\hspace{1em}}{
  $\bigtriangledown$\medskip\\}
\newenvironment{proof-attempt}{\noindent{\bf Proof Attempt}
  \hspace*{1em}}{\qed\bigskip\\}

\abstract{
\pagenumbering{arabic}
\setcounter{page}{3}
Reinforcement learning (RL) provides a general framework for sequential decision making under explicit objectives. In its classical form, RL studies how an agent should act to maximise long-term reward in a dynamic environment. In richer settings, however, the problem extends beyond a single agent and fixed environment: intelligent behavior may require strategic interaction with other agents, adaptation to uncertainty, and the ability to reason over complex, high-dimensional worlds. This thesis studies reinforcement learning from two complementary perspectives: RL algorithms in games, and RL in the era of foundation models.

The first part focuses on multi-agent reinforcement learning in games. It examines how incentives, policies, and equilibrium concepts interact in competitive and general-sum environments, spanning canonical two-player zero-sum games, large-scale video-game domains, and multi-player settings with more general strategic structure. Together, these works investigate both the algorithmic foundations of learning in multi-agent systems and the practical behavior of RL methods in complex interactive environments.

The second part studies reinforcement learning with generative and foundation models, motivated by the idea that broad prior knowledge can substantially enrich sequential decision making. In this setting, pretrained generative models and learned world models serve not only as representation tools, but also as structured priors for planning, control, and policy optimization. The thesis develops diffusion-based world models, investigates reinforcement learning for efficient video generation, explores expressive generative models as policy classes, and studies interactive video world models in which actions shape future observations. It further addresses long-horizon sequential modeling through architectures that incorporate memory into world modeling.

Taken together, these contributions present a unified view of reinforcement learning as the study of objective-driven adaptation in complex sequential domains. From strategic games to generative world models, the thesis highlights how reinforcement learning connects algorithmic decision making, environment modeling, and the emerging capabilities of large-scale foundation models, offering a broader perspective on the principles underlying intelligent behavior.

}

\acknowledgements{
Princeton is a place both profound and elegant—quiet in its outward form, yet endlessly rich in thought. It is a place made for study, for reflection, and for the slow shaping of one’s character. Over the past four years, it has witnessed not only my Ph.D. journey, but also a period of growth far beyond academics. What made this journey truly meaningful were the people I encountered along the way. With them, I shared not only the demands of research, but also conversations, friendship, and life itself. Their support and companionship sustained me through this long and often challenging path. I am also deeply grateful to the administrative staff of the Department of Electrical and Computer Engineering and across the Princeton campus. With patience, kindness, and care, they quietly supported us through the years, like devoted gardeners tending the ground on which others may grow.

Among my most cherished memories of Princeton are the evenings spent around a dinner table, where meals became occasions for thought as much as for rest. We brought riddles, brain teasers, half-formed ideas, and sudden curiosities to one another, and what began as casual gatherings often unfolded into lively exchanges of mind and imagination. Those moments reminded me that intellectual life does not belong only to offices, classrooms, or papers—it also lives in friendship, laughter, and conversation, and in the simple joy of thinking together.

I would like to express my deepest gratitude to my advisor, Professor Chi Jin, who has been the most important mentor in my Ph.D. journey. He stood with us through every stage of research: from the earliest brainstorming of ideas, to experiment design, project organization, and the many rounds of paper polishing that carried us through submission deadlines. From him, I learned what true rigor in research means—not only in broad vision, but in the precision of every detail, down to the value of a single parameter or the choice of a single word in a manuscript. He built for me a solid foundation in reinforcement learning and optimization, while continually opening promising directions for exploration. Even more importantly, he taught me how to recognize the taste of good research and how to judge the importance of a problem. Under his guidance, my way of thinking grew from theoretical understanding into a broader research capability that includes both conceptual depth and empirical judgment. At the same time, he gave me the freedom to explore independently and to grow into my own style as a researcher. His optimism, generosity, and kindness have also shaped the spirit of our group in lasting ways. I cannot fully express how much I appreciate his guidance and support.

I am also deeply thankful to the other members of my dissertation committee—Professor Benjamin Eysenbach, Professor Pramod Viswanath, and Professor Mengdi Wang. Each of them is a leading scholar in their respective field, and I have learned a great deal from their work, their insights, and their feedback. I feel very fortunate to have received their support and guidance throughout the dissertation process.

I would also like to thank my wonderful friends: Qinghua Liu, Xiao Ma, Zhou Lu, Yuanhao Wang, Ahmed Khaled, Wenzhe Li, Seth Karten, Yixiao Chen, Yuqi Nie, Andy Su, Allen Ren, Yuheng Zheng, Kexin Jin, Qinqin Nan Li, Anjian Li, Zhengyuan Shang, Yaqian Tang, Shouda Wang, Xinhao Liu, Yubin Lin, Yong Lin, Xiang Ji, Zixu Zhang, Cindy Wu, Qiang Zhang, Jiachen Wang, Kaixuan Huang, Tianle Cai, Eshaan Nichani, Hongjie Wang, Tong Wu, Cheng Gao, Jikai Hou, Chengzhuo Ni, Sulin Liu, Lihan Zha, Kaiqu Liang, Yulai Zhao, Beining Han, Yaqian Tang, Yuchen Liu, Zhiyang Yuan, Haoyu Wu, Haoyu Zhao, Jimmy Yang, Runzhe Wang, Yifan Li, Yue Wu, Xuezhou Zhang, Xiaonan Zhu, Chengshuai Shi, Xinran Liang, Ziran Yang, and Qinxin Yan. Thank you for the time we shared at Princeton, for your friendship, and for the many conversations, exchanges, and moments of mutual encouragement that made these years far richer than they otherwise would have been.

Finally, and most importantly, I want to thank my parents. Thank you for raising me, shaping me, and supporting me with unwavering love. Your influence has been present in everything I have done, and your support has been the foundation beneath every step of this journey. Whatever I have achieved, I owe first to you.
}

\dedication{The greatest mystery in human lies not in intelligence, \\but in humanity and consciousness.}

\fi  

\begin{document}

\makefrontmatter


\pagenumbering{arabic}
\setcounter{page}{16}
\part{Preclude} \label{part0:preclude}
\chapter{Introduction to Reinforcement Learning}
\section{Reinforcement Learning at AGI Era}
Reinforcement learning (RL) provides a general mathematical framework for designing incentive-driven decision-making systems. This perspective is increasingly important for artificial general intelligence (AGI), where an intelligent system must act over time under explicit objectives, adapt to changing circumstances, and coordinate with a complex environment. At a conceptual level, RL is also deeply connected to studies in human physiology and neuroscience, where dopamine-related signals are often interpreted as carrying information about reward, motivation, and learning. These connections suggest that incentive-based adaptation is not only an engineering principle, but also a fundamental mechanism for intelligent behaviour.

The central idea of RL is to specify what the agent should want, rather than prescribing exactly how it should behave. A desired outcome for the agent is encoded quantitatively by a \emph{reward function}, which evaluates the consequences of actions. The agent itself follows a decision-making rule, called a \emph{policy}, and learning consists of improving this policy so as to perform optimally with respect to the objective induced by the reward. In the simplest single-agent setting, optimality often means maximising expected cumulative reward. In richer multi-agent environments, however, the relevant notion of optimality may instead take the form of a Nash equilibrium or other equilibrium concepts, depending on how different agents interact and how incentives are coupled.

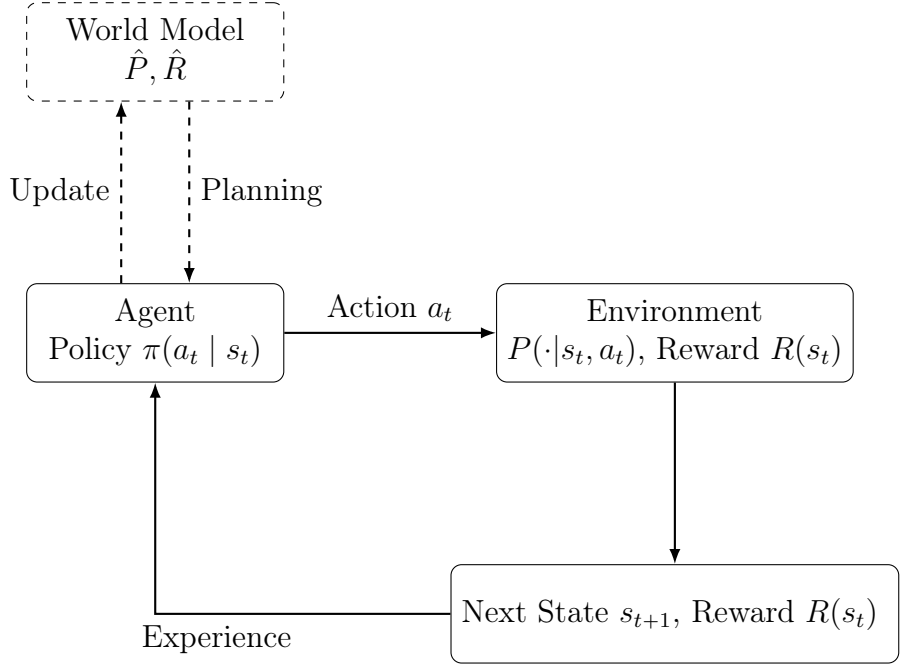
\begin{figure}[t]
    \centering
    \begin{tikzpicture}[
        node distance=2.4cm and 2.8cm,
        box/.style={
            draw,
            rounded corners,
            minimum width=3.4cm,
            minimum height=1.3cm,
            align=center
        },
        arrow/.style={
            -{Latex[length=2mm]},
            thick
        }
    ]

    \node[box] (agent) {Agent\\Policy $\pi(a_t\mid s_t)$};
    \node[box, dashed, above=of agent] (world) {World Model\\$\hat{P}, \hat{R}$};
    \node[box, right=of agent] (env) {Environment\\$P(\cdot|s_t, a_t)$, Reward $R(s_t)$};
    \node[box, below=of env] (feedback) {Next State $s_{t+1}$, Reward $R(s_t)$ };

    \draw[arrow, dashed] ([xshift=-0.45cm]agent.north) -- node[left] {Update} ([xshift=-0.45cm]world.south);
    \draw[arrow, dashed] ([xshift=0.45cm]world.south) -- node[right] {Planning} ([xshift=0.45cm]agent.north);
    \draw[arrow] (agent) -- node[above] {Action $a_t$} (env);
    \draw[arrow] (env) -- (feedback);
    \draw[arrow] (feedback.west) -| node[pos=0.25, below left] {Experience} (agent.south);

    \end{tikzpicture}
    \caption{A general reinforcement learning system: the agent interacts with the environment to optimize its policy for reward maximization, while optionally using a learned world model to predict dynamics and rewards, support planning, and improve policy optimisation.}
    \label{fig:rl_system}
\end{figure}

Although the reward function may be simple, the strategy that emerges from it can be highly nontrivial. This complexity is driven largely by the dynamic nature of the problem: actions affect future states, the environment may evolve stochastically, and in multi-agent settings the behaviours of other agents also shape the decision landscape. As a result, modelling the environment becomes a central issue. In classical RL, this is captured by the \emph{state transition model}, which describes how the system evolves in response to actions and uncertainty.

Recent progress in foundation models has opened a new direction for RL by enabling the injection of broad prior knowledge into the decision-making process. Instead of learning entirely from scratch, an agent can leverage pretrained representations and structured knowledge about the world to explore more sensibly and optimise its policy more sample efficiently and with better generalization. From this perspective, \emph{world modelling} can be viewed as environment modelling enriched by the capabilities of foundation models. A strong world model can support planning, improve sample efficiency, and better align optimisation with the agent's objective. At the same time, these benefits depend critically on the accuracy and computational efficiency of the world model itself, which makes the design of appropriate architectures and algorithms a central challenge.

\section{Single-Agent RL}
Through trial and error, an RL agent attempts to find the optimal policy to maximise its long-term reward. 
This process is  formulated by Markov Decision Processes. 

\subsection{Markov Decision Process}
\begin{definition}[Markov Decision Process]
An MDP can be described  by a tuple of key elements $ \langle \mathcal{S}, \mathcal{A}, P, R, \gamma  \rangle$:
\begin{itemize}
	\item $\mathcal{S}$: the set of environmental states.
		\item $\mathcal{A}$: the set of agent's possible actions.
	\item $P: \mathcal{S} \times \mathcal{A} \rightarrow \Delta(\mathcal{S})$: for each time step $t \in \mathcal{N}$, given agent's action $a\in \mathcal{A}$, the transition probability from a state $s \in \mathcal{S}$ to  the state in the next time step $ s' \in \mathcal{S}$.
	\item $R: \mathcal{S} \times \mathcal{A} \times \mathcal{S} \rightarrow \mathbb{R}$: the reward function that returns a scalar value to the agent for a transition from $s$ to $s'$ as a result of action $a$. The rewards have absolute values uniformly bounded by $R_{\text{max}}$. 

	\item $\gamma \in [0, 1]$ is the discount factor that represents the value of time.  
\end{itemize}	
\label{def:mdp}
\end{definition}
At each time step $t$, the environment has a state $s_t$. The learning agent observes this state\footnote{If the agent cannot fully observe the environment state, it be it becomes the partially-observable setting, shorten as POMDP.} and executes an action $a_t$. The action makes the environment transition into the next state $s_{t+1} \sim P(\cdot | s_t, a_t)$, and the new environment returns an immediate reward $R(s_t, a_t, s_{t+1})$ to the agent.  
The reward function  can  be also written as $R:\mathcal{S}\times \mathcal{A}\rightarrow \mathbb{R}$, which is interchangeable with $R: \mathcal{S} \times \mathcal{A} \times \mathcal{S} \rightarrow \mathbb{R}$. 

\subsection{Policy and Value Functions}
 Given above MDP setup, the goal of the agent is to solve the MDP: to find the optimal policy that maximises the reward over time.
Mathematically, one common objective is for the agent to find a Markovian (i.e., the input depends on only the current state) and stationary (i.e., function form is time-independent) policy function $\pi: \mathcal{S} \rightarrow \Delta(\mathcal{A})$, with $\Delta(\cdot)$ denoting the probability simplex.

Formally, a \emph{Markov policy} depends only on the current state, and can be written as
\begin{align}
\pi:\mathcal{S}\to \Delta(\mathcal{A}))\end{align}
for discrete stochastic case or $\pi:\mathcal{S}\to\mathcal{A}$ in the deterministic case. In contrast, a \emph{general} or \emph{non-Markovian} policy may depend on the whole history available at time $h$. If we denote the history by
\begin{align}
    H_h=(s_1,a_1,r_1,\dots,s_{h-1},a_{h-1},r_{h-1},s_h),
\end{align}
then a general policy is a mapping
\begin{align}
    \pi_h:H_h\to \Delta(\mathcal{A}),
\end{align}
or equivalently $\pi_h(a_h\mid H_h)$ for a stochastic policy.

The optimal Markov policy exists as long as the transition function and the reward
function are both Markovian and stationary, so by default we use $\pi$ to denote the Markov policy without further clarification.

The policy takes sequential actions such that the discounted cumulative reward is maximised in this infinite-horizon MDP in this infinite-horizon MDP:  
 \begin{equation}
\mathcal{J}(\pi)=\mathbb{E}_{s_{t+1} \sim P(\cdot | s_t, a_t)}\left[\sum_{t=0}^{\infty} \gamma^{t} R\left(s_{t}, a_{t}, s_{t+1}\right) \Big| a_{t} \sim \pi\left(\cdot \mid s_{t}\right), s_{0}\right].	
\label{eq:rlobj}
 \end{equation}
 Another common mathematical objective of an MDP is to maximise the time-average reward for finite-horizon MDP:
 \begin{equation}
 \mathcal{J}(\pi)=\mathbb{E}_{s_{t+1}\sim P(\cdot | s_t, a_t)}\left[\frac{1}{T} \sum_{t=0}^{T-1} R(s_t, a_t, s_{t+1}) \Big| a_{t} \sim \pi\left(\cdot \mid s_{t}\right), s_{0}\right].
 	 \label{eq:time_average_reward}
 \end{equation}
We will adopt the infinite-horizon MDP setting by default.

Based on the objective function of Eq. (\ref{eq:rlobj}), under a given policy $\pi$, we can define the state-action function (namely, the Q-function, which determines the expected return from undertaking action $a$ in state $s$) and the value function (which determines the return associated with the policy in state $s$) as:
  \begin{align}
Q^{\pi}(s, a) &=  \mathbb{E}^{\pi} \left[\sum_{t \geq 0} \gamma^{t} R\left(s_{t}, a_{t}, s_{t+1}\right) \Big| a_{0}=a, s_{0}=s\right] 	, \forall s \in \mathcal{S}, a\in \mathcal{A}
\label{eq:q_func} \\
V^{\pi}(s) &=  \mathbb{E}^{\pi} \left[\sum_{t \geq 0} \gamma^{t} R\left(s_{t}, a_{t}, s_{t+1}\right) \Big| s_{0}=s\right], \forall s \in \mathcal{S} 	
\label{eq:v_func} 
 \end{align}
where $\mathbb{E}^{\pi}$ is the expectation under the probability measure $\mathcal{P}^{\pi}$  over the set of infinitely long state-action trajectories $\tau=(s_0, a_0, s_1, a_1, ...)$ and  where $\mathcal{P}^{\pi}$  is induced by  state transition probability $P$, the policy $\pi$,  the initial state $s$ and initial action $a$ (in the case of the Q-function). 
The connection between the Q-function and value function is $V^{\pi}(s) = \mathbb{E}_{a\sim \pi(\cdot|s)}[Q^{\pi}(s, a)]$ and $Q^{\pi} = \mathbb{E}_{s'\sim P(\cdot| s, a)}[R(s, a, s') + V^{\pi}(s')]$.

\subsection{Value Iteration}

\subsubsection{Value-Based Method}
For all MDPs with finite states and actions, there exists at least one deterministic stationary optimal policy. With the known transition function $P(s'|s,a)$, this becomes a dynamic programming problem.
One of the most fundamental dynamic programming methods for computing the optimal value is \emph{value iteration} as Alg.~\ref{alg:value_iteration}. Starting from an arbitrary initial value function $V_0$, value iteration repeatedly applies the Bellman optimality operator:
\begin{align}
    V_{k+1}(s)
    = \max_{a}\sum_{s'}P(s'|s,a)\left(R(s,a,s')+\gamma V_k(s')\right).
    \label{eq:value_iteration}
\end{align}
After convergence, an optimal policy can be recovered by acting greedily with respect to the converged value function:
\begin{align}
    \pi^*(s)\in \arg\max_a \sum_{s'}P(s'|s,a)\left(R(s,a,s')+\gamma V^*(s')\right).
\end{align}

\begin{algorithm}[t]
\caption{Value Iteration}
\label{alg:value_iteration}
\begin{algorithmic}[1]
\STATE \textbf{Input}: transition model $P$, reward function $R$, discount factor $\gamma$, tolerance $\varepsilon$
\STATE Initialize $V_0(s)$ arbitrarily for all $s\in\mathcal{S}$
\FOR{$k=0,1,2,\ldots$}
    \STATE $V_{k+1}(s)\leftarrow \max_{a}\sum_{s'}P(s'|s,a)\left(R(s,a,s')+\gamma V_k(s')\right), \quad \forall s\in\mathcal{S}$
    \IF{$\|V_{k+1}-V_k\|_\infty<\varepsilon$}
        \STATE break
    \ENDIF
\ENDFOR
\STATE \textbf{Return}: $V_{k+1}$ and $\pi^*(s)\in \arg\max_a \sum_{s'}P(s'|s,a)\left(R(s,a,s')+\gamma V_{k+1}(s')\right)$
\end{algorithmic}
\end{algorithm}

Without knowing the transition function $P$, Q-learning is introduced to find the optimal Q-function $Q^*$ that maximises Eq. (\ref{eq:q_func}). Correspondingly, the optimal policy can be derived from the Q-function by taking the greedy action $\pi^*(s)=\arg\max_a Q^*(s, a)$. The classic Q-learning algorithm approximates $Q^*$ by $\hat{Q}$, and updates its value via temporal-difference learning:
\begin{equation}
    \hat{Q}(s_t,a_t)\leftarrow \hat{Q}(s_t,a_t)+\alpha\Big(R_t+\gamma \max_{a\in\mathcal{A}}\hat{Q}(s_{t+1},a)-\hat{Q}(s_t,a_t)\Big).
    \label{eq:q_learning}
\end{equation}

Researchers applied neural networks as a function approximator for the Q-function in updating Eq. (\ref{eq:q_learning}). Specifically, DQN optimises the following equation: 
\begin{equation}
	\min_\theta \mathbb{E}_{(s_t, a_t, R_t, s_{t+1})\sim \mathcal{D}}\left[\left(R_t + \gamma \max_{a\in\mathbb{A}}Q_{\theta^{-}}\left(s_{t+1}, a \right) - Q_\theta \left(s_t, a_t\right) \right)^2\right].
	\label{eq:dqn}
\end{equation}
The neural network parameters $\theta$ is fitted by drawing i.i.d. samples from the replay buffer  $\mathcal{D}$ and then updating in a supervised learning fashion. $Q_{\theta^{-}}$ is a slowly updated target network that helps stabilise training.

\subsubsection{Bellman Operator}
Using the MDP notation introduced above, the Bellman operator maps a value function to its one-step look-ahead update under the transition kernel $P$ and reward function $R$. For a fixed policy $\pi$, the Bellman expectation operator is defined by
\begin{align}
    (\mathcal{B}^{\pi}V)(s)
    &= \mathbb{E}_{a\sim \pi(\cdot\mid s),\, s'\sim P(\cdot\mid s,a)}
    \left[R(s,a,s') + \gamma V(s')\right] \\
    &= \sum_{a}\pi(a\mid s)\sum_{s'}P(s'\mid s,a)\left(R(s,a,s') + \gamma V(s')\right).
\end{align}
and its shorthand form is $\mathcal{B}Q(s,a) = \sum_{s^\prime}P_{ss^\prime}^a(R_{ss^\prime}^a+\gamma\max_{a^\prime} Q(s^\prime, a^\prime) )$.
In particular, the value function defined in the previous subsection satisfies the Bellman fixed-point equation
\begin{align}
    V^{\pi} = \mathcal{B}^{\pi}V^{\pi}.
\end{align}
For optimal control, we replace the policy average by a maximisation over actions and obtain the Bellman optimality operator
\begin{align}
    (\mathcal{B}V)(s)
    = \max_{a}\sum_{s'}P(s'\mid s,a)\left(R(s,a,s') + \gamma V(s')\right),
\end{align}
which is the same operator later written in the shorthand form $\mathcal{B}V(s)=\max_a(R_s^a+\gamma \sum_{s^\prime}P_{ss^\prime}^a V(s^\prime))$. Likewise, for action-values, we have the Bellman optimality operator:
\begin{align}
    (\mathcal{B}Q)(s,a)
    = \sum_{s'}P(s'\mid s,a)\left(R(s,a,s')+\gamma \max_{a'}Q(s',a')\right).
\end{align}
and its shorthand form is $\mathcal{B}Q(s,a) = \sum_{s^\prime}P_{ss^\prime}^a(R_{ss^\prime}^a+\gamma\max_{a^\prime} Q(s^\prime, a^\prime) )$.

\subsubsection{Convergence of Bellman Operator}

\begin{definition}[Metric Space] A metric space $(\mathcal{X}, d)$ is a set $\mathcal{X}$ with a metric $d$ defined to measure the distance between any two elements of the set $\mathcal{X}$. The distance measure in the metric space needs to satisfy the following properties:
\begin{itemize}
    \item Identity: $d(x, x)=0, \forall x\in\mathcal{X}$;
    \item Non-negativity: $d(x, y)\geq=0, \forall x,y\in\mathcal{X}$;
    \item Symmetry: $d(x,y) = d(y, x), \forall x,y\in\mathcal{X}$;
    \item Triangular Inequality: $d(x,z)\leq d(x,y)+d(y,z)$.
\end{itemize}
\label{def:metric_space}
\end{definition}

\begin{definition}[Cauchy Sequence]
In metric space $(\mathcal{X}, d)$, we take a subset (sequence) $\mathcal{X}_c=\{x_1, x_2, x_3, ..., x_n\}\subseteq \mathcal{X}$, if $\forall \epsilon>0, \exists N>0, \forall a, b>N, d(x_a, x_b),\epsilon$, then the sequence $\mathcal{X}_c$ is a Cauchy sequence.  
\end{definition}

\begin{definition}[Complete Metric Space] A metric space $(\mathcal{X}, d)$ is complete if every possible Cauchy sequence of the elements in the set $\mathcal{X}$ converges to an element that belongs to the set $\mathcal{X}$, \emph{i.e.}, the convergence limit of every Cauchy sequence of the elements of the set lies in the set itself.
\label{def:complete_metric_space}
\end{definition}

\begin{theorem}[Contractor/Contraction Mapping]
A function $f$ is defined on a metric space $(\mathcal{X}, d)$, and satisfies for $\forall x_1, x_2, \exists \gamma\in[0,1), d(f(x_1), f(x_2))\leq \gamma d(x_1, x_2)$, the the function (operator) $f$ is contraction mapping or contractor.
\end{theorem}

\begin{theorem}[Banach Fixed Point Theorem]
Let $(\mathcal{X}, d)$ be a complete metric space and a function $f: \mathcal{X}\rightarrow \mathcal{X}$ be a contractor, then $f$ has a unique fixed point $x^*\in\mathcal{X}$ (\emph{i.e.}, $f(x^*)=x^*$) such that the sequence $f(f(f(...f(x))))$ converges to $x^*$.
\label{theorem:banach}
\end{theorem}

\begin{lemma}[Bellman Optimal Operator]
\begin{align}
    \mathcal{B}V(s) = \max_a(R_s^a+\gamma \sum_{s^\prime}P_{ss^\prime}^a V(s^\prime))
\end{align}
\end{lemma}

Prove the convergence of value iteration is equivalent of proving the Bellman optimal operator satisfies the Banach fixed point theorem. In this case, the metric space $(\mathcal{X}, d)$ has $\mathcal{X}:=\{V(s)|V(s)\in\mathbb{R}\}$, and $d:=||\mathcal{X}||_\infty=\max|x_i|, x_i\in\mathcal{X}$. It is easy to see this metric space is complete.

\begin{theorem}[Convergence of Bellman Optimal Operator-$V^*$]
Optimal $V^*$ value is a fixed point of the Bellman contraction operator $\mathcal{B}$, which is applied on general $V:\mathcal{S}\times\mathcal{A}\rightarrow \mathbb{R}$.
The operator is a contractor in the sup-norm, \emph{i.e.},
\begin{align}
    ||\mathcal{B}V_1-\mathcal{B}V_2||_\infty \leq \gamma ||V_1 - V_2||_\infty
\end{align}
\end{theorem}
\begin{proof}
\begin{align}
    ||\mathcal{B}V_1-\mathcal{B}V_2||_\infty &= \max_{s}| \max_a(R_s^a + \gamma \sum_{s^\prime_1}P_{ss^\prime}^a V_1(s^\prime_1)) - \max_a(R_s^a + \gamma \sum_{s^\prime_2}P_{ss^\prime}^a V_2(s^\prime_2)) |\\
    &= \max_{s,a} \gamma|\sum_{s^\prime}P_{ss^\prime}^a (V_1(s^\prime) - V_2(s^\prime)) | \\
    &\leq\max_{s,a} \gamma \sum_{s^\prime}P_{ss^\prime}^a |V_1(s^\prime) - V_2(s^\prime) |  \text{ (by triangular inequality)} \\
    &\leq\max_{s,a} \gamma \max_{s^\prime} |V_1(s^\prime) - V_2(s^\prime) |   \text{ (assign probability 1 to } \max_{s_\prime} )\\
    &\leq \max_{s,a} \gamma ||V_1 - V_2 ||_\infty \\
    &=\gamma ||V_1-V_2||_\infty
\end{align}
\end{proof}

\begin{theorem}[Convergence of Bellman Optimal Operator-$Q^*$]
Optimal $Q^*$ value is a fixed point of the Bellman contraction operator $\mathcal{B}$, which is applied on general $Q:\mathcal{S}\times\mathcal{A}\rightarrow \mathbb{R}$ as: (neglect the ``*'' in following for simplicity)
\begin{align}
     \mathcal{B}Q(s,a) = \sum_{s^\prime}P_{ss^\prime}^a(R_{ss^\prime}^a+\gamma\max_{a^\prime} Q(s^\prime, a^\prime) )
     \label{eq:bellman_q}
\end{align}
The operator is a contractor in the sup-norm, \emph{i.e.},
\begin{align}
    ||\mathcal{B}Q_1-\mathcal{B}Q_2||_\infty \leq \gamma ||Q_1 - Q_2||_\infty
\end{align}
\label{thm:converge_bellman_q}
\end{theorem}
\begin{proof}
\begin{align}
    ||\mathcal{B}Q_1-\mathcal{B}Q_2||_\infty &= \max_{s,a}| \sum_{s^\prime}P_{ss^\prime}^a(R_{ss^\prime}^a+\gamma \max_{a^\prime_1}Q_1(s^\prime, a^\prime_1)-R_{ss^\prime}^a-\gamma \max_{a^\prime_2}Q_2(s^\prime, a^\prime_2)
    ) |\\
    &= \max_{s,a} \gamma|\sum_{s^\prime}P_{ss^\prime}^a (\max_{a^\prime_1}Q_1(s^\prime, a^\prime_1) - \max_{a^\prime_2}Q_2(s^\prime, a^\prime_2)) | \\
    &\leq \max_{s,a} \gamma \sum_{s^\prime}P_{ss^\prime}^a  |\max_{a^\prime_1}Q_1(s^\prime, a^\prime_1) - \max_{a^\prime_2}Q_2(s^\prime, a^\prime_2)|  \text{ (by triangular inequality)} \\
    &\leq \max_{s,a} \gamma \sum_{s^\prime}P_{ss^\prime}^a  \max_{a^\prime} |Q_1(s^\prime, a^\prime) - Q_2(s^\prime, a^\prime) |   \text{ (by convexity of max function)} \\
    &\leq \max_{s,a} \gamma  \max_{a^\prime}|Q_1(\hat{s}^\prime, a^\prime) - Q_2(\hat{s}^\prime, a^\prime) |   \text{ (assign probability 1 to } \max_{s_\prime} ) \\
    &\leq \max_{s,a} \gamma ||Q_1 - Q_2 ||_\infty \\
    &=\gamma ||Q_1-Q_2||_\infty
\end{align}
\end{proof}

Not only the optimal Bellman operators have contraction property, but also the general Bellman operators. The proofs are similar and omitted here.

\begin{remark}
Specifically, in the contraction of normal Bellman operators, if one of the value function is the unbiased value function, \emph{e.g.}, $V^\pi_2=\mathcal{B} V^\pi_2$ or $Q^\pi_2=\mathcal{B} Q^\pi_2$, another one is the estimation ($V^\pi_1$ or $Q^\pi_1$), the convergence theorems say that the estimated value function will converge to the true value function. 
\end{remark}

\subsubsection{Convergence of $Q$-Learning}

\begin{theorem}[Convergence of $Q$-Learning]
Consider a finite discounted MDP with bounded rewards, and the $Q$-learning update
\begin{align}
    Q_{t+1}(s_t,a_t)
    =
    (1-\alpha_t(s_t,a_t))Q_t(s_t,a_t)
    + \alpha_t(s_t,a_t)\left[r_t + \gamma \max_{a'} Q_t(s_{t+1},a')\right],
\end{align}
while for $(s,a)\neq (s_t,a_t)$ we keep
\begin{align}
    Q_{t+1}(s,a)=Q_t(s,a).
\end{align}
Assume:
\begin{itemize}
    \item every state-action pair $(s,a)$ is visited infinitely often;
    \item the reward is uniformly bounded;
    \item the learning rates satisfy the Robbins-Monro conditions
    \begin{align}
        \sum_{t=0}^{\infty}\alpha_t(s,a)=\infty,
        \qquad
        \sum_{t=0}^{\infty}\alpha_t^2(s,a)<\infty,
        \qquad \forall (s,a)\in \mathcal{S}\times\mathcal{A}.
    \end{align}
\end{itemize}
Then the iterates $Q_t$ converge to the optimal action-value function $Q^*$ almost surely:
\begin{align}
    Q_t(s,a)\xrightarrow[t\to\infty]{a.s.} Q^*(s,a), \qquad \forall (s,a)\in\mathcal{S}\times\mathcal{A},
\end{align}
where $Q^*$ is the unique fixed point of the Bellman optimality operator
\begin{align}
    (\mathcal{B}Q)(s,a)=\sum_{s'}P(s'|s,a)\left(R(s,a,s')+\gamma \max_{a'}Q(s',a')\right).
\end{align}
\label{thm:q_learning_convergence}
\end{theorem}

\begin{remark}
This theorem shows that $Q$-learning is a stochastic approximation procedure for solving the Bellman optimality fixed-point equation. The contraction of $\mathcal{B}$ guarantees uniqueness of $Q^*$, while the martingale-difference noise induced by sampling transitions is controlled by the diminishing step sizes.
\end{remark}

\subsection{Policy Gradient}
\label{sec:single_pg}

\subsubsection{Policy-Based Method}

Policy-based methods are designed to directly search over the policy space to find the optimal policy $\pi^*$. 
One can parameterise the policy expression $\pi^* \approx \pi_\theta(\cdot | s)$ and update the parameter $\theta$ in the direction that maximises the cumulative reward $\theta \leftarrow \theta + \alpha \nabla_\theta V^{\pi_\theta}(s)$ to find the optimal policy.  However, the gradient will depend on the unknown effects of policy changes on the state distribution. The famous policy gradient (PG) theorem \citep{sutton2000policy} derives an analytical solution that does not involve the state distribution, that is:     
\begin{equation}
 \nabla_\theta V^{\pi_\theta}(s) =	\mathbb{E}_{s\sim \mu^{\pi_{\theta}}(\cdot), a\sim \pi_{\theta}(\cdot| s)}\Big[ \nabla_\theta \log \pi_\theta (a|s) \cdot Q^{\pi_\theta}(s, a) \Big]
 \label{eq:pg}
\end{equation}
where $\mu^{\pi_{\theta}}$ is the state occupancy measure under policy $\pi_\theta$ and $\nabla\log \pi_\theta (a|s)$ is the updating score of the policy. When the policy is deterministic and the action set is continuous, one obtains the deterministic policy gradient (DPG) theorem \citep{silver2014deterministic}  as 
\begin{equation}
 \nabla_\theta V^{\pi_\theta}(s) =	\mathbb{E}_{s\sim \mu^{\pi_{\theta}}(\cdot)}\Big[ \nabla_\theta \pi_\theta (a|s) \cdot \nabla_a Q^{\pi_\theta}(s, a) \big|_{a=\pi_\theta(s)} \Big].
 \label{eq:dpg}
\end{equation}

A classic implementation of the PG theorem is REINFORCE \citep{williams1992simple}, which uses a sample return $R_t = \sum_{i=t}^{T}\gamma^{i-t}r_i$ to estimate $Q^{\pi_\theta}$. 
Alternatively, one can use a model of $Q_\omega$ (also called \emph{critic}) to approximate the true $Q^{\pi_\theta}$ and update the parameter $\omega$ via TD learning. This approach gives rise to the famous  actor-critic methods \citep{konda2000actor,peters2008natural}. Important variants of actor-critic methods include trust-region methods \citep{schulman2015trust, schulman2017proximal},  soft actor-critic methods \citep{haarnoja2018soft}, and  deep deterministic policy gradient (DDPG) methods  \citep{lillicrap2015continuous}.

\subsubsection{Policy Gradient Theorem}

In this subsection, we write the policy objective as
\begin{align}
    J(\theta) := \mathbb{E}_{s_0\sim p_0}\left[V^{\pi_{\theta}}(s_0)\right].
\end{align}
Here $\mu^{\pi_{\theta}}$ denotes the discounted state occupancy measure induced by policy $\pi_{\theta}$, and $\pi_{\theta}(a\mid s)$ denotes a stochastic policy parameterized by $\theta$. For the reparameterization form, we assume that an action can be generated by first sampling $\epsilon\sim p(\epsilon)$ from a fixed noise distribution and then applying a differentiable transformation
\begin{align}
    a = f_{\theta}(\epsilon; s).
\end{align}
The conditional density induced by this sampling procedure is just the policy $\pi_{\theta}(a\mid s)$.

\begin{assumption}\label{assum1}
$\mathcal{S}$ and $\mathcal{A}$ are closed and bounded.
\end{assumption}

\begin{assumption}\label{assum2}
$P(s' \mid s,a)$, $f_{\theta}(\epsilon; s)$, $\pi_{\theta}(a\mid s)$, $p(\epsilon)$, $R(s,a,s')$, $p_0(s)$ and their derivatives are continuous in all variables $s$, $a$, $s'$, $\theta$, and $\epsilon$.
\end{assumption}

\begin{remark}
The two assumptions above allow us to exchange derivatives and integrals, and the order of multiple integrations, using Fubini's theorem and Leibniz integral rule.
\end{remark}

\begin{theorem}[Stochastic Policy Gradient (SPG) Theorem]\label{thm_sto}
Suppose that the MDP satisfies Assumption~\ref{assum1} and \ref{assum2}, then
\begin{align*}
\nabla_{\theta} J(\theta)
&= \mathbb{E}_{s\sim \mu^{\pi_{\theta}},\, a\sim \pi_{\theta}(\cdot\mid s)}
\left[\nabla_{\theta}\log \pi_{\theta}(a\mid s)\, Q^{\pi_{\theta}}(s,a)\right] \\
&= \mathbb{E}_{s\sim \mu^{\pi_{\theta}},\, a\sim \pi_{\theta}(\cdot\mid s),\, s'\sim P(\cdot\mid s,a)}
\left[\nabla_{\theta}\log \pi_{\theta}(a\mid s)\left(R(s,a,s')+\gamma V^{\pi_{\theta}}(s')\right)\right].
\end{align*}
\end{theorem}

\begin{corollary}[Baseline Subtraction Stochastic Policy Gradient]
\begin{align*}
\nabla_{\theta} J(\theta)
&= \mathbb{E}_{s\sim \mu^{\pi_{\theta}},\, a\sim \pi_{\theta}(\cdot\mid s),\, s'\sim P(\cdot\mid s,a)}
\left[\nabla_{\theta}\log \pi_{\theta}(a\mid s)\left(R(s,a,s')+\gamma V^{\pi_{\theta}}(s')-V^{\pi_{\theta}}(s)\right)\right] \\
&= \mathbb{E}_{s\sim \mu^{\pi_{\theta}},\, a\sim \pi_{\theta}(\cdot\mid s)}
\left[\nabla_{\theta}\log \pi_{\theta}(a\mid s)\, A^{\pi_{\theta}}(s,a)\right].
\end{align*}
\end{corollary}
where $A^{\pi_{\theta}}(s,a)=Q^{\pi_{\theta}}(s,a)-V^{\pi_{\theta}}(s)$ is the advantage function.
Both of these are unbiased, but baseline subtraction reduces the variance in gradient estimation.

\begin{theorem}[Reparameterization Policy Gradient Theorem]\label{thm_repara}
Suppose that the MDP satisfies Assumption~\ref{assum1} and \ref{assum2}, then
\begin{align*}
\nabla_{\theta} J(\theta)
&= \mathbb{E}_{s\sim \mu^{\pi_{\theta}},\, \epsilon\sim p}
\left[\nabla_{\theta} f_{\theta}(\epsilon;s)\nabla_{a}Q^{\pi_{\theta}}(s,a)\big|_{a=f_{\theta}(\epsilon;s)}\right].
\end{align*}
\end{theorem}

\begin{proof}
By the policy gradient theorem, we have
\begin{align*}
\nabla_{\theta} J(\theta)
&= \int \mu^{\pi_{\theta}}(s)\pi_{\theta}(a\mid s)Q^{\pi_{\theta}}(s,a)\nabla_{\theta}\log \pi_{\theta}(a\mid s)\,\mathrm{d}a\,\mathrm{d}s.
\end{align*}

Thus
\begin{align*}
&\nabla_{\theta} J(\theta) \\
=& \int \mu^{\pi_{\theta}}(s)\pi_{\theta}(a\mid s)Q^{\pi_{\theta}}(s,a)\nabla_{\theta}\log \pi_{\theta}(a\mid s)\,\mathrm{d}a\,\mathrm{d}s \\
=& \int \mu^{\pi_{\theta}}(s) \left( \int Q^{\pi_{\theta}}(s,a) \nabla_{\theta} \pi_{\theta}(a\mid s) \,\mathrm{d}a \right) \,\mathrm{d}s \\
=& \int \mu^{\pi_{\theta}}(s) \left[ \int \nabla_{\theta} \left( Q^{\pi_{\theta}}(s,a) \pi_{\theta}(a\mid s) \right) \,\mathrm{d}a - \int \pi_{\theta}(a\mid s) \nabla_{\theta} Q^{\pi_{\theta}}(s,a) \,\mathrm{d}a \right] \,\mathrm{d}s \\
=& \int \mu^{\pi_{\theta}}(s) \left[\nabla_{\theta} \left(\int Q^{\pi_{\theta}}(s,a) \pi_{\theta}(a\mid s) \,\mathrm{d}a \right) - \int \pi_{\theta}(a\mid s) \nabla_{\theta} Q^{\pi_{\theta}}(s,a) \,\mathrm{d}a \right] \,\mathrm{d}s \\
=& \int \mu^{\pi_{\theta}}(s) \left[\nabla_{\theta} \left(\int p(\epsilon) Q^{\pi_{\theta}}(s,f_{\theta}(\epsilon; s)) \,\mathrm{d}\epsilon \right) - \int p(\epsilon) \nabla_{\theta} Q^{\pi_{\theta}}(s,a)\big|_{a=f_{\theta}(\epsilon; s)} \,\mathrm{d}\epsilon \right] \,\mathrm{d}s \\
&\qquad \text{(by reparameterization)} \\
=& \int \mu^{\pi_{\theta}}(s) \left[\int p(\epsilon) \left(\nabla_{\theta} f_{\theta}(\epsilon; s) \nabla_{a} Q^{\pi_{\theta}}(s,a)\big|_{a=f_{\theta}(\epsilon; s)} + \nabla_{\theta} Q^{\pi_{\theta}}(s,a)\big|_{a=f_{\theta}(\epsilon; s)} \right) \,\mathrm{d}\epsilon \right.\\
&\qquad\qquad\left. - \int p(\epsilon) \nabla_{\theta} Q^{\pi_{\theta}}(s,a)\big|_{a=f_{\theta}(\epsilon; s)} \,\mathrm{d}\epsilon \right] \,\mathrm{d}s \\
=& \int \mu^{\pi_{\theta}}(s)\, p(\epsilon) \nabla_{\theta} f_{\theta}(\epsilon; s) \nabla_{a} Q^{\pi_{\theta}}(s,a)\big|_{a=f_{\theta}(\epsilon; s)} \,\mathrm{d}\epsilon \,\mathrm{d}s.
\end{align*}
\end{proof}

\begin{remark}
This theorem is a direct application of reparameterization to the gradient of the policy objective. It is understood to be known but is not formally presented or derived for policy gradients in any existing work.
We include it here for completeness as well as a useful theoretical tool.
\end{remark}

\begin{corollary}[Deterministic Policy Gradient Theorem]\label{cor_dpg}
Suppose that the MDP satisfies Assumption~\ref{assum1} and \ref{assum2}, for a deterministic policy $\mu_{\theta}(s)$, we have
\begin{align*}
\nabla_{\theta} J(\theta)
&= \mathbb{E}_{s\sim \mu^{\mu_{\theta}}}\left[\nabla_{\theta}\mu_{\theta}(s)\nabla_{a}Q^{\mu_{\theta}}(s,a)\big|_{a=\mu_{\theta}(s)}\right].
\end{align*}
\end{corollary}

\begin{proof}
Let $p(\epsilon)$ be the delta function. Thus $\int p(\epsilon) f_{\theta}(\epsilon; s)\,\mathrm{d}\epsilon = f_{\theta}(0;s)$. Furthermore, let $f_{\theta}(0;s)=\mu_{\theta}(s)$. By Theorem~\ref{thm_repara},
\begin{align*}
\nabla_{\theta} J(\theta)
=& \int \mu^{\mu_{\theta}}(s) \left(\int p(\epsilon) \nabla_{\theta} f_{\theta}(\epsilon; s) \nabla_{a}Q^{\mu_{\theta}}(s,a)\big|_{a=f_{\theta}(\epsilon; s)} \,\mathrm{d}\epsilon \right) \,\mathrm{d}s \\
=& \int \mu^{\mu_{\theta}}(s) \nabla_{\theta} f_{\theta}(0;s) \nabla_{a}Q^{\mu_{\theta}}(s,a)\big|_{a=f_{\theta}(0;s)} \,\mathrm{d}s \\
=& \int \mu^{\mu_{\theta}}(s) \nabla_{\theta} \mu_{\theta}(s) \nabla_{a}Q^{\mu_{\theta}}(s,a)\big|_{a=\mu_{\theta}(s)} \,\mathrm{d}s.
\end{align*}
\end{proof}

\begin{remark}
Both DPG and DDPG are proposed based on this theorem, which is first proved by~\citet{silver2014deterministic}. It is limited to deterministic policies and thus can be deduced as a corollary of Theorem~\ref{thm_repara}, which is applicable to stochastic policies as well.
\end{remark}

\begin{corollary}[Entropy-regularized Reparameterization Policy Gradient Theorem]\label{cor_sac}
Consider the entropy-regularized values
\begin{align*}
V^{\pi_{\theta}}(s_0) &= \mathbb{E}_{\pi}\left[\sum_{t=0}^{\infty} \gamma^{t}\left(R(s_t, a_t, s_{t+1}) + \alpha \mathcal{H}(\pi_{\theta}(\cdot \mid s_t))\right)\right], \\
Q^{\pi_{\theta}}(s_0, a_0) &= \mathbb{E}_{\pi}\left[\sum_{t=0}^{\infty} \gamma^{t}R(s_t, a_t, s_{t+1}) + \alpha \sum_{t=1}^{\infty} \gamma^{t}\mathcal{H}(\pi_{\theta}(\cdot \mid s_t))\right],
\end{align*}
where
\begin{align*}
\mathcal{H}(p) = - \int p(x)\log p(x)\,\mathrm{d}x
\end{align*}
is the differential entropy for probability density function $p(x)$, and $\alpha$ is a positive constant. Suppose that the MDP satisfies Assumption~\ref{assum1} and \ref{assum2}, then
\begin{align*}
\nabla_{\theta} J(\theta)
&= \mathbb{E}_{s\sim \mu^{\pi_{\theta}},\, \epsilon\sim p}
\left[\nabla_{\theta}f_{\theta}(\epsilon; s)\nabla_{a}Q^{\pi_{\theta}}(s, a)\big|_{a=f_{\theta}(\epsilon; s)} - \alpha \nabla_{\theta}\log \pi_{\theta}(f_{\theta}(\epsilon; s)\mid s)\right].
\end{align*}
\end{corollary}

\section{Game-Theoretical Multi-Agent RL}

Single-agent RL studies decision making against an environment whose dynamics are fixed and do not strategically respond to the learner's behavior. In multi-agent RL, by contrast, part of the environment is replaced by other learning or decision-making agents. As a result, the return of each agent depends not only on the state transition model and its own policy, but also on the policies chosen by the other agents. This interdependence naturally gives rise to a game-theoretical structure.

From this perspective, multi-agent RL can be viewed as extending MDPs to interactive decision processes with multiple players, joint actions, and possibly different reward functions. Compared with the single-agent objective in Eq. (\ref{eq:rlobj}), where one optimizes a single policy against a passive environment, the multi-agent objective is defined over a tuple of policies $(\pi^1,\ldots,\pi^M)$. For player $i$, a general discounted-return objective can be written as
\begin{equation}
J_i(\pi^1,\ldots,\pi^M)=\mathbb{E}_{\pi^1,\ldots,\pi^M}\left[\sum_{t=0}^{\infty}\gamma^t r_i(s_t,a_t^1,\ldots,a_t^M)\right].
\end{equation}
The optimization problem is therefore no longer simply $\max_{\pi} J(\pi)$, but rather a strategic problem in which each player optimizes its own objective while accounting for the policies of others. Depending on the reward structure, one studies solution concepts such as equilibrium policies, minimax policies in zero-sum games, or cooperative optima in team settings. In this section, we first introduce minimax optimization as the mathematical core of competitive interaction, then review zero-sum games, followed by their sequential extension to Markov games, and finally briefly discuss general-sum and cooperative settings.

\paragraph{Minimax Optimization}
The mathematical core of zero-sum game solving is minimax optimization. In its most basic form, one optimizes an objective $f(x,y)$ with two competing variables $x\in\mathbb{R}^m$ and $y\in\mathbb{R}^n$:
\begin{equation}
    \min_{x\in\mathbb{R}^m} \max_{y\in\mathbb{R}^n} f(x,y)
    \label{eq:minimax_opt}
\end{equation}
where $f(x,y):\mathbb{R}^m\times \mathbb{R}^n \rightarrow \mathbb{R}$ is not necessarily convex-concave. When $f(\cdot, y)$ is convex for every fixed $y$ and $f(x, \cdot)$ is concave for every fixed $x$, the problem is called convex-concave. Many game-theoretic learning problems, especially in multi-agent reinforcement learning, can be understood as finding stable solutions to such minimax structures.

\subsection{Zero-Sum Game}
Zero-sum games provide the canonical game-theoretic setting in which the minimax structure in Eq. (\ref{eq:minimax_opt}) arises naturally. They model fully competitive interactions, where one player's gain is exactly the other player's loss, so the total utility sums to zero.

For a two-player zero-sum game, let $\mu$ denote the policy of the maximizing player and $\nu$ denote the policy of the minimizing player, and let $R(\mu,\nu)$ denote the payoff to the maximizing player under strategy pair $(\mu,\nu)$. A central result is the minimax theorem \citep{von1945theory}, which states that
\begin{equation}\label{eq:minimax}
\max_{\mu}\min_{\nu} \mathbb{E}\Big[R\big(\mu, \nu\big)\Big]
=
\min_{\nu}\max_{\mu} \mathbb{E}\Big[R\big(\mu, \nu\big)\Big].
\end{equation}
The minimax theorem implies that the game has a well-defined value, and a Nash equilibrium is a natural solution concept: when a player uses an equilibrium strategy, the opponent cannot improve against it by unilateral deviation.

\begin{definition}[Nash Equilibrium]
For a two-player zero-sum game, a strategy pair $(\mu^\star,\nu^\star)$ is called a Nash equilibrium if neither player can improve its objective by unilateral deviation, that is,
\begin{align}
    \mathbb{E}\big[R(\mu,\nu^\star)\big] \le \mathbb{E}\big[R(\mu^\star,\nu^\star)\big] \le \mathbb{E}\big[R(\mu^\star,\nu)\big],
    \qquad \forall \mu,\nu.
\end{align}
\end{definition}

\subsection{Markov Game}

\begin{definition}[Markov Game]
A finite-horizon Markov game can be viewed as the sequential extension of a stage game, where players repeatedly interact over $H$ steps and the environment state evolves according to a Markov transition rule. In the general multi-player case, a Markov game is specified by
\begin{align}
    \left\langle H, \mathcal{S}, \{\mathcal{A}_i\}_{i=1}^M, P, \{R_i\}_{i=1}^M \right\rangle,
\end{align}
where $H$ is the horizon, $\mathcal{S}$ is the state space, $\mathcal{A}_i$ is the action space of player $i$, $P=\{P_h\}_{h\in[H]}$ is a collection of transition kernels
\begin{align}
    P_h:\mathcal{S}\times \mathcal{A}_1\times\cdots\times \mathcal{A}_M \rightarrow \Delta(\mathcal{S}),
\end{align}
and $R_i=\{R_{i,h}\}_{h\in[H]}$ is the reward function of player $i$ with
\begin{align}
    R_{i,h}:\mathcal{S}\times \mathcal{A}_1\times\cdots\times \mathcal{A}_M \rightarrow \mathbb{R}.
\end{align}
\end{definition}

\subsection{Two-Player Zero-Sum Markov Game}
As a special case of Markov game, we define two-player zero-sum Markov game as follows. 
\begin{definition}[Two-Player Zero-Sum Markov Game]
It is specified by the tuple
\begin{align}
    \langle H, \mathcal{S}, \mathcal{A}, \mathcal{B}, P, R \rangle,
\end{align}
where $\mathcal{A}$ and $\mathcal{B}$ are the action spaces of the max-player and min-player respectively, $P=\{P_h\}_{h\in[H]}$ is a collection of transition kernels
\begin{align}
    P_h:\mathcal{S}\times\mathcal{A}\times\mathcal{B}\rightarrow \Delta(\mathcal{S}),
\end{align}
and $R=\{R_h\}_{h\in[H]}$ is the reward function of the max-player with
\begin{align}
    R_h:\mathcal{S}\times\mathcal{A}\times\mathcal{B}\rightarrow \mathbb{R}.
\end{align}
The min-player receives reward $-R_h$, so the total reward sums to zero at each step.
\end{definition}


\paragraph{Policy and Value Functions}
A (Markov) policy $\mu$ of the max-player is a collection of functions $\{\mu_h:\mathcal{S}\rightarrow \Delta(\mathcal{A})\}_{h\in[H]}$, and similarly a policy $\nu$ of the min-player is a collection $\{\nu_h:\mathcal{S}\rightarrow \Delta(\mathcal{B})\}_{h\in[H]}$. We write $\mu_h(a\mid s)$ and $\nu_h(b\mid s)$ for the corresponding action probabilities at step $h$.

We use $V^{\mu, \nu}_{h} \colon \mathcal{S} \to \mathbb{R}$ to denote the value
function at step $h$ under policy $\mu$ and $\nu$, so that
$V^{\mu, \nu}_{h}(s)$ gives the expected cumulative rewards
received under policy $\mu$ and $\nu$, starting from $s$ at step $h$:
\begin{equation} \label{eq:V_value}
\begin{aligned}
	 V^{\mu, \nu}_{h}(s) 
\defeq \mathbb{E}_{\mu, \nu}\bigg[\sum_{h' =
        h}^H R_{h'}(s_{h'}, a_{h'}, b_{h'}) \bigg| s_h = s\bigg].
\end{aligned}
\end{equation}
We also define $Q^{\mu, \nu}_h:\mathcal{S} \times \mathcal{A} \times \mathcal{B} \to \mathbb{R}$ to be the action-value function at step $h$ so that
$Q^{\mu, \nu}_{h}(s, a, b)$ gives the cumulative rewards received under policy $\mu$ and $\nu$, starting from
$(s, a, b)$ at step $h$:
\begin{equation} \label{eq:Q_value}
\begin{aligned}
  & \quad Q^{\mu, \nu}_{h}(s, a, b) \\
  & \defeq \mathbb{E}_{\mu,
    \nu}\bigg[\sum_{h' = h}^H R_{h'}(s_{h'},  a_{h'}, b_{h'})
  \big| s_h = s, a_h = a, b_h = b\bigg].
\end{aligned}
\end{equation}
For simplicity, we define the transition operator
\begin{align}
    [P_h V](s, a, b) \defeq \mathbb{E}_{s' \sim P_h(\cdot\mid s, a, b)}V(s')
\end{align}
for any value function $V$. We also use
\begin{align}
    [\mathcal{D}_{\mu_h\times \nu_h} Q](s) \defeq \mathbb{E}_{a\sim \mu_h(\cdot\mid s),\, b\sim \nu_h(\cdot\mid s)} Q(s, a, b)
\end{align}
for any action-value function $Q$. By definition of value functions, we have the Bellman
equation
\begin{equation}
\left\{
	\begin{aligned}
&  Q^{\mu, \nu}_{h}(s, a, b) =
  (R_h + P_h V^{\mu, \nu}_{h+1})(s, a, b),\\
 &    V^{\mu, \nu}_{h}(s)
  =  (\mathcal{D}_{\mu_h\times\nu_h} Q^{\mu, \nu}_h)(s),
\end{aligned}
\right.
\label{eq:bellman}
\end{equation}
for all $(s, a, b, h) \in \mathcal{S} \times \mathcal{A} \times \mathcal{B} \times [H]$, and at the $(H+1)^{\text{th}}$ step we have $V^{\mu, \nu}_{H+1}(s) = 0$ for all $s \in \mathcal{S}$.



\paragraph{Best Response and Nash Equilibrium.}
For any policy of the max-player $\mu$, there exists a \emph{best response} of the min-player, which is a policy
$\nu^\dagger(\mu)$ satisfying $V_h^{\mu, \nu^\dagger(\mu)}(s) = \inf_{\nu} V_h^{\mu, \nu}(s)$ for
any $(s, h) \in \mathcal{S} \times [H]$. 
We denote $V_h^{\mu, \dagger} \defeq V_h^{\mu, \nu^\dagger(\mu)}$. According to the previous Bellman equation \eqref{eq:bellman}, we further have the best response defined in a dynamic programming manner (which is applied in the experimental section):
\begin{align}
    Q^{\mu, \dagger}_h(s,a,b) &= (R_h+P_hV^{\mu, \dagger}_{h+1})(s,a,b) \label{eq:best_response_q}\\
    V^{\mu, \dagger}_h(s) &= \inf_{\nu}(\mathcal{D}_{\mu_h\times \nu_h}Q_h^{\mu, \dagger})(s)
    \label{eq:best_response_v}
\end{align}

By symmetry, we
can also define $\mu^\dagger(\nu)$ and $V_h^{\dagger, \nu}$.  
It is further known that there exist policies $\mu^\star$, $\nu^\star$
that are optimal against the best responses of the opponents, in the sense that
\begin{equation*}
\left\{
\begin{aligned}
	  &V^{\mu^\star, \dagger}_h(s) = \sup_{\mu}
      V^{\mu, \dagger}_h(s), \\
       &V^{\dagger, \nu^\star}_h(s) = \inf_{\nu}
      V^{\dagger, \nu}_h(s),\end{aligned}
      \right.
\end{equation*}
for all $(s,h) \in \mathcal{S} \times [H]$.
We call these optimal strategies $(\mu^\star,\nu^\star)$ the Nash equilibrium of the Markov game, which satisfies the following minimax equation 
\footnote{The minimax theorem here is different from the one for matrix games, i.e. $\max_\phi\min_\psi \phi\trans A\psi = \min_\psi\max_\phi \phi\trans A\psi$ for any matrix $A$, since here $V^{\mu, \nu}_h(s)$ is in general not bilinear in $\mu, \nu$.}:
\begin{equation*}
\textstyle \sup_{\mu} \inf_{\nu} V^{\mu, \nu}_h(s) = V^{\mu^\star, \nu^\star}_h(s) = \inf_{\nu} \sup_{\mu} V^{\mu, \nu}_h(s).
\end{equation*}
Intuitively, a Nash equilibrium gives a solution in which no player has anything to gain by changing only her own policy. 


\paragraph{Learning Objective.}

We measure the suboptimality of any pair of general policies $(\hat{\mu}, \hat{\nu})$ using the gap between their performance and the performance of the optimal strategy (i.e., Nash equilibrium) when playing against the best responses respectively:
\begin{equation*}
\begin{aligned}
&	  V^{\dagger, \hat{\nu}}_1(s_1) - V^{\hat{\mu}, \dagger}_1(s_1)  \\
	  =& \brac{V^{\dagger, \hat{\nu}}_1(s_1) - V^{*}_1(s_1)} 
	  +  \brac{V^{*}_1(s_1) -   V^{\hat{\mu},\dagger}_1(s_1)}.
\end{aligned}
\end{equation*}

\begin{definition}[$\alpha$-approximate Nash equilibrium] \label{def:epsilon_Nash} A pair of general policies $(\hat{\mu},\hat{\nu})$ is an \textbf{$\alpha$-approximate Nash equilibrium}, if $V^{\dagger, \hat{\nu}}_1(s_1) - V^{\hat{\mu}, \dagger}_1(s_1) \le \alpha$.
\end{definition}

In value iteration for Markov games with simulator, the estimated transition model is:
\begin{align}
    \hat{P}_h(s^\prime\mid s,a,b) = \frac{1}{n}\sum_{i=1}^n\mathbbm{1}[s^\prime_i=s^\prime]
\end{align}

Nash equilibrium policies $\hat{\mu}$ can be derived with the following:

By Bellman optimality equation,
\begin{align}
    Q_h^*(s,a,b) &= R_h(s,a,b)+(\hat{P}_h V^*_{h+1})(s,a,b)\\
    V^*_h(s)&=  \min_{\nu_h}\max_{\mu_h}\sum_{a,b}\mu_h(a|s)\nu_h(b|s)Q^*_h(s,a,b)
\end{align}

The Nash equilibrium $(\hat{\mu}, \hat{\nu})$ can be calculated by:
\begin{align}
    (\hat{\mu}_h(\cdot|s), \hat{\nu}_h(\cdot|s)) = \arg\min_{\nu_h}\arg\max_{\mu_h}\sum_{a,b}\mu_h(a|s)\nu_h(b|s)Q_h^*(s,a,b)
\end{align}
for each $(s, h)$.

\begin{theorem}
For value iteration (VI) with simulator in two-player zero-sum Markov game, with $n\ge \frac{cH^4 \iota}{\epsilon^2}, \iota=\log\frac{2HSA}{p}$, with probability at least $1-p$, the policy $(\hat{\mu}, \hat{\nu})$ resulting from VI will be $\epsilon-$optimal.
\begin{align}
    V_1^{\dagger, \hat{\nu}}(s_1) - V_1^{\hat{\mu}, \dagger}(s_1)\le \epsilon
\end{align}
\end{theorem}

\subsection{Markov Potential Game}
Markov potential games form an important structured subclass of general-sum Markov games. While a general-sum game may involve conflicting objectives across players, a potential game admits a single scalar \textit{potential function} that tracks every player's incentive for unilateral deviation. This structure is useful because it converts a strategic equilibrium problem into an optimization problem over the potential function.

\subsubsection{Potential Game}
We first recall the stage-game notion. Let $N$ be the number of players, let $\mathcal{A}_i$ be the action space of player $i$, and write the joint action as $\mathbf{a}=(a_i,a_{-i})\in \mathcal{A}_1\times\cdots\times\mathcal{A}_N$. For a fixed state $s$, a state-dependent stage game is called a potential game if unilateral changes in a player's reward are exactly matched by the change in a common potential function.

\begin{definition}[Stage Potential Game]
For a stage game with reward functions $\{R_i(s,\mathbf{a})\}_{i=1}^N$, it is a state potential game if there exists a function $\Phi(s,\mathbf{a})$ such that for every player $i$, every state $s$, every $a_i,\tilde{a}_i\in\mathcal{A}_i$, and every fixed $a_{-i}$,
\begin{align}
    R_i(s,\tilde{a}_i,a_{-i})-R_i(s,a_i,a_{-i})
    =
    \Phi(s,\tilde{a}_i,a_{-i})-\Phi(s,a_i,a_{-i}).
\end{align}
\end{definition}

\subsubsection{Markov Potential Game}
The dynamic counterpart is obtained by extending a stage potential game with Markov state transitions, so that one-step payoff comparisons are replaced by comparisons of long-term value functions induced by policies.

\begin{definition}[Markov Potential Game]
Consider a general-sum Markov game with players $i\in[N]$, policies $\boldsymbol{\pi}=(\pi_1,\ldots,\pi_N)$, and player-wise value functions $V_i^{\boldsymbol{\pi}}(s)$. The game is called a \textbf{Markov potential game} if there exists a potential function $\Phi^{\boldsymbol{\pi}}(s)$ such that for every player $i$, every state $s$, every $\pi_i,\tilde{\pi}_i\in\Pi_i$, and every fixed profile $\pi_{-i}\in \Pi_{-i}$,
\begin{align}
    V_i^{\pi_i,\pi_{-i}}(s)-V_i^{\tilde{\pi}_i,\pi_{-i}}(s)
    =
    \Phi^{\pi_i,\pi_{-i}}(s)-\Phi^{\tilde{\pi}_i,\pi_{-i}}(s).
\end{align}
\end{definition}

This definition means that every player's incentive for unilateral deviation is perfectly aligned with the change of the same scalar potential. Consequently, maximizing the potential is compatible with reaching a Nash equilibrium.

\begin{proposition}
If $\boldsymbol{\pi}^\star$ maximizes the potential function $\Phi^{\boldsymbol{\pi}}(s)$, then $\boldsymbol{\pi}^\star$ is a Nash equilibrium of the Markov potential game.
\end{proposition}

\begin{theorem}
For finite Markov potential games, there exists at least one deterministic Nash equilibrium.
\end{theorem}

\paragraph{Sufficient Conditions.}
Several structural conditions imply that a Markov game is a Markov potential game. Typical examples include:
\begin{itemize}
    \item each state-wise stage game is a potential game and the transition structure preserves this alignment across states;
    \item separable rewards together with state-independent transitions;
    \item action-independent transitions combined with a state-wise potential-game reward structure.
\end{itemize}
These conditions illustrate the main design principle behind Markov potential games: strategic interactions may be decentralized across players, but the induced incentives can still be summarized by a single global potential function.

\chapter{Organization}

This thesis is organized into two technical parts that develop a common theme from complementary directions. Part~I studies reinforcement learning algorithms in games, where the central challenge is strategic decision making in multi-agent environments. Part~II studies reinforcement learning in the era of foundation models, where pretrained generative models and learned world models provide rich priors for sequential decision making. Together, the two parts connect the algorithmic foundations of reinforcement learning with the emerging role of world modeling in large-scale generative and interactive systems.

Part~I begins with Chapter~\ref{ch:two_player_zero_sum} \emph{Two-Player Zero-Sum Game}, which develops reinforcement learning algorithms in the canonical adversarial setting and establishes the basic connection between incentives, policies, and equilibrium behaviour. It is followed by Chapter~\ref{ch:zero_sum_video} \emph{Zero-Sum Video Game}, which brings these ideas into richer video-game environments and studies how competitive reinforcement learning behaves in practice at scale. The part then concludes with Chapter~\ref{chap:rl_general_sum} \emph{Multi-Player General-Sum Game}, which moves beyond pure competition to more general strategic interaction, where the relevant notion of optimality is no longer a single winner but an equilibrium shaped by the incentives of multiple agents.

Part~II turns to reinforcement learning with foundation models, motivated by the idea that broad prior knowledge can make optimisation more efficient and better aligned with complex objectives. Chapter~\ref{ch:diffusion_world_model} \emph{Diffusion World Model} studies how diffusion-based generative models can serve as learned environment models for planning and control. Chapter~\ref{ch:consistency_policy} \emph{Consistency Models as Reinforcement Learning Policy} explores expressive generative model classes as policies, showing how advances in generative modeling can directly enrich the action space of reinforcement learning. Next, Chapter~\ref{ch:rl_video} \emph{RL in Few-Step Video Generation} then investigates how reinforcement learning can improve efficient video generation when the generator itself is a powerful pretrained model. Chapter~\ref{ch:video_world_model} \emph{Video World Model} further develops the world-modeling perspective in  interactive settings, where the world model must reflect how actions shape future visual observations. Finally, Chaper~\ref{ch:world_model_mem} \emph{World Model with Memory} addresses long-horizon sequence modeling, providing architectural tools that support more capable world models for sequential generation and interaction.

Taken together, these chapters present a unified view of reinforcement learning as the study of objective-driven sequential decision making, from algorithmic questions in strategic games to foundation-model-based systems that rely on learned world models to represent, predict, and optimise complex environments.
\part{Reinforcement Learning in Games} \label{part1:rl_in_games}
\chapter{Two-Player Zero-Sum Game\label{ch:two_player_zero_sum}}
\begin{center}
\begin{quote}
This section is based on paper ``\textit{A Deep Reinforcement Learning Approach for Finding Non-Exploitable Strategies in Two-Player Atari Games}''~\cite{ding2022deep} written in collaboration with Dijia Su, Qinghua Liu and Chi Jin.
\end{quote}
\end{center}

\section{Introduction}
\label{intro}

Reinforcement learning (RL) in multi-agent systems has succeeded in many challenging
tasks, including Go \citep{go}, hide-and-seek \citep{baker2019emergent},
Starcraft \citep{vinyals2017starcraft}, Dota \citep{berner2019dota}, Poker
\citep{heinrich2016deep, brown2019superhuman, zha2021douzero}, and board games
\citep{lanctot2019openspiel, serrino2019finding}. Excluding the systems for
Poker, a large number of these works measure their success in terms of performance
against fixed agents, average human players or experts in a few shots. A distinguishing
feature of games is that the opponents can further model the learner's behaviors,
adapt their strategies, and exploit the learner's weakness. It is highly unclear
whether the policies found by many of these multi-agent systems remain viable
against the adversarial exploitation of the opponents.

In this paper, we consider two-player zero-sum \emph{Markov games} (MGs), and our
objective is to find the \emph{Nash Equilibrium} (NE)~\citep{nash1950equilibrium}.
By definition, the NE strategy is a stationary point where no player has the incentive
to deviate from its current strategy. Due to the minimax theorem, the NE strategy
for one player is also the best solution when facing against the best response of
the opponent. That is, NE is a natural solution that is free from the exploitation
by adversarial opponents.

The concepts of Nash equilibrium and non-exploitability have been well studied in
the community of learning \emph{extensive-form games} (EFGs) such as Poker
\citep{heinrich2016deep, brown2019superhuman, zha2021douzero, mcaleer2021xdo}.
Distinct from EFGs which feature tree-structured transition dynamics and do not
\emph{efficiently} represent the games where multiple states in the past may lead
to the same state in the future, this paper focuses on MGs with general
transition structure, and leverages the Markov structures. Another line of prior
works \citep{heinrich2016deep, lanctot2017unified} directly combine the best-response-based
algorithm for finding NE in normal-form games, such as fictitious play~\citep{brown1951iterative}
and double oracle~\citep{mcmahan2003planning}, with single-agent deep RL algorithm
such as DQN~\citep{mnih2013playing} and PPO~\citep{schulman2017proximal} for finding
the best-response. While these approaches can be applied to MGs, they do not utilize
the fine structure of MGs beyond treating it as normal-form games, which leads
to the significant inefficiency in learning (as shown in both tabular and Atari experiments
of this paper).

This paper proposes two novel, end-to-end deep reinforcement learning algorithms
for learning the Nash equilibrium of two-player zero-sum Markov games---Nash Deep
Q-Network (\nd), and its variant Nash Deep Q-Network with Exploiter (\nde). \nd combines
the recent theoretical progress for learning Nash equilibria of tabular Markov games
\citep{hu2003nash, bai2020provable, liu2021sharp, jin2021power}, with well-known
single-agent DRL algorithm DQN \citep{mnih2013playing} for addressing continuous
state space and function approximation. \nde is a variant of \nd which explicitly
train an exploiting opponent during its learning. The exploiting opponent
stimulates the exploration for the main agent. Both algorithms are the practical
variants of theoretical algorithms which are guaranteed to converge to Nash equilibria
in the basic tabular setting.

\begin{figure}[tp]
    \centering
    \includegraphics[width=\columnwidth]{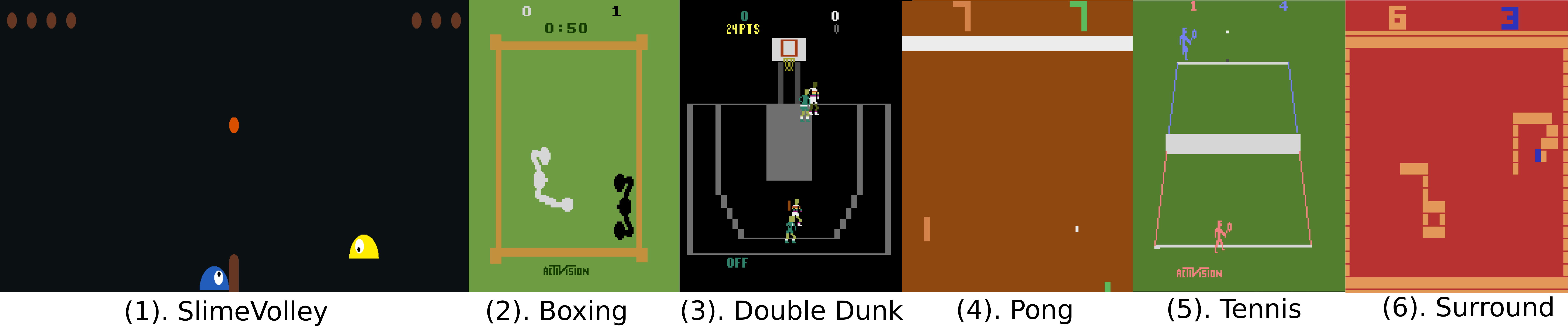}
    \caption{Screen shots of the six two-player video games.}
    \label{fig:games}
\end{figure}

Experimental evaluations are conducted on both tabular Markov games and two-player
video games, to show the effectiveness and robustness of the proposed algorithms.
As shown in Fig.~\ref{fig:games}, the video games in our experiments include
five two-player Atari games in PettingZoo library \citep{pettingzoo, Bellemare_2013}
and a benchmark environment Slime Volley-Ball \citep{slimevolleygym}. Due to the
constraints of computational resource, we consider the RAM-based version of
Atari games, and truncate the length of each game to 300 steps. We test the
performance by training adversarial opponents using DQN that directly exploit
the learner's policies. Our experiments in both settings show that our algorithms
significantly outperform standard algorithms for MARL including Neural Fictitious
Self-Play (NFSP) \citep{heinrich2016deep} and Policy Space Response Oracle (PSRO)
\citep{lanctot2017unified}, in terms of the robustness against adversarial
exploitation.

\section{Related Works}
\label{sec:related} In this section, we review the related works.

\textbf{MARL in cooperative games.} A rich literature of MARL has been focused
on the cooperative setting, where all players share the same objective, and seek
to maximize this objective jointly. Many empirical or theoretical algorithms have
been developed, such as VDN~\cite{sunehag2017value}, COMA~\cite{foerster2018counterfactual},
QMIX~\cite{rashid2018qmix}, MADDPG~\cite{lowe2017multi}, MAPPO~\cite{yu2021surprising}.
In contrast, our paper focus on the competitive setting and the key challenge is
to find a policy that is non-exploitable. Most of the algorithms designed for the
cooperative games do not have mechanisms to handle the adversarial exploitation of
the opponents.

\textbf{MARL in IIEFGs.} This line of work \cite{lanctot2009monte, heinrich2015fictitious, brown2019superhuman, farina2020stochastic, mcaleer2021xdo, kozuno2021model, bai2022near}
focuses on learning Imperfect Information Extensive-Form Games (IIEFGs). This line
of results focuses on finding Nash equilibrium, which is non-exploitable. However,
a majority of these results focus on the settings and applications (such as Poker)
where state space is discrete. More importantly, comparing to MGs, the model of IIEFGs
made a strong assumption of transition which must be tree-formed. The results of
IIEFG typically scales super-linearly with respect to the number of information
sets, which in general grows exponentially with the horizon length of the game.

\textbf{MARL in zero-sum MGs.} There has not been much prior empirical efforts
to design algorithms specializing in solving zero-sum MGs. However, there is a
rich class of empirical algorithms that can be directly applied to this setting.
These algorithms involve combining single-agent RL algorithm, such as deep Q-network
(DQN)~\citep{mnih2013playing} or proximal policy optimisation (PPO)~\citep{schulman2017proximal},
with best-response-based Nash equilibrium finding algorithm for \emph{normal-form}
games, including fictitious play (FP)~\citep{brown1951iterative}, double oracle (DO)~\citep{mcmahan2003planning},
and many others. A few other examples such as neural fictitious self-play (NFSP)~\citep{heinrich2016deep},
policy space response oracles (PSRO)~\citep{lanctot2017unified}, online double
oracle~\citep{dinh2021online} and prioritized fictitious self-play \citep{vinyals2019grandmaster}
also in general fall into this class of algorithms or their variants. These algorithms
call single-agent RL algorithm to compute the best response of the current ``meta-strategy'',
and then use the best-response-based algorithm for normal-form games to compute
a new ``meta-strategy''. However, these algorithms inherently treat MGs as
normal-form games, do not efficiently utilize the finer structure within MGs.
Specifically, for normal-form game, FP has a convergence rate exponential in the
number of actions, while DO is linear. However, a direct converting of Markov game
to normal-form game will generate a new action space \textbf{exponential} in
horizon, number of states and number of actions in original Markov game. This, as
shown in our experiments, leads to significant inefficiency in scaling up with
size of the MGs.

On the other hand, there has been rich studies on two-player zero-sum Markov games
from the theoretical perspectives. Many of these works \cite{hu2003nash, bai2020provable, bai2020near, liu2021sharp, jin2021v}
focused on the tabular setting, which requires the numbers of states and actions
to be finite. These algorithms are proved to converge to the Nash equilibria
policies in a number of samples that is \textbf{polynomial} in the number of states,
actions, horizon (or the discount coefficient), and the target accuracy. \reb{Among those, Nash Q-learning \cite{hu2003nash} is one of the earliest works along this line of research, which provably converges to NE for general-sum games under the assumption that the NE is unique for each stage game during learning process.}
On the other hand, \reb{\golf \cite{jin2021power} is another theoretical work with provable polynomial convergence for two-player zero-sum Markov games.}
Our Nash DQN algorithm is designed based on the provable tabular algorithm Nash
Value Iteration \cite{liu2021sharp}, which is a natural extension of value iteration
algorithm from single-agent setting to the multi-agent setting. \reb{For better understanding, we provide a detailed comparison of similarities and differences of Nash Q-learning, Nash Value Iteration, \golf and Nash DQN}
in the Sec. \ref{app:sec_compare_tabular_algs}. \cite{xie2020learning}
considers Markov games with linear function approximation. There are a few theoretical
works on studying zero-sum Markov games with general function approximation \cite{jin2021power,
huang2021towards}, which include neural network function approximation as
special cases. However, these algorithms are sample-efficient, but not computationally
efficient. They require solving optimistic policies with complicated confidence
sets as constraints, which can not be run in practice.

\textbf{Single-agent deep reinforcement learning.} Deep RL algorithms have been
demonstrated effectively for optimizing polices in the single-agent setting. Off-policy
algorithms like deep Q-network (DQN)~\citep{mnih2013playing} and its variants
have shown great superhuman performances on real-time Atari video games~\citep{badia2020agent57}.
On-policy algorithms like proximal policy optimisation (PPO)~\citep{schulman2017proximal}
has been widely applied for complex tasks with both discrete and continuous action
spaces. We refer readers to \citep{sutton2018reinforcement, arulkumaran2017deep, lillicrap2015continuous, dong2020deep}
and the references therein for more algorithms and details about single-agent deep
RL.

\section{Preliminaries}
\label{sec:prelim} In this paper, we consider Markov Games \citep[MGs,][]{shapley1953stochastic, littman1994markov},
which generalizes standard Markov Decision Processes (MDPs) into multi-player
settings. Each player has its own utility and optimize its policy to maximize the
utility. We consider a special setting in MG called two-player zero-sum games, which
has a competitive relationship between the two players.


More concretely, consider a infinite-horizon discounted version of two-player zero-sum
MG, which is denoted as $\rm{MG}(\cS, \cA, \cB, \P, r, \gamma)$. $\mathcal{S}$ is
the state space, $\mathcal{A}$ and $\mathcal{B}$ are the action spaces for the
max-player and min-player respectively. $\P$ is the state transition distribution,
and $\P ( \cdot | s, a, b)$ is the distribution of the next state given the
current state $s$ and action pair $(a, b)$. $r\colon \cS \times \cA \times \cB \to
\mathbb{R}$ is the reward function. In the zero-sum setting, the reward is the gain
for the max-player and the loss for the min-player due to the zero-sum payoff structure.
$\gamma\in[0,1]$ is the discount factor. At each step, the two players will
observe the state $s \in \cS$ and choose their actions $a \in \cA$ and
$b \in \cB$ independently and simultaneously. The action from their opponent can
be observed after they take the actions, and the reward $r(s, a, b)$ will be
received ($r$ for max-player and $-r$ for min-player). The environment then transit
to the next state $s'\sim\P(\cdot | s, a, b)$.

\paragraph{Policy, value function.}
We define the policy and value functions for each player. For the max-player,
the (Markov) policy is a map $\mu: \cS \rightarrow \Delta_{\cA}$. Here we only consider
discrete action space, so $\Delta_{\cA}$ is the probability simplex over action
set $\cA$. Similarly, the policy for the min-player is
$\nu: \cS \rightarrow \Delta_{\cB}$.

$V^{\mu, \nu}\colon \cS \to \mathbb{R}$ represents the value function evaluated
with policies $\mu$ and $\nu$, which can be expanded as the expected cumulative reward
starting from the state $s$,
\begin{equation}
    \label{eq:V_value}
    \begin{aligned}
        V^{\mu, \nu}(s) \defeq \E_{\mu, \nu}\bigg[\sum_{h = 1}^{\infty} \gamma^{h-1}r(s_{h}, a_{h}, b_{h}) \bigg| s_{1} = s\bigg].
    \end{aligned}
\end{equation}
Correspondingly, $Q^{\mu, \nu}:\cS \times \cA \times \cB \to \mathbb{R}$ is the state-action
value function evaluated with policies $\mu$ and $\nu$, which can also be
expanded as expected cumulative rewards as:
\begin{equation}
    \label{eq:Q_value}
    \begin{aligned}
         & \quad Q^{\mu, \nu}(s, a, b) \defeq \E_{\mu, \nu}\bigg[\sum_{h = 1}^{\infty} \gamma^{h-1}r(s_{h}, a_{h}, b_{h}) \big| s_{1} = s, a_{1} = a, b_{1} = b\bigg].
    \end{aligned}
\end{equation}
In this paper we also use a simplified notation for convenience,
$[\P V](s, a, b) \defeq \E_{s' \sim \P(\cdot|s, a, b)}V(s')$, where $\P$ as the transition
function can be viewed as an operator. Similarly, we denote $[\D_{\pi} Q](s) \defeq
\E_{(a, b) \sim \pi(\cdot, \cdot|s)}Q(s, a, b)$ for any state-action value function.
In this way, the Bellman equation for two-player MG can be written as:
\begin{equation}
    Q^{\mu, \nu}(s, a, b) = (r + \gamma \P V^{\mu, \nu})(s, a, b), \quad V^{\mu,
    \nu}(s) = (\D_{\mu\times\nu}Q^{\mu, \nu})(s), \label{eq:bellman}
\end{equation}
for all $(s, a, b) \in \cS \times \cA \times \cB$.

\paragraph{Best response and Nash equilibrium.}
In two-player games, if the other player always play a fixed Markov policy, optimizing
over learner's policy is the same as optimizing over the policy of single agent in
MDP (with other players' polices as a part of the environment). For two-player cases,
given the max-player's policy $\mu$, there exists a \emph{best response} of the
min-player, which is a policy $\nu^{\dagger}(\mu)$ satisfying $V^{\mu, \nu^\dagger(\mu)}
(s) = \inf_{\nu}V^{\mu, \nu}(s)$ for any $s \in \cS$. We simplify the notation
as: $V^{\mu, \dagger}\defeq V^{\mu, \nu^\dagger(\mu)}$. Similar best response
for a given min-player's policy $\nu$ also exists as $\mu^{\dagger}(\nu)$
satisfying $V^{\dagger, \nu}=\sup_{\mu} V^{\mu, \nu}$. By leveraging the Bellman
equation Eq.~\eqref{eq:bellman}, the best response can be derived with dynamic
programming,
\begin{equation}
    \label{eq:best_response_v}Q^{\mu, \dagger}(s,a,b) = (r+\gamma \mathbb{P}V^{\mu,
    \dagger})(s,a,b), \quad V^{\mu, \dagger}(s) = \inf_{\nu}(\mathbb{D}_{\mu\times
    \nu}Q^{\mu, \dagger})(s)
\end{equation}

The \emph{Nash equilibrium} (NE) is defined as a pair of policies $(\mu^{\star},\nu
^{\star})$ serving as the optimal against the best responses of the opponents,
indicating:
\begin{equation}
    \begin{aligned}
        V^{\mu^\star, \dagger}(s) = \sup_{\mu}V^{\mu, \dagger}(s), \quad V^{\dagger, \nu^\star}(s) = \inf_{\nu}V^{\dagger, \nu}(s),
    \end{aligned}
\end{equation}
for all $s \in \cS$. The existence of NE is shown by previous work~\cite{filar2012competitive}.
Furthermore, NE strategies satisfy the following minimax equation:
\begin{equation}
    \textstyle \sup_{\mu}\inf_{\nu}V^{\mu, \nu}(s) = V^{\mu^\star, \nu^\star}(s)
    = \inf_{\nu}\sup_{\mu}V^{\mu, \nu}(s).
\end{equation}
which is similar as the normal-form game but without the bilinear structure of the
payoff matrix. NE strategies are the ones where no player has incentive to
change its own strategy. The value functions of $(\mu^{\star},\nu^{\star})$ is denoted
as $V^{\star}$ and $Q^{\star}$, which satisfy the following Bellman optimality
equation:
\begin{equation}
    \label{eq:nash_q}
    \begin{aligned}
        Q^{\star}(s,a,b)                                                      & = (r+\gamma\mathbb{P}V^{\mu, \dagger})(s,a,b)                                                                                \\
        V^{\star}(s) = \sup_{\mu \in \Delta_{\cA}}\inf_{\nu \in \Delta_{\cB}} & (\D_{\mu \times \nu}Q^{\star})(s) = \inf_{\nu \in \Delta_{\cB}}\sup_{\mu \in \Delta_{\cA}}(\D_{\mu \times \nu}Q^{\star})(s).
    \end{aligned}
\end{equation}

\paragraph{Learning Objective.}
The \emph{exploitability} of policy $(\hat{\mu}, \hat{\nu})$ can be defined as
the difference in values comparing to Nash strategies when playing against their
best response. Formally, the exploitability of the max-player can be defined as
$V^{\star}(s_{1}) - V^{\hat{\mu},\dagger}(s_{1})$ while the exploitability of
the min-player is defined as $V^{\dagger, \hat{\nu}}(s_{1}) - V^{\star}(s_{1})$.
We define the total suboptimality of $(\hat{\mu}, \hat{\nu})$ simply as the summation
of the exploitability of both players
\begin{equation}
    \label{eq:subopt}V^{\dagger, \hat{\nu}}(s_{1}) - V^{\hat{\mu}, \dagger}(s_{1}
    ) = \brac{V^{\dagger, \hat{\nu}}(s_1) - V^{\star}(s_1)}+ \brac{V^{\star}(s_1) - V^{\hat{\mu},\dagger}(s_1)}
    .
\end{equation}
This quantity is also known as the duality gap in the literature of MGs, which
can be viewed as a distance measure to Nash equilibria. We note that the duality
gap of Nash equilibria is equal to zero. Furthermore, all video games we conduct
experiments on are symmetric to two players, which implies that
$V^{\star}(s_{1}) = 0$.

\paragraph{\nv and Nash Q-Learning.}

The model-based tabular algorithm for MGs---\nv, computes the near-optimal policy
by performing Bellman optimality update Eq.~\eqref{eq:nash_q} with $(\P, r)$ replaced
by their empirical estimates using samples. The empirical estimates of the entry
$\hat{P}(s'|s, a, b)$ in transition matrix is computed by how many times
$(s, a, b, s')$ is visited divided by how many times $(s, a, b)$ is visited. Please
see Algorithm \ref{alg:sec_nvi} in Sec. for more details.
Similar to the relation between Q-learning and VI in the single-agent setting, Nash
Q-Learning \citep{hu2003nash} is the model-free version of \nv, which performs incremental
update
\[
    Q(s, a, b)\leftarrow (1-\alpha) Q(s, a, b) + \alpha(r + \gamma \cdot{\rm Nash}
    (Q(s', \cdot, \cdot))
\]
whenever a new sample $(s, a, b, r, s')$ is observed.

\section{Methodology}

To learn the Nash equilibria of two-player zero-sum Markov games, this paper proposes
two novel, end-to-end deep MARL algorithms---\nd and \nde. \nd combines single-agent
DQN \citep{mnih2013playing} with \nv \citep{liu2021sharp}---a provable algorithm
for tabular Markov games. \nde is a variant of \nd by explicitly training an
adversarial opponent during the learning phase to encourage the exploration of
the learning agent. \reb{The details of \nv is discussed in Sec. \ref{app:sec_nvi}.}

\subsection{\nd}
\label{subsec:nash_dqn}

\begin{algorithm}
    [t]
    \caption{Nash Deep Q-Network (\nd)}
    \begin{algorithmic}
        [1] \STATE Initialize replay buffer $\cD=\emptyset$, counter $i=0$, Q-network
        $Q_{\phi}$ \STATE Initialize target network parameters:
        $\phi^{\text{target}}\leftarrow \phi$. \FOR{episode $k=1,\ldots,K$}
        \STATE reset the environment and observe $s_{1}$. \FOR{$t=1,\ldots,H$}
        \STATE {\color{blue}\% collect data} \STATE sample actions
        $(a_{t}, b_{t})$ from $\begin{cases}
            \text{Uniform}(\cA \times \cB)                                 & \quad \text{with probability~}\epsilon \\
            (\mu_{t}, \nu_{t})=\textit{Nash}(Q_{\phi}(s_{t},\cdot, \cdot)) & \quad \text{otherwise.}
        \end{cases}$ \STATE execute actions $(a_{t}, b_{t})$, observe reward $r_{t}$,
        next state $s_{t+1}$. \STATE store data sample $(s_{t},a_{t},b_{t},r_{t},
        s_{t+1})$ into $\cD$ \STATE {\color{blue}\% update Q-network} \STATE randomly
        sample minibatch $\cM \subset \{1, \ldots, |\mathcal{D}|\}$. \FOR{all $j \in \minibatch$}
        \STATE compute $(\hat{\mu}, \hat{\nu})=\textit{Nash}(Q_{\phi^{\text{target}}}
        (s_{j+1}, \cdot, \cdot))$ \STATE \alglinelabel{line:target_nash_DQN} set
        $y_{j} = r_{j} + \gamma \hat{\mu}\trans Q_{\phi^{\text{target}}}(s_{j+1},
        \cdot, \cdot) \hat{\nu}$. \ENDFOR \STATE Perform $m$ steps of GD on loss
        $\sum_{j \in \minibatch}(y_{j}-Q_{\phi}(s_{j},a_{j},b_{j}))^{2}$ to update
        $\phi$. \alglinelabel{line:square_loss_nd} \STATE {\color{blue}\% update target network}
        \STATE $i=i+1$; if $i\%N=0$: $\phi^{\text{target}}\leftarrow \phi$.
        \ENDFOR \ENDFOR
    \end{algorithmic}
    \label{alg:nash_dqn}
\end{algorithm}

We describe \nd~in Algorithm \ref{alg:nash_dqn} which incorporates neural networks
into the tabular \nv \citep{liu2021sharp} algorithm for approximating the Q-value
function. Similar to the single-agent DQN, \nd~maintains two networks in the
training process: the $Q$-network and its target network, which are
parameterized by $\phi$ and $\phi^{\text{target}}$, respectively. In each
episode, \nd~executes the following two main steps:

$\bullet$ \textbf{Data collection:} \nd adopts the $\epsilon$-greedy strategy
for exploration. At each state $s_{t}$, with probability $\epsilon$, both players
take random actions; otherwise, they will sample actions $(a_{t},b_{t})$ from the
Nash equilibrium of its Q-value matrix (i.e.,
$\textit{Nash}(Q_{\phi}(s_{t},\cdot, \cdot))$, see \eqref{eq:nash_subroutine}).
After that, we add the collected data into the experience replay buffer $\cD$.

$\bullet$ \textbf{Model update}: We first randomly sample a batch of data $\cM$ from
replay buffer $\cD$, and then perform $m$-steps gradient descent to update
$\phi$ using the loss below
\[
    \textstyle \sum_{j \in\cM}\left(Q_{\phi}(s_{j},a_{j},b_{j})-y_{j}\right)^{2},
\]
where target $y_{j}$ is computed according to line 14. We adopt the convention
of setting $Q_{\phi^{\text{target}}}(s_{j+1}, \cdot, \cdot) = 0$ for all
terminal state $s_{j+1}$.

Here, $\textit{Nash}(\cdot)$ is the NE subroutine for normal-form games, which
takes a payoff matrix $\A \in \R^{A \times B}$ as input and outputs one of its
Nash equilibria $(\mu^{\star}, \nu^{\star})$. In math, we have:
\begin{equation}
    \label{eq:nash_subroutine}(\mu^{\star}, \nu^{\star}) = \textit{Nash}(\A) \quad
    \text{if and only if}\quad \forall \mu, \nu, ~~ \mu\trans \A \nu^{\star} \le
    (\mu^{\star})\trans \A \nu^{\star} \le (\mu^{\star})\trans \A \nu.
\end{equation}
There are several off-the-shelf libraries to implement this $\textit{Nash}$ subroutine.
After comparing the performance of several different implementations (see Sec.~\ref{sec:choose_solver}
for details), we found the ECOS library \citep{domahidi2013ecos} works the best,
which from now on is set as the default choice in our algorithms.

Regarding the choice of target value update (line \ref{line:target_nash_DQN}),
one can view it as a Monte Carlo estimate of
\begin{equation}
    \label{eq:nash_dqn_target_expectation}\textstyle r(s_{j},a_{j},b_{j})+ \gamma
    \E_{s'\sim\P(\cdot\mid s_j,a_j,b_j)}[ \max_{\hat\mu\in\Delta_A}\min_{\hat\nu\in\Delta_B}
    \hat\mu^{\intercal} Q_{\phi^{\text{target}}}(s', \cdot, \cdot)\hat\nu].
\end{equation}
Intuitively, we aim to approximate the Q-value function of Nash equilibria $Q^{\star}$
by our Q-network $Q_{\phi}$. Recall that $Q^{\star}$ is the \emph{unique} solution
of the Bellman optimality equations:
\begin{equation}
    \textstyle \label{eq:Qstar}\forall (s,a,b), \quad Q^{\star}(s,a,b) = r(s,a,b)
    + \gamma\E_{{s}'\sim\P(\cdot\mid s,a,b)}[ \max_{\hat\mu\in\Delta_A}\min_{\hat\nu\in\Delta_B}
    \hat\mu^{\intercal} Q^{\star}({s}', \cdot, \cdot)\hat\nu].
\end{equation}
As a result, by performing gradient descent on $\phi$ to minimize the square loss
as in line \ref{line:square_loss_nd}, $Q_{\phi}$ will decrease its Bellman error,
and eventually converge to $Q^{\star}$, as more samples are collected. Finally,
we remark that the target Nash $Q$-network ($Q_{\phi^{\text{target}}}$) is
updated in a delayed manner as DQN to stabilize the training process.

\subsection{\nde}
\label{subsec:nash_dqn_exploiter}

\begin{algorithm}
    [t]
    \caption{Nash Deep Q-Network with Exploiter (\nde)}
    \begin{algorithmic}
        [1] \STATE Initialize replay buffer $\cD=\emptyset$, counter $i=0$, Q-network
        $Q_{\phi}$, exploiter network $\tilde{Q}_{\psi}$. \STATE Initialize target
        network parameters: $\phi^{\text{target}}\leftarrow \phi$, $\psi^{\text{target}}
        \leftarrow \psi$. \FOR{episode $k=1,\ldots,K$} \STATE reset the environment
        and observe $s_{1}$. \FOR{$t=1,\ldots,H$} \STATE {\color{blue}\% collect data}
        \STATE sample actions $(a_{t}, b_{t})$ from $\begin{cases}
            \text{Uniform}(\cA \times \cB)                                             & \quad \text{w.p.~}\epsilon \\
            (\mu_{t}, \nu_{t}) \text{~computed according to \eqref{eq:compute_policy}} & \quad \text{otherwise.}
        \end{cases}$ \STATE execute actions $(a_{t}, b_{t})$, observe reward
        $r_{t}$, next state $s_{t+1}$. \STATE store data sample
        $(s_{t},a_{t},b_{t},r_{t},s_{t+1})$ into $\cD$. \STATE {\color{blue}\% update Q-network and exploiter network}
        \STATE randomly sample minibatch $\minibatch \subset \{1, \ldots, |\cD|\}$.
        \FOR{all $j \in \mathcal{M}$} \STATE compute $(\hat{\mu}, \hat{\nu})=\textit
        {Nash}(Q_{\phi^{\text{target}}}(s_{j+1}, \cdot, \cdot))$ \STATE set $y_{j}
        = r_{j} + \gamma \hat{\mu}\trans Q_{\phi^{\text{target}}}(s_{j+1}, \cdot,
        \cdot)\hat{\nu}$. \STATE set
        $\tilde{y}_{j} = r_{j}+ \gamma \min_{b\in\mathcal{B}}\hat{\mu}\trans \tilde
        Q_{\psi^\text{target}}(s_{j+1},\cdot,b)$
        \alglinelabel{line:exploiter_target} \ENDFOR \STATE Perform $m_{1}$ steps
        of GD on loss $\sum_{j \in \minibatch}(y_{j}-Q_{\phi}(s_{j},a_{j},b_{j}))
        ^{2}$ to update $\phi$. \STATE Perform $m_{2}$ steps of GD on loss
        $\sum_{j \in \minibatch}(\tilde{y}_{j}-\tilde Q_{\psi}(s_{j},a_{j},b_{j})
        )^{2}$
        to update $\psi$. \alglinelabel{line:square_loss_nde} \STATE
        {\color{blue}\% update target network} \STATE $i=i+1$; if $i\%N=0$:
        $\phi^{\text{target}}\leftarrow \phi$, $\psi^{\text{target}}\leftarrow \psi$.
        \ENDFOR \ENDFOR
    \end{algorithmic}
    \label{alg:nash_dqn_exp}
\end{algorithm}

\nd relies on the $\epsilon$-greedy strategy for exploration. To improve the exploration
efficiency, we propose a variant of \nd --- \nde, which additionally introduces an
exploiter in the training procedure. By constantly exploiting the weakness of
the main agent, the exploiter forces the main agent to play the part of the games
she is still not good at, and thus helps the main agent improve and discover
more effective strategies.

We describe \nde in Algorithm \ref{alg:nash_dqn_exp}. We let the main agent maintain
a Q-network $Q_{\phi}$ and let the exploiter maintain a separate value network $\tilde
{Q}_{\psi}$, both are functions of $(s, a, b)$. We make two key modifications from
\nd. First, in the data collection phase, at state $s_{t}$, we no longer choose both
$\mu_{t},\nu_{t}$ to be the Nash equilibrium computed from $Q_{\phi}$. Instead, we
only choose $\mu_{t}$ to be the Nash strategy of $Q_{\phi}$ but pick the policy of
the exploiter $\nu_{t}$ to be the best response of $\mu_{t}$ under the exploiter's
$Q$-network $\tilde{Q}_{\psi}$. Formally,
\begin{align}
    \label{eq:compute_policy} & (\mu_{t}, \cdot) = \textit{Nash}(Q_{\phi}(s_{t},\cdot, \cdot))                            \\
                              & \nu_{t} = \argmin_{\nu} \mu_{t}\trans\tilde{Q}_{\psi}(s_{t}, \cdot, \cdot) \nu. \nonumber
\end{align}
In the model update phase, \nde follows exactly the same rule as \nd to update
$Q_{\phi}$ and $Q_{\phi^\text{target}}$, the $Q$-networks of the main agent.
However, for the update of the exploiter networks, \nde utilizes a different
regression target in the loss function as specified in line
\ref{line:exploiter_target}. One can view the target as a Monte Carlo estimate of
\begin{equation}
    \begin{aligned}
        \textstyle & r(s_{j},a_{j},b_{j})+\gamma \E_{s'\sim\P(\cdot\mid s_j,a_j,b_j)}\left[ \min_{b} \hat\mu(s')^{\intercal} \tilde{Q}_{\psi^{\text{target}}}(s', \cdot, b)\right] \label{eq:nash_dqn_exploiter_target_expectation}
    \end{aligned}
\end{equation}
We set the target in this way because we aim to approximate $Q^{\hat\mu,\dagger}$,
which is the value of the current policy of the main player $\hat{\mu}$ against
its best response, using our exploiter network $\tilde{Q}_{\psi}$. Recall that
$Q^{\hat\mu,\dagger}$ satisfies the following Bellman equations for the best
response:
\begin{equation}
    \textstyle \label{eq:Qdagger}\forall (s,a,b), \quad Q^{\hat\mu,\dagger}(s,a,b
    ) = r(s,a,b) + \gamma\E_{{s}'\sim\P(\cdot\mid s,a,b)}[ \min_{b}\hat\mu(s')^{\intercal}
    Q^{\hat\mu,\dagger}({s}', \cdot, b)].
\end{equation}
Therefore, by performing gradient descent on $\psi$ to minimize the square loss
as in line \ref{line:square_loss_nde}, $\tilde{Q}_{\psi}$ will decrease its (best
response version of) Bellman error, and approximate $Q^{\hat\mu,\dagger}$.

\subsection{Theoretical Justification}
\label{sec:theo_just}

With special choices of Q-network architecture $Q_{\phi}$, minibatch size
$|\minibatch|$ and number of steps for GD $m$, both our algorithms \nd and \nde reduce
to the $\epsilon$-greedy version of standard algorithm \nv \citep{liu2021sharp}
and \nve \citep{jin2021power} for learning tabular Markov games, where the
numbers of states and actions are finite and small. Please see Sec. \ref{sec:alg_tab_markov}
for a detailed discussion on the connections of these algorithms.

When replacing the $\epsilon$-greedy exploration with optimistic exploration (typically
in the form of constructing upper confidence bounds), both \nv and \nve are
guaranteed to efficiently find the Nash equilibria of MGs in the tabular
settings.

\begin{theorem}[\citep{liu2021sharp,jin2021power}]
    \label{thm:theory}For tabular Markov games, the optimistic versions of both \nv
    and \nve can find $\epsilon$-approximate Nash equilibria in $\text{poly}(S, A
    , B, (1-\gamma)^{-1}, \epsilon^{-1}, \log(1/\delta))$ steps with probability
    at least $1-\delta$. Here $S$ is the size of states, $A, B$ are the size of two
    players' actions respectively, and $\gamma$ is the discount factor.
\end{theorem}

We defer the proof of Theorem \ref{thm:theory} to Sec. \ref{sec:alg_tab_markov}.
We highlight that, in contrast, existing deep MARL algorithms such as NFSP
\citep{heinrich2016deep} or PSRO \citep{lanctot2017unified} are incapable of
efficient learning of Nash equilibria with a polynomial convergence rate for
tabular Markov games. Our simulation results reveal that they are indeed highly inefficient
in finding Nash equilibria.

\section{Experiments}
The experimental evaluations are conducted on randomly generated tabular Markov games
and two-player video games on Slime Volley-Ball \citep{slimevolleygym} and
PettingZoo Atari~\citep{pettingzoo}. We tested the performance of proposed
methods as well as the baseline algorithms in both (a) the basic tabular form
without function approximation (only in tabular environments); and (b) full versions
with deep neural networks (in both tabular environments and video games). For (a),
we measure the exploitability by computing the exact best response using the ground
truth transition and reward function. This is only feasible in the tabular
environment. For (b), we measure the exploitability by training single-agent DQN
(exploiter) against the learned policy to directly exploit it.

\subsection{Baselines}
For benchmarking purpose, we have the following baselines with deep neural network
function approximation for scalable tests:
\begin{itemize}
    \item \textbf{Self-Play} (SP): each agent learns to play the best response strategy
        against the fixed opponent strategy alternatively, i.e., iterative best
        response.

    \item \textbf{Fictitious Self-Play} (FSP)~\cite{heinrich2015fictitious}: each
        agent learns a best-response strategy against the episodic average of its
        opponent's historical strategy set, and save it to its own strategy set.

    \item \textbf{Neural Fictitious Self-Play} (NSFP)~\cite{heinrich2016deep}: an
        neural network approximation of FSP, a policy network is explicitly
        maintained to imitate the historical behaviours by an agent, and the learner
        learns the best response against it.

    \item \textbf{Policy Space Response Oracles} (PSRO)~\cite{lanctot2017unified}:
        we adopt a version based on double oracle (DO), each agent learns the
        best response against a meta-Nash strategy of its opponent's strategy
        set, and add the learned strategy to its own strategy set.
\end{itemize}

For tabular case without function approximation, SP, FSP and DO are implemented with
Q-learning as the base learning agents for finding best responses. For tabular
case with function approximation and video games, all four baseline methods use DQN
as the basic agent for learning the best-response strategies. The pseudo-codes
for algorithms SP, FSP, DO are provided in Sec.~\ref{sec:alg_tab_markov}.

\subsection{Tabular Markov Game}
\textbf{Tabular forms without function approximation.} We first evaluate methods
(1) SP, (2) FSP, (3) PSRO, (4) \nd and (5) \nde without function approximation (i.e.,
w/o neural network) on the tabular Markov games. They reduce to methods (1) SP,
(2) FSP, (3) DO, (4) Nash value iteration (\nv) and (5) Nash value iteration
with exploiter (\nve), correspondingly. For SP, FSP and DO, we adopt Q-learning
as a subroutine for finding the best response policies. As tabular versions of our
deep MARL algorithms, \nv and \nve also use $\textit{Nash}$ subroutine for
calculating NE in normal-form games, with $\epsilon=0.5$ for $\epsilon$-greedy exploration.
The pseudo-codes for \nv and \nve are provided in Sec.
\ref{sec:alg_tab_markov}.

We randomly generated the tabular Markov games, which has discrete state space $\mathcal{S}$,
discrete action spaces $\mathcal{A}, \mathcal{B}$ for two players and the
horizon $H$\footnote{We encode the horizon into the state space in order to use
the algorithm designed for the discounted setting.} The state transition probability
function $\{\mathcal{T}_{h}:\mathcal{S}\times \mathcal{A}\times \mathcal{S}\rightarrow
[0,1], h\in[H]\}$ and reward function $\{R_{h}:\mathcal{S}\times \mathcal{A}\times
\mathcal{S}\rightarrow [-1,1], h\in[H]\}$ are both i.i.d sampled uniformly over their
ranges.
\begin{figure}[htbp]
    \centering
    \includegraphics[width=0.45\columnwidth]{
        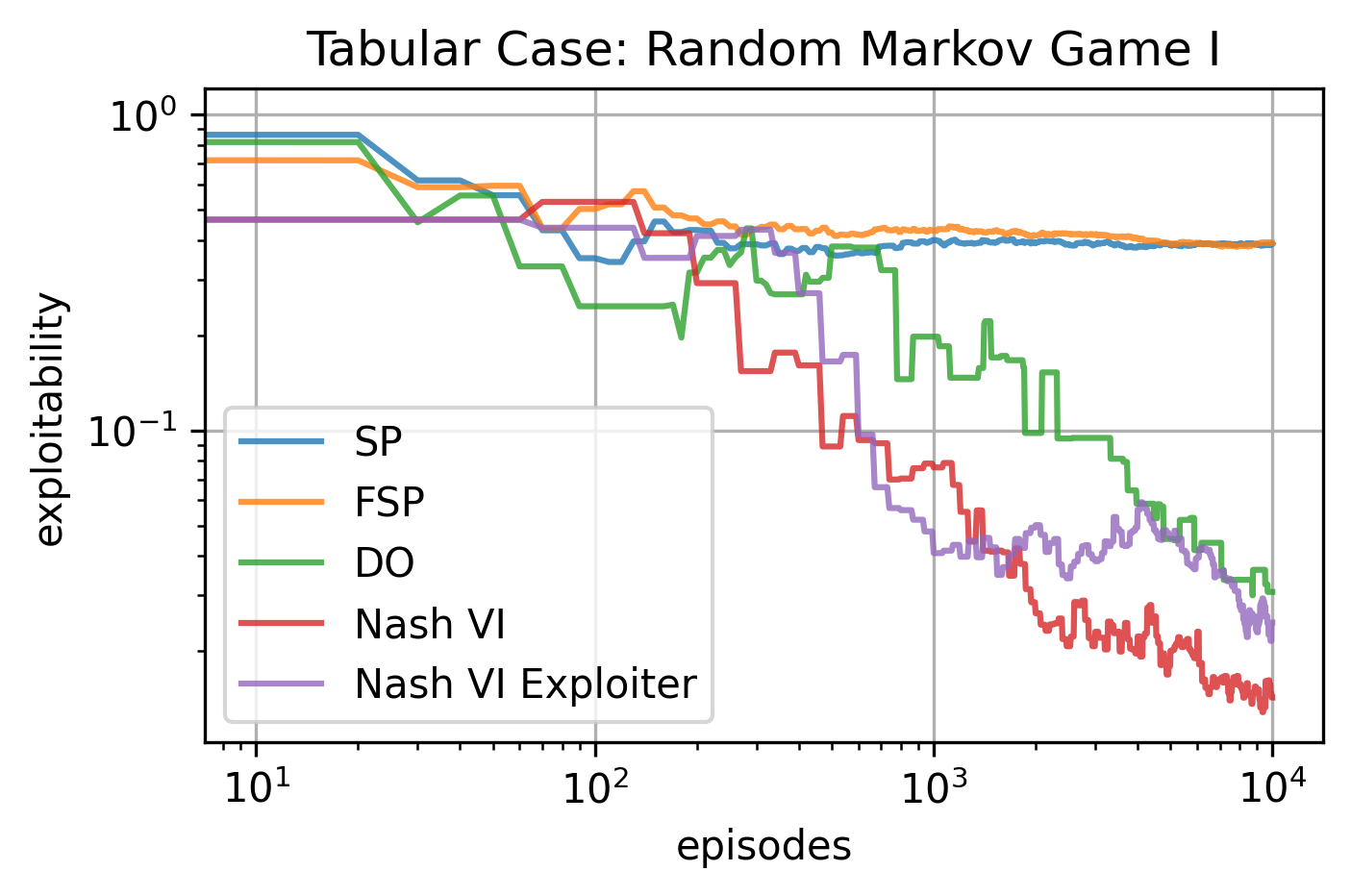
    }
    \hspace{0.1in}
    \includegraphics[width=0.45\columnwidth]{
        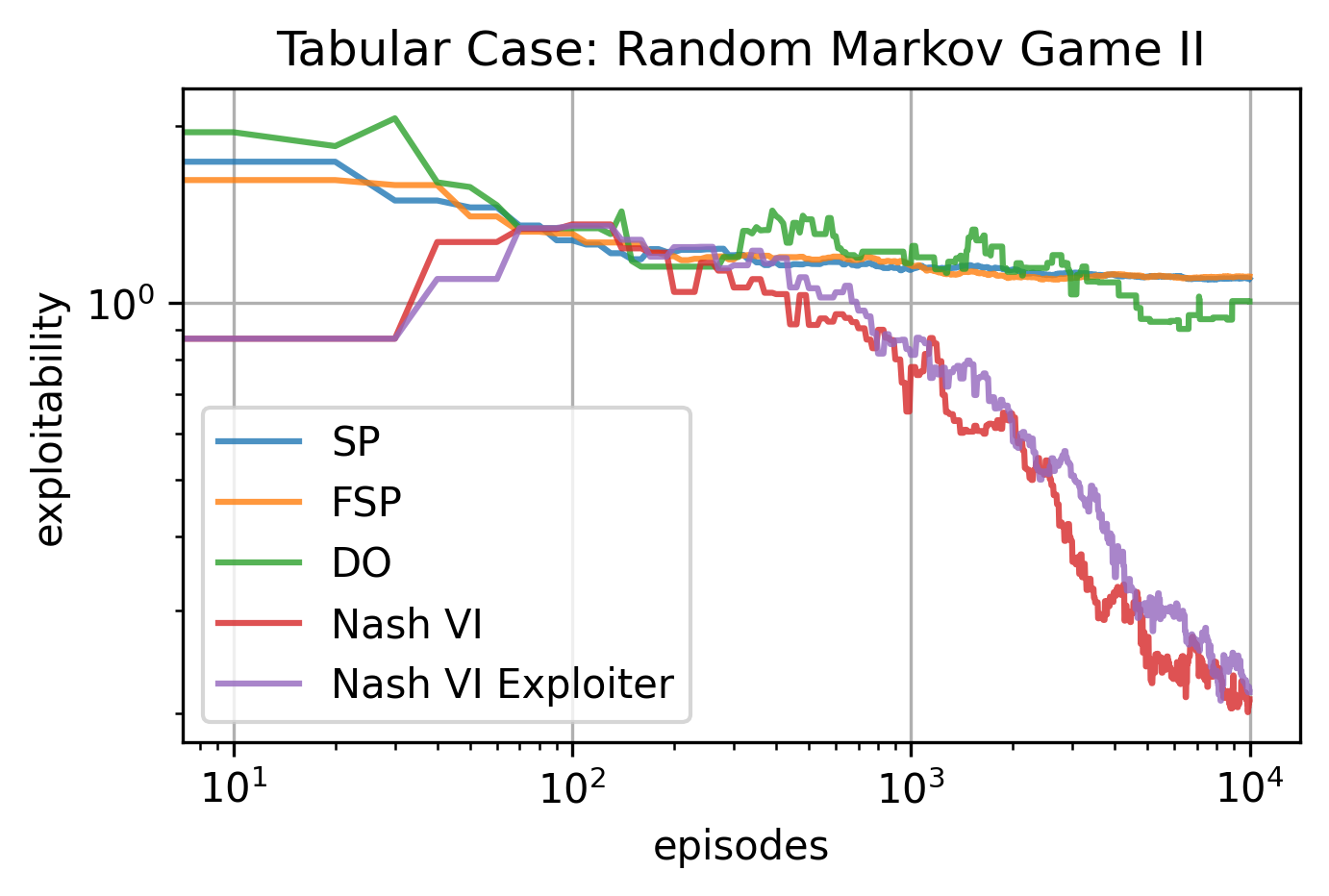
    }
    \caption{Tabular case experiments on two randomly generated Markov games.}
    \label{fig:tabular_nvi} \vskip -.05in
\end{figure}
As shown in Fig.~\ref{fig:tabular_nvi}, we tested on two randomly generated
Markov games of different sizes: I.
$|\mathcal{S}|=|\mathcal{A}|=|\mathcal{B}|=H=3$; II.
$|\mathcal{S}|=|\mathcal{A}|=|\mathcal{B}|=H=6$. The exploitability is calculated
according to Eq.~\eqref{eq:subopt}, which can be solved with dynamic programming
in the tabular cases with known transition and reward functions. Our two proposed
algorithms without function approximation show significant speedup for decreasing
the exploitability compared against other baselines, especially for the larger
environment (II). This aligns with our theoretical justification as in Section \ref{sec:theo_just}.

\textbf{With neural networks as function approximation.} In this set of
experimentation, we add neural networks as function approximators. We evaluate
\nd and \nde on the same tabular MG environments I and II. We setup the neural-network
versions of baseline methods---FSP, NFSP and PSRO, using the same set of hyperparameters and the same training configurations.
During training, the model checkpoints are saved at different stages for each method
and reloaded for exploitation test. Each method is trained for a total of
$5\times 10^{4}$ episodes to get the final model, against which the exploiter is
trained for $3\times 10^{4}$ episodes as the exploitation test. The exploiter is
a DQN agent trained from scratch with the same hyperparameters as the base
agents (for finding best responses) in FSP, NFSP, PSRO. We empirically measure the
exploitability of the learner's policy as follows: we first compute smoothed
version of cumulative utility achieved by the exploiter at each episode (the smoothing
is conducted by averaging over a small set of neighboring episodes); we then report
the highest smoothed cumulative utility of the exploiter as an approximation for
the exploitability. We also test the effectiveness of using DQN exploiter to
measure the exploitability, by training it against the oracle Nash strategies (i.e.,
the ground-truth Nash equilibria). Results for two tabular environments are
displayed in Table~\ref{tab:tabular_exploit_compare128}. As shown in Table~\ref{tab:tabular_exploit_compare128},
both \nd and \nde outperform all other methods by a significant margin. The negative
exploitability values of the Oracle Nash strategy indicates that the DQN-based exploiter
is not able to find the exact theoretical best response. Nevertheless, it approximates
the best response very well. The reported exploitability of Oracle Nash is very
close to zero, which justifies the effectiveness of using DQN for exploitation tests.

\begin{table*}
    [htbp]
    \centering
    \caption{\reb{Approximate exploitability} (lower is better) in two tabular Markov
    games}
    \resizebox{\textwidth}{!}{ 
    \begin{tabular}{ c|>{\centering\arraybackslash}m{22pt}|>{\centering\arraybackslash}m{22pt}|>{\centering\arraybackslash}m{30pt}|>{\centering\arraybackslash}m{30pt}|>{\centering\arraybackslash}m{50pt}|>{\centering\arraybackslash}m{90pt}|>{\centering\arraybackslash}m{50pt}
    }
        \hline
        \backslashbox{Env}{Method} & SP      & FSP     & NFSP    & PSRO    & \textbf{Nash DQN}            & \textbf{Nash DQN Exploiter}  & Oracle Nash \\
        \hline
        Tabular Env I              & $0.448$ & $0.379$ & $0.317$ & $0.134$ & $0.096$                      & $\color{blue}\mathbf{0.020}$ & $-0.027$    \\
        \hline
        Tabular Env II             & $1.239$ & $0.694$ & $0.379$ & $0.569$ & $\color{blue}\textbf{0.017}$ & $0.071$                      & $-0.082$    \\
        \hline
    \end{tabular}
    } \vskip -.2in \label{tab:tabular_exploit_compare128}
\end{table*}

\subsection{Two-Player Video Game}
\label{sec:video_game} To evaluate the scalability and robustness of the proposed
method, we examine all algorithms in five two-player Atari environments~\citep{Bellemare_2013}
in PettingZoo library~\citep{pettingzoo} (\textit{Boxing-v1}, \textit{Double
Dunk-v2}, \textit{Pong-v2}, \textit{Tennis-v2}\footnote{The original \textit{Tennis-v2}
environment in PettingZoo is not zero-sum, a reward wrapper is applied to make
it.}, \textit{Surround-v1}) and in environment \textit{SlimeVolley-v0} in a public
available benchmark named Slime Volley-Ball~\citep{slimevolleygym}, as shown in Fig.~\ref{fig:games}.
The algorithms tested for this setting include: (1) SP, (2) FSP, (3) NFSP, (4) PSRO,
(5) \nd and (6) \nde. To speed up the experiments, each environment is truncated
to 300 steps per episode for both training and exploitation. For Atari games,
the observation is based on RAM and normalized in range $[0,1]$.

Similar to experiments in the tabular environment with function approximation, the
exploitation test (using single-agent DQN) is conducted to evaluate the learned models.
Ideally, if an agent learns the perfect Nash equilibrium strategy, then by
definition, we shall expect the agent to be perfectly non-exploitable (i.e.,
with even the strongest exploiter only capable of achieving her cumulative
utility at most zero in symmetric games). To carry out the experiment, we first trained
all the methods for $5\times 10^{4}$ episodes. After the methods are fully
trained, we take their final models or certain distributions of historical
strategies (uniform for FSP and meta-Nash for PSRO), and the train separate exploiters
playing against those learned strategies. We instantiate a DQN agent as
exploiter using the same set of hyperparameters and network architectures to
learn from scratch against the fixed trained checkpoint. The resulting learning curve
in the exploitation test illustrates the degree of exploitation. An exploiter
reward greater than zero indicates that the agent has been exploited since the
games are symmetric. The model with lower exploiter reward means is more difficult
to be exploited (is thus better).

\begin{figure}[htp]
    \centering
    \includegraphics[width=\columnwidth]{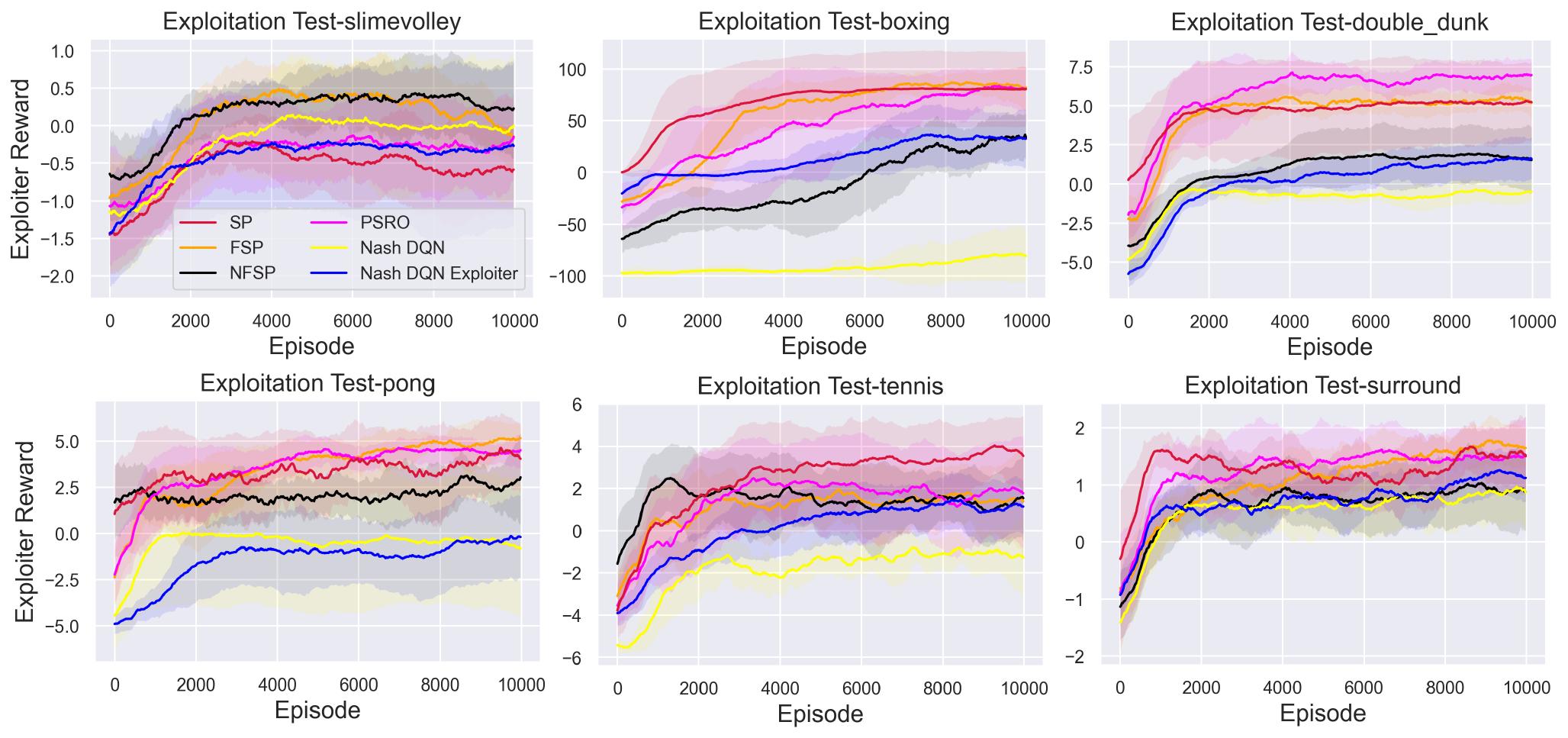}
    \caption{The exploiter learning curves for exploitation tests on six two-player
    zero-sum video games.}
    \label{fig:atari_nvi}
\end{figure}

\begin{table*}
    [htbp]
    \centering
    \caption{\reb{Approximate exploitability} (lower is better) for six two-player
    video games.}
    \resizebox{\textwidth}{!}{ 
    \begin{tabular}{>{\centering\arraybackslash}m{60pt}|>{\centering\arraybackslash}m{30pt}|>{\centering\arraybackslash}m{30pt}|>{\centering\arraybackslash}m{30pt}|>{\centering\arraybackslash}m{30pt}|>{\centering\arraybackslash}m{47pt}|>{\centering\arraybackslash}m{89pt}}
        \toprule \multicolumn{1}{c|}{\backslashbox{Env}{Method}} & \multicolumn{1}{c|}{SP} & \multicolumn{1}{c|}{FSP} & \multicolumn{1}{c|}{NFSP} & \multicolumn{1}{c|}{PSRO} & \multicolumn{1}{c|}{\textbf{Nash DQN}} & \multicolumn{1}{c}{\textbf{Nash DQN Exploiter}} \\
        \hline
        SlimeVolley                                              & $-0.099$                & $0.316$                  & $0.643$                   & $-0.762$                  & $-0.396$                               & $\color{blue}\mathbf{-0.821}$                   \\
        \hline
        Boxing                                                   & $99.475$                & $93.683$                 & $24.544$                  & $66.891$                  & $\color{blue}\mathbf{-90.254}$         & $28.271$                                        \\
        \hline
        Double Dunk                                              & $3.801$                 & $5.920$                  & $0.445$                   & $6.960$                   & $\color{blue}\mathbf{-0.539}$          & $1.367$                                         \\
        \hline
        Pong                                                     & $4.207$                 & $5.196$                  & $4.238$                   & $4.693$                   & $\color{blue}\mathbf{-3.336}$          & $-1.524$                                        \\
        \hline
        Tennis                                                   & $4.176$                 & $2.355$                  & $2.643$                   & $2.943$                   & $\color{blue}\mathbf{-0.425}$          & $0.069$                                         \\
        \hline
        Surround                                                 & $1.782$                 & $1.574$                  & $1.594$                   & $1.603$                   & $\color{blue}\mathbf{0.904}$           & $1.462$                                         \\
        \bottomrule
    \end{tabular}
    } \vskip -.1in \label{tab:atari_exploit}
\end{table*}

Fig.~\ref{fig:atari_nvi} and Table~\ref{tab:atari_exploit} show the exploitation
results of all algorithms and baselines. Each method is trained for five random seeds,
and the model for each random seed is further exploited with DQN exploiter for
$10^{4}$ episodes under three random initializations. For each method and environment,
Table~\ref{tab:atari_exploit} displays the best performing models with its
corresponding exploitability. Complete results are provided in Sec. Sec.~\ref{app:sec_results_video}.
The values in the Table~\ref{tab:atari_exploit} is the maximum of smoothed exploiter
reward in the exploitation test of Fig.~\ref{fig:atari_nvi}. The baseline methods
SP, FSP, NFSP, PSRO do not perform well in most games. This shows the challenge for
finding approximate Nash equilibrium strategies for these games. \nd
demonstrates significant advantages over other methods across all six games. Except
for \textit{Surround} environment, \nd achieves non-positive exploiter rewards for
five environments, which demonstrates the non-exploitability of the policies
learned by \nd. \nde also shows unexploitable performance on \textit{SlimeVolley}
and \textit{Pong} environments. Different environments show different levels of
difficulties to find a non-exploitable model. \textit{SlimeVolley} is relatively
easy with almost all methods achieving exploitability close to zero. \textit{Surround}
is generally hard to resolve due to the inherent complexity of the game. We believe
that solving Surround requires more advanced exploration technique to boost its performance.

\begin{figure}
    \begin{center}
        \includegraphics[width=0.9\columnwidth]{
            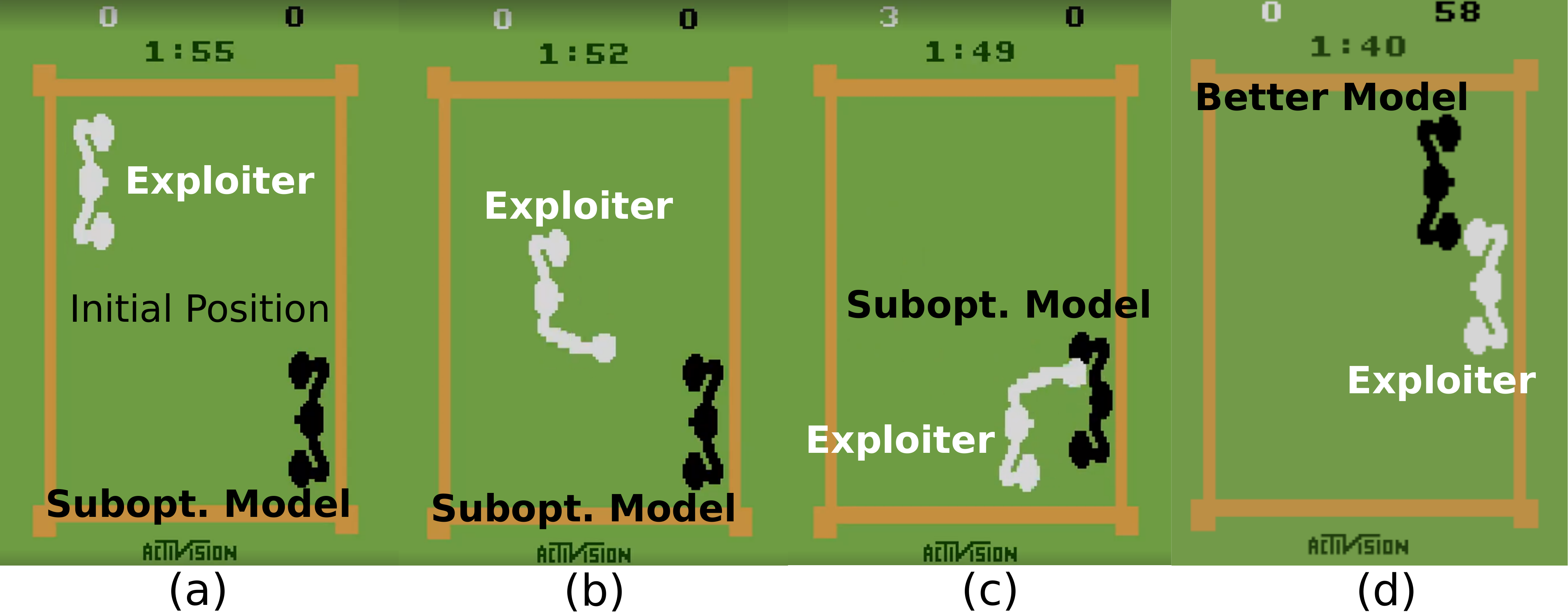
        }
        \caption{The key frames in \textit{Boxing} exploitation test: (a-c)
        shows a sub-optimal model exploited by the exploiter. (d) shows our proposed
        algorithms learn hard-to-exploit policy robust against the turning-around
        strategy of the exploiter.}
        \label{fig:exploit_box}
    \end{center}
\end{figure}

Interestingly, we observe that in the exploitation test for \textit{Boxing}
environment, baseline methods such as SP sometimes produce a policy that keeps staying
at the corner of the ground. As shown in Fig.~\ref{fig:exploit_box}, the black
agent uses the learned suboptimal model by SP algorithm, which tries to avoid any
touch with the white opponent (a-b). Such policy (always hide in a corner) is
not bad when playing against average player or AI whose policies may not have
considered this extreme cases and thus unable to even locate the black agent. However,
this policy is very vulnerable to exploitation. Once the exploiter explores the way
to touch the black agent, (c) our exploiter learns to heavily exploit such
policy in a short time. On the contrary, our algorithm \nd and \nde will never
learn such easy-to-exploit policies. The models learned with \nd and \nde are usually
aggressively approaching the exploiter and directly fighting against it, which
is found to be harder to exploit in this game. Moreover, our policies are
further robust to a turn-around strategy by the exploiter, as shown in Fig.~\ref{fig:exploit_box}
(d).

\begin{figure}[htbp]
    \centering
    \includegraphics[width=\columnwidth]{
        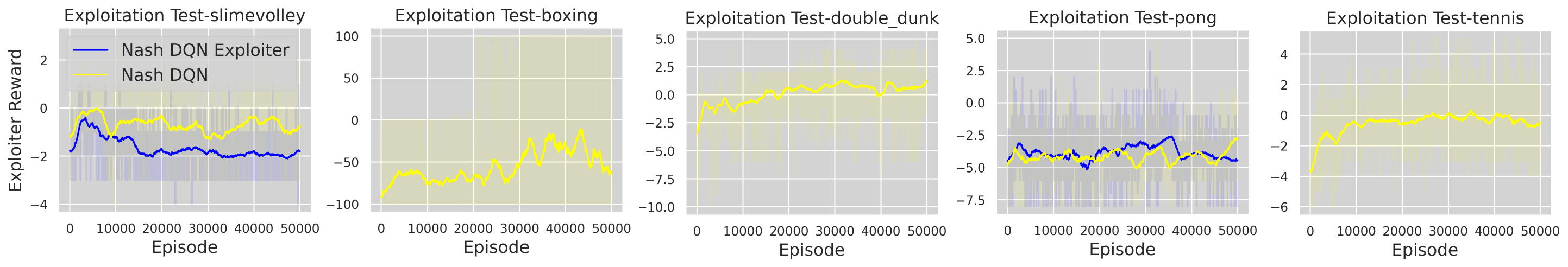
    }
    \caption{The exploiter learning curves for longer exploitation tests on five
    video games.}
    \label{fig:exploit_longer} \vskip -.15in
\end{figure}

To address the possibility insufficient exploitation on our models, we exploit the
models for longer time ($5\times 10^{4}$ episodes) for those methods in Table~\ref{tab:atari_exploit} , and the
results are shown in Fig.~\ref{fig:exploit_longer}. Except for the \textit{Double
Dunk} environment, the \nd and \nde models are still hard to be exploited on
four environments even for long enough exploitation. The difficulty of \textit{Double
Dunk} is that each agent needs to control a team of two players to compete
through team collaboration, which might require a longer training time to further
improve the learned policies.

\section{Comparison of Nash Solvers for Normal-Form Game}
\label{sec:choose_solver} In this section we show the Nash equilibrium solvers (\emph{i.e.},
Nash solving subroutine $\textit{Nash}$) for zero-sum normal-form games, which
is an important subroutine for the proposed algorithms. Since NE for zero-sum normal-form
games can be solved by linear programming, some packages involving linear
programming functions like ECOS, PuLP and Scipy can be leveraged. We also
implement a solver based on an iterative algorithm--multiplicative weights update
(MWU), which is detailed in Sec. Sec.~\ref{sec:mwu}.

\subsection{Multiplicative Weights Update}
\label{sec:mwu} The MWU algorithm~\cite{bailey2018multiplicative, daskalakis2018last}
is no-regret in online learning setting, which can be used for solving NE in two-player
zero-sum normal-form games. Given a payoff matrix $A$ (from the max-player's
perspective), the NE strategies will be solved by iteratively applying MWU. Specifically,
for $n$-th iteration, state $s$ and actions $(a_{i}, b_{j})$ for the max-player $\mu$
and min-player $\nu$ respectively, $i, j$ are the entry indices of discrete
action space, then the update rule of the action probabilities for two players are:
\begin{align}
    \mu^{(n+1)}(a_{i}|s) = \mu^{(n)}(a_{i}|s)\frac{e^{\eta (A \boldsymbol{\nu}^{(n)\intercal}(s))_i}}{\sum_{i^\prime}\mu^{(n)}(a_{i^\prime}; s) e^{\eta (A \boldsymbol{\nu}^{(n)\intercal}(s))_{i^\prime}}}   \\
    \nu^{(n+1)}(b_{j}|s) = \nu^{(n)}(b_{j}|s)\frac{e^{-\eta (A \boldsymbol{\mu}^{(n)\intercal}(s))_j}}{\sum_{j^\prime}\nu^{(n)}(b_{j^\prime}; s) e^{-\eta (A \boldsymbol{\mu}^{(n)\intercal}(s))_{j^\prime}}}
\end{align}
where $\eta$ is the learning rate. By iteratively updating each action entry of the
strategies with respect to the payoff matrix, MWU is provably converging to NE.

\subsection{Comparison}
\begin{table}[htbp]
    \caption{Comparison of different Nash solvers for zero-sum matrix game ($6\times
    6$ random matrices).}
    \label{tab:compare_ne_solver}
    \begin{threeparttable}
        \begin{tabular}{cccc}
            \toprule    & \multirow{1}{*}{Solver}      & \multicolumn{1}{c}{Time per Sample (s)} & \multicolumn{1}{c}{Solvability}  \\
            \cline{2-4} & Nashpy (equilibria)\tnote{1} & $0.751$                                 & all                              \\
            \cline{2-4} & Nashpy (equilibrium)         & $0.0016$                                & not for some degenerate matrices \\
            \cline{2-4} & ECOS\tnote{2}                & $0.0015$                                & all                              \\
            \cline{2-4} & MWU (single)                 & $0.008$                                 & all (but less accurate)          \\
            \cline{2-4} & MWU (parallel)               & fast, depends on batch size\tnote{3}    & all (but less accurate)          \\
            \cline{2-4} & CVXPY                        & $0.009$                                 & all                              \\
            \cline{2-4} & PuLP                         & $0.020$                                 & all                              \\
            \cline{2-4} & Scipy                        & $-$                                     & not for some                     \\
            \cline{2-4} & Gurobipy                     & $~0.01$                                 & not for some                     \\
            \midrule
        \end{tabular}
        \begin{tablenotes}
            \footnotesize \item[1] Nashpy (\href{https://github.com/drvinceknight/Nashpy}{https://github.com/drvinceknight/Nashpy})
            can be adopted to achieve two versions of NE solvers: one returns a
            single NE, and another returns all Nash equilibria for given payoff matrices.
            \item[2] ECOS (\href{https://github.com/embotech/ecos}{https://github.com/embotech/ecos})
            is a package for solving convex second-order cone programs. \item[3]
            MWU solver is self-implemented for solving either a single payoff matrix
            or solving a batch of matrices in parallel. \item[4] CVXPY (\href{https://github.com/cvxpy/cvxpy}{https://github.com/cvxpy/cvxpy})
            is a Python package for convex optimization. \item[5] PuLP (\href{https://github.com/coin-or/pulp}{https://github.com/coin-or/pulp})
            is a linear programming package with Python. \item[6] Scipy (\textit{scipy.optimize.linprog()})
            is a general package for numerical operations. \item[7] Gurobipy (\href{https://www.gurobi.com/}{https://www.gurobi.com/})
            is a package for linear and quadratic optimization.
        \end{tablenotes}
    \end{threeparttable}
\end{table}

We conduct experiments to compare different solvers for NE subroutine, including
Nashpy (for single Nash \reb{equilibrium} or all Nash equilibria), ECOS, MWU (single
or parallel), CVXPY, PuLP, Scipy, Gurobipy. The test is conducted on a Dell XPS
15 laptop with only CPU computation. The code will be released after the review process
(anonymous during review process). Experiments are evaluated on $6\times 6$
random matrices and averaged over 1000 samples. The zero-sum property of the generated
Markov games is guaranteed by generating each random matrix as one player's payoff
and take the negative values as its opponent's payoff.

As in Table.~\ref{tab:compare_ne_solver}, the solvability indicates whether the solver
can solve all possible randomly generated matrices (zero-sum). Nashpy for
solving all equilibria cannot handle some degenerate matrices. Scipy and Gurobipy
also cannot solve for some payoff matrices. Other solvers can solve all random
payoff matrices in our tests but with different solving speed and accuracy. The solvability
is essential for the program since the values within the payoff matrix can be arbitrary
as a result of applying function approximation. The results support our choice
of using the ECOS-based solver as the default Nash solving subroutine for the
proposed algorithms, due to its speed and robustness for solving random matrices.
ECOS is originally built for solving convex second-order cone programs, which covers
linear programming (LP) problem. It tries to transform the input matrices to be
Scipy sparse matrices and speeds up the solving procedure. By formulating the NE
solving as LP on normal-form game, we can plug in the ECOS solver to get the
solution. Some constraints like positiveness and constant sum need to be handled
carefully. Other solvers like Nashpy (equilibrium) and MWU (parallel) can
achieve a similar level of computational time, but less preferred due to either not
being able to solve some matrices or less accurate results. Specifically, for the
case with a large batch size and a small number of inner-loop iterations for MWU,
MWU can be faster than ECOS. However, the accuracy of MWU depends on the number
of iterations for update~\cite{bailey2018multiplicative, daskalakis2018last}. More
iterations lead to more accurate approximation but also longer computational time
for MWU method. Empirically, we find that the accuracy of the subroutine solver is
critical for our proposed algorithms with function approximation, especially in
video games with long horizons. \reb{Moreover, although we have already selected the best solver through comparisons, the computational time of the solvers used in each inference or update step still account for a considerable portion. This leaves some space for improvement of running-time efficiency. }


\section{Algorithms on Tabular Markov Games}
\label{sec:alg_tab_markov} In this section, we provide further details on connections
of our algorithms to tabular algorithms \nv and \nve in Sec.\ref{sec:connections},
and prove the theoretical guarantees of the latter two algorithms in Sec.\ref{sec:proof_of_theory}.
We then provide all the pseudo-codes for subroutines and algorithms used in this
paper. In particular, we introduce the pseudo-codes for important subroutines in
Sec.~\ref{app:sec_sub}, and then the pseudo-codes for several algorithms: self-play
(SP, \ref{app:sec_sp}), fictitious self-play (FSP, \ref{app:sec_fsp}), double
oracle (DO, \ref{app:sec_do}), Nash value iteration (\nv, \ref{app:sec_nvi}) and
Nash value iteration with exploiter (\nve, \ref{app:sec_nvie}). SP, FSP and DO
are the baseline methods in experimental comparisons, while \nv and \nve are the
tabular version of our proposed algorithms \nd and \nde, without function
approximation.


\subsection{Nash Algorithm Subroutines}
\label{app:sec_sub}

\begin{algorithm}
    [htbp]
    \caption{{$\textit{Meta-Nash}$}: Meta-Nash Equilibrium Solving Subroutine}
    \begin{algorithmic}
        [1] \INPUT two strategy sets $\mu,\nu$; evaluation iterations $N$ \STATE
        Initialize payoff matrix: $M_{i,j}=0, i\in[|\mu|], j\in[|\nu|]$ \FOR{$\mu_{i}\in \mu$}
        \FOR{$\nu_{i}\in \nu$} \FOR{episodes $k=1,\ldots,N$} \STATE Rollout
        policies $\mu_{i}, \nu_{j}$ to get episodic reward $r_{k}$ \STATE
        $M_{i,j}=\frac{1}{N}\sum_{k=1}^{N} r_{k}$ \ENDFOR \ENDFOR \ENDFOR \STATE
        $(\rho_{\mu}, \rho_{\nu})={\textit{Nash}}(M)$ \STATE Return $\rho_{\mu}$
        or $\rho_{\nu}$
    \end{algorithmic}
    \label{alg:meta_nash}
\end{algorithm}

Before introducing the pseudo-code for each algorithm, we summarize several subroutines
-- { $\textit{Nash}$}, { $\textit{Meta-Nash}$}, { $\textit{Best-Response}$}
and { $\textit{Best-Respons--Value}$} -- applied in the algorithms. These
subroutines are marked in {magenta} color in the this and the following sections.

{ $\textit{Nash}$}: As a NE solving subroutine for normal-form games, it returns
the NE strategy given the payoff matrix as the input. Specifically it uses the solvers
introduced in Sec. Sec.~\ref{sec:choose_solver}, and ECOS is the default choice
in our experiments.

{ $\textit{Meta-Nash}$}: As a meta-Nash solving subroutine (Algorithm~\ref{alg:meta_nash}),
it returns the one-side meta NE strategy given two strategy \textbf{sets}: $\mu=\{
\mu_{1}, \cdots, \mu_{i}, \cdots\},\nu=\{\nu_{1}, \cdots, \nu_{i}, \cdots\}$. A
one-by-one matching for each pair of polices
$(\mu_{i}, \nu_{j}), i\in [|\mu|], j\in[|\nu|]$ is evaluated in the game to get
an estimated payoff matrix, with the average episodic return as the estimated payoff
values of two players for each entry in the payoff matrix. The
{ $\textit{Nash}$} subroutine is called to solve the meta-Nash strategies. It is
applied in DO algorithm, which is detailed in Sec.~\ref{app:sec_do}.

\begin{algorithm}
    [htbp]
    \caption{{$\textit{Best-Response}$ I}: Best Response Subroutine in Markov Game
    (known transition, reward functions)}
    \label{alg:br1}
    \begin{algorithmic}
        [1] \INPUT mixture policy $\rho_{\mu}$ as a distribution over $\{\mu^{0},
        \mu^{1},\dots, \mu^{n}\}$ \STATE Initialize non-Markovian policies $\hat{\mu}
        =\{\hat{\mu}_{h}\}, \hat{\nu}=\{\hat{\nu}_{h}\}, h\in[H], \mu_{h}:(\mathcal{S}
        \times\mathcal{A\times\mathcal{B}})^{(h-1)}\times\mathcal{S}\times\mathcal{A}
        \rightarrow [0,1], \nu_{h}:(\mathcal{S}\times\mathcal{A\times\mathcal{B}}
        )^{(h-1)}\times\mathcal{S}\times\mathcal{B}\rightarrow [0,1]$ \STATE Initialize
        $Q$ table for non-Markovian policies $\hat{\mu}, \hat{\nu}$,
        $Q=\{Q_{h}\}, h\in[H], Q_{h}:(\mathcal{S}\times\mathcal{A}\times\mathcal{B}
        )^{h}\rightarrow [0,1]$
        \STATE Initialize $V$ table for non-Markovian policies
        $\hat{\mu}, \hat{\nu}$, $V=\{V_{h}\}, h\in[H], V_{h}:(\mathcal{S}\times\mathcal{A}
        \times\mathcal{B})^{(h-1)}\times\mathcal{S}\rightarrow [0,1]$ \FOR{$h=1,\ldots,H$}
        \STATE For all $\tau_{h-1}$:
        \begin{align}
            Q^{\mu, \dagger}_{h}(\tau_{h-1}, s_{h},a_{h},b_{h}) & =\sum_{s^\prime\in\mathcal{S}}\mathbb{P}_{h}(s_{h+1}|s_{h},a_{h},b_{h})[r_{h}(s_{h}, a_{h}, b_{h})+V^{\hat{\mu}, \dagger}_{h+1}(\tau_{h}, s_{h+1})]              \\
            V^{\hat{\mu}, \dagger}_{h}(\tau_{h-1}, s_{h})       & = \min_{\nu_h}\hat{\mu}_{h}(\cdot|\tau_{h-1}, s_{h}) Q_{h}^{\hat{\mu}, \dagger}(\tau_{h-1}, s_{h}, \cdot, \cdot)\nu_{h}^{\intercal}(\cdot|\tau_{h-1}, s_{h})     \\
            \hat{\nu}_{h}(\tau_{h-1}, s_{h})                    & = \arg\min_{\nu_h}\hat{\mu}_{h}(\cdot|\tau_{h-1}, s_{h}) Q_{h}^{\hat{\mu}, \dagger}(\tau_{h-1}, s_{h}, \cdot, \cdot)\nu_{h}^{\intercal}(\cdot|\tau_{h-1}, s_{h})
        \end{align}
        where
        \begin{equation*}
            \hat{\mu}_{h}(a_{h}|\tau_{h-1}, s_{h}):=\frac{\sum_{i} \mu_{h}^{i}(a_{h}|
            s_{h})\rho(i)\Pi_{t^\prime=1}^{h-1}\mu^{i}_{t^\prime}(a_{t^\prime}| s_{t^\prime})}{\sum_{j}
            \rho(j)\Pi_{t^\prime=1}^{h-1}\mu^{j}_{t^\prime}(a_{t^\prime}| s_{t^\prime})}
        \end{equation*}
        \ENDFOR \STATE Return $\hat{\nu}$ or $V_{1}^{\hat{\mu}, \dagger}(s_{1})$
        \\
        {\color{blue}\% $\hat{\mu}$ is the posterior policy of non-Markovian mixture $\mu$, $\hat{\nu}$ is the best response of it}
    \end{algorithmic}
\end{algorithm}

\begin{algorithm}
    [htbp]
    \caption{{$\textit{Best-Response}$ II}: Best Response Subroutine in Markov Game
    ($Q$-learning based, unknown transition, reward functions)}
    \label{alg:br2}
    \begin{algorithmic}
        [1] \INPUT mixture policy $\rho_{\mu}$ as a distribution over $\{\mu^{0},
        \mu^{1},\dots, \mu^{n}\}$; best response $Q$-learning iterations $N$; soft
        update coefficient $\alpha$ \STATE Initialize the $Q=\{Q_{h}|h\in[H]\}\in
        \mathbb{R}^{|\mathcal{S}| \times |\mathcal{B}|}$ table for the best response
        player, \FOR{episodes $k=1,\ldots,N$} \STATE Sample policy $\mu_{k}\sim\rho
        _{\mu}$ \FOR{$t = 1,\ldots,H$} \STATE {\color{blue}\% collect data}
        \STATE Sample greedy action $a_{t}\sim \mu_{k}(\cdot|s_{t})$ \STATE With
        $\epsilon$ probability, sample random action $b_{t}$; \STATE Otherwise, sample
        greedy action $b_{t}\sim\nu(\cdot|s_{t})$ according to $Q$ \STATE Rollout
        environment to get sample $(s_{t},a_{t},b_{t},r_{t},\text{done},s_{t+1})$
        \quad($r_{t}$ is for the learning player) \STATE
        {\color{blue}\% update best response Q-value} \IF{not done} \STATE $Q^{\text{target}}
        _{t}(s_{t}, b_{t})=r_{t}+ V_{t+1}(s_{t+1})$ \STATE where $V_{t+1}(s_{t+1}
        )=\max_{b'}Q_{t+1}(s_{t+1}, b')$ \ELSE \STATE $Q_{t}^{\text{target}}(s_{t}
        , b_{t})=r_{t}$ \ENDIF \STATE $Q_{t}(s_{t}, b_{t})\leftarrow \alpha\cdot
        Q_{t}^{\text{target}}(s_{t}, b_{t})+(1-\alpha)\cdot Q_{t}(s_{t}, b_{t})$
        \IF{done} \STATE break \ENDIF \ENDFOR \ENDFOR \STATE Represent $Q$ as a
        greedy policy $\hat{\nu}$ \STATE Return $\hat{\nu}$
    \end{algorithmic}
\end{algorithm}

{ $\textit{Best-Response}$}: As a best response subroutine, it returns the best
response strategy of the given strategy, which satisfies Eq.~\eqref{eq:best_response_v}.
To be noticed, the best response we discuss here is the best response of a meta-distribution
$\rho_{\mu}$ over a strategy set $\{\mu^{0}, \mu^{1},\dots, \mu^{n}\}$, which covers
the case of best response against a single strategy by just making the distribution
one-hot. We use this setting for the convenience of being applied in SP, FSP, DO
algorithms. Here we discuss two types of best response subroutine that are used at
different positions in the algorithms: (1) {$\textit{Best-Response}$ I} (as Algorithm~\ref{alg:br1})
is a best response subroutine with oracle transition and reward function of the
game, which is used for evaluating the exploitability of the model after training;
(2) {$\textit{Best-Response}$ II} (as Algorithm~\ref{alg:br2}) is a best
response subroutine with $Q$-learning agent for approximating the best response,
without knowing the true transition and reward function of the game. It is used
in the procedure of methods based on iterative best response, like SP, FSP, DO.
We claim here for the following sections, by default,
{ $\textit{Best-Response}$} will use {$\textit{Best-Response}$ II}, and { $\textit{Best-Response-Value}$}
will use {$\textit{Best-Response}$ I}.

{ $\textit{Best-Respons---Value}$}: It has the same procedure as
${\textit{Best-Response}}$ as a best response subroutine, but returns the average
value of the initial states as $V_{1}^{\hat{\mu}, \dagger}(s_{1})$ in Eq.~\eqref{eq:subopt}
with the given strategy $\hat{\mu}$. Since the best response value estimation is
used in evaluating the exploitability of a certain strategy, it by default adopts
{$\textit{Best-Response}$ I} (Algorithm~\ref{alg:br1}) as an oracle process,
which returns the ground-truth best response values because of knowing the
transition and reward functions.

\subsection{\reb{Nash Q-Learning}}
\label{app:sec_nash_q_l} \reb{ The pseudo-code for Nash Q-Learning and Nash Q-Learning with Exploiter are shown in Algorithm.\ref{alg:nash_q_l} and \ref{alg:nash_q_l_exploiter} below. 

\begin{algorithm}[htbp] \caption{Nash Q-Learning} \label{alg:nash_q_l} \begin{algorithmic}[1] \STATE Initialize $Q:\mathcal{S}\times\mathcal{A}\times\mathcal{B}\rightarrow \mathbb{R}$, given $\epsilon, \gamma, \alpha$. \FOR{$k = 1,\ldots,K$} \FOR{$t = 1,\ldots,H$} \STATE {\color{blue}\% collect data} \STATE With $\epsilon$ probability, sample random actions $a_{t}, b_{t}$; \STATE Otherwise, $a_{t}\sim\mu(\cdot|s_{t}),b_{t}\sim\nu(\cdot|s_{t}), (\mu(\cdot|s_{t}),\nu(\cdot|s_{t}) )={\textit{Nash}}(Q(s_{t}, \cdot, \cdot))$ \STATE Rollout environment to get sample $(s_{t},a_{t},b_{t},r_{t},\text{done},s_{t+1})$ \STATE {\color{blue}\% update Q-value} \IF{not done} \STATE Compute $(\hat{\mu}, \hat{\nu})={\textit{Nash}}(Q(s_{t+1}, \cdot, \cdot))$ \STATE Set $Q^{\text{target}}(s_{t}, a_{t}, b_{t}) = r_{t} + \gamma \hat{\mu}\trans Q(s_{t+1}, \cdot, \cdot)\hat{\nu}$. \ELSE \STATE Set $Q^{\text{target}}(s_{t}, a_{t}, b_{t})=r_{t}$ \ENDIF \STATE $Q(s_{t}, a_{t}, b_{t})\leftarrow \alpha\cdot Q^{\text{target}}(s_{t}, a_{t}, b_{t})+(1-\alpha)\cdot Q(s_{t}, a_{t}, b_{t})$ \IF{done} \STATE break \ENDIF \ENDFOR \ENDFOR\end{algorithmic}\end{algorithm} }

\begin{algorithm}[htbp]
\caption{Nash Q-learning with Exploiter} \label{alg:nash_q_l_exploiter}
\begin{algorithmic}
	\STATE Initialize $Q, \tilde{Q}:\mathcal{S}\times\mathcal{A}\times\mathcal{B}\rightarrow \mathbb{R}$, given $\epsilon, \gamma, \alpha$.
\FOR{$k = 1,\ldots,K$}
\FOR{$t = 1,\ldots,H$}
\STATE With $\epsilon$ probability, sample random actions $a_t, b_t$;
\STATE Otherwise, $a_t\sim\mu(\cdot|s_t),b_t\sim\tilde{\nu}(\cdot|s_t)$,
\STATE where $(\mu(\cdot|s_t),\nu(\cdot|s_t) )=\textit{Nash}(Q(s_{t}, \cdot, \cdot)), \tilde{\nu}(\cdot|s_t) = \text{One-Hot}(\argmin_b \E_{a\sim \mu_t} \tilde{Q}(s_t, a, b))$.
\STATE Rollout environment to get sample $(s_t,a_t,b_t,r_t,\text{done},s_{t+1})$
\STATE {\color{blue}{Nash Q-value update}}
\IF{not done} \STATE $Q^\text{target}(s_t, a_t, b_t)=r_t+\gamma V^\text{Nash}(s_{t+1})$,
\STATE where $V^\text{Nash}(s_{t+1})=\textit{Nash}(Q(s_{t+1}, \cdot, \cdot))$.
\STATE $\tilde{Q}^\text{target}(s_t, a_t, b_t)=r_t+\gamma \min_{b'\in\mathcal{B}} \mu(s_{t+1})^\intercal \tilde{Q}(s_{t+1}, \cdot, b')$,
\STATE where $(\mu, \nu)=\textit{Nash}(Q(s_{t+1}, \cdot, \cdot))$
\ELSE \STATE $Q^\text{target}(s_t, a_t, b_t)=r_t$
\STATE $\tilde{Q}^\text{target}(s_t, a_t, b_t)=r_t$
\ENDIF
\STATE $Q(s_t, a_t, b_t)\leftarrow \alpha\cdot Q^\text{target}(s_t, a_t, b_t)+(1-\alpha)\cdot Q(s_t, a_t, b_t)$
\STATE $\tilde{Q}(s_t, a_t, b_t)\leftarrow \alpha\cdot \tilde{Q}^\text{target}(s_t, a_t, b_t)+(1-\alpha)\cdot \tilde{Q}(s_t, a_t, b_t)$
\IF{done} \STATE break
\ENDIF
\ENDFOR
\ENDFOR
\STATE {\# Note: $\textit{Nash}$ is a NE solving subroutine}
\end{algorithmic}
\end{algorithm}

\subsection{Self-play}
\label{app:sec_sp} The pseudo-code for self-play is shown in Algorithm \ref{alg:sp}.

\begin{algorithm}
    [htbp]
    \caption{Self-play for Markov Game}
    \begin{algorithmic}
        [1] \STATE Initialize policies $\mu^{0}=\{\mu_{h}\}, \nu^{0}=\{\nu_{h}\},
        h\in[H]$ \STATE Initialize policy sets:
        $\mu=\{\mu^{0}\}, \nu=\{\nu^{0}\}$ \STATE Initialize meta-strategies: $\rho
        _{\mu}=[1.], \rho_{\nu}=[1.]$ \FOR{$t=1,\ldots,T$} \IF{$t\%2==0$} \STATE
        $\nu^{t} ={\textit{Best-Response}}(\rho_{\mu}, \mu)$ \STATE $\nu=\nu\bigcup
        \{\nu^{t}\}$ \STATE $\rho_{\nu}=(0,\dots, 1)$ as a one-hot vector with only
        1 for the last entry \ELSE \STATE $\mu^{t} ={\textit{Best-Response}}(\rho
        _{\nu}, \nu)$ \STATE $\mu=\mu\bigcup\{\mu^{t}\}$ \STATE $\rho_{\mu}=(0,\dots
        , 1)$ as a one-hot vector with only 1 for the last entry \ENDIF \STATE exploitability
        = ${\textit{Best-Response-Value}}(\rho_{\mu},\mu)+{\textit{Best-Response-Value}}
        (\rho_{\nu},\nu)$ \ENDFOR \STATE Return $\mu, \nu$
    \end{algorithmic}
    \label{alg:sp}
\end{algorithm}

\subsection{Fictitious Self-play}
\label{app:sec_fsp} The pseudo-code for fictitious self-play is shown in
Algorithm.\ref{alg:fsp}. We use uniform($\cdot$) to denote a uniform
distribution over the policy set.

\begin{algorithm}
    [htbp]
    \caption{Fictitious Self-play for Markov Game}
    \label{alg:fsp}
    \begin{algorithmic}
        [1] \STATE Initialize policies $\mu^{0}=\{\mu_{h}\}, \nu^{0}=\{\nu_{h}\},
        h\in[H]$ \STATE Initialize policy sets:
        $\mu=\{\mu^{0}\}, \nu=\{\nu^{0}\}$ \STATE Initialize meta-strategies: $\rho
        _{\mu}=[1.], \rho_{\nu}=[1.]$

        \FOR{$t=1,\ldots,T$} \IF{$t\%2==0$} \STATE
        $\nu^{t} ={\textit{Best-Response}}(\rho_{\mu}, \mu)$ \STATE
        $\nu=\nu\bigcup\{\nu^{t}\}$ \STATE $\rho_{\nu}=\text{Uniform}(\nu)$
        \ELSE \STATE $\mu^{t} ={\textit{Best-Response}}(\rho_{\nu}, \nu)$
        \STATE $\mu=\mu\bigcup\{\mu^{t}\}$ \STATE
        $\rho_{\mu}=\text{Uniform}(\mu)$ \ENDIF \STATE exploitability =
        ${\textit{Best-Response-Value}}(\rho_{\mu},\mu)+{\textit{Best-Response-Value}}
        (\rho_{\nu},\nu)$
        \ENDFOR \STATE Return $\mu, \rho_{\mu}, \nu, \rho_{\nu}$
    \end{algorithmic}
\end{algorithm}

\begin{algorithm}
    [htbp]
    \caption{Double Oracle for Markov Game}
    \label{alg:sec_do}
    \begin{algorithmic}
        [1] \STATE Initialize policies $\mu^{0}=\{\mu_{h}\}, \nu^{0}=\{\nu_{h}\},
        h\in[H]$ \STATE Initialize policy sets:
        $\mu=\{\mu^{0}\}, \nu=\{\nu^{0}\}$ \STATE Initialize meta-strategies: $\rho
        _{\mu}=[1.], \rho_{\nu}=[1.]$

        \FOR{$t=1,\ldots,T$} \IF{$t\%2==0$} \STATE
        $\nu^{t}={\textit{Best-Response}}(\rho_{\mu}, \mu)$ \STATE
        $\nu=\nu\bigcup\{\nu^{t}\}$ \STATE
        $\rho_{\nu}={\textit{Meta-Nash}}(\nu, \mu)$ \ELSE \STATE
        $\mu^{t}={\textit{Best-Response}}(\rho_{\nu}, \nu)$ \STATE
        $\mu=\mu\bigcup\{\mu^{t}\}$ \STATE
        $\rho_{\mu}={\textit{Meta-Nash}}(\nu, \mu)$ \ENDIF \STATE
        exploitability =
        ${\textit{Best-Response-Value}}(\rho_{\mu}, \mu)+{\textit{Best-Response-Value}}
        (\rho_{\nu},\nu)$
        \ENDFOR \STATE Return $\mu, \rho_{\mu}, \nu, \rho_{\nu}$
    \end{algorithmic}
\end{algorithm}

\subsection{Double Oracle}
\label{app:sec_do} The pseudo-code for double oracle is shown in Algorithm.\ref{alg:sec_do}.

\subsection{\nv}
\label{app:sec_nvi} The pseudo-code for Nash value iteration (\nv) is shown in
Algorithm~\ref{alg:sec_nvi}. Different from \nd (as Algorithm~\ref{alg:nash_dqn}),
for tabular Markov games, the $Q$ network is changed to be the $Q$ table and updated
in a tabular manner (as Algorithm~\ref{alg:sec_nvi} line~\ref{line:update_nash_vi}),
given the estimated transition function $\tilde{\mathbb{P}}$ and reward function
$\tilde{r}$. The target $Q$ is not used.
Since \nv is applied for tabular Markov games, here we write the pseudo-code in an
episodic setting without the reward discount factor, which is slightly different
from Sec.~\ref{subsec:nash_dqn}.

\begin{algorithm}
    [htbp]
    \caption{Nash Value Iteration \label{alg:sec_nvi} (\nv, $\epsilon$-greedy
    sample version)}
    \begin{algorithmic}
        [1] \STATE Initialize $Q=\{Q_{h}\},h\in[H],Q_{h}:\mathcal{S}_{h}\times\mathcal{A}
        _{h}\times\mathcal{B}_{h}\rightarrow \mathbb{R}$, buffer $\cD=\phi$, given
        $\epsilon$, update interval $p$. \FOR{$k = 1,\ldots,K$} \FOR{$t = 1,\ldots,H$}
        \STATE {\color{blue}\% collect data} \STATE With $\epsilon$ probability,
        sample random actions $a_{t}, b_{t}$; \STATE Otherwise,
        $a_{t}\sim\mu_{t}(\cdot|s_{t}),b_{t}\sim\nu_{t}(\cdot|s_{t}), (\mu_{t}(\cdot
        |s_{t}),\nu_{t}(\cdot|s_{t}) )={\textit{Nash}}(Q_{t}(s_{t}, \cdot, \cdot)
        )$. \STATE Rollout environment to get sample $(s_{t},a_{t},b_{t},r_{t},\text{done}
        ,s_{t+1})$ and store in $\cD$.
        \STATE {\color{blue}\% update Q-value} \IF{$|\cD|\%p=0$} \FOR {$\forall (s,a,b, s')\in\mathcal{S}_{h}\times\mathcal{A}_{h}\times\mathcal{B}_{h}\times\mathcal{S}_{h+1}, h\in [H]$}
        \STATE Estimate $\tilde{\mathbb{P}}_{h}(s_{h+1}=s'|s_{h}=s,a_{h}=a,b_{h}=
        b)=\frac{1}{n}\sum_{i=1}^{n} \mathbbm{1}(s_{h+1}=s'_{i}), (s,a,b,s_{i}')\in
        \cD$. \STATE Estimate
        $\tilde{r}_{h}(s_{h}=s,a_{h}=a,b_{h}=b)=\frac{1}{m}\sum_{i=1}^{m} r_{i}(s
        ,a,b), (s,a,b,r_{i})\in \cD$. \STATE \alglinelabel{line:update_nash_vi}
        $Q_{h}(s, a, b)=\tilde{r}_{h}(s,a,b)+ (\tilde{\mathbb{P}}_{h}V_{h+1}^{\hat{\mu}_{h+1},
        \hat{\nu}_{h+1}})(s,a, b)\cdot \mathbb{I}[s' \text{ is non-terminal}]$,

        \STATE where
        $(\hat{\mu}_{h+1}, \hat{\nu}_{h+1})={\textit{Nash}}(Q_{h+1})$. \ENDFOR \ENDIF
        \IF{done} \STATE break \ENDIF \ENDFOR \ENDFOR
    \end{algorithmic}
\end{algorithm}

\subsection{\nve}
\label{app:sec_nvie} The pseudo-code for Nash value iteration with Exploiter (\nve)
is shown in Algorithm.\ref{alg:sec_nvie}. Different from \nde (as Algorithm~\ref{alg:nash_dqn_exp}),
for tabular Markov games, the $Q$ network and exploiter $\tilde{Q}$ network are changed
to be $Q$ tables and updated in a tabular manner (as Algorithm~\ref{alg:sec_nvie}
line~\ref{line:update_nash_vi_exp} and line~\ref{line:update_nash_vi_exp_tilde}),
given the estimated transition function $\tilde{\mathbb{P}}$ and reward function
$\tilde{r}$. The target $Q$ and target $\tilde{Q}$ are not used. Since \nve is
applied for tabular Markov games, here we write the pseudo-code in an episodic setting
without the reward discount factor, which is slightly different from Sec.~\ref{subsec:nash_dqn_exploiter}.

\begin{algorithm}
    [htbp]
    \caption{Nash Value Iteration with Exploiter (\nve, $\epsilon$-greedy sample
    version)}
    \label{alg:sec_nvie}
    \begin{algorithmic}
        [1] \STATE Initialize $Q=\{Q_{h}\}, \tilde{Q}=\{\tilde{Q}_{h}\},h\in[H],\tilde
        {Q}_{h},Q_{h}:\mathcal{S}_{h}\times\mathcal{A}_{h}\times\mathcal{B}_{h}\rightarrow
        \mathbb{R}$, buffer $\cD=\phi$, given $\epsilon$, update interval $p$. \FOR{$k = 1,\ldots,K$}
        \FOR{$t = 1,\ldots,H$} \STATE {\color{blue}\% collect data} \STATE With $\epsilon$
        probability, sample random actions $a_{t}, b_{t}$; \STATE Otherwise,
        $a_{t}\sim\mu_{t}(\cdot|s_{t}),b_{t}\sim\tilde{\nu}_{t}(\cdot|s_{t})$, \STATE
        $(\mu_{t}(\cdot|s_{t}),\nu_{t}(\cdot|s_{t}) )={\textit{Nash}}(Q(s_{t}, \cdot
        , \cdot)), \tilde{\nu}_{t}(\cdot|s_{t}) = \arg\min_{\nu} \mu_{t}^{\intercal}
        (\cdot|s_{t}) \tilde{Q}_{t}(s_{t}, \cdot, \cdot)\nu$. \STATE Rollout
        environment to get sample
        $(s_{t},a_{t},b_{t},r_{t},\text{done},s_{t+1})$ and store in $\cD$.
        \STATE {\color{blue}\% update Q-value} \IF{$|\mathcal{D}|\%p=0$} \FOR
        {$\forall (s,a,b, s')\in\mathcal{S}_{h}\times\mathcal{A}_{h}\times\mathcal{B}_{h}\times\mathcal{S}_{h+1}, h\in [H]$}
        \STATE Estimate
        $\tilde{\mathbb{P}}_{h}(s_{h+1}=s'|s_{h}=s,a_{h}=a,b_{h}=b)=\frac{1}{n}\sum
        _{i=1}^{n} \mathbbm{1}(s_{h+1}=s'_{i}), (s,a,b,s_{i}')\in \cD$. \STATE Estimate
        $\tilde{r}_{h}(s_{h}=s,a_{h}=a,b_{h}=b)=\frac{1}{m}\sum_{i=1}^{m} r_{i}(s
        ,a,b), (s,a,b,r_{i})\in \cD$. \STATE \alglinelabel{line:update_nash_vi_exp}
        $Q_{h}(s, a, b)=\tilde{r}_{h}(s,a,b)+ (\tilde{\mathbb{P}}_{h}V_{h+1}^{\hat{\mu}_{h+1},
        \hat{\nu}_{h+1}})(s,a, b)\cdot \mathbb{I}[s' \text{ is non-terminal}]$,
        \STATE where $(\hat{\mu}_{h+1}, \hat{\nu}_{h+1})={\textit{Nash}}(Q_{h+1})$.
        \STATE \alglinelabel{line:update_nash_vi_exp_tilde} $\tilde{Q}_{h}(s, a,
        b)=\tilde{r}_{h}(s,a,b)+ (\tilde{\mathbb{P}}_{h}V_{h+1}^{\text{Exploit}})
        (s,a, b)$,

        \STATE where $V_{h+1}^{\text{Exploit}}(s')=
        \begin{cases}
            \min_{b'\in\mathcal{B}_{h+1}}{\hat{\mu}}_{h+1}(s')^{\intercal} \tilde{Q}_{h+1}(s', \cdot, b') & \text{for non-terminal $s'$} \\
            0                                                                                             & \text{for terminal $s'$}
        \end{cases}$. \ENDFOR \ENDIF \IF{done} \STATE break \ENDIF \ENDFOR \ENDFOR
    \end{algorithmic}
\end{algorithm}




\subsection{\reb{Comparisons of Nash Series Algorithms}}
\paragraph{Connections of \nd, \nde to tabular algorithms}
\label{sec:connections} We first note that the $\epsilon$-greedy version of \nv and
\nve algorithms (as shown in Section \ref{app:sec_nvi}, \ref{app:sec_nvie}), are
simply the optimistic Nash-VI algorithm in \cite{liu2021sharp} and \golf algorithm
in \cite{jin2021power} when applied to the tabular setting, with optimistic
exploration replaced by $\epsilon$-greedy exploitation.

Comparing our algorithms \nd (Algorithm \ref{alg:nash_dqn}) and \nde (Algorithm \ref{alg:nash_dqn_exp})
with the $\epsilon$-greedy version of \nv (Algorithm \ref{alg:sec_nvi}) and \nve
(Algorithm \ref{alg:sec_nvie}), we notice that, besides the minor difference
between episodic setting versus infinite horizon discounted setting, the latter two
algorithms are special cases of the former two algorithms when
\begin{enumerate}
    \item specialize the neural network structure to represent a table of values
        for each state-action pairs (i.e. specialize both algorithms to the tabular
        setting);

    \item let the minibatch $\mathcal{M}$ to contain all previous data (i.e.,
        use the full batch $\cD$);

    \item let the number of gradient step $m$ to be sufficiently large so that GD
        finds the minimizer of the objective function;

    \item let $N=1$, that is update the target network at every iterations.
\end{enumerate}

We remark that the use of small minibatch size, and small gradient steps are to speed
up the training in practice beyond tabular settings. The delay update of the
target networks is used to stabilize the training process.

\paragraph{Nash Series Algorithms}
\label{app:sec_compare_tabular_algs} \reb{ We will detail the essential similarities and differences of the four algorithms \nv, Nash Q-Learning, \golf and \nd from four aspects: model-based/model-free, update manner, replay buffer, and exploration method. \begin{itemize}\item \nv: model-based; update using full batch, no soft update; there is a buffer containing all samples so far; $\epsilon$-greedy exploration.

\item Nash Q-Learning: model-free; update using stochastic gradient for each sample, using soft update $Q\leftarrow (1-\alpha) Q + \alpha Q_{\text{target}}$;no replay buffer; $\epsilon$-greedy exploration.

\item \golf: model-based; using an optimistic way of updating policy and exploiter within a confidence set; there is a buffer containing all samples so far; a different behavior policy for exploration compared with $\epsilon$-greedy exploration.

\item \nd: model-free; minibatch stochastic gradient update, using Mean Squared Error($Q, Q_{\text{target}}$) for gradient-based update; there is a buffer containing all samples so far; $\epsilon$-greedy exploration.\end{itemize} From these similarities and differences, we can see that three theoretical algorithms Nash-VI, Nash Q-learning and GOLF-with-exploiter have slight differences in details, Nash-DQN can be viewed as practical approximation of both \textit{Nash-VI} and Nash Q-learning.}

\subsection{Proof of Theorem \ref{thm:theory}}
\label{sec:proof_of_theory} The result of optimistic Nash-VI algorithm in
\cite{liu2021sharp}, and the result of \golf algorithm in \cite{jin2021power} (when
specialized to the tabular setting) already prove that both optimistic versions of
\nv and \nve can find $\epsilon$-approximate Nash equilibria for \textbf{episodic}
Markov games in $\text{poly}(S, A, B, H, \epsilon^{-1}, \log(1/\delta))$ steps
with probability at least $1-\delta$. Here $H$ is the horizon length of the episodic
Markov games.

To convert the episodic results to the infinite-horizon discounted setting in this
paper, we can simply truncate the infinite-horizon games up to
$H = \frac{1}{1-\gamma}\ln \frac{2}{(1-\gamma)\epsilon}$ steps so that the
remaining cumulative reward is at most
\begin{equation*}
    \sum_{h=H}^{\infty} \gamma^{h} = \frac{\gamma^{H}}{1-\gamma}\le \frac{e^{-(1-\gamma)H}}{1-\gamma}
    \le \frac{\epsilon}{2}
\end{equation*}
which is smaller than the target accuracy. To further address the non-stationarity
of the value and policy in the the episodic setting (which requires both value
and policy to depends on not only the state, but also the steps), we can augment
the state space $s$ to $(s, h)$ to include step information (up to the
truncation point $H$) in the state space. Now, we are ready to apply the episodic
results to the infinite horizon discounted setting, which shows that both
optimistic versions of \nv and \nve can find $\epsilon$-approximate Nash
equilibria for \textbf{infinite-horizon discounted} Markov games in $\text{poly}(
S, A, B, (1-\gamma)^{-1}, \epsilon^{-1}, \log(1/\delta))$ steps with probability
at least $1-\delta$. Here $\gamma$ is the discount coefficient.

\chapter{Zero-Sum Video Game\label{ch:zero_sum_video}}
\begin{center}
\begin{quote}
This section is based on paper ``\textit{FightLadder: A Benchmark for Competitive Multi-Agent Reinforcement Learning}''~\cite{li2024fightladder} written in collaboration with Wenzhe Li, Seth Karten and Chi Jin, previously published at ICML 2024.
\end{quote}
\end{center}

\section{Introduction}
As an active branch of artificial intelligence (AI), deep reinforcement learning (DRL) has achieved significant success in various domains, including, but not limited to, strategic games~\cite{silver2016mastering,li2020suphx,moravvcik2017deepstack,vinyals2019grandmaster,berner2019dota}, robotics control~\cite{lillicrap2015continuous,andrychowicz2020learning,brohan2022rt}, and large language models alignment~\cite{ouyang2022training}.
Underpinning these rapid advances are not only the development of sample-efficient RL algorithms but also the availability of well-designed benchmarks. These benchmarks provide environmental platforms, unify evaluation protocols, enable comparisons of state-of-the-art methods, motivate improved solutions, and guide practical applications.
As an example, policy proximal optimization (PPO)~\cite{schulman2017proximal} demonstrates its superior performance across different single-agent RL benchmarks, hence being considered as one of the most widely adopted single-agent RL algorithms~\cite{andrychowicz2020matters}. 
In the realm of multi-agent reinforcement learning (MARL), while a series of benchmarks have also been proposed, most of them focus on fully cooperative settings. For competitive environments, some platforms simulate games with tabular representations and relatively simple dynamics, such as board games, while others, based on complex game engines, require significant computational resources and expert knowledge, such as Starcraft II and DOTA. To advance research on competitive multi-agent reinforcement learning (MARL) and transform game-theoretical results into practical applications, a fully competitive game platform that strikes the right balance between complexity, efficiency, and generality is urgently needed.

\begin{figure}[!t]
\begin{center}
\centerline{\includegraphics[width=\textwidth]{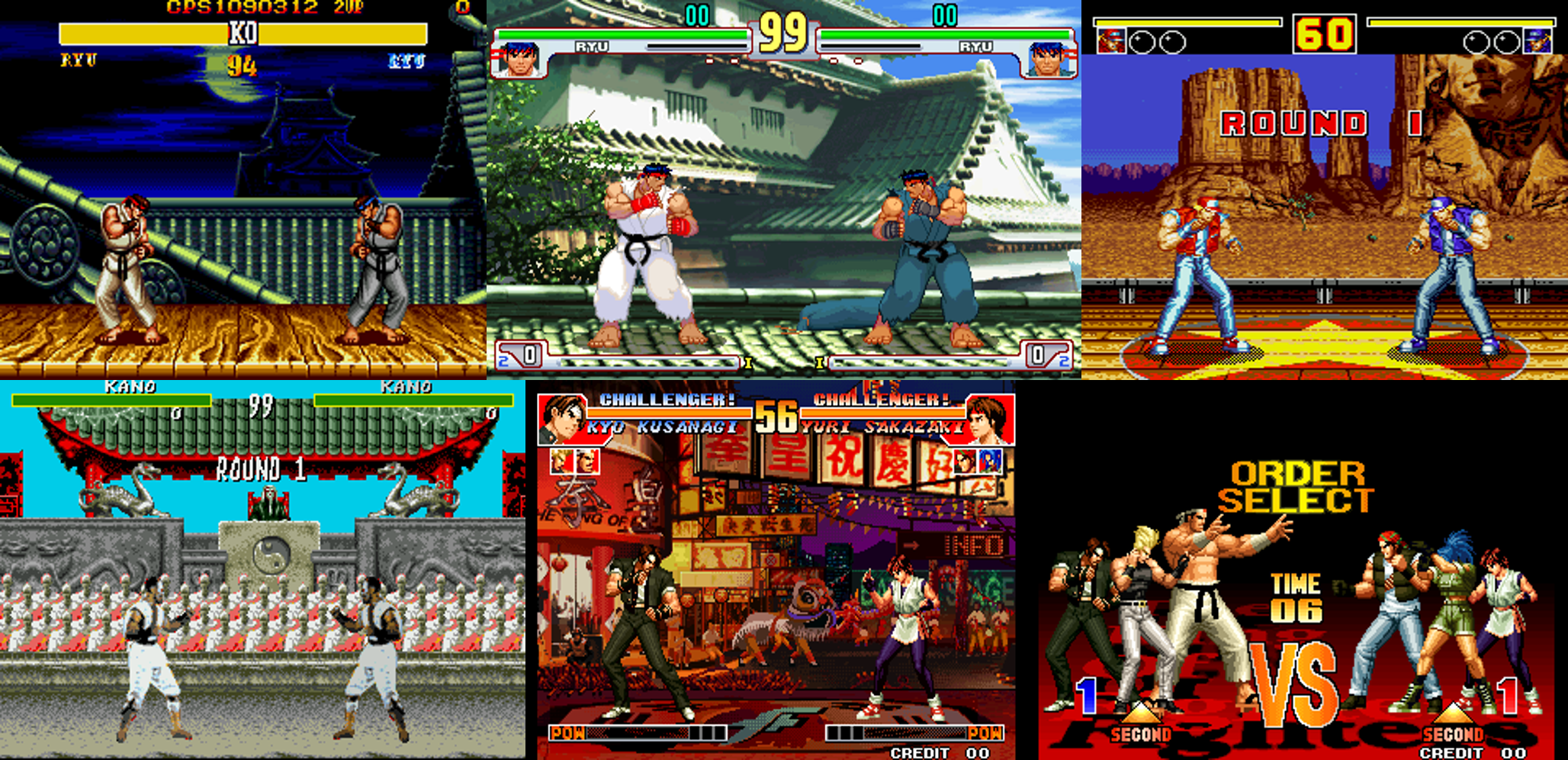}}
\vskip -0.1in
\caption{\benchname{} currently supports various cross-platform video fighting games: \textit{Street Fighter II} (Genesis platform), \textit{Street Fighter III} (Arcade platform), \textit{Fatal Fury 2} (Genesis platform), \textit{Mortal Kombat} (Genesis platform), and \textit{The King of Fighters '97} (Neo Geo platform).}
\label{fig:scenarios}
\end{center}
\vskip -0.4in
\end{figure}

Multi-agent games are known to be more challenging than single-agent ones due to the additional non-stationarity introduced by the interactions with other players. Among different types of interactions, fully competitive settings can be rather difficult. People have a long history of designing and playing competitive games, as well as building strong AI opponents
to make the game more challenging and hence intriguing. Previous AI research has investigated the solutions of competitive games using RL, but mostly for small-scale games like Backgammon~\cite{tesauro1995temporal} or other board games~\cite{schrittwieser2020mastering, brown2018superhuman, brown2019superhuman}. Moreover, this line of work mostly uses state vectors as inputs, which is arguably easier than directly learning from raw pixel inputs that commonly appear in most popular video games. 
In contrast, this paper considers fighting games, which feature rich policy space, and significant depth in strategy --- including catching specific timing, counter-attack by exploiting the stiffness of the opponents, managing energy resources, etc. Moreover, these games also have a rather large number of characters with distinct move-sets which add another layer of complexity for AI agents to master the game.
As a result, we are motivated to build a platform for a series of fighting games, with image inputs and complex fighting dynamics, to serve as a challenging competitive multi-player platform for the broad AI research community.

Apart from the game platform, the evaluation criteria and benchmark results for certain game settings are essential for boosting the field. MARL has been greatly investigated in the past few years for solving multi-player games, from both theoretical and empirical perspectives. A large number of algorithms have been proposed according to specific settings~\cite{sunehag2017value, yu2022surprising, lowe2017multi, silver2018general, lanctot2017unified, vinyals2019grandmaster, ding2022deep}. Nonetheless, for competitive game settings, there is a lack of unified evaluation criteria with thorough comparisons among different approaches.

In this work, we present \benchname{}, a competitive two-player games benchmark. Our contributions are three-fold: We build the \benchname{} platform to support five two-player fighting games, with ease to extend to other games in the future. The games support various observation spaces involving rendered images. Based on prior work, we provide implementations of the most popular algorithms for solving these competitive games, including an AlphaStar league training algorithm~\cite{vinyals2019grandmaster} and policy space response oracle~\cite{lanctot2017unified}. Furthermore, a unified evaluation framework with \textit{Elo rating} and \textit{exploitability} tests are provided alongside the game platforms and algorithm library. We report experimental results using the above toolkits to serve as the baselines for two-player competitive game settings. 
One important challenge of MARL is its diverse nature, which includes collaborative games, competitive games, two-player games, and multiplayer games, all of which have rather different problem structures, properties, and solution concepts. While it is promising to develop a unified solution that addresses them all together, in this work, we empirically demonstrate that  \textit{to some extent, existing methods are still limited in solving competitive two-player zero-sum games alone when combined with visual input, rich strategy space, and lack of extensive human demonstration}. We hope that \benchname{}, which particularly focuses on this fundamental two-player setting, can serve as a stepping stone for the research community to develop effective self-play style algorithms to tackle it first before moving on to even more complicated scenarios, and inspire future directions that involve more types of interactions.

\section{Related Work}



\paragraph{MARL Environments.}
MARL environments can be categorized into three types according to the payoff structure of the game: \textit{fully cooperative}, \textit{fully competitive}, and \textit{general}.


Existing environments for \textit{fully cooperative} games are designed for various scenarios, including simulated games like MAMuJoCo~\cite{peng2021facmac}, card games like Hanabi~\cite{bard2020hanabi}, video games like small-scale StarCraft SMAC~\cite{samvelyan2019starcraft} and Google Research Football~\cite{kurach2020google}, as well as practical scenarios like Traffic Junction~\cite{sukhbaatar2016learning} in a grid world, Flatland~\cite{mohanty2020flatland} for railway networks, network load balancing~\cite{yao2022learning} and CityFlow~\cite{zhang2019cityflow} for city traffic. Cooperative environments feature a single reward function shared by all agents, which makes them distinct from competitive games.

On the other hand, the \textit{fully competitive} game benchmarks are relatively underdeveloped. Prior competitive environments are either on games with low-dimensional or discrete state space such as Pommerman~\cite{resnick2018pommerman} and board games \cite{tesauro1995temporal,schrittwieser2020mastering, brown2018superhuman}; or complex games with image input that require a significant amount of computational resources, such as Starcraft II~\cite{vinyals2019grandmaster} or DOTA~\cite{berner2019dota}. The fighting game environments proposed in this paper strike the right balance between complexity, efficiency, and generality. A few previous works also have explored fighting games: \cite{go2023phase} focuses on developing an algorithm for a single fighting game---street fighter, as opposed to this paper which provides an environment that supports various fighting games. While \cite{palmas2022diambra} provides a platform for fighting games, most of its efforts have been focused on the single-agent setting. It lacks explicit criteria for two-player scenarios with adaptive opponents, and does not provide a benchmark evaluating existing competitive MARL algorithms. \cite{khan2022darefightingice} focuses on fighting games in the blind setting where agents have to rely on acoustic inputs to play.


Finally, there are also a number of environments for \textit{general} multiagent games that feature both cooperation and competition, including MPE~\cite{mordatch2018emergence}, MAgent~\cite{zheng2018magent}, Hide-and-Seek~\cite{baker2019emergent}, DMLab2D~\cite{beattie2020deepmind}, Arena~\cite{song2020arena}, Smarts~\cite{zhou2020smarts}, Neural MMO~\cite{suarez2021neural}, PettingZoo~\cite{terry2021pettingzoo}, MATE~\cite{pan2022mate}, etc. Generic multi-agent general-sum games are rather challenging to evaluate --- even the optimal solution concepts remain elusive. In contrast, the fully competitive setting considered in this paper presents clear game-theoretic properties and well-defined solution concepts. We also remark that while a number of the platforms above support several fully competitive games, they did not provide carefully designed evaluation toolkits as well as extensive baselines for competitive MARL algorithms.

\vspace{-0.2in}
\paragraph{MARL Algorithms and Evaluation.}
To solve multi-agent learning tasks, researchers have proposed algorithms and built libraries for ease of usage and evaluation. PyMARL~\cite{samvelyan2019starcraft} is an initial MARL library built for solving SMAC tasks, while PyMARL2~\cite{hu2021rethinking} extends PyMARL with QMIX~\cite{rashid2020monotonic}. EPyMARL~\cite{papoudakis2020benchmarking} is also an extension of PyMARL, as a unified library for cooperative games supporting different learning paradigms including centralized and decentralized learning, value decomposition, etc. MARLlib~\cite{hu2023marllib} includes major cooperative MARL algorithms like VDN~\cite{sunehag2017value}, MAPPO~\cite{yu2022surprising}, MADDPG~\cite{lowe2017multi}, etc. More recent libraries include Pantheonrl~\cite{sarkar2022pantheonrl}, MAlib~\cite{zhou2023malib}, etc. These libraries mainly support MARL algorithms for cooperative games, lacking support for solving competitive games. 

On the other hand, there is a line of research for solving competitive games with algorithms like self-play~\cite{silver2018general}, fictitious play~\cite{brown1951iterative}, Nash Q-learning~\cite{hu2003nash, ding2022deep}, double oracle~\cite{mcmahan2003planning}, policy space response oracle (PSRO)~\cite{lanctot2017unified} and league training~\cite{vinyals2019grandmaster}. A unified benchmark remains missing to compare and evaluate the efficiency these algorithms on the same set of tasks, especially when combined with deep RL. This paper addresses this issue in the fully competitive setting. We concentrate on two-player zero-sum games, and propose a platform for fighting-style fully competitive games, along with a baseline implementation and evaluation of popular algorithms.


\section{Multi-Agent Reinforcement Learning}
\label{sec:marl}
\benchname{} is designed to motivate novel algorithms for fully competitive two-player games in the domains of MARL and game theory. 
Markov Games (MGs) \cite{shapley1953stochastic} generalize single-player Markov Decision Processes (MDPs) into multi-player settings. Each player has its own utility and optimizes its policy to maximize the utility. The two-player zero-sum setting in MG represents a competitive relationship between the two players. With a shaped dense reward, the games can be generalized to general-sum. 


We denote a finite-horizon two-player general-sum partially observable MG as $\mathrm{POMG}(\mathcal{S}, \mathcal{O}, \mathcal{A}, \mathcal{B}, \mathbb{P}, \mathbb{O}, \{r\}_{i=1}^2, H)$. $\mathcal{S}$ is the state space, which can be partially observable and transformed through an observation emission function $\mathbb{O}$: $\mathcal{S}\rightarrow \mathcal{O}$ to the observation space $\mathcal{O}$. $\mathcal{A}$ and $\mathcal{B} $ are action spaces for two players, respectively. $\mathbb{P}( \cdot | s, a, b)$ is the state transition distribution, $r_i: \mathcal{S} \times \mathcal{A} \times \mathcal{B} \rightarrow \mathbb{R}$ is the reward function for the $i$-th player. In the zero-sum setting, two reward functions satisfy the zero-sum payoff structure $r_1+r_2=0$. $H$ is the horizon length. We denote the policies of two players as $\mu$ and $\nu$, respectively. $V_i^{\mu, \nu} \colon \cS \to \mathbb{R}$ represents the value function for player $i$ evaluated with policies $\mu$ and $\nu$, which can be expanded as the expected cumulative reward starting from the state $s$, 
\begin{equation*} 
\textstyle	 V_i^{\mu, \nu}(s) 
\defeq \E_{\mu, \nu}\big[\sum_{h =
        1}^\infty r_i(s_{h}, a_{h}, b_{h}) \big| s_1 = s\big].
\end{equation*}
In zero-sum games, we have $V_1^{\mu, \nu}(s)=-V_2^{\mu, \nu}(s), \forall s\in \mathcal{S}$ and define $V^{\mu, \nu}(s)=V_1^{\mu, \nu}(s)$ for simplicity.


\begin{definition}[Best Response]
For any policy of the first player $\mu$, there exists a \emph{best response} (BR) against it from the second player, which is a policy
$\nu^\dagger(\mu)$ satisfying $V_{2,h}^{\mu, \nu^\dagger(\mu)}(s) = \max_{\nu} V_{2, h}^{\mu, \nu}(s)$ for
any $(s, h) \in \cS \times [H]$. 
We denote $V_{2,h}^{\mu, \dagger} \defeq V_{2,h}^{\mu, \nu^\dagger(\mu)}$ for simplification. $V_{2,h}^{\mu, \nu}(s)$ is the value function of the second player. BR against the second player can be defined similarly.
\end{definition}

\begin{definition}[Nash Equilibrium]
    The \emph{Nash equilibrium} (NE) in zero-sum setting is defined as a pair of policies $(\mu^\star,\nu^\star)$ 
satisfying the following minimax equation:
\begin{equation*}
\max_{\mu} \min_{\nu} V^{\mu, \nu}(s) = V^{\mu^\star, \nu^\star}(s) = \min_{\nu} \max_{\mu} V^{\mu, \nu}(s).
\end{equation*}
\end{definition}

\begin{definition}[Exploitability]
\label{def:exploitability}
The exploitability for a policy $\mu$ of the first player is defined as 
$V_2^{\mu, \dagger}(s_1) - V_2^{\mu^\star, \nu^\star}(s_1) $, i.e., the value of its BR policy $\nu^\dagger(\mu)$ or the suboptimality gap from the NE value.
The exploitability of the other side policy $\nu$ can be defined accordingly.
\end{definition}

Note that NE strategies will always lead to zero exploitability, thus approaching the non-exploitable strategies is a reasonable pursuit for the game.



\section{\benchname{}} \label{sec:scenarios}
In this section, we present technical details of \benchname{}. In the following part, we first introduce different game settings of \benchname{}, followed by elaborating elements of MGs corresponding to the environment, and conclude with highlighting features of our benchmark.

\subsection{Scenarios}

\benchname{} provides a flexible interface between modern game emulators~\cite{murphy2013hacking, nichol2018retro} and algorithm developers. Thanks to its flexibility, \benchname{} can support a wide range of classical fighting games over the past decades, including Street Fighter, Mortal Kombat, Fatal Fury, and The King of Fighters, some of which are still very popular nowadays. Figure~\ref{fig:scenarios} shows screenshots of several fighting games provided by \benchname{}. With this diverse set of supported games, we can benchmark algorithms on various fighting scenarios differing in backgrounds, characters, and moving dynamics, which can further motivate novel algorithms that are general rather than overfitting to one specific game. For better readability and clarity, we would use Street Fighter as an example for illustration and evaluation in the rest of the paper. The other fighting games are very similar, and readers could refer to Sec.~\ref{appx:scenarios} for more details. We name each scenario in the form \emph{[game alias]\_[character left]\_vs\_[character right]}, for example \emph{sf\_ryu\_vs\_ryu} in Street Fighter.

While \benchname{} mainly focuses on the competitive two-player setting, the nature of fighting games allows it to be seamlessly deployed to the single-player scenario where the agent's task is to compete against a built-in game AI (e.g., \emph{sf\_ryu\_vs\_ryu(cpu)}). Under this single-player setting, users have the freedom to choose characters and set up the difficulty of the scripted AI opponent. Moreover, our benchmark also supports training in a much more challenging full-game scenario (e.g., \emph{sf\_ryu\_full\_game}), where the agent needs to defeat all 12 characters controlled by computers with the difficulty progressively increasing. As we shall see in later experiments, this scenario could also serve as a sanity check for our baseline algorithms to see whether they could learn effective behaviors from the environment.

\subsection{State and Observations}
We define the state space $\mathcal{S}$ as the complete set of attributes stored in the game emulator after each step of action. Same as human players, the agent is not allowed to access the underlying full state but can only access the observation space $\mathcal{O}$ of pixels, which forms a 128$\times$100 RGB image corresponding to the rendered screen. This image includes the position and movement of both sides of the players, as well as the hit-point bar and the round timer on the top of the screen. At every step, a configurable number of images are stacked as the input of the agent. 

While we use pixels as default observations, we also provide an interface for users to access additional information about the game status, including position, hit-point, and exact countdown number for agents on both sides. Users can leverage these attributes to better understand the agent's behavior or augment feature representations. More details are provided in Sec.~\ref{appx:scenarios}.

\subsection{Action Space}
\begin{figure}[t]
\hskip -0.1in
\includegraphics[width=0.5\columnwidth]{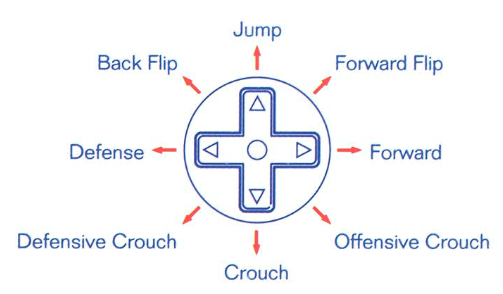}
\includegraphics[width=0.5\columnwidth]{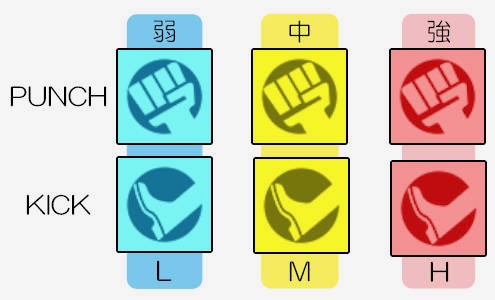}
\vskip -0.1in
\caption{Motion and attack action spaces of fighting games. Images are adapted from Instruction Manual of Street Fighter II.}
\label{fig:actions}
\end{figure}

In fighting games, two players share the same action space $\mathcal{A}$.
The native \emph{human action space} $\mathcal{A}_\text{human}$ is designed to mimic the joystick control of arcade games, which is a 12-dimensional binary space (['B', 'A', 'MODE', 'START', 'UP', 'DOWN', 'LEFT', 'RIGHT', 'C', 'Y', 'X', 'Z']) with each dimension representing a button being pressed or not. Note that due to the nature of fighting game engines, this space contains many redundant actions that are invalid, for instance, moving in opposite directions or moving and attacking at the same moment. To filter out these redundant actions and to construct a more structured space, we develop a categorical \emph{transformed action space} 
$\mathcal{A}_\text{trans}$ through an encoding function $F:\mathcal{A}_\text{human} \rightarrow \mathcal{A}_\text{trans} $. Specifically, $\mathcal{A}_\text{trans}$ is the joint set of a direction move set $\mathcal{A}_\text{motion}$=\{defense, forward, jump, crouch, back flip, front flip, offensive crouch, defensive crouch\} and an attack move set $\mathcal{A}_\text{attack}$=\{light punch, medium punch, hard punch, light kick, medium kick, hard kick\}, as shown in Figure~\ref{fig:actions}. Each action will remain a number of frames according to users' configuration. The games also have special techniques called \textit{close attack}, i.e., Throws and Holds, which can be applied in certain regions near the opponent.

\begin{figure}[t]
\begin{center}
\includegraphics[width=0.1\columnwidth]{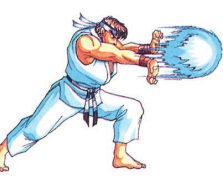}
\includegraphics[width=0.1\columnwidth]{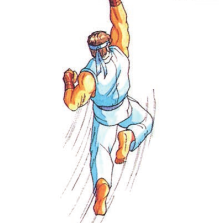}
\includegraphics[width=0.2\columnwidth]{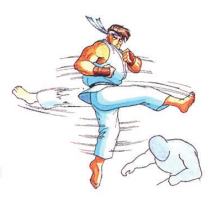}
\caption{Example of special moves for character Ryu in StreetFighter II (left to right): Fireball, Dragon Punch, Hurricane Kick. Images are adapted from Instruction Manual of Street Fighter II.}
\label{fig:special_move}
\end{center}
\end{figure}

In addition to the standard move set, one signature element of fighting games is \textit{special moves}, which is a kind of powerful attack or maneuver that requires the player to follow a specific action sequence (i.e., sequential keys combination, or combination of key holding and key pressing), with an example depicted in Figure~\ref{fig:special_move}. These moves usually have special properties (e.g., invincibility frames, larger coverage, etc.) and play a critical role in the strategy and depth of the game. They are especially useful for higher levels of play, from which players could create complex combos and outperform opponents. However, we observe that learning to perform special moves from scratch can be challenging to baseline algorithms, as it requires the agent to memorize frames and actions in previous steps and accurately perform the next action in the action sequence of special moves. Moreover, the special moves can be different from character to character, which increases the difficulty of the game. Therefore, to alleviate this challenge, we also include hard-coded special move lists as one part of the action space so that the agent can directly access special moves with one single action. 


\subsection{Rewards}
\paragraph{Sparse Reward.}
Both sides of the agents are to maximize their win rate for each round of the game. The \emph{sparse reward} $r_\text{sparse}$ assigns +1 for the winner and -1 for the loser at the end of each episode. In the sparse reward setting, all fighting games are two-player zero-sum
games, which are theoretically guaranteed to exist at least one Nash Equilibrium~\cite{filar2012competitive}, which directly induces a pair of non-exploitable policies.

\paragraph{Win Rate.}
For two players $A$ and $B$, policy $\pi_A$ winning against policy $\pi_B$ can be defined as a reward relationship $r^{A}_\text{sparse}(\pi_A, \pi_B) > r^{B}_\text{sparse}(\pi_A, \pi_B)$ in a single match, with $r^{A}_\text{sparse}$ and $r^{B}_\text{sparse}$ as the sparse reward for players A and B in the zero-sum setting. The win rate is defined as the probability of winning as $p(\pi_A\succ \pi_B)$.


\paragraph{Shaped Dense Reward.} While sparse reward is straightforward for evaluation, we discover that baseline algorithms could not effectively learn to behave well from such a sparse signal. To address this issue, we introduce a \emph{shaped dense reward} $r_\text{dense}$ for training, which is a weighted sum of the hit-point damage inflicted by the agent on the opponent and the damage it receives, together with a bonus (penalty) for winning (losing) the game. Specific format of this reward refers to Sec.~\ref{appx:dense_reward}. The dense reward $r_\text{dense}$ is chosen to coincide with the win rate of the policy, such that $\pi_A\succ \pi_B$ will always lead to $ r^{A}_\text{dense}(\pi_A, \pi_B) > r^{B}_\text{dense}(\pi_A, \pi_B)$ in expectation. The dense reward also offers some flexibility,
that the user can control the agent's aggressiveness by configuring the weighing scales in the reward function.

\subsection{Features}
We remark on the following features of the proposed benchmark that could benefit MARL research.

\paragraph{Rich Strategy Space.} 
One key feature of our benchmark is the rich strategy space as the nature of fighting games, which is particularly beneficial to the development of game-theoretical algorithms. To name a few, fighting games require players to consider \textbf{(a) character diversity:} each character has a unique skill set with different strengths and weaknesses, so one needs to master the strategy and counter-strategy of all possible opponents, and even reason how to select and order characters when they have the freedom to do so; \textbf{(b) complexity of mechanics:} fighting games are designed with sophisticate mechanics such as invincibility frame, hitboxes, and combo systems, which are challenging for micromanagement of characters; and \textbf{(c) adversarial opponents:} opponents may progressively adapt their policies to players' policies, thus finding non-exploitable policies is crucial in mastering fighting games. 

\paragraph{Various Difficulty Levels.}
FightLadder provides several kinds of scenarios: single-player mode against one CPU player (e.g., \emph{sf\_ryu\_vs\_ryu(cpu)}), single-player mode full game (e.g., \emph{sf\_ryu\_full\_game}), two-player mode (e.g., \emph{sf\_ryu\_vs\_ryu}), team mode (supported in some games such as The King of Fighters). The difficulty levels are increasing in this order, as two-player mode (no CPU) introduces additional non-stationary (opponents can be adaptive), and team mode offers a richer strategy space. Moreover, FightLadder supports specifying \textbf{arbitrary difficulty levels} of CPUs and \textbf{arbitrary characters} for both the player and its opponent. This enriches the features of our platform and the diversity of strategy space.

\paragraph{Computational Efficiency.}
\benchname{} also enjoys efficient computation for its usage, and the comparison with several other popular game environments is shown in Table~\ref{tab:fps}. The frame rate is 13 times faster than SMACv2, with one-fourth usage of the memory. While it is less efficient than \benchname{} is the PettingZoo Atari, it provides more game complexity. The balance of complexity and low computational cost is important for evaluating algorithms at scale. 

\begin{table}[t]
\vskip -0.1in
  \centering
  \caption{FPS and memory usage of several open-sourced platforms.}
  \vskip 0.1in
  \label{tab:fps}
  \resizebox{0.6\columnwidth}{!}{
  \begin{tabular}{lccc}
    \toprule
    \textbf{Environment} & \textbf{Speed (FPS)} & \textbf{Memory (MB)} \\
    \midrule
    \benchname{} (Ours) & 1935.76 & 195.46 \\
    SMACv2 & 146.72 & 876.96 \\
    PettingZoo Atari & 6268.18 & 32.13\\
    DMLab2D & 1144.27 & 47.41 \\
    \bottomrule
  \end{tabular}
  }
\vskip -0.1in
\end{table}

\paragraph{Fidelity and Popularity.}
\benchname{} allows testing agents in full-length fighting games with an interface similar to human perception, thus providing a high-fidelity evaluation of competitive RL algorithms. Moreover, fighting games have been gaining popularity since they were released, making it easier to test the learned RL agents against human expert players. 

\paragraph{Open-Source and Compatibility.}
\benchname{} is designed for the broad RL research community, so we make efforts to improve the ease of usage and make it accessible to all potential users. It is compatible with the Gym~\cite{brockman2016openai} interface so that users can leverage off-the-shelf RL algorithms implementation. 

\paragraph{Customization, Extension, and Flexibility.}
\benchname{} is extremely flexible for configuration and extension. For customization, the users can customize action spaces (human/transformed action), reward functions (sparse/tunable shaped dense reward), number of frames to be observed per step, as well as access to additional information to help training. Moreover, our platform is built upon popular modern game emulators so that it is easy to extend to other games not provided by us. Specifically, it supports \href{https://github.com/openai/retro}{Gym Retro} and \href{https://github.com/M-J-Murray/MAMEToolkit}{MAMEToolkit}, which already support a wide range of games. This extension capability of
diverse games is provided by our platform with minimal engineering efforts. Please check our \href{https://sites.google.com/view/fightladder/home}{open-source project}\footnote{\href{https://sites.google.com/view/fightladder/home}{https://sites.google.com/view/fightladder/home}} for more details.

\section{Evaluation Metrics} \label{sec:metrics}
\paragraph{Versus Built-In Game AIs.} 
Directly competing with the built-in AIs of the games provides a straightforward way of measuring policy performance.
Typically, fighting games offer a hierarchical structure of levels, enabling players to adjust the difficulty setting (for example, Street Fighter features eight distinct levels). This structure allows for the empirical evaluation of the policy against the game's scripted AI at varying levels of challenge. It is important to acknowledge, however, that the limitations associated with hard-coded adversaries restrict the extent to which this metric can accurately reflect the policy's real capability. For brevity, we shall refer to such agents as CPU.

\paragraph{Elo Ratings.} 
The skills of agents can be ranked through the FIDE rating system~\cite{elo1978rating}, which is an incremental learning system that increases the Elo of winners and decreases the Elo of losers.
The larger the difference in Elo between players $A$ and $B$, the higher the probability that the player with the higher Elo, $A$, beats the player with the lower Elo, $B$. The Elo score calculation takes the following procedures:

First, the probability of player $A$ winning is estimated with,
\begin{equation*}
   p_A:= p(\pi_A \succ \pi_B) = ( 1.0 + 10^{\frac{\text{Elo}_B - \text{Elo}_A}{400}} )^{-1}.
\end{equation*}
Then the Elo rating for player $A$ as $\text{Elo}_A$ will be updated with following formula:
\begin{equation*}
    \text{Elo}_A = \text{Elo}_A + k\cdot (\mathds{1} [\text{winner} = A] - p_A),
\end{equation*}
where $k$ is a constant of update rate.
The update is symmetric for player $B$, as well as any other player in the ranking system.


\paragraph{Versus AI Exploiters.} 
As discussed in Section~\ref{sec:marl}, exploitability (as Definition~\ref{def:exploitability}) measures the distance of a policy to the Nash equilibrium of the game. Specifically, the exploitability of a policy $\mu$ is measured by the win rate of its BR policy $\nu^\dagger(\mu)$ against $\mu$, since $V^{\mu^\star, \nu^\star}(s_1)=0$ for symmetric zero-sum game and  $V_2^{\mu, \dagger}(s_1)= 1 \cdot p(\nu \succ\mu) + 0 \cdot p(\nu \preceq \mu) = p(\nu \succ\mu)$ for sparse reward setting.
In practice, we can use any single-agent deep RL algorithm as an exploiter to approximately learn the BR policy $\nu^\dagger(\mu)$. For fair comparisons, we should use one consistent exploiter (same RL algorithm with same configurations) to evaluate the exploitability of different baselines.

\paragraph{Versus Human Players.} 
While Definition~\ref{def:exploitability} is a general metric to measure exploitability, it may be limited to the capability of deep RL algorithms in usage. Therefore, we also provide an interface for human players such that they can play with any learned model with convenience. This feature will show the strengths and weaknesses of agents directly and visibly, and motivate developers to improve their algorithms to be more non-exploitable in general. Given the remarkable success of modern RL algorithms outperforming expert human players in various video games~\cite{mnih2013playing, vinyals2019grandmaster, berner2019dota}, we believe that \benchname{} will emerge as a promising platform for the broad competitive MARL community and researchers will eventually build AI agents that could beat world champions in a much richer set of strategic games with significantly less engineering efforts.

\section{\benchname{}-Baselines} \label{sec:baselines}
For the convenience of the community to evaluate existing methods and new algorithms on FightLadder platform, we open-source the implementation of several state-of-the-art (SOTA) competitive MARL algorithms,
including independent learning~\cite{de2020independent}, two-timescale learning~\cite{daskalakis2020independent}, fictitious self-play~\cite{pmlr-v37-heinrich15}, policy-space response oracle~\cite{lanctot2017unified} and league training~\cite{vinyals2019grandmaster}. Our codebase supports decentralized learning across multiple GPUs, and it is built upon Stable-Baselines3~\cite{stable-baselines3} so that users can leverage off-the-shelf implementations of RL algorithms. We choose proximal policy optimization (PPO)~\cite{schulman2017proximal} as the backbone policy optimization algorithm in our experiments. More details of baseline algorithms refer to Sec.~\ref{appx:baselines}.


\begin{figure*}[htbp]
\begin{center}
\centerline{\includegraphics[width=\textwidth]{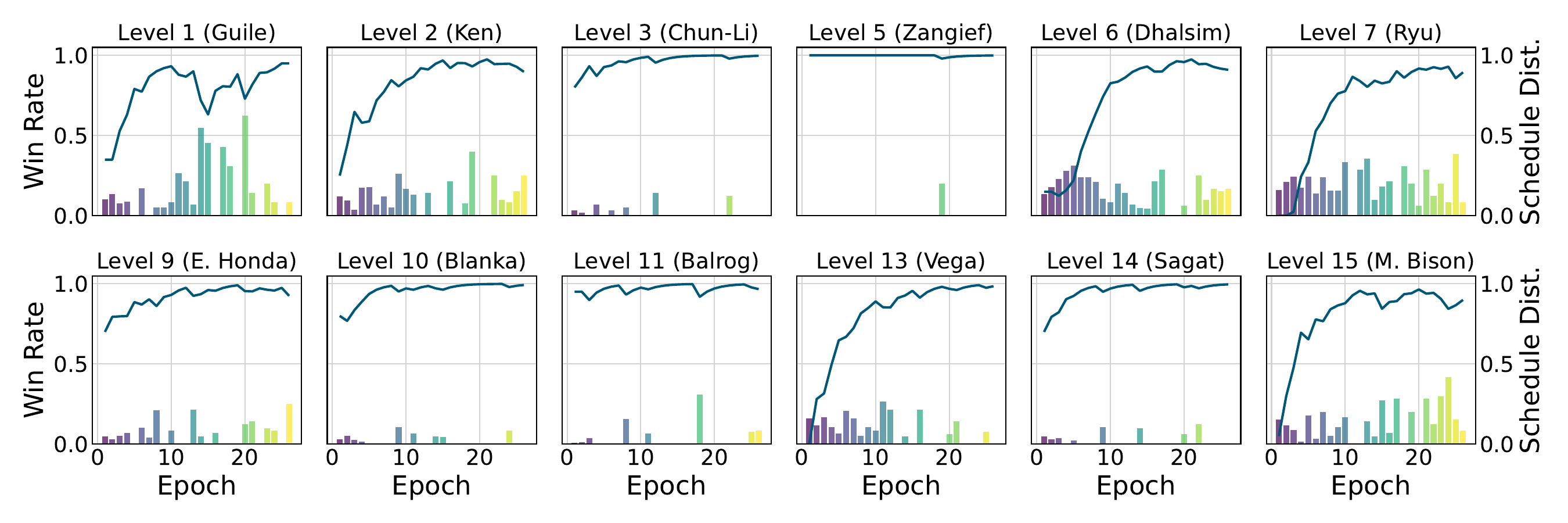}}
\caption{The win rate curves and the scheduling distribution bar plot in \emph{sf\_ryu\_full\_game} via the proposed PPO with curriculum learning. Opponents of different characters are marked with different levels. Levels 4, 8, and 12 are omitted as they are bonus levels without fighting.}
\label{fig:1p_win_rate}
\end{center}
\vskip -0.3in
\end{figure*}

\section{Results}
In this section, we provide benchmark results on a selected game in \benchname{}--the Street Fighter. We aim to answer the following questions through our benchmark: \textbf{(a)} Can existing RL algorithms solve the full video game in the single-player scenario? \textbf{(b)} How does the performance of state-of-the-art baseline algorithms in the two-player competitive setting compare? and \textbf{(c)} Does multi-agent training help to improve the non-exploitability? 

\subsection{Single-Player Full Video Game}
To answer question \textbf{(a)}, we evaluate PPO's performance in the scenario \emph{sf\_ryu\_full\_game} with human action space as a feasibility check. As mentioned in Section~\ref{sec:scenarios}, this scenario requires the agent to learn a generalizable policy to compete against all different characters with increasing difficulty levels. \textit{Curriculum learning} is applied to train the policy from easy to hard cases. Furthermore, to improve learning efficiency we develop a curriculum scheduler for opponent sampling to match with the learner after each epoch. More specifically, for the current learner $L$ with policy $\pi_L$, we sample its opponent $C$ from the entire character set $\mathcal{C}$, with the following inverse-weight scheduling distribution: $$C\sim \Delta(\mathcal{C})\propto 1-p(\pi_L\succ \pi_C),$$ where $p(\pi_L\succ \pi_C)$ is the win rate of the learner against the opponent and $\Delta(\cdot)$ is the simplex. Intuitively, such a curriculum will encourage the agent to focus on the hardest opponents, similarly to prioritized experience play~\cite{schaul2015prioritized}.

Figure~\ref{fig:1p_win_rate} shows the performance of our proposed method during training. With 20 epochs of training (each epoch involves 10M training steps competing with opponents sampled from the curriculum scheduler in parallel), the agent is capable of defeating characters at each level with a win rate close to 1. In addition to beating each character with a high probability, the trained policy can complete the full video game with over 0.6 win rate, outperforming human players with hours of playing experience. This result shows that existing RL algorithms can already learn a well-behaved policy to solve the full single-player video game, which provides a good starting point for exploring the multi-agent setting.

As an additional experiment, we also test the inclusion of hard-coded special move lists in this setting with exactly the same algorithm implementation. Although it could be easier for the agent to learn more offensive policies, significant improvement in the overall win rate is not observed. It indicates that the agents without encoded special moves can also effectively learn policies against CPUs. Constantly playing special moves will lead to a vulnerable situation for the agent, whereas the defensive strategy also matters greatly in the game. Moreover, given that an experienced human player can perform special moves easily (by executing the action sequences almost instantly), we do not think that hard-coded special move lists will become the advantage of trained agents over human players.

\subsection{Performance of Two-Player Baseline Algorithms} \label{sec:2p_exp}
\begin{figure}[!h]
    \centering
    \includegraphics[trim=5cm 0.5cm 5cm 0.5cm,clip,width=\columnwidth]{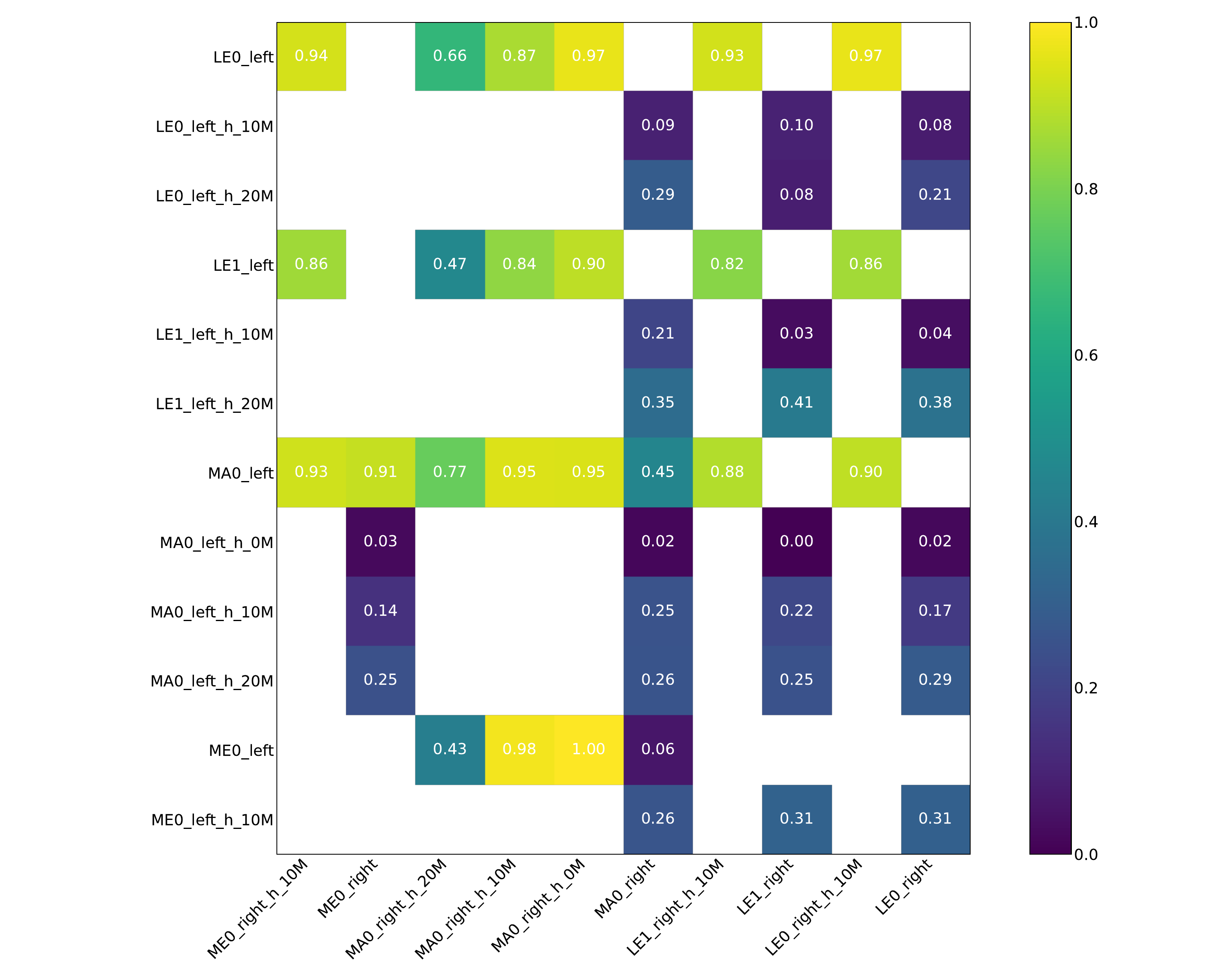}
    \caption{The payoff matrix for each pair of agents at a certain stage of League training. For league training, there is one main agent (MA), two league exploiters (LE0, LE1), and one main exploiter (ME) for each side (left or right). The name of each row indicates the agent information as \texttt{Character\_Side\_Checkpoint}. \texttt{Checkpoint=h\_xM} represents a historical version of agent saved at \texttt{x} million steps. The value indicates the win rate of the left (row) player against the right (column) player. For instance, \texttt{ME0\_right} wins all \texttt{MA0\_left\_h\_xM} with high probability, indicating that main exploiters in the league can fully exploit previous main agents. Also the high win rate of \texttt{MA0\_left} against all right agents (except \texttt{MA0\_right}) shows that the main agent at current steps outperforms other agents in the league.}
    \label{fig:league_payoff_main}
\end{figure}

\begin{figure}[t]
\vskip -0.1in
    \centering
    \includegraphics[width=0.6\columnwidth]{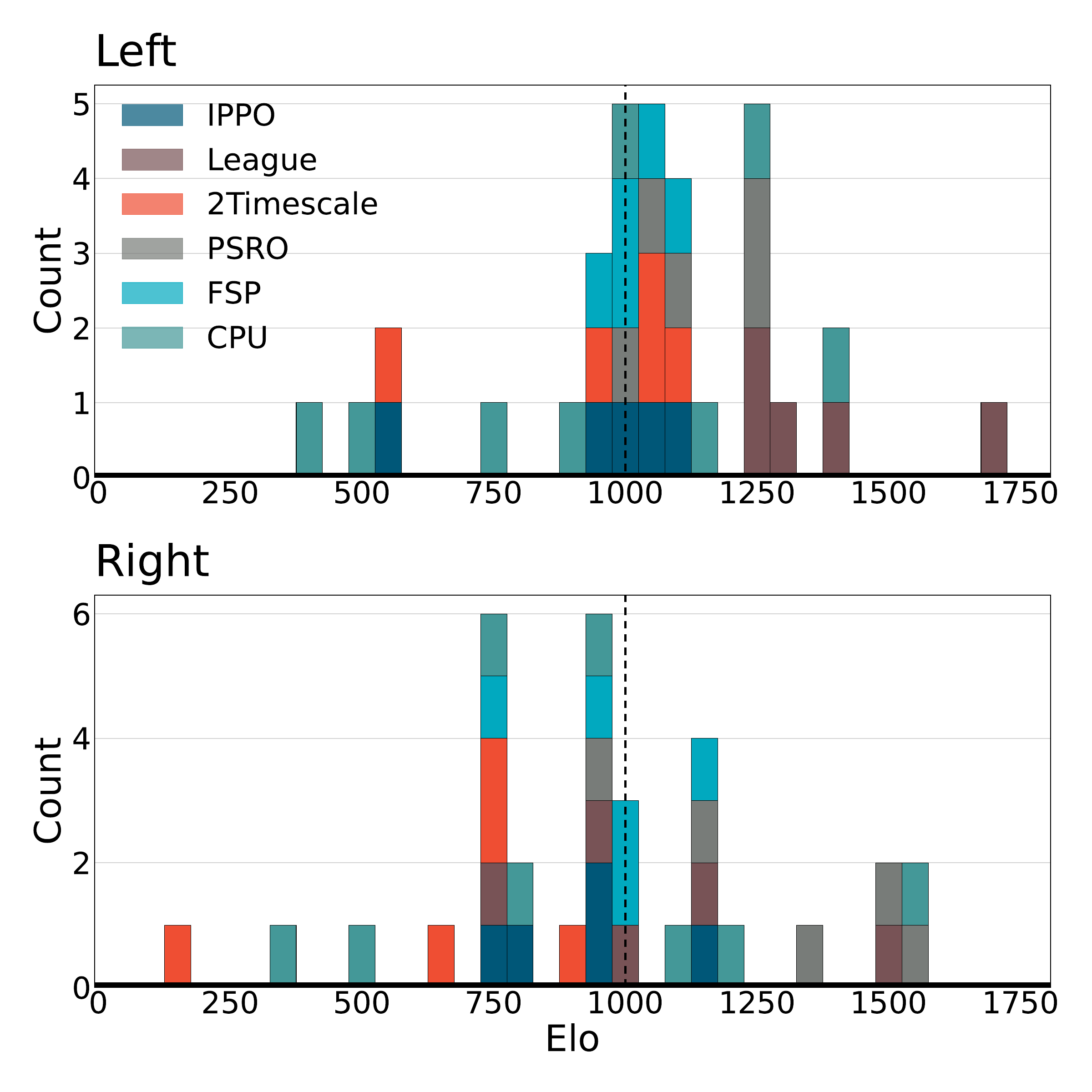}
    \vskip -0.2in
    \caption{The distribution of Elo ratings for top ten agents from each baseline.}
    \label{fig:top_elo}
\vskip -0.2in
\end{figure}

To answer question \textbf{(b)}, we evaluate five SOTA algorithms mentioned in Section~\ref{sec:baselines}: independent PPO (IPPO), two-timescale IPPO (2Timescale), fictitious self-play (FSP), policy-space response oracles (PSRO), and league training (League) in the scenario \emph{sf\_ryu\_vs\_ryu}. IPPO and 2Timescale can be categorized into the independent learning paradigm, while FSP, PSRO, and League can be categorized into the population-based learning paradigm. For each algorithm, we initialize the population of agents with a pretrained policy in \emph{sf\_ryu\_vs\_ryu(cpu)} against the most difficult CPU\footnote{We do not pre-train in \emph{sf\_ryu\_full\_game} as \emph{sf\_ryu\_vs\_ryu} does not require skills to compete with other characters rather than Ryu.}. We use the transformed actions $\mathcal{A}_\text{trans}$ with hard-coded special moves to unleash the full potential for agents. As a fair comparison, we use the same codebase (\benchname{}-Baselines) and fix the hyperparameters of the backbone PPO algorithm. We train IPPO and 2Timescale for approximately 50M steps until the Elos saturate across all three seeds, FSP and PSRO for approximately 250M steps, and League for approximately 700M steps due to a larger population. A slice of the league during the league training process is visualized in Figure~\ref{fig:league_payoff_main}.

For each algorithm, we report the training Elos of agents in the population during the course of training, respectively. The results are shown in Sec.~\ref{appx:indiv_elo}, which reveal that all baseline algorithms are improving their policies at the onset of training. Subsequently, IPPO and 2Timescale gradually converge and oscillate around the peak Elos, where FSP, PSRO, and League continue to increase their scores. This suggests that IPPO and 2Timescale may suffer from optimization issues during training and population-based methods may be more suitable for policy learning in fighting games.

To compare different baseline algorithms, we select the top ten agents (five on each left or right side) from each algorithm to form a new population, and compute the test Elos for this group of agents and CPU policies. We report the highest Elos for each algorithm in Table~\ref{tab:top_elo} and the distribution of these agents' Elos in Figure~\ref{fig:top_elo}, where we find that League and PSRO significantly outperform other baselines, and population-based methods deliver better results than independent learning counterparts, which is aligned with our previous observation inspecting Elos of baselines individually. On the other hand, we notice that CPU policies may defeat most of the agents in this group except for a few best-performing agents, suggesting that it is still very challenging for existing SOTA algorithms to reach an advanced or superhuman level of performance in these fighting games. We also noticed that two sides of agents reveal asymmetric strengths in terms of Elos in both individual evaluation for each algorithm (Sec.~\ref{appx:indiv_elo}, Figure~\ref{fig:ippo_elo}-\ref{fig:league_elo}) and overall evaluations across algorithms (Table~\ref{tab:top_elo}). Such an imbalance may result from various factors, for instance, optimizing instability, variance from the population or Elos computation, etc, and can be an interesting research question for future work.


\begin{table}[htbp]
  \centering
  \caption{Comparison of training steps and the best Elo ratings among baselines, with CPU's Elos as references.}
  \label{tab:top_elo}
  \resizebox{0.7\columnwidth}{!}{
  \begin{tabular}{lcc}
    \toprule
    \textbf{Method} & \textbf{Training Steps} (Left/Right) & \textbf{Elo} (Left/Right) \\
    \midrule
    IPPO & 46M / 46M & 1082 / 1164 \\
    League & 647M / 630M & \textbf{1682} / \textbf{1503} \\
    2Timescale & 51M / 46M & 1080 / 919 \\
    PSRO & 176M / 161M & 1262 / \textbf{1517} \\
    FSP & 262M / 244M & 1079 / 1150 \\
    \textcolor{gray}{CPU} & \textcolor{gray}{N/A} & \textcolor{gray}{1395} / \textcolor{gray}{1541} \\
    \bottomrule
  \end{tabular}}
\end{table}



\subsection{Non-Exploitability of Trained Agents}
\begin{table}[t]
  \vskip -0.1in
  \centering
  \caption{Comparison of methods' exploitability. A lower number indicates the evaluated policy is more robust to exploitation.}
  \label{tab:exploitability}
  \vskip 0.1in
  \begin{tabular}{lcc}
    \toprule
    \textbf{Method} & \textbf{Exploitability} (Left/Right) \\
    \midrule
    IPPO       & 0.96 $\pm$ 0.03 / 0.91 $\pm$ 0.03 \\
    League     & \textbf{0.94 $\pm$ 0.05} / 0.94 $\pm$ 0.00 \\
    2Timescale & 0.96 $\pm$ 0.02 / 0.90 $\pm$ 0.05 \\
    PSRO       & 0.97 $\pm$ 0.02 / \textbf{0.88 $\pm$ 0.05} \\
    FSP        & 1.00 $\pm$ 0.00 / 0.95 $\pm$ 0.01 \\
    PPO        & 0.99 $\pm$ 0.02 / 0.99 $\pm$ 0.01 \\
    \bottomrule
  \end{tabular}
  \vskip -0.2in
\end{table}

To answer question \textbf{(c)}, we measure the non-exploitability of baseline algorithms according to the evaluation approaches proposed in Section~\ref{sec:metrics}. More specifically, we choose models with the highest Elos from each two-player baseline algorithm respectively, and compare their exploitability with the single-player pretrained model used for initializing the population-based methods in Section~\ref{sec:2p_exp}.

The practical exploitability is calculated by setting the trained policy fixed on one side, and deploying a PPO agent on the other side as an exploiter. The PPO exploiter will be trained until convergence, and the success rate of the exploiter is the estimated exploitability of the original policy, according to Definition~\ref{def:exploitability}.

\paragraph{Single-Agent RL Exploiters.} 
We use PPO as the algorithm for training exploiters, given its decent performance in both single-player and two-player scenarios shown in previous experiments. Table~\ref{tab:exploitability} shows the exploitability of comparing methods evaluated across three seeds, from which we observe that the single-player pretrained policy via PPO is easier to exploit and suffers from higher exploitability than almost all selected policies from two-player baselines. Therefore, this result indicates that two-player learning algorithms such as League and PSRO can help to improve the robustness of learned policies. On the other hand, the PPO exploiter eventually learns to beat policies from all baselines (with a win rate greater than 0.5), which means that none of these algorithms can result in the exact Nash equilibrium policies, or even close to it. Therefore, closing this gap is a challenging direction for future research.

\paragraph{Human Players as Exploiters.}
In addition to exploiting the learned models with RL algorithms, we also attempt to exploit their policies with human effort. During human evaluations, the evaluated models reveal some robustness to human players (e.g., defend when a human player attacks), but some simple strategies (e.g., defensive posture combined with low kicks at proper timing) could still defeat them rather consistently.

Therefore, based on two exploiting experiments, we observe that \textit{existing competitive MARL algorithms are found hard to learn non-exploitable strategies in competitive fighting games like Street Fighter}, thus raising a new challenge for the research community.

\section{Details of FightLadder}

\subsection{Dense Reward}
\label{appx:dense_reward}
The shaped dense reward for the $i$-th agent at step $t$ is defined as follows:
\begin{align}
    r_{i,t} = \alpha\left[\lambda(\text{HP}_{-i,t-1}-\text{HP}_{-i,t})-(\text{HP}_{i,t-1}-\text{HP}_{i,t}) + r_{i,\text{bonus}}\right],
\end{align}
where $\alpha$ is a scaling factor, $\text{HP}_{i,t}$ denotes agent $i$'s hit-point at step $t$ and $\lambda$ control the aggressiveness of learned agents, and $-i$ denotes the opponent agent. At the end of the game, the agent $i$ will receive a bonus reward $r_{i,\text{bonus}}$, which is positively correlated to $\text{HP}_i$ if it wins and negatively correlated to $\text{HP}_{-i}$ if it loses. By default, we choose $\lambda=3$ in SF2, FF2, and MK, and $\lambda=1$ in SF3 and KOF97, for the consideration of practical performances.

\subsection{Game Settings} \label{appx:scenarios}
Table~\ref{tab:games} illustrates the observation, action, and rewards as well as other elements in the environment for all supported games --- Street Fighter II (SF2), Fatal Fury 2 (FF2), Mortal Kombat (MK), Street Fighter III (SF3), and The King of Fighters '97 (KOF97).

\begin{table}[!h]
  \vskip -0.1in
  \centering
  \caption{Specification of supported games in \benchname{}.}
  \label{tab:games}
  \vskip 0.1in
  \small
  \resizebox{\textwidth}{!}{
  \begin{tabular}{lccccc}
    \toprule
    & \textbf{SF2} & \textbf{FF2} & \textbf{MK} & \textbf{SF3} & \textbf{KOF97} \\
    \midrule
    Observation (Pixels) & 100$\times$128$\times$3 & 112$\times$128$\times$3 & 112$\times$160$\times$3 & 112$\times$192$\times$3 & 112$\times$192$\times$3 \\
    Human Action Supported & Yes & Yes & Yes & Yes & Yes \\
    Transformed Action Supported & Yes & Yes & Yes & No & No \\
    Shaped Dense Reward & Yes & Yes & Yes & Yes & Yes  \\
    Default Frames Per Step & 8 & 8 & 8 & 3 & 3 \\
    Default Frames Stacked\tablefootnote{We uniformly sample the stacked frames as observations to improve the computational efficiency.} & 12 & 12 & 12 & 9 & 9 \\
    \multirow{2}{*}{Additional Available Info} & HPs, Countdown, & HPs, Countdown & HPs, Countdown, & HPs & HPs, Countdown, \\ 
    & Scoreboard, Positions & & Scoreboard & & Positions, Power Status \\
    \bottomrule
  \end{tabular}
  }
\end{table}

\subsection{Comparison of MARL Game Platforms} \label{appx:compare_platforms}
Table~\ref{tab:environment_comparison} compares our FightLadder with several popular MARL game platforms mostly focusing on competitive settings, in terms of observation space, action space, whether baseline methods are included and the number of agents in games. For the observation space, `Continuous' indicates a vector-form latent state information of the game with continuous numerical values, and `Image' indicates visual RGB information as observations. PommerMan~\cite{resnick2018pommerman} uses grid environments therefore its observation only has discrete values. For the action space, most of the games only involves discrete action values except for Arena~\cite{song2020arena}. For the number of agents in these platforms, MPE provide diverse competitive settings like 1v1, 1v$N$, 1v1v1 and so on. MAgent includes 1 million agents competing againts each other, and for Neural MMO~\cite{suarez2021neural} the number of agents is 256 or 1024. The team mode in our FightLadder and Arena supports the competitive settings of two teams, where each team includes multiple characters to be controlled by one team policy or separate agent policies.

\begin{table}[h!]
\centering
\caption{Comparison of popular MARL game platforms.}
\vskip 0.1in
\resizebox{\textwidth}{!}{
\begin{tabular}{ccccc}
\hline
\textbf{Env} & \textbf{Observation Space} & \textbf{Action Space} & \textbf{Baselines} & \textbf{\# Agents} \\ \hline
MPE \cite{mordatch2018emergence} & Continuous & Discrete & Yes & 1v1, 1v$N$ and 1v1v1... \\ \hline
MAgent \cite{zheng2018magent} & Continuous+Image & Discrete & Yes & 1 million \\ \hline
Arena \cite{song2020arena} & Continuous+Image & Continuous/Discrete & Yes & 1v1, $N$v$N$ and team mode \\ \hline
Neural MMO \cite{suarez2021neural} & Continuous & Discrete & Yes & 256 and 1024 \\ \hline
PettingZoo Atari \cite{terry2021pettingzoo} & Continuous+Image & Discrete & No & 1v1 \\ \hline
PommerMan \cite{resnick2018pommerman} & Discrete & Discrete & No & 2v2 \\ \hline
FightLadder (Ours) & Continuous+Image & Discrete & Yes & 1v1 and team mode \\ \hline
\end{tabular}
}
\label{tab:environment_comparison}
\end{table}

\section{Baseline Algorithms of \benchname{}-Baselines} \label{appx:baselines}

\paragraph{Independent Learning (IPPO).} 
Independent learning is a straightforward extension of single-agent RL into MARL. It decomposes the joint optimization into individual ones for each agent while regarding all other agents as part of the environment. It can be implemented easily by simultaneously running single-agent RL algorithms for each player.
Theoretically, this independent learning paradigm suffers from suboptimality~\cite{tan1993multi, foerster2018counterfactual}, because the environment becomes non-stationary while other agents are updating their policies.  
However, recent work~\cite{de2020independent, yu2022surprising} finds that with modest hyperparameter tuning, IPPO can serve as a strong baseline compared to other state-of-the-art algorithms in some cooperative MARL tasks. 

\paragraph{Two-timescale Learning (2Timescale).}
Two-timescale learning follows the independent learning paradigm, but requires two players to update gradients according to the two-timescale rule, i.e., one player uses a much smaller step size than the other one.
As a result of this modification, two-timescale learning enjoys some nice theoretical properties --- it is proven that under some mild assumptions, independent policy gradient algorithms satisfying two-timescale converge to a Nash equilibrium in two-player zero-sum stochastic games~\cite{daskalakis2020independent}.

\paragraph{Population-Based Methods.}

\begin{algorithm}[htbp]
\caption{Population-Based Methods for MGs}
\label{alg:population}
\begin{algorithmic}[1]
\STATE Initialize policies $\mu^0=\{\mu_h\}, \nu^0=\{\nu_h\}, h\in[H]$
\STATE Initialize policy sets: $\mu=\{\mu^0\}, \nu=\{\nu^0\}$
\STATE Initialize meta-strategies: $\rho_\mu=[1.], \rho_\nu=[1.]$

\FOR{$t=1,\ldots,T$}
    \IF{$t\%2==0$}
     \STATE  $\nu^t = \textsc{Best\_Response}(\rho_\mu, \mu)$
     \STATE $\nu=\nu\bigcup\{\nu^t\}$
     \STATE Update $\rho_\nu$ according to specific algorithms
    \ELSE
    \STATE  $\mu^t = \textsc{Best\_Response}(\rho_\nu, \nu)$
    \STATE $\mu=\mu\bigcup\{\mu^t\}$
    \STATE Update $\rho_\mu$ according to specific algorithms
    \ENDIF
\ENDFOR
\STATE Return $\mu, \rho_\mu, \nu, \rho_\nu$
\end{algorithmic}
\end{algorithm}

The independent learning framework is only training agents against the current version of their opponents, which may fail or converge slowly due to the lack of diversity~\cite{dresher2016advances}. Population-based methods are proposed to increase policy diversity by maintaining a pool of policies in previous iterations, and using them as a curriculum to update the current policy. More specifically, for $t$-th update, the agent $\mu^t$ plays with previous versions of its opponent $\Tilde{\nu}$ sampled from the meta-strategy $\rho_\nu$, which is a distribution over $\nu^0,\nu^1,\dots,\nu^{t-1}$. Algorithm~\ref{alg:population} presents the pseudo-code for general population-based methods. With different choices of sampling distribution, we can recover several state-of-the-art baselines:
\begin{itemize}
    \item \textbf{Fictitious Self-Play (FSP)}~\cite{pmlr-v37-heinrich15}, where $\rho_\nu$ is the uniform distribution $\text{Uniform}(\nu^0,\nu^1,\dots,\nu^{t-1})$.
    \item \textbf{Policy-Space Response Oracles (PSRO),} where $(\Tilde{\mu}, \Tilde{\nu})$ are sampled from the meta-strategy $(\rho_\mu, \rho_\nu)$ by solving Nash equilibrium of the payoff matrix game between $\mu^0,\mu^1,\dots,\mu^{t-1}$ and $\nu^0,\nu^1,\dots,\nu^{t-1}$~\cite{lanctot2017unified}.
    \item \textbf{League Training (League),} where three types of agents --- main agents, league exploiters, and main exploiters, are introduced into the population. Main agents train against themselves as well as all previous versions of agents in the population; league exploiters train against all previous agents; and main exploiters optimize the best response of main agents. Each type of agent adopts a different sampling distribution which is a mixture of self-play and prioritized fictitious self-play. We refer readers to~\cite{vinyals2019grandmaster} for more implementation details.
\end{itemize}

\subsection{Training Details}
\label{subsec:app_train_detail}
\begin{figure}[!h]
    \centering
    \includegraphics[width=0.49\textwidth]{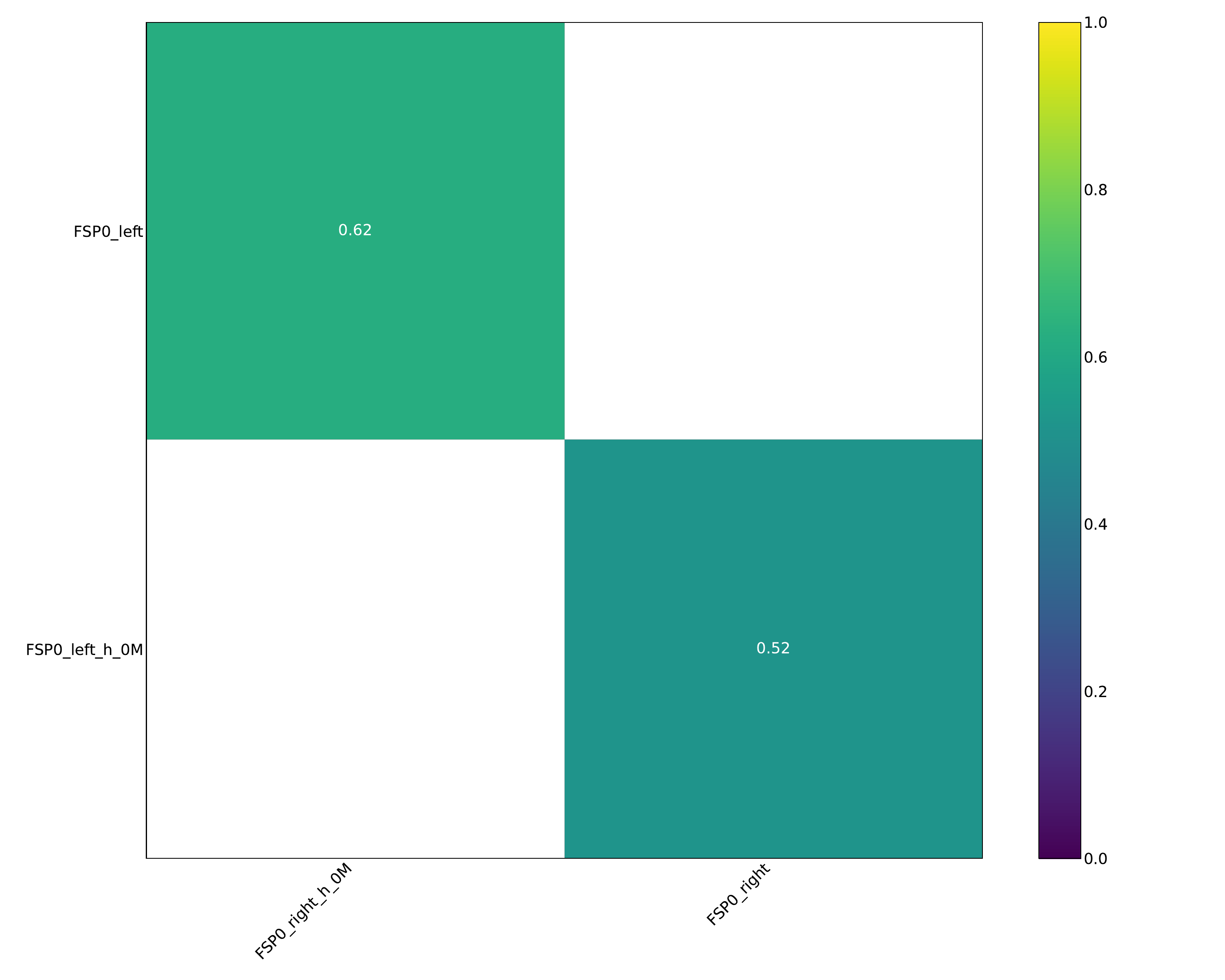}
    \includegraphics[width=0.49\textwidth]{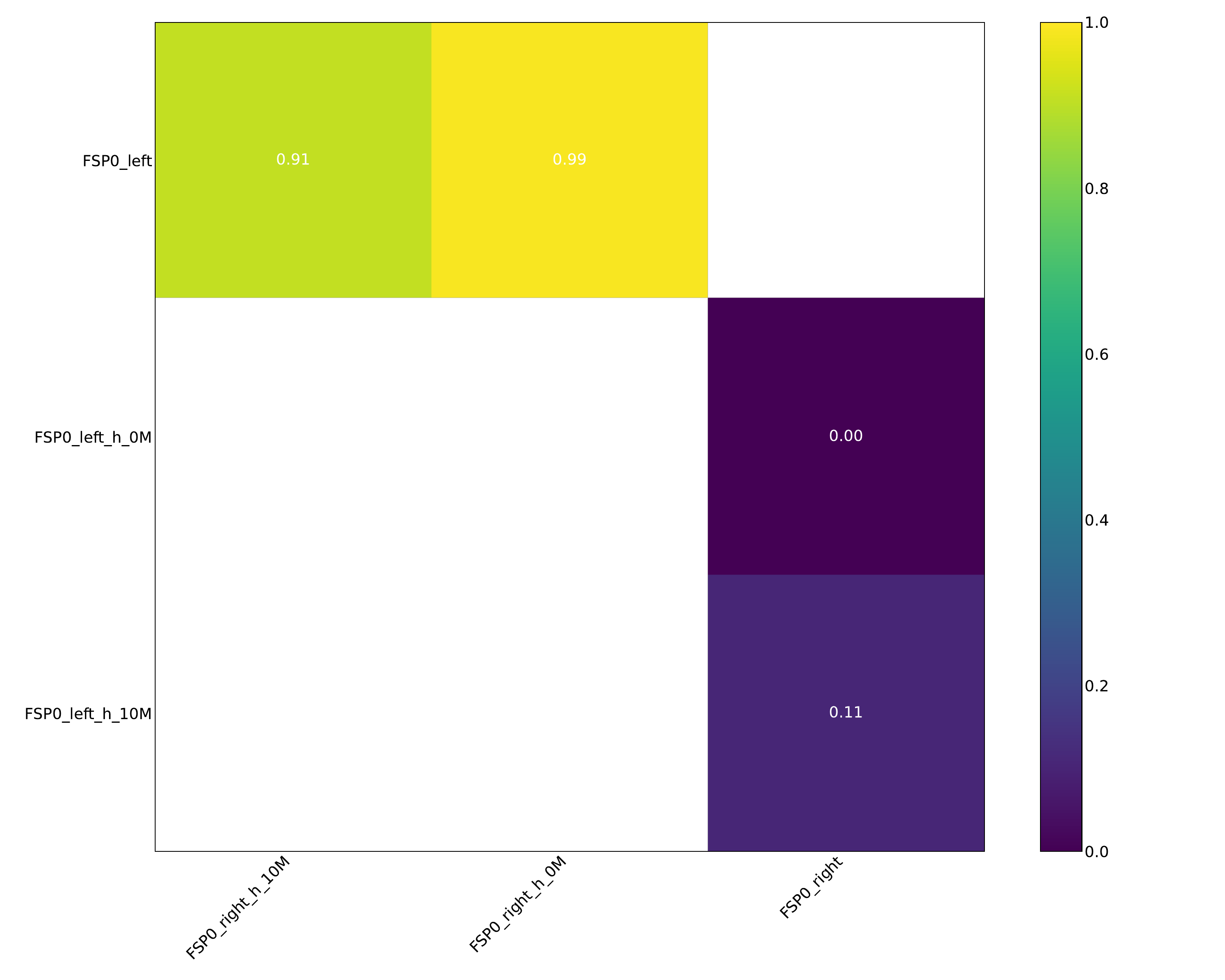}
    \includegraphics[width=0.49\textwidth]{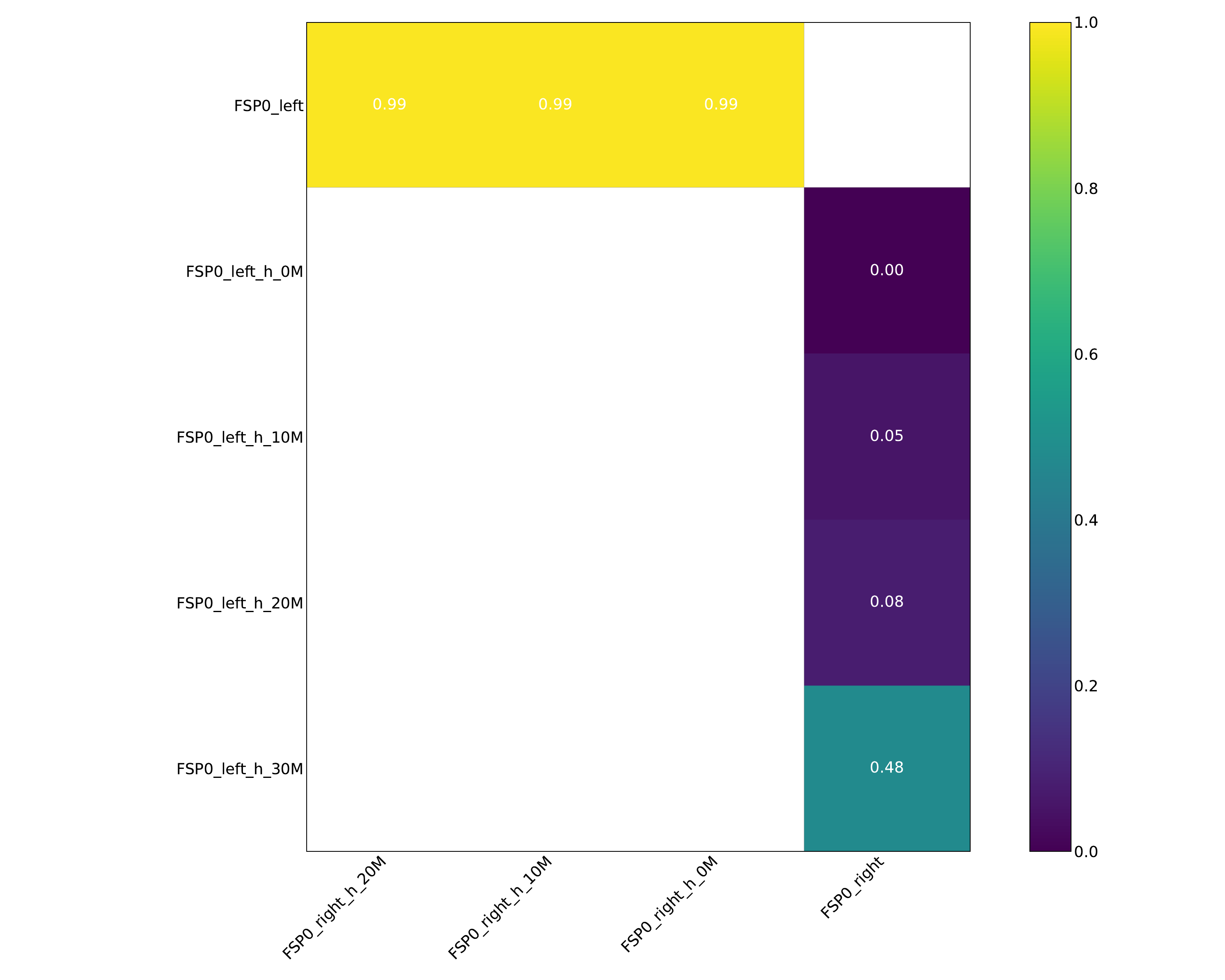}
    \includegraphics[width=0.49\textwidth]{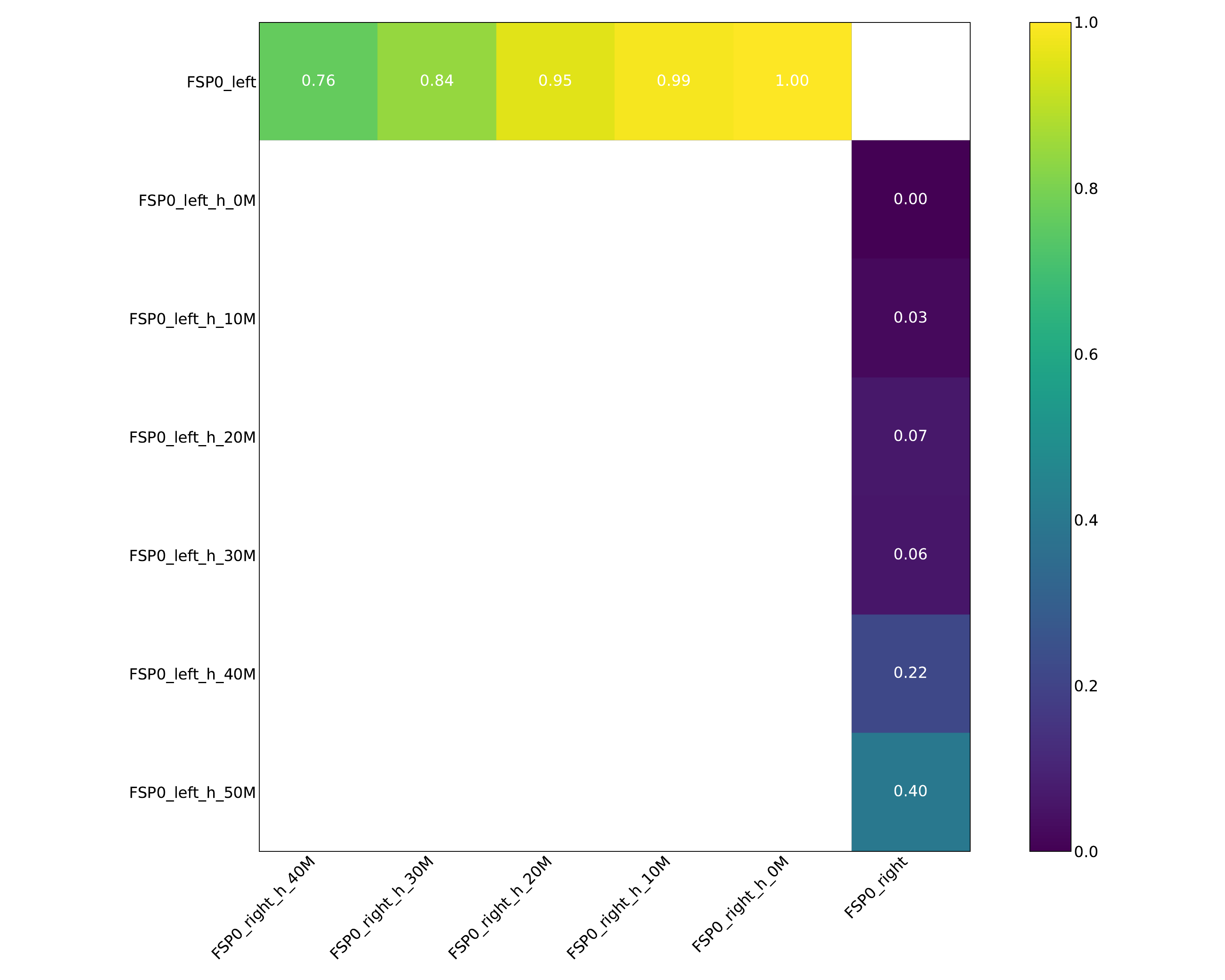}
    \includegraphics[width=0.49\textwidth]{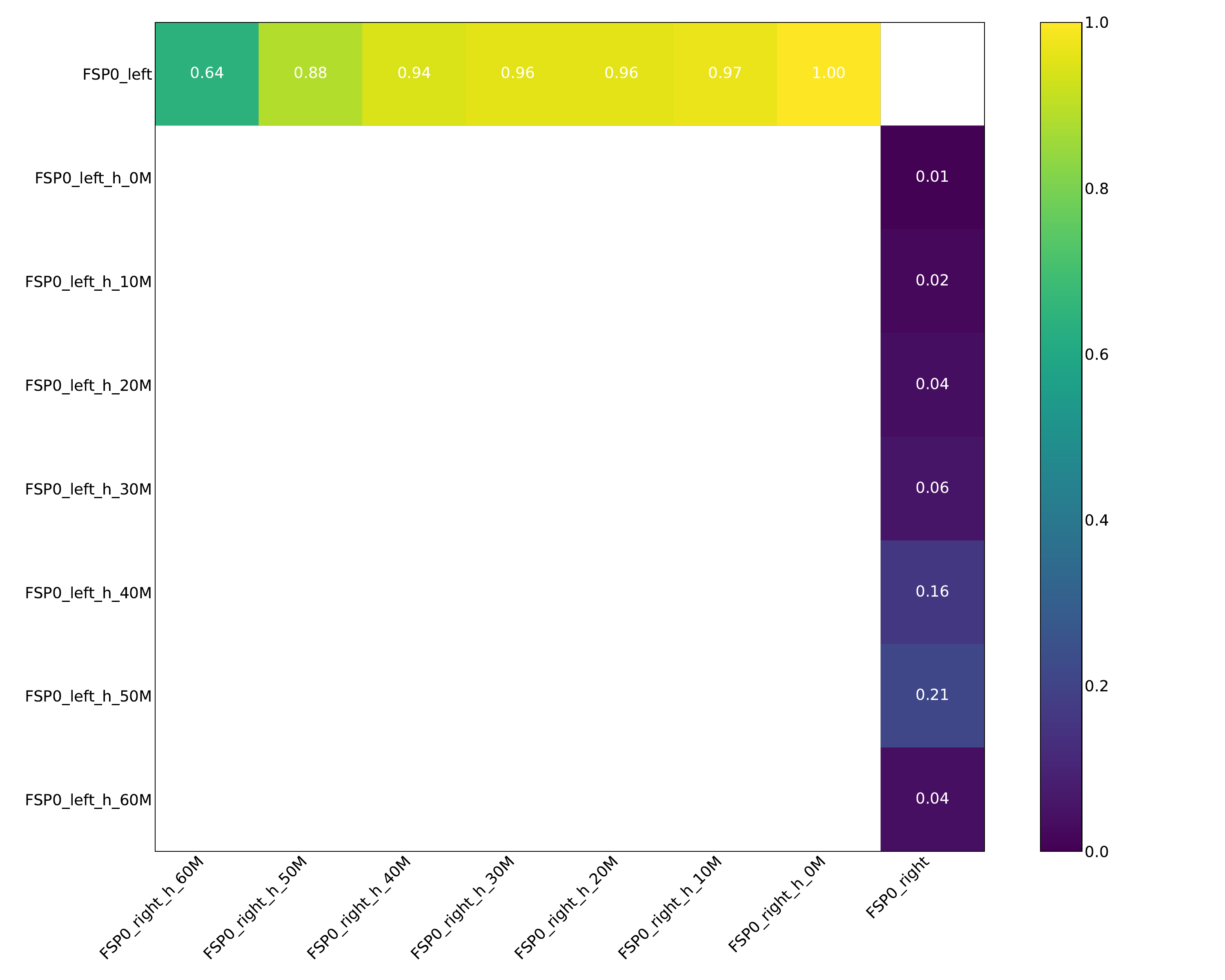}
    \includegraphics[width=0.49\textwidth]{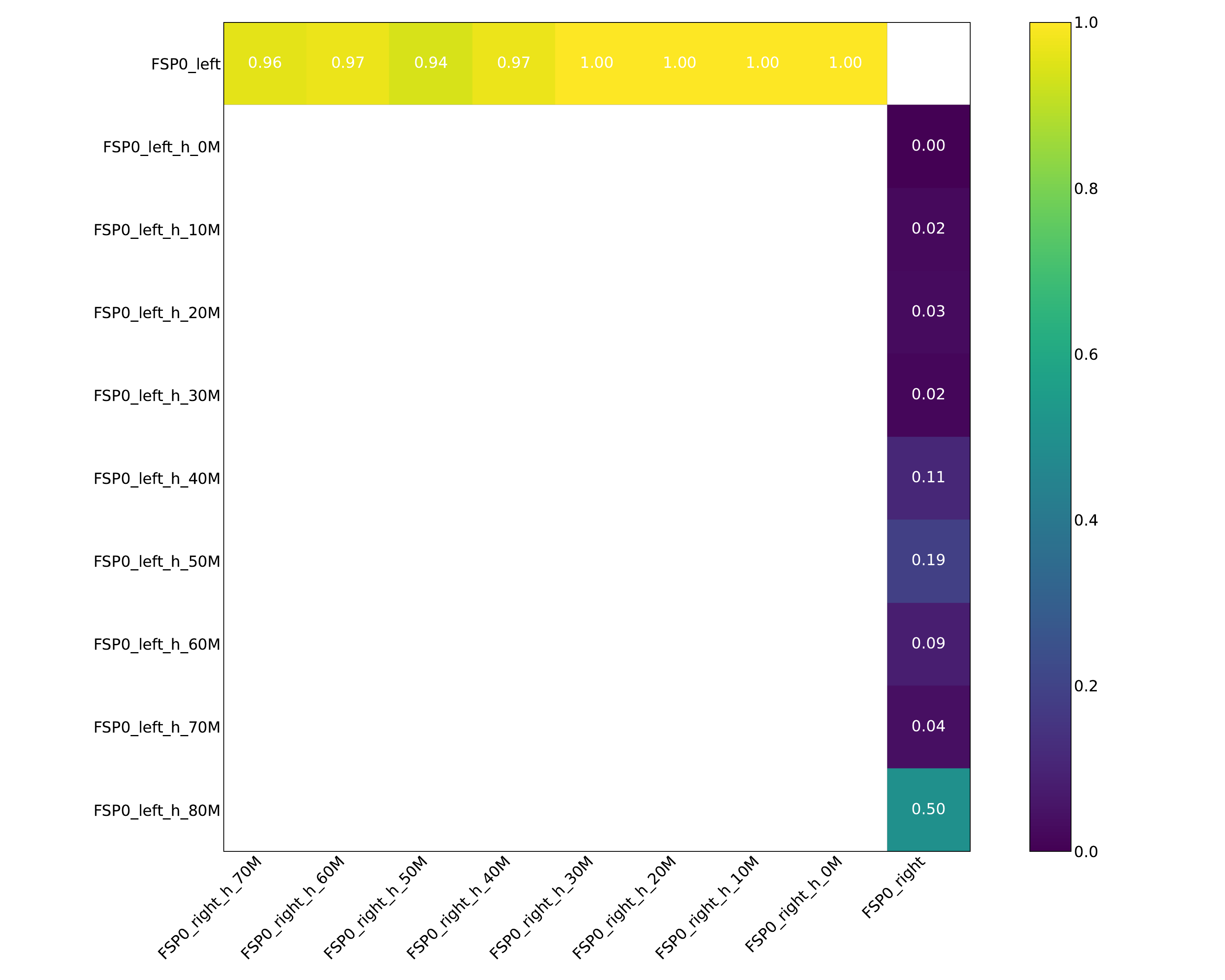}
    \caption{FSP details (training order from top left to bottom right): For FSP, there is one agent for each side (left or right). The name of each row indicates the agent information as \texttt{Character\_Side\_Checkpoint}. \texttt{Checkpoint=h\_xM} represents a previous version of agent saved at \texttt{x} million steps. The value indicates the win rate of the left (row) player against the right (column) player.}
    \label{fig:fsp_payoff}
\end{figure}

\begin{figure}[!h]
    \centering
    \includegraphics[width=0.49\textwidth]{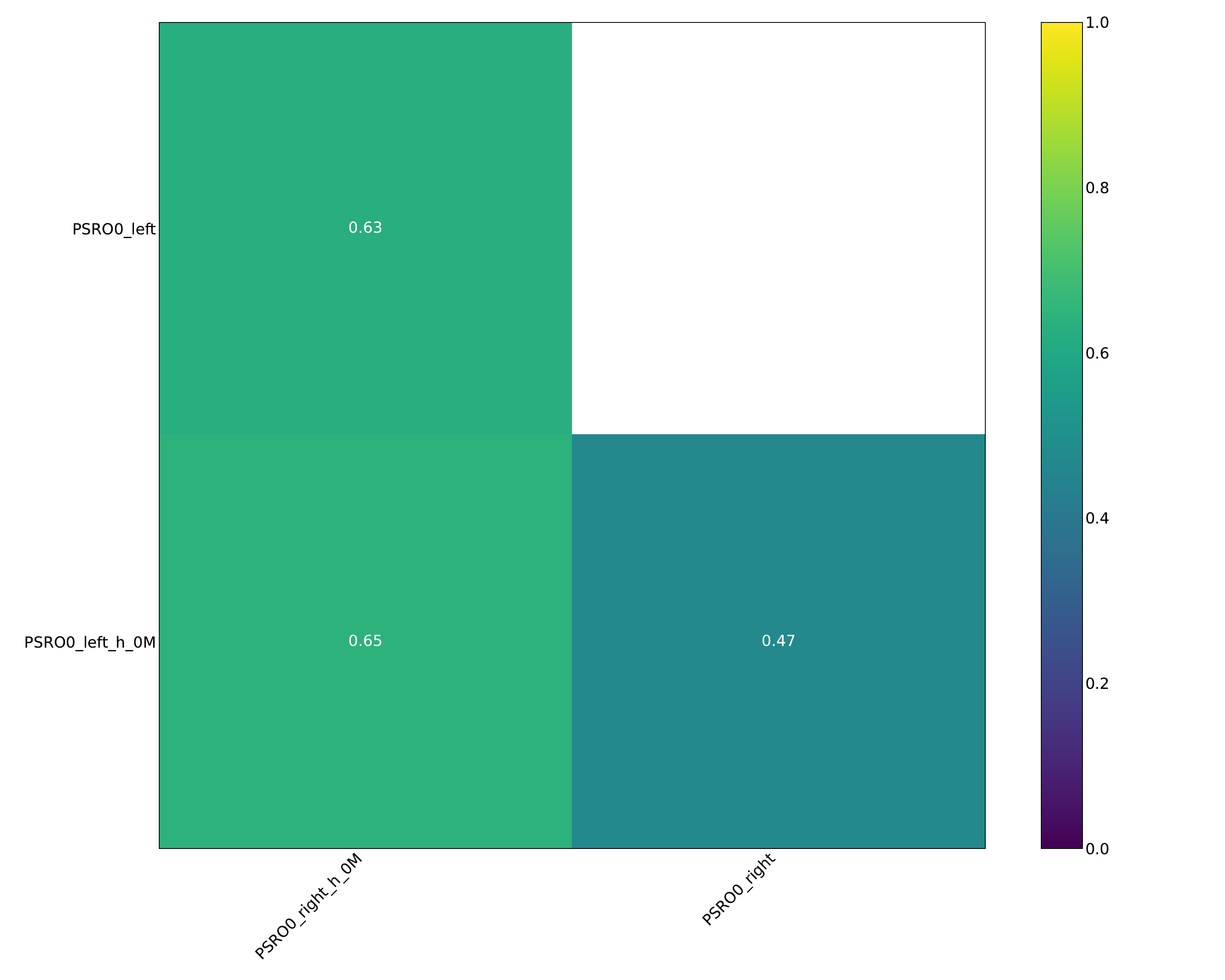}
    \includegraphics[width=0.49\textwidth]{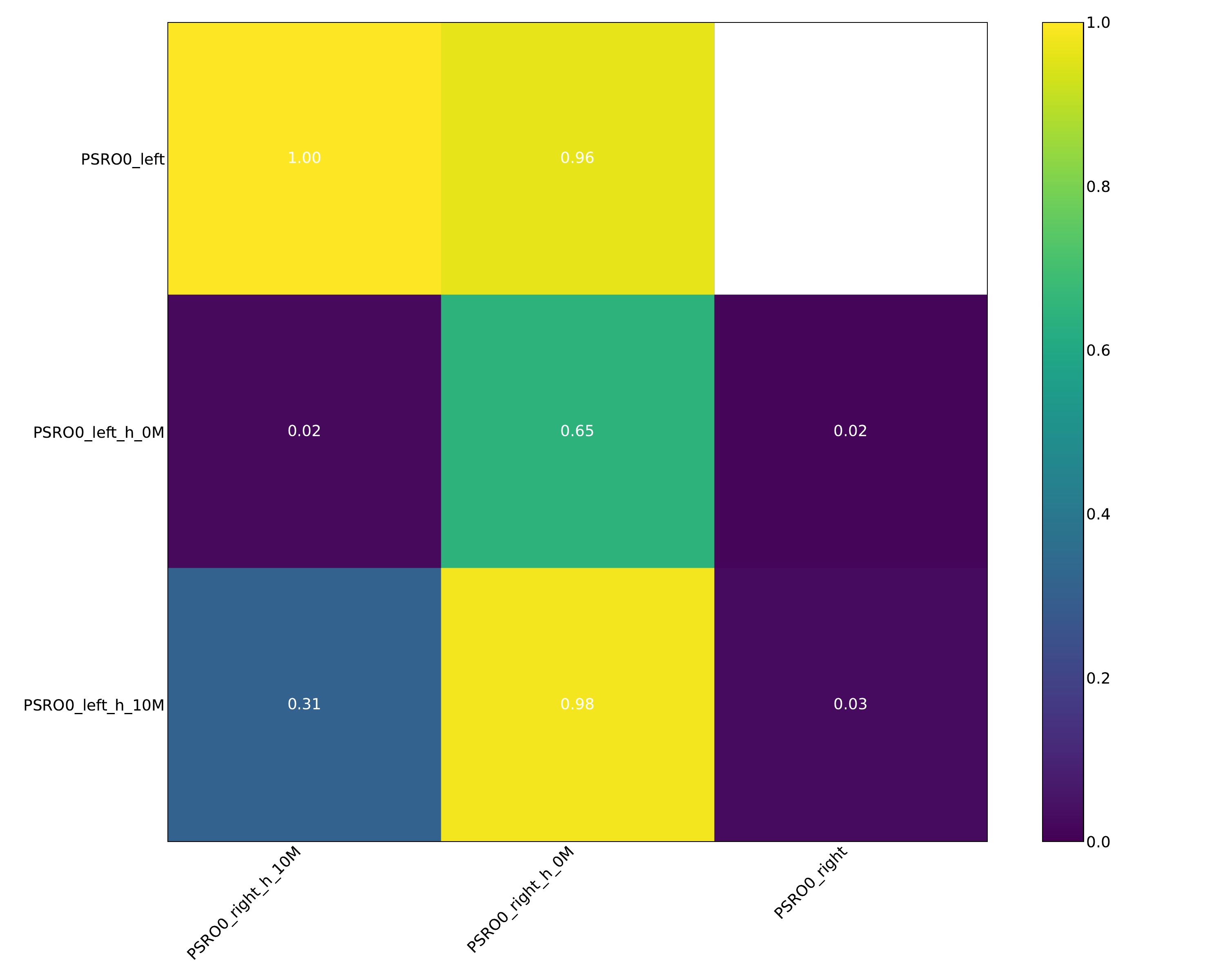}
    \includegraphics[width=0.49\textwidth]{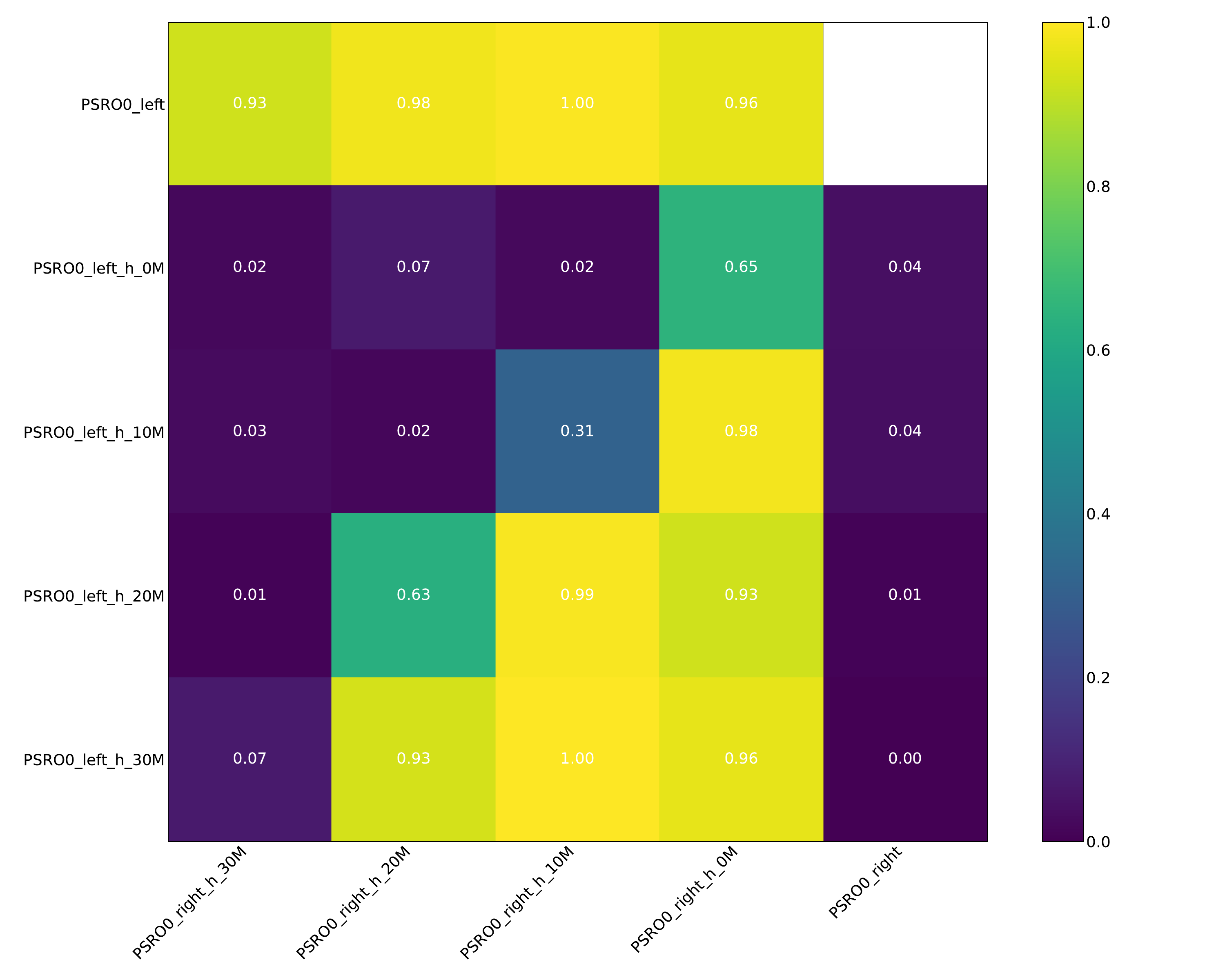}
    \includegraphics[width=0.49\textwidth]{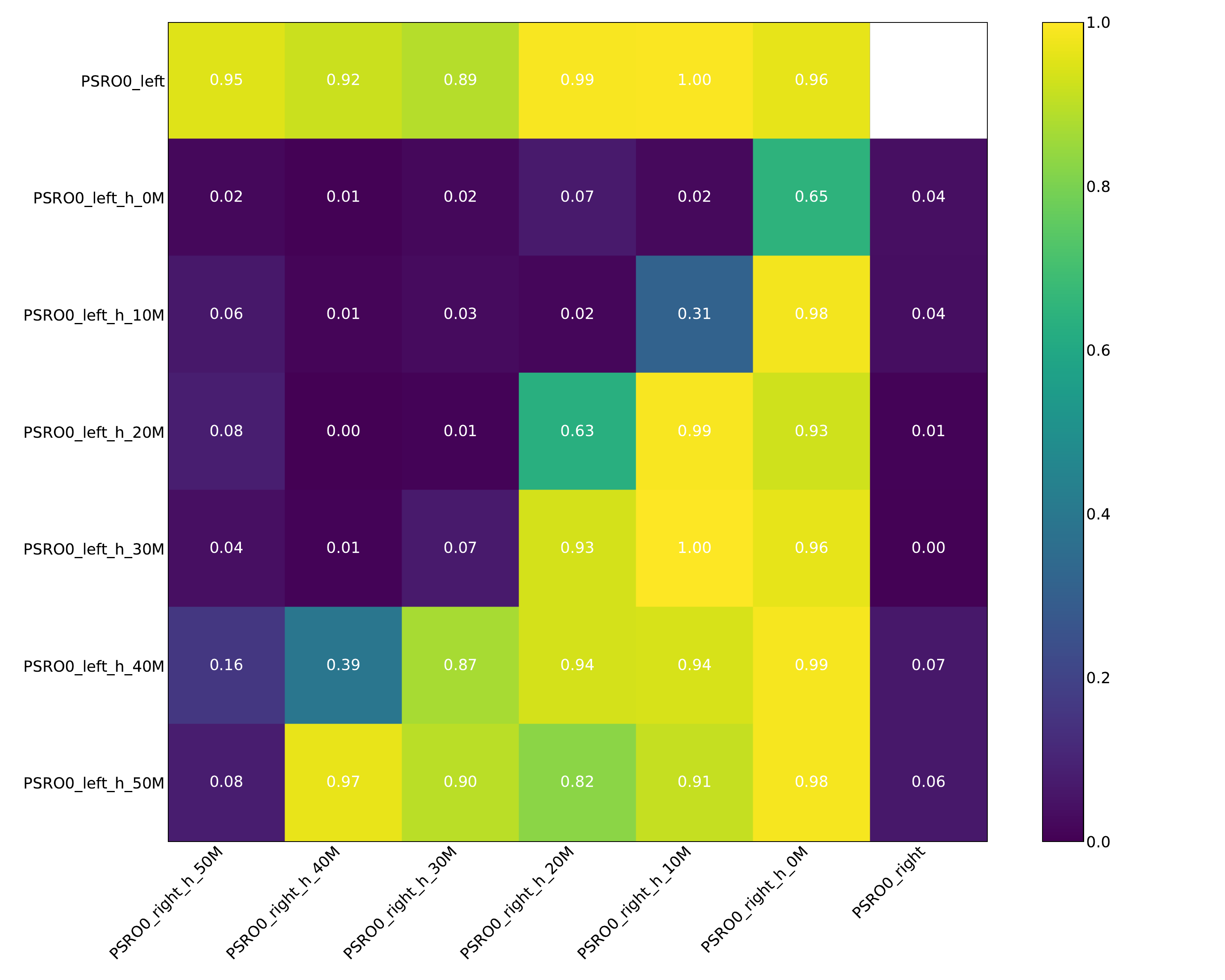}
    \includegraphics[width=0.49\textwidth]{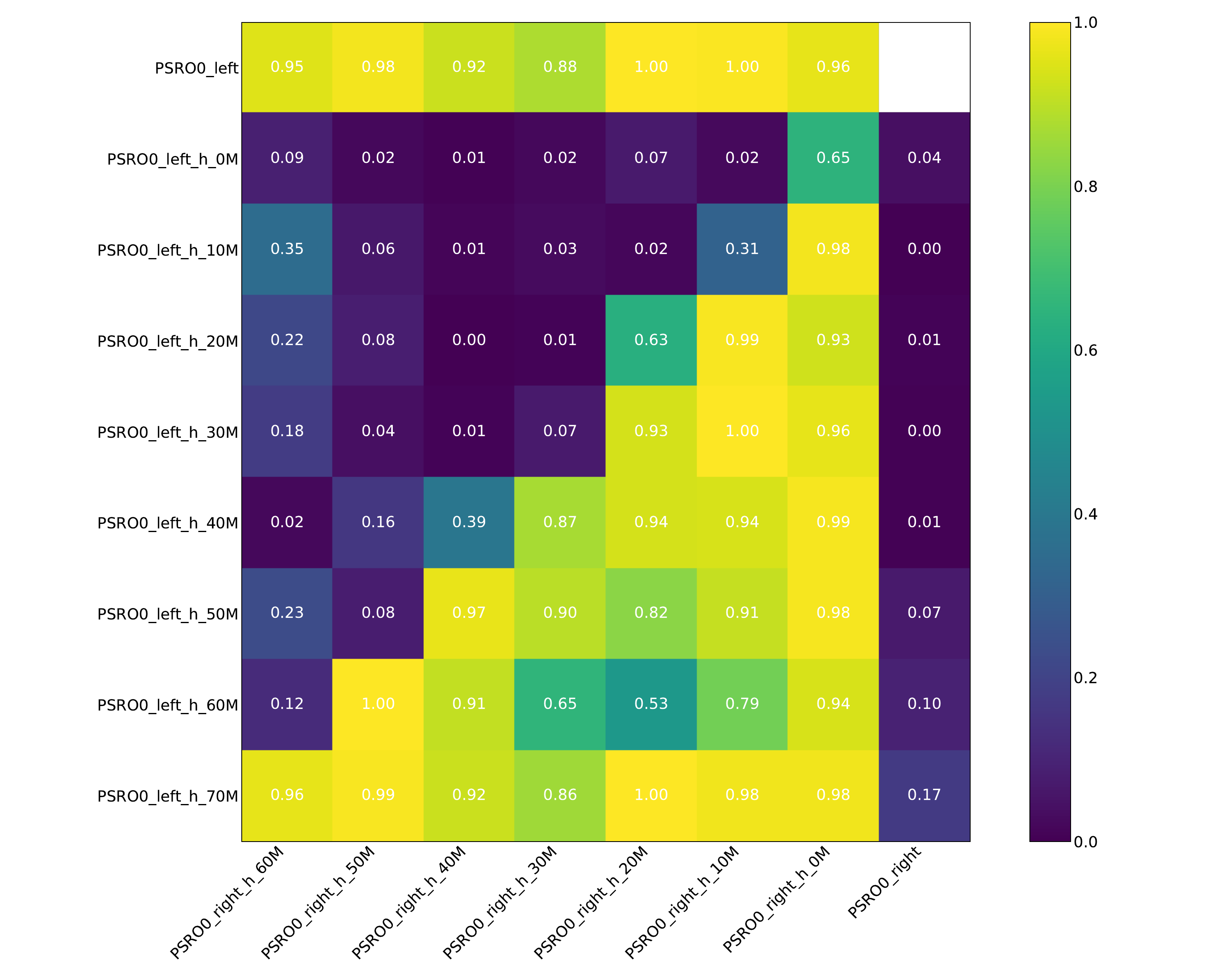}
    \includegraphics[width=0.49\textwidth]{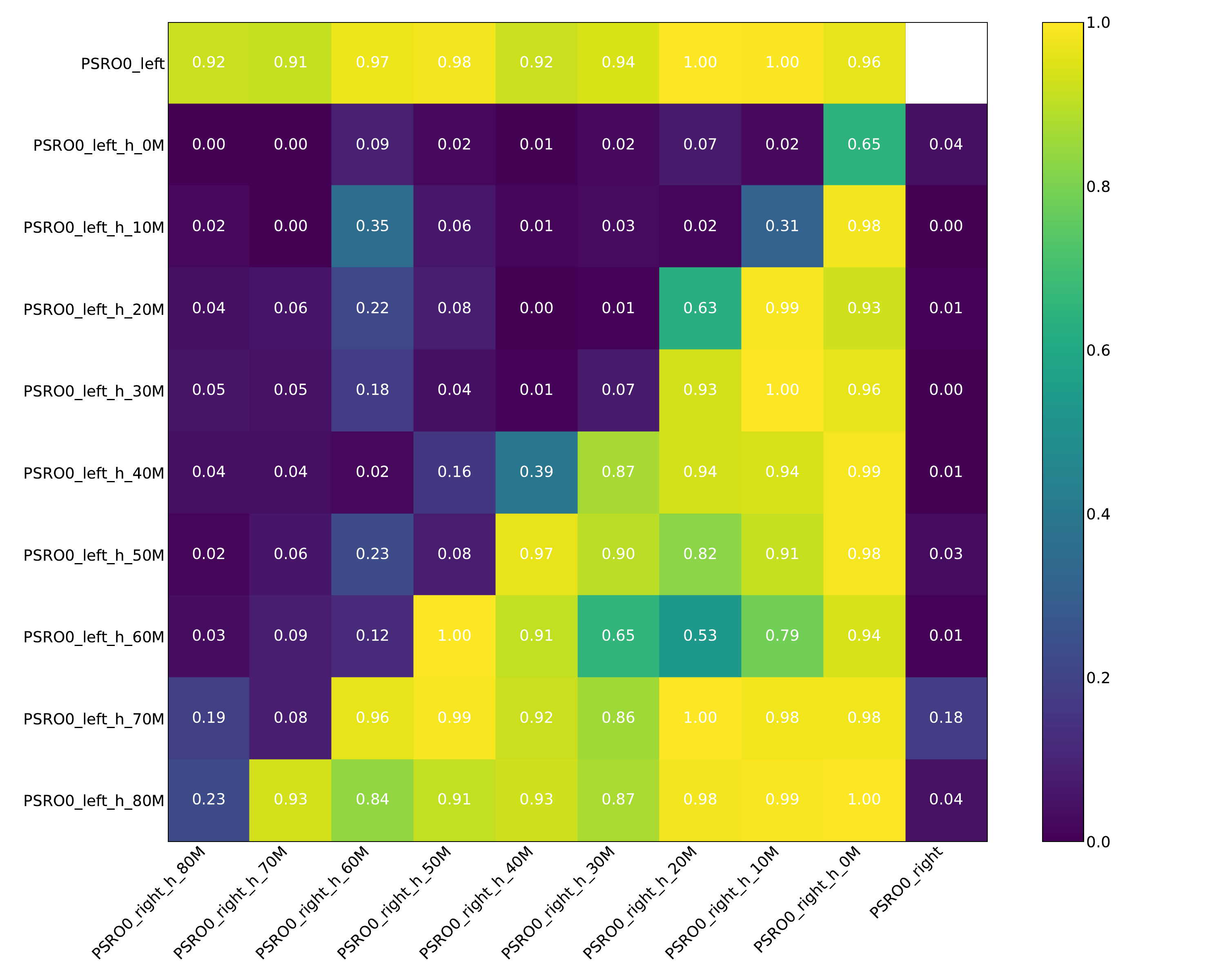}
    \caption{PSRO details (training order from top left to bottom right): For PSRO, there is one agent for each side (left or right). The name of each row indicates the agent information as \texttt{Character\_Side\_Checkpoint}. \texttt{Checkpoint=h\_xM} represents a previous version of agent saved at \texttt{x} million steps. The value indicates the win rate of the left (row) player against the right (column) player.}
    \label{fig:psro_payoff}
\end{figure}

\begin{figure}[!h]
    \centering
    \includegraphics[width=0.49\textwidth]{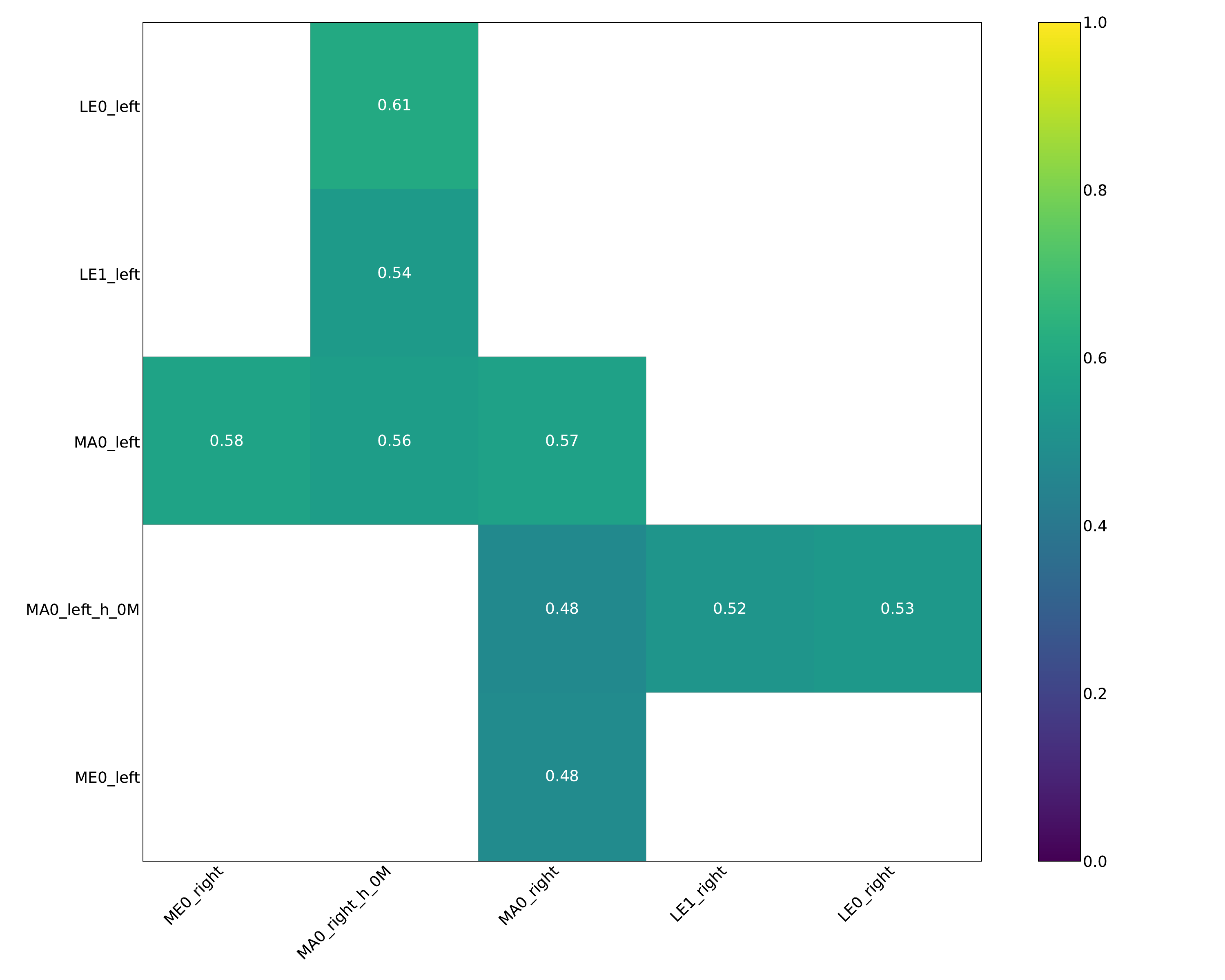}
    \includegraphics[width=0.49\textwidth]{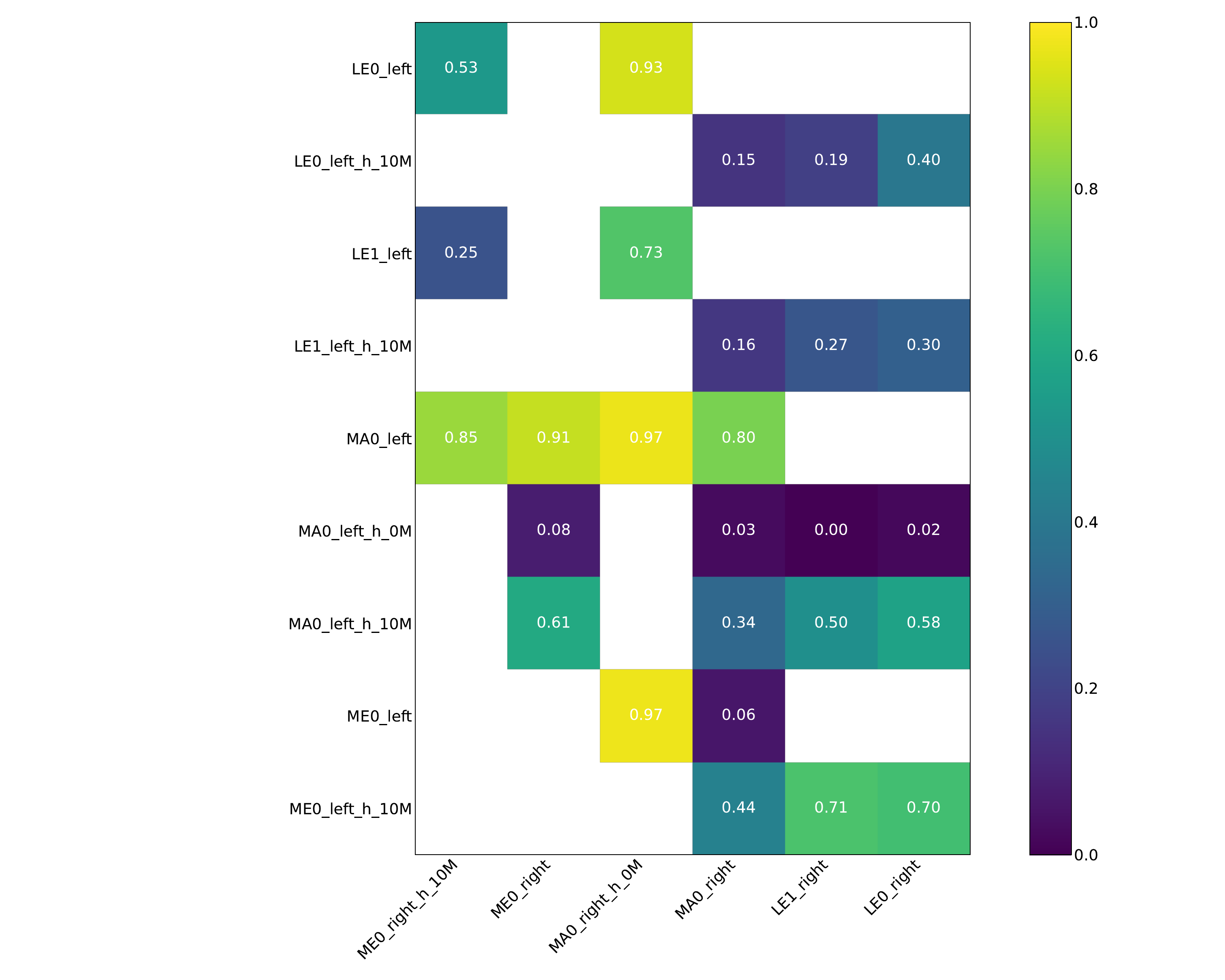}
    \includegraphics[width=0.49\textwidth]{part1/research2/figures/payoff/payoff_lg_3.pdf}
    \includegraphics[width=0.49\textwidth]{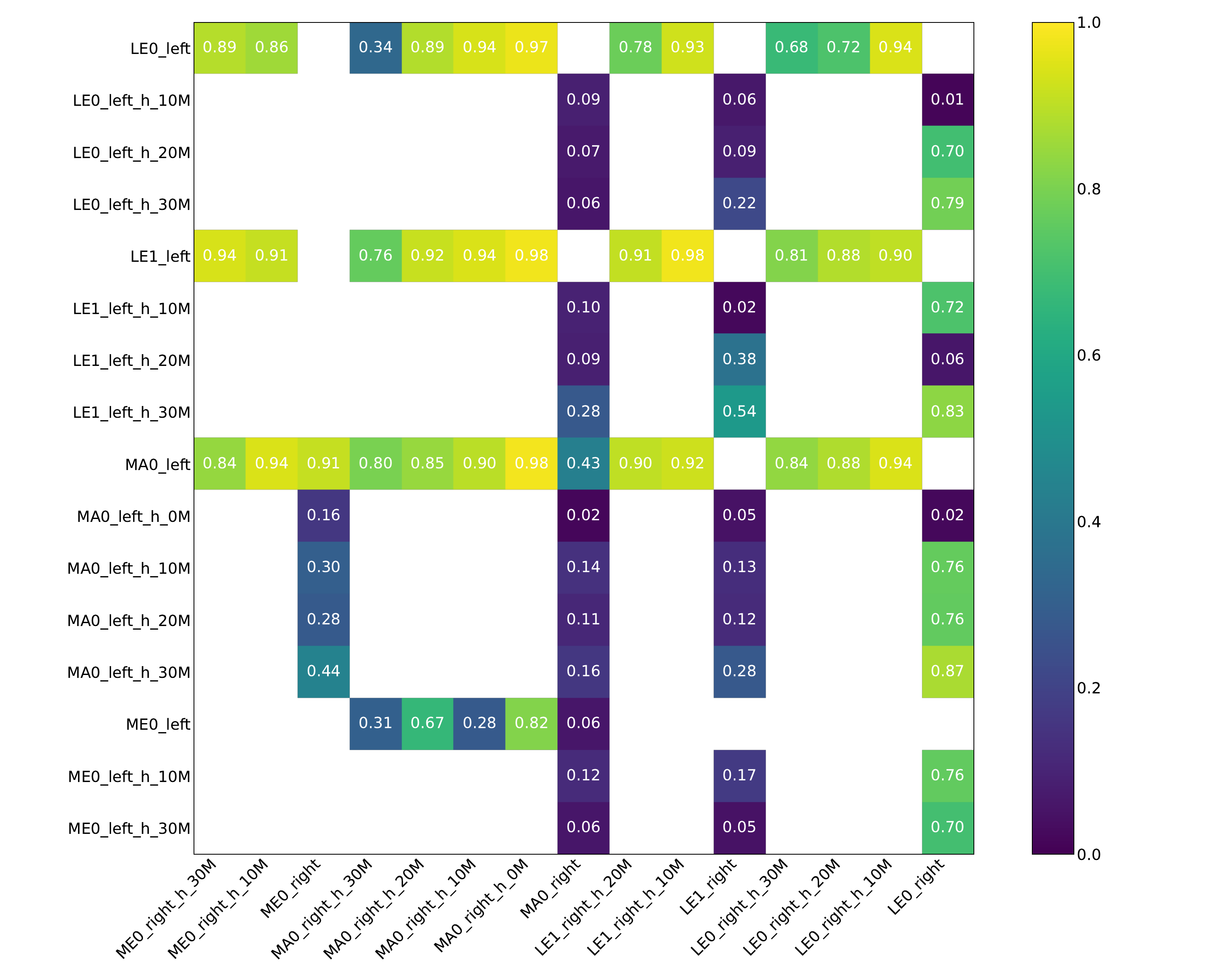}
    \includegraphics[width=0.49\textwidth]{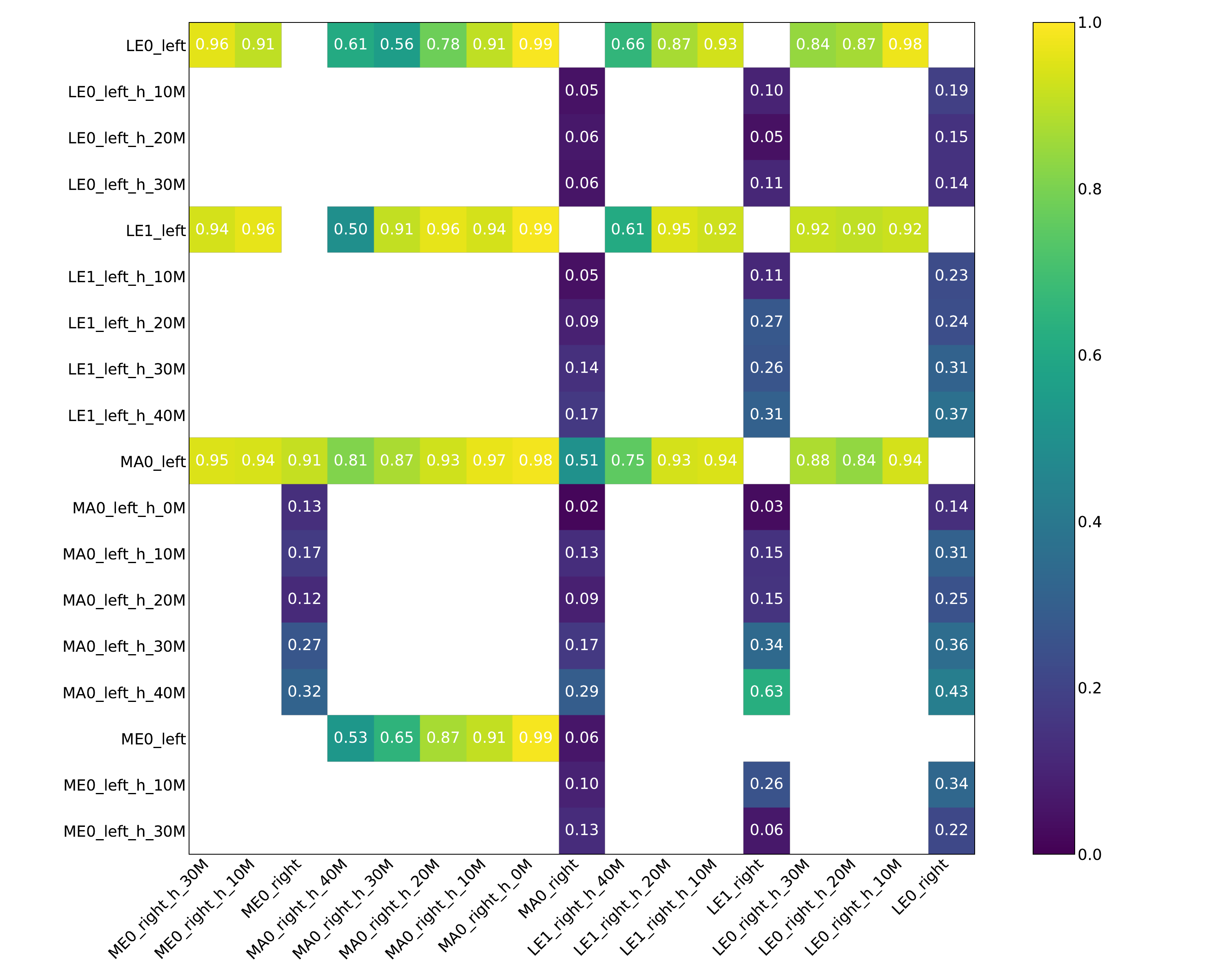}
    \includegraphics[width=0.49\textwidth]{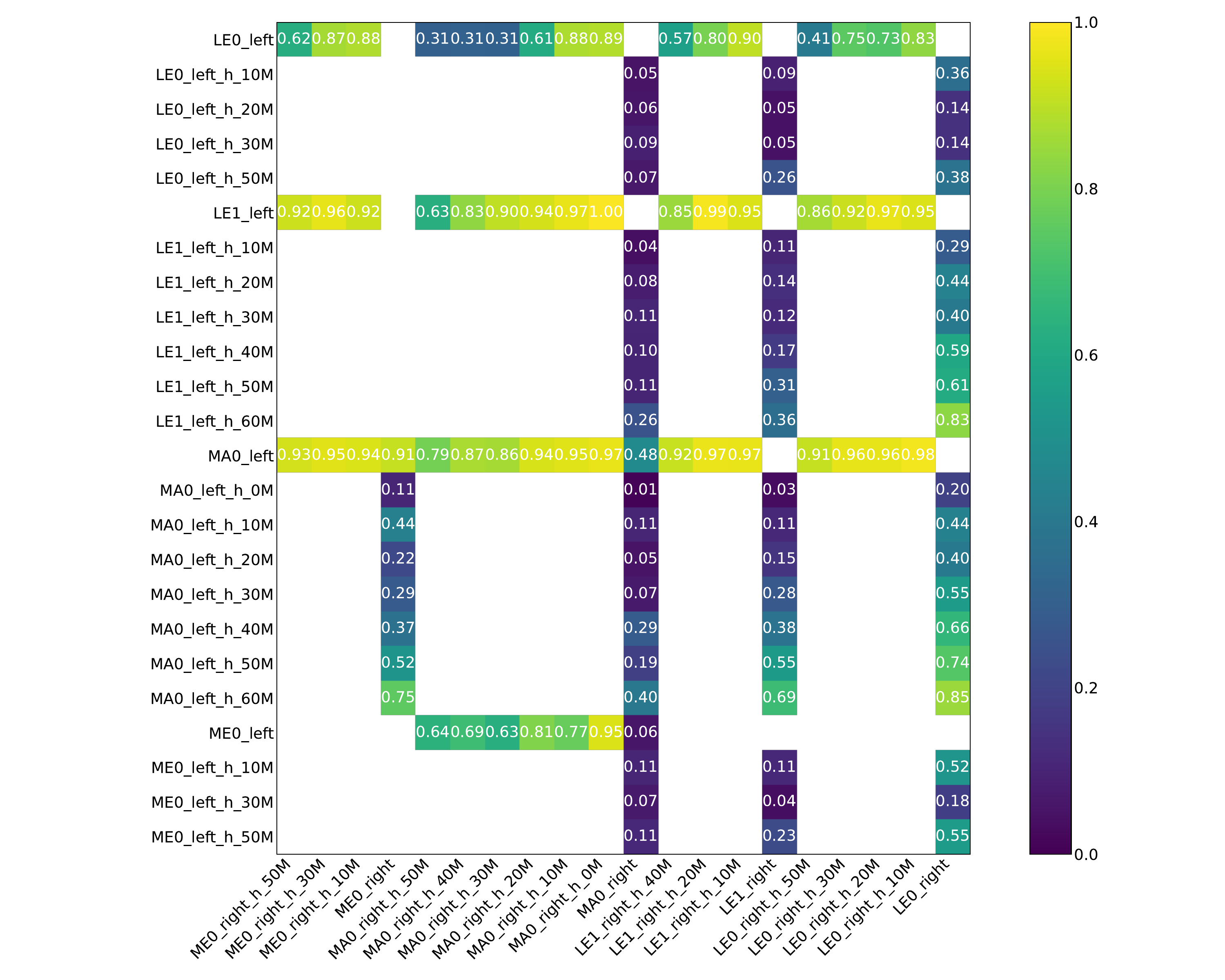}
    \caption{League training details (training order from top left to bottom right): For league training, there is one main agent (MA), two league exploiters (LE0, LE1), and one main exploiter (ME) for each side (left or right). The name of each row indicates the agent information as \texttt{Character\_Side\_Checkpoint}. \texttt{Checkpoint=h\_xM} represents a previous version of agent saved at \texttt{x} million steps. The value indicates the win rate of the left (row) player against the right (column) player.}
    \label{fig:league_payoff}
\end{figure}

Figure~\ref{fig:fsp_payoff}, \ref{fig:psro_payoff}, and \ref{fig:league_payoff} report the payoff matrix of policies within the population for FSP, PSRO, and League, respectively, with the value representing the win rate of the left player against the right player. We trained all our agents on one server with 192 CPUs and 8 A6000 GPUs.

\section{Individual Elo Results} \label{appx:indiv_elo}

\begin{figure}[!h]
    \centering
    \includegraphics[width=\textwidth]{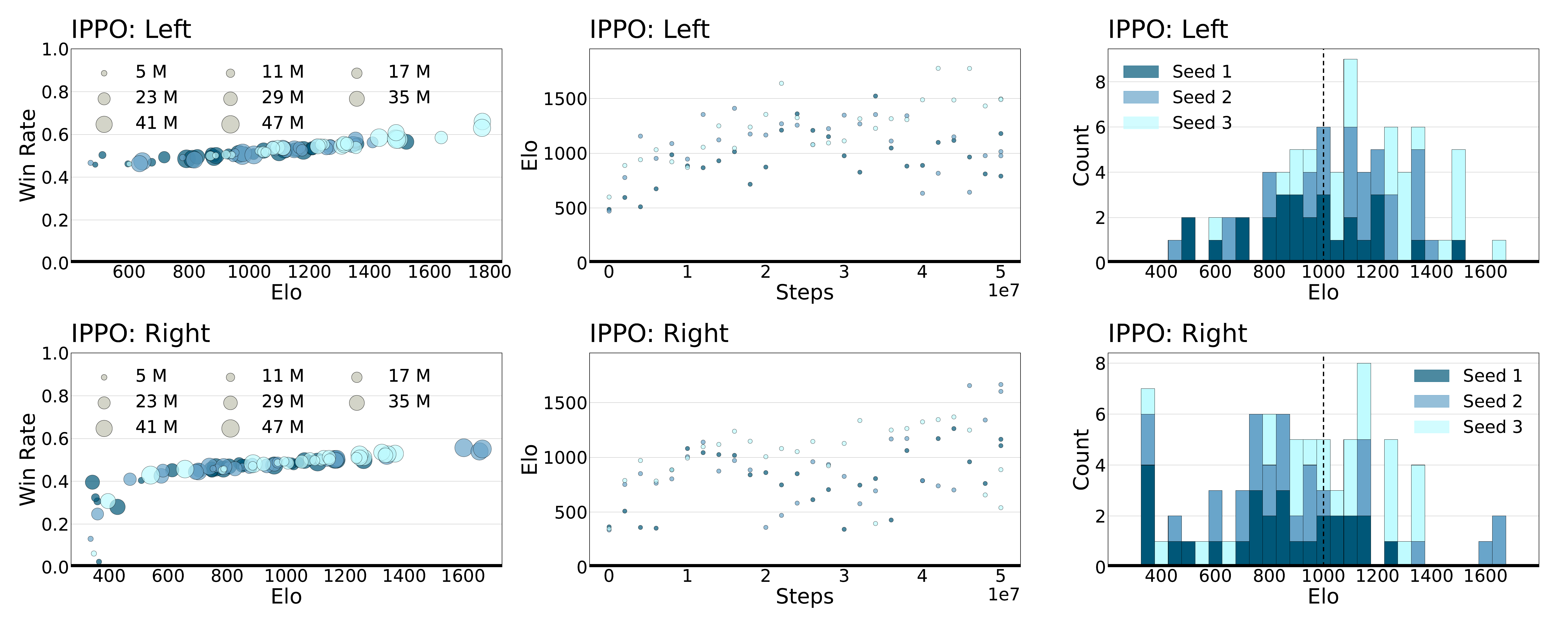}
    \caption{The Elo rating for the population of agents trained with IPPO algorithm. The upper three plots are for left-side player and the bottom three are for the right-side player. The Elo rating is plotted against the winning rate over matched policies (left figures), training steps (middle figures) and the number of policies (right figures).}
    \label{fig:ippo_elo}
\end{figure}

\begin{figure}[!h]
    \centering
    \includegraphics[width=\textwidth]{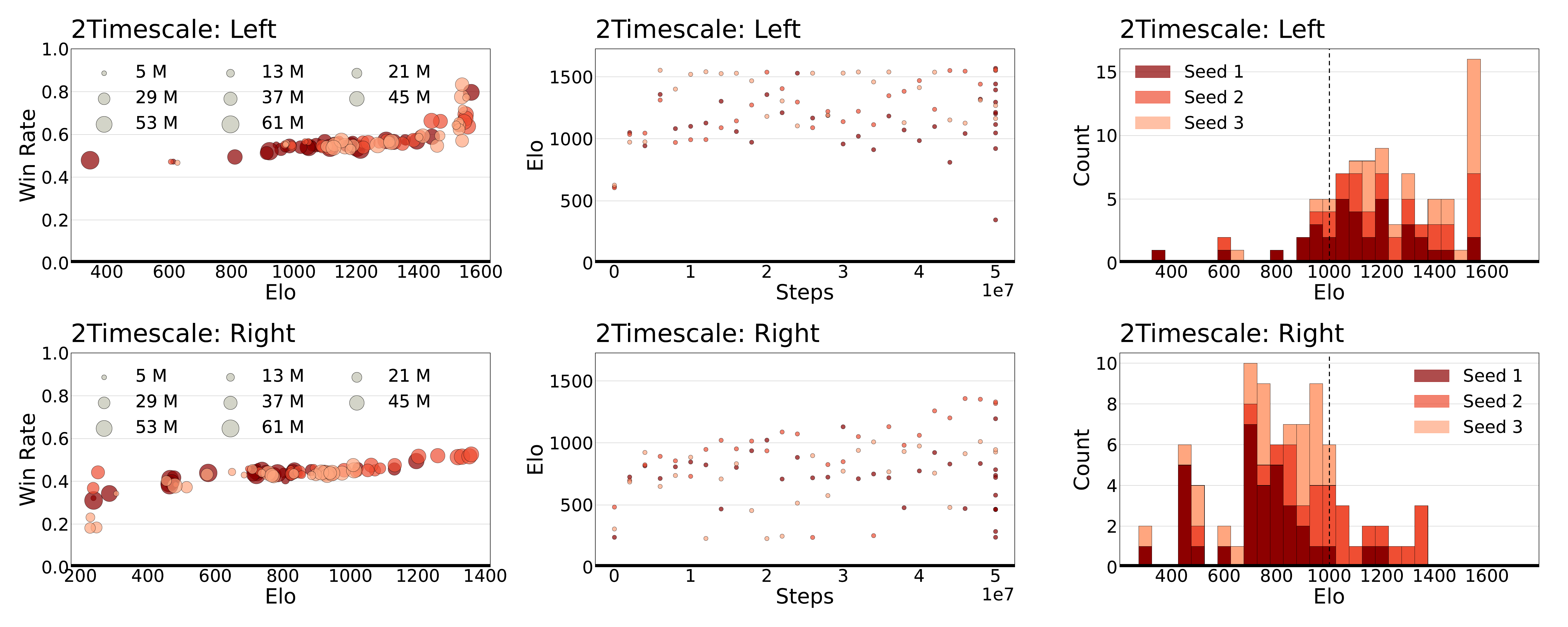}
    \caption{The Elo rating for the population of agents trained with 2Timescale algorithm. }
    \label{fig:2timescale_elo}
\end{figure}

\begin{figure}[!h]
    \centering
    \includegraphics[width=\textwidth]{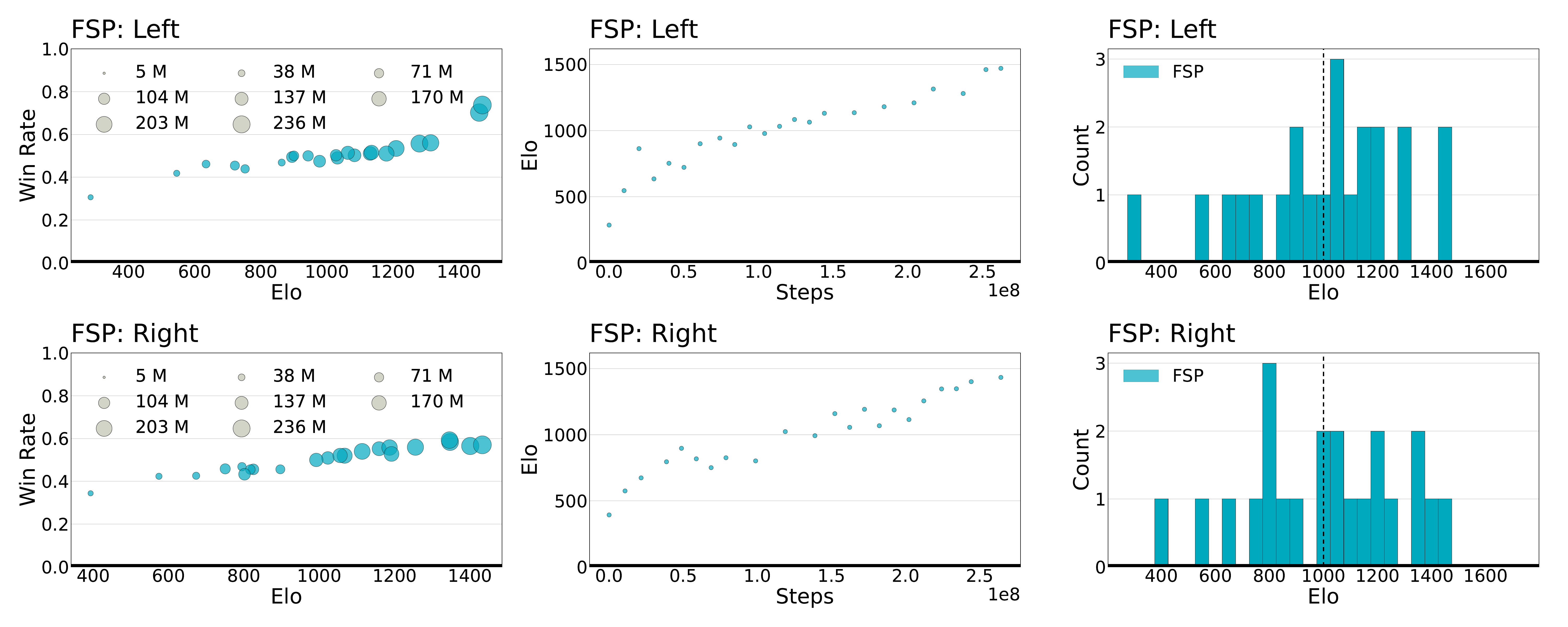}
    \caption{The Elo rating for the population of agents trained with FSP algorithm. }
    \label{fig:fsp_elo}
\end{figure}

\begin{figure}[!h]
    \centering
    \includegraphics[width=\textwidth]{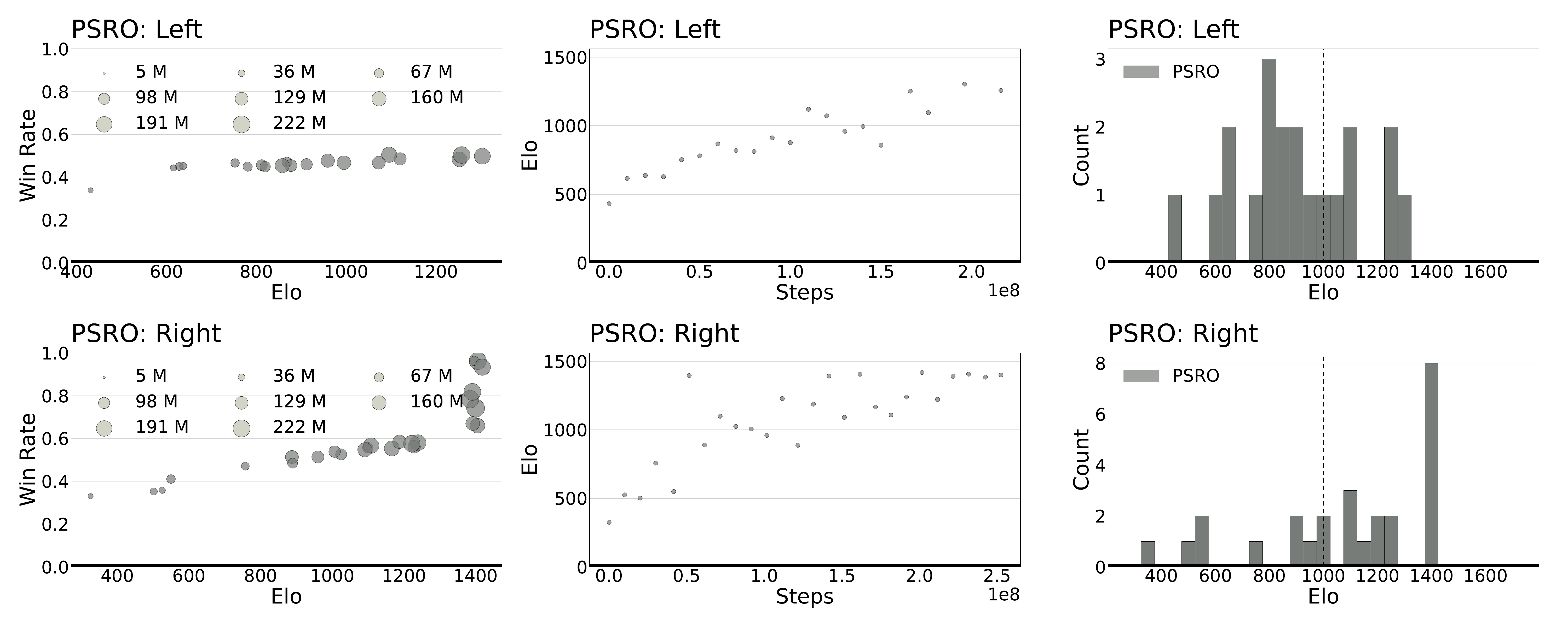}
    \caption{The Elo rating for the population of agents trained with PSRO algorithm. }
    \label{fig:psro_elo}
\end{figure}

\begin{figure}[!h]
    \centering
    \includegraphics[width=\textwidth]{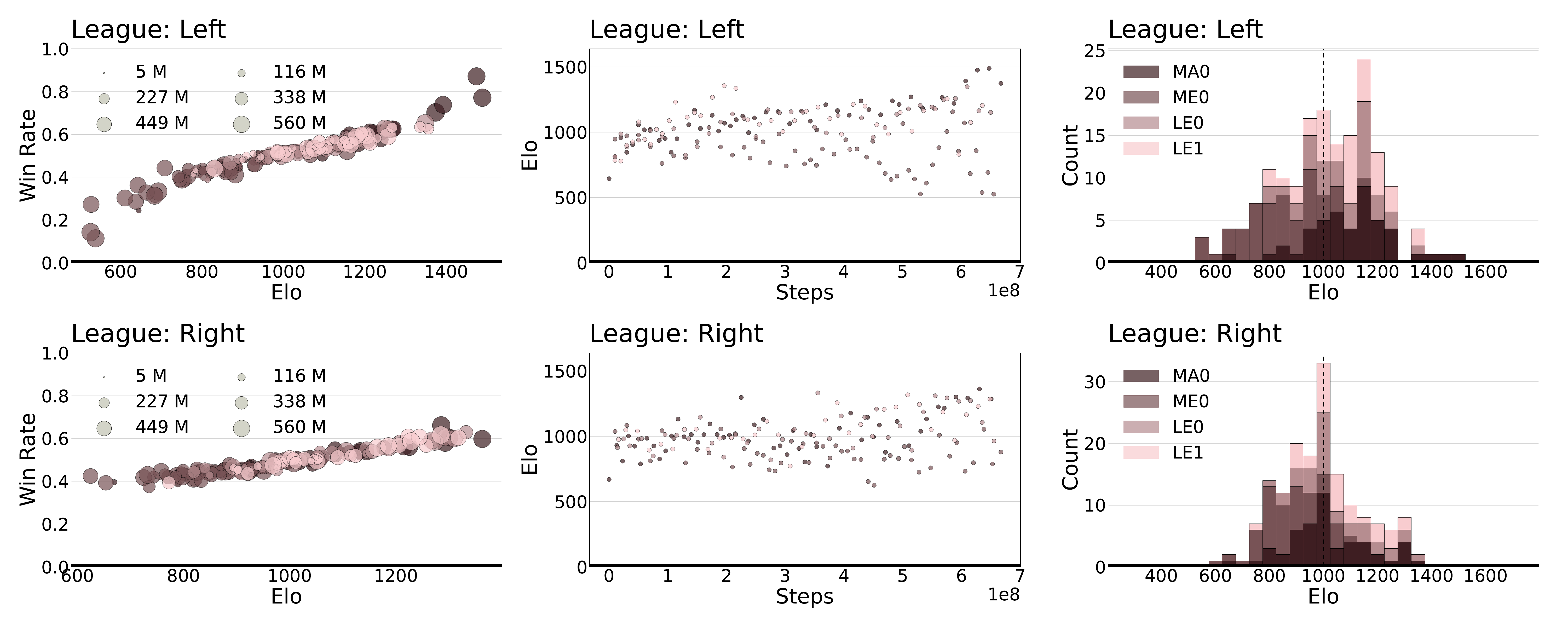}
    \caption{The Elo rating for the population of agents trained with League training. }
    \label{fig:league_elo}
\end{figure}

Figure~\ref{fig:ippo_elo} visualizes the Elo distribution of the IPPO population from three complementary views: matched winning rate, training progress, and policy index. 
Figure~\ref{fig:2timescale_elo} reports the corresponding Elo evolution for the 2Timescale method.
Figure~\ref{fig:fsp_elo} shows the Elo profile of agents produced by FSP. Figure~\ref{fig:psro_elo} presents the Elo results for PSRO. Figure~\ref{fig:league_elo} summarizes the Elo behavior of the League-trained population.

\chapter{Multi-Player General-Sum Game}\label{chap:rl_general_sum}
\begin{center}
\begin{quote}
This section is based on paper ``\textit{Learning Distributed and Fair Policies for Network Load Balancing as Markov Potential Game}''~\cite{yao2022learning} written in collaboration with Zhiyuan Yao, previously published at NeurIPS 2022.
\end{quote}
\end{center}

\section{Introduction}
\label{sec:intro}

In cloud data centers (DCs) and distributed networking systems, servers are deployed on infrastructures with multiple processors to provide scalable services~\cite{dragoni2017microservices}.
To optimise workload distribution and reduce additional queuing delay, load balancers (LBs) play a significant role in such systems.
State-of-the-art network LBs rely on heuristic mechanisms~\cite{maglev, desmouceaux20186lb, incab2018} under the low-latency and high-throughput constraints of the data plane.
However, these heuristics are not adaptive to dynamic environments and require human interventions, which can lead to most painful mistakes in the cloud -- mis-configurations.
RL approaches have shown performance gains in distributed system and networking problems~\cite{auto2018sigcomm, decima2018, drl-udn-2019, sivakumar2019mvfst}, yet applying RL on the network load balancing problem is challenging.

First, unlike traditional workload distribution or task scheduling problem~\cite{auto2018sigcomm, decima2018}, network LBs have limited observations over the system, including task sizes and actual server load states.
Being aware of only the number of tasks they have distributed, servers can be overloaded by collided elephant tasks and have degraded quality of service (QoS).

Second, to guarantee high service availability in the cloud, multiple LBs are deployed in DCs. Network traffic is split among all LBs.
This multi-agent setup makes LBs have only partial observation over the system.

Third, modern DCs are based on heterogeneous hardware and elastic infrastructures~\cite{kumar2020fast}, where server capacities vary.
It is challenging to assign correct weights to servers according to their actual processing capacities, and this process conventionally requires human intervention -- which can lead to error-prone configurations~\cite{maglev,incab2018}.

\noindent\begin{minipage}{0.55\textwidth}
\begin{algorithm}[H]
	\footnotesize
	\caption{LB System Transition Protocol}\label{alg:model-transition}
	\begin{algorithmic}[1]
	    \STATE Initialise server load, $X_j(0) \gets 0, \forall j\in[N]$
		\FOR {each time step $t$}
		    \FOR {each LB agent $i\in[M]$}
		        \STATE Choose action $\alpha_{ij}(t)$ for coming tasks $w_i(t)$
		    \ENDFOR
		    \FOR {each server $j$}
		        \STATE Update workload:\\
		        $X_j(t)= X_j(t-1)+\sum_{i=1}^M w_i(t)\alpha_{ij}(t)-v_j(t-1)$
		    \ENDFOR
		\ENDFOR
	\end{algorithmic}
\end{algorithm}
\end{minipage}
\begin{minipage}{0.4\textwidth}
\vspace{.3cm}
    \centering
    \includegraphics[width=\columnwidth]{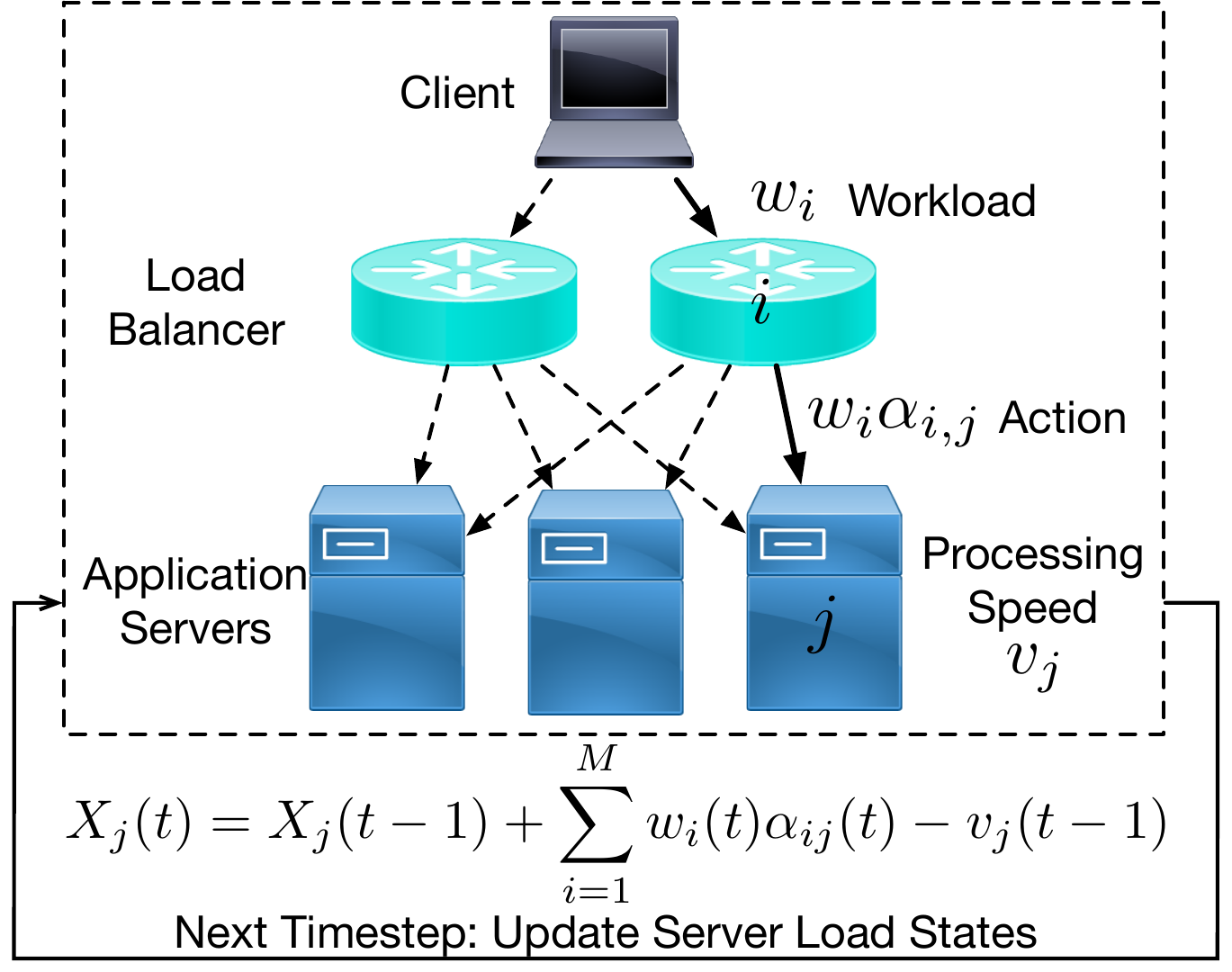}
	\captionof{figure}{Network load balancing.}
	\label{fig:intro}
\end{minipage}
\vspace{.3cm}

Last but not least, given the low-latency and high-throughput constraints in the distributed networking setup, the interactive training procedure of RL models and the centralised-training-decentralised-execution (CTDE) scheme~\cite{foerster2018counterfactual} can incur additional communication and management overhead.

In this paper, we study the network load balancing problem in multi-agent game theoretical approach, by formulating it as a Markov potential game through specifying the proper reward function, namely variance-based fairness.
We propose a distributed Multi-Agent RL (MARL) network load balancing mechanism that is able to exploit asynchronous actions based only on local observations and inferences.
Load balancing performance gains are evaluated based on both event-based simulations and real-world experiments\footnote{Source code and data of both simulation and real-world experiment are open-sourced at \href{https://github.com/ZhiyuanYaoJ/MARLLB}{https://github.com/ZhiyuanYaoJ/MARLLB}.}.




\section{Related Work}
\label{sec:background}

\begin{figure}[tbp]
	\centering
	\begin{subfigure}{0.47\columnwidth}
		\centering
		\includegraphics[height=1.2in]{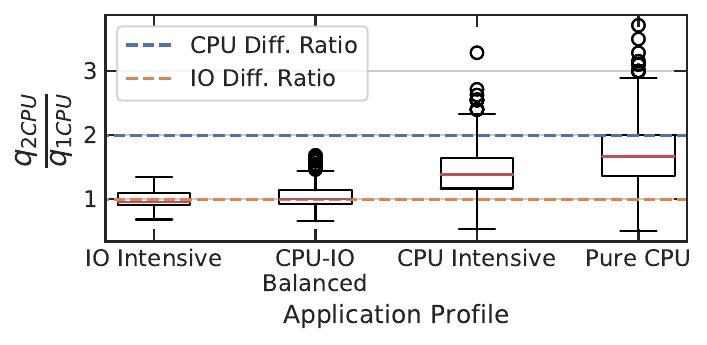}
		\caption{It is hard to accurately estimate the actual server processing speeds since it depends on both provisioned resources, and application profiles.}
		\label{fig:motivation-multi-stage}
	\end{subfigure}
	\hspace{.05in}
	\begin{subfigure}{0.5\columnwidth}
		\centering
		\includegraphics[height=1.2in]{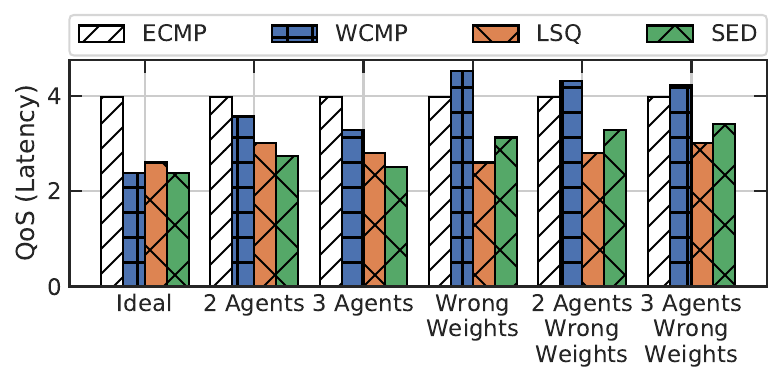}
		\caption{The performance of existing network load balancing algorithms degrades when observation becomes partial with multi-agents and weights are mis-configured.}
		\label{fig:motivation-inaccurate}
	\end{subfigure}
	
	\caption{Existing network load balancing algorithms are sub-optimal under real-world setups.}
	\label{fig:motivation}
	\vskip -.1in
\end{figure}

\textbf{Network Load Balancing Algorithms.} The main goal of network LBs is to \textit{fairly} distribute workloads across servers.
The system transition protocol of network load balancing system is described in Alg.~\ref{alg:model-transition} and depicted in Fig.~\ref{fig:intro}.
Existing load balancing algorithms are sensitive to partial observations and inaccurate server weights.
Equal-Cost Multi-Path (ECMP) LBs randomly assign servers to new requests~\cite{glb2018, faild2018, silkroad2017}, which makes them agnostic to server load state differences.
Weighted-Cost Multi-Path (WCMP) LBs assign weights to servers proportional to their provisioned resources (\eg CPU power)~\cite{maglev, concury2020, prism2020}.
However, the statically assigned weights may not correspond to the actual server processing capacity.
As depicted in Fig.~\ref{fig:motivation-multi-stage}, servers with the same IO speed yet different CPU capacities have different actual processing speed when applications have different resource requirements.
Active WCMP (AWCMP) is a variant of WCMP and it periodically probe server utilisation information (CPU/memory/IO usage)~\cite{spotlight2018, incab2018}.
However, active probing can cause delayed observations and incur additional control messages, which degrades the performance of distributed networking systems.
Local Shortest Queue (LSQ) assigns new requests to the server with the minimal number of ongoing networking connections that are \textit{locally} observed~\cite{twf2020, cheetah2020}.
It does not concern server processing capacity differences.
Shortest Expected Delay (SED) derives the ``expected delay'' as locally observed server queue length divided by statically configured server processing speed.
However, LSQ and SED are sensitive to partial observations and misconfigurations.
As depicted in Fig.~\ref{fig:motivation-inaccurate}, the QoS performance of each load balancing algorithm degrades from the ideal setup (global observations and accurate server weight configurations) when network traffic is split across multiple LBs or server weights are mis-configured\footnote{The stochastic Markov model of the simulation is detailed in Sec.~\ref{app:model-basic}}, which prevails in real-world cloud DCs.

In this paper, we propose a distributed MARL-based load balancing algorithm that considers dynamically changing queue lengths (\eg sub-ms in modern DC networks~\cite{guo2015pingmesh}), and autonomously adapts to actual server processing capacities, with no additional communications among LB agents or servers.

\textbf{Markov Potential Games.} 
A potential game (PG)~\cite{monderer1996potential, sandholm2001potential, marden2009cooperative, candogan2011flows} has a special function called \textit{potential function}, which specifies a property that any individual deviation of the action for one player will change the value of its own and the potential function equivalently. A desirable property of PG is that pure NE always exists and coincides with the maximum of potential function in norm-form setting. Self-play~\cite{fudenberg1998theory} is provably converged for PG.
Markov games (MG) is an extension of normal-form game to a multi-step sequential setting. A combination of PG and MG yields the Markov potential games (MPG)~\cite{fox2022independent, leonardos2021global}, where pure NE is also proved to exist. Some algorithms~\cite{macua2018learning, mguni2021learning, fox2022independent} lying in the intersection of game theory and reinforcement learning are proposed for MPG.
For example, independent nature policy gradient is proved to converge to Nash equilibrium (NE) for MPG~\cite{fox2022independent}. 


\begin{table}[tbp]
\centering
\caption{Trade-offs among the probing frequency, measurement quality, and communication overhead.}
\label{tab:communication-overhead}
\begin{tabular}{cc|c|c|c|c|c}
\hline
\multicolumn{2}{c|}{Probing Frequency (/s)}                                                                                    & 2.22  & 2.86  & 4.00  & 6.67  & 20.00  \\ \hline
\multicolumn{1}{c|}{\multirow{2}{*}{RMSE}}                                                                      & CPU (\%)     & 48.33 & 44.56 & 39.84 & 32.65 & 21.97  \\
\multicolumn{1}{c|}{}                                                                                           & \#Job        & 2.07  & 1.85  & 1.61  & 1.31  & 0.91   \\ \hline
\multicolumn{1}{c|}{\multirow{2}{*}{Spearman's Corr.}}                                                          & CPU (\%)     & 0.28  & 0.40  & 0.52  & 0.68  & 0.85   \\
\multicolumn{1}{c|}{}                                                                                           & \#Job        & 0.47  & 0.56  & 0.66  & 0.77  & 0.89   \\ \hline
\multicolumn{1}{c|}{\multirow{2}{*}{\begin{tabular}[c]{@{}c@{}}Communication\\ Overhead (kbps)\end{tabular}}} & 2LB-7server  & 2.15  & 2.76  & 3.86  & 6.44  & 9.32   \\
\multicolumn{1}{c|}{}                                                                                           & 6LB-20server & 18.40 & 23.66 & 33.12 & 55.20 & 165.60 \\ \hline
\end{tabular}
\end{table}

\textbf{Multi-Agent RL}. MARL~\cite{yang2020overview} has been viewed as an important avenue for solving different types of games in recent years. For cooperative settings, a line of work based on joint-value factorisation have been proposed, involving VDN~\cite{sunehag2017value}, COMA~\cite{foerster2018counterfactual}, MADDPG~\cite{lowe2017multi}, and QMIX~\cite{rashid2018qmix}. For these works, a global reward is assigned to players within the team, but individual policies are optimised to execute individual actions, known as the CTDE setting.
MPG satisfies the assumptions of the value decomposition approach, with the well-specified potential function as the joint rewards. However, deploying CTDE RL models in real-world distributed system incurs additional communication latency and management overhead for synchronising agents and aggregating trajectories.
These additional management and communication overheads can incur substantial performance degradation -- constrained throughput and increased latency -- especially in data center networks.
As listed in Table~\ref{tab:communication-overhead}, when we use active probing to measure server utilisation information, higher probing frequencies give improved measurement quality--in terms of CPU usage and number of on-going jobs on the servers.
However, higher probing frequencies also incur increased communication overhead, especially in large-scale data center networks.
The detailed experimental setups, as well as both qualitative and quantitative analysis of the impact of communication overhead, are described in Sec.~\ref{app:results-ablation-comm}.
By leveraging the special structure of MPG, independent learning approach can be more efficient due to the decomposition of the joint state and action spaces, which is leveraged in the proposed methods. Methods like MATRPO~\cite{li2020multi}, IPPO~\cite{de2020independent} follow a fully decentralised setting, but for general cooperative games.

In terms of the distribution fairness, FEN~\cite{jiang2019learning} is proposed as a decentralised approach for fair reward distribution in multi-agent systems. They defined the fairness as coefficient of variation and decompose it for each individual agent. Another work~\cite{zimmer2021learning} proposes a decentralised learning method for fair policies in cooperative games. However, the decentralised learning manner in these methods are not well justified, while in this paper the load balancing problem is formally characterised as a MPG and the effectiveness of distributed training is verified.

\section{Methods}
\subsection{Problem Description}

We formulate the load balancing problem into a discrete-time dynamic game with strong distributed and concurrent settings, where no centralised control mechanism exists among agents.
We let $M$ denote the number of LB agents ($[M]$ denotes the set of LB agents $\{1, \dots, M\}$) and $N$ denote the number of servers ($[N]$ denotes the set of servers $\{1, \dots, N\}$).
At each time step (or round) $t \in H$ in a horizon $H$ of the game, each LB agent $i$ receives a workload $w_i(t) \in W$, where $W$ is the workload distribution, and the LB agent assigns a server to the task using its load balancing policy $\pi_i \in \Pi$, where $\Pi$ is the load balancing policy profile.
At each time-step $t$, a LB agent $i$ takes an action $a_i(t) = \{a_{ij}(t)\}_{j=1}^{N}$, according to which the tasks $w_i(t)$ are assigned with distribution $\alpha_i(t)$. $\alpha_{ij}(t)$ is the probability mass of assigning tasks to server $j$, $ \sum_{j=1}^{N} \alpha_{ij}(t) = 1$.
Therefore, at each time step, the workload assigned to server $j$ by the $i$-th LB is $w_i(t)\alpha_{ij}(t)$.
During each time interval, each server $j$ is capable of processing a certain amount of workload $v_j$ based on the property of each server (\eg provisioned resources including CPU, memory, \etc).
We have server load state (remaining workload to process) $X_j(T) = \sum_{t=0}^{T}\max\{0, \sum_{i=1}^M w_i(t)\alpha_{ij}(t) - v_j\} = \max\{0, \sum_{t=0}^{T}\sum_{i=1}^{M}w_i(t)\alpha_{ij}(t) - v_{j}T\} = \sum_{i=1}^{M}X_{ij}(T)$\footnote{$X_{ij}(T) = \sum_{t=0}^{T}\max\{0, w_i(t)\alpha_{ij}(t) - \frac{v_j}{M}\}$}.
Let $l_{j}$ denote the time for a server $j$ to process all remaining workloads, which is also the potential queuing time for new-coming tasks, $l_j(t) = \frac{X_j(t-1)+\sum_{i=1}^{M}w_i(t)\alpha_{ij}(t)}{v_j} = \frac{\sum_{i=1}^{M}X_{ij}(t-1)+w_i(t)\alpha_{ij}(t)}{v_j} = \sum_{i=1}^{M}l_{ij}(t)$.
Then transition from time step $t$ to time step $t+1$ is given in Alg.~\ref{alg:model-transition}.
Reward: $r_{i}(t) = R(\boldsymbol{l}(t), a_i(t), \delta_i(t))$, where $R$ is the reward function, $\boldsymbol{l}(t) = \sum_{j=1}^{N} l_j(t) = \sum_{i=1}^{M} l_{i}(t)$ denotes the estimated remaining time to process on each server, and $\delta_i(t)$ is a random variable that makes the process stochastic.
\begin{definition}(Makespan)
In the selfish load balancing problem, the makespan is defined as:
{\small
\begin{align}
 \text{MS} = \max_j(l_j), l_j = \sum_i l_{ij}
 \label{eq:makespan}
\end{align}}
\end{definition}
The network load balancing problem is featured as multi-commodity flow problems and is NP-hard, which makes it hard to solve with trivial algorithmic solution within micro-second level~\cite{sen2013scalable}.
This problem can be formulated as a constrained optimisation problem for minimizing the makespan over an horizon $t \in [H]$:
{\small
\begin{align}
    minimize \sum_{t=h}^{H}&\max_j l_j(t) \\[-5pt]
    s.t. \quad
    l_{j}(t)=&\frac{\sum_{i=1}^M (X_{ij}(t-1)+w_{i}(t)\alpha_{ij}(t))}{v_{j}}, \quad \sum_{i=1}^{M}w_i(t) \le \sum_{j=1}^{N}v_j,   \quad w_i, v_j\in(0, +\infty) \label{eq:lb_cons1}\\
    X_{ij}(T)&=\sum_{t=0}^{T} \max\{0, w_{i}(t)\alpha_{ij}(t) - \frac{v_j}{M}\},  \quad\sum_{j=1}^{N}\alpha_{ij}(t) = 1,  \quad \alpha_{ij} \in [0, 1] \label{eq:lb_cons2}
\end{align}}

In modern realistic network load balancing system, the arrival of network requests is usually unpredictable in both its arriving rate and the expected workload, which introduces large stochasticity into the problem. Moreover, due to the existence of noisy measurements and partial observations, the estimation of makespan can be inaccurate, which indicates the actual server load states or processing capacities are not correctly captured. Instant collisions of elephant workloads or bursts of mouse workloads often happen, which do not indicate server processing capacity thus misleading the observation. To solve this issue, we introduce \emph{fairness} as an alternative of the original objective makespan. Specifically, makespan is estimated on a per-server level, while the estimation of fairness can be decomposed to the LB level, which allows evaluating the individual LB performance without general loss. This is more natural in load balancing system due to the partial observability of LBs.

\subsection{Distribution Fairness}
\label{sec:fairness}
We mainly introduce two types of load balancing distribution fairness: (1) variance-based fairness (VBF) and (2) product-based fairness (PBF). It will be proved that optimization over either fairness will be sufficient but not necessary for minimising the makespan.

\begin{definition}(Variance-based Fairness)
\label{def:vbf}
For a vector of time to finish all remaining jobs $\boldsymbol{l}=[l_{1}, \dots, l_{N}]$ on each server $j\in[N]$, let $\overline{\boldsymbol{l}}(t) = \frac{1}{N}\sum_{j=1}^{N}\sum_{i=1}^{M}l_{ij}(t)$, the variance-based fairness for workload distribution is just the negative sample variance of the job time, which is defined as:
{\small
\begin{align}
    F(\boldsymbol{l}) = -\frac{1}{N}\sum_{j=1}^{N}\bigg(l_j(t)-\overline{\boldsymbol{l}}(t)\bigg)^2 = -\frac{1}{N}\sum_{j=1}^{N}l_j^2(t)+\overline{\boldsymbol{l}}^2(t).
\end{align}}
VBF defined per LB is: $F_i(\boldsymbol{l}_i) = -\frac{1}{N}\sum_{j=1}^{N}l_{ij}^2(t)+\overline{\boldsymbol{l}}_{i}^2(t)$, where $\overline{\boldsymbol{l}}_i(t) =  \frac{1}{N}\sum_{j=1}^{N}l_{ij}(t)$.
\end{definition}

\begin{lemma} The VBF for load balancing system satisfies the following property:
\label{lem:vbf}
{\small
\begin{align}
    F_i^{\pi_i, -\pi_i}(\boldsymbol{l}_i)-F_i^{\tilde{\pi}_i, -\pi_i}(\tilde{\boldsymbol{l}}_i) = F^{\pi_i, -\pi_i}(\boldsymbol{l})-F^{\tilde{\pi}_i, -{\pi}_i}(\tilde{\boldsymbol{l}})
\end{align}}
\end{lemma}
This property makes VBF a good choice for the reward function in load balancing tasks. We will see more discussions in later sections. Proof of the lemma is provided in Sec.~\ref{sec:app_vbf}.

\begin{proposition}
\label{prop:var_fairness}
Maximising the VBF is sufficient for minimising the makespan, subjective to the load balancing problem constraints (Eq.~\eqref{eq:lb_cons1} and \eqref{eq:lb_cons2}): $\max F(\boldsymbol{l}) \Rightarrow  \min \max_j(l_j)$. 
This also holds for per-LB VBF as $\max F_i(\boldsymbol{l}_i) \Rightarrow  \min \max_j(\boldsymbol{l}_i)$.
\end{proposition}

\begin{definition}(Product-based Fairness~\cite{yao2022reinforced}) For a vector of time to finish all remaining jobs $\boldsymbol{l}=[l_{1}, \dots, l_{N}]$ on each server $j\in[N]$, the product-based fairness for workload distribution is defined as: $F(\boldsymbol{l}) = F([l_1, \dots, l_N]) = \prod_{j \in [N]}\frac{l_j}{\max(\boldsymbol{l})}$.
PBF defined per LB is: $    F_i(\boldsymbol{l}_i) = F([l_{i1}, \dots, l_{iN}]) = \prod_{j \in [N]}\frac{l_{ij}}{\max(\boldsymbol{l}_i)}$.
\end{definition}

\begin{proposition}
\label{prop:pro_fairness}
Maximising the product-based fairness is sufficient for minimising the makespan, subjective to the load balancing problem constraints (Eq.~\eqref{eq:lb_cons1} and \eqref{eq:lb_cons2}): $\max F(\boldsymbol{l}) \Rightarrow  \min \max(\boldsymbol{l})$.
\end{proposition}
Proofs of proposition \ref{prop:var_fairness} and \ref{prop:pro_fairness} are in Sec.~\ref{sec:app_vbf_pbf}. From proposition~\ref{prop:var_fairness} and \ref{prop:pro_fairness}, we know that the two types of fairness can serve as an effective alternative objective for optimising the makespan, which will be leveraged in our proposed MARL method as valid reward functions.


\subsection{Game Theory Framework}

Markov game is defined as $\mathcal{MG}(H, M, \mathcal{S}, \mathcal{A}_{\times M}, \mathbb{P}, r_{\times M})$, where $H$ is the horizon of the game, $M$ is the number of player in the game, $\mathcal{S}$ is the state space, $\mathcal{A}_{\times M}$ is the joint action space of all players, $\mathcal{A}_i$ is the action space of player $i$, $\mathbb{P}=\{\mathbb{P}_h\}, h\in[H]$ is a collection of transition probability matrices $\mathbb{P}_h: \mathcal{S}\times \mathcal{A}_{\times M} \rightarrow \Pr(\mathcal{S})$, $r_{\times M}=\{r_i|i\in[M]\}, r_i:\mathcal{S}\times\mathcal{A}_{\times M}\rightarrow \mathbb{R}$ is the reward function for $i$-th player given the joint actions.  The stochastic policy space for the $i$-th player in $\mathcal{MG}$ is defined as $\Pi_i: \mathcal{S} \rightarrow \Pr(\mathcal{A}_i)$, $\Pi=\{\Pi_i\}, i\in[M]$.

For the Markov game $\mathcal{MG}$, the state value function $V_{i, h}^{\boldsymbol{\pi}}: \mathcal{S}\rightarrow \mathbb{R}$ and state-action value function $Q_{i, h}^{\boldsymbol{\pi}}: \mathcal{S}\times \mathcal{A}\rightarrow \mathbb{R}$ for the $i$-th player at step $h$ under policy $\boldsymbol{\pi}\in\Pi_{\times M}$ is defined as:
\begin{equation} \label{eq:V_value}
{\small\begin{aligned}
	 V_{i, h}^{\boldsymbol{\pi}}(s):= \mathbb{E}_{\boldsymbol{\pi},\mathbb{P}}\bigg[\sum_{h' =
        h}^H r_{i, h'}(s_{h'}, \boldsymbol{a}_{h'}) \bigg| s_h = s\bigg], 	 Q_{i, h}^{\boldsymbol{\pi}}(s, \boldsymbol{a}):= \mathbb{E}_{\boldsymbol{\pi},\mathbb{P}}\bigg[\sum_{h' =
        h}^H r_{i, h'}(s_{h'}, \boldsymbol{a}_{h'}) \bigg| s_h = s, a_h=\boldsymbol{a}\bigg].
\end{aligned}}
\end{equation}

\begin{definition}($\epsilon$-approximate Nash equilibrium) Given a Markov game $\mathcal{MG}$ with tuples $(H, M, \mathcal{S}, \mathcal{A}_{\times M}, \mathbb{P}, \Pi_{\times M}, r_{\times M})$, let $\pi_{-i}$ be the policies of the players except for the $i$-th player, the policies $(\pi_i^*, \pi_{-i}^*)$ is an $\epsilon$-Nash equilibrium if $\forall i\in[M], \exists \epsilon>0$, 
{\small
\begin{align}
V_i^{\pi^*_i, \pi^*_{-i}}(s)\ge V_i^{\pi_i, \pi^*_{-i}}(s)-\epsilon, \forall \pi_i \in \Pi_i.
\end{align}}
If $\epsilon=0$, it is an exact Nash equilibrium.
\end{definition}

\begin{definition}(Markov Potential Game) A Markov game $\mathcal{M}(H, M, \mathcal{S}, \mathcal{A}_{\times M}, \mathbb{P}, \Pi_{\times M}, r_{\times M})$ is a Markov potential game (MPG) if $\forall i\in[M], \pi_i, \tilde{\pi}_i\in\Pi_i, \pi_{-i}\in\Pi_{-i}, s\in\mathcal{S}$, 
{\small
\begin{align}
V_i^{\pi_i, \pi_{-i}}(s) - V_i^{\tilde{\pi}_i, \pi_{-i}}(s)=\phi^{\pi_i, \pi_{-i}}(s)-\phi^{\tilde{\pi}_i, \pi_{-i}}(s),
\end{align}}
where $\phi(\cdot)$ is the potential function independent of the player index. 
\end{definition}

\begin{lemma} Pure NE (PNE) always exists for PG, local maximisers of potential function are PNE. PNE also exists for MPG. \cite{monderer1996potential}
\label{lem:pne}
\end{lemma}

\begin{theorem}
\label{thm:mpg_vbf}
Multi-agent load balancing is MPG with the VBF $F_i(\boldsymbol{l}_i)$ as the reward $r_i$ for each LB agent $i\in[M]$, then suppose for $\forall s\in \mathcal{S}$ at step $h\in[H]$, the potential function is time-cumulative total fairness: $\phi^{\pi_i, -\pi_i}(s)=\sum_{t =h}^H F^{\pi_i, -\pi_i}(\boldsymbol{l}(t))$.
\end{theorem}
The proof of the theorem is based on Lemma~\ref{lem:vbf}, and it's provided in Sec.~\ref{sec:app_vbf_pbf}.
This theorem is essential for establishing our method, since it proves that multi-agent load balancing problem can be formulated as a MPG with the time-cumulative VBF as its potential function. Also, the choice of per-LB VBF as reward function for individual agent is critical for making it MPG, it is easy to verify that PBF cannot guarantee such property. From Lemma~\ref{lem:pne} we know the maximiser of potential function is the NE of MPG, and from proposition~\ref{prop:var_fairness} it is known that maximising the VBF gives the sufficient condition for minimising the makespan. Therefore, an effective independent optimisation with respect to the individual reward function specified in the above theorem will lead the minimiser of makespan for load balancing tasks. The effective independent optimisation here means the NE of MPG is achieved.

\subsection{Distributed Method}
\label{sec:rl}


\begin{figure}[t]
	\centering
	\centerline{\includegraphics[width=\columnwidth]{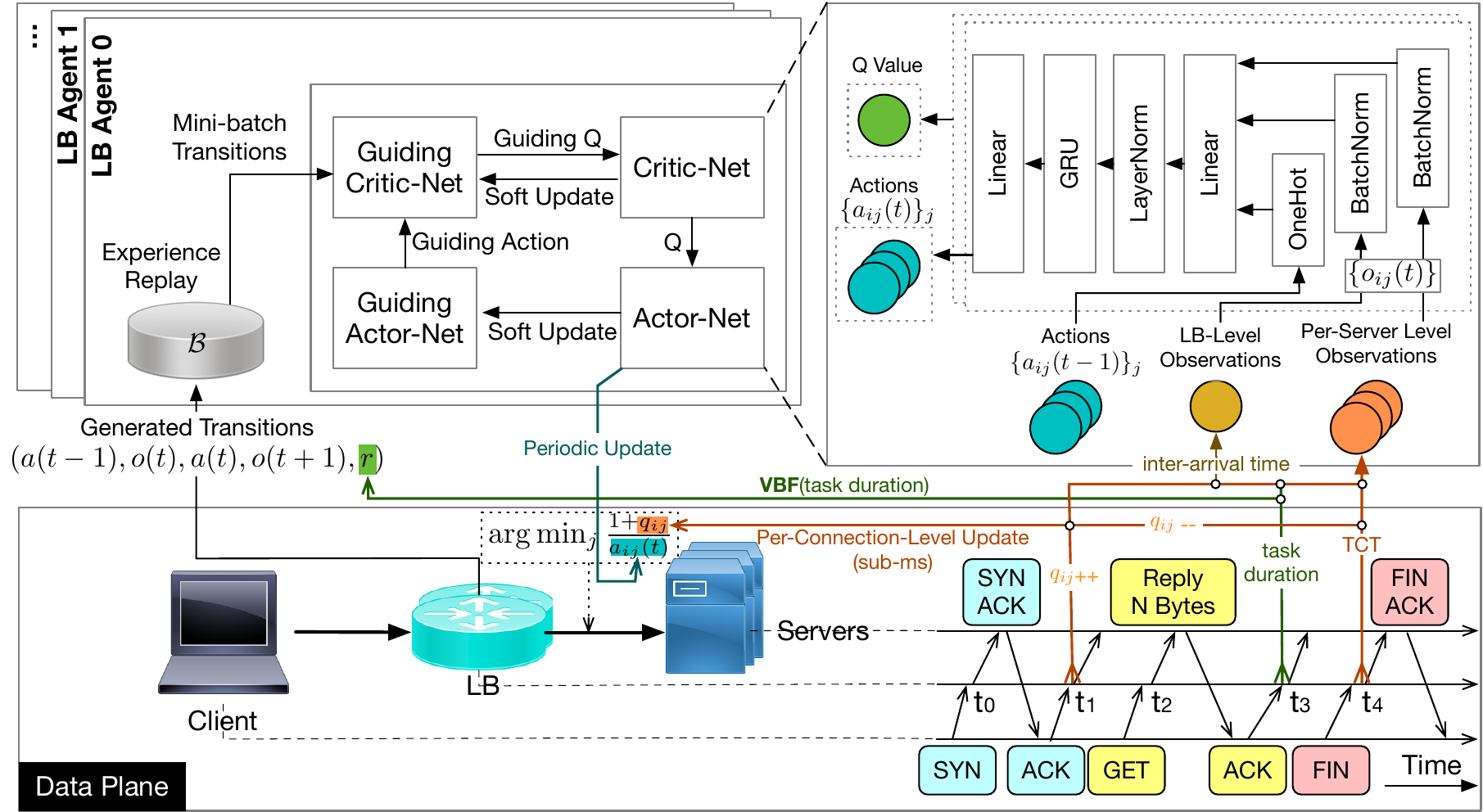}}
	\caption{Overview of the proposed distributed MARL framework for network LB. }
	\label{fig:design}
\end{figure}

With the above analysis, the load balancing problem can be formulated as an episodic version of multi-player partially observable Markov game, which we denote as $\mathcal{POMG}(H, M, \mathcal{S}, \mathcal{O}_{\times M}, \mathbb{O}_{\times M}, \mathcal{A}_{\times M}, \mathbb{P}, r_{\times M})$, where $M, H, \mathcal{S},\mathcal{A}_{\times M}$ and $\mathbb{P}$ follow the same definitions as in Markov game $\mathcal{MG}$, $\mathcal{O}_{\times M}$ contains the observation space $O_i$ for each player, $\mathbb{O}=\{\mathbb{O}_h\}, h\in[H]$ is a collection of observation emission matrices, $\mathbb{O}_{i, h}: \mathcal{S} \rightarrow \Pr(\mathcal{O}_{i})$, $r_{\times M}=\{r_i|i\in[M]\}, r_i:\mathcal{O}_i\times\mathcal{A}_{\times M}\rightarrow \mathbb{R}$ is the reward function for $i$-th LB agent given the joint actions. The stochastic policy space for the $i$-th agent in $\mathcal{POMG}$ is defined as $\Pi_i: \mathcal{O}_i \rightarrow \Pr(\mathcal{A}_i)$. As discussed in Sec.~\ref{sec:background}, the partial observability comes from the fundamental configuration of network LBs in DC networks, which allows LBs to observe only a partial of network traffic and does not give LBs information about the tasks (\eg expected workload) distributed from each LB. The reward functions in our experiments are variants of distribution fairness introduced in Sec.~\ref{sec:fairness}. The potential functions can be defined accordingly based on the two fairness indices. The overview of the proposed distributed MARL framework is shown in Fig.~\ref{fig:design}.

In MPG, independent policy gradient allows finding the maximum of the potential function, which is the PNE for the game. This inspires us to leverage the policy optimisation in a decomposed manner, \emph{i.e.}, distributed RL for policy learning of each LB agent. However, due to the partial observability of the system and the challenge of directly estimating the makespan (Eq.~\eqref{eq:makespan}), each agent cannot have a direct access to the global potential function. To address this problem, the aforementioned fairness (Sec.~\ref{sec:fairness}) can be deployed as the reward function for each agent, which makes the value function as a valid alternative for the potential function as an objective. This also transforms the joint objective (makespan or potential) to individual objectives (per LB fairness) for each agent. Proposition~\ref{prop:var_fairness} and \ref{prop:pro_fairness} verify that optimising towards these fairness indices is sufficient for minimising the makespan.

\begin{algorithm}[tbp]
\footnotesize
\caption{Distributed LB for MPG}
\label{alg:dec_lb_mpg}
\begin{algorithmic}[1]
\STATE \textbf{Initialise:}
\STATE \quad LB policy $\pi_{\theta_i}$ and critic $Q_{\phi_i}$ networks, replay buffer $\mathcal{B}_i, \forall i \in [M]$;
\STATE \quad server processing speed function $v_j, \forall j \in [N]$;
\STATE \quad initial observed instant queue length on server $j$ by the $i$-th LB: $q_{ij}=0,  \forall i\in[M], j\in[N]$.
\WHILE {not converge}
\STATE Reset server load state $X_j(1) \gets 0, \forall j\in[N]$
\STATE Each LB agent $i$ ($i\in[M]$) receives individual observation ${o}_i(1)$
\FOR {$t=1,\dots, H$}
    \STATE Initialise distributed workload $m_{ij}, w_{i}(t) \gets 0, i\in[M], j\in[N]$ 
    \STATE Get actions $a_{i}(t) \gets \{a_{ij}(t)\}_{j=1}^N = \pi_{\theta_i}({o}_{i}(t)), i\in[M]$
    
    \FOR {job $\tilde{w}$ arrived at LB $i$ between timestep [$t$, $t+1$)}
        \STATE LB $i$ assigns $\tilde{w}$ to server $j = \arg \min_{k\in[N]}\frac{q_{ik}(t)+1}{a_{ik}(t)}$ \alglinelabel{line:job_update_start}
        \STATE $m_{ij} \gets m_{ij}+\tilde{w}$, $w_{i}(t)\gets w_{i}(t)+ \tilde{w}$
        \STATE $\alpha_{ij}(t) \gets \frac{m_{ij}}{w_{i}(t)}$ \alglinelabel{line:job_update_end}
    \ENDFOR
    
    \FOR {each server $j$}
        \STATE Update workload: {$X_{ij}(t+1) \gets \max\{X_{ij}(t)+w_{i}(t)\alpha_{ij}(t) - \frac{v_j}{M}, 0\}$}
        \STATE $X_{j}(t+1) \gets \sum_{i=1}^{M}X_{ij}(t)$
    \ENDFOR
    \STATE Each agent receives individual reward $r_i(t)$
	\STATE Each agent $i$ collects observation ${o}_i(t+1), i\in[M]$
    \STATE Update replay buffer: $\mathcal{B}_i=\mathcal{B}_i\bigcup (a_i(t-1), {o}_i(t), a_i(t), r_i(t), {o}_i(t+1)), i\in[M]$
\ENDFOR
\STATE Update critics with gradients: $\nabla_{\phi_i}\mathbb{E}_{(o_i,a_i, r_i, o'_i)\sim \mathcal{B}_i}\bigg[\bigg(Q_{\phi_i}(o_i,a_i)-r_i-\gamma V_{\tilde{\phi}_i}(o_i^\prime)\bigg)^2\bigg]$\\
\STATE where $V_{\tilde{\phi}_i}(o_i^\prime)=\mathbb{E}_{ (o_i^\prime, a'_i)\sim\mathcal{B}_i}[Q_{\tilde{\phi}_i}(o_i^\prime, a_i^\prime)-\alpha\log\pi_{\theta_i}(a_i^\prime|o_i^\prime)], i\in[M]$
\STATE Update policies with gradients: -$\nabla_{\theta_i}\mathbb{E}_{o_i\sim\mathcal{B}_i}[\mathbb{E}_{a\sim\pi_{\theta_i}}[\alpha\log\pi_{\theta_i}(a_i|o_i)-Q_{\phi_i}(o_i,a_i)]], i\in[M]$
\ENDWHILE
\RETURN final models of learning agents
\end{algorithmic}
\end{algorithm}

Alg.~\ref{alg:dec_lb_mpg} shows the proposed distributed LB for load balancing problem, which is a partially observable MPG. The distributed policy optimisation is based on Soft Actor-Critic (SAC)~\cite{haarnoja2018soft} algorithm, which is a type of maximum-entropy RL method. It optimises the objective $\mathbb{E}[\sum_t \gamma^t r_t+\alpha \mathcal{H(\pi_\theta)}]$, whereas $\mathcal{H}(\cdot)$ is the entropy of the policy $\pi_\theta$. Specifically, the critic $Q$ network is updated with gradient $\nabla_\phi\mathbb{E}_{o,a}\bigg[\bigg(Q_\phi(o,a)-r(o,a)-\gamma \mathbb{E}_{o^\prime}[V_{\tilde{\phi}}(o^\prime)]\bigg)^2\bigg]$, where $V_{\tilde{\phi}}(o^\prime)=\mathbb{E}_{a^\prime}[Q_{\tilde{\phi}}(o^\prime, a^\prime)-\alpha\log\pi_\theta(a^\prime|o^\prime)]$ and $Q_{\tilde{\phi}}$ is the target $Q$ network; the actor policy $\pi_\theta$ is updated with the gradient $\nabla_\theta\mathbb{E}_o[\mathbb{E}_{a\sim\pi_\theta}[\alpha\log\pi_\theta(a|o)-Q_\phi(o,a)]]$. Other key elements of RL methods involve the observation, action and reward function, which are detailed as following.

\textbf{Observation.} 
Each LB agent partially observes over the traffic that traverses through itself, including per-server-level and LB-level measurements. For each LB, per-server-level observations consist of -- for each server -- the number of on going tasks, and sampled task duration and task completion time (TCT).
Specifically, in Alg.~\ref{alg:dec_lb_mpg} line \ref{line:job_update_start}-\ref{line:job_update_end}, $w_{i}$ is the coming workload on servers assigned by $i$-th LB, and it is not observable for LB. $q_{ik}+1$ is the locally observed number of tasks on $k$-th server by $i$-th LB, due to the real-world constraints of limited observability at the Transport layer. The ``+1'' is for taking into account the new-coming task. 
Observations of task duration and TCT samples, along with LB-level measurements which sample the task inter-arrival time as an indication of overall system load state, are reduced to 5 scalars -- \ie average, 90th-percentile, standard deviation, discounted average and weighted discounted average\footnote{Discounted average weights are computed as $0.9^{t^\prime-t}$, where $t$ is the sample timestamp and $t^\prime$ is the moment of calculating the reduced scalar.} -- as inputs for LB agents.

\textbf{Action.}
To bridge the different timing constraints between the control plane and data plane, each LB agent assigns the $j$-th server to newly arrived tasks using the ratio of two factors, $\arg \min_{k\in[N]} \frac{q_{ik}+1}{a_{ik}}$, where the number of on-going tasks $q_{ik}$ helps track dynamic system server occupation at per-connection level -- which allows making load balancing decision at $\mu$s-level speed -- and $a_{ik}$ is the periodically updated RL-inferred server processing speed. As in line \ref{line:job_update_end} of Alg.~\ref{alg:dec_lb_mpg}, $\alpha_{ij}(t)$ is a statistical estimation of workload assignment distribution at time interval $[t, t+1)$.

\textbf{Reward.} The individual reward for distributed MPG LB is chosen as the VBF (as Def.~\ref{def:vbf}) of the discounted average of sampled task duration measured on each LB agent, such that the LB group jointly optimise towards the potential function defined in Eq.~\eqref{thm:mpg_vbf}.
Task duration information is gathered as the time interval between the end of connection initialisation (\eg $3$-way handshake for TCP traffic) and the acknowledgement to the first data packet (\eg the first ACK packet for TCP traffic).
Given the limited and partial observability of LB agents, task duration information approximates the remaining workload $\boldsymbol{l}$ by measuring the queuing and processing delay for new-coming tasks on each server.
This PBF- and MS-based rewards are also implemented for CTDE MARL algorithm as a comparison.

\textbf{Model.} The architecture of the proposed RL framework is depicted in Fig.~\ref{fig:design}.
Each LB agent consists of a replay buffer, and a pair of actor-critic networks, whose architecture is depicted on the top right.
There is also a pair of guiding actor-critic networks, with the same network architectures but updated in a delayed and soft manner.
Each LB agent takes observations $o_i(t)$ extracted from the data plane (\eg numbers of ongoing tasks $\{q_{ij}\}$, task duration, TCT) and actions from previous timestep $a_{i}(t-1)$ as inputs, and periodically generates new actions $a_{i}(t)$, which is used to update the server assignment function $\arg \min_{j\in[N]} \frac{q_{ij}+1}{a_{ij}}$ in the data plane. The gated recurrent units (GRU)~\cite{chung2014empirical} are applied for all agents to leverage the sequential history information for handling partial observability. 

\section{Evaluation}
\label{sec:experiment}

We developed (i) an event-based simulator to study the distance between the NE achieved by the proposed algorithm and the NE achieved by the theoretical optimal load balancing policy (with perfect observation), and (ii) a realistic testbed on physical servers in a DC network providing Apache web services, with real-world network traffic~\cite{wiki_traces}, to evaluate the real-world performance of the proposed algorithm, in comparison with in-production state-of-the-art LB~\cite{maglev}.

\begin{figure}[tbp]
	\centering
	\begin{subfigure}{0.4\columnwidth}
		\centering
		\includegraphics[height=1.3in]{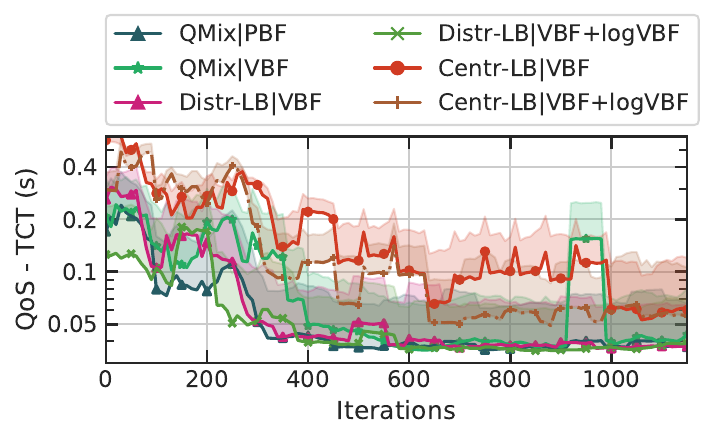}
		\vskip -.1in
		\caption{Learning curves.}
		\label{fig:eval-train}
	\end{subfigure}
	\hspace{.05in}
	\begin{subfigure}{0.55\columnwidth}
		\centering
		\includegraphics[height=1.3in]{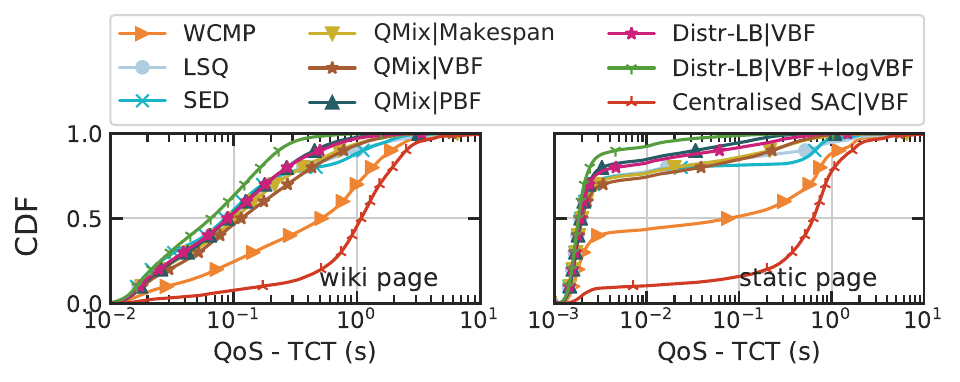}
		\vskip -.1in
		\caption{CDF of TCT.}
		\label{fig:eval-cdf}
	\end{subfigure}
	\caption{Experimental results show that the proposed distributed RL framework using proposed VBF as rewards converges and effectively achieves better load balancing performance (lower TCT and better QoS) than existing LB algorithms and CTDE RL algorithms.}
	\label{fig:eval}
\end{figure}

\begin{table}[tbp]
    \scriptsize
    \centering
    \caption{Comparison of average QoS (s) in moderate-scale real-world network setup.}
    \begin{tabular}{c|c|c|c|c|c}
    \toprule
    \multicolumn{2}{c}{\multirow{2}{*}{Method}} &  \multicolumn{2}{|c|}{Period III ($758.787$ queries/s)} & \multicolumn{2}{c}{Period IV ($784.522$ queries/s)} \\
    \cline{3-6}
     \multicolumn{2}{c|}{} & \multicolumn{1}{c|}{Wiki} & \multicolumn{1}{c|}{Static} & \multicolumn{1}{c|}{Wiki} & \multicolumn{1}{c}{Static} \\
    \hline
     \multicolumn{2}{c|}{WCMP} &  $0.412\pm0.101$ & $0.134\pm0.059$ & $ 0.834\pm0.323$ & $0.492\pm0.276 $ \\ 
      \multicolumn{2}{c|}{LSQ} & $0.620\pm0.442$ & $0.339\pm0.316$ & $0.357\pm0.373 $ & $0.173\pm0.299 $  \\ 
      \multicolumn{2}{c|}{SED} &  $0.215\pm0.210$ & $0.051\pm0.081$ & $0.346\pm0.496$ &$0.169\pm0.330 $ \\ \cline{1-2}
      \multirow{2}{*}{\textbf{RLB-SAC}~\cite{yao2022reinforced}} 
      & Jain & $0.193\pm0.073$ & $0.026\pm0.022$ & $0.204\pm0.084$ & $0.039\pm0.047$ \\
      & G    & $0.149\pm0.049$ & $0.015\pm0.011$ & $0.155\pm0.052$ & $0.011\pm0.011$ \\	\cline{1-2}
      \multirow{3}{*}{\textbf{QMix-LB}} & MS &  $0.217\pm0.157$ & $0.048\pm0.069$ &$0.263\pm0.202 $ &$0.073\pm0.092 $  \\ 
    & VBF &  $0.141\pm0.025$ & $0.008\pm0.004$ &$ 0.286\pm0.162$ & $0.068\pm0.066 $ \\ 
      & PBF &  $0.211\pm0.153$ & $0.047\pm0.078$ & $0.181\pm0.042 $& $0.018\pm0.009 $ \\ \cline{1-2}
    \multirow{2}{*}{\makecell{\textbf{Distr-LB}\\(this paper)}} & VBF &  $0.159\pm0.054$ & $0.017\pm0.009$ & $0.196\pm0.091 $& $ 0.032\pm0.033$\\ 
    & VBF+$\log$VBF &  $\mathbf{0.108\pm0.022}$ & $\mathbf{0.004\pm0.001}$ & $\mathbf{0.104\pm0.013}$ & $\mathbf{0.006\pm0.003} $\\ \cline{1-2}
    \multirow{2}{*}{\textbf{Centr-LB}} & VBF & $1.068\pm0.386$ & $0.570\pm0.378$ & $ 1.378\pm0.377$& $ 0.867\pm0.350$\\ 
    & VBF+$\log$VBF & $0.759\pm0.254$ & $0.306\pm0.222$ & $ 1.013\pm0.168$&$ 0.520\pm0.167$ \\
    \bottomrule
    \end{tabular}
    \label{tab:compare_small_scale_518}
\end{table}

\textbf{Moderate-Scale Real-World Testbed:} As depicted in Fig.~\ref{fig:eval-train}, in a moderate-scale real-world DC network setup with $2$ LB agents and $7$ servers, after $120$ episodes of training, the proposed distributed LB (Distr-LB) algorithm is able to learn from the environment based on VBF as rewards, and it converges to offer better QoS than QMix.
Centralised RL agent (Centr-LB) has difficulties to learn within $120$ episodes because of the increased state and action space.
An empirical finding is that, by adding a log term to the VBF-based reward for Distr-LB, we help LB agents to become more sensitive to close-to-$0$ VBF during training ($\nabla_x \log f(x)>\nabla_xf(x) \text{ when } f(x)< 1$), therefore achieving better load balancing performance.
As depicted in Fig.~\ref{fig:eval-cdf}, when comparing with in-production LB algorithms (WCMP, LSQ, SED), Distr-LB shows clear performance gains and reduced TCT for both types of web pages -- Wikipedia pages require to query SQL databases thus they are more CPU-intensive, while static pages are IO-intensive.
The comparison of average TCT using different LB algorithms is shown in Table~\ref{tab:compare_small_scale_518}.
The proposed Distr-LB also shows superior performance than the RL-based solution (RLB-SAC)~\cite{yao2022reinforced} because of (i) a well designed MARL framework, and (ii) the use of recurrent neural network to handle load balancing problem as a sequential problem.

\begin{table}[tbp]
\scriptsize
\centering
\caption{Comparison of average QoS (s) in moderate-scale simulator for different types of applications.}
\begin{tabular}{c|c|c|c|c}
\toprule
\multicolumn{2}{c|}{} & \multicolumn{1}{c|}{50\%-CPU+50\%-IO} &  \multicolumn{1}{c|}{75\%-CPU+25\%-IO} & \multicolumn{1}{c}{100\%-CPU} \\ \hline
\multicolumn{2}{c|}{Oracle} & $6.437\pm1.006 $ & $1.469\pm0.102 $ & $1.291\pm0.075 $ \\ \cline{1-2}
\multirow{2}{*}{\textbf{QMix-LB}} & PBF & $ 10.230\pm0.108$ & $1.828\pm0.054 $ & $ 2.200\pm0.288$ \\ 
 & VBF & $10.936\pm0.470 $  & $2.023\pm0.255 $ & $2.125\pm0.074 $ \\ \cline{1-2}
\multirow{2}{*}{\makecell{\textbf{Distr-LB}\\(this paper)}} & VBF& $ 10.335\pm0.362$ &$\mathbf{1.695\pm0.104} $  & $ \mathbf{1.643\pm0.016}$ \\ 
& VBF+$\log$VBF & $ \mathbf{8.797\pm0.459}$ &$1.873\pm0.328 $  & $ 2.004\pm0.042$ \\
\bottomrule
\end{tabular}
\label{tab:simulation-optimal-distance}
\end{table}

\begin{table}[tbp]
    \centering
    \caption{Comparison of average QoS (s) in large-scale real-world network setup.}
    \resizebox{\columnwidth}{!}{ 
    \begin{tabular}{c|c|c|c|c|c}
    \toprule
    \multicolumn{2}{c}{\multirow{2}{*}{Method}} &  \multicolumn{2}{|c|}{Period I ($2022.855$ queries/s)} & \multicolumn{2}{c}{Period II ($2071.129$ queries/s)} \\
    \cline{3-6}
    \multicolumn{2}{c|}{} & \multicolumn{1}{c|}{Wiki} & \multicolumn{1}{c|}{Static} & \multicolumn{1}{c|}{Wiki} & \multicolumn{1}{c}{Static} \\
    \hline
     \multicolumn{2}{c|}{WCMP}  &        $0.473\pm0.102$ & $0.194\pm0.090$ & $0.460\pm0.241$ & $0.239\pm0.212$ \\ 
      \multicolumn{2}{c|}{LSQ}  &        $0.266\pm0.127$ & $0.063\pm0.065$ & $0.218\pm0.246$ & $0.082\pm0.152$  \\ 
      \multicolumn{2}{c|}{SED}  &        $0.169\pm0.062$ & $0.020\pm0.025$ & $0.166\pm0.141$ & $0.050\pm0.070$ \\
      \multicolumn{2}{c|}{RLB-SAC-G\cite{yao2022reinforced}}  &  $0.182\pm0.049$ & $0.013\pm0.009$ & $0.111\pm0.029$ & $0.010\pm0.009$ \\ \cline{1-2}
    \multirow{2}{*}{\textbf{QMix-LB}} 
    & VBF                       &  $0.181\pm0.062$ & $0.019\pm0.020$ & $0.188\pm0.147$ & $0.052\pm0.075$ \\ 
    & PBF                       &  $0.210\pm0.041$ & $0.013\pm0.006$ & $0.104\pm0.009$ & $0.005\pm0.003 $ \\ \cline{1-2}
    \multirow{2}{*}{\makecell{\textbf{Distr-LB}\\(this paper)}} 
    & VBF                       & $0.228\pm0.055$ & $0.019\pm0.011$ & $0.174\pm0.102$& $ 0.035\pm0.039$\\ 
    & VBF+$\log$VBF             & $\mathbf{0.161\pm0.033}$ & $\mathbf{0.008\pm0.003}$ & $\mathbf{0.094\pm0.015}$ & $\mathbf{0.004\pm0.001}$\\ 
    \bottomrule
    \end{tabular}
    }
    \label{tab:compare_large_scale}
\end{table}

\textbf{NE Gap Evaluation with Simulation:} To evaluate the gap between the performance of Distr-LB and the theoretical optimal policy, we implement in the simulator an Oracle LB, which has perfect observation (inaccessible in real world) over the system and minimises makespan for each load balancing decision. 
Table~\ref{tab:simulation-optimal-distance} shows that, for different types of applications, Distr-LB is able to achieve closer-to-optimal performance than QMix.
As the simulator is implemented based on the load balancing model formulated in this paper, our theoretical analysis can be directly applied, and VBF -- as a potential function -- helps independent cooperative LB agents to achieve good performance.
The additional $log$ term shows empirical performance gains in real-world system, yet it is not necessarily the case in these simulation results.
On one hand, the generated traffic of tasks in the simulation has higher expected workload ($>1$s mean and stddev), while the $log$ terms is more sensitive to close-to-$0$ variances, which is the case in real-world experimental setups.
On the other hand, though the simulator models the formulated LB problem, it fails to captures the complexity in the real-world system -- \eg Apache backlog, multi-processing optimisation, context switching, multi-level cache, network queues \etc
For instance, batch processing~\cite{vpp} helps reduce cache and instruction misses, yet yields similar processing time for different tasks, thus the variance of task processing delay decreases and becomes closer to $0$ in real-world system. 
The additional $log$ term exaggerates the low variance differences to better evaluate load balancing decisions.

\begin{table}[tbp]
\scriptsize
\centering
\caption{Comparison of $99$-th percentile QoS (s) of Wiki pages under different traffic rates using large-scale real-world setup.}
\begin{tabularx}{\textwidth}{c|c|X|X|X|X|X|X|X|X|X}
\toprule
 \multicolumn{2}{c|}{\multirow{2}{*}{Method}} & \multicolumn{9}{c}{Traffic Rate (queries/s)}  \\ \cline{3-11}
 \multicolumn{2}{c|}{} & 731.534 & 1097.3  & 1463.067 & 1828.834  & 2194.601 & 2377.484  & 2560.368 & 2743.251  & 2926.135\\ \hline
 \multicolumn{2}{c|}{\multirow{2}{*}{LSQ}} & 0.175\newline$\pm$0.015  & 0.212\newline$\pm$0.025  & 0.249\newline$\pm$0.043 & 0.342\newline$\pm$0.121  & 0.827\newline$\pm$0.572  & 2.103\newline$\pm$0.654 & 10.662\newline$\pm$2.557  & 17.656\newline$\pm$0.714 & 17.999\newline$\pm$0.253\\ \hline
 \multicolumn{2}{c|}{\multirow{2}{*}{SED}}  & 0.201\newline$\pm$0.022  & 0.261\newline$\pm$0.079  & 0.322\newline$\pm$0.099 & 0.360\newline$\pm$0.088  & 0.618\newline$\pm$0.268  & 2.175\newline$\pm$1.328 &  11.444\newline$\pm$3.861 & 22.086\newline$\pm$4.892 & 22.727\newline$\pm$5.632 \\ \hline
 \multirow{4}{*}{\makecell{\textbf{Distr-LB}\\(this paper)}}  &  \multirow{2}{*}{VBF} & \textbf{0.160\newline$\pm$0.010}  & \textbf{0.205\newline$\pm$0.036} & \textbf{0.248\newline$\pm$0.086}  & \textbf{0.284\newline$\pm$0.113}  & 0.567\newline$\pm$0.306  & \textbf{1.276\newline$\pm$0.647} & 7.005\newline$\pm$1.147 & 10.560\newline$\pm$1.042 & 15.745\newline$\pm$0.254 \\ \cline{2-11}
 & \multirow{2}{*}{VBF+$\log$VBF}  &  0.161\newline$\pm$0.008 & 0.216\newline$\pm$0.052  & 0.249\newline$\pm$0.068 & 0.348\newline$\pm$0.122  & \textbf{0.439\newline$\pm$0.121}  & 1.533\newline$\pm$0.670  &  \textbf{4.427\newline$\pm$0.443} & \textbf{9.391\newline$\pm$0.329} & \textbf{15.347\newline$\pm$0.572} \\ 
\bottomrule
\end{tabularx}
\vskip -.1in
\label{tab:99_qos_wiki}
\end{table}

\begin{table}[tbp]
\scriptsize
\centering
\caption{Comparison of $99$-th percentile QoS (s) of static pages under different traffic rates using large-scale real-world setup.}
\begin{tabularx}{\textwidth}{c|c|X|X|X|X|X|X|X|X|X}
\toprule
 \multicolumn{2}{c|}{\multirow{2}{*}{Method}} & \multicolumn{9}{c}{Traffic Rate (queries/s)}  \\ \cline{3-11}
 \multicolumn{2}{c|}{} & 731.534 & 1097.3  & 1463.067 & 1828.834  & 2194.601 & 2377.484  & 2560.368 & 2743.251  & 2926.135\\ \hline
 \multicolumn{2}{c|}{\multirow{2}{*}{LSQ}} & 0.014\newline$\pm$0.001  & 0.015\newline$\pm$0.000  & 0.015\newline$\pm$0.000 & 0.018\newline$\pm$0.003  & 0.217\newline$\pm$0.305  & 0.856\newline$\pm$0.554 & 11.066\newline$\pm$3.095  & 16.874\newline$\pm$0.391 & 17.155\newline$\pm$0.217\\ \hline
 \multicolumn{2}{c|}{\multirow{2}{*}{SED}}  & 0.014 \newline$\pm$0.000  & 0.015\newline$\pm$0.000  & 0.016\newline$\pm$0.001 & 0.018\newline$\pm$0.001  & 0.071\newline$\pm$0.066  & 1.252\newline$\pm$1.489 &  11.272\newline$\pm$3.975 & 21.941\newline$\pm$5.970 & 20.708\newline$\pm$5.423 \\ \hline
 \multirow{4}{*}{\makecell{\textbf{Distr-LB}\\(this paper)}}  &  \multirow{2}{*}{VBF} & 0.014\newline$\pm$0.000  & 0.015\newline$\pm$0.000 & 0.016\newline$\pm$0.001  & \textbf{0.017 \newline$\pm$0.000}  & \textbf{0.041\newline$\pm$0.025}  & \textbf{0.338 \newline$\pm$0.364} & 6.670\newline$\pm$1.152 & 9.743\newline$\pm$0.863 & 15.506\newline$\pm$0.056 \\ \cline{2-11}
 & \multirow{2}{*}{VBF+$\log$VBF}  &  0.014\newline$\pm$0.000 & 0.015\newline$\pm$0.001  & 0.016 \newline$\pm$0.000 & 0.018\newline$\pm$0.002  & 0.072\newline$\pm$0.087  & 0.465\newline$\pm$0.403  &  \textbf{3.970\newline$\pm$0.545} & \textbf{8.782\newline$\pm$0.187} & \textbf{15.095\newline$\pm$0.497} \\ 
\bottomrule
\end{tabularx}
\vskip -.1in
\label{tab:99_qos_static}
\end{table}

\textbf{Large-Scale Real-World Testbed:} To evaluate the performance of Distr-LB in large-scale DC networks in real world, we scale up the real-world testbed to have $6$ LB agents and $20$ servers and apply heavier network traffic ($>2000$ queries/s) to evaluate the performance of the LB algorithms that achieved the best performance in moderate scale setups, in comparison with in-production LB algorithms.
The test results after $200$ episodes of training are shown in Table~\ref{tab:compare_large_scale}, where Distr-LB achieves the best performance in all cases.
QMix also outperforms in-production LB algorithms. But as a CTDE algorithm, similar to the Centr-LB, it requires agents to communicate their trajectories, which -- after $200$ episodes of training -- become $221$MiB communication overhead at the end of each episode (episodic training), whereas $95\%$-percentile per-destination-rack flow rate is less than $1$MiB/s~\cite{facebook-dc-traffic}.

\textbf{Scaling Experiments:} Using the same large-scale real-world testbed with $6$ LB agents and $20$ servers, we conduct scaling experiments by applying network traces with different traffic rates, comparing $4$ LB methods with the best performances.
The $99$-th percentile QoS for both Wiki and static pages are shown in Table~\ref{tab:99_qos_wiki},~\ref{tab:99_qos_static}.
As listed in Table~\ref{tab:99_qos_wiki} and~\ref{tab:99_qos_static}, under low traffic rates, when servers are all under utilised, the advantage of our proposed Distr-LB is not obvious because all resources are over-provisioned. 
With the increase of traffic rates (till servers are $100\%$ saturated), our methods outperforms the best classical LB methods.




\section{Stochastic Markov Model}
\label{app:model-basic}

The simulation results of Fig.~\ref{fig:motivation-inaccurate} is based on a basic load balancing setup of $2$ servers with different processing capacities $\frac{v_1}{v_2} = 2$ (\ie server $1$ is 2x faster than server $2$).
Each server has a queue of size $Q$, such that $0\leq l_1, l_2 \leq Q$.
Traffic arrivals and departures are modeled as Poisson processes with rates $\lambda$ (observed traffic), $\gamma$ (unobserved traffic), and $v_1$, $v_2$.
With sufficiently short timeslots, it can be assumed that only one arrival or departure (at most) happen at a given timeslot (\ie $\sum_{i=1}^2 (\lambda_i + \gamma_i + v_i) \leq 1$); the system is then Markovian with the state $(l_1, l_2)$, departure rates $(\mu_1, \mu_2)$, and arrival rates $(\lambda_1, \lambda_2, \gamma_1, \gamma_2)$.
For simplicity and stability, the system works at {\em nominal} capacity (\ie $\lambda + \gamma = v$).
With $q_i(n)_{l_i}$ denoting the probability (or probability density function), of server $q_i$ to have a queue length of $l_i$ at time-step $n$, the transition of server occupations between two time-steps can be described as, for $0 < l_i < Q$ (corner cases are treated accordingly):
{\small
\begin{align*}
	q_i(n)_{l_i} - q_i(n-1)_{l_i} = ( \lambda_i + \gamma_i ) \cdot q_i(n-1)_{l_i-1} + v_i \cdot s_i(n-1)_{l_i+1} - (\lambda_i+\gamma_i+v_i) \cdot q_i(n-1)_{l_i}.
\end{align*}
}%

The QoS performance of each load balancing algorithm in Fig.~\ref{fig:motivation-inaccurate} is measured as the weighted service duration of a connection ($\sum_{i\in\{1, 2\}} \frac{l_i}{l_1+l_2} \frac{l_i}{\mu_i}$), under different configurations.
When the LB has accurate observations and configurations (observing $100\%$ traffic -- \ie $\gamma = 0$ -- and assigning server weights based on actual processing speeds $\frac{w_1}{w_2}=\frac{v_1}{v_2}=2$), WCMP and SED have the best performance.
When the LB observes only partial network traffic ($50\%-Q$ and $33\%-Q$ corresponds to $\gamma = \lambda$, $\gamma = 2*\lambda$, respectively) and the rest of the network traffic is uniformly split between the two servers ($\gamma_1 = \gamma_2$), LSQ and SED outperform WCMP, which is agnostic to instant server occupancy.
However, partial traffic observation also degrades the performance of LSQ and SED.
When LBs have inaccurate server weights ($\sim W$ \ie in case of mis-configuration, $\frac{w_1}{w_2}=\frac{1}{2}$, while $\frac{\mu_1}{\mu_2}=2$), WCMP and SED exhibit degraded performance even when the LB agent sees all the traffic ($\gamma = 0$).
Taking both server queue lengths and processing speeds into account, SED makes more informed load balancing decisions, yet its performance risks being degraded by both partial observations on server queue lengths and inaccurate server weights.

\section{Analysis of Distribution Fairness}
\label{sec:app_vbf_pbf}
\paragraph{Analysis of Variance-Based Fairness}
\label{sec:app_vbf}
\begin{lemma} The VBF for load balancing system satisfies the following property:
\begin{align}
    F_i^{\pi_i, -\pi_i}(\boldsymbol{l}_i)-F_i^{\tilde{\pi}_i, -\pi_i}(\tilde{\boldsymbol{l}}_i) = F^{\pi_i, -\pi_i}(\boldsymbol{l})-F^{\tilde{\pi}_i, -{\pi}_i}(\tilde{\boldsymbol{l}})
\end{align}
\end{lemma}

\begin{proof}
From the definition of the variance-based fairness (as Def.~\ref{def:vbf}) we have the following for $\forall i\in[M], j\in[N]$,
\begin{align}
    F^{\pi_i, -\pi_i}(\boldsymbol{l}) &= -\frac{1}{N}\sum_{j=1}^{N}(l_j-\overline{\boldsymbol{l}})^2\\
    F_i^{\pi_i, -\pi_i}(\boldsymbol{l}_i)&= -\frac{1}{N}\sum_{j=1}^{N}(l_{ij}-\overline{l_i})^2 \quad (\overline{l_i}=\frac{1}{N}\sum_{j=1}^{N}l_{ij})
\end{align}
By indexing the agent $i$ as the one to change its strategy and slightly abusing notation, denote $l_j = l_{ij} + l_{-ij}$, where $l_{-ij} = \sum_{k\neq i}l_{kj}$.
\begin{align}
     F^{\pi_i, -\pi_i}(\boldsymbol{l})&=-\frac{1}{N}\sum_{j=1}^{N}(l_{ij}+l_{-ij}-\overline{(l_{i}+l_{-i})})^2 \quad (\text{where }\overline{(l_{i}+l_{-i})}=\frac{1}{N}\sum_j(l_{ij}+l_{-ij}))\\
     &=-\frac{1}{N}\sum_{j=1}^{N}[l_{ij}+l_{-ij}-(\overline{l}_{i}+\overline{l}_{-i})]^2\\
    &=-\frac{1}{N}\sum_{j=1}^{N}[(l_{ij}-\overline{l}_{i})^2+(l_{-ij}-\overline{l}_{-i})^2-2(l_{ij}-\overline{l}_{i})(l_{-ij}-\overline{l}_{-i})]\\
    &=-\frac{1}{N}\sum_{j=1}^{N}(l_{ij}-\overline{l_{i}})^2-\frac{1}{N}\sum_{j=1}^{N}[(l_{-ij}-\overline{l}_{-i})^2-\frac{2}{N}\sum_{j=1}^{N}(l_{ij}-\overline{l}_{i})(l_{-ij}-\overline{l}_{-i})]\\
    &=F_i^{\pi_i, -\pi_i}(\boldsymbol{l}_i)-\frac{1}{N}\sum_{j=1}^{N}(l_{-ij}-\overline{l}_{-i})^2 \quad (\sum_{j=1}^{N}(l_{ij}-\overline{l}_i)=0)
\end{align}
where the second term is a common term not depend on the changing policy $\pi_i$. Therefore, the second term will be cancelled in $ F^{\pi_i, -\pi_i}(\boldsymbol{l})-F^{\tilde{\pi}_i, -\pi_i}(\tilde{\boldsymbol{l}})=F_i^{\pi_i, -\pi_i}(\boldsymbol{l}_i)-F_i^{\tilde{\pi}_i, -\pi_i}(\tilde{\boldsymbol{l}}_i)$, thus finishes the proof.
\end{proof}

\begin{proposition}
\label{prop:vbf-proof}
Maximising the VBF is sufficient for minimising the makespan, subjective to the load balancing problem constraints (Eq.~\eqref{eq:lb_cons1} and \eqref{eq:lb_cons2}):
\begin{align}
    \max F(\boldsymbol{l}) \Rightarrow  \min \max_j(l_j)
\end{align}
this also holds for per-LB VBF as $\max F_i(\boldsymbol{l}_i) \Rightarrow  \min \max_j(\boldsymbol{l}_i)$.
\end{proposition}

\begin{proof}
Given the stability constraint in Eq.~\eqref{eq:lb_cons1} $\sum_{i=1}^{M}w_i(t)\leq\sum_{j=1}^{N}v_j$, we denote the total amount of workload in the system $C=\sum_{j=1}^{N}l_j$, and $l_k=\max_{j\in[N]}l_j$.
Based on the constraint in Eq.~\eqref{eq:lb_cons2}, we have $C \geq 0$, $l_j(t) \geq 0$.
\begin{align}
\max F(\boldsymbol{l}) &\Leftrightarrow \min -F(\boldsymbol{l})\\
-F(\boldsymbol{l}) &= \frac{1}{N}\sum_{j=1}^{N}((l_j)-\overline{\boldsymbol{l}})^2\\
&=\frac{1}{N}\sum_{j=1}^{N}(l_j - \frac{C}{N})^2\\
&=\frac{1}{N}\sum_{j=1}^{N}l^2_j - \frac{2C}{N^2}\sum_{j=1}^{N}l_j+\frac{C^2}{N^2}\\
&=\frac{1}{N}\sum_{j=1}^{N}l^2_j-\frac{C^2}{N^2}\\
&\le [(\max_j l_j)^2-\frac{C^2}{N^2}] \quad (\text{by means inequality})
\end{align}
with the equivalence achieved when $l_j=l_k, \forall j\neq k, j \in [N]$ holds.
Therefore,
\begin{align}
    \max F(\boldsymbol{l}) &\Rightarrow \min (l_k)^2-\frac{C^2}{N^2}  \\
    & \Leftrightarrow \min l_k\\
    &\Leftrightarrow \min \max_{j\in[n]} l_j
\end{align}
and the condition is sufficient but not necessary because $\min (l_k)^2-\frac{C^2}{N^2}$ is essentially minimizing the upper bound of $-F(\boldsymbol{l})$.
\end{proof}

\paragraph{Analysis of Product-Based Fairness}
\label{sec:app_pbf}
\begin{proposition}
Maximising the product-based fairness is sufficient for minimising the makespan, subjective to the load balancing problem constraints (Eq.~\eqref{eq:lb_cons1} and \eqref{eq:lb_cons2}):
\begin{align}
    \max F(\boldsymbol{l}) \Rightarrow  \min \max(\boldsymbol{l})
\end{align}
\end{proposition}

\begin{proof}
For a vector of workloads $\boldsymbol{l}=[l_1, \dots, l_N]$ on each server $j\in[N]$, by the definition of fairness, 
\begin{align}
    \max F(\boldsymbol{l}) &= \max \frac{\prod_{j\in[N]} l_j}{\max_{k\prime\in[N]} l_{k^\prime}}
\end{align}
WLOG, let $l_k = \max_{k^\prime\in[N]} l_{k^\prime}$, then,
\begin{align}
     \max F(\boldsymbol{l}) = \max \prod_{j\in[N], j \neq k} l_j
\end{align}
Similar to the proof of Proposition~\ref{prop:vbf-proof}, given the stability constraint in Eq.~\eqref{eq:lb_cons1} $\sum_{i=1}^{M}w_i(t)\leq\sum_{j=1}^{N}v_j$, we denote the total amount of workload in the system $C=\sum_{j=1}^{N}l_j$.
Based on the constraint in Eq.~\eqref{eq:lb_cons2}, we have $C \geq 0$, $l_j(t) \geq 0$.
By means inequality,
\begin{align}
    \left(\prod_{j\in[N], j \neq k}l_j\right)^{\frac{1}{N-1}} \leq \frac{\sum_{j\in[N], j \neq k}l_j}{N-1} = \frac{C-l_k}{N-1}.
\end{align}
with the equivalence achieved when $l_i=l_j, \forall i,j\neq k, i,j\in[N]$ holds.
Therefore,
\begin{align}
    \max F(\boldsymbol{l}) &\Rightarrow \max \frac{C-l_k}{N-1} \\
    & \Leftrightarrow \min l_k\\
    &\Leftrightarrow \min \max_{j\in[N]} l_j
\end{align}
The inverse may not hold since $\max \frac{C-l_k}{N-1}$ does not indicates $\max F(\boldsymbol{l})$, so maximising the linear product-based fairness is sufficient but not necessary for minimising the makespan. This finishes the proof.
\end{proof}

\paragraph{Variance-Based Fairness for MPG}
\label{sec:app_vbf_mpg}
\begin{theorem}
Multi-agent load balancing is MPG with the VBF $F_i(\boldsymbol{l}_i)$ as the reward $r_i$ for each LB agent $i\in[M]$, then suppose for $\forall s\in \mathcal{S}$ at step $h\in[H]$, the potential function is time-cumulative total fairness: $\phi^{\pi_i, -\pi_i}(s)=\sum_{t =h}^H F^{\pi_i, -\pi_i}(\boldsymbol{l}(t))$.
\end{theorem}
\begin{proof}
\begin{align}
    V_i^{\pi_i, \pi_{-i}}(s) - V_i^{\tilde{\pi}_i, \pi_{-i}}(s)&=\mathbb{E}_{\pi_i, \pi_{-i}}\bigg[\sum_{t =
        h}^H r_{i, t}(s_{t}, \boldsymbol{a}_{t}) \bigg| s_h = s\bigg] - \mathbb{E}_{\tilde{\pi}_i, \pi_{-i}}\bigg[\sum_{t =
        h}^H r_{i, t}(s_{t}, \tilde{a}_{i, t}, a_{-i, t}) \bigg| s_h = s\bigg]\\
        &=\mathbb{E}_{{\pi}_i, \pi_{-i}}\bigg[\sum_{t =
        h}^H F_i(\boldsymbol{l}_i(t))\bigg] - \mathbb{E}_{\tilde{\pi}_i, \pi_{-i}}\bigg[\sum_{t =
        h}^H F_i(\tilde{\boldsymbol{l}}_i(t))\bigg]\\
        &=\sum_{t =
        h}^H \bigg(F^{\pi_i, -\pi_i}(\boldsymbol{l})-F^{\tilde{\pi}_i, -\pi_i}(\tilde{\boldsymbol{l}})\bigg) \quad (\text{Lemma \ref{lem:vbf}})\\
        &=\phi^{\pi_i, -\pi_i}(s)-\phi^{\tilde{\pi}_i, -\pi_i}(s)
\end{align}
Notice that $s$ is the ground truth state of the environment, therefore involving the expected time $\boldsymbol{l}$ to finish remaining jobs.
\end{proof}

\begin{lemma} NE for MPG is $\epsilon$-approximate NE for $\epsilon$-approximate MPG.~\cite{ali2019reinforcement}

\begin{proof}
We know NE $(\pi^*_i, \pi^*_{-i})$ for MPG,
\begin{align}
    V_i^{\pi^*_i, \pi^*_{-i}}(s) - V_i^{\tilde{\pi}_i, \pi^*_{-i}}(s)=\phi^{\pi^*_i, {\pi}^*_{-i}}(s)-\phi^{\tilde{\pi}_i, \pi^*_{-i}}(s) \ge 0
\end{align}
the policies can be $\epsilon$-approximate NE for another game with a different value function $\widehat{V}$ but the same potential function,
\begin{align}
    \widehat{V}_i^{\pi^*_i, \pi^*_{-i}}(s) - \widehat{V}_i^{\tilde{\pi}_i, \pi^*_{-i}}(s) \ge \epsilon, \forall i\in[N], \tilde{\pi}_i\in\Pi_i, s\in\mathcal{S}
\end{align}
thus,
\begin{align}
   \bigg|\bigg( \widehat{V}_i^{\pi^*_i, \pi^*_{-i}}(s) - \widehat{V}_i^{\tilde{\pi}_i, \pi^*_{-i}}(s) \bigg)-\bigg( \phi^{\pi^*, \pi^*_{-i}}(s)-\phi^{\tilde{\pi}, \pi^*_{-i}}(s)\bigg)\bigg| \le \epsilon
\end{align}
which satisfies the definition of $\epsilon$-approximate MPG.
\end{proof}
\end{lemma}

\part{Reinforcement Learning in Foundation Models} \label{part2:rl_in_foundation_models}
\chapter{Diffusion World Model\label{ch:diffusion_world_model}}
\begin{center}
\begin{quote}
This section is based on paper ``\textit{Diffusion World Model: Future Modeling Beyond Step-by-Step Rollout for Offline Reinforcement Learning}''~\cite{ding2024diffusion} written in collaboration with Amy Zhang, Yuandong Tian and Qinqing Zheng, previously published at ICLR 2024 GenAI4DM Workshop.
\end{quote}
\end{center}

\section{Introduction}
\label{sec:intro}

\begin{wrapfigure}[9]{r}{0.5\textwidth}
    \centering
    \includegraphics[width=0.5\columnwidth]{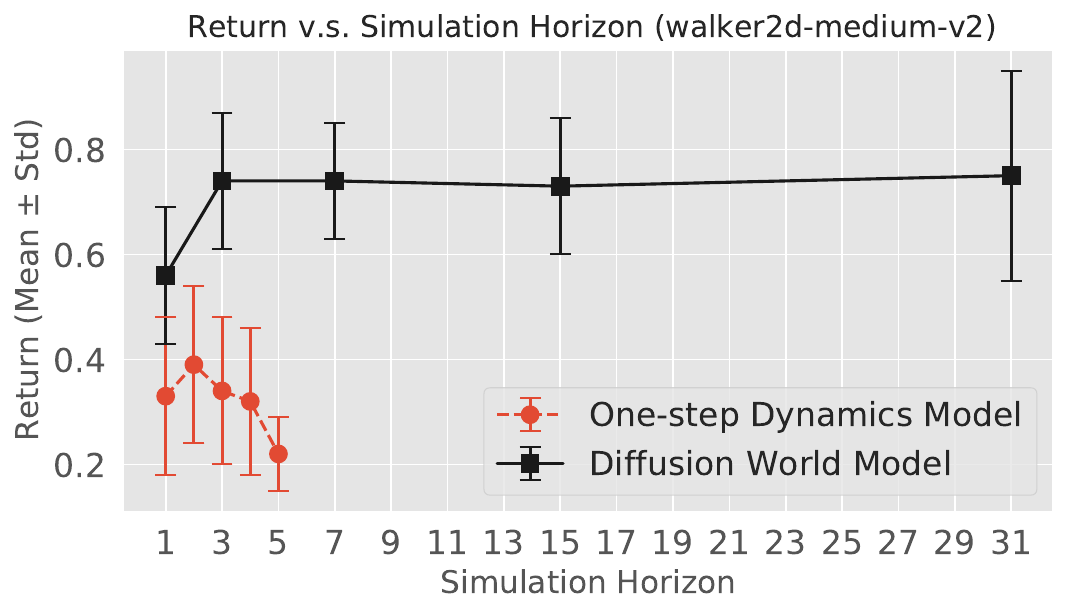}
    \caption{\small The return of TD3+BC trained using diffusion world model and one-step dynamics model.}
    \label{fig:compounding_error_comparison}
\end{wrapfigure}

World models are foundational blocks of AI systems to perform planning and reasoning~\cite{ha2018world}. They serve as simulators of real environments that predict the future outcome of certain actions will produce, and policies can be derived from them. Representative example usages of them in model-based reinforcement learning (MBRL) include action searching~\cite{schrittwieser2020mastering, ye2021mastering}, policy optimization within such simulators~\cite{dean2020sample, feinberg2018model, hafner2019dream, sutton1991dyna}, or a combination of both~\cite{chitnis2023iql, hansen2023td, hansen2022temporal}.

The prediction accuracy of a world model is critical to the final performance of model-based RL approaches.
Traditional MB methods builds a one-step dynamics model $p_{\text{one}}(s_{t+1}, r_t | s_t, a_t)$ that predicts reward $r_t$ and next state $s_{t+1}$ based on the current state $s_t$ and the current action $a_t$~\cite{hafner2019dream, hafner2020mastering, hafner2023mastering, hansen2022modem, janner2019trust, kaiser2019model,  kidambi2020morel, yu2020mopo}. When planning for multiple steps into the future, $p_\text{one}$ is recursively invoked, leading to a rapid accumulation of errors and unreliable predictions for long-horizon rollouts.
Figure~\ref{fig:compounding_error_comparison} plots the performance of an MB approach with a one-step dynamics model. The return quickly collapses as the rollout length increases, highlighting the issue of \emph{compounding errors} for such models~\cite{asadi2019combating, lambert2022investigating, xiao2019learning}.  

Recently, there has been growing interest in utilizing sequence modeling techniques to solve decision making problems, as seen in various studies~\cite{ajay2022conditional, chen2021decision, janner2022planning, janner2021offline,  micheli2022transformers, robine2023transformer, zheng2023guided,  zheng2022online}. Under this theme, a number of works have proposed Transformer based~\cite{chen2022transdreamer, micheli2022transformers, robine2023transformer} or diffusion model based~\cite{alonso2024diffusion, janner2022planning, lu2023synthetic,rigter2023world, yang2023learning, zhang2023learning} dynamics models, or closely related approaches. As we will review at the end of this section, as well as in Section~\ref{sec:related},
while most existing approaches leverage these sequence models as dynamics models for planning, they model one-step future outcome $s_{t+1}$ and $r_t$ using information of current and previous steps. At planning time, they still plan step by step. 
This raises an intriguing question that our paper seeks to answer:
\begin{quote}
    \emph{Can sequence modeling tools effectively reduce the compounding error in long-horizon prediction via jointly predicting multiple steps into the future?}
\end{quote}
In this paper, we introduce \textbf{\emph{Diffusion World Model}} (DWM). Conditioning on current state $s_t$, action $a_t$, and expected return $g_t$, DWM simultaneously predicts \textit{multistep future states and rewards}. Namely, it models $p_\theta(r_{t}, s_{t+1}, r_{t+1}, \ldots, s_{t+T-1}, r_{t+T-1} | s_t, a_t, g_t)$ where $T$ is the sequence length of the diffusion model.  For planning $H~(H < T)$  steps into the future, DWM only needs to be called once, whereas the traditional one-step model $p_\text{one}$ needs to be invoked $H$ times. This greatly reduces the source of compounding error. As illustrated in Figure~\ref{fig:compounding_error_comparison}, diffusion world model is robust to long-horizon simulation, where the performance does not deteriorate even with simulation horizon $31$. \looseness=-1

To verify the proposed DWM, we consider the \textbf{offline RL} setup, where the objective is to learn a policy from a static dataset without online interactions. The detachment from online training circumvents the side effects of exploration and allows us to investigate the quality of world models thoroughly. We propose a generic Dyna-type~\cite{sutton1991dyna} model-based framework. In brief, 
we first train a diffusion world model using the offline dataset, then train a policy using imagined data generated by the diffusion world model, in an actor-critic manner. Particularly, to generate the target value for training the critic, we introduce \emph{\textbf{Diffusion Model Based Value Expansion} (Diffusion-MVE)} that uses diffusion world model generated future trajectories to simulate the return up to a chosen horizon. As we will elaborate later, \emph{Diffusion-MVE can be interpreted as a value regularization for offline RL through generative modeling, or alternatively, a way to conduct offline Q-learning with synthetic data}.  Our framework is flexible to carry any MF actor-critic RL method of choice, and the output policy is efficient at inference time, as the world model does not intervene with action generation. \looseness=-1

\begin{table}[t]
    \centering
    \scriptsize
    \setlength{\tabcolsep}{2pt}
    \begingroup
    \everymath{\scriptstyle}
    \begin{tabularx}{\textwidth}{@{}>{\raggedright\arraybackslash}p{2.2cm}>{\centering\arraybackslash}p{1.9cm} >{\raggedright\arraybackslash\hsize=1.35\hsize}X >{\raggedright\arraybackslash\hsize=0.65\hsize}X >{\raggedright\arraybackslash}p{2.5cm}@{}}
    \toprule
      Method   &  RL Setup & Diffusion Model & Model Usages & Action Prediction\\
      \midrule
         SynthER~\cite{lu2023synthetic} & Offline/Online  & $p(s_t, a_t, s_{t+1}, r_t)$ & transition-level data augmentation & MF methods\\
         \midrule
         DWMs~\cite{alonso2024diffusion} & Online & $p(s_{t+1}|s_t, a_t, \ldots, s_{t-T+1}, a_{t-T+1})$ & step-by-step planning & REINFORCE~\cite{williams1992simple}\\
         \midrule
         PolyGrad~\cite{rigter2023world} & Online& $p(r_t, s_{t+1}, \ldots, s_{t+T-1}, r_{t+T-1} | s_t, a_t, \ldots, a_{t+T-1})$ & generate on-policy trajectories for policy optimization & stochastic Langevine dynamics\\
         \midrule
         PGD~\cite{jackson2024policy} & Offline & $p(s_t, a_t, r_t, \ldots, s_{t+T-1}, a_{t+T-1}, r_{t+T-1}, s_{t+T})$ & generate on-policy trajectories with policy gradient guidance for data augmentation & MF offline methods\\
         \midrule
         UniSim$^*$~\cite{yang2023learning} & Offline & $p(s_{t+1} | s_t, a_t)$  & step-by-step planning & REINFORCE~\cite{williams1992simple} \\
         \midrule
         Diffuser~\cite{janner2022planning} & Offline & $p(a_t, \ldots, s_{t+T-1}, a_{t+T-1}|s_t)$ & extract $a_t$ from the sample & extract from the sample \\
         \midrule
         DD~\cite{ajay2022conditional} &Offline &$p(s_{t+1}, \ldots, s_{t+T-1} | s_t, g_t)$  & extract  $s_{t+1}$ from the sample & inverse dynamics model\\
         \midrule
         \textbf{DWM (ours)} &  Offline &$p(r_t, s_{t+1}, r_{t+1}, \ldots, s_{t+T-1}, r_{t+T-1} | s_t, a_t, g_t)$ & multistep planning & MF offline methods\\
    \bottomrule
    \multicolumn{5}{@{}>{\raggedright\arraybackslash}p{\dimexpr\textwidth-2\tabcolsep}@{}}{\footnotesize *The observation of UniSim might contain multiple frames, yet the nature of their diffusion model is still a one-step model.} \\
    \end{tabularx}
    \endgroup
    \caption{A comparison of representative diffusion-model based MBRL methods. }
    \label{tab:diffusion_mbrl_comparison}
\end{table}

Empirically, we benchmark diffusion-based and traditional one-step world models on 9 locomotion tasks from the D4RL datasets~\cite{fu2020d4rl}, where all the tasks are in continuous action and observation spaces.
The predominant results are:
\begin{enumerate}[leftmargin=1em]
    \item Our results confirm that DWM outperform one-step models, where DWM-based algorithms achieves a $44\%$ performance gain. 
    \item We further consider a variant of our approach where the diffusion model is substituted with a Transformer architecture~\cite{vaswani2017attention}. Although Transformer is a sequence model, its inherent autoregressive structure is more prone to compounding error. We confirm that DWM-based algorithms surpass Transformer-based algorithms with a $37.5\%$ performance gain. 
    \item We also compare our algorithm with Decision Diffuser~\cite{ajay2022conditional}, a closely related model-based offline RL method that simulates the state-only trajectory, while predicting actions using an inverse dynamics model. The performance of the two methods are comparable. 

    \item 
    Meanwhile, due to inevitable modeling error, MB methods typically exhibit worse final performance compared with their model-free (MF) counterparts that directly learn policies from interacting with the true environment. 
    Our results show that DWM-based MB algorithms is comparable to or even slightly outperforming its MF counterparts. We believe this stimulates us to conduct research in the space of model-based RL approaches, which come with an advantage of sample efficiency~\cite{dean2020sample, deisenroth2013survey} and thus are potentially more suitable for practical real-world problems.
\end{enumerate}

\paragraph{Key Differences with Other Diffusion-Based Offline RL Methods}
More recently, various forms of diffusion models like~\cite{ajay2022conditional, alonso2024diffusion, jackson2024policy, janner2022planning, lu2023synthetic, rigter2023world, yang2023learning, zhang2023learning} 
have been introduced for world modeling and related works. These works have targeted different data setups (offline or online RL), and utilize diffusion models to model different types of data distributions. When applying to downstream RL tasks, they also have distinct ways to derive a policy.  While Section~\ref{sec:related} will review them in details, we summarize our key distinctions from these works in Table~\ref{tab:diffusion_mbrl_comparison}.

\section{Related Work}
\label{sec:related}
\paragraph{Model-Based RL}
One popular MB technique is action searching.
Using the world model, one simulates the outcomes of candidate actions, which are sampled from proposal distributions or policy priors~\cite{nagabandi2018neural, williams2015model}, and search for the optimal one. This type of approaches
has been successfully applied to games like Atari and Go~\cite{schrittwieser2020mastering, ye2021mastering} and continuous control problems with pixel observations~\cite{hafner2019learning}.
Alternatively, we can optimize the policy through interactions with the world model. This idea originally comes from the Dyna algorithm~\cite{sutton1991dyna}. The primary differences between works in this regime lie in their usages of the model-generated data. For example, Dyna-Q~\cite{sutton1990integrated} and MBPO~\cite{janner2019trust} augment the true environment data by world model generated transitions, and then conduct MF algorithms on either augmented or generated dataset. \citet{feinberg2018model} proposes to improve the value estimation by unrolling the policy within the world model up to a certain horizon. The Dreamer series of work~\cite{hafner2019dream, hafner2020mastering, hafner2023mastering} use the rollout data for both value estimation and policy learning. More recently, \citet{hansen2023td, hansen2022temporal, chitnis2023iql} combine both techniques to solve continuous control problems. As we cannot go over all the MB approaches, we refer readers to \citet{wang2019benchmarking, amos2021model} for more comprehensive review and benchmarks of them. 

Most of the aforementioned approaches rely on simple one-step world models $p_\text{one}(s_{t+1}, r_t|s_t, a_t)$. The Dreamer series of work~\cite{hafner2019dream, hafner2020mastering, hafner2023mastering} use recurrent neural networks (RNN) to engage in past information for predicting the next state.
Lately, \citet{robine2023transformer, micheli2022transformers, chen2022transdreamer} have independently proposed Transformer-based world models as a replacement of RNN. \citet{janner2020gamma} uses a generative model to learn the occupancy measure over future states, which can perform long-horizon rollout with a single forward pass.
\looseness=-1

\paragraph{Offline RL} 
Directly applying online RL methods to offline RL usually leads to poor performance. The failures are typically attributed to the extrapolation error~\cite{fujimoto2019off}.
To address this issue, a number of conservatism notions have been introduced to encourage the policy to stay close to the offline data. For model-free methods, these notions are applied to the value functions~\cite{kumar2020conservative, kostrikov2021offline, garg2023extreme} or to the policies~\cite{wu2019behavior, jaques2019way, kumar2019stabilizing, fujimoto2021minimalist}. Conservatism has also been incorporated into MB techniques through modified MDPs. 
For instance, MOPO~\cite{yu2020mopo} builds upon MBPO and relabels the predicted reward when generating transitions. It subtracts the uncertainty of the world model's prediction from the predicted reward, thereby softly promoting state-action pairs with low-uncertainty outcome. 
In a similar vein, MOReL~\cite{kidambi2020morel} trains policies using a constructed pessimistic MDP with terminal state. 
The agent will be moved to the terminal state if the prediction uncertainty of the world model is high,
and will receive a negative reward as a penalty. 

\paragraph{Sequence Modeling for RL}
There is a surge of recent research interest in applying sequence modeling tools to RL problems.
\cite{chen2021decision, janner2021offline} first consider the offline trajectories as autoregressive sequences and model them using Transformer architectures~\cite{vaswani2017attention}. This has inspired a line of follow-up research, including~\cite{meng2021offline, lee2022multi}. Normalizing flows like diffusion model~\cite{ho2020denoising, sohl2015deep, song2020score}, flow matching~\cite{lipman2022flow} and consistency model~\cite{song2023consistency} have also been incorporated into various RL algorithms, see e.g., ~\cite{chi2023diffusion, ding2023consistency, du2023learning, hansen2023idql, jia2023chain, mishra2023reorientdiff, wang2022diffusion, xu2023controllable}. 
Several recent works have utilized the diffusion model (DM) for world modeling in a variety of ways. 
Here we discuss them and highlight
the key differences between our approach and theirs, see also Table~\ref{tab:diffusion_mbrl_comparison}. 
\citet{alonso2024diffusion} trains a DM-based one-step dynamics model, which predicts the next single state $s_{t+1}$, conditioning on past states $s_{t-T}, \ldots, s_t$ and actions $a_{t-T}, \ldots, a_t$. 
This concept is similarly applied in UniSim~\cite{yang2023learning}. In essence, these models still plan step by step while incorporating information from preivous steps, whereas our model plans multiple future steps at once.  Similarly, \citet{zhang2023learning} trains a discretized DM with masked and noisy input. Despite still predicting step by step at inference time, this work mainly focuses on prediction tasks and does not conduct RL experiments. 
SynthER~\cite{lu2023synthetic} is in the same spirit as MBPO~\cite{janner2019trust}, which models the collected transition-level data distribution via an unconditioned diffusion model, and augments the training dataset by its samples. We focus on simulating the future trajectory for enhancing the model-based value estimation, and our diffusion model is conditioning on $s_t$ and $a_t$.
PolyGRAD~\cite{rigter2023world} learns a DM to predict a sequence of future states $s_{t+1}, \ldots, s_{t+T-1}$  and rewards $r_t, \ldots, r_{t+T-1}$, conditioning on the initial state $s_t$ and corresponding actions $a_{t}, \ldots, a_{t+T-1}$. Given that the actions are also unknown, PolyGRAD alternates between predicting the actions (via stochastic Langevin dynamics using policy score) and denoising the states and rewards during the DM's sampling process. This approach results in generating on-policy trajectories. In contrast, our approach is off-policy, since it does not interact with the policy during the sampling process. 
Policy-guided diffusion (PGD)~\cite{jackson2024policy} shares the same intention as PolyGrad, which is to generate on-policy trajectories. To achieve this, it trains an unconditioned DM using the offline dataset but samples from it under the (classifier) guidance of policy-gradient.
Diffuser~\cite{janner2022planning} and Decision Diffuser (DD)~\cite{ajay2022conditional} are most close to our work, as they also predict future trajectories. However, the modeling details and the usage of generated trajectories significantly differs. Diffuser trains an unconditioned model that predicts both states and actions, resulting in a policy that uses the generated next action directly. DD models state-only future trajectories conditioning on $s_t$, while we model future states and rewards conditioning on $s_t$ and $a_t$. DD predicts the action by an inverse dynamics model given current state and predicted next state, hence the diffusion model needs to be invoked at inference time. Our approach, instead, can connect with any MF offline RL methods that is fast to execute for inference.

\section{Preliminaries}
\label{sec:preliminary}
\paragraph{Offline RL.}
We consider an infinite-horizon Markov decision process (MDP) defined by $(\mathcal{S}, \mathcal{A}, R, P, p_0, \gamma)$,
where $\mathcal{S}$ is the state space, $\mathcal{A}$ is the action space. Let $\Delta(\mathcal{S})$ be the probability simplex of
the state space. $R: \mathcal{S} \times \mathcal{A} \mapsto \R$ is a deterministic reward function,
$P: \mathcal{S} \times \mathcal{A} \mapsto \Delta(\mathcal{S})$ defines the probability distribution of transition, $p_0: \mathcal{S} \mapsto \Delta(\mathcal{S})$ defines the distribution of initial state $s_0$,
and $\gamma \in (0,1)$ is the discount function. The task of RL is to learn a policy $\pi: \mathcal{S} \mapsto \mathcal{A}$ 
that maximizes its return $J(\pi) = \E_{s_0 \sim p_0(s), a_t \sim \pi(\cdot | s_t), s_{t+1} \sim P(\cdot|s_t, a_t) } \left[ \sum_{t=0}^\infty \gamma^t R(s_t, a_t) \right]$.
Given a trajectory $\tau=\set{s_0, a_0, r_0, \ldots, s_{|\tau|}, a_{|\tau|}, r_{|\tau|}}$, where $|\tau|$ is the total number of timesteps, the return-to-go (RTG) at timestep
$t$ is $g_t = \sum_{t'=t}^{|\tau|} \gamma^{t'-t} r_{t'}$.
In offline RL, we are constrained to learn a policy solely from a static dataset generated by certain unknown policies. Throughout this paper,
we use 
$\mathcal{D}_\text{offline}$ to denote the offline data distribution and use
$D_\text{offline}$ to denote the offline dataset. \looseness=-1

\paragraph{Diffusion Model.}
Diffusion probabilistic models~\cite{ho2020denoising, sohl2015deep, song2020score} are generative models that create samples from noises by an iterative denoising process. It defines a fixed Markov chain, called the \emph{forward} or \emph{diffusion process}, that iteratively adds Gaussian noise to $x^{(k)}$ starting from a data point $x^{(0)}$:
   $x^{(k+1)}|x^{(k)} \sim \mathcal{N}\left(\sqrt{1 - \beta_{k}}x^{(k)}, \beta_{k}\mathbf{I}\right), \; 0 \leq k \leq K-1.$
As the number of diffusion steps $K \rightarrow \infty$, $x^{(K)}$ essentially becomes a random noise. We learn the corresponding \emph{reverse process} that transforms random noise to data point:
    $x^{(k-1)}|x^{(k)} \sim \mathcal{N}\left(\mu_\theta(x^{(k)}), \Sigma_\theta(x^{(k)}) \right), \; 1 \leq k \leq K.$
Sampling from a diffusion model amounts to first sampling a random noise $x^{(K)} \sim \mathcal{N}(0, \mathbf{I})$ then running the reverse process. 
Let $\varphi(z;\mu, \Sigma)$ denote the density function of a random variable $z \sim \mathcal{N}(\mu, \Sigma)$. To learn the reverse process, we parameterize
    $p_\theta(x^{(k-1)}|x^{(k)}) = \varphi\left(x^{(k-1)};\mu_\theta(x^{(k)}), \Sigma_\theta(x^{(k)}) \right), \; 1 \leq k \leq K,$
and optimize the variational lower bound of the marginal likelihood $p_\theta(x^{(0:K)})$.
There are multiple equivalent ways to optimize the lower bound~\cite{kingma2021variational}, 
and we take the noise prediction route as follows. One can rewrite 
$ x^{(k)} = \sqrt{\bar{\alpha}_k} x^{(0)} + \sqrt{1 - \bar{\alpha}_k} \eps$,
where $\bar{\alpha}_k = \prod_{k'=1}^K (1 -\beta_{k'})$, and $\eps \sim \mathcal{N}(0, \mathbf{I})$ is the noise injected for $x^{(k)}$ (before scaling). We then parameterize a neural network $\eps_\theta(x^{(k)}, k)$ to predict $\eps$ injected for $x^{(k)}$. Moreover, a conditional variable $y$ can be easily added into both processes via formulating the corresponding density functions $q(x^{(k+1)}|x^{(k)}, y)$ and $p_\theta(x^{(k-1)}|x^{(k)}, y)$, respectively. We further deploy classifier-free guidance~\cite{ho2022classifier} to promote the conditional information, which essentially learns both conditioned and unconditioned noise predictors. More precisely, we optimize the following loss function:\looseness=-1
\begin{equation}
    \E_{(x^{(0)}, y), k, \eps, b } \norm{ \eps_\theta\big(x^{(k)}(x^{(0)}, \eps), k, (1-b)\cdot y + b \cdot \varnothing \big) - \eps }^2_2,
\label{eq:diffusion_loss}
\end{equation}
where $x^{(0)}$ and $y$ are the true data point and conditional information sampled from data distribution,
$\eps \sim \mathcal{N}(0, \mathbf{I})$ is the injected noise,
$k$ is the diffusion step sampled uniformly between $1$ and $K$,
$b \sim \text{Bernoulli}(p_\text{uncond})$ is used to indicate whether we will use null condition,
and finally, $x^{(k)} = \sqrt{\bar{\alpha}_k} x^{(0)} + \sqrt{1 - \bar{\alpha}_k} \eps$.
Algorithm~\ref{algo:diffusion_sampling_general} details how to sample from a guided diffusion model. In section~\ref{sec:method}, we shall introduce the form of $x^{(0)}$ and $y$ in the context of offline RL, and discuss how we utilize diffusion models to ease planning. \looseness=-1

\section{Diffusion World Model}
\label{sec:method}

In this section, we introduce a general recipe for model-based offline RL with diffusion world model. 
Our framework consists of two training stages, which we will detail in Section~\ref{sec:method_diffusion} and~\ref{sec:method_mbrl}, respectively.
In the first stage, we train a diffusion model to predict a sequence of future states and rewards, conditioning on the current 
state, action and target return. Next, we train an offline policy using an actor-critic method, where we utilize the pretrained diffusion model for model-based value estimation.
Algorithm~\ref{algo:dwm_training}-\ref{algo:diffusion_mbrl} presents this framework with a simple actor-critic algorithm with delayed updates, where we assume a deterministic offline policy. 
Our framework can be easily extended in a variety of ways.
First, we can generalize it to account for stochastic policies. 
Moreover, the actor-critic algorithm we present is of the simplest form. It can be extended to combine with various existing offline learning algorithms. In Section~\ref{sec:expr}, we discuss three instantiations of
Algorithm~\ref{algo:diffusion_mbrl}, which embeds TD3+BC~\cite{fujimoto2021minimalist}, IQL~\cite{kostrikov2021offline} and Q-learning with pessimistic reward~\cite{yu2020mopo} respectively. \looseness=-1

\begin{algorithm}[t]
\caption{Diffusion World Model Training}
\label{algo:dwm_training}
\begin{algorithmic}[1]
\STATE \textbf{Hyperparameters}: number of diffusion steps $K$, null conditioning probability $p_\text{uncond}$, noise parameters $\bar{\alpha}_k$, $k\in[K]$.
\WHILE{not converged}
    \STATE Sample a length-$T$ subtrajectory from $D_\text{offline}$: $x^{(0)} \leftarrow (s_t, a_t, r_t, s_{t+1}, r_{t+1}, \ldots, s_{t+T-1}, r_{t+T-1})$.
    \STATE Compute RTG $g_t \leftarrow \sum_{h=0}^{T-1} \gamma^{h} r_{t+h}$.
    \STATE {\color{blue}\% optimize DWM via Eq.~\eqref{eq:diffusion_loss}}
    \STATE Sample $\eps \sim \mathcal{N}(0, \mathbf{I})$ and $k \in [K]$ uniformly.
    \STATE Compute $x^{(k)} \leftarrow \sqrt{\bar{\alpha}_k} x^{(0)} + \sqrt{1 - \bar{\alpha}_k}\, \eps$.
    \STATE Set $y \leftarrow \varnothing$ with probability $p_\text{uncond}$, otherwise $y \leftarrow g_t$.
    \STATE Take gradient step on $\nabla_\theta \big\| \eps_\theta(x^{(0)}, k, y) - \eps \big\|^2_2$.
\ENDWHILE
\STATE \textbf{Return}: diffusion world model $p_\theta$.
\end{algorithmic}
\end{algorithm}

\begin{algorithm}[t]
\caption{General Actor-Critic Framework for Offline Model-Based RL with DWM}
\label{algo:diffusion_mbrl}
\begin{algorithmic}[1]
\STATE \textbf{Input}: pretrained diffusion world model $p_\theta$.
\STATE \textbf{Hyperparameters}: rollout length $H$, conditioning RTG $g_\text{eval}$, guidance parameter $\omega$, target network update frequency $n$.
\STATE Initialize actor and critic networks $\pi_\psi$, $Q_\phi$.
\STATE Initialize target network weights: $\bar{\psi} \leftarrow \psi$, $\bar{\phi} \leftarrow \phi$.
\WHILE{not converged}
    \STATE Sample state-action pair $(s_t, a_t)$ from $D_\text{offline}$.
    \STATE {\color{blue}\% diffusion model value expansion}
    \STATE Sample $\hat{r}_t, \hat{s}_{t+1}, \hat{r}_{t+1}, \ldots, \hat{s}_{t+T-1}, \hat{r}_{t+T-1} \sim p_\theta(\cdot | s_t, a_t, g_\text{eval})$ with guidance parameter $\omega$.
    \STATE Compute the target $Q$ value:
    $$
    y \leftarrow \sum_{h=0}^{H-1} \gamma^{h} \hat{r}_{t+h} + \gamma^H Q_{\bar{\phi}}(\hat{s}_{t+H}, \pi_{\bar{\psi}}(\hat{s}_{t+H}))
    $$
    \STATE {\color{blue}\% update the critic}
    \STATE Update the critic: $\phi \leftarrow \phi - \eta \nabla_\phi \norm{Q_\phi(s_t, a_t) - y}^2_2$
    \STATE {\color{blue}\% update the actor}
    \STATE Update the actor: $\psi \leftarrow \psi  + \eta \nabla_\psi Q_{\phi}(s_t, \pi_\psi(s_t))$
    \STATE {\color{blue}\% update the target networks}
    \IF{iteration $\mathrm{mod}~n=0$}
        \STATE $\bar{\phi} \leftarrow \bar{\phi} + w (\phi - \bar{\phi})$
        \STATE $\bar{\psi} \leftarrow \bar{\psi} + w (\psi - \bar{\psi})$
    \ENDIF
\ENDWHILE
\end{algorithmic}
\end{algorithm}

\subsection{Conditional Diffusion Model}
\label{sec:method_diffusion}
We train a return-conditioned diffusion model $p_\theta(x^{(0)}|s_t, a_t, y)$
on length-$T$ subtrajectories,  where the conditioning variable is the RTG of a subtrajectory. That is, $y=g_t$ and $x^{(0)} = (r_t, s_{t+1}, r_{t+1}, \ldots, s_{t+T-1}, r_{t+T-1})$. 
As introduced in Section~\ref{sec:preliminary}, we employ classifier-free guidance to promote the role of RTG. 
Stage 1 of Algorithm~\ref{algo:diffusion_mbrl} describes
the training procedure in detail.
For the actual usage of the trained diffusion model in the second stage of our pipeline, 
we predict future $T-1$ states and rewards based on a target RTG $g_\text{eval}$ and also current state $s_t$ and action $a_t$. These predicted states and rewards are used to facilitate the value estimation in policy training, see Section~\ref{sec:method_mbrl}.
As the future actions are not needed, we do not model them in our world model.
To enable the conditioning of $s_t$ and $a_t$, we slightly adjust the standard sampling procedure (Algorithm~\ref{algo:diffusion_sampling_general}),
where we fix $s_t$ and $a_t$ as conditioning for every denoising step in the reverse process, see Algorithm~\ref{algo:diffusion_sampling_rl}.
\looseness=-1

\subsection{Model-Based RL with Diffusion World Model}
\label{sec:method_mbrl}
As summarized in Algorithm~\ref{algo:diffusion_mbrl}, we propose an actor-critic algorithm where the critic is trained on synthetic data generated by the diffusion model, and the actor is then trained with policy evaluation based on the critic. In a nutshell, we estimate the $Q$-value by the sum of a short-term return, simulated by the DWM, and a long-return value, estimated by a proxy state-action value function $\hat{Q}$ learned through temporal difference (TD) learning. It is worth noting that in our framework, DWM only intervenes the critic training, and Algorithm~\ref{algo:diffusion_mbrl} is general to connect with any MF value-based algorithms. We shall present 3 different instantiations of it in Section~\ref{sec:expr}.
\begin{definition}[\textbf{$H$-step Diffusion Model Value Expansion}]
\label{def:dmve}
Let $(s_t, a_t)$ be a state-action pair. Sample $\hat{r}_t, \hat{s}_{t+1}$$, \hat{r}_{t+1}, \ldots,$$ \hat{s}_{t+T-1},\hat{r}_{t+T-1}$ from the diffusion model $p_\theta(\cdot|s_t, a_t, g_\text{eval})$. Let $H$ be the simulation horizon, where $H < T$. The $H$-step \emph{diffusion model value expansion} estimate of the value of $(s_t, a_t)$ is given by
\begin{equation}
    \hat{Q}_\text{diff}^H(s_t, a_t) = \mathbb{E}_{p_\theta(\cdot|s_t, a_t, g_\text{eval})}\left[\textstyle \sum_{h=0}^{H-1} \gamma^{h} \hat{r}_{t+h} + \gamma^H \hat{Q}(\hat{s}_{t+H}, \hat{a}_{t+H})\right],
\end{equation}
where $\hat{a}_{t+H} = \pi(\hat{s}_{t+H})$ and $\hat{Q}(\hat{s}_{t+H}, \hat{a}_{t+H})$ is the proxy value for the final state-action pair. \looseness=-1
\end{definition}
We employ this expansion to compute the target value in TD learning, see Algorithm~\ref{algo:diffusion_mbrl}. 
This mechanism is key to the success of our algorithm and has several appealing properties. \looseness=-1
\begin{enumerate}[leftmargin=1em]
    \item In deploying the standard model-based value expansion~(MVE, \citet{feinberg2018model}),
 the imagined trajectory is derived by recursively querying the one-step dynamics model $p_\text{one}(s_{t+1}, r_t | s_t, a_t)$, which is the root cause of error accumulation. As an advantage over MVE, our DWM generates the imagined trajectory (without actions) as a whole. 
    \item More interestingly, MVE uses the policy predicted action $\hat{a}_t = \pi(\hat{s}_t)$ when querying $p_\text{one}$. This can be viewed as an on-policy value estimation of $\pi$ in a simulated environment. In contrast, Diffusion-MVE operates in an off-policy manner, as $\pi$ does not influence the sampling process. As we will explore in Section~\ref{sec:expr}, the off-policy diffusion-MVE excels in offline RL, significantly surpassing the performance of one-step-MVE. We will now delve into two interpretations of this, each from a unique perspective. 
\end{enumerate}
\paragraph{(a)} Our approach can be viewed as a policy iteration algorithm, alternating between policy evaluation (Algorithm~\ref{algo:diffusion_mbrl} line 7-9) and policy improvement (line 10) steps. Here, $\hat{Q}^H_\text{diff}$ is the estimator of the policy value function $Q^\pi$, with adjustable lookahead horizon $H$ and pessimistic or optimistic estimation through changing $g_\text{eval}$. Parameterized $Q_\phi$ is optimized towards the target $\hat{Q}^H_\text{diff}$ through a mean squared error.
In the context of offline RL, TD learning often lead to overestimation of $Q^\pi$~\cite{thrun2014issues, kumar2020conservative}. This is because $\pi$ might produce out-of-distribution actions, leading to erroneous values for $\hat{Q}$, and the policy is defined to maximize $\hat{Q}$. Such overestimation negatively impacts the generalization capability of the resulting policy when it is deployed online. To mitigate this, a broad spectrum of offline RL methods 
apply various forms of regularization to the value function~\cite{garg2023extreme, kostrikov2021offline, kumar2020conservative}, to ensure the resulting policy remains close to the data. As the DWM is trained exclusively on offline data, it can be seen as a synthesis of the behavior policy that generates the offline dataset. In other words, diffusion-MVE introduces a type of \emph{\textbf{value regularization for offline RL through generative modeling}}. 

Moreover, our approach significantly differs from existing value pessimism notions. One challenge of offline RL is that the behavior policy that generates the offline dataset is often of low-to-moderate quality, 
so that the resulting dataset might only contain trajectories with low-to-moderate returns. As a result, many regularization techniques introduced for offline RL are often \emph{overly pessimistic}~\cite{ghasemipour2022so, nakamoto2023cal}. To address this issue, we typically condition on large out-of-distribution (OOD) values of $g_\text{eval}$ when sampling from the DWM. Putting differently, we ask the DWM to output an imagined trajectory under an \emph{optimistic goal}.  

\paragraph{(b)} Alternatively, we can also view the approach as an offline Q-learning algorithm~\cite{watkins1992q}, where $\hat{Q}$ is estimating the optimal value function $Q^*$ using off-policy data. Again, the off-policy data is generated by the diffusion model, conditioning on OOD RTG values. In essence, our approach can be characterized as \emph{\textbf{offline $Q$-learning on synthetic data}}.

\paragraph{Comparison with Transformer-based World Models.}  Curious readers may wonder about the key
distinctions between DMW and existing Transformer-based world models~\cite{chen2022transdreamer, micheli2022transformers, robine2023transformer}.
These models, given the current state $s_t$ and action $a_t$,
leverage the autoregressive structure of Transformer to incorporate past information to predict $s_{t+1}$. 
To forecast multiple steps into the future, they must make iterated predictions. 
In contrast, DWM makes long-horizon predictions in a single query. It is worth noting that it is entirely possible
to substitute the diffusion model in our work with a Transformer, and we justify our design choice in Section~\ref{sec:expr_ablation}.\looseness=-1

\section{Experiments}
\label{sec:expr}

Our experiments are designed to answer the following questions. (1) Compared with the one-step dynamics model, does DWM effectively reduces the compounding error and lead to better performance in MBRL? (2) How does the proposed Algorithm~\ref{algo:diffusion_mbrl} compare with other diffusion model model-based methods, and (3) their model-free counterparts?

To answer these questions, we consider three instantiations of Algorithm~\ref{algo:diffusion_mbrl}, where we integrate TD3+BC~\cite{fujimoto2021minimalist}, IQL~\cite{kostrikov2021offline} and Q-learning with pessimistic reward (which we refer to as PQL) as the offline RL algorithm in the second stage. These algorithms come with different conservatism notions defined on the action (TD3+BC), the value function (IQL), and the reward (PQL), respectively.  Specifically, the PQL algorithm is inspired by the MOPO algorithm~\cite{yu2020mopo}, where we penalize the world model predicted reward by the uncertainty of its prediction. Nonetheless, it is distinct from MOPO in the critic learning. MOPO uses standard TD learning on model-generated transitions, whereas we use MVE or Diff-MVE for value estimation. In the sequel, we refer to our algorithms as DWM-TD3BC, DWM-IQL and DWM-PQL respectively. For DWM-IQL, we have observed performance enhancement using a variant of Diff-MVE based on the $\lambda$-return technique~\cite{schulman2015high}, therefore we incorporate it as a default feature. Detailed descriptions of these algorithms are deferred to Sec.~\ref{sec:app_instantiations}. We present the comparisons in Section~\ref{sec:expr_dwm_vs_onestep}-\ref{sec:expr_dwm_vs_mf}, and ablate the design choices we made for DWM in Section~\ref{sec:expr_ablation}. In Sec.~\ref{subsec:add_baselines}, we conduct experimental comparison with additional baselines including data augmentation~\cite{lu2023synthetic} and autoregressive diffusion~\cite{rigter2023world} methods. 

\paragraph{Benchmark and Hyperparameters.} 
We conduct experiments on 9 datasets of locomotion tasks from the D4RL~\cite{fu2020d4rl} benchmark, and report the obtained normalized return (0-1 with 1 as expert performance). Throughout the paper, we train each algorithm for 5 instances with different random seeds, and evaluate them for 10 episodes. All reported values are means and standard deviations aggregated over 5 random seeds. 
We set the sequence length of DWM to be $T=8$ (discussed in Section~\ref{sec:expr_dwm_vs_onestep}). The number of diffusion steps is $K=5$ for training. For DWM inference, an accelerated inference technique is applied with a reduced number of diffusion steps $N=3$, as detailed in Section~\ref{sec:expr_ablation}.
The training and sampling details of DWM refer to Sec.~\ref{subsec:app_diffusion_model}, and the training details of each offline algorithm refer to Sec.~\ref{subsec:app_mf_mb_details}. We further conduct extensive experiments on sparse-reward tasks, and results are detailed in Sec.~\ref{sec:add_sparse_reward}.

\subsection{DWM v.s. One-Step Dynamics Model}
\label{sec:expr_dwm_vs_onestep}
We first investigate the effectiveness of DWM in reducing the compounding error for MBRL, and compare it with the counterparts using one-step dynamics model. Next, we evaluate the performance of our proposed Algorithm~\ref{algo:diffusion_mbrl} and the one-step dynamics model counterparts, where we substitute DWM by one-step dynamics models and use standard MVE. We call these baselines OneStep-TD3BC, OneStep-IQL and OneStep-PQL, correspondingly. 

\begin{figure}[t]
     \centering
    \includegraphics[width=1\columnwidth]{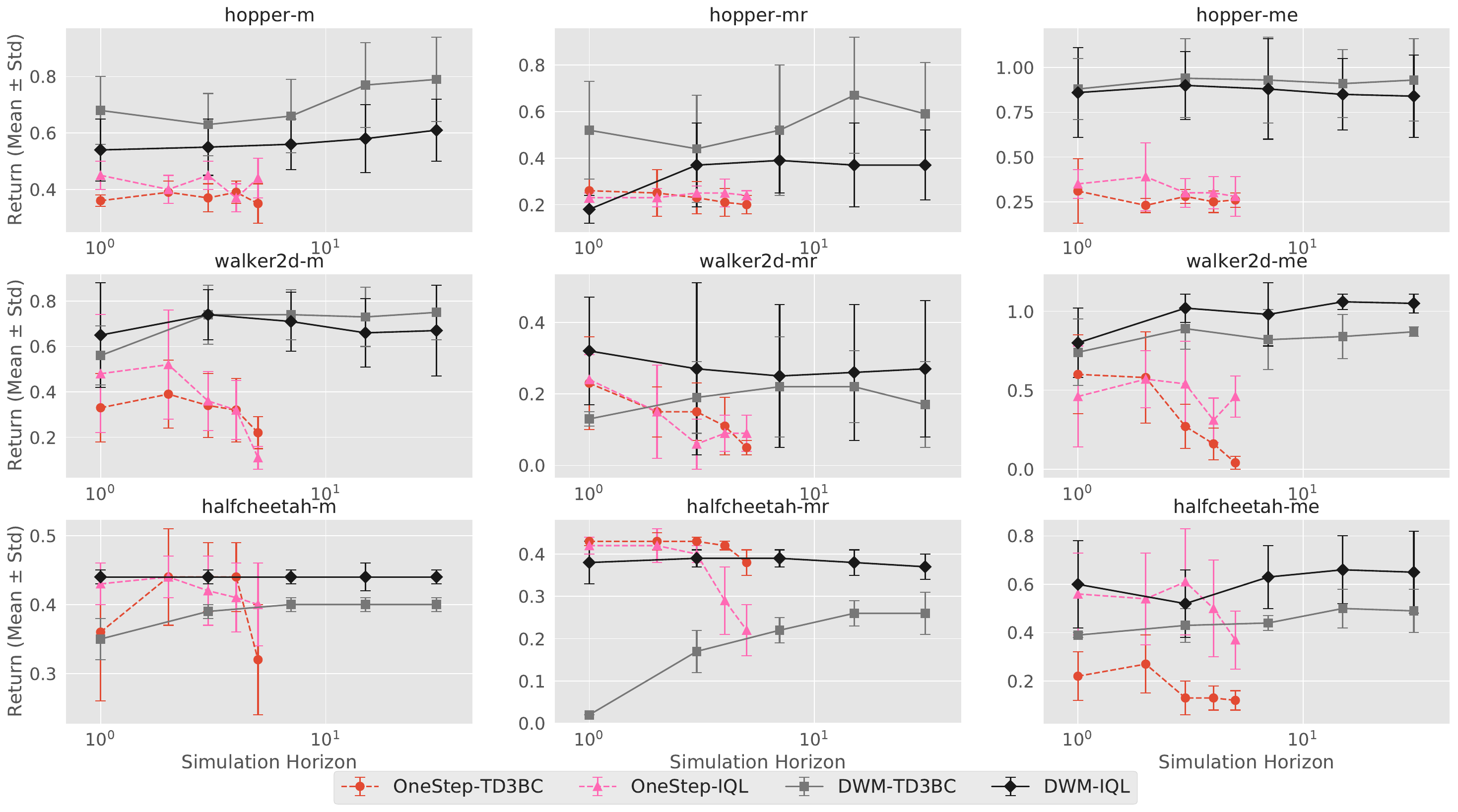}
    \caption{Performances of Algorithm~\ref{algo:diffusion_mbrl} with DWM and one-step models, using different simulation horizons. 
    The x-axis has range $[1, 31]$ in a logarithm scale.}
    \label{fig:one_step_compare}
\end{figure}
\paragraph{Long Horizon Planning and Compounding Error Comparison.} To explore the response of different world models to long simulation horizons, we compare the performance
DWM methods (DWM-TD3BC and DWM-IQL) with their one-step counterparts (OneStep-TD3BC and OneStep-IQL) when the simulation horizon $H$ used in policy training changes.
To explore the limit of DWM models, we train another set of DWMs with longer sequence length $T=32$ and investigate the performance of downstream RL algorithms for $H\in\{1,3,7,15,31\}$.
The algorithms with one-step dynamics models have simulation horizon from 1 to 5. Figure~\ref{fig:one_step_compare} plots the results across 9 tasks.
OneStep-IQL and OneStep-TD3BC exhibit a clearly performance drop as the simulation horizon increases. For most tasks,
their performances peak with relatively short simulation horizons, like one or two. This suggests that longer model-based rollout with one-step dynamics models suffer from severe compounding errors. On the contrary, DWM-TD3BC and DWM-IQL maintain relatively high returns without significant performance degradation, even using horizon length 31.
Note that in the final result Table~\ref{tab:compare_mbrl}, we report results using DWM with sequence length $T=8$, because the performance gain of using $T=32$ is marginal. See Sec.~\ref{app:seq_length_dwm} for details. We additionally conduct experiments on analyzing the compounding error for DWM and one-step model predictions. The results in Sec.~\ref{subsec:compounding} indicate the superior performance of DWM in reducing the compounding errors, which verifies our hypothesis. \looseness=-1

\paragraph{Offline RL Performance.} Table~\ref{tab:compare_mbrl} reports the performance of
Algorithm~\ref{algo:diffusion_mbrl} using DWM and one-step dynamics models on the D4RL datasets. We sweep over the simulation horizon $H\in\{1,3,5,7\}$ and a set of evaluation RTG values. The RTG values we search vary across environments, see Table~\ref{tab:hyperparam}. The predominant trends we found are: \emph{the proposed DWM significantly outperforms the one-step counterparts, with a notable $44\%$ performance gain.} 
This is attributed to the strong expressivity of diffusion models and the prediction of entire sequences all at once, which circumvents the compounding error issue in multistep rollout. This point will be further discussed in the studies of simulation horizon as next paragraph.

\begin{table*}[t]
\centering
\resizebox{\textwidth}{!}{
\begin{tabular}{c|ccc|ccc}
\toprule
 Env.  &  OneStep-TD3BC & OneStep-IQL & OneStep-PQL & DWM-TD3BC & DWM-IQL & DWM-PQL  \\  \midrule
hopper-m  &  0.39 $\pm$ 0.04 & 0.45 $\pm$ 0.05 & 0.63 $\pm$ 0.12 & \textbf{0.65 $\pm$ 0.10} & 0.54 $\pm$ 0.11 & 0.50 $\pm$ 0.09  \\
walker2d-m  &  0.39 $\pm$ 0.15 & 0.52 $\pm$ 0.24 & 0.74 $\pm$ 0.14 & 0.70 $\pm$ 0.15 & 0.76 $\pm$ 0.05 & \textbf{0.79 $\pm$ 0.08}\\
halfcheetah-m &  0.44 $\pm$ 0.05 & 0.44 $\pm$ 0.03 & 0.45 $\pm$ 0.01 & \textbf{0.46 $\pm$ 0.01} & 0.44 $\pm$ 0.01 & 0.44 $\pm$ 0.01   \\
hopper-mr  &  0.26 $\pm$ 0.05 & 0.25 $\pm$ 0.03 & 0.32 $\pm$ 0.03 & 0.53 $\pm$ 0.09 & \textbf{0.61 $\pm$ 0.13} & 0.39 $\pm$ 0.03\\
walker2d-mr &  0.23 $\pm$ 0.13 & 0.24 $\pm$ 0.07 & 0.62 $\pm$ 0.22 & 0.46 $\pm$ 0.19 & 0.35 $\pm$ 0.14 & 0.35 $\pm$ 0.13 \\
halfcheetah-mr & \textbf{0.43 $\pm$ 0.01} & 0.42 $\pm$ 0.02 & 0.42 $\pm$ 0.01 & \textbf{0.43 $\pm$ 0.01} & 0.41 $\pm$ 0.01 & \textbf{0.43 $\pm$ 0.01} \\
hopper-me &  0.31 $\pm$ 0.18 & 0.39 $\pm$ 0.19 & 0.43 $\pm$ 0.18 & \textbf{1.03 $\pm$ 0.14} & 0.90 $\pm$ 0.25 & 0.80 $\pm$ 0.18 \\
walker2d-me &  0.60 $\pm$ 0.25 & 0.57 $\pm$ 0.18 & 0.61 $\pm$ 0.22 & \textbf{1.10 $\pm$ 0.00} & 1.04 $\pm$ 0.10 & \textbf{1.10 $\pm$ 0.01} \\
halfcheetah-me &  0.27 $\pm$ 0.12 & 0.61 $\pm$ 0.22 & 0.61 $\pm$ 0.22 & \textbf{0.75 $\pm$ 0.16} & 0.71 $\pm$ 0.14 & 0.69 $\pm$ 0.13 \\ \hline
\multirow{2}{*}{Average} &  0.368 $\pm$ 0.105 & 0.432 $\pm$ 0.115 & 0.537 $\pm$ 0.128 & \textbf{0.679 $\pm$ 0.098} & 0.641 $\pm$ 0.117 & 0.610 $\pm$ 0.080   \\
\cline{2-7}
& \multicolumn{3}{c|}{0.446$\pm$0.116} & \multicolumn{3}{c}{\textbf{0.643$\pm$0.07}} \\\bottomrule
\end{tabular}
}
\caption{\small Comparison of MB methods with one-step model versus DWM on the D4RL dataset.}
\label{tab:compare_mbrl}
\end{table*}

\subsection{DWM v.s. Decision Diffuser}
\label{sec:expr_dwm_vs_other_mb}

We further compare DWM-TD3BC (the best-performing DWM-based algorithms) with Decision Diffuser (DD)~\cite{ajay2022conditional}, another closely related approach that also use diffusion models to model the trajectory in the offline dataset.
As noted in Section~\ref{sec:intro}, our approach is significantly different from theirs, though. DWM conditions on both state $s_t$ and action $a_t$, where DD only conditions on $s_t$. More importantly, we train a downstream model-free policy using imagined rollout, whereas DD predicts the action via an inverse dynamics model, using current state $s_t$ and predicted next state $\hat{s}_{t+1}$. This means, at inference time, DD needs to generate the whole trajectory, which is computationally inefficient. On the contrary, DWM based approaches are efficient as we do not need to sample from the trained DWM anymore. Table~\ref{tab:compare_diffusion_dd} reports the performance and the inference time of DWM-TD3BC and DD. The inference time is averaged over $600$ evaluation episodes. The performance of DWM-TD3BC is comparable to Decision Diffuser, and it enjoys $4.6$x faster inference speed. We anticipate the difference in speed amplifies for higher dimensional problems.
\begin{table*}
    \centering
\resizebox{\textwidth}{!}{
\small
\begin{tabular}{c|cc | cccc}
\toprule
 Env.  &  \multicolumn{2}{c|}{DD} & \multicolumn{2}{c}{DWM-TD3BC}  \\  
  & Normalized Return & Inference Time (sec) & Normalized Return & Inference Time (sec)\\
  \midrule
hopper-m  & 0.49 $\pm$ 0.07 & 4.11 $\pm$ 4.64 & \textbf{0.65 $\pm$ 0.10} & \textbf{1.18 $\pm$ 0.51}\\
walker2d-m & 0.67 $\pm$ 0.16  & 8.09 $\pm$ 1.24 & 0.70 $\pm$ 0.15 &  \textbf{2.00 $\pm$ 0.63} \\ 
halfcheetah-m & \textbf{0.49 $\pm$ 0.01}  &8.18 $\pm$ 3.77 & 0.46 $\pm$ 0.01 &  \textbf{1.81 $\pm$ 0.54} \\
hopper-mr  & \textbf{0.66 $\pm$ 0.15}  & 6.21 $\pm$ 4.21  & 0.53 $\pm$ 0.09 & \textbf{0.64 $\pm$ 0.52}\\
walker2d-mr  & 0.44 $\pm$ 0.26 &5.94 $\pm$ 4.32 & \textbf{0.46 $\pm$ 0.19} & \textbf{0.83 $\pm$ 0.45} \\
halfcheetah-mr  & 0.38 $\pm$ 0.06 &7.46 $\pm$ 9.72  & \textbf{0.43 $\pm$ 0.01} & \textbf{0.60 $\pm$ 0.17} \\
hopper-me  & \textbf{1.06 $\pm$ 0.11} & 8.82 $\pm$ 2.96 & 1.03$\pm$ 0.14 & \textbf{2.62 $\pm$ 1.22} \\
walker2d-me  & 0.99 $\pm$ 0.15  & 9.26 $\pm$ 1.30 &\textbf{1.10 $\pm$ 0.00}  & \textbf{3.60  $\pm$ 3.90} \\
halfcheetah-me  & \textbf{0.91 $\pm$ 0.01} &9.53 $\pm$ 2.50  & 0.75 $\pm$ 0.16 &  \textbf{3.77 $\pm$ 2.43} \\ \hline
{Average} & \textbf{0.677$\pm$0.109} & 7.531 $\pm$ 3.651 & \textbf{0.679$\pm$0.098} & \textbf{1.620 $\pm$ 1.379} \\
\bottomrule
\end{tabular}
}
\caption{\small The performance of DWM-TD3BC and Decision Diffuser (DD) are comparable, while DWM-TD3BC is $4.6$x faster than DD. }
\label{tab:compare_diffusion_dd}
\end{table*}

\subsection{DWM v.s. Model-Free Counterparts}
\label{sec:expr_dwm_vs_mf} 

Finally, we compare DWM-based algorithms with their MF counterparts, namely, TD3+BC vs DWM-TD3BC, and IQL vs DWM-IQL. Table~\ref{tab:dwm_vs_mf} reports the results. For each group of comparison, we highlight the winning performance. 

\begin{table}
\centering
\resizebox{0.8\textwidth}{!}{
\begin{tabular}{c|cc|cc}
\toprule
 Env.          &  TD3+BC & DWM-TD3BC                 & IQL  & DWM-IQL \\  \midrule
hopper-m       & 0.58 $\pm$ 0.11 &  \textbf{0.65 $\pm$ 0.10}  & 0.48 $\pm$ 0.08 & \textbf{0.54 $\pm$ 0.11} \\
walker2d-m     & \textbf{0.77 $\pm$ 0.09} &  0.70 $\pm$ 0.15  & 0.75 $\pm$ 0.15 & \textbf{0.76 $\pm$ 0.05} \\
halfcheetah-m  & \textbf{0.47 $\pm$ 0.01} &  0.46 $\pm$ 0.01 & \textbf{0.46 $\pm$ 0.07}  & 0.44 $\pm$ 0.01\\
hopper-mr      & 0.53 $\pm$ 0.19 & \textbf{0.53 $\pm$ 0.09}  & 0.25 $\pm$ 0.02& \textbf{0.61 $\pm$ 0.13}\\
walker2d-mr    & \textbf{0.75 $\pm$ 0.19} & 0.46 $\pm$ 0.19&  \textbf{0.48 $\pm$ 0.23}  & 0.35 $\pm$ 0.14 \\
halfcheetah-mr & \textbf{0.43 $\pm$ 0.01}  & \textbf{0.43 $\pm$ 0.01}& \textbf{0.44 $\pm$ 0.01}  & 0.41 $\pm$ 0.01\\
hopper-me      & 0.90 $\pm$ 0.28  & \textbf{1.03 $\pm$ 0.14} & 0.86 $\pm$ 0.22& \textbf{0.90 $\pm$ 0.25}\\
walker2d-me    & 1.08 $\pm$ 0.01  & \textbf{1.10 $\pm$ 0.00} & \textbf{1.09 $\pm$ 0.00} & 1.04 $\pm$ 0.10 \\
halfcheetah-me & 0.73 $\pm$ 0.16  &\textbf{0.75 $\pm$ 0.16} & 0.60 $\pm$ 0.23  & \textbf{0.71 $\pm$ 0.14}\\ \hline
Average        & \textbf{0.693 $\pm$ 0.116} & 0.679 $\pm$ 0.098 & 0.601 $\pm$ 0.112  & \textbf{0.641 $\pm$ 0.117}\\
\bottomrule
\end{tabular}
}
\caption{\small Performance of DWM methods and its MF counterparts.}
\label{tab:dwm_vs_mf}
\end{table}
We can see DWM-IQL outperforms IQL, DWM-TD3BC is comparable to TD3+BC. 
Different from MF algorithms with ground-truth samples, MB algorithms inevitably suffers additional modeling errors from approximating the dynamics. It is worth noting that MB algorithms using traditional one-step dynamics model significantly underperforms the MF counterparts (results reported in Table~\ref{tab:compare_mbrl}), while DWM alleviates the downside of dynamics modeling through reducing compounding errors. \looseness=-1

\subsection{Ablation Studies}
\label{sec:expr_ablation}

\paragraph{Diffusion Model v.s. Transformer.}
Algorithm~\ref{algo:diffusion_mbrl} is capable of accommodating various types of sequence models, including
Transformer~\cite{vaswani2017attention}, one of the most successful sequence models.
However, analogous to the compounding error issue for one-step dynamics model, 
Transformer is subject to inherent error accumulation due to its autoregressive structure.
Therefore, we hypothesize Transformer will underperform and choose diffusion model.
\begin{table}
\centering
\resizebox{0.8\textwidth}{!}{
\begin{tabular}{c|cc|cc}
\toprule
\small
 Env. & \multicolumn{1}{c}{\textbf{T-TD3BC}} & \multicolumn{1}{c|}{\textbf{T-IQL}} & \multicolumn{1}{c}{\textbf{DWM-TD3BC}} & \multicolumn{1}{c}{\textbf{DWM-IQL}} \\ \midrule  
hopper-m  & 0.58 $\pm$ 0.08 & 0.55 $\pm$ 0.08 & \textbf{0.65 $\pm$ 0.10} & 0.54 $\pm$ 0.11 \\
walker2d-m & 0.60 $\pm$ 0.16 & 0.72 $\pm$ 0.12  & 0.70 $\pm$ 0.15 &  \textbf{0.76 $\pm$ 0.05}\\
halfcheetah-m & 0.42 $\pm$ 0.03 & 0.43 $\pm$ 0.01  & \textbf{0.46 $\pm$ 0.01} & 0.44 $\pm$ 0.01    \\
hopper-mr & 0.25 $\pm$ 0.06 & 0.26 $\pm$ 0.09  & 0.53 $\pm$ 0.09 & \textbf{0.61 $\pm$ 0.13} \\
walker2d-mr & 0.13 $\pm$ 0.06 & 0.23 $\pm$ 0.12 & \textbf{0.46 $\pm$ 0.19} & 0.35 $\pm$ 0.14  \\
halfcheetah-mr & 0.40 $\pm$ 0.01 & 0.39 $\pm$ 0.01 & \textbf{0.43 $\pm$ 0.01} & 0.41 $\pm$ 0.01 \\
hopper-me & 0.66 $\pm$ 0.25 & 0.62 $\pm$ 0.16  & \textbf{1.03$\pm$ 0.14} & 0.90 $\pm$ 0.25  \\
walker2d-me & 0.58 $\pm$ 0.15 & 1.03 $\pm$ 0.09 & \textbf{1.10 $\pm$ 0.00} & 1.04 $\pm$ 0.10\\
halfcheetah-me & 0.36 $\pm$ 0.17 & 0.44 $\pm$ 0.08 & \textbf{0.75 $\pm$ 0.16} & 0.71 $\pm$ 0.14  \\ \hline
Avg. & 0.442$\pm$0.101 &  0.519$\pm$0.084 & \textbf{0.679 $\pm$ 0.098} & 0.641 $\pm$ 0.117   \\
\bottomrule
\end{tabular}
}
\caption{\small Performance of Algorithm~\ref{algo:diffusion_mbrl} using DWM and Transformer-based world models. }
\label{tab:rollout_compare_dt}
\end{table}
To verify this hypothesis, we replace the diffusion model with Transformer in our proposed algorithms,
and compare the resulting performance with DWM methods. We particularly consider the combination with
TD3+BC and IQL, where we call the obtained algorithms T-TD3BC and T-IQL.
We test T-TD3BC and T-IQL with parameter sweeping over simulation horizon $H\in\{1,3,5,7\}$, as the same as DWM methods. For the evaluation RTG, we take the value used in Decision Transformer~\cite{chen2021decision} and apply the same normalization as used for DWM.
As shown in Table~\ref{tab:rollout_compare_dt}, DWM consistently outperforms Transformer-based world models across offline RL algorithm instantiations and environments. The experiment details refer to Sec.~\ref{subsec:app_transformer}. \looseness=-1

\paragraph{Additional Ablation Experiments.} Sec.~\ref{app:diffusion_steps} discusses the number of diffusion steps we use in training DWM and trajectory sampling. Sec.~\ref{app:ood_rtg} discusses the evaluation RTG values we use when sampling from the DWM. Sec.~\ref{app:lambda_return} ablates the $\lambda$-return technique we incorporate for DWM-IQL. Last, we also investigate the effects of fine-tuning DWM with relabelled RTGs~\cite{yamagata2023q}. We have found this technique is of limited utility, so we did not include it in the final design for the simplicity of our algorithm. See the results in Sec.~\ref{app:rtg_relabel}.


\section{Implementation Details}
\subsection{Diffusion World Model}
\label{subsec:app_diffusion_model}
We summarize the architecture and hyperparameters used for our experiments. 
For all the experiments, we use our own PyTorch implementation that is heavily influenced by the following codebases: 

Decision Diffuser~\cite{ajay2022conditional} \hskip5pt \url{https://github.com/anuragajay/decision-diffuser} \\
Diffuser~\cite{janner2022planning} \hskip5pt \url{https://github.com/jannerm/diffuser/}\\
SSORL~\cite{zheng2023semi} \hskip5pt \url{https://github.com/facebookresearch/ssorl/}


\paragraph{Architecture.}
As introduced in Section~\ref{sec:method_diffusion}, the diffusion world model $p_\theta$ used in this paper is chosen to model a length-$T$
subtrajecotires $(s_t, a_t, r_t, s_{t+1}, r_{t+1}, \dots, s_{t+T-1}, r_{t+T-1})$. At inference time,
it predicts the subsequent subtrajecotry of $T-1$ steps, conditioning on initial state-action pair $(s_t, a_t)$ and target RTG $y=g_t$:
\begin{align}
    \hat{r}_t, \hat{s}_{t+1}, \hat{r}_{t+1}, \dots, \hat{s}_{t+T-1}, \hat{r} _{t+T-1} \sim p_\theta(\cdot| s_t, a_t, y=g_t).
\end{align}
There are two reasons we choose not to model future actions in the sequence. First, our proposed diffusion model value expansion (Definition~\ref{def:dmve}) does not require the action information for future steps. Second, previous work have found that modeling continuous action through diffusion is less accurate~\cite{ajay2022conditional}.

Throughout the paper, we train guided diffusion models for state-reward sequences of length $T=8$. The number of diffusion steps is $K=5$.
The probability of null conditioning $p_\text{uncond}$ is set to $0.25$, and the batch size is $64$.
We use the cosine noise schedule proposed by \citet{nichol2021improved}.
The discount factor is $\gamma=0.99$, and we normalize the discounted RTG by a task-specific reward scale, which is $400$ for Hopper, $550$ for Walker, and $1200$ for Halfcheetah tasks.

Following~\citet{ajay2022conditional}, our noise predictor $\eps_\theta$ is a temporal U-net~\cite{janner2022planning, ronneberger2015u} that consists of 6 repeated residual blocks, where each block consists of 2 temporal convolutions followed by the group norm~\cite{wu2018group}
and a final Mish nonlinearity activation~\cite{mish2019self}.
The diffusion step $k$ is first transformed to its sinusoidal position encoding and projected to a latent space via a 2-layer MLP, and the RTG value is transformed into its latent embedding via a 3-layer MLP. In our diffusion world model, the initial action $a_t$ as additional condition is also transformed into latent embedding via a 3-layer MLP, and further concatenated with the embeddings of the diffusion step and RTG. 

\paragraph{Optimization.} We optimize our model by the Adam optimizer with a learning rate $1 \times 10^{-4}$ for all the datasets. The final model parameter $\bar{\theta}$ we consider is an exponential moving average (EMA) of the obtained parameters over the course of training. For every $10$ iteration, we update 
$\bar{\theta} = \beta \bar{\theta} + (1- \beta) \theta$,
where the exponential decay parameter $\beta = 0.995$. We train the diffusion model for $2 \times 10^6$ iterations.

\paragraph{Sampling with Guidance.}
\label{appendix:sampling}
To sample from the diffusion model, we need to first sample a random noise $x{(K)} \sim \mathcal{N}(0, \mathbf{I})$ and then run the reverse process.
Algorithm~\ref{algo:diffusion_sampling_general} presents the general process of sampling from a diffusion model trained under classifier-free guidance.

In the context of offline RL, the diffusion world model generates future states and rewards based on the current state $s_t$, the current action $a_t$ and the target return $g_\text{eval}$, see Section~\ref{sec:method}. Therefore, the sampling process is slightly different from Algorithm~\ref{algo:diffusion_sampling_general}, as we need to constrain the initial state and initial action to be $s_t$ and $a_t$, respectively. The adapted algorithm is summarized in Algorithm~\ref{algo:diffusion_sampling_rl}.

Following \citet{ajay2022conditional}, we apply the low temperature sampling technique for diffusion models. The temperature is set to be $\alpha=0.5$ for sampling at each
diffusion step from Gaussian $\mathcal{N}(\hat{\mu}_\theta, \alpha^2\hat{\Sigma}_\theta)$, with $\hat{\mu}_\theta$ and $\hat{\Sigma}_\theta$ being the predicted mean and covariance.

\paragraph{Accelerated Inference.}
\label{subsec:app_acc}
Algorithm~\ref{algo:diffusion_sampling_general} and \ref{algo:diffusion_sampling_rl} run the full reverse process, 
Building on top of them, we further apply the stride sampling technique as in~\citet{nichol2021improved} to speed up sampling process.
Formally, in the full reverse process, we generates $x^{(k-1)}$ by $x^{(k)}$ one by one, from $k=K$ till $k=1$:
\begin{equation}
    x^{(k-1)} = \frac{\sqrt{\bar{\alpha}_{k-1}}\beta_k}{1-\bar{\alpha}_k}\hat{x}^{(0)} + \frac{\sqrt{\alpha_k}(1-\bar{\alpha}_{k-1})}{1-\bar{\alpha}_k}x^{(k)}+ \sigma_k\varepsilon, \;\; \varepsilon\sim\mathcal{N}(0, \mathbf{I}),
    \label{eq:diffusion_reverse_computing}
\end{equation}
where $\hat{x}^{(0)}$ is the prediction of the true data point of Algorithm~\ref{algo:diffusion_sampling_general}),
$\sigma_k = \sqrt{\dfrac{\beta_k(1 - \bar{\alpha}_{k-1})}{1 - \bar{\alpha}_k}}$ is the standard deviation of noise at step $k$ (line 8 of in Algorithm~\ref{algo:diffusion_sampling_general}). We note that $\bar{\alpha}_k = \prod_{k'=1}^K (1 -\beta_{k'})$ where the noise schedule $\set{\beta_k}_{k=1}^K$ is predefined, see Section~\ref{sec:preliminary} and Sec.~\ref{subsec:app_diffusion_model}.

Running a full reverse process amounts to evaluating Equation~\eqref{eq:diffusion_reverse_computing} for $K$ times, which is time consuming. 
To speed up sampling, we choose $N$ diffusion steps equally spaced between $1$ and $K$, namely, $\tau_1, \ldots, \tau_N$, where $\tau_N=K$. We then evaluate Equation~\eqref{eq:diffusion_reverse_computing} for the chosen steps $\tau_1, \ldots, \tau_N$.
This effectively reduces the inference time to approximately $N/K$ of the original. In our experiments, we train the diffusion model with $K=5$ diffusion steps and sample with $N=3$ inference steps, see Section~\ref{sec:expr_ablation} for a justification of this number.

\begin{algorithm}[t]
\caption{Sampling from Guided Diffusion Models}
\label{algo:diffusion_sampling_general}
\begin{algorithmic}[1]
\STATE \textbf{Input:} trained noise prediction model $\eps_\theta$, conditioning parameter $y$, guidance parameter $\omega$, number of diffusion steps $K$
\STATE Sample $x^{(K)} \sim \mathcal{N}(0, \mathbf{I})$
\FOR{$k = K,\ldots,1$}
    \STATE $\hat{\eps} \leftarrow \omega \cdot \eps_\theta(x^{(k)}, k, y) + (1-\omega) \cdot \eps_\theta(x^{(k)}, k, \varnothing)$
    \STATE \textcolor{blue}{\% estimate true data point $x^{(0)}$}
    \STATE $\hat{x}^{(0)} \leftarrow \dfrac{1}{\sqrt{\bar{\alpha}_k}}\big(x^{(k)} - \sqrt{1-\bar{\alpha}_k}\hat{\eps}\big)$
    \STATE \textcolor{blue}{\% sample from the posterior $q(x^{(k-1)} | x^{(k)}, x^{(0)})$}
    \STATE \textcolor{blue}{\% See Equation (6) and (7) of \citet{ho2020denoising}}
    \STATE $\hat{\mu} \leftarrow \dfrac{\sqrt{\bar{\alpha}_{k-1}}\beta_k}{1 - \bar{\alpha}_k} \hat{x}^{(0)} + \dfrac{\sqrt{\alpha_k}(1 - \bar{\alpha}_{k-1})}{1 - \bar{\alpha}_k}x^{(k)}$
    \STATE $\hat{\Sigma} \leftarrow \dfrac{\beta_k(1 - \bar{\alpha}_{k-1})}{1 - \bar{\alpha}_k}\mathbf{I}$
    \STATE Sample $x^{(k-1)} \sim \mathcal{N}(\hat{\mu}, \hat{\Sigma})$
\ENDFOR
\STATE \textbf{Output:} $x^{(0)}$
\end{algorithmic}
\end{algorithm}

\begin{algorithm}[t]
\caption{Diffusion World Model Sampling}
\label{algo:diffusion_sampling_rl}
\begin{algorithmic}[1]
\STATE \textbf{Input:} trained noise prediction model $\eps_\theta$, initial state $s_t$, initial action $a_t$, target return $g_\text{eval}$, guidance parameter $\omega$, number of diffusion steps $K$
\STATE Sample $x^{(K)} \sim \mathcal{N}(0, \mathbf{I})$
\STATE \textcolor{blue}{\% apply conditioning of $s_t$ and $a_t$}
\STATE $x^{(K)}[0:\text{dim}(s_t) + \text{dim}(a_t)] \leftarrow \text{concatenate}(s_t, a_t)$
\FOR{$k = K,\ldots,1$}
    \STATE $\hat{\eps} \leftarrow \omega \cdot \eps_\theta(x^{(k)}, k, g_\text{eval}) + (1-\omega) \cdot \eps_\theta(x^{(k)}, k, \varnothing)$
    \STATE \textcolor{blue}{\% estimate true data point $x^{(0)}$}
    \STATE $\hat{x}^{(0)} \leftarrow \dfrac{1}{\sqrt{\bar{\alpha}_k}}\big(x^{(k)} - \sqrt{1-\bar{\alpha}_k}\hat{\eps}\big)$
    \STATE \textcolor{blue}{\% Sample from the posterior distribution $q(x^{(k-1)} | x^{(k)}, x^{(0)})$}
    \STATE \textcolor{blue}{\% See Equation (6) and (7) of \citet{ho2020denoising}}
    \STATE $\hat{\mu} \leftarrow \dfrac{\sqrt{\bar{\alpha}_{k-1}} \beta_k}{1 - \bar{\alpha}_k} \hat{x}^{(0)} + \dfrac{\sqrt{\alpha_k}(1 - \bar{\alpha}_{k-1})}{1 - \bar{\alpha}_k}x^{(k)}$
    \STATE $\hat{\Sigma} \leftarrow \dfrac{\beta_k(1 - \bar{\alpha}_{k-1})}{1 - \bar{\alpha}_k}\mathbf{I}$
    \STATE Sample $x^{(k-1)} \sim \mathcal{N}(\hat{\mu}, \hat{\Sigma})$
    \STATE \textcolor{blue}{\% apply conditioning of $s_t$ and $a_t$}
    \STATE $x^{(k-1)}[0:\text{dim}(s_t) + \text{dim}(a_t)] \leftarrow \text{concatenate}(s_t, a_t)$
\ENDFOR
\STATE \textbf{Output:} $x^{(0)}$
\end{algorithmic}
\end{algorithm}

\subsection{One-step Dynamics Model}
\label{sec:app_onestep}
The traditional one-step dynamics model $f_\theta(s_{t+1},r_t|s_t,a_t)$ is typically represented by a parameterized probability distribution over the state and reward spaces, and optimized through log-likelihood maximization of the single-step transitions:
\begin{align}
    \max_\theta \mathbb{E}_{(s_t,a_t,r_t,s_{t+1}) \sim \mathcal{D}_\text{offline}}\left[\log f_\theta(s_{t+1},r_t|s_t,a_t) \right],
\end{align}
where $(s_t,a_t,r_t,s_{t+1})$ is sampled from the offline data distribution $\mathcal{D}_\text{offline}$.
As in \citet{kidambi2020morel}, we model $f_\theta$ as a Gaussian distribution $\mathcal{N}(\mu_\theta, \Sigma_\theta)$, where the mean $\mu_\theta$ and the diagonal covariance matrix $\Sigma_\theta$ are parameterized by two 4-layer MLP neural networks with 256 hidden units per layer. We use the ReLU activation function for hidden layers. The final layer of $\Sigma_\theta$ is activated by a SoftPlus function to ensure validity. 
We train the dynamics models for $1\times 10^6$ iterations, using the Adam optimizer with learning rate $1\times 10^{-4}$.

\section{Diffusion World Model Based Offline RL Methods}
\label{sec:app_instantiations}
In Section~\ref{sec:expr}, we consider 3 instantiations of Algorithm~\ref{algo:diffusion_mbrl} where we integrate TD3+BC, IQL, Q-learning with pessimistic reward (PQL) into our framework. These algorithms are specifically designed for offline RL, with \textit{conservatism} notions defined on actions (TD3+BC), value function (IQL), and rewards (PQL) respectively. In the sequel, we refer to our instantiations as DWM-TD3BC, DWM-IQL and DWM-PQL. The detailed implementation of them will be introduced below.

\subsection{DWM-TD3BC: TD3+BC with Diffusion World Model}

Building on top of the TD3 algorithm~\cite{fujimoto2018addressing}, TD3+BC~\cite{fujimoto2021minimalist} employs explicit behavior cloning regularization to learn a deterministic policy. The algorithm works as follows.

The critic training follows the TD3 algorithm exactly. We learn two critic networks $Q_{\phi_1}$ and $Q_{\phi_2}$ as double Q-learning~\cite{fujimoto2018addressing} through TD learning.
The target value in TD learning for a transition $(s,a, r, s')$ is given by:
\begin{equation}
    y = r + \gamma \min_{i\in\{1,2\}}Q_{\phi_i}\big(s', a'=\text{Clip}(\pi_{\bar{\psi}}(s')+\eps, -C, C)\big),
    \label{eq:target_value_td3bc}
\end{equation}
where $\eps \sim \mathcal{N}(0, \sigma^2I)$ is a random noise, $\pi_{\bar{\psi}}$ is the target policy and Clip($\cdot$) is an operator that bounds the value of the action vector to be within $[-C, C]$ for each dimension. Both $Q_{\phi_1}$ and $Q_{\phi_2}$ will regress into the target value $y$ by minimizing the mean squared error (MSE), which amounts to solving the following problem: 
\begin{equation}
    \min_{\phi_i} \mathbb{E}_{(s,a,r,s')\sim \mathcal{D}_\text{offline}}\left[ \left( y(r,s')- Q_{\phi_i}(s,a) \right)^2\right], \; i\in\{1,2\}.
 \label{eq:q_loss_td3bc}
\end{equation}
 
For training the policy, TD3+BC optimizes the following regularized problem:
\begin{align}
   \max_\psi \E_{(s,a)\sim \mathcal{D}_\text{offline} } \left[ \lambda Q_{\phi_1}(s,\pi_\psi(s)) - \norm{a-\pi_\psi(s)}^2 \right],
   \label{eq:td3bc_policy}
\end{align}
which $\mathcal{D}_\text{offline}$ is the offline data distribution. Without the behavior cloning regularization term $\norm{a-\pi_\psi(s)}^2$, the above problem reduces to the objective corresponding to the deterministic policy gradient theorem~\cite{silver2014deterministic}. Note that $\pi$ is always trying to maximize one fixed proxy value function.

Both the updates of target policy $\pi_{\bar{\psi}}$ and the target critic networks $Q_{\bar{\phi}_i}$ are delayed in TD3+BC. The whole algorithm is summarized in Algorithm~\ref{algo:mbtd3bc}.

\begin{algorithm}[t]
\caption{DWM-TD3BC}
\label{algo:mbtd3bc}
\begin{algorithmic}[1]
\STATE \textbf{Input:} offline dataset $\mathcal{D}_\text{offline}$, pretrained diffusion world model $p_\theta$, simulation horizon $H$, conditioning RTG $g_\text{eval}$, policy/target update frequency $n$, coefficient $\lambda$, action perturbation/clipping parameters: $\sigma$, $C$

\STATE Initialize actor and critic networks: $\pi_\psi$, $Q_{\phi_1}$, $Q_{\phi_2}$
\STATE Initialize target network weights: $\bar{\psi} \leftarrow \psi$, $\bar{\phi}_1 \leftarrow \phi_1$, $\bar{\phi}_2 \leftarrow \phi_2$
\FOR{$i=1,2,\ldots$ until convergence}
    \STATE Sample state-action pair $(s_t, a_t)$ from $\mathcal{D}_\text{offline}$
    \STATE {\color{blue}\% diffusion model value expansion}
    \STATE Sample $\hat{r}_t, \hat{s}_{t+1}, \hat{r}_{t+1}, \ldots, \hat{s}_{t+T-1}, \hat{r}_{t+T-1} \sim p_\theta(\cdot \mid s_t, a_t, g_\text{eval})$
    \STATE Sample $\eps \sim \mathcal{N}(0, \sigma^2 I)$
    \STATE $\hat{a}^\eps_{t+H} \leftarrow \text{Clip}(\pi_{\bar{\psi}}(\hat{s}_{t+H}) + \eps, -C, C)$
    \STATE Compute the target Q-value: $y = \sum_{h=0}^{H-1} \gamma^{h}\hat{r}_{t+h} + \gamma^H \min_{i\in\{1,2\}} Q_{\bar{\phi}_i}(\hat{s}_{t+H}, \hat{a}^\eps_{t+H})$
    \STATE {\color{blue}\% update critic networks}
    \STATE $\phi_1 \leftarrow \phi_1 - \eta \nabla_{\phi_1} \lVert Q_{\phi_1}(s_t, a_t) - y \rVert_2^2$
    \STATE $\phi_2 \leftarrow \phi_2 - \eta \nabla_{\phi_2} \lVert Q_{\phi_2}(s_t, a_t) - y \rVert_2^2$
    \STATE {\color{blue}\% update actor and target networks (delayed)}
    \IF{$i\ \text{mod}\ n = 0$}
        \STATE $\psi \leftarrow \psi + \eta \nabla_\psi\left( \lambda Q_{\phi_1}(s_t, \pi_\psi(s_t)) - \lVert a_t - \pi_\psi(s_t) \rVert^2 \right)$ \hfill {\color{blue}\% update actor}
        \STATE $\bar{\phi}_1 \leftarrow \bar{\phi}_1 + w\cdot (\phi - \bar{\phi}_1)$
        \STATE $\bar{\phi}_2 \leftarrow \bar{\phi}_2 + w\cdot (\phi - \bar{\phi}_2)$
        \STATE $\bar{\psi} \leftarrow \bar{\psi} + w\cdot (\psi - \bar{\psi})$
    \ENDIF
\ENDFOR
\STATE \textbf{Output:} $\pi_\psi$
\end{algorithmic}
\end{algorithm}

\subsection{DWM-IQL: IQL with Diffusion World Model}
IQL~\cite{kostrikov2021offline} applies pessimistic value estimation on offline dataset. In addition to the double Q functions used in TD3+BC,
IQL leverages an additional state-value function $V_\xi(s)$,
which is estimated through expectile regression:
\begin{align}
   \min_\xi \E_{(s,a)\sim \mathcal{D}_\text{offline}} \left[ 
   L^\tau 
   \left( 
   \min_{i\in\{1,2\}} Q_{\bar{\phi}_i}(s,a)-V_\xi(s) 
   \right)
   \right],
   \label{eq:iql_v}
\end{align}
where $L^\tau(u)=|\tau - \mathbbm{1}_{u<0}|u^2$ with hyperparameter $\tau\in(0.5, 1)$,
As $\tau \rightarrow 1$, $V_\xi(s)$ is essentially estimating the maximum value of $Q(s,a)$. This can be viewed as implicitly performing the policy improvement step, without an explicit policy. Using a hyperparameter $\tau < 1$ regularizes the value estimation (of an implicit policy) and thus mitigates the overestimation issue of $Q$ function. The $Q$ function is updated also using Eq.~\eqref{eq:q_loss_td3bc} but with target $y=r+\gamma V_\xi(s')$. Finally, given the $Q$ and the $V$ functions, the policy is extracted by Advantage Weighted Regression~\cite{peng2019advantage}, i.e., solving
\begin{align}
    \max_\psi \mathbb{E}_{(s,a)\sim\mathcal{D}_\text{offline}}\left[\exp \left( \beta (Q_\phi(s,a)-V_\xi(s))\right) \log \pi_\psi(a|s) \right].
    \label{eq:iql_policy}
\end{align}
The update of the target critic networks $Q_{\bar{\phi}_i}$ are delayed in IQL. The whole algorithm is summarzied in Algorithm~\ref{algo:mbiql}.

\begin{algorithm}[t]
\caption{DWM-IQL}
\label{algo:mbiql}
\begin{algorithmic}[1]
\STATE \textbf{Inputs:} offline dataset $\mathcal{D}_\text{offline}$, pretrained diffusion world model $p_\theta$, simulation horizon $H$, conditioning RTG $g_\text{eval}$, target network update frequency $n$, expectile loss parameter $\tau$
\STATE Initialize actor, critic, and value networks $\pi_\psi$, $Q_{\phi_1}$, $Q_{\phi_2}$, $V_\xi$
\STATE Initialize target networks: $\bar{\phi}_1 \leftarrow \phi_1$, $\bar{\phi}_2 \leftarrow \phi_2$
\FOR{iteration $i = 1, 2, \ldots$ until convergence}
    \STATE Sample state-action pair $(s_t, a_t)$ from $\mathcal{D}_\text{offline}$
    \STATE {\color{blue}\% diffusion model value expansion}
    \STATE Sample $\hat{r}_t, \hat{s}_{t+1}, \hat{r}_{t+1}, \ldots, \hat{s}_{t+T-1}, \hat{r}_{t+T-1} \sim p_\theta(\cdot \mid s_t, a_t, g_\text{eval})$
    \STATE Compute target Q-value: $y = \sum_{h=0}^{H-1} \gamma^{h} \hat{r}_{t+h} + \gamma^H V_\xi(\hat{s}_{t+H})$
    \STATE {\color{blue}\% update value network}
    \STATE $\xi \leftarrow \xi - \eta \nabla_{\xi} L^\tau(\min_{i\in\{1,2\}} Q_{\bar{\phi}_i}(s, a) - V_\xi(s))$
    \STATE {\color{blue}\% update critic (Q networks)}
    \STATE $\phi_1 \leftarrow \phi_1 - \eta \nabla_{\phi_1} \left\| Q_{\phi_1}(s_t, a_t) - y \right\|_2^2$
    \STATE $\phi_2 \leftarrow \phi_2 - \eta \nabla_{\phi_2} \left\| Q_{\phi_2}(s_t, a_t) - y \right\|_2^2$
    \IF{$i~\bmod~n = 0$}
        \STATE {\color{blue}\% update actor}
        \STATE $\psi \leftarrow \psi + \eta \nabla_\psi \exp\left( \beta \left( \min_{i \in \{1,2\}} Q_{\phi_i}(s, a) - V_\xi(s) \right) \right) \log \pi_\psi(a \mid s)$
        \STATE {\color{blue}\% update target networks}
        \STATE $\bar{\phi}_1 \leftarrow \bar{\phi}_1 + w (\phi - \bar{\phi}_1)$
        \STATE $\bar{\phi}_2 \leftarrow \bar{\phi}_2 + w (\phi - \bar{\phi}_2)$
    \ENDIF
\ENDFOR
\STATE \textbf{Output:} $\pi_\psi$
\end{algorithmic}
\end{algorithm}

\subsection{DWM-PQL: Pessimistic Q-learning with Diffusion World Model}
Previous offline RL algorithms like MOPO~\cite{yu2020mopo} have applied the conservatism notion directly to the reward function, which we referred to as pessimistic Q-learning (PQL) in this paper. Specifically, the original algorithm proposed by \citet{yu2020mopo} learns an ensemble of $m$ one-step dynamics models $\{p_{\theta_i}\}_{i\in[m]}$, and use a modified reward
\begin{equation}
    \tilde{r}(s,a)=\hat{r}(s,a) - \kappa u(s, a|p_{\theta_1}, \ldots, p_{\theta_m})
\end{equation}
for learning the $Q$ functions, where $\hat{r}$ is the mean prediction from the ensemble, $u(s, a|p_{\theta_1}, \ldots, p_{\theta_m})$ is a measurement of prediction uncertainty using the ensemble. 

\citet{yu2020mopo} parameterize each dynamics model by a Gaussian distribution, and measure the prediction uncertainty using the maximum of the Frobenious norm of each covariance matrix. Since the diffusion model does not have such parameterization, and it is computationally daunting to train an ensemble of diffusion models, we propose an alternative uncertainty measurement similar to the one used in MoRel~\cite{kidambi2020morel}.

Given $(s_t, a_t)$ and $g_\text{eval}$, we randomly sample $m$ sequences from the DWM, namely,
\begin{equation}
    \hat{r}^i_t, \hat{s}^i_{t+1}, \hat{r}^i_{t+1}, \ldots, \hat{s}^i_{t+T-1},\hat{r}^i_{t+T-1}, \;\; i \in [m].
\end{equation}
Then, we take the 1st sample as the DWM output with modified reward:
\begin{equation}
\label{eq:morel_uncertainty}
\tilde{r}_{t'} = \sum_{i=1}^m\frac{1}{m}\hat{r}_{t'}^i - \kappa \max_{i\in[m], j \in [m]} \left( \norm{\hat{r}^i_{t'} - \hat{r}^j_{t'} }^2 + \norm{\hat{s}^i_{t'+1} - \hat{s}^j_{t'+1} }^2_2 \right), \;\; t' = t, \ldots, t+T-2.
 \end{equation}   
This provides an efficient way to construct uncertainty-penalized rewards for each timestep along the diffusion predicted trajectories. Note that this does not apply to the reward predicted for the last timestep.
The rest of the algorithm follows IQL but using MSE loss instead of expectile loss for updating the value network. 

The DWM-PQL algorithm is summarized in Algorithm~\ref{algo:pqld}

\begin{algorithm}[htbp]
\caption{DWM-PQL}
\label{algo:pqld}
\begin{algorithmic}[1]

\STATE \textbf{Inputs:} offline dataset $\mathcal{D}_\text{offline}$, pretrained diffusion world model $p_\theta$, simulation horizon $H$, conditioning RTG $g_\text{eval}$, target network update frequency $n$, pessimism coefficient $\lambda$, number of samples for uncertainty estimation $m$
\STATE Initialize actor, critic, and value networks $\pi_\psi$, $Q_{\phi_1}$, $Q_{\phi_2}$, $V_\xi$
\STATE Initialize target networks: $\bar{\phi}_1 \leftarrow \phi_1$, $\bar{\phi}_2 \leftarrow \phi_2$
\FOR{iteration $i=1, 2, \ldots$ until convergence}
    \STATE Sample state-action pair $(s_t, a_t)$ from $\mathcal{D}_\text{offline}$
    \STATE {\color{blue}\% diffusion model value expansion}
    \STATE Sample $m$ subtrajectories $\{\hat{r}^j_t, \hat{s}^j_{t+1}, \hat{r}^j_{t+1}, \ldots, \hat{s}^j_{t+T-1}, \hat{r}^j_{t+T-1}\}_{j=1}^m \sim p_\theta(\cdot \mid s_t, a_t, g_\text{eval})$
    \STATE Modify the rewards of the first subtrajectory as in Eq.~\eqref{eq:morel_uncertainty}: $\tilde{r}_t, \hat{s}_{t+1}, \tilde{r}_{t+1}, \ldots, \hat{s}_{t+T-1}, \hat{r}_{t+T-1}$
    \STATE Compute target Q-value: $y = \sum_{h=0}^{H-1} \gamma^{h} \tilde{r}_{t+h} + \gamma^H V_\xi(\hat{s}^1_{t+H})$
    \STATE {\color{blue}\% update value network}
    \STATE $\xi \leftarrow \xi - \eta \nabla_{\xi} \left\| \min_{i\in\{1,2\}} Q_{\bar{\phi}_i}(s, a) - V_\xi(s) \right\|_2^2$
    \STATE {\color{blue}\% update critic (Q networks)}
    \STATE $\phi_1 \leftarrow \phi_1 - \eta \nabla_{\phi_1} \left\| Q_{\phi_1}(s_t, a_t) - y \right\|_2^2$
    \STATE $\phi_2 \leftarrow \phi_2 - \eta \nabla_{\phi_2} \left\| Q_{\phi_2}(s_t, a_t) - y \right\|_2^2$
    \IF{$i~\bmod~n = 0$}
        \STATE {\color{blue}\% update actor}
        \STATE $\psi \leftarrow \psi + \eta \nabla_\psi \exp\left( \beta \left( \min_{i \in \{1,2\}} Q_{\phi_i}(s, a) - V_\xi(s) \right) \right) \log \pi_\psi(a \mid s)$
        \STATE {\color{blue}\% update target networks}
        \STATE $\bar{\phi}_1 \leftarrow \bar{\phi}_1 + w (\phi - \bar{\phi}_1)$
        \STATE $\bar{\phi}_2 \leftarrow \bar{\phi}_2 + w (\phi - \bar{\phi}_2)$
    \ENDIF
\ENDFOR
\STATE \textbf{Output:} $\pi_\psi$
\end{algorithmic}
\end{algorithm}

For baseline methods with one-step dynamics model, the imagined trajectories starting from sample $(s_t, a_t)\sim \mathcal{D}_\text{offline}$ are derived by recursively sample from the one-step dynamics model $f_\theta(\cdot|s,a)$ and policy $\pi_\psi(\cdot|s)$: $\hat{\tau}(s_t, a_t) \sim (f_\theta \circ \pi_\psi)^{H-1}(s_t, a_t)$. By keeping the rest same as above, it produces MBRL methods with one-step dynamics, namely O-IQL, O-TD3BC and O-PQL.

\section{Training and Evaluation Details of Offline RL Algorithms}
\label{subsec:app_mf_mb_details}
\subsection{Common Settings}
We conduct primary tests on TD3+BC and IQL for selecting the best practices for data normalization. Based on the results,
TD3+BC, O-TD3BC and DWM-TD3BC applies observation normalization, while other algorithms (O-PQL, DWM-PQL, IQL, O-IQL and DWM-IQL) applies both observation and reward normalization.

All models are trained on NVIDIA Tesla V100 PCle GPU devices. For training 200 epochs (1000 iterations per epoch), model-free algorithms like TD3+BC and IQL typically takes around 2000 seconds, DWM model-based algorithms like DWM-TD3BC and DWM-IQL typically takes around 18000 seconds, transformer model-based algorithms like T-TD3BC and T-IQL typically takes around  8000 seconds, one-step model-based algorithms like O-TD3BC and O-IQL typically takes around 2300 seconds. The specific computational time varies from task to task.

All algorithms are trained with a batch size of 128 using a fixed set of pretrained dynamics models (one-step and diffusion). The discount factor is set as $\gamma=0.99$ for all data.

\subsection{Model-Free Algorithms}
TD3+BC and IQL are trained for $1\times 10^6$ iterations, with learning rate $3\times 10^{-4}$ for actor, critic and value networks. The actor, critic, and value networks are all parameterized by 3-layer MLPs with 256 hidden units per layer. We use the ReLU activation function for each hidden layer. IQL learns a stochastic policy which outputs a Tanh-Normal distribution, while TD3+BC has a deterministic policy with Tanh output activation. The hyperparameters for TD3+BC and IQL are provided in Table~\ref{tab:mf_hyperparam}. \looseness=-1

\begin{table}[h]
\centering
\begin{tabular}{cc|cc}
\toprule
\multicolumn{2}{c|}{TD3+BC} & \multicolumn{2}{c}{IQL}  \\ \hline
policy noise & 0.2 & expectile & 0.7  \\
noise clip & 0.5 & $\beta$ & 3.0\\
policy update frequency & 2 & max weight & 100.0 \\
target update frequence & 2 & policy update frequence & 1 \\
$\alpha$ & 2.5 & advantage normalization & False\\ 
EMA $w$ & 0.005 & EMA $w$ & 0.005  \\
\bottomrule
\end{tabular}
 \caption{Hyperparameters for training TD3+BC and IQL.}
\label{tab:mf_hyperparam}
\end{table}

The baseline DD~\cite{ajay2022conditional} algorithm uses diffusion models trained with sequence length $T=32$ and number of diffusion steps $K=5$.
It requires additionally training an inverse dynamics model (IDM) for action prediction, which is parameterized by a 3-layer MLP with 1024 hidden units for each hidden layer and ReLU activation function. The dropout rate for the MLP is 0.1. The IDMs are trained for $2\times 10^6$ iterations for each environment.
For a fair comparison with the other DWM methods, DD uses $N=3$ internal sampling steps as DWM. We search over the same range of evaluation RTG $g_\text{eval}$ for DD and the other DWM methods.

\subsection{Model-Based Algorithms}
DWM-TD3BC, DWM-IQL and DWM-PQL are trained for $5\times 10^5$ iterations. Table~\ref{tab:hyperparam} summarizes the hyperparameters we search for each experiment. The other hyperparameters and network architectures are the same as original TD3+BC and IQL in above sections. DWM-IQL with $\lambda$-return takes $\lambda=0.95$, following \citet{hafner2023mastering}.

The counterparts with one-step dynamics models are trained for $2\times 10^5$ iterations due to a relatively fast convergence from our empirical observation. Most of the hyperparameters also follow TD3+BC and IQL. The PQL-type algorithms (O-PQL and DWM-PQL) further search the pessimism coefficient $\kappa$ (defined in Eq.~\eqref{eq:morel_uncertainty}) among $\{0.01, 0.1, 1.0\}$.

\begin{table}[H]
\centering
\begin{tabular}{cccc}
\toprule
Env & Evaluation RTG &$H$ \\ \hline
hopper-medium-v2 & [0.6, 0.7, 0.8]  & [1,3,5,7]\\
walker2d-medium-v2 &  [0.6, 0.7, 0.8]  & [1,3,5,7] \\
halfcheetah-medium-v2 &  [0.4, 0.5, 0.6]   & [1,3,5,7]\\
hopper-medium-replay-v2 & [0.6, 0.7, 0.8]  & [1,3,5,7]\\
walker2d-medium-replay-v2 & [0.6, 0.7, 0.8]  & [1,3,5,7]\\
halfcheetah-medium-replay-v2 &  [0.4, 0.5, 0.6]  & [1,3,5,7]\\
hopper-medium-expert-v2 & [0.7, 0.8, 0.9]  & [1,3,5,7] \\
walker2d-medium-expert-v2 & [0.8, 0.9, 1.0]   & [1,3,5,7]\\
halfcheetah-medium-expert-v2 &  [0.6, 0.7, 0.8] & [1,3,5,7] \\
\bottomrule
\end{tabular}
 \caption{List of the hyperparameters we search for DWM methods.}
\label{tab:hyperparam}
\end{table}

\section{Additional Experiments}
\vspace{-.5cm}
\subsection{Detailed Results of Long Horizon Planning with DWM}
\label{subsec:app_rollout_length}
The section provides the detailed results of the experiments for long horizon planning with DWM in Section~\ref{sec:expr_dwm_vs_onestep}. Table~\ref{tab:rollout_length} summarizes the normalized returns (with means and standard deviations) of DWM-IQL and DWM-TD3BC for different simulation horizons $\{1,3,7,15, 31\}$. 

\begin{table}[H]
\centering
\resizebox{0.58\columnwidth}{!}{
\begin{tabular}{cccc}
\toprule
& & \multicolumn{2}{c}{Return (mean$\pm$std)} \\ \cline{3-4}
Env. & Simulation Horizon & DWM-IQL & DWM-TD3BC \\ \hline
\multirow{5}{*}{\textbf{hopper-medium-v2}} 
& 1 & 0.54 ± 0.11 & 0.68 ± 0.12\\
& 3 & 0.55 ± 0.10  & 0.63 ± 0.11\\
& 7 & 0.56 ± 0.09  & 0.66 ± 0.13 \\
& 15 & 0.58 ± 0.12  & 0.77 ± 0.15\\
& 31 & 0.61 ± 0.11  & 0.79 ± 0.15\\ \hline
\multirow{5}{*}{\textbf{walker2d-medium-v2}} 
& 1 & 0.65 ± 0.23 & 0.56 ± 0.13 \\
& 3 & 0.74 ± 0.11  & 0.74 ± 0.13 \\
& 7 & 0.71 ± 0.13  & 0.74 ± 0.11 \\
& 15 & 0.66 ± 0.15 & 0.73 ± 0.13\\
& 31 & 0.67 ± 0.20 & 0.75 ± 0.12\\\hline
\multirow{5}{*}{\textbf{halfcheetah-medium-v2}} 
& 1 & 0.44 ± 0.01 & 0.35 ± 0.03 \\
& 3 & 0.44 ± 0.01  & 0.39 ± 0.01 \\
& 7 & 0.44 ± 0.01  & 0.40 ± 0.01\\
& 15 & 0.44 ± 0.02   & 0.40 ± 0.01 \\
& 31 & 0.44 ± 0.01 & 0.40 ± 0.01 \\\hline
\multirow{5}{*}{\textbf{hopper-medium-replay-v2}} 
& 1 & 0.18 ± 0.06 & 0.52 ± 0.21 \\
& 3 & 0.37 ± 0.18  & 0.44 ± 0.23 \\
& 7 & 0.39 ± 0.14  & 0.52 ± 0.28 \\
& 15 & 0.37 ± 0.18   & 0.67 ± 0.25 \\
& 31 & 0.37 ± 0.15 & 0.59 ± 0.22 \\\hline
\multirow{5}{*}{\textbf{walker2d-medium-replay-v2}} 
& 1 & 0.32 ± 0.15 & 0.13 ± 0.02 \\
& 3 & 0.27 ± 0.24  & 0.19 ± 0.10 \\
& 7 & 0.25 ± 0.20  & 0.22 ± 0.14 \\
& 15 & 0.26 ± 0.19   & 0.22 ± 0.10 \\
& 31 & 0.27 ± 0.19 & 0.17 ± 0.12  \\\hline
\multirow{5}{*}{\textbf{halfcheetah-medium-replay-v2}}  
& 1 & 0.38 ± 0.05 & 0.02 ± 0.00\\
& 3 & 0.39 ± 0.02  & 0.17 ± 0.05  \\
& 7 & 0.39 ± 0.02  & 0.22 ± 0.03 \\
& 15 & 0.38 ± 0.03   & 0.26 ± 0.03\\
& 31 & 0.37 ± 0.03 & 0.26 ± 0.05 \\\hline
\multirow{5}{*}{\textbf{hopper-medium-expert-v2}} 
& 1 & 0.86 ± 0.25 & 0.88 ± 0.17\\
& 3 & 0.90 ± 0.19 & 0.94 ± 0.22 \\
& 7 & 0.88 ± 0.28  & 0.93 ± 0.24\\
& 15 & 0.85 ± 0.20  & 0.91 ± 0.19 \\
& 31 & 0.84 ± 0.23 & 0.93 ± 0.23 \\\hline
\multirow{5}{*}{\textbf{walker2d-medium-expert-v2}} 
& 1 & 0.80 ± 0.22 &  0.74 ± 0.21\\
& 3 & 1.02 ± 0.09  & 0.89 ± 0.13 \\
& 7 & 0.98 ± 0.2  & 0.82 ± 0.19  \\
& 15 & 1.06 ± 0.05 & 0.84 ± 0.14 \\
& 31 & 1.05 ± 0.06 & 0.87 ± 0.03 \\\hline
\multirow{5}{*}{\textbf{halfcheetah-medium-expert-v2}} 
& 1 & 0.60 ± 0.18 & 0.39 ± 0.01\\
& 3 & 0.52 ± 0.14  & 0.43 ± 0.07 \\
& 7 & 0.63 ± 0.13  & 0.44 ± 0.03\\
& 15 & 0.66 ± 0.14   & 0.50 ± 0.08\\
& 31 & 0.65 ± 0.17 & 0.49 ± 0.09 \\
\bottomrule
\end{tabular}
}
 \caption{Comparison of the normalized returns with different simulation horizons for DWM-TD3BC and DWM-IQL. The reported values are the best performances across different RTG values (listed in Table~\ref{tab:hyperparam}).}
\label{tab:rollout_length}
\end{table}

\subsection{World Modeling: Prediction Error Analysis}
\label{subsec:compounding}
An additional experiment is conducted to evaluate the prediction errors of the observations and rewards with DWM under simulation horizon $H=8$, for an example task \textit{walker2d-medium-expert-v2}.

We randomly sample a subsequence $(s_t,\dots, s_{t+7})$ from the offline dataset, and let DWM predict the subsequent states and rewards, conditioned on the first state $s_t$ and true action $a_t$.
For the one-step model, the model iteratively predicts the reward $r_t$ and next state $s_{t+1}$, conditioned on the current state $s_t$ (which is predicted in the previous step) and true action $a_t$.

We report the mean squared error (MSE) between the predicted samples and the  ground truth for each rollout timestep in Table ~\ref{tab:obs_rew_compounding}. Each method is evaluated with five models and 100 sequences per model, and the mean and standard deviations are reported.  The average prediction errors over the entire sequence are also calculated in the last row.  It shows the significant reduction of prediction errors in sequence modeling by using DWM over traditional one-step models, especially when the prediction timestep is large.

\begin{table}[H]
\centering
\resizebox{\textwidth}{!}{
\begin{tabular}{c||c|c||c|c}
\hline
\multirow{2}{*}{\textbf{Step}} & \multicolumn{2}{c||}{\textbf{One-step Model}} & \multicolumn{2}{c}{\textbf{DWM}} \\
\cline{2-5}
 & \textbf{Observation} & \textbf{Reward} & \textbf{Observation} & \textbf{Reward} \\
\hline
1 & $0.0000 \pm 0.0000$ & $8.05e-05 \pm 0.00011$ & $0.0000 \pm 0.0000$ & $8.89e-05 \pm 0.00018$ \\
2 & $0.0363 \pm 0.0455$ & $2.03e-04 \pm 0.00031$ & $0.1050 \pm 0.1668$ & $1.50e-04 \pm 0.00020$ \\
3 & $0.1576 \pm 0.2308$ & $4.73e-04 \pm 0.00084$ & $0.4173 \pm 0.3902$ & $2.34e-04 \pm 0.00021$ \\
4 & $0.3503 \pm 0.3547$ & $7.68e-04 \pm 0.00148$ & $0.4525 \pm 0.5021$ & $2.33e-04 \pm 0.00018$ \\
5 & $0.6173 \pm 0.4945$ & $1.13e-03 \pm 0.00222$ & $0.4796 \pm 0.5035$ & $2.71e-04 \pm 0.00039$ \\
6 & $0.9185 \pm 0.8678$ & $2.00e-03 \pm 0.00417$ & $0.4854 \pm 0.4759$ & $4.01e-04 \pm 0.00102$ \\
7 & $1.2394 \pm 0.9788$ & $3.40e-03 \pm 0.00757$ & $0.5353 \pm 0.5076$ & $3.89e-04 \pm 0.00094$ \\
8 & $1.4890 \pm 0.9612$ & $4.47e-03 \pm 0.00844$ & $0.5146 \pm 0.5814$ & $3.77e-04 \pm 0.00064$ \\
\hline
\textbf{Average} & $0.6010 \pm 0.4916$ & $0.0015 \pm 0.0031$ & $0.3737 \pm 0.3984$ & $0.0003 \pm 0.0004$ \\
\hline
\end{tabular}
}
\caption{Comparison of observation and prediction error (MSE) over rollout timesteps ($H=8$) for one-step model and DWM for \textit{walker2d-medium-expert-v2}.}
\label{tab:obs_rew_compounding}
\end{table}

\subsection{Transformer-based World Model}
\label{subsec:app_transformer}
Following the same protocol as DWM Algorithm~\ref{algo:diffusion_mbrl}, the Transformer model is trained to predict future state-reward sequences, conditioning on the initial state-action pair. We use a 4-layer transformer architecture with 4 attention heads, similar to the one in \citet{zheng2022online}. Specifically, all the actions except for the first one are masked out as zeros in the state-action-reward sequences.
Distinct from the original DT~\cite{chen2021decision} where the loss function only contains the action prediction error, here the Transformer is trained with state and reward prediction loss. The Transformers are trained with optimizers and hyperparameters following ODT~\cite{zheng2022online}.
The evaluation RTG for Transformers takes values $3600/400=9.0, 5000/550\approx 9.1, 6000/1200=5.0$ for hopper, walker2d and halfcheetah environments, respectively. The complete results of T-TD3BC and T-IQL are provided in Table~\ref{tab:dt_td3bc} and Table~\ref{tab:dt_iql} respectively.

\begin{table}[H]
\centering
\small
\begin{tabular}{c|cccc}
\toprule
& \multicolumn{4}{c}{Simulation Horizon} \\
Env & 1 & 3 & 5 & 7   \\  \hline
hopper-m  & 0.50$\pm$0.05 & 0.57$\pm$0.08 & 0.58$\pm$0.08 & 0.57$\pm$0.08 \\
walker2d-m & 0.36$\pm$0.15 &  0.40$\pm$0.20 & 0.60$\pm$0.16 & 0.53$\pm$0.17  \\
halfcheetah-m & 0.18$\pm$0.07 &  0.41 $\pm$0.03 & 0.38$\pm$0.08  &  0.42$\pm$0.03   \\
hopper-mr &   0.24$\pm$0.01 & 0.23$\pm$0.05 & 0.25$\pm$0.06 & 0.22$\pm$0.08  \\ 
walker2d-mr &  0.12 $\pm$0.04 & 0.09$\pm$0.05 & 0.13$\pm$0.06 & 0.12$\pm$0.05 \\
halfcheetah-mr &  0.40$\pm$0.01 & 0.39$\pm$0.02 & 0.39$\pm$0.03 & 0.39$\pm$0.02  \\
hopper-me &  0.41$\pm$0.13 & 0.57$\pm$0.19 & 0.66$\pm$0.25 & 0.52$\pm$0.15 \\
walker2d-me & 0.34$\pm$0.22 & 0.58$\pm$0.15 & 0.58$\pm$0.26 & 0.46$\pm$0.37  \\
halfcheetah-me &  0.14$\pm$0.06 & 0.31$\pm$0.09 & 0.36$\pm$0.17 & 0.29$\pm$0.12 \\ 
\bottomrule
\end{tabular}
\caption{The normalized returns of T-TD3BC.}
\label{tab:dt_td3bc}
\end{table}

\begin{table}[H]
\small
\centering
\begin{tabular}{c|cccc}
\toprule
& \multicolumn{4}{c}{Simulation Horizon} \\
Env & 1 & 3 & 5 & 7   \\  \hline
hopper-m  & 0.48$\pm$0.08 & 0.54$\pm$0.10 & 0.55$\pm$0.08 & 0.51$\pm$0.09 \\
walker2d-m & 0.54$\pm$0.18 &  0.62$\pm$0.19 & 0.72$\pm$0.12 & 0.72$\pm$0.14  \\
halfcheetah-m & 0.42$\pm$0.03 & 0.42$\pm$0.02 & 0.43$\pm$0.01 & 0.43$\pm$0.01     \\
hopper-mr & 0.17$\pm$0.05 & 0.24$\pm$0.09 & 0.26$\pm$0.09 & 0.20$\pm$0.07\\
walker2d-mr & 0.17$\pm$0.12 & 0.17$\pm$0.14 & 0.23$\pm$0.12 & 0.16$\pm$0.11\\
halfcheetah-mr & 0.38$\pm$0.04 & 0.39$\pm$0.01 & 0.38$\pm$0.04 & 0.39$\pm$ 0.03\\
hopper-me &  0.62$\pm$0.16 & 0.59$\pm$0.21 & 0.47$\pm$0.21 &  0.47$\pm$0.21 \\
walker2d-me & 0.67$\pm$0.23 & 0.87$\pm$0.21 & 1.03$\pm$0.09 & 0.71$\pm$0.22 \\
halfcheetah-me & 0.39$\pm$0.19 & 0.43$\pm$0.13 & 0.44$\pm$0.08 &  0.43$\pm$0.09 \\
\bottomrule
\end{tabular}
\caption{The normalized returns of T-IQL.}
\label{tab:dt_iql}
\end{table}

\subsection{Additional Baselines}
\label{subsec:add_baselines}
\subsubsection{Data Augmentation}
\label{subsec:data_aug}
As a data augmentation (DA) method, SynTHER~\cite{lu2023synthetic} learns an unconditional diffusion model at the transition level, where the generated data are used for augmenting the training distribution. We conduct additional experiments for DWM and SynTHER-type data augmentation for this section.

The two data augmentation methods based on TD3+BC and IQL are referred to as DA-TD3BC and DA-IQL. For the sake of fair comparison, for DA-TD3BC and DA-IQL, the models are trained in the same manner as DWM with the same parameter sweeping for RTG values and horizons. The results show that the DWM consistently performs better than the DA algorithms across all tasks, for both TD3+BC and IQL.

For the sake of fair comparison with DWM, we use the same data preprocessing as DWM for this experiment, which is different from state reward normalizations as the original SynTHER paper. 
In our experiments, we use CDF normalizers to transform each dimension of the state vector to $[-1, 1]$ independently, \emph{i.e.}, making the data uniform over each dimension by transforming with marginal CDFs. Specifically, 
we transform the raw reward $r_\text{raw}$ to $r = 2(r_\text{raw}-r_\text{min})/(r_\text{max}-r_\text{min})-1$, where $r_\text{min}$ and $r_\text{max}$
are max and min raw reward of the offline dataset.
SynTHER applies ``whitening" that makes each (non-terminal) continuous dimension mean 0 and std 1, and the terminal states are rounded without normalization. To enable fast sampling, we use a very low diffusion steps: 5 at training and 3 at testing. The original SynthER paper uses 128 steps, which requires more computational time for sample generation. We use a set of consistent parameters like model sizes and batch sizes cross all the environments, the same as our previous experiments.

\begin{table}[ht]
\centering
\begin{tabular}{c|cc|cc}
\toprule
Env. & DA-TD3BC & DWM-TD3BC & DA-IQL & DWM-IQL \\
\midrule
hopper-m & 0.65 $\pm$ 0.10 & \textbf{0.65 $\pm$ 0.10} & 0.51 $\pm$ 0.10 & 0.54 $\pm$ 0.11 \\
walker2d-m & 0.63 $\pm$ 0.18 & \textbf{0.70 $\pm$ 0.15} & 0.74 $\pm$ 0.09 & 0.76 $\pm$ 0.05 \\
halfcheetah-m & 0.44 $\pm$ 0.01 & \textbf{0.46 $\pm$ 0.01} & 0.44 $\pm$ 0.01 & 0.44 $\pm$ 0.01 \\
hopper-mr & 0.53 $\pm$ 0.09 & 0.53 $\pm$ 0.09 & 0.25 $\pm$ 0.04 & \textbf{0.61 $\pm$ 0.13} \\
walker2d-mr & 0.37 $\pm$ 0.22 & \textbf{0.46 $\pm$ 0.19} & 0.42 $\pm$ 0.24 & 0.35 $\pm$ 0.14 \\
halfcheetah-mr & \textbf{0.43 $\pm$ 0.01} & \textbf{0.43 $\pm$ 0.01} & 0.42 $\pm$ 0.04 & 0.41 $\pm$ 0.01 \\
hopper-me & \textbf{1.03 $\pm$ 0.14} & \textbf{1.03 $\pm$ 0.14} & 0.55 $\pm$ 0.19 & 0.90 $\pm$ 0.25 \\
walker2d-me & 1.09 $\pm$ 0.04 & \textbf{1.10 $\pm$ 0.00} & 0.76 $\pm$ 0.13 & 1.04 $\pm$ 0.10 \\
halfcheetah-me & 0.72 $\pm$ 0.14 & \textbf{0.75 $\pm$ 0.16} & 0.62 $\pm$ 0.14 & 0.71 $\pm$ 0.14 \\
\midrule
Average & 0.654 $\pm$ 0.103 & \textbf{0.679 $\pm$ 0.098} & 0.523 $\pm$ 0.109 & 0.641 $\pm$ 0.117 \\
\bottomrule
\end{tabular}
\caption{\small Comparison of our DWM method and data-augmentation (DA) methods on the D4RL dataset. Results are aggregated over 5 random seeds. }
\label{tab:data_aug_compare}
\end{table}

\subsubsection{Autoregressive Diffusion}
We conduct experiments on Autoregressive Diffusion (AD) mentioned in previous work~\cite{rigter2023world} as an additional Baseline. AD is essentially a one-step model using diffusion instead of MLP, which can be autoregressively rolled out and used for Model Based Value Expansion (MVE). We find that AD is computationally inefficient to be practically integrated into the MVE framework. We have checked the wallclock time (in seconds) for sampling a batch of 128 future trajectories with different values of horizon, for the \textit{walker2d-medium-v2} environment.

We set the number of trajectories to be 128 because this is the batch size we use for training RL agents. Results are averaged over 100 trials. For both approaches we use diffusion models of the same model size, where the number of sampling diffusion steps is 3 (to enable fast inference). This experiment is conducted on a A6000 GPU and time unit is second. 

The results are displayed in Table~\ref{tab:compare_ad}. The sampling time of DWM is a constant because it's a sequence model, and in practice we diffuse the whole sequence and take a part of it according to $H$; while the sampling time of AD scale linearly as $H$ increases. When $H=7$, the sampling time is roughly 
$6.67\times$ compared with DWM. The MB methods we reported in the paper are trained for $5\times 10^5$ iterations (see Section D.3). That means, even only generating trajectories will take 27.5 hours if we use AD (as opposed to $\sim4$ hours for DWM).This suggests that AD is too computationally expensive to be incorporated into the MBRL framework.

\begin{table}[htbp]
\centering
\begin{tabular}{l|c|c|c|c}
\hline
\textbf{Method} & \textbf{H=1} & \textbf{H=3} & \textbf{H=5} & \textbf{H=7} \\ \hline
DWM (trained with $T=8$) & 0.031 & 0.031 & 0.031 & 0.031 \\ \hline
Autoregressive Diffusion & 0.030 & 0.087 & 0.145 & 0.198 \\ \hline
\end{tabular}
\caption{Comparison AD and DWM for sampling time (seconds) under different horizon $H$ values}
\label{tab:compare_ad}
\end{table}

\subsection{Additional Environments: Sparse-reward Tasks}
\label{sec:add_sparse_reward}
To further verify the method on various tasks, we conduct experiments on the maze-type tasks with sparse rewards. The methods include TD3+BC, IQL, the DWM counterparts and DD. The training and evaluation protocol follow exactly the same as the main experiments in Sec.~\ref{subsec:app_mf_mb_details}. The results are summarized in Table~\ref{tab:maze}, which shows the superior performance of DWM-based algorithms in sparse-reward settings.

\begin{table}[ht]
\centering
\begin{tabular}{c|ccccc}
\toprule
Env. & DWM-TD3BC & DWM-IQL & TD3+BC & IQL & DD\\
\midrule
maze2d-umaze   & $0.36\pm0.23$ & $0.39\pm0.29$ & $0.05\pm0.15$ & $0.08\pm0.16$ & $0.40\pm0.52$ \\ \hline
maze2d-medium  & $0.57\pm0.50$ & $0.41\pm0.09$ & $0.00\pm0.006$ & $0.10\pm0.10$ & $0.20\pm0.19$ \\ \hline
maze2d-large   & $0.28\pm0.13$ & $0.11\pm0.13$ & $-0.01\pm0.03$ & $0.01\pm0.07$ & $0.02\pm0.07$ \\ \hline
antmaze-umaze  & $0.86\pm0.29$ & $0.66\pm0.47$ & $0.58\pm0.49$ & $0.64\pm0.48$ & $0.31\pm0.45$ \\
\bottomrule
\end{tabular}
\caption{\small Comparison of different methods on sparse-reward tasks: three \textit{maze2d} tasks and one \textit{antmaze} task.}
\label{tab:maze}
\end{table}

\subsection{Ablation: Number of Diffusion Steps for Training and Inference} 
\label{app:diffusion_steps}
The number of training diffusion steps $K$ can heavily influence the modeling quality, where a larger value of $K$ generally leads to better performance.
At the same time, sampling from the diffusion models is recognized as a slow procedure, as it involves $K$ internal denoising steps.
We apply the \textit{stride sampling} technique~\cite{nichol2021improved} to accelerate the sampling process with reduced internal steps $N$, see Sec.~\ref{subsec:app_acc} for more details.
However, the sampling speed comes at the cost of quality. It is important to strike a balance between inference speed and prediction accuracy.
We investigate how to choose the number of $K$ and $N$ to significantly accelerate sampling without sacrificing model performance. \looseness=-1

\begin{figure}[t]
    \centering
    \includegraphics[width=0.65\columnwidth]{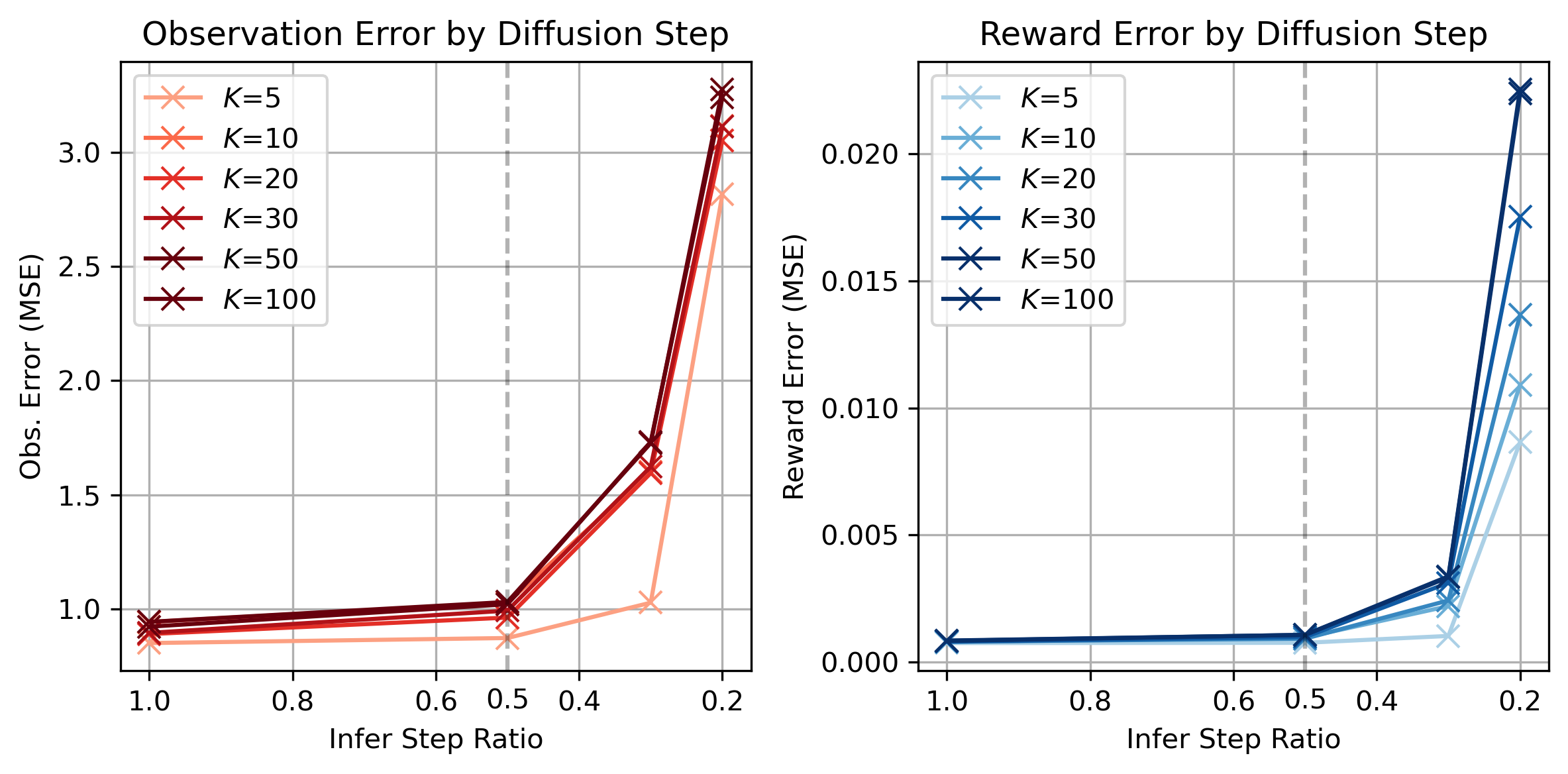}
    \caption{
    Average observation and reward prediction errors (across 9 tasks and simulation horizon $H\in[7]$) for DWM  DWM trained with $T=8$ and different diffusion steps $K$, as the inference step ratio $r_\text{ratio}$ changes.
    }
    \label{fig:prediction_error_T8}

    \includegraphics[width=0.65\columnwidth]{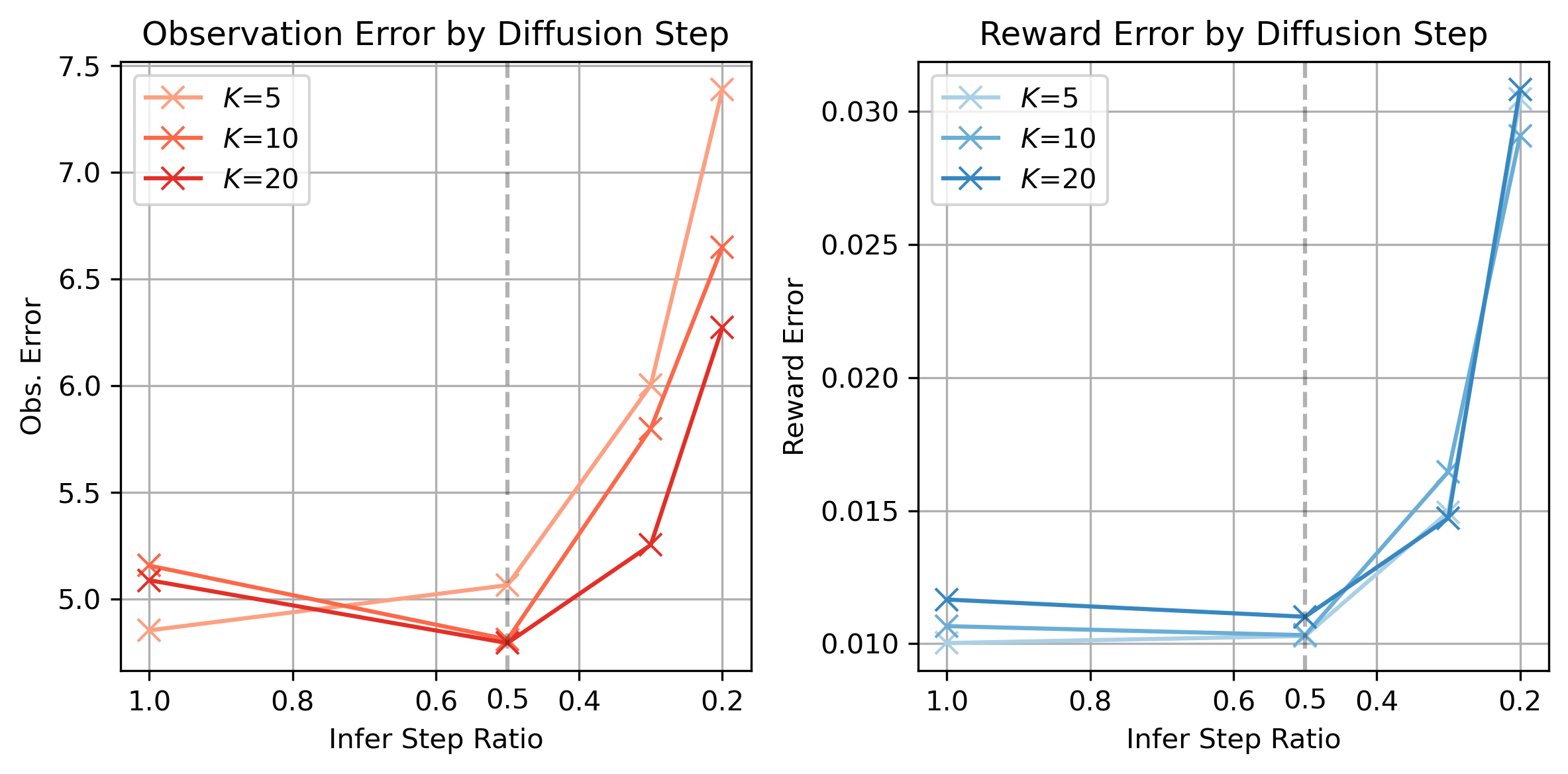}
    \caption{Average observation and reward prediction errors (across 9 tasks and simulation horizon $H \in [31]$) for DWM trained with $T=32$ and different diffusion steps $K$, as the inference step ratio $r_\text{ratio}$ changes.}
    \label{fig:prediction_error_T32}
\end{figure}

We train DWM with different numbers of diffusion steps $K\in\{5,10,20,30,50,100\}$, where the sequence length is $T=8$. We set four inference step ratios $r_\text{infer}\in\{0.2, 0.3, 0.5, 1.0\}$ and use $N=\lceil r_\text{infer}\cdot K \rceil$ internal steps in stride sampling. Figure~\ref{fig:prediction_error_T8} reports the prediction errors of DMW for both observation and reward sequences, defined in Equation~\eqref{eq:avg_pred_error}. We note that the prediction error depends on the evaluation RTG, and we report the best results across multiple values of it (listed in Table~\ref{tab:hyperparam}). An important observation is that $r_\text{infer}=0.5$ is a critical ridge for distinguishing the performances with different inference steps, where $N < K/2$ hurts the prediction accuracy significantly. 
Moreover, within the regime $r_\text{infer} \geq 0.5$, a small diffusion steps $K=5$ performs roughly the same as larger values.
Therefore, we choose $K=5$ and $r_\text{infer}=0.5$ for our main experiments, which leads to the number of sampling steps $N=3$. We have also repeated the above experiments for DWM with longer sequence length $T=32$. The results also support the choice $r_\text{infer}=0.5$ but favors $K=10$, see Figure~\ref{fig:prediction_error_T32}.

\subsubsection{Details Results}
Let $\tau$ denote a length-$T$ subtrajectory $(s_t, a_t, r_t, s_{t+1}, \ldots, s_{t+T-1}, r_{t+T-1})$.
The average prediction errors of a DWM $p_\theta$ for states and rewards along sequences are defined as:
\begin{align}
    \bar{\epsilon}_s = \E_{ \tau \sim \mathcal{D}_\text{offline}, \hat{s}_t\sim p_\theta(\cdot|s_1, a_1, g_\text{eval})}\left[\frac{1}{T}\sum_{t'=t}^{t+T-1} \norm{\hat{s}_{t'} - s_{t'}}^2 \right],  \; \text{and}\nonumber \\
    \bar{\epsilon}_r = \E_{\tau \sim \mathcal{D}_\text{offline}, \hat{s}_t\sim p_\theta(\cdot|s_1, a_1, g_\text{eval})}\left[\frac{1}{T}\sum_{t'=t}^{t+T-1} \norm{\hat{r}_{t'} - r_{t'}}^2 \right]. \nonumber \\
  \label{eq:avg_pred_error}
\end{align}
We first note that all the prediction errors depend on the evaluation RTG $g_\text{eval}$. For the ease of clean presentation, we search over multiple values of $g_\text{eval}$, listed in Table~\ref{tab:hyperparam}, and report the best results.

In addition to Figure~\ref{fig:prediction_error_T8}, the average prediction errors for diffusion models with $T=32$ (longer sequence) and diffusion steps $K\in\{5, 10, 20\}$ are shown in Figure~\ref{fig:prediction_error_T32}. 
Based on the results, $K=10$ and $r_\text{infer}=0.5$ are selected to strike a good balance between prediction accuracy and inference speed. The corresponding numerical results are listed in Table~\ref{tab:prediction_error_T8} and \ref{tab:prediction_error_T32}.

\begin{table}[H]
\centering
\small
\begin{tabular}{ccccc}
\toprule
Diffusion Step $K$ & Infer Step Ratio $r_\text{ratio}$ & Infer Step $N$ & Obs. Error & Reward Error \\
\midrule
\multirow{4}{*}{5} & 0.2 & 1 & 2.815 & 0.009 \\
& 0.3 & 2 & 1.028 & 0.001 \\
& 0.5 & 3 &  0.873 & 0.001 \\
& 1.0 & 5 & 0.851 & 0.001 \\ \hline
\multirow{4}{*}{10} & 0.2  & 1 & 3.114 & 0.011 \\
& 0.3  & 2 & 1.601 & 0.002 \\
& 0.5  & 3 & 1.028 & 0.001 \\
& 1.0  & 5 & 0.943 & 0.001 \\ \hline
\multirow{4}{*}{20} & 0.2  & 1 & 3.052 & 0.014 \\
& 0.3  & 2 & 1.595 & 0.002 \\
& 0.5  & 3 & 0.963 & 0.001 \\
& 1.0  & 5 & 0.890 & 0.001 \\ \hline
\multirow{4}{*}{30} & 0.2  & 1 & 3.112 & 0.018 \\
& 0.3  & 2 & 1.623 & 0.003 \\
& 0.5  & 3 & 0.993 & 0.001 \\
& 1.0  & 5 & 0.896 & 0.001 \\ \hline
\multirow{4}{*}{50} & 0.2  & 1 & 3.275 & 0.022 \\
& 0.3  & 2 & 1.726 & 0.003 \\
& 0.5  & 3 & 1.031 & 0.001 \\
& 1.0  & 5 & 0.944 & 0.001 \\ \hline
\multirow{4}{*}{100} & 0.2  & 1 & 3.239 & 0.023 \\
& 0.3  & 2 & 1.732 & 0.003 \\
& 0.5  & 3 & 1.021 & 0.001 \\
& 1.0  & 5 & 0.923 & 0.001 \\
\bottomrule
\end{tabular}
\caption{The average (across tasks and simulation horizon $H\in[7]$) observation and reward prediction errors for DWM with $T=8$ and different inference steps $N=\lceil r_\text{infer}\cdot K \rceil$.}
\label{tab:prediction_error_T8}
\end{table}

\begin{table}[H]
\centering
\small
\begin{tabular}{ccccc}
\toprule
Diffusion Step $K$ & Infer Step Ratio $r_\text{infer}$ & Infer Step $N$ & Obs. Error & Reward Error \\
\midrule
\multirow{4}{*}{5} & 0.2 & 1 & 7.390 & 0.030 \\
& 0.3  & 2 & 6.003 & 0.015 \\
& 0.5 & 3 & 5.065 & 0.010 \\
& 1.0  & 5 & 4.853 & 0.010 \\ \hline
\multirow{4}{*}{10} & 0.2  & 1 & 6.650 & 0.029 \\
& 0.3  & 2 & 5.799 & 0.016 \\
& 0.5  & 3 & 4.811 & 0.010 \\
& 1.0  & 5 & 5.157 & 0.011 \\ \hline
\multirow{4}{*}{20} & 0.2  & 1 & 6.273 & 0.031 \\
& 0.3  & 2 & 5.254 & 0.015 \\
& 0.5  & 3 & 4.794 & 0.011 \\
& 1.0  & 5 & 5.088 & 0.012 \\
\bottomrule
\end{tabular}
\caption{The average (across tasks and simulation horizon $H\in[31]$) observation and reward prediction errors for DWM with $T=32$ and different inference steps $N=\lceil r_\text{infer}\cdot K \rceil$.}
\label{tab:prediction_error_T32}
\end{table}

\subsection{Ablation: Sequence Length of Diffusion World Model}
\label{app:seq_length_dwm}
We further compare the average performances of algorithms with DWM trained with sequence length $T=8$ and $T=32$.
Table~\ref{tab:rollout_compare} presents average best return across 9 tasks (searched over RTG values and simulation horizon $H$). 
Even though DWM is robust to long-horizon simulation and in certain cases we have found the optimal $H$ is larger than $8$, 
we found using $T=32$ improves the performance of DWM-IQL, but slightly hurts the performance of DWM-TD3BC. 
\begin{table}[H]
\centering
\small
\begin{tabular}{cc|cc}
\toprule
\multicolumn{2}{c|}{DWM-TD3BC} & \multicolumn{2}{c}{DWM-IQL (w/o $\lambda$)}\\
T=8 & T=32 & T=8 & T=32    \\  \hline
0.68 $\pm$ 0.10 & 0.60 $\pm$ 0.12 & 0.57 $\pm$ 0.09 & 0.61$\pm$ 0.10\\
\bottomrule
\end{tabular}
\caption{The average performance of DWM algorithms across 9 tasks, using DWM with different sequence lengths.}
\label{tab:rollout_compare}
\end{table}
Therefore, we choose $T=8$ for our main experiments.

\subsection{Ablation: OOD Evaluation RTG Values} 
\label{app:ood_rtg}
\begin{figure}[htbp]
    \centering
    \includegraphics[height=0.2\columnwidth]{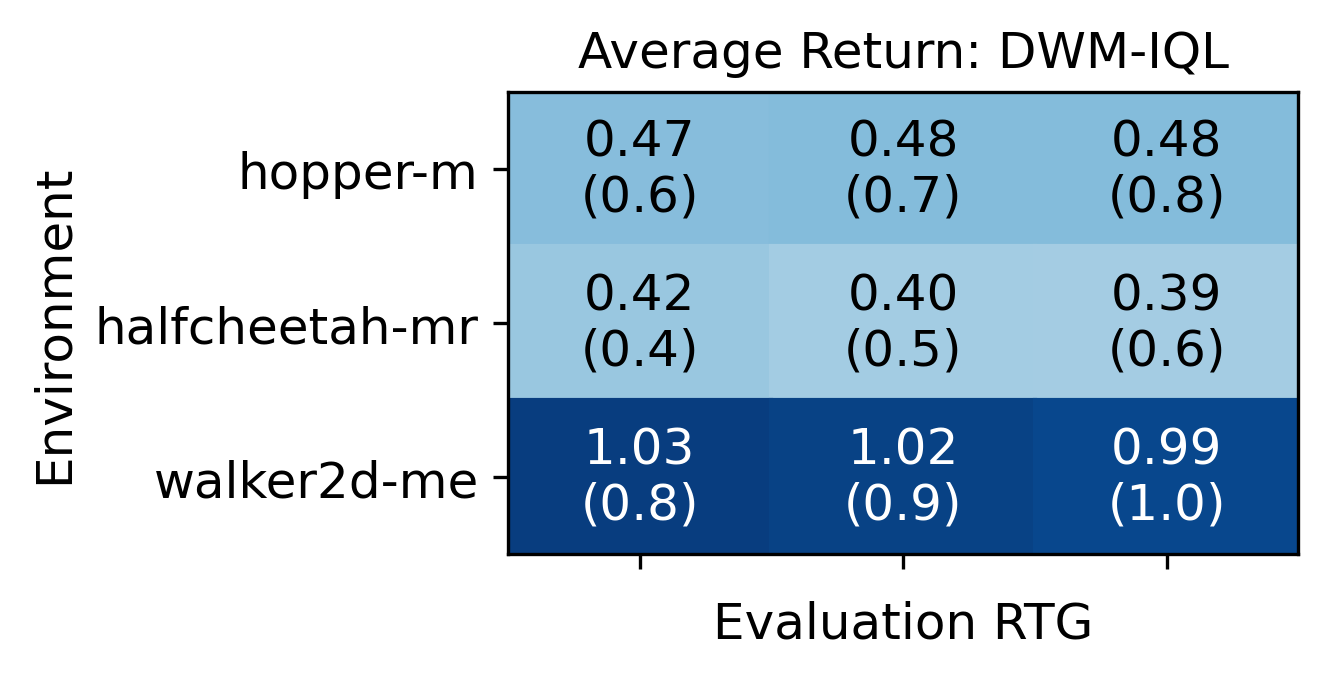}
    \includegraphics[height=0.2\columnwidth]{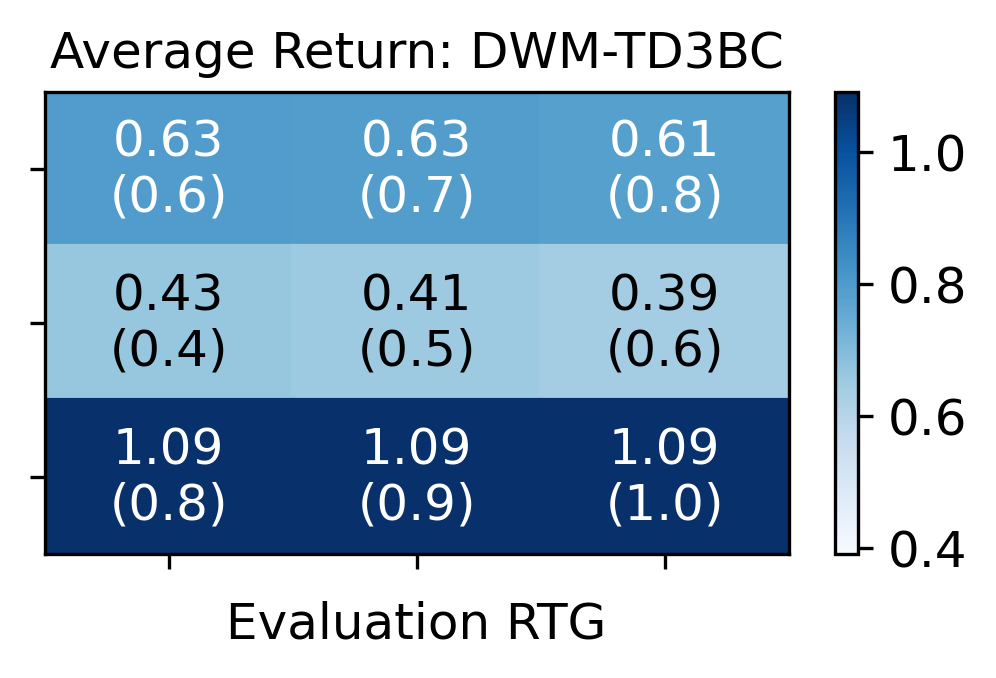}
    \caption{Comparison of DWM methods using different evaluation RTG values (displayed in parenthesis).}
    \label{fig:rtg}
\end{figure}

We found that the evaluation RTG values play a critical role in determining the performance of our algorithm. Our preliminary experiments on trajectory preidction have suggested that in distribution evaluation RTGs underperforms OOD RTGs, see Sec.~\ref{subsec:app_in_vs_ood}. Figure~\ref{fig:rtg} reports the return of DWM-IQL and DWM-TD3BC across 3 tasks, with different values of $g_\text{eval}$\footnote{We note that the return and RTG are normalized in different ways: the return computed by the D4RL benchmark is undiscounted and normalized by the performance of one SAC policy, whereas the RTG we use in training is discounted and normalized by hand-selected constants.}. We report the results averaged over different simulation horizons 1, 3, 5 and 7. 
The compared RTG values are different for each task, but are all OOD. Sec.~\ref{subsec:app_data_dist} shows the distributions of training RTGs for each task.  The results show that the actual return does not always match with the specified $g_\text{eval}$. This is a well-known issue of return-conditioned RL methods~\cite{emmons2021rvs, zheng2022online, nguyen2022conserweightive}. Nonetheless, 
OOD evaluation RTGs generally performs well. Figure~\ref{fig:rtg} shows both DWM-TD3BC and DWM-IQL are robust to OOD evaluation RTGs. We emphasize the reported return is averaged over training instances with different simulation horizons, where the peak performance, reported in Table~\ref{tab:compare_mbrl} is higher. Our intuition is to encourage the diffusion model to take
an optimistic view of the future return for the current state. On the other hand, the evaluation RTG cannot be overly high. As shown in task \textit{halfcheetah-mr}, increasing RTG $g_\text{eval}>0.4$ will further decrease the actual performances for both methods. The optimal RTG values vary from task to task, and the complete experiment results are provided below.

\subsubsection{Offline Data Distribution}
\label{subsec:app_data_dist}
\begin{figure}[H]
    \includegraphics[width=0.3\columnwidth]{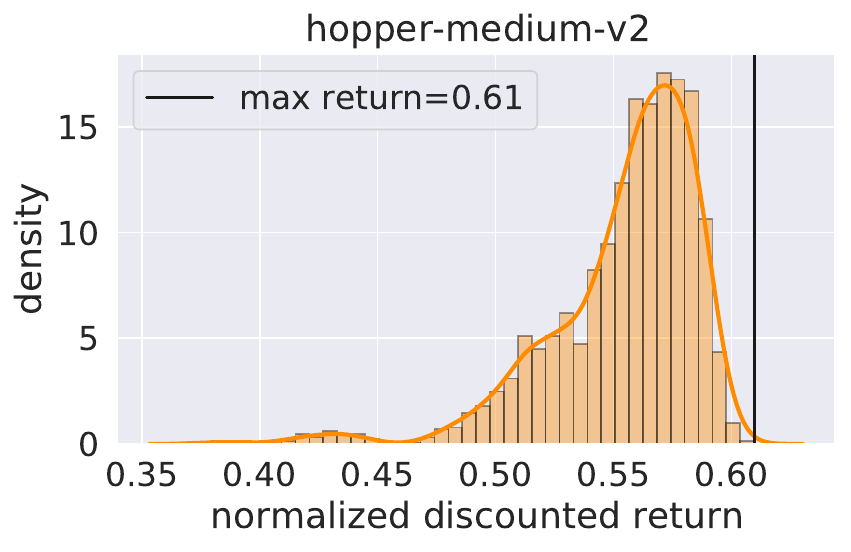}
    \includegraphics[width=0.3\columnwidth]{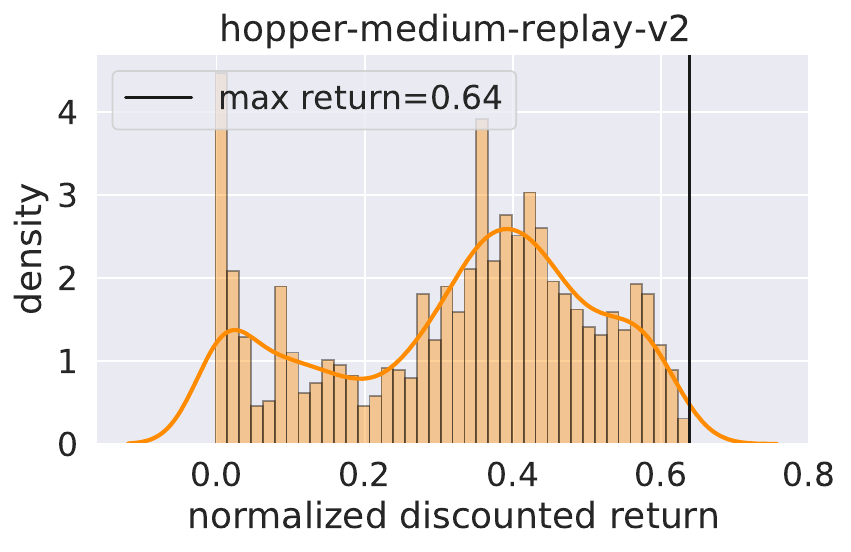}
    \includegraphics[width=0.3\columnwidth]{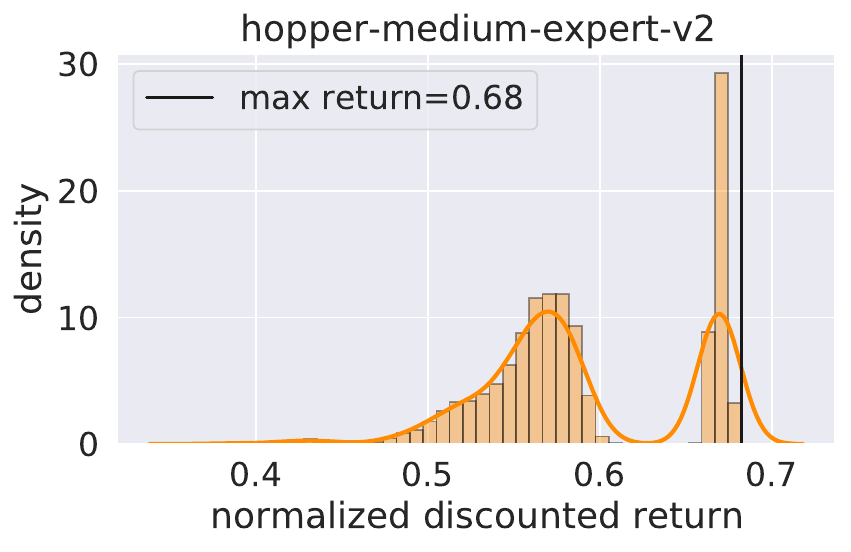}
    \includegraphics[width=0.3\columnwidth]{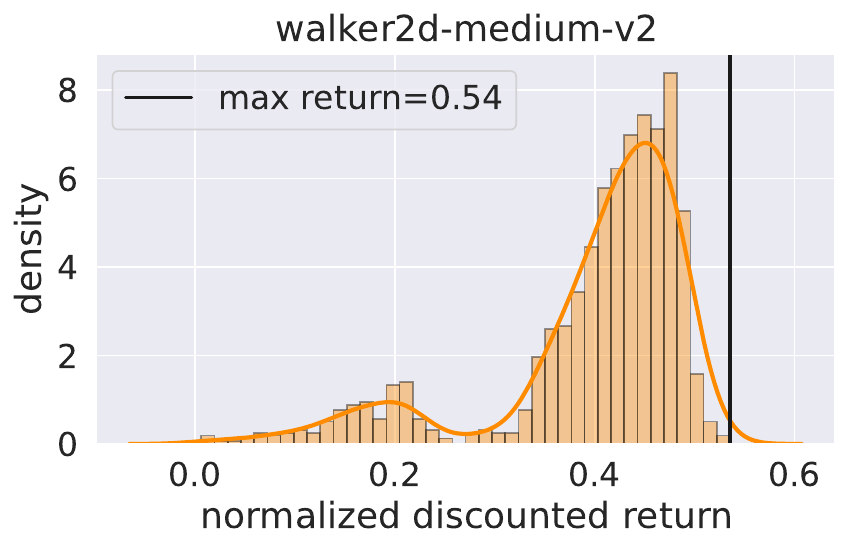}
    \includegraphics[width=0.3\columnwidth]{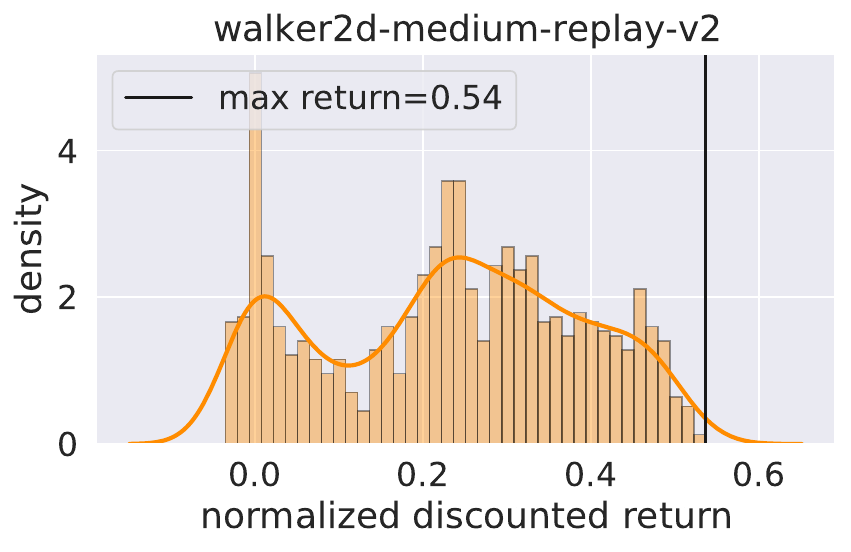}
    \includegraphics[width=0.3\columnwidth]{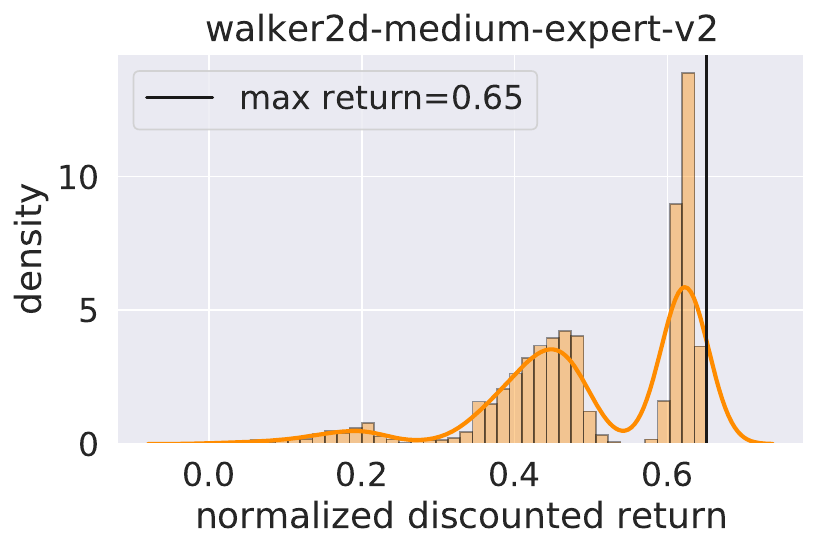}
    \includegraphics[width=0.3\columnwidth]{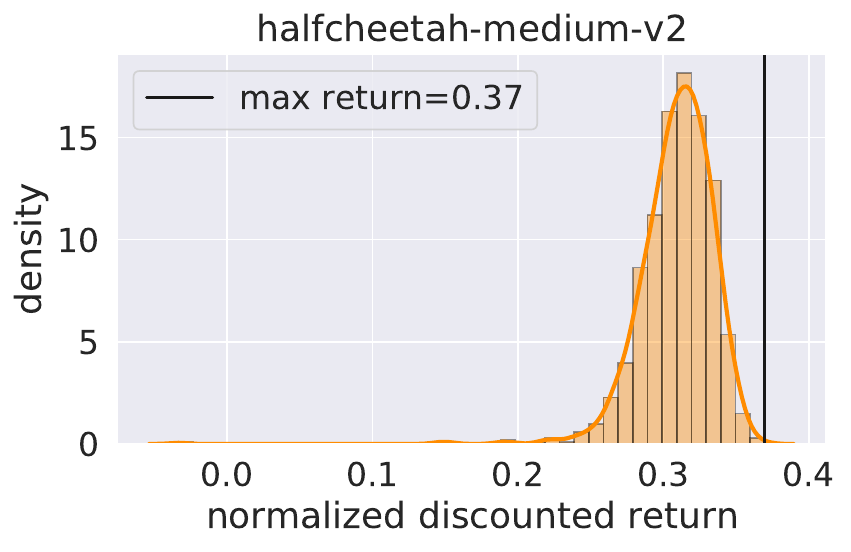}
    \includegraphics[width=0.33\columnwidth]{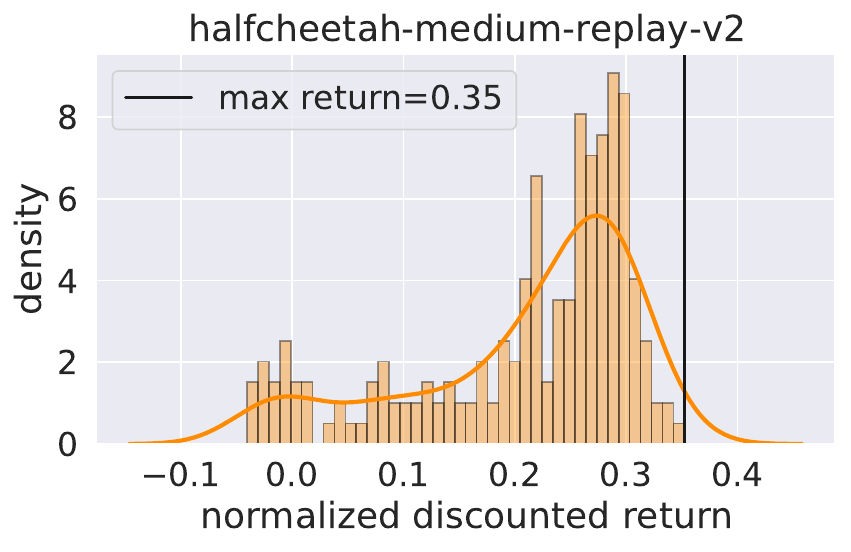}
    \includegraphics[width=0.34\columnwidth]{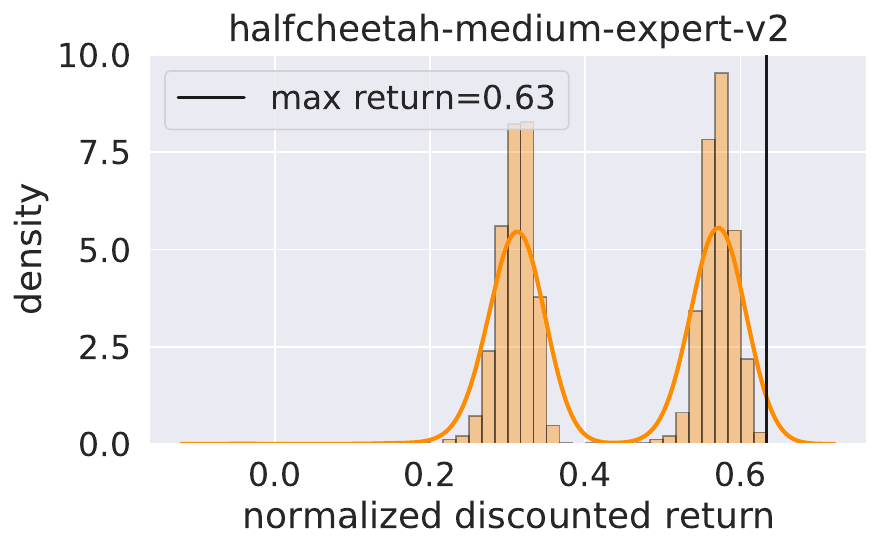}
    \caption{Normalized discounted returns for each environment.}
    \label{fig:norm_return}
\end{figure}

The normalized discounted return (as RTG labels in training) for the entire D4RL dataset over the nine tasks are analyzed in Fig.~\ref{fig:norm_return}. Compared with RTG values in our experiments as Table~\ref{tab:hyperparam}, 
the data maximum is usually smaller than the evaluation RTG values that leads to higher performances, as observed in our empirical experiments.

\subsubsection{In-Distribution v.s. Out-of-Distribution RTG}
\label{subsec:app_in_vs_ood}
\begin{figure}[H]
    \includegraphics[width=\columnwidth]{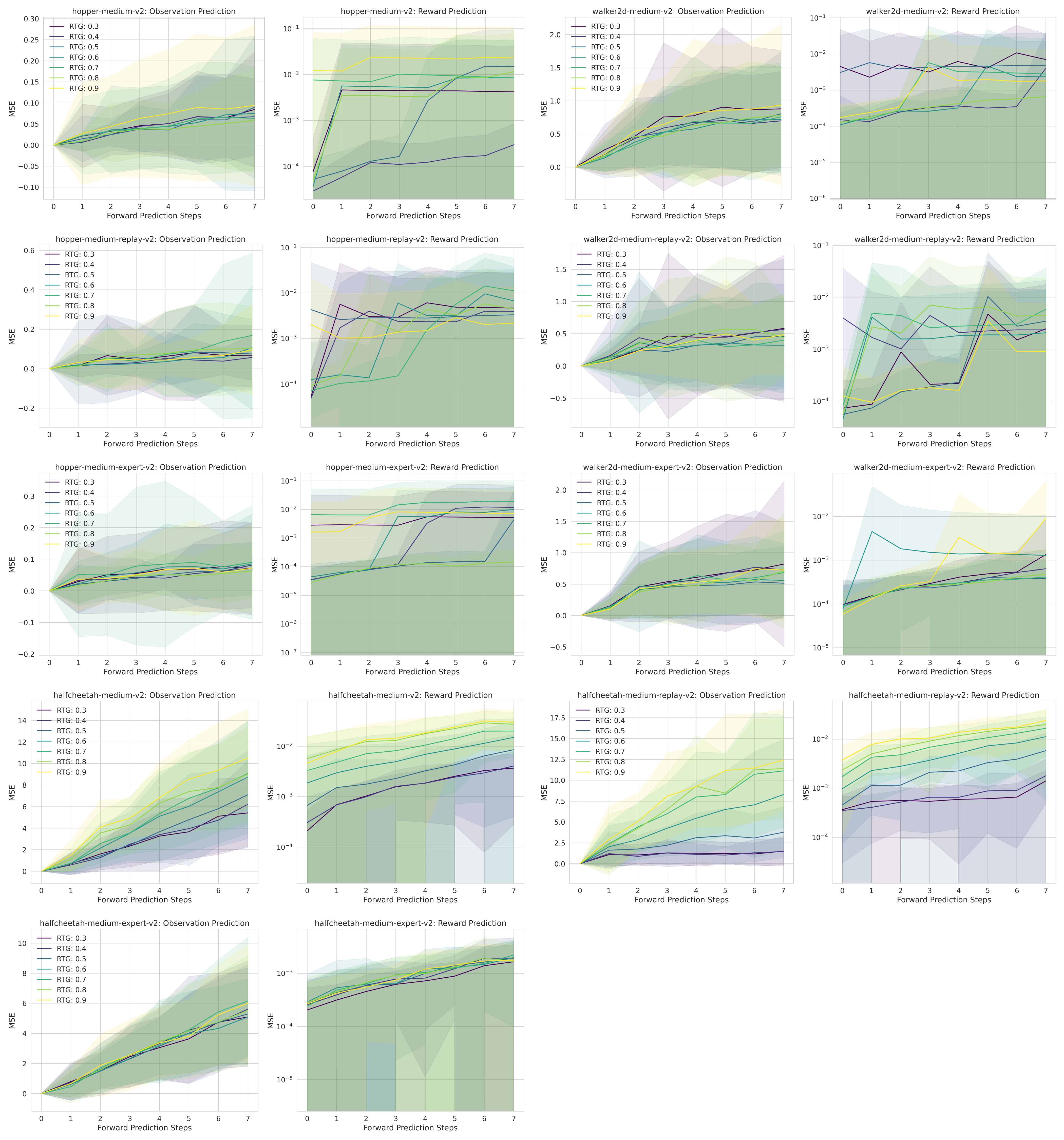}
    \caption{The breakdown prediction errors of DWM at each prediction timestep with different RTGs. The DWM is trained with $T=8$ and $K=5$.}
    \label{fig:breakdown_prediction_error_T8}
\end{figure}

\begin{figure}[H]
    \includegraphics[width=\columnwidth]{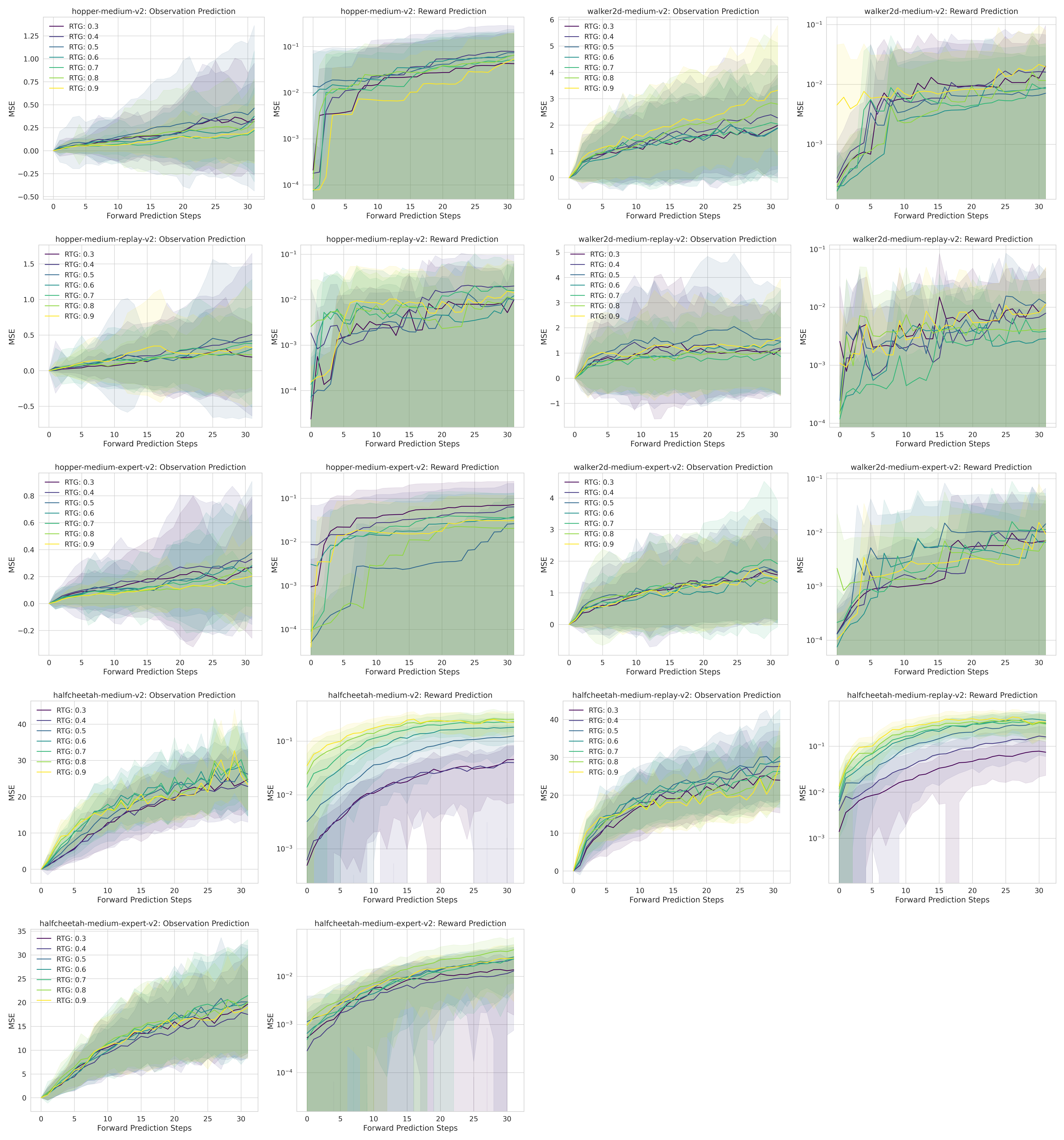}
    \caption{The breakdown prediction errors of DWM at each prediction timestep with different RTG.  The DWM is trained with $T=32$ and $K=10$.}
    \label{fig:breakdown_prediction_error_T32}
\end{figure}

Analogous to Equation~\eqref{eq:avg_pred_error}, we can define the breakdown prediction errors for each timestep $t'$, $t \leq t' \leq t+T-1$.
Figure~\ref{fig:breakdown_prediction_error_T8} and Figure~\ref{fig:breakdown_prediction_error_T32} plot the results,
using different values of $g_\text{eval}$. The OOD RTGs generally perform better.

It is worth noting that the prediction error naturally grows as the horizon increases. Intuitively, given a fixed environment, the initial state of the D4RL datasets are very similar, whereas the subsequent states after multiple timesteps become quite different.

\subsection{Ablation: $\lambda$-Return Value Estimation}
\label{app:lambda_return}
The Dreamer series of work~\cite{hafner2019dream, hafner2020mastering, hafner2023mastering} applies the $\lambda$-return technique \cite{schulman2015high} for value estimation, used the imagined trajectory.
This technique can be seamlessly embedded into our framework as a modification of the standard Diff-MVE.
More precisely, given a state-action pair $(s_t, a_t)$ sampled from the offline dataset,
we recursively compute the $\lambda$-target value for $h=H, \ldots, 0$:
\begin{align}
  \hat{Q}^\lambda_{t+h} = \hat{r}_{t+h} + \gamma \begin{cases} (1-\lambda)Q_{\bar{\phi}}(\hat{s}_{t+h+1}, \pi_{\bar{\psi}}(\hat{s}_{t+h+1})) + \lambda \hat{Q}^\lambda_{t+h+1} & \text{ if }h<H  \\ Q_{\bar{\phi}}(\hat{s}_{t+H}, \pi_{\bar{\psi}}(\hat{s}_{t+H})) & \text{ if }h=H \end{cases} 
\label{eq:lambda_return}
\end{align}
using DWM predicted states $\set{\hat{s}_{t+h}}_{h=0}^H$ and rewards $\{\hat{r}_t\}_{h=0}^H$.
We can use $\hat{Q}^\lambda_t$ as the target $Q$ value for TD learning, as a modification of  Algorithm~\ref{algo:diffusion_mbrl}. 
For algorithms that also learn the state-only value function, like IQL, the $Q_{\bar{\phi}}$ function can be replaced by the $V_{\bar{\psi}}$ function.
Worth noting, Equation~\eqref{eq:lambda_return} reduces to the vanilla Diff-MVE when $\lambda=1$. \looseness=-1

We conduct experiments to compare the vanilla diff-MVE and the $\lambda$-return variant for DWM-TD3BC and DWM-IQL, using $\lambda=0.95$. 
We search over RTG values (specified in Sec. Table~\ref{tab:hyperparam})  and simulation horizons $1,3,5,7$. The results are summarized in Table~\ref{tab:compare_lambda_return}. The $\lambda$-return technique
is beneficial for DWM-IQL, but harmful for DWM-TD3BC. We speculate that since Equation~\eqref{eq:lambda_return} iteratively invokes the $Q_{\bar{\phi}}$ or the $V_{\bar{\psi}}$ function, it favors approaches with more accurate value estimations. While IQL regularizes the value functions, TD3+BC only 
has policy regularization and is shown to be more prone to the value over-estimation issue in our experiments. 
Based on these results, we incorporated the $\lambda$-return technique into DWM-IQL, but let DWM-TD3BC use the vanilla Diff-MVE.
We let DWM-PQL uses the vanilla Diff-MVE for the sake of algorithmic simplicity.

\begin{table}[htbp]
\centering
\small
\begin{tabular}{c|cc|cc}
\toprule
 & \multicolumn{2}{c|}{\textbf{DWM-TD3BC}} & \multicolumn{2}{c}{\textbf{DWM-IQL}}  \\  
 Env. & w/o $\lambda$ &  w/ $\lambda$  & w/o $\lambda$ &  w/ $\lambda$  \\  \midrule
hopper-m  & 0.65 $\pm$ 0.10 & \textbf{0.68 $\pm$ 0.13} & 0.50 $\pm$ 0.08 & 0.54 $\pm$ 0.11 \\
walker2d-m & 0.70 $\pm$ 0.15 & 0.74 $\pm$ 0.08  & 0.62 $\pm$ 0.19 & \textbf{0.76 $\pm$ 0.05}\\
halfcheetah-m & \textbf{0.46 $\pm$ 0.01} & 0.40 $\pm$ 0.01  & \textbf{0.46 $\pm$ 0.01} & 0.44 $\pm$ 0.01    \\
hopper-mr & 0.53 $\pm$ 0.09 & 0.50 $\pm$ 0.23  & 0.29 $\pm$ 0.04 & \textbf{0.61 $\pm$ 0.13} \\
walker2d-mr & \textbf{0.46 $\pm$ 0.19} & 0.23 $\pm$ 0.10 & 0.27 $\pm$ 0.09 &  0.35 $\pm$ 0.14  \\
halfcheetah-mr & \textbf{0.43 $\pm$ 0.01} & 0.39 $\pm$ 0.02 & \textbf{0.43 $\pm$ 0.01} & 0.41 $\pm$ 0.01 \\
hopper-me & 1.03 $\pm$ 0.14 & \textbf{1.05 $\pm$ 0.16}  & 0.78 $\pm$ 0.24 & 0.90 $\pm$ 0.25  \\
walker2d-me & \textbf{1.10 $\pm$ 0.00} & 0.89 $\pm$ 0.13 & 1.08 $\pm$ 0.03 & 1.04 $\pm$ 0.10 \\
halfcheetah-me & \textbf{0.75 $\pm$ 0.16} & 0.71 $\pm$ 0.22 & 0.73 $\pm$ 0.14 & 0.74 $\pm$ 0.16  \\ \hline
Avg. & \textbf{0.68} &  0.62 & 0.57 & 0.64   \\
\bottomrule
\end{tabular}
\caption{Comparison of the performance of DWM methods using vanilla Diff-MVE and the $\lambda$-return variant. }
\label{tab:compare_lambda_return}
\end{table}

\subsection{Ablation: RTG Relabeling and Model Fine-tuning}
\label{app:rtg_relabel}
Unlike dynamic programming in traditional RL, sequential modeling methods like diffusion models and DT are suspected to fail to stitch suboptimal trajectories. RTG relabeling is proposed to alleviate this problem for DT~\cite{yamagata2023q}, through iteratively relabeling RTG $g$ from training dataset to be: 
\begin{equation}
    \tilde{g}_t = r_t + \gamma \max (g_{t+1}, \hat{V}(s_{t+1})) = \max(g_t, r_t + \hat{V}(s_{t+1}),
    \label{eq:rtg_relabel_discounted}
\end{equation}
where the $\hat{V}$ function is separately learned from the offline dataset using CQL~\cite{kumar2020conservative}, and the max operator is used to prevent underestimation due to the pessimism of $\hat{V}$. The original formulation by \citet{yamagata2023q} does not include the $\gamma$ term as DT uses undiscounted RTG, i.e., $\gamma=1$.

We apply the RTG relabeling for DWM fine-tuning in the policy learning phase of vanilla DWM-IQL algorithm, without the $\lambda$-return technique. The value function $\hat{V}$ comes from the IQL algorithm. We take the first $10\%$ steps of the entire policy learning as warm up steps, where we do not apply RTG relabeling. This is because 
$\hat{V}$ can be inaccurate at the beginning of training. The modified algorithm DWM-IQL(R) achieves an average score of $0.61$, improved over score $0.57$ for DWM-IQL(w/o $\lambda$), under exactly the same training and test settings. Results are provided in Table~\ref{tab:compare_rtg_relabel}. Nonetheless, the improvement is of limited unity compared with the $\lambda$-return, thus we do not include it in the final design.

\begin{table}[H]
\centering
\small
\begin{tabular}{ccccc}
\toprule
Env & DWM-IQL(w/o $\lambda$) & DWM-IQL(w/ RTG Relabel)   \\  \hline
hopper-m  & 0.50 $\pm$ 0.08   & 0.59 ± 0.13\\
walker2d-m & 0.62 $\pm$ 0.19   & 0.65 ± 0.17 \\
halfcheetah-m & 0.46 $\pm$ 0.01  & 0.47 ± 0.01   \\
hopper-mr & 0.29 $\pm$ 0.04  & 0.27 ± 0.02 \\
walker2d-mr & 0.27 $\pm$ 0.09  & 0.32 ± 0.15 \\
halfcheetah-mr & 0.43 $\pm$ 0.01  & 0.43 ± 0.02 \\
hopper-me & 0.78 $\pm$ 0.24 & 0.88 ± 0.26 \\
walker2d-me & 1.08 $\pm$ 0.03  & 1.1 ± 0.0 \\
halfcheetah-me& 0.73 $\pm$ 0.14  & 0.79 ± 0.10 \\ \hline
Avg. & 0.57 & 0.61   \\
\bottomrule
\end{tabular}
\caption{The results of finetuning DMW via RTG relabeling in the policy training phase: normalized return (mean $\pm$ std) }
\label{tab:compare_rtg_relabel}
\end{table}


\chapter{Consistency Models as Reinforcement Learning Policy\label{ch:consistency_policy}}
\begin{center}
\begin{quote}
This section is based on paper ``\textit{Consistency Models as a Rich and Efficient Policy Class for Reinforcement Learning}''~\cite{ding2023consistency} written in collaboration with Chi Jin, previously published at ICLR 2024.
\end{quote}
\end{center}

\section{Introduction}
Parameterized policy representation is an important component for policy-based deep reinforcement learning (DRL)~\citep{sutton2018reinforcement, arulkumaran2017deep, dong2020deep}. Prior works have developed a variety of policy parameterization methods. For discrete action space, existing policy parameterization includes Softmax action preferences~\citep{sutton2018reinforcement}, Gumbel-Softmax for categorical distributions~\citep{jang2016categorical}, decision trees~\citep{frosst2017distilling, ding2020cdt}, etc. For continuous action space, the most typical choice is unimodal Gaussian distribution. However, in practice the demonstration dataset often encompasses samples from a mixture of behavior policies. To capture the multi-modality in data distribution, Gaussian mixture model (GMM)~\citep{jacobs1991adaptive, ren2021probabilistic}, variational auto-encoders (VAE)~\citep{kingma2013auto, kumar2019stabilizing}, denoising diffusion probabilistic model (DDPM)~\citep{ho2020denoising, song2020score, wang2022diffusion, chi2023diffusion, hansen2023idql, venkatraman2023reasoning} are adopted as policy representation. 

The desiderata for policy representation in DRL includes: 1. The strong expressiveness of the function class is found to be critical for modeling multi-modal data distribution in offline RL~\citep{wang2022diffusion} or imitation learning (IL)~\citep{chi2023diffusion}; 2. Differentiability of the model is usually required for ease of optimization with stochastic gradient descent; 3. Computational and time efficiency for sampling can be essential for RL agents learning from interactions with environments. Previous works with action diffusion models (\emph{i.e.}, diffusion policy) testify the expressiveness of diffusion models for multi-modal action distributions~\citep{wang2022diffusion,chi2023diffusion, hansen2023idql, janner2022planning, ajay2022conditional}. Although GMM and VAE also capture multi-modality, diffusion models with large sampling steps are found to be more expressive for IL and offline RL scenarios~\citep{wang2022diffusion,chi2023diffusion}. However, it is known that the diffusion model with progressive denoising over a large number of steps can lead to slow sampling speed. The action inference can be a critical bottleneck for online RL heavily depends on sampling from environments. A direct usage of \textit{diffusion policies for online settings with policy gradient for optimization requires backpropagating through the diffusion networks for the number of sampling steps}, which is not scalable for its \textit{large time consumption and memory occupancy}. Consistency models~\citep{song2023consistency} based on probability flow ordinary differential equation (ODE) is proposed as a rescue with comparable performances as diffusion models but much less computational time, which allows few-step generation process thus significantly reduce the time consumption at inference stage.

This paper takes the first step adapting the consistency model--an expressive yet efficient generative model--as policy representation for DRL. The consistency policy is embedded in both behavioral cloning (BC) method and an actor-critic (AC) algorithm, namely Consistency-BC and Consistency-AC. Experimental evaluation includes three typical RL settings: offline, offline-to-online and online. Policies with two generative models--diffusion model and consistency model--are thoroughly compared in all three settings on D4RL dataset~\citep{fu2020d4rl}. For offline RL, we propose a new loss scaling for stabilizing the training process of consistency policy with policy regularization, and demonstrate the expressiveness of two generative policy models. This is illustrated by showing BC with an expressive model like diffusion or consistency provides fairly good policies outperforming some previous offline RL methods. The performances are further improved by leveraging the actor-critic style algorithm with necessary policy regularization to avoid generating out-of-distribution actions. The fast sampling process of the consistency policy not only helps to reduce the training time, \emph{e.g.}, by $43\%$ for offline BC, but more importantly, improves the time efficiency for online interaction in the environments by accelerating action inference. For offline-to-online setting with initialized models trained on offline dataset and online setting with learning from scratch, the consistency policy shows comparable or even higher performances than the diffusion policy in some tasks, using significantly shorter wall-clock time for training and inference. The source code is available\footnote{\href{https://github.com/quantumiracle/Consistency_Model_For_Reinforcement_Learning}{https://github.com/quantumiracle/Consistency\_Model\_For\_Reinforcement\_Learning}}. 

\section{Related Works}

\paragraph{Offline and Offline-to-Online RL.}
The \textit{offline} RL is the problem of policy optimization with a fixed dataset. It is well known for suffering from the value overestimation problem for out-of-distribution states and actions from the dataset. Existing methods for solving this issue fall into categories of (1) explicitly constraining the learning policy with offline data using batch constraining, behavior cloning (BC) or divergence constraints (\emph{e.g.}, Kullback-Leibler, maximum mean discrepancy), including algorithms Batch-Constrained deep Q-learning (BCQ)~\citep{fujimoto2019off}, TD3+BC~\citep{fujimoto2021minimalist}, Onestep RL~\citep{brandfonbrener2021offline}, Advantage Weighted Actor-Critic (AWAC)~\citep{nair2020awac}, Bootstrapping Error Accumulation Reduction (BEAR)~\citep{kumar2019stabilizing}, BRAC~\citep{wu2019behavior}, Diffusion Q-learning (Diffusion QL)~\citep{wang2022diffusion}, Extreme Q-learning ($\mathcal{X}$-QL)~\citep{garg2023extreme} and Actor-Restricted Q-learning (ARQ)~\citep{goo2022know}, or (2) implicit regularization with pessimistic value estimation, like Conservative Q-learning (CQL)~\citep{kumar2020conservative}, Random Ensemble Mixture (REM)~\citep{agarwal2020optimistic}, Implicit Q-learning  (IQL)~\citep{kostrikov2021offline}, Implicit Diffusion Q-learning (IDQL)~\cite{hansen2023idql}, Model-based Offline Policy Optimization (MOPO)~\citep{yu2020mopo}, etc. 
MoRel~\citep{kidambi2020morel} is a model-based offline RL algorithm constructing pessimistic MDP for learning conservative policies, which does not clearly fall into above two categories. 

\textit{Offline-to-online} RL usually suffers from a catastrophic degraded performance at initial online training stage, due to the distribution shift of training samples. Previous research has studied online fine-tuning with offline data or pre-trained policies, including Hybrid Q learning~\citep{song2022hybrid}, RLPD~\citep{ball2023efficient}, Cal-QL~\citep{nakamoto2023cal}, Action-free Guide~\citep{zhu2023guiding}, Actor-Critic Alignment (ACA)~\citep{yu2023actor} and \cite{lee2022offline}.

\paragraph{Score-based Generative Model for RL.}
For policy representation in RL, recent work also uses Denoising Diffusion Probabilistic Models (DDPM)~\citep{ho2020denoising, song2020score}, which we loosely refer to as the diffusion model (original diffusion model traces back to \cite{sohl2015deep}) in this paper, to capture the multi-modal distributions in offline dataset. Diffusion QL~\citep{wang2022diffusion} uses diffusion model for policy representation in the Q-learning+BC approach. Implicit Diffusion Q-learning (IDQL)~\citep{hansen2023idql} is a variant of IQL using diffusion policy. Diffusion policies~\citep{chi2023diffusion} applies diffusion models for policy representation under imitation learning settings in robotics domain. Diffuser~\citep{janner2022planning} and Decision Diffuser~\citep{ajay2022conditional} combines decision transformer architecture with diffusion models for model-based reinforcement learning from offline dataset. Diffusion policies are also used for goal-conditioned imitation learning~\citep{reuss2023goal} and human behavior imitation~\citep{pearce2023imitating}. Q-guided Policy Optimization (QGPO)~\citep{lu2023contrastive} proposes a new formulation for intermediate guidance in diffusion sampling process. Latent Diffusion-Constrained Q-Learning (LDCQ)~\citep{venkatraman2023reasoning} proposes to apply latent diffusion model with a batch-constrained Q value to handle the stitching issue and the extrapolation errors for offline dataset.

\section{Preliminaries}
\subsection{Offline and Online RL}
For RL, we define a Markov decision process $(\mathcal{S}, \mathcal{A}, R, \mathcal{T}, \rho_0, \gamma)$, where $\mathcal{S}$ is the state space, $\mathcal{A}$ is the action space, $R(s,a):\mathcal{S}\times \mathcal{A}\rightarrow \mathbb{R}$ is the reward function, $\mathcal{T}(s'|s,a):\mathcal{S}\times \mathcal{A}\rightarrow \Pr(\mathcal{S})$ is the stochastic transition function, $\rho_0(s_0):\mathcal{S}\rightarrow \Pr(\mathcal{S})$ is the initial state distribution, and $\gamma\in[0,1]$ is the discount factor for value estimation. A stochastic policy $\pi(a|s):\mathcal{S}\rightarrow \Pr(\mathcal{A})$ determines the action $a\in\mathcal{A}$ for the agent to take given its current state $s\in\mathcal{S}$, and the optimization objective for the policy is its discounted cumulative reward: $\mathbb{E}_\pi[\sum_{t=0}^\infty \gamma^t r(s_t, a_t)]$. For offline RL, there exist a dataset $\mathcal{D}=\{(s,a,r,s',\text{done})\}$ collected with some behavior policies $\pi^b$, and the current policy $\pi$ is set to be optimized with $\mathcal{D}$. For online RL, the agent is allowed collect samples through interacting with the environment to compose an online training dataset $\tilde{\mathcal{D}}$ for optimizing its policy. We consider parameterized policy representation as $\pi_\theta$.

\subsection{Consistency Model}
The diffusion model~\citep{ho2020denoising, song2020score} solves the multi-modal distribution matching problem with a stochastic differential equation (SDE), while the consistency model~\citep{song2023consistency} solves an equivalent probability flow ordinary differential equation (ODE): $\frac{d\mathbf{x}_\tau}{d\tau}=-\tau \nabla \log p_\tau(\mathbf{x})$ with $p_\tau(\mathbf{x})=p_\text{data}(\mathbf{x})\otimes \mathcal{N}(\mathbf{0}, \tau^2\mathbf{I})$ for time period $\tau\in[0, T]$, where $p_\text{data}(\mathbf{x})$ is the data distribution. The reverse process along the solution trajectory $\{\hat{\mathbf{x}}_\tau\}_{\tau\in[\epsilon, T]}$ of this ODE is the data generation process from initial random samples $\hat{\mathbf{x}}_T\sim\mathcal{N}(\mathbf{0}, T^2\mathbf{I})$, with $\epsilon$ as a small constant close to $0$ for handling numerical problem at the boundary. For speeding up the sampling process from a diffusion model, consistency model shrinks the required number of sampling steps to a much smaller value than the diffusion model, without hurting the model generation performance much. Specifically, it approximates a parameterized consistency function $f_\theta: (\mathbf{x}_\tau, \tau)\rightarrow \mathbf{x}_\epsilon$, which is defined as a map from the noisy sample $\mathbf{x}_\tau$ at step $\tau$ back to the original sample $\mathbf{x}_\epsilon$, instead of applying a step-by-step denoising function $p_\theta(\mathbf{x}_{\tau-1}|\mathbf{x}_\tau)$ as the reverse diffusion process in diffusion model. 
The training and inference details of consistency model refer to Sec.~\ref{app:consist_details}.
For modeling the conditional distribution with condition variable $c$, the consistency function is changed to be $f_\theta(c, \mathbf{x}_\tau, \tau)$, which is sightly different from original consistency model.

\section{Consistency Model as RL Policy}
The consistency model as policy representation in RL can be formulated in the following way. To map the consistency model to a policy in MDP, we set:
\begin{equation}
    c\triangleq s, \quad \mathbf{x}\triangleq a, \quad p_\text{data}(\mathbf{x})\triangleq p_\mathcal{D}(a|s), \quad \pi_\theta(s)\triangleq \texttt{Consistency\_Inference}(s;f_\theta)
\end{equation}
where $p_\mathcal{D}(a|s)$ is the action-state conditional distribution from offline dataset $\mathcal{D}$.
\paragraph{Consistency Action Inference.}
By setting the condition variable $c$ as state $s$ and generated variable $\mathbf{x}$ as action $a$, the consistency function $f_\theta$ can be used for generating actions from states following the conditional distribution of the dataset, \emph{i.e.}, a behavior RL policy. The parameterized policy $\pi_\theta$ is defined implicitly in terms of $f_\theta$, with which an action $a$ conditioned on state $s$ can be generated following the \texttt{Consistency\_Inference} as Alg.~\ref{alg:forward_a} with predetermined $\{\tau_n |n\in[N]\}$ sequence. During the inference process, a trained consistency model $f_\theta(s, \hat{a}_{\tau_n}, \tau_n)$ iteratively predicts denoised samples from the noisy inputs $\hat{a}_{\tau_n}=a +\sqrt{\tau_n^2-\epsilon^2}z$ along the probability flow ODE trajectory at step $n\in[N]$, with Gaussian noise $z\sim \mathcal{N}(\mathbf{0}, \mathbf{I})$. $\{\tau_n |n\in[N]\}$ is a sub-sequence of time points on a certain time period $[\epsilon, {T}]$ with $\tau_1=\epsilon, \tau_N=T$. For inference, the sub-sequence is a linspace of $[\epsilon, {T}]$ with $(N-1)$ sub-intervals.  
A single-step version of \texttt{Consistency\_Inference} can be achieved by just set $\{\tau_n|n=0,1\}=\{\epsilon, {T}\}$. Notice that $T$ here is the time horizon for denoising process in the consistency model instead of the episode length of the sample trajectory.

\paragraph{Consistency Behavior Cloning.}
With the offline dataset $\mathcal{D}$, the conditional consistency model as policy can be trained with loss by adapting the original~\citep{song2023consistency}:
\begin{equation}
    \mathcal{L}_c(\theta) = \mathbb{E}_{n\sim\mathcal{U}(1,N-1), (s, a)\sim\mathcal{D}, z\sim\mathcal{N}(\mathbf{0},\mathbf{I})}\Big[\lambda(\tau_n)d\big(f_\theta(s, a_{\tau_{n+1}}, \tau_{n+1}), f_{\theta^\intercal}(s, a_{\tau_n}, \tau_n)\big)\Big]
    \label{eq:consis_loss}
\end{equation}
where $\lambda(\cdot)$ is a step-dependent weight function, $a_{\tau_n}=a +\tau_n z$ and $d(\cdot, \cdot)$ is the distance metric. $f_{\theta^\intercal}$ is exponential moving average of $f_\theta$ for stabilizing the target estimation in training. In classical actor-critic algorithm, there exists the same delayed update of the policy network $\pi_{\theta^\intercal}$ (\emph{i.e.}, $f_{\theta^\intercal}$) for estimating target $Q$-values, which is set to coincide with the target in estimating the consistency loss. The setting for $\tau_n$ is detailed in Sec.~\ref{app:consist_details}.
Pseudo-code of Consistency BC refers to Alg.~\ref{alg:cbc}.

\begin{figure}[t]
\begin{minipage}{0.49\textwidth}
\begin{algorithm}[H]
\caption{Consistency Action Inference}
\begin{algorithmic}[1]
\STATE \textbf{Input}: $s, f_\theta, N, \{\tau_n\}_{n\in[N]}$
\STATE Initial $a\leftarrow f_\theta(s, \hat{a}_T, T), \hat{a}_T\sim\mathcal{N}(\mathbf{0},T^2\mathbf{I})$
\FOR{$n=N-1$ to $2$}
    \STATE $\hat{a}_{\tau_{n}}\leftarrow a +\sqrt{\tau_n^2-\epsilon^2}z$, $z\sim \mathcal{N}(\mathbf{0}, \mathbf{I})$
    \STATE $a\leftarrow f_\theta(s, \hat{a}_{\tau_{n}}, \tau_n)$
\ENDFOR
\STATE \textbf{Return}: $a$
\end{algorithmic}
\label{alg:forward_a}
\end{algorithm}
\begin{algorithm}[H]
\caption{Consistency Behavior Cloning}
\begin{algorithmic}[1]
\STATE \textbf{Input}: offline dataset $\mathcal{D}$
\STATE Initialize consistency policy $\pi_\theta$, target $\theta^\intercal \leftarrow \theta$
\FOR{iterations $k = 1,\ldots,K$}
    \STATE Update policy $\pi_\theta$ (with model $f_\theta)$ using loss $\mathcal{L}_c(\theta)$ as Eq.~\ref{eq:consis_loss};
    \STATE Update target: $\theta^\intercal \leftarrow \alpha \theta^\intercal +(1-\alpha)\theta$
\ENDFOR
\end{algorithmic}
\label{alg:cbc}
\end{algorithm}
\end{minipage}
\hfill
\begin{minipage}{0.49\textwidth}
\begin{algorithm}[H]
\caption{Offline Consistency Actor-Critic}
\begin{algorithmic}[1]
\STATE \textbf{Input}: offline dataset $\mathcal{D}$
\STATE Initialize consistency policy network $\pi_\theta$, critic networks $Q_{\phi_1}, Q_{\phi_2}$
\STATE Initialize target network parameters: $\theta^\intercal \leftarrow \theta$, $\phi_1^\intercal \leftarrow \phi_1, \phi_2^\intercal \leftarrow \phi_2$
\FOR{policy training iterations $k = 1,\ldots,K$}
    \STATE Sample minibatch $\mathcal{B}=\{(s,a,r,s')\}\subseteq\mathcal{D}$;
    \STATE {\color{blue}\% Q-value Update}
    \STATE Update $Q_{\phi_1}, Q_{\phi_2}$ with Eq.~\ref{eq:double_q_loss};
    \STATE {\color{blue}\% Policy Update}
    \STATE Update policy $\pi_\theta$ (with model $f_\theta)$ via Eq.~\ref{eq:policy_loss};
    \STATE {\color{blue}\% Target Update}
    \STATE Update target: $\theta^\intercal \leftarrow \alpha \theta^\intercal +(1-\alpha)\theta, \phi_i^\intercal \leftarrow \alpha \phi_i^\intercal + (1-\alpha)\phi_i, i\in\{1,2\}$;
\ENDFOR
\STATE \textbf{Return}: $\pi_\theta, Q_{\phi_1}, Q_{\phi_2}$
\end{algorithmic}
\label{alg:cac}
\end{algorithm}
\end{minipage}
\end{figure}

\paragraph{Consistency Actor-Critic.}
As as estimation of the state-action value of current policy, the parameterized $Q_\phi(s,a)$ function can be learned with the double Q-learning loss~\citep{fujimoto2018addressing} with batched data $\mathcal{B}\subseteq \mathcal{D}$:
\begin{align}
    \mathcal{L}(\phi)=\mathbb{E}_{(s,a,s')\sim \mathcal{B}, a'\sim\pi_{{\theta}^\intercal}(\cdot|s')}\bigg[\Big(\big(r(s,a)+\gamma \min_{i\in\{1,2\}}Q_{\phi^\intercal_i}(s',a')\big)- Q_{\phi_i}(s,a)\Big)^2\bigg]
    \label{eq:double_q_loss}
\end{align}
with $Q_{\phi^\intercal_i}$ as a delayed update of $Q_{\phi_i}, i\in\{1,2\}$ for stabilizing training.

The regularized policy $\pi_\theta$ on offline dataset is learned with a combination of policy gradient through maximizing the expected $Q_\phi(s,a)$ function and a behavior cloning regularization with consistency loss $\mathcal{L}_c(\theta)$:
\begin{align}
    \mathcal{L}(\theta)&=\mathcal{L}_c(\theta) + \eta \mathcal{L}_q(\theta) \label{eq:policy_loss} \\
    \text{where }\mathcal{L}_q(\theta)&=-\mathbb{E}_{s\sim\mathcal{B}, a\sim \pi_\theta(s)}\big[Q_\phi(s,a)\big]
    \label{eq:max_q_loss}
\end{align}
where $a\sim\pi_\theta(s)$ is action inference from the consistency policy as Alg.\ref{alg:forward_a}. It can be noticed that the actions generated with $N$ denoising steps will produce the policy gradients through the $Q_\phi(s,a)$ in above equation, thus it also backpropagates through $f_\theta$ for $N$ times in the gradient descent procedure, which can lead to additional time consumption apart from the multi-step model inference. Therefore, reducing the denoising steps $N$ can be critical for the speed of this type of models as RL policies. The consistency actor-critic (Consistency-AC) algorithm is provided in pseudo-code Alg.~\ref{alg:cac}.

\paragraph{Loss Scaling.} The consistency loss as Eq.~\ref{eq:consis_loss} matches the denoised predictions from two consecutive timesteps $\tau_n$ and $\tau_{n+1}$. Due to the usage of $N(k)$ schedule (detailed in Sec.~\ref{app:consist_details}), their difference $|\tau_{n+1}-\tau_n|$ decreases as the training iteration $k$ increases (thus $N(k)$ also increases), which allows the consistency model to have a coarse-to-fine matching process across different time scales. This also leads to a decreasing loss value $\mathcal{L}(\theta;k)$ as $k$ increases from 1 to $K$ since the predictions from smaller time intervals are easier to match.
Actually, the loss $\mathcal{L}_c(\theta)$ changes drastically across several magnitudes within an epoch, which leads to severe imbalance with the second loss term $\mathcal{L}_q(\theta)$ in Eq.~\ref{eq:policy_loss}. The coefficient $\eta$ is a constant hyperparameter independent of $k$, so it cannot help to alleviate this issue. Although original consistency model applies $\lambda(\tau_n)\equiv 1$ for image generation, we empirically find that in offline RL this imbalance of two loss terms can hurt the effect of policy regularization in some tasks, as evidenced by ablation studies in Sec.~\ref{sec:offline_cac}. To solve this issue, we propose a $k$-dependent weighting mechanism to balance the values of two loss terms. This is found to improve the performances of this policy regularization method with consistency model on offline RL. 
Specifically, $\lambda(\cdot)$ in Eq.~\ref{eq:consis_loss} is chosen to be: $\lambda(\tau_{n}, \tau_{n+1};k)=\frac{\xi}{|\tau_{n+1}(k) - \tau_{n}(k)|}$
where $\xi$ is set according to tasks (or absorbed in $\eta$). The denominator captures the loss scale at iteration $k$ conveniently. 

\section{Experimental Evaluation}

\begin{table}[b]
 \caption{The average scores of vanilla BC (with Gaussian), Consistency-BC, Diffusion-BC and several offline RL baselines on D4RL Gym, AntMaze, Adroit, and Kitchen tasks are shown. For Consistency-BC and Diffusion-BC, each cell has two values: one for offline model selection and another (in brackets) for online model selection. Each result is averaged over five random seeds with standard deviations reported. The bold values are the highest among each row.}
    \label{tab:offline_bc}
    \centering
    \resizebox{\textwidth}{!}{
    \begin{tblr}{
    colspec = {l||*{3}{c}|*{6}{c}},
    row{1, 12, 20, 24} = {font=\bfseries}
    }
    \toprule
        Gym Tasks & BC & Consistency-BC & Diffusion-BC & AWAC & Diffuser & MoRel & Onestep RL & TD3+BC & DT \\
        \hline
        halfcheetah-m & 42.6 & $31.0\pm0.4$ ($46.2\pm0.4$) & $45.4\pm1.8$ ($46.3\pm0.2$) & 43.5 & 44.2 & 42.1 & \textbf{48.4} & 48.3 & 42.6   \\
        hopper-m & 52.9 & $71.7\pm8.0$ ($78.3\pm2.6$) & $65.3\pm5.8$ ($71.1\pm5.5$) & 57.0 & 58.5 & \textbf{95.4} & 59.6 & 59.3 & 67.6 \\ 
        walker2d-m & 75.3 & $83.1\pm0.3$ ($84.1\pm0.3$)  & $81.2\pm1.7$ ($84.3\pm0.5$) & 72.4 & 79.7 & 77.8 & 81.8 & \textbf{83.7} & 74.0 \\
        halfcheetah-mr & 36.6 & $34.4\pm5.3$ ($45.4\pm0.7$) & $41.7\pm0.4$ ($44.1\pm0.3$) & 40.5 & \textbf{42.2} & 40.2 & 38.1 & 44.6 & 36.6   \\
        hopper-mr & 18.1 & $\textbf{99.7}\pm0.5$ ($100.4\pm0.6$) & $67.9\pm28.1$ ($99.1\pm2.3$) & 37.2 & 96.8 & 93.6 & 97.5 & 60.9 & 82.7 \\
        walker2d-mr & 26.0 & $73.3\pm5.7$ ($80.8\pm2.4$) & $77.5\pm4.7$ ($80.8\pm4.5$) & 27.0 & 61.2 & 49.8 & 49.5 & \textbf{81.8} & 66.6  \\
        halfcheetah-me & 55.2 & $32.7\pm1.2$ ($39.6\pm3.4$) & $90.8\pm1.1$ ($93.5\pm0.4$) & 42.8 & 79.8 & 53.3 & \textbf{93.4} & 90.7 & 86.8 \\ 
        hopper-me & 52.5 & $90.6\pm9.3$ ($96.8\pm4.6$) & ${107.6}\pm4.3$ ($111.7\pm0.3$) & 55.8 & 107.2 & \textbf{108.7} & 103.3 & 98.0 & 107.6  \\ 
        walker2d-me & 107.5 & $110.4\pm0.7$ ($111.6\pm0.7$) & $108.9\pm0.6$ ($110.5\pm0.5$) & 74.5 & 108.4 & 95.6 & \textbf{113.0} & 110.1 & 108.1   \\ 
        \hline
        \textbf{Average} & 51.9 & 69.7 (75.9) & \textbf{76.3} (82.4) & 50.1 & 75.3 & 72.9 & 76.1 & 75.3 & 74.7   \\
    \bottomrule
    \toprule
        AntMaze Tasks & BC & Consistency-BC & Diffusion-BC & AWAC & BCQ & BEAR & Onestep RL & TD3+BC & DT \\ \hline
        antmaze-u & 54.6 & $75.8\pm4.0$ ($87.0\pm4.5$) & $71.8\pm8.2$ ($76.8\pm3.9$) & 56.7 & \textbf{78.9} & 73.0 & 64.3 & 78.6 & 59.2  \\
        antmaze-ud & 45.6 & $\textbf{77.6}\pm6.3$ ($82.4\pm3.4$) & $61.2\pm9.4$ ($78.8\pm7.0$) & 49.3 & 55.0 & 61.0 & 60.7 &  {71.4} & 53.0  \\
        antmaze-mp & 0.0 & $\textbf{56.8}\pm$30.1 ($71.6\pm14.5$) & $43.4\pm37.8$ ($56.8\pm34.5$) & 0.0 & 0.0 & 0.0 & 0.3 & 10.6 & 0.0  \\
        antmaze-md & 0.0 & $\textbf{31.6}\pm22.4$ ($66.0\pm6.5$) & $29.8\pm36.3$ ($69.4\pm12.3$) & 0.7 & 0.0 & 8.0 & 0.0 & 3.0 & 0.0  \\
        antmaze-lp & 0.0 & ${10.2}\pm4.6$ ($15.0\pm3.8$) & $\textbf{14.6}\pm11.2$ ($22.4\pm5.8$) & 0.0 & 6.7 & 0.0 & 0.0 & 0.2 & 0.0  \\
        antmaze-ld & 0.0 & ${12.8}\pm8.2$ ($19.8\pm4.0$) & $\textbf{26.6}\pm10.7$ ($33.0\pm8.2$) & 1.0 & 2.2 & 0.0 & 0.0 & 0.0 & 0.0   \\ \hline
        \textbf{Average} & 16.7 & \textbf{44.1} (57.0) & 41.2 (53.3) & 18.0 & 23.8 & 23.7 & 20.9 & 27.3 & 18.7  \\
    \bottomrule
    \toprule
        Adroit Tasks & BC & Consistency-BC & Diffusion-BC & SAC & BCQ & BEAR & BRAC-p & BRAC-v & REM  \\ \hline
        pen-human-v1 & 25.8 & $52.4\pm13.7$ ($63.7\pm7.4$) & $61.1\pm5.9$ ($66.7\pm4.9$) & 4.3 & \textbf{68.9} & -1.0 & 8.1 & 0.6 & 5.4 \\
        pen-cloned-v1 & 38.3 & $33.4\pm6.0$ ($51.9\pm6.6$)  & $\textbf{57.6}\pm9.5$ ($62.7\pm6.1$) & -0.8 & 44.0 & 26.5 & 1.6 & -2.5 & -1.0 \\ \hline
        \textbf{Average}  & 32.1 & 42.9 (57.8) & \textbf{59.4} (64.7) & 1.8 & 56.5 & 12.8 & 4.9 & -1.0 & 2.2  \\ 
    \bottomrule
    \toprule
        Kitchen Tasks & BC & Consistency-BC & Diffusion-BC & SAC & BCQ & BEAR & BRAC-p & BRAC-v & AWR  \\ \hline
        kitchen-c & 33.8 &  $45.2\pm5.0$ ($50.9\pm3.6$) & $\textbf{76.5}\pm8.9$ ($87.3\pm6.8$) & 15.0 & 8.1 & 0.0 & 0.0 & 0.0 & 0.0   \\
        kitchen-p & 33.8 & $22.6\pm3.8$ ($23.8\pm2.8$)  & $\textbf{50.3}\pm3.0$ ($52.9\pm1.6$) & 0.0 & 18.9 & 13.1 & 0.0 & 0.0 & 15.4  \\
        kitchen-m & 47.5 & $23.5\pm1.8$ ($24.3\pm1.3$) & $\textbf{56.5}\pm6.6$ ($64.7\pm4.6$) & 2.5 & 8.1 & 47.2 & 0.0 & 0.0 & 10.6  \\ \hline
        \textbf{Average} & 38.4 & 30.4 (33.0) & \textbf{61.1} & 5.8 & 11.7 & 20.1 & 0.0 & 0.0 & 8.7  \\ 
    \bottomrule
    \end{tblr}}
\end{table}

To evaluate the expressiveness and computational efficiency of the proposed consistency policy and corresponding algorithms, we conduct experiments on four task suites (Gym, AntMaze, Adroit, Kitchen) in D4RL benchmarks under three canonical RL settings: offline (Sec.~\ref{sec:offline_cbc}, Sec.~\ref{sec:offline_cac}), offline-to-online and online (Sec.~\ref{sec:online_cac}). It is known that the D4RL offline dataset can exhibit multi-modality since the samples may be collected with a mixture of polices or along various sub-optimal trajectories, which makes the expressiveness of policy representation critical~\citep{fu2020d4rl, wang2022diffusion}. For offline RL, the generative models as policies are evaluated with both behavior cloning (Consistency-BC, Diffusion-BC) and actor-critic type (Consistency-AC, Diffusion-QL) algorithms, in terms of both performances and computational time. The Diffusion-QL is also an actor-critic algorithm though with name QL. Variants of Consistency-AC are compared as ablation studies and the best performances are reported. For offline-to-online and online RL settings, the learning curves and final results are compared for different methods. For evaluation, each model is evaluated over 10 episodes for Gym tasks and 100 episodes for other tasks, following the settings in previous work~\citep{wang2022diffusion}. By default, the consistency policy applies the number of denoising steps $N=2$ with a saturated performances on most of D4RL tasks, while diffusion policy uses $N=5$~\citep{wang2022diffusion}. Effects of different choices of $N$ are discussed in Sec.~\ref{sec:offline_cac}.

\subsection{Offline RL: Behavior Cloning}
\label{sec:offline_cbc}

\textbf{Empirical finding 1:} \textit{By behavior cloning alone (without any RL component), using an expressive policy representation with multi-modality like the consistency or diffusion model achieves performances comparable to many existing popular offline RL methods. Learning consistency policy requires much less computation than learning diffusion policy.}

The proposed method \textbf{Consistency-BC} follows Alg.~\ref{alg:cbc} with consistency policy for behavior cloning, and \textbf{Diffusion-BC} is by replacing the policy representation with a diffusion model and replace the policy loss with the diffusion model training loss $\mathcal{L}_d(\theta)$ as specified in paper~\citep{wang2022diffusion}. Results for classic BC with Gaussian policies and previous offline RL baselines, including AWAC~\citep{nair2020awac}, Diffuser~\citep{janner2022planning}, MoRel~\citep{kidambi2020morel}, Onestep RL~\citep{brandfonbrener2021offline}, TD3+BC~\citep{fujimoto2021minimalist}, Decision Transformer (DT)~\citep{chen2021decision}, BCQ~\citep{fujimoto2019off}, BEAR~\citep{kumar2019stabilizing}, BRAC~\citep{wu2019behavior} and REM~\citep{agarwal2020optimistic}, are adopted from previous paper~\citep{wang2022diffusion}. SAC~\citep{haarnoja2018soft} is the algorithm used for collecting data in D4RL Gym tasks.

\begin{wrapfigure}{r}{0.52\textwidth}
	\centering\includegraphics[width=0.5\textwidth]{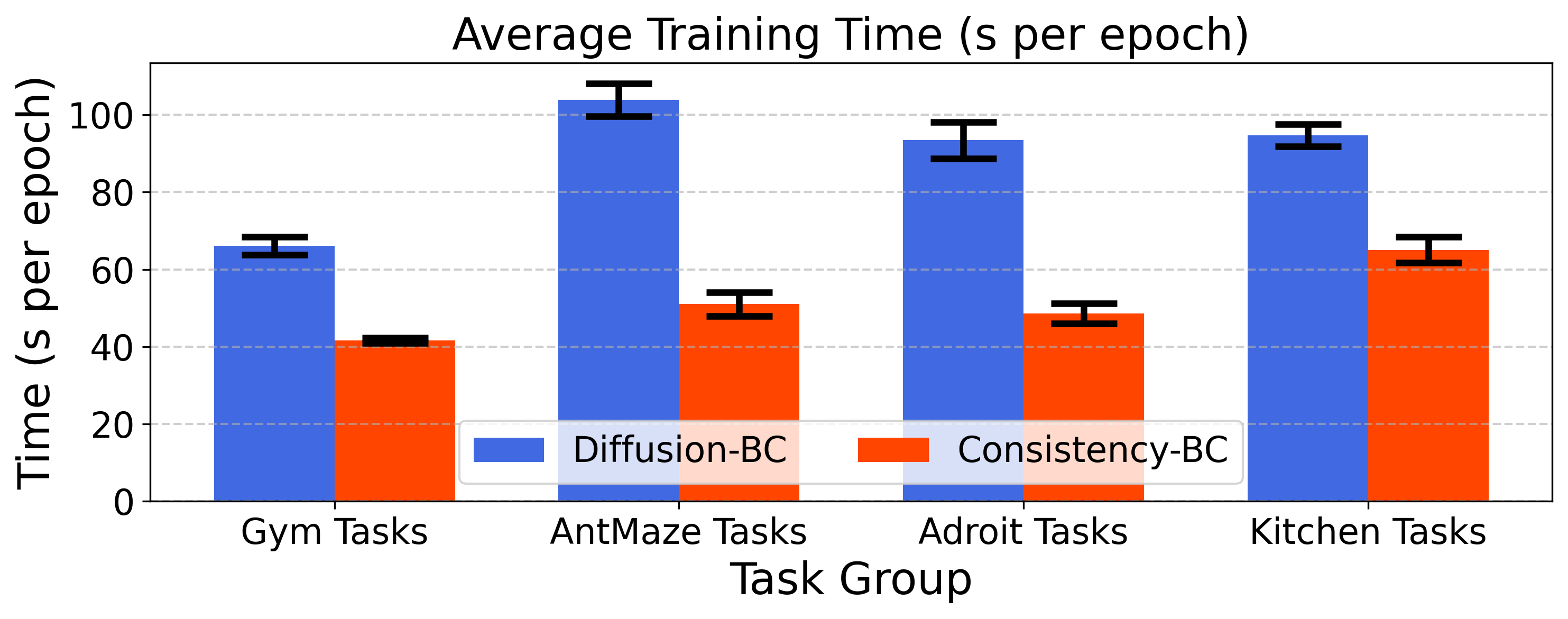}
    	\caption{Average training time (seconds per epoch) for Consistency-BC and Diffusion-BC across tasks.}
	\label{fig:offline_bc_time}
\end{wrapfigure}
Results from Tab.~\ref{tab:offline_bc} show the advantage of using multi-modal policy representation for offline RL even only with the BC method. For reference purpose, the values in the brackets allow for online evaluation to achieve the best model selection from the set of trained models, which serve as the maximal possible values for the standard offline selection without leveraging online evaluation. Compared with vanilla BC using the Gaussian distribution for policies, Consistency-BC with multi-modality outperforms it on $14/20$ tasks, and Diffusion-BC has better or equivalent performance as BC for $20/20$ tasks. Through leveraging multi-modal representation in BC, the improvement of normalized scores averaged over tasks is significant, and this is mainly caused by the multi-modality within the offline dataset by mixing over policies. Moreover, compared with previous offline RL baselines, which do not just apply BC, the Consistency-BC and Diffusion-BC show comparable performances, and even superior performances for tasks like \textit{walker2d-medium-v2}, \textit{hopper-medium-replay-v2}, \textit{walker2d-medium-replay-v2}, \textit{walker2d-medium-expert-v2} in Gym tasks, most of AntMaze, Adroit and Kitchen tasks. The consistency policy is slightly less expressive than the diffusion policy, which is within our expectation due to its heavy reduction on the sampling steps. However, the consistency policy shows higher computational efficiency than diffusion policy as compared in Fig.~\ref{fig:offline_bc_time}, with an average reduction of $\mathbf{42.97\%}$ computational time across 20 tasks. Detailed computational time for each task is provided in Sec.~\ref{app:offline_bc}, Tab.~\ref{tab:offline_train_time}.

\subsection{Offline RL: Consistency Actor-Critic}
\label{sec:offline_cac}

\begin{table}[htbp]
 \caption{The performance of Consistency-AC and SOTA baselines on D4RL Gym, AntMaze, Adroit and Kitchen tasks for offline RL setting. For Consistency-AC and Diffusion-QL, each cell has two values: one for offline model selection and another (in brackets) for online model selection. The bold values are the highest among each row. }
    \label{tab:offline_cac}
    \centering
    \resizebox{\textwidth}{!}{
    \begin{tblr}{
    colspec = {l||*{6}{c}|c},
    row{1} = {font=\bfseries}
    }
    \toprule
         Tasks & CQL & IQL & $\mathcal{X}$-QL & ARQ & IDQL-A  & Diffusion-QL &  Consistency-AC  \\
        \hline
        halfcheetah-m & 44.0 &  47.4 & 48.3 & 45 {\small $\pm$ 0.3}  &  51.0 & {51.1} {\small $\pm$ 0.5} (51.5 {\small $\pm$ 0.3}) & \textbf{69.1} {\small $\pm$ 0.7} (71.9 {\small $\pm$ 0.8}) \\
        hopper-m & 58.5 & 66.3 & 74.2 & 61 {\small $\pm$ 0.4} & 65.4 & \textbf{90.5} {\small $\pm$ 4.6} (96.6 {\small $\pm$ 3.4})  & 80.7 {\small $\pm$10.5} (99.7 {\small $\pm$2.3}) \\ 
        walker2d-m & 72.5 & 78.3 & 84.2 & 81 {\small $\pm$ 0.7} & 82.5 & \textbf{87.0} {\small $\pm$ 0.9} (87.3 {\small $\pm$ 0.5}) & 83.1 {\small $\pm$0.3} (84.1 {\small $\pm$0.3})\\
        halfcheetah-mr & 45.5  & 44.2 & 45.2 & 42 {\small $\pm$ 0.3} & 45.9 & {47.8} {\small $\pm$ 0.3} (48.3 {\small $\pm$ 0.2}) & \textbf{58.7} {\small $\pm$3.9} (62.7 {\small $\pm$0.6})\\
        hopper-mr & 95.0 & 94.7  & 100.7 & 81 {\small $\pm$ 24.2} & 92.1 & \textbf{101.3} {\small $\pm$ 0.6} (102.0 {\small $\pm$ 0.4}) & 99.7 {\small $\pm$ 0.5} (100.4 {\small $\pm$ 0.6})\\
        walker2d-mr & 77.2 & 73.9 & 82.2 & 66 {\small $\pm$ 7.0} & 85.1 & \textbf{95.5} {\small $\pm$ 1.5} (98.0 {\small $\pm$ 0.5})  & 79.5 {\small $\pm$ 3.6} (83.0 {\small $\pm$ 1.5})\\
        halfcheetah-me & 91.6 & 86.7  & 94.2 & 91 {\small $\pm$ 0.7} & 95.9 & \textbf{96.8} {\small $\pm$ 0.3} (97.2 {\small $\pm$ 0.4}) & 84.3 {\small $\pm$ 4.1} (89.2 {\small $\pm$ 3.3})\\ 
        hopper-me & 105.4 & 91.5 & \textbf{111.2} & 110 {\small $\pm$ 0.9} & 108.6 & 111.1 {\small $\pm$ 1.3}  (112.3 {\small $\pm$ 0.8}) & 100.4 {\small $\pm$ 3.5} (106.0 {\small $\pm$ 1.3})\\ 
        walker2d-me & 108.8 & 109.6 & \textbf{112.7 }& 109 {\small $\pm$ 0.5} & \textbf{112.7} & {110.1} {\small $\pm$ 0.3} (111.2 {\small $\pm$ 0.9}) & 110.4 {\small $\pm$ 0.7} (111.6 {\small $\pm$ 0.7})\\ 
        \hline
        \textbf{Average} & 77.6 & 77.0 & 83.7 & 76.2 & 82.1 & \textbf{87.9} (89.3) & 85.1 (89.8) \\
    \bottomrule
        antmaze-u  & 74.0 & 87.5 & \textbf{93.8} & 97 {\small $\pm$ 0.8} & 94.0 & 93.4 {\small $\pm$ 3.4} (96.0 {\small $\pm$ 3.3}) & 75.8 {\small $\pm$ 1.6} (85.6 {\small $\pm$ 3.9})\\
        antmaze-ud & \textbf{84.0} & 62.2 & 82.0 & 62 {\small $\pm$ 12.1} & 80.2 & {66.2} {\small $\pm$ 8.6} (84.0 {\small $\pm$ 10.1}) & 77.6 {\small $\pm$ 6.3} (82.4 {\small $\pm$ 3.4})\\
        antmaze-mp & 61.2 & 71.2 & 76.0 & 80 {\small $\pm$ 8.3} & \textbf{84.5} & 76.6 {\small $\pm$ 10.8} (79.8 {\small $\pm$ 8.7})  & 56.8 {\small $\pm$ 30.1} (71.6 {\small $\pm$ 14.5})\\ \hline
        \textbf{Average} & 73.1 & 73.6 & 83.9 & 79.7 & \textbf{86.2} & 78.7 (86.6) & 70.1 (79.9)\\
    \bottomrule
        pen-human-v1 & 35.2 & 71.5 & - & 45 {\small $\pm$ 5.2} (v0) & -& \textbf{72.8} {\small $\pm$ 9.6} (75.7 {\small $\pm$ 9.0})  & 63.4 {\small $\pm$ 7.7} (67.9 {\small $\pm$ 5.3})\\
        pen-cloned-v1 & 27.2 & 37.3 & - & 50 {\small $\pm$ 7.1} (v0) & -& \textbf{57.3} {\small $\pm$ 11.9} (60.8 {\small $\pm$ 11.8}) & 50.1 {\small $\pm$ 2.2} (53.7 {\small $\pm$ 3.4})\\ \hline
        \textbf{Average} & 31.2 &  54.4 & - & 47.5 & -& \textbf{65.1} (68.3) & 56.8 (60.8)\\ 
    \bottomrule
        kitchen-c & 43.8 & 62.5 & 82.4 & 37 {\small $\pm$ 14.2} &- & \textbf{84.0} {\small $\pm$ 7.4} (84.5 {\small $\pm$ 6.1}) & 51.9 {\small $\pm$ 6.0} (67.6 {\small $\pm$ 2.7}) \\
        kitchen-p  & 49.8 & 46.3 & \textbf{73.7} & 50 {\small $\pm$ 5.0} &- & 60.5 {\small $\pm$ 6.9} (63.7 {\small $\pm$ 5.2})& 38.2 {\small $\pm$ 1.8} (39.8 {\small $\pm$ 1.6}) \\
        kitchen-m  & 51.0  & 51.0 & 62.5 & 39 {\small $\pm$ 9.4} &- & \textbf{62.6} {\small $\pm$ 5.1} (66.6 {\small $\pm$ 3.3})& 45.8 {\small $\pm$ 1.5} (46.7 {\small $\pm$ 0.9})\\ \hline
        \textbf{Average} & 48.2 & 53.3 & \textbf{72.9} & 42.0 & - & 69.0 (71.6) & 45.3 (51.4) \\ 
    \bottomrule
    \toprule
     \textbf{Total Average} & 66.2 & 69.6 & - & 67.4 & - & \textbf{80.3} (83.2) & 72.1 (77.9)  \\
     \bottomrule
    \end{tblr}}
\end{table}

\textbf{Empirical finding 2:} \textit{Replacing diffusion model with consistency model in TD3-BC type algorithm for offline RL will lead to speed up of model training and inference, with slightly worse performances while still outperforming some other baselines.}

For offline RL, the proposed method \textbf{Consistency-AC} follows Alg.~\ref{alg:cac} with consistency model for policy representation, and the consistency policy is embedded in an actor-critic algorithm with BC policy regularization to avoid generating out-of-distribution actions. Results for previous baselines, including CQL~\citep{kumar2020conservative}, IQL~\citep{kostrikov2021offline}, $\mathcal{X}$-QL~\citep{garg2023extreme}, ARQ~\citep{goo2022know}, IDQL-A~\citep{hansen2023idql} and Diffusion-QL~\citep{wang2022diffusion} are adopted from results reported in corresponding papers.

Results from Tab.~\ref{tab:offline_cac} show the average normalized scores of different methods across five random seeds, with standard deviations reported for Diffusion-QL and Consistency-AC. The results in Tab.~\ref{tab:offline_cac} are directly comparable with the results in Tab.~\ref{tab:offline_bc} since they follow the same offline RL setting.

Tab.~\ref{tab:offline_cac} shows that although Consistency-AC achieves a slightly lower average score ($72.1$) than Diffusion-QL ($80.3$), it outperforms the other baselines in most of the tasks, like Gym and Adroit. The AntMaze tasks are found to be hard for the Consistency-AC method, we conjecture that this is potentially caused by the sparse reward signals in the dataset, which makes the difficulty of Q-learning become more of a bottleneck than modeling the multi-modal distributions with behavior cloning.
The conservativeness of the $Q$-value estimation might be important but orthogonal to the proposed consistency policy. Considering the reduction of denoising steps in the training and inference stages of Consistency-AC, it can be regarded as a trade-off between the computational efficiency and the approximation accuracy of multi-modal distribution, which will be discussed as following.

\paragraph{Computational Time.}

\begin{table}[htbp]
\resizebox{\columnwidth}{!}{ 
\footnotesize
\begin{tabular}{c||c|ccc}
\toprule
       \textbf{Method} &  $N$ &\multirow{1}{*}{\textbf{Training Time (s per epoch)}} & \multicolumn{1}{c}{\textbf{Inference Time (ms per sample)}} & \multicolumn{1}{c}{\textbf{Avg. Norm Score}} \\ \hline
     \multirow{6}{*}{Diffusion-QL}
     & 50 &  206.44$\pm$16.70        &  30.65$\pm$2.10  &  - \\ \cline{2-5}
     & 20 &  108.65$\pm$2.85       &  13.04$\pm$0.90  &  $109.2\pm1.1$ ($111.1\pm1.9$) \\ \cline{2-5}
     & 10 &  $76.54\pm{10.74} $        &  $6.87\pm0.55$  & $108.6\pm0.6$ ($112.5\pm0.2$)  \\ \cline{2-5}
     & \cellcolor{gray!20}5 &  \cellcolor{gray!20}$57.06\pm{19.16} $        &  \cellcolor{gray!20}$3.76\pm0.29$  &  \cellcolor{gray!20}$108.2\pm5.6$ ($112.3\pm0.2$) \\ \cline{2-5}
     & 2 &  $31.59\pm{10.32} $        &  $1.96\pm0.10$  & $53.6\pm16.6$ ($103.5\pm10.0$)  \\ \cline{2-5}
     & 1 &  $30.23\pm{8.75} $        &  $1.37\pm0.09$  &  $2.8\pm1.5$ ($13.1\pm12.5$)  \\ \hline
     \multirow{6}{*}{Consistency-AC}
     &  50 &  $150.84\pm{31.02} $        &  $26.50\pm1.92$  &  - \\ \cline{2-5}
    &  20 &  $76.22\pm{9.92} $        &  $11.12\pm0.77$  & $101.3\pm6.3$ ($107.3\pm0.2$)  \\ \cline{2-5}
      &  10 &  $54.04\pm{4.43} $        &  $5.95\pm0.44$  & $98.4\pm4.3$ ($107.1\pm4.0$)  \\ \cline{2-5}
     &  5 &  $40.79\pm{2.79} $        &  $3.39\pm0.29$  & $101.4\pm4.7$ ($110.1\pm1.6$)  \\ \cline{2-5}
     & \cellcolor{gray!20}2 &  \cellcolor{gray!20}$31.94\pm{1.55} $        &  \cellcolor{gray!20}$1.84\pm0.21$  &  \cellcolor{gray!20}$102.4\pm3.0$ ($106.2\pm1.6$) \\ \cline{2-5}
     & 1 &  $28.51\pm{1.78} $        &  $1.23\pm0.11$  &  $6.2\pm5.4$ ($19.1\pm9.3$) \\ \hline
\end{tabular}
}
\caption{Comparison of computational time for two methods with different denoising steps $N$ on the task \textit{hopper-medium-expert-v2}. The gray lines apply default $N$ values for two models.}
\label{tab:time}
\end{table}

\begin{figure}[htbp]
	\centering
	\includegraphics[width=0.8\textwidth]{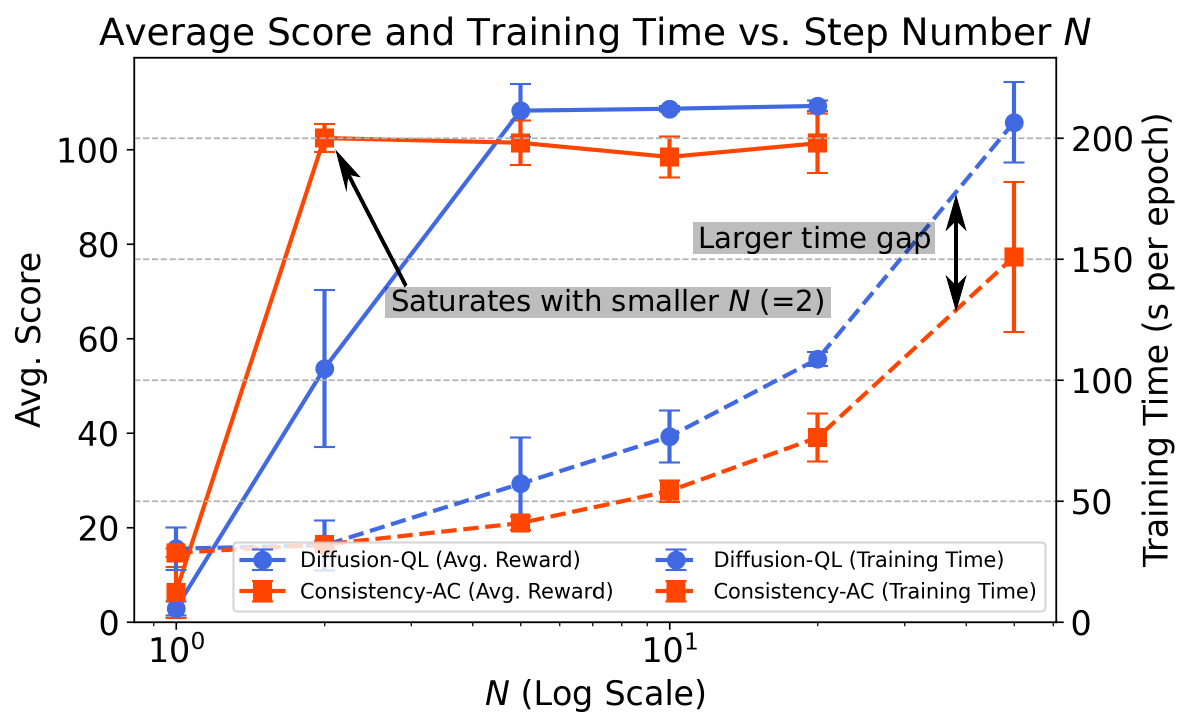}
	\caption{The average normalized scores and training time versus $N$ for two models on \textit{hopper-medium-expert}.}
	\label{fig:test_t}
\end{figure}

To evaluate the computational efficiency of Consistency-AC and Diffusion-QL with different denoising steps $N$, we conduct experiments for evaluating the training and inference time for $N\in\{1,2,5,10,20,50\}$ on the \textit{hopper-medium-expert-v2} environment. As generative models based on probability flow, both the consistency model and the diffusion model require the computational time directly dependent on the number of denoising steps $N$, and consistency model~\citep{song2023consistency} by design requires a smaller number of steps for achieving similar generative performances as the diffusion model. The results are summarized in Tab.~\ref{tab:time} for both the training time (seconds per epoch) and the inference time (milliseconds per sample) with the model training using different denoising steps $N$, as well as the average normalized scores for models trained after 2000 epochs with each $N$. Each cell contains the mean and standard deviation over five random seeds. Consistency-AC saturates its performance with only $N=2$ while Diffusion-QL saturates at $N=5$, which consumes about $1.786\times$ more training time while yielding a slightly better performance ($1.057\times$). 
The ``-'' in the table with $N=50$ indicates a missing value of the average score due to exceeding the limited time (72 hours) for the job.  
Moreover, as shown in Fig.~\ref{fig:test_t}, Consistency-AC has better scaling laws than Diffusion-QL for both training and inference in time consumption with increasing $N$, which is further testified by the linear fitting results in Sec.~\ref{app:offline_bc}, Fig.~\ref{fig:offline_fit_t}.

\begin{figure}[htbp]   
	\centering\includegraphics[width=0.9\textwidth]{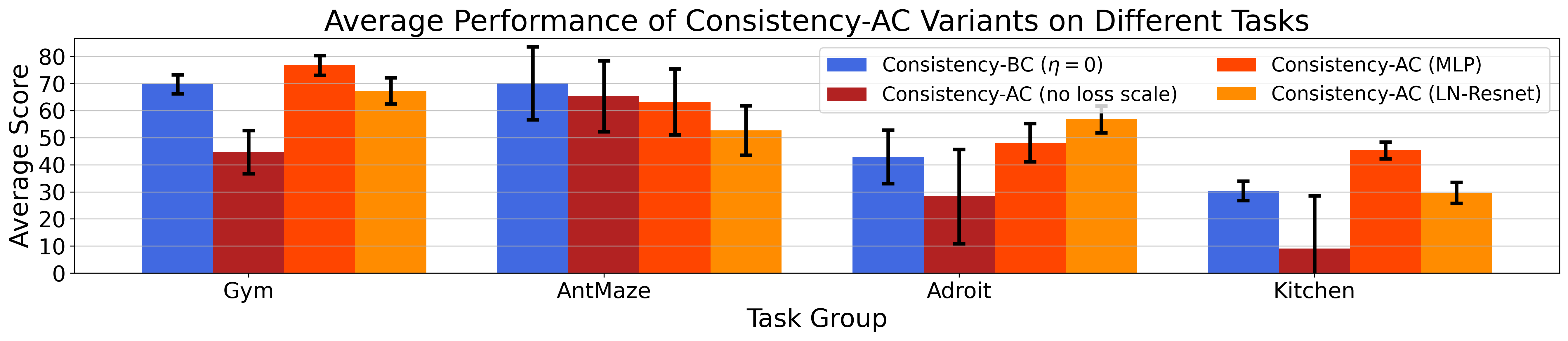}
    	\caption{Comparison of variants of Consistency-AC across tasks in offline RL setting.}
	\label{fig:offline_cac_ablation}
\end{figure}

\paragraph{Ablation Studies.}
\cite{hansen2023idql} proposes to use residual networks with layer normalization for network parameterization in diffusion policy, namely LN-Resnet, which is also tested for consistency policy in our experiments. As an ablation study, we compare different variants of Consistency-AC for offline RL setting, including (1) Consistency-BC by setting $\eta=0$ and without using loss scaling ($\lambda(\tau_n)\equiv 1$ in Eq.~\ref{eq:consis_loss}); (2) only without loss scaling; (3) the standard setting with multi-layer perceptrons (MLP) networks for the parameterization of $f_\theta$; 
(4) the LN-Resnet parameterization of $f_\theta$. These variants can be regarded as various hyperparameters or training settings for the proposed Consistency-AC algorithm, and the reported results in Tab.~\ref{tab:offline_cac} are the best choices among these variants. The comparison results for four variants across four task domains are summarized in Fig.~\ref{fig:offline_cac_ablation}. Detailed results for this ablation study are shown in Sec.~\ref{app:var_cac}. We find that LN-Resnet does not consistently improve over MLP across tasks for the consistency model but benefits mainly for the Adroit tasks. Without loss scaling, the performance degrades significantly (by $37.8\%$ on average) for most tasks, although for some specific tasks (\emph{e.g.}, AntMaze) it may improve the performance a bit without loss scaling. For most tasks except for AntMaze, Consistency-BC cannot achieve the best performances and the Q-learning loss $\mathcal{L}_q(\theta)$ (as Eq.~\ref{eq:max_q_loss}) with proper scaling helps to further improve the scores. 

\subsection{Offline-to-Online and Online RL}
\label{sec:online_cac}
\textbf{Empirical finding 3:} \textit{Consistency policy has a close but slightly worse performance than diffusion policy for offline-to-online RL, but a significant improvement of computational efficiency.} 

For online RL, we consider both online learning from scratch and the offline-to-online setting with the model trained on offline dataset as an initialization for online fine-tuning. As discussed in previous Sec.~\ref{sec:offline_cac}, the offline model can be selected in either an online or offline manner, respectively by model evaluation with or without online experience. Both types of models are used for initializing the policy and value models at the beginning of online fine-tuning. For online fine-tuning, it follows the standard actor-critic algorithm, that the $Q$ value is updated with Eq.~\ref{eq:double_q_loss} using online data, and the policy is updated with the Q-learning loss $\mathcal{L}_q(\theta)$ only as Eq.~\ref{eq:max_q_loss}. The algorithms use $\epsilon$-greedy for exploration with a decaying schedule. Pseudo-codes for offline-to-online and online Consistency-AC are provided in Sec.~\ref{app:online_algs}.

\begin{table}[htbp]
\caption{Comparison of normalized scores (last epoch) for methods in offline-to-online and online RL.}
\label{tab:off2on_rl}
\resizebox{\columnwidth}{!}{ 
\footnotesize
\begin{tabular}{c|ccccc|cc}
\toprule
         &  \multicolumn{5}{c|}{\textbf{Offline-to-Online}} & \multicolumn{2}{c}{\textbf{Online}}   \\ \hline
       \textbf{Gym Tasks} & {SAC} & {AWAC} & {ACA} & {Diffusion-QL} &  {Consistency-AC}  & {Diffusion-QL} & {Consistency-AC}  \\ \hline
     halfcheetah-m & 75.2 & 50.5 &  66.6 &  $\mathbf{99.6}\pm 2.3$ ($99.8\pm1.6$)     &  $98.7\pm1.8$ ($97.3\pm2.9$) &   $47.3\pm 2.9$   &   $55.1\pm 7.0$   \\ \hline
     hopper-m & 73.4 & \textbf{97.5} & 96.5 &  $77.2\pm25.6$ ($60.0\pm11.8$)   &  $60.5\pm8.6$ ($61.8\pm26.6$)&   $82.8\pm 30.6$&   $86.3\pm 28.4$\\ \hline
     walker2d-m  & 79.6 & 1.9 & 74.7 &  $\mathbf{118.3}\pm5.8$  ($117.5\pm5.9$)       &  $108.9\pm3.0$ ($107.9\pm10.5$) &   $77.0\pm 25.7$&   $69.4\pm 38.9$   \\ \hline
     halfcheetah-mr & 68.9 & 46.8 & 59.0 &  $\mathbf{96.3}\pm3.9$ ($97.6\pm1.2$)     &  $80.7\pm10.5$ ($82.3\pm9.4$) &   $43.5\pm 5.7$&   $56.5\pm 8.0$ \\ \hline
     hopper-mr  & 74.0 & \textbf{96.0} & 85.5 & $68.4\pm20.3$ ($90.6\pm24.0$)    &  $74.6\pm25.1$ ($63.4\pm16.7$) &   $94.0\pm 12.2$&   $75.8\pm 26.8$  \\ \hline
     walker2d-mr & 85.4 & 80.8 & 85.2 &  $95.7\pm18.8$ ($105.5\pm13.7$)       &  $\mathbf{102.0}\pm11.6$ ($96.5\pm17.9$)  &   $87.8\pm 29.0$&   $69.0\pm 42.3$\\ \hline
     halfcheetah-me & 82.2 & 68.8  & 93.7 &  $\mathbf{103.9}\pm2.2$ ($102.9\pm1.8$)    &  $99.6\pm4.1$ ($95.1\pm9.7$) &   $39.7\pm 3.6$&   $56.7\pm 5.8$ \\ \hline
     hopper-me  & 65.4 & 73.1  & \textbf{98.0} &  $71.7\pm31.1$ ($67.9\pm18.6$)        &  $65.4\pm5.7$ ($54.7\pm28.4$) &   $62.5\pm 22.2$&   $78.6\pm 14.6$ \\ \hline
     walker2d-me & 87.2 & 45.2 & 110.5  &  $\mathbf{117.0}\pm6.3$ ($111.2\pm10.6$)       &  $101.8\pm13.3$ ($89.2\pm16.2$)  &   $74.6\pm 39.0$&   $86.2\pm 27.8$\\ \hline
     \textbf{Average}  & 76.8 & 62.3 & $85.5$        & $\mathbf{94.2}$   &   $88.0$&   $67.7$ & 70.4 \\ \midrule
\end{tabular}
}
\end{table}

Tab.~\ref{tab:off2on_rl} summarizes the quantitative results for average scores achieved with Consistency-AC and Diffusion-QL across five random seeds for two settings over 9 Gym tasks, as well as offline-to-online baseline methods SAC, AWAC and ACA~\citep{yu2023actor}. Both the Consistency-AC and Diffusion-QL are pre-trained on the offline dataset for 2000 epochs. Each model is trained for one million steps for online fine-tuning or learning from scratch. The results for SAC, AWAC and ACA are adopted from the paper~\citep{yu2023actor} with each model fine-tuned for 100k online steps. Each cell has two values: one for offline model selection as initialization and another (in brackets) for online model selection as initialization. The normalized scores are slightly lower for Consistency-AC compared with Diffusion-QL in offline-to-online settings, but higher in online RL from scratch. On average, the two methods achieve lower values in online setting than the offline-to-online setting, which testifies the improvement of learning efficiency by initializing with pre-trained generative policy models. However, since the training is set to have a fixed overall timesteps and using the same learning rate $1\times10^{-5}$, the purely online models do not converge to its optimal performances yet. The learning rate is chosen for the fine-tuning setting, and the purpose is not to show online RL can achieve scores higher than 100 with sufficient training but to compare with the offline-to-online setting, for demonstrating the effectiveness of initialized models with offline pre-training. 

\textbf{Empirical finding 4:} \textit{Consistency policy could outperform diffusion policy for online RL setting mostly in computational efficiency and sometimes in sample efficiency, especially for hard tasks.} 

\begin{figure}[htbp]   
	\centering\includegraphics[width=\textwidth]{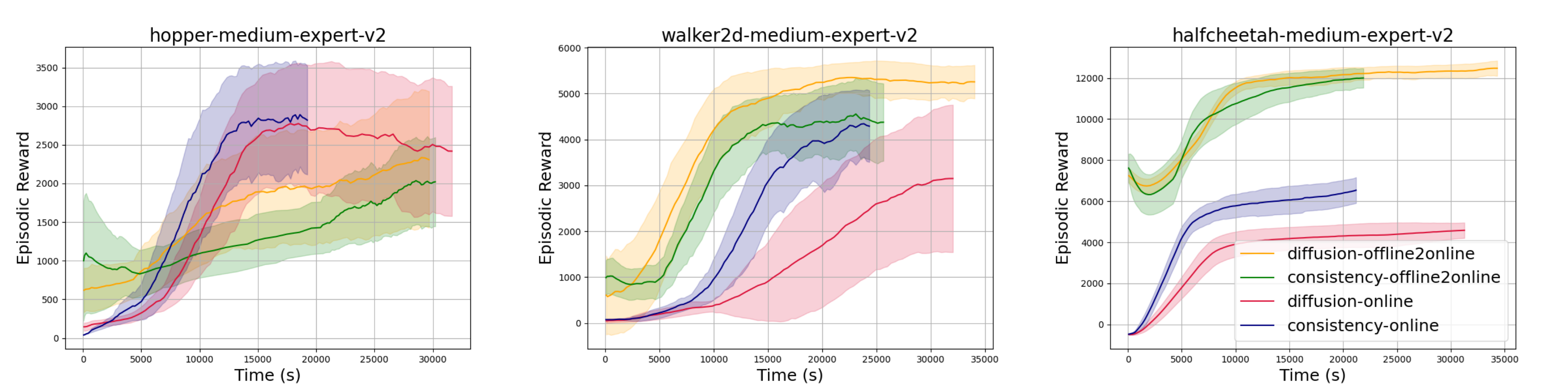}
    	\caption{Learning curves of Diffusion-QL and Consistency-AC for online RL and offline-to-online RL with offline model selection in time axis (all trained with one million environment steps). Each curve is smoothed and averaged over five random seeds, and shaded regions show the $95\%$ confidence interval.}
	\label{fig:compare_offline_load_offline_time_three}
\end{figure}

Fig.~\ref{fig:compare_offline_load_offline_time_three} shows the learning curves of Consistency-AC and Diffusion-QL for both offline-to-online and online RL settings with one million online training steps on three example tasks. Different methods consume different time to finish the entire training. The diagrams are plotted with x-axis being the wall-clock time, therefore the curves exhibit different lengths. For most tasks, the consistency policy has comparable performances with the diffusion policy and a significantly shorter time to finish the entire online training. The offline-to-online methods are usually more sample efficient than the online methods except for three \textit{hopper} tasks, which are relatively easy to learn a good policy. For the online setting, the consistency policies demonstrate significantly more efficient learning than the diffusion policies, especially for more complex tasks like \textit{halfcheetah}. Consistency policies show a sharper score-increasing slope for $8/9$ tasks than the diffusion policies. Our conjecture is that the expressiveness of a model is more essential in offline setting than online setting. For a deterministic optimal policy in MDP, overly expressive policy models like diffusion may hinder the convergence in online setting by being too explorative. For offline-to-online setting, this advantage is less obvious presumably due to the lower initial performances of consistency policies from the offline pre-training. We refer to Sec.~\ref{app:online_time}, Tab.~\ref{tab:online_train_time} and Fig.~\ref{fig:online_train_t} for more analysis of the training time for two methods.




\section{Consistency Model Training and Inference Details}
\label{app:consist_details}

\paragraph{Training.}
The consistency model $f_\theta$ for modeling data distribution $p_\text{data}(\mathbf{x})$ has the loss function~\citep{song2023consistency}:
\begin{equation}
    \mathcal{L}_c(\theta) = \mathbb{E}_{n\sim\mathcal{U}(1,N-1), \mathbf{x}\sim p_\text{data}(\mathbf{x}), \mathbf{z}\sim\mathcal{N}(\mathbf{0},\mathbf{I})}\Big[\lambda(\tau_n)d\big(f_\theta(\mathbf{x} + \tau_{n+1}\mathbf{z}, \tau_{n+1}), f_{\theta^\intercal}(\mathbf{x} + \tau_{n}\mathbf{z}, \tau_n)\big)\Big]
    \label{eq:consis_loss_ori}
\end{equation}
where $d(\cdot, \cdot)$ is the distance metric and we use $l_2$ distance $d(\mathbf{x},\mathbf{y})= \lVert \mathbf{x} -\mathbf{y}\rVert_2^2$.
For training, the sub-sequence $\{\tau_n |n\in[N]\}$ is different from inference, and it follows the Karras boundary~\citep{karras2022elucidating} schedule: $\tau_n=\big(\epsilon^{1/\rho}+\frac{n-1}{N-1}(\tau_N^{1/\rho}-\epsilon^{1/\rho})\big)^\rho$. The schedule function $N(k)=\left\lceil\sqrt{\frac{k}{K}((s_1+1)^2-s_0^2)+s_0^2}-1\right\rceil+1$ with $k$ as the current training iteration of a total $K$ iterations within one epoch\footnote{Our experiments use constants $\epsilon=0.002, T=80; \rho=7; s_0=2, s_1=150$ following~\citet{song2023consistency}}. 

\paragraph{Inference.}
After training, the consistency model $f_\theta$ can be used for generating samples given initial noisy input $\hat{\mathbf{x}}_T\sim\mathcal{N}(\mathbf{0}, T^2\mathbf{I})$, following either single-step sampling $\mathbf{x}=f_\theta(\hat{\mathbf{x}}_T, T)$, or multistep sampling by iteratively calculating $\mathbf{x}=f_\theta(\hat{\mathbf{x}}_{\tau_n}, \tau_n)$ with $\hat{\mathbf{x}}_{\tau_n}=\mathbf{x}+\sqrt{\tau_n^2-\epsilon^2}\mathbf{z}$ following a given time sequence $\{\tau_n|n\in[N]\}$.
For inference, the time sequence is a linspace of $[\epsilon, {T}]$ with $(N-1)$ sub-intervals as: $\tau_n=\frac{n-1}{N-1}(T-\epsilon)+\epsilon, n\in[N]$.

\section{Offline RL Experiment Details}

\subsection{Computational Time}
\label{app:offline_bc}
\paragraph{Overall Training Time.} Tab.~\ref{tab:offline_train_time} shows the comparison of Diffusion-BC and Consistency-BC in terms of the computational time during training for D4RL Gym, AntMaze, Adroit and Kitchen tasks. Each result is averaged over five random seeds with standard deviations reported. Since different environments are trained for various numbers of total epochs, the comparison is based on per-epoch time consumption. The two methods use the same batch size and number of iterations within each epoch, as well as the same network architecture.

\begin{table}[H]
\caption{The training time (seconds per epoch) for two BC methods on D4RL Gym, AntMaze, Adroit and Kitchen tasks.}
\label{tab:offline_train_time}
\centering
\footnotesize
\begin{tblr}{
colspec = {l||*{1}{c}|c},
row{1} = {font=\bfseries}
}
\toprule
Tasks & Diffusion-BC & Consistency-BC \\
\hline
halfcheetah-m & $67.93\pm2.00$ & $43.07\pm 0.61$ \\
hopper-m & $61.00\pm1.58$ & $38.00\pm0.49$ \\
walker2d-m & $68.17\pm1.51$ & $43.58\pm0.90$ \\
halfcheetah-mr & $67.07\pm2.07$ & $42.75\pm0.83$ \\
hopper-mr & $64.89\pm3.29$ & $38.03\pm0.47$ \\
walker2d-mr & $66.11\pm1.69$ & $42.70\pm0.54$ \\
halfcheetah-me & $67.73\pm1.78$ & $43.60\pm0.86$ \\
hopper-me & $63.04\pm4.25$ & $38.56\pm0.56$ \\
walker2d-me & $69.10\pm2.83$ & $43.88\pm0.72$ \\
\hline
\textbf{Average} & $66.12\pm2.33$ & $41.57\pm0.66$ \\
\bottomrule
antmaze-u & $97.13\pm4.21$ & $47.88\pm2.59$ \\
antmaze-ud & $104.83\pm4.50$ & $47.20\pm2.23$ \\
antmaze-mp & $109.66\pm3.82$ & $57.92\pm4.46$ \\ 
antmaze-md & $112.25\pm2.41$ & $51.80\pm2.59$  \\
antmaze-lp & $113.15\pm1.01$ & $56.52\pm3.17$  \\
antmaze-ld & $118.62\pm3.42$ & $53.89\pm2.74$ \\  \hline
\textbf{Average} & $109.27\pm3.44$ & $52.54\pm3.05$ \\
\bottomrule
pen-human-v1 & $92.20\pm3.17$ & $46.94\pm2.92$ \\
pen-cloned-v1 & $94.64\pm6.16$ & $50.33\pm2.23$ \\ \hline
\textbf{Average} & $93.42\pm4.67$ & $48.64\pm2.58$ \\
\bottomrule
kitchen-c & $96.77\pm2.74$ & $66.71\pm4.64$ \\
kitchen-p & $94.25\pm2.27$ & $61.85\pm2.74$ \\
kitchen-m & $93.02\pm3.49$ & $66.60\pm2.59$ \\ \hline
\textbf{Average} & $94.68\pm2.83$ & $65.05\pm3.32$ \\
\bottomrule
\toprule
\textbf{Total Average} & $86.08\pm3.16$ & $49.09\pm2.34$ \\
\bottomrule
\end{tblr}
\end{table}

\begin{figure}[htbp]   
	\centering\includegraphics[width=0.49\textwidth]{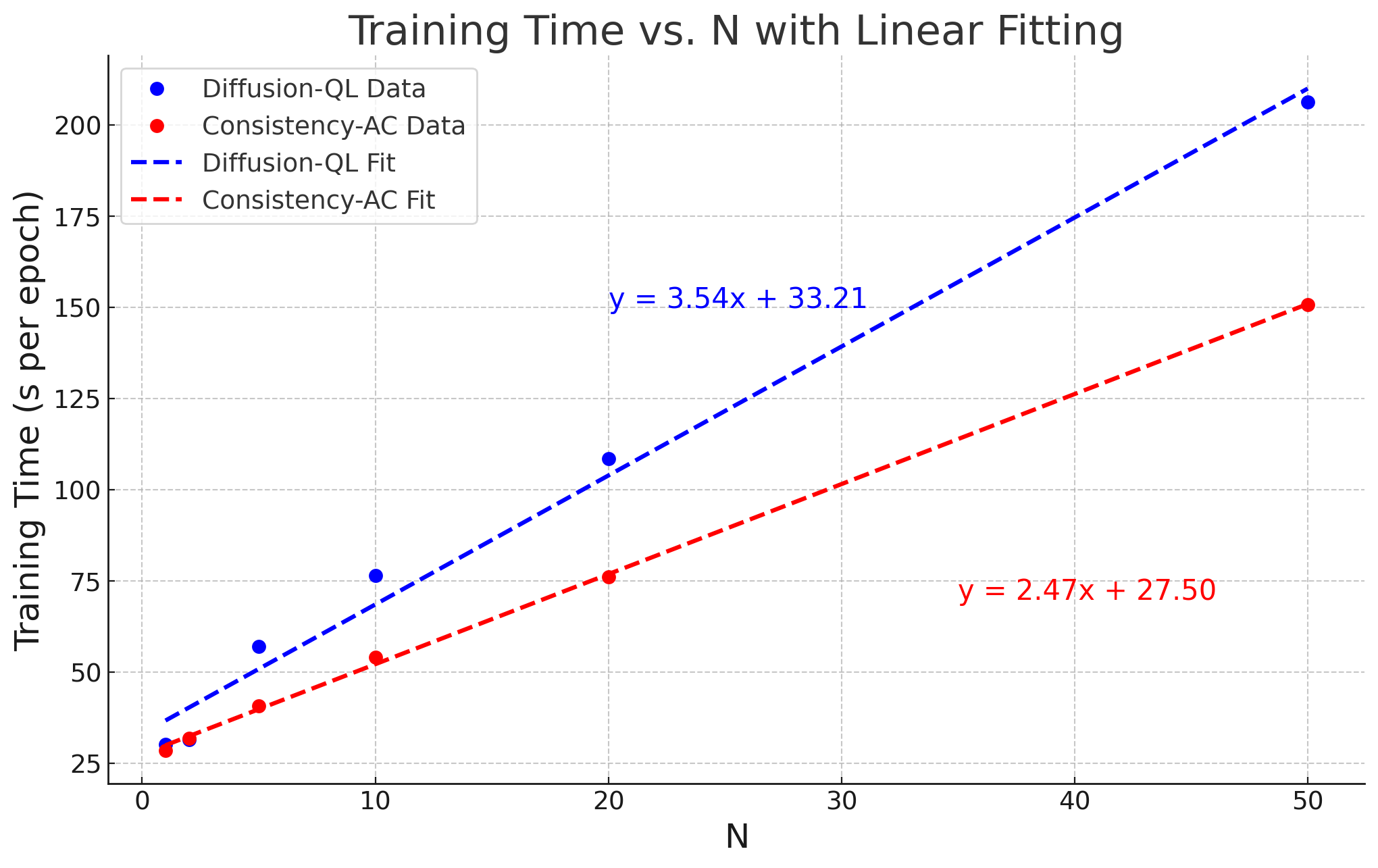}
        \includegraphics[width=0.49\textwidth]{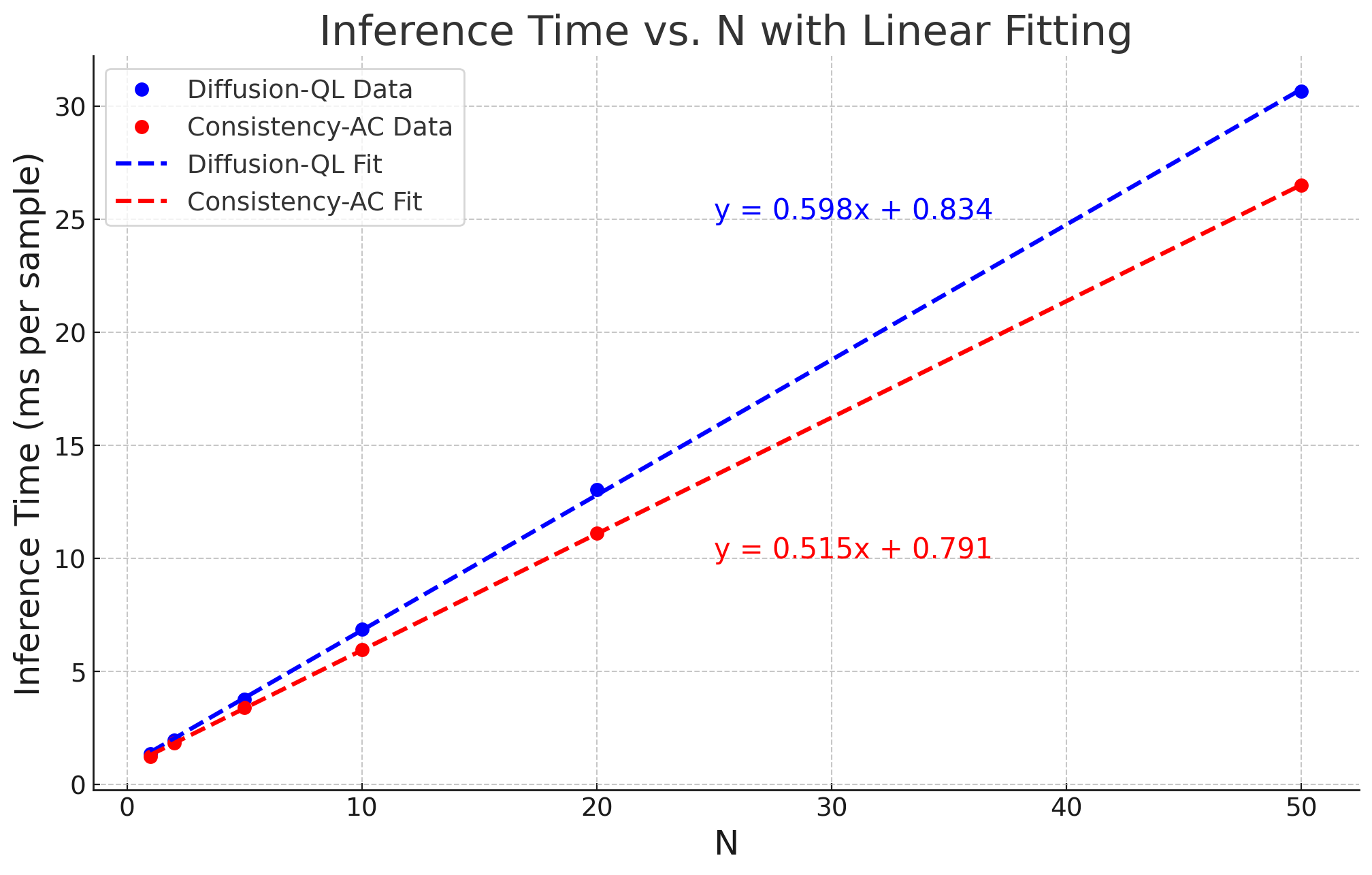}
    	\caption{The training time (left) and inference time (right) versus denoising steps $N$ for Diffusion-QL and Consistency-AC in offline RL, evaluated on \textit{hopper-medium-expert-v2} environment.}
	\label{fig:offline_fit_t}
\end{figure}

\paragraph{Scaling Law.}
Fig.~\ref{fig:offline_fit_t} further shows the scaling laws of training time and inference time with increasing $N$ for Diffusion-QL and Consistency-AC in offline RL setting, based on results in Tab.~\ref{tab:time} for environment \textit{hopper-medium-expert-v2}. Notice that the coefficients of Consistency-AC are smaller than Diffusion-QL in both training (2.47 vs. 3.54) and inference (0.515 vs. 0.598), which indicates smaller time consumption with increasing $N$. 

\subsection{Ablation Studies}
\label{app:var_cac}
Four variants of Consistency-AC are compared for offline RL setting, including (1) Consistency-BC by setting $\eta=0$ and without using loss scaling ($\lambda(\tau_n)\equiv 1$ in Eq.~\ref{eq:consis_loss}); (2) only without loss scaling; (3) the standard setting with MLP networks for the parameterization of $f_\theta$; (4) the LN-Resnet parameterization of $f_\theta$. These variants can be regarded as various hyperparameters or training settings for the proposed Consistency-AC algorithm. The average scores for five random seeds over D4RL Gym, AntMaze, Adroit and Kitchen are shown in Tab.~\ref{tab:offline_cac_variants}.

\begin{table}[htbp]
 \caption{The performance of Consistency-AC variants on D4RL Gym, AntMaze, Adroit and Kitchen tasks for offline RL setting. Each cell has two values: one for offline model selection and another (in brackets) for online model selection. Each result is averaged over five random seeds with standard deviations reported.}
    \label{tab:offline_cac_variants}
    \centering
    \resizebox{\textwidth}{!}{
    \begin{tblr}{
    colspec = {l||cccc},
    row{1, 12, 17, 21} = {font=\bfseries}
    }
    \toprule
        Gym Tasks & Consistency-BC ($\eta=0$) & Consistency-AC (no loss scale) & Consistency-AC (MLP) & Consistency-AC (LN-Resnet) \\
        \hline
        halfcheetah-m &  $31.0\pm0.4$ ($46.2\pm0.4$) & $69.1\pm0.7$ ($71.9\pm0.8$) & $50.1\pm0.4$ ($50.4\pm0.2$) & $50.6\pm0.3$ ($50.9\pm0.2$) \\
        hopper-m & $71.7\pm8.0$ ($78.3\pm2.6$) & $80.7\pm10.5$ ($99.7\pm2.3$) & $78.0\pm3.9$ ($86.4\pm4.0)$ & $74.1\pm6.7$ ($83.7\pm8.5$)\\ 
        walker2d-m &  $83.1\pm0.3$ ($84.1\pm0.3$) & $5.5\pm1.7$ ($21.2\pm1.6$) & $63.0\pm5.2$ ($75.0\pm1.8)$& $66.2\pm5.2$ ($75.4\pm1.9$)\\
        halfcheetah-mr & $34.4\pm5.3$ ($45.4\pm0.7$) & $58.7\pm3.9$ ($62.7\pm0.6$) & $47.3\pm0.2$ ($47.8\pm0.3)$& $47.8\pm0.3$ ($48.4\pm0.1$)\\
        hopper-mr & ${99.7}\pm0.5$ ($100.4\pm0.6$) & $80.2\pm9.0$ ($103.4\pm1.2$) & $94.5\pm6.4$ ($100.9\pm0.2)$& $98.7\pm2.9$ ($100.6\pm0.3$)\\
        walker2d-mr & $73.3\pm5.7$ ($80.8\pm2.4$) & $72.3\pm15.4$ ($105.1\pm1.6$) & $76.8\pm5.5$ ($86.1\pm1.2)$& $79.5\pm3.6$ ($83.0\pm1.5$)\\
        halfcheetah-me & $32.7\pm1.2$ ($39.6\pm3.4$) & $22.6\pm10.4$ ($55.2\pm11.6$)& $84.3\pm4.1$ ($89.2\pm3.3)$& $61.7\pm13.6$ ($68.4\pm6.7$)\\ 
        hopper-me & $90.6\pm9.3$ ($96.8\pm4.6$) & $10.1\pm16.2$ ($10.8\pm15.6$)& $100.4\pm3.5$ ($106.0\pm1.3)$& $43.1\pm5.7$ ($54.5\pm11.2$)\\ 
        walker2d-me & $110.4\pm0.7$ ($111.6\pm0.7$) & $2.7\pm4.3$ ($14.9\pm6.9$)& $91.1\pm3.4$ ($97.7\pm3.2)$& $84.1\pm5.1$ ($97.5\pm1.6$)\\ 
        \hline
        \textbf{Average}  & 69.7 (75.9) & 44.7 (60.5)  & 76.7 (82.2) & 67.3 (73.6) \\
    \bottomrule
    \toprule
        AntMaze Tasks & Consistency-BC ($\eta=0$) & Consistency-AC (no loss scale) & Consistency-AC (MLP) & Consistency-AC (LN-Resnet)\\ \hline
        antmaze-u  &  $75.8\pm4.0$ ($87.0\pm4.5$) & $75.4\pm5.8$ ($82.6\pm3.8$) & $68.8\pm2.3$ ($82.2\pm4.7$) & $75.8\pm1.6$ ($85.6\pm3.9$)\\
        antmaze-ud  & ${77.6}\pm6.3$ ($82.4\pm3.4$) & $75.2\pm6.6$ ($80.2\pm2.8$) & $68.6\pm4.4$ ($78.4\pm1.1$) & $72.4\pm3.5$ ($81.2\pm1.9$)\\
        antmaze-mp & $56.8\pm30.1$ ($71.6\pm14.5$) & $45.2\pm26.9$ ($73.2\pm8.4$) & $52.2\pm29.8$ ($70.4\pm7.1$)& $10.0\pm22.4$ ($59.4\pm12.8$)\\ \hline
        \textbf{Average} & 70.1 (80.3)  & 65.3 (78.7) & 63.2 (77.0) & 52.7 (75.4)\\
    \bottomrule
    \toprule
        Adroit Tasks & Consistency-BC ($\eta=0$) & Consistency-AC (no loss scale) & Consistency-AC (MLP) & Consistency-AC (LN-Resnet)\\ \hline
        pen-human-v1 & $52.4\pm13.7$ ($63.7\pm7.4$) & $8.4\pm24.0$ ($22.1\pm20.5$)  & $60.6\pm10.2$ ($66.6\pm7.5$)& $63.4\pm7.7$ ($67.9\pm5.3$)\\
        pen-cloned-v1 & $33.4\pm6.0$ ($51.9\pm6.6$) & $48.2\pm10.8$ ($58.2\pm12.6$) & $35.8\pm3.9$ ($40.5\pm2.6$) & $50.1\pm2.2$ ($53.7\pm3.4$)\\ \hline
        \textbf{Average} & 42.9 (57.8) & 28.3 (40.2) & 48.2 (53.6) & 56.8 (60.8) \\ 
    \bottomrule
    \toprule
        Kitchen Tasks & Consistency-BC ($\eta=0$) & Consistency-AC (no loss scale)  & Consistency-AC (MLP) & Consistency-AC (LN-Resnet)\\ \hline
        kitchen-c  &   $45.2\pm5.0$ ($50.9\pm3.6$) & $10.0\pm20.1$ ($25.5\pm24.6$) & $51.9\pm6.0$ ($67.6\pm2.7$) & $36.9\pm3.2$ ($38.0\pm2.5$)\\
        kitchen-p  &  $22.6\pm3.8$ ($23.8\pm2.8$) & $7.7\pm16.9$ ($17.0\pm14.5$) & $38.2\pm1.8$ ($39.8\pm1.6$) & $25.8\pm5.5$ ($28.6\pm2.7$)\\
        kitchen-m  &  $23.5\pm1.8$ ($24.3\pm1.3$) & $9.7\pm21.3$ ($15.8\pm20.2$) & $45.8\pm1.5$ ($46.7\pm0.9$)& $26.0\pm3.0$ ($28.8\pm2.1$)\\ \hline
        \textbf{Average} & 30.4 (33.0) & 9.1 (19.4) & 45.3 (51.4) & 29.6 (31.8) \\ 
    \bottomrule
    \toprule
     \textbf{Total Average}  & $59.7$ ($67.0$)  &  $40.1$ ($54.1$) & $64.5$ ($72.5$) & $56.8$ ($65.0$)   \\
     \bottomrule
    \end{tblr}}
\end{table}

\newpage
\section{Offline-to-Online and Online RL Details}
\label{app:off2on_rl}

\subsection{Algorithms}
\label{app:online_algs}
\begin{algorithm}[H]
\caption{Offline-to-Online Consistency Actor-Critic}
\begin{algorithmic}[1]
\STATE \textbf{Input}: offline pretrained policy $\pi_\theta$ and critic networks $Q_{\phi_1}, Q_{\phi_2}$
\STATE Initialize online dataset $\tilde{\mathcal{D}}=\emptyset$, target network parameters: $\theta^\intercal \leftarrow \theta$, $\phi_1^\intercal \leftarrow \phi_1, \phi_2^\intercal \leftarrow \phi_2$
\FOR{episode $j = 1,\ldots,M$}
    \STATE Reset the environment and observe $s_1$.
    \FOR{$t=1,\ldots,H$}
        \STATE {\color{blue}\% Collect Samples}
        \STATE Infer action $a_t$ based on $s_t$ with consistency policy $\pi_\theta$ by Alg.~\ref{alg:forward_a}.
        \STATE Execute actions $a_t$, observe reward $r_t$, next state $s_{t+1}$.
        \STATE Store data sample $(s_t,a_t,r_t,s_{t+1})$ into $\tilde{\mathcal{D}}$.
        \STATE Sample minibatch $\mathcal{B}=\{(s,a,r,s')\}\subseteq\tilde{\mathcal{D}}$;
        \STATE {\color{blue}\% Q-value Update}
        \STATE Update $Q_{\phi_1}, Q_{\phi_2}$ with Eq.~\ref{eq:double_q_loss};
        \STATE {\color{blue}\% Policy Update}
        \STATE Update policy $\pi_\theta$ (with model $f_\theta)$ via loss $\mathcal{L}_q(\theta)$ as Eq.~\ref{eq:max_q_loss};
        \STATE {\color{blue}\% Target Update}
        \STATE Update target: $\theta^\intercal \leftarrow \alpha \theta^\intercal +(1-\alpha)\theta, \phi_i^\intercal \leftarrow \alpha \phi_i^\intercal + (1-\alpha)\phi_i, i\in\{1,2\}$;
    \ENDFOR
\ENDFOR
\end{algorithmic}
\label{alg:off2on_cac}
\end{algorithm}
\begin{algorithm}[H]
\caption{Online Consistency Actor-Critic}
\begin{algorithmic}[1]
\STATE Initialize policy $\pi_\theta$ and critic networks $Q_{\phi_1}, Q_{\phi_2}$
\STATE Initialize online dataset $\tilde{\mathcal{D}}=\emptyset$, target network parameters: $\theta^\intercal \leftarrow \theta$, $\phi_1^\intercal \leftarrow \phi_1, \phi_2^\intercal \leftarrow \phi_2$
\FOR{episode $j = 1,\ldots,M$}
    \STATE Reset the environment and observe $s_1$.
    \FOR{$t=1,\ldots,H$}
        \STATE {\color{blue}\% Collect Samples}
        \STATE Infer action $a_t$ based on $s_t$ with consistency policy $\pi_\theta$ by Alg.~\ref{alg:forward_a}.
        \STATE Execute actions $a_t$, observe reward $r_t$, next state $s_{t+1}$.
        \STATE Store data sample $(s_t,a_t,r_t,s_{t+1})$ into $\tilde{\mathcal{D}}$.
        \STATE Sample minibatch $\mathcal{B}=\{(s,a,r,s')\}\subseteq\tilde{\mathcal{D}}$;
        \STATE {\color{blue}\% Q-value Update}
        \STATE Update $Q_{\phi_1}, Q_{\phi_2}$ with Eq.~\ref{eq:double_q_loss};
        \STATE {\color{blue}\% Policy Update}
        \STATE Update policy $\pi_\theta$ (with model $f_\theta)$ via loss $\mathcal{L}_q(\theta)$ as Eq.~\ref{eq:max_q_loss};
        \STATE {\color{blue}\% Target Update}
        \STATE Update target: $\theta^\intercal \leftarrow \alpha \theta^\intercal +(1-\alpha)\theta, \phi_i^\intercal \leftarrow \alpha \phi_i^\intercal + (1-\alpha)\phi_i, i\in\{1,2\}$;
    \ENDFOR
\ENDFOR
\end{algorithmic}
\label{alg:on_cac}
\end{algorithm}


\subsection{Computational Time}
\label{app:online_time}
\begin{figure}[htbp]   
	\centering\includegraphics[width=0.55\textwidth]{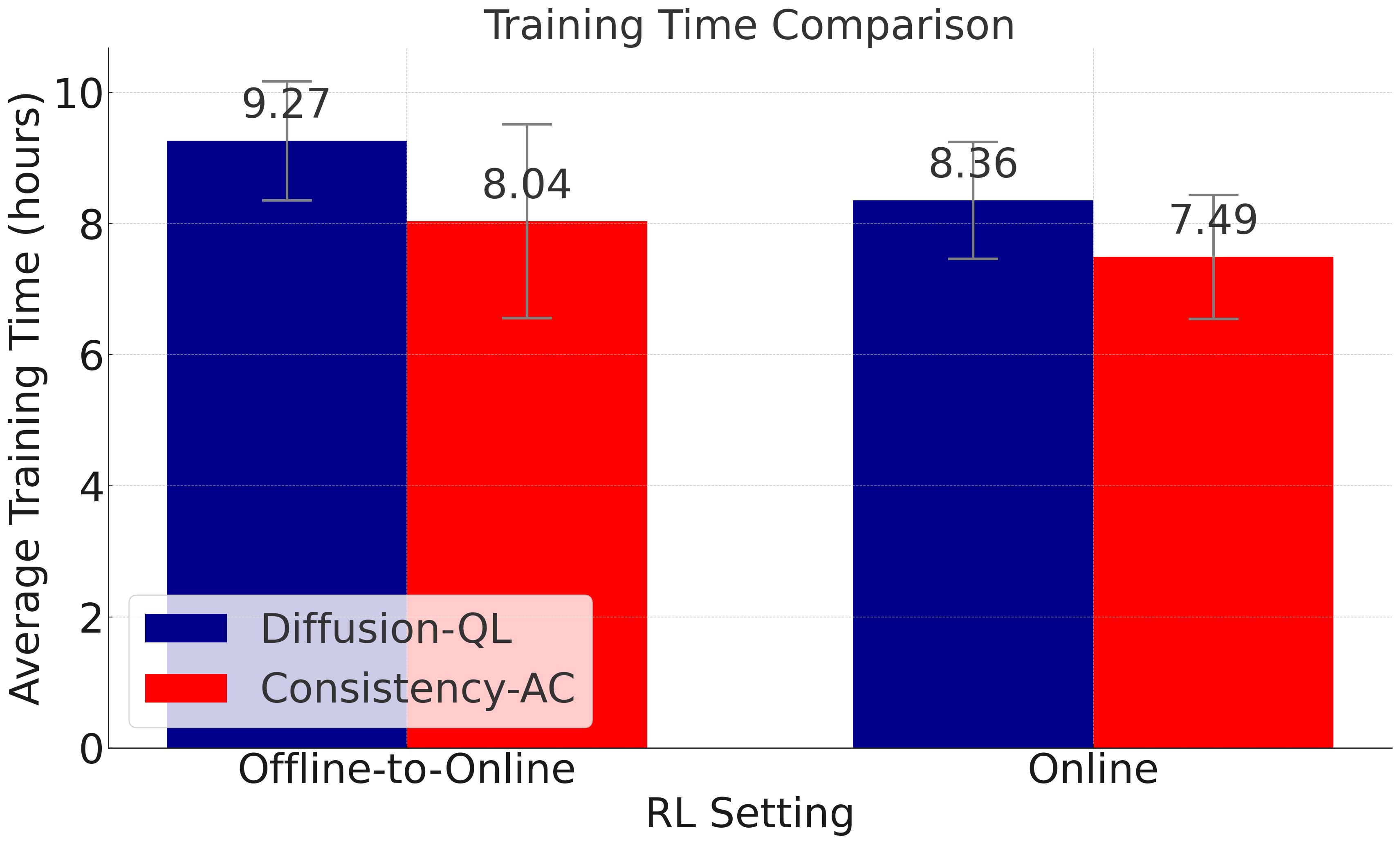}
    	\caption{The average training time (hours) for offline-to-online and online training with Diffusion-QL and Consistency-AC on 9 Gym tasks.}
	\label{fig:online_train_t}
\end{figure}

The overall training time for one million environment steps using Diffusion-QL and Consistency-AC in offline-to-online and online RL settings is shown in Tab.~\ref{tab:online_train_time}, with the average training time for each setting summarized in Fig.~\ref{fig:online_train_t}.
The reduction of computational time in online setting is less significant than the offline setting (as Fig.~\ref{fig:offline_bc_time}) because there is a large portion of time consumed by the environment simulation steps following the agent's action inference. The improvement of model inference and update will not affect the environment simulation time. 
\begin{table}[H]
\caption{The overall training time (hours) for offline-to-online and online settings on Gym tasks.}
\label{tab:online_train_time}
\centering
\begin{tabular}{c|cccc}
\toprule
& \multicolumn{2}{c}{\textbf{Offline-to-Online}}  & \multicolumn{2}{c}{\textbf{Online}} \\ \hline 
\textbf{Gym Tasks} & Diffusion-QL & Consistency-AC & Diffusion-QL & Consistency-AC \\
\hline
halfcheetah-m & $11.55\pm4.08$ & $9.60\pm 2.33$ & $9.09\pm0.88$ & $8.77\pm0.96$ \\
hopper-m & $8.97\pm1.48$ & $6.90\pm0.79$ & $8.06\pm0.88$ & $6.99\pm0.95$\\
walker2d-m & $9.17\pm0.29$ & $8.19\pm1.91$ & $8.23\pm1.04$ & $6.98\pm0.89$\\
halfcheetah-mr & $9.18\pm0.22$ & $7.72\pm0.89$ & $8.72\pm0.88$ & $7.76\pm1.02$\\
hopper-mr & $8.22\pm0.20$ & $7.26\pm1.69$ & $8.12\pm0.81$ & $6.92\pm0.78$\\
walker2d-mr & $8.85\pm0.22$ & $7.32\pm0.88$ & $8.07\pm1.01$ & $7.05\pm1.05$\\
halfcheetah-me & $9.24\pm0.21$ & $8.46\pm1.96$ & $8.54\pm0.74$ & $7.65\pm1.02$\\
hopper-me & $8.28\pm0.20$ & $7.47\pm0.57$ & $8.02\pm1.07$ & $7.01\pm0.99$\\
walker2d-me & $9.93\pm1.27$ & $9.41\pm2.26$ & $8.35\pm0.73$ & $8.29\pm0.86$\\
\hline
\textbf{Average} & $9.27\pm0.91$ & $8.04\pm1.48$ & $8.36\pm0.89$ & $7.49\pm0.95$\\
\bottomrule
\end{tabular}
\end{table}

\chapter{Reinforcement Learning in Few-Step Video Generation\label{ch:rl_video}}
\begin{center}
\begin{quote}
This section is based on paper ``\textit{DOLLAR: Few-Step Video Generation via Distillation and Latent Reward Optimization}''~\cite{ding2024dollar} written in collaboration with Chi Jin, Difan Liu, Haitian Zheng, Krishna Kumar Singh, Qiang Zhang, Yan Kang, Zhe Lin and Yuchen Liu, previously published at ICCV 2025.
\end{quote}
\end{center}

\section{Introduction}
\label{sec:intro}

Diffusion probabilistic models~\cite{sohl2015deep, song2019generative, ho2020denoising, song2021score} have recently revolutionized generative modeling in continuous domains.  With remarkable expressive power and flexibility across diverse data formats and modalities, diffusion models have significant breakthroughs in tasks such as text-to-image and text-to-video (T2V) generation. However, despite substantial improvements in generation quality, the efficiency of diffusion models remains a limiting factor in practical applications due to the inherently large number of iterative sampling steps.  This efficiency challenge is exacerbated in video generative modeling, where the higher-dimensional space demands larger model sizes, more extensive training data, larger input and output tensors, and more sampling iterations. Furthermore, practical applications often require generation qualities that may differ from the training distribution—such as higher aesthetic standards or diverse stylistic choices—necessitating efficient post-training adjustments or fine-tuning to meet specific requirements while managing the substantial cost of pre-training.

\begin{figure*}[htb]
    \centering
\includegraphics[width=\textwidth]{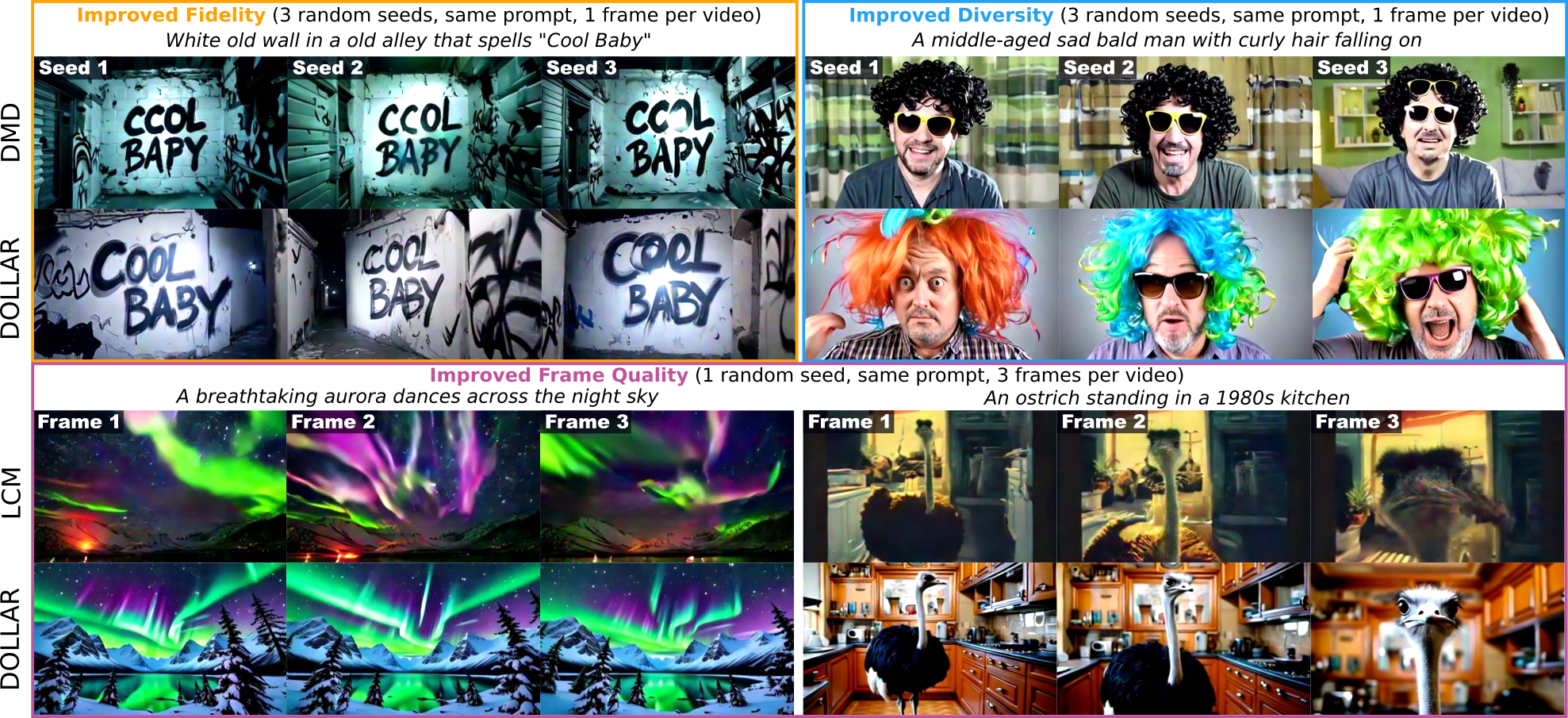}
    \caption{By incorporating variational score distillation, consistency distillation and latent reward fine-tuning, our DOLLAR method with 4-inference steps increases sample diversity and fidelity over DMD~\cite{yin2024one}, generates higher quality videos than LCM~\cite{luo2023latent,wang2023videolcm}, and achieves $\times15.6$ acceleration compared with teacher.
    }
    \label{fig:teaser}
\end{figure*}

To address the efficiency challenges in diffusion models, model distillation~\cite{polino2018model, luhman2021knowledge, salimans2022progressive} has been widely researched across various models and domains. Score distillation, specifically, aims to improve efficiency in 3D~\cite{wang2023score, poole2022dreamfusion, wang2024prolificdreamer} and image synthesis~\cite{yin2024one, luo2024diff, xie2024distillation, salimans2024multistep} by aligning the distribution between teacher and student diffusion models. However, despite achieving high sample fidelity, it often encounters model collapse issues~\cite{yin2024one,lu2024simplifying}. Another approach, consistency distillation (CD)~\cite{song2023consistency}, seeks to ensure consistent sample predictions along the diffusion trajectory. While CD promotes greater sample diversity, it has limitations: it tends to lower sample fidelity and can produce overly smooth outputs in large-scale T2V applications.

A further challenge with distillation methods is that the student's performance is typically upper-bounded by the teacher model. Previous efforts to address this limitation have involved integrating variational score distillation (VSD)~\cite{wang2024prolificdreamer, yin2024improved} or consistency distillation~\cite{kim2023consistency} with GAN~\cite{goodfellow2014generative} loss, which modestly enhances sample fidelity within the training distribution, limited by the sparsity of discriminative signals of adversarial training. Consequently, the generated samples may still fall short in capturing nuanced visual quality details and text-to-image alignment, both of which require denser feedback signals. Recently developed image or video reward models offer promising potential to address this gap, providing richer signals for fine-grained improvements in generation quality.
 
In this work, we address the limitations of consistency distillation (CD) by incorporating a larger number of teacher denoising steps and combining CD with variational score distillation (VSD) to produce high-quality, diverse samples with a few-step model after distillation. However, this alone does not suffice to outperform the teacher model or reliably meet specific preferences for downstream applications, as generated samples may still face challenges in visual quality and text-to-video alignment. While model post-training with a high-quality dataset is a potential solution, it is often costly to implement. To overcome these limitations, we further introduce an efficient reward model fine-tuning method that enhances the student model beyond the teacher's capabilities and aligns it with any pre-defined requirements through tailored reward metrics. The improved performance is shown in Fig.~\ref{fig:teaser}. 


We propose learning a dual reward model within the latent space, guided by the pixel-space reward model, and utilize the gradients from this latent reward model (LRM) to fine-tune the diffusion model directly. This approach combines the strengths of reward-gradient methods in pixel space and stochastic policy gradient methods, offering several advantages: (1) it harnesses the rich gradient information from the latent reward model, enabling efficient and effective tuning; (2) it does not require the original reward model to be differentiable, broadening applicability to a variety of reward models; (3) it significantly reduces computational and memory costs during fine-tuning by eliminating the need for backpropagation through large pixel-space reward models and the decoder. The LRM approach is versatile, accommodating various reward types—including image, video, text-image, and text-video rewards—thereby enhancing practical usability.

In summary, our contributions are threefold: (1) We introduce a diffusion model distillation method that combines VSD and CD losses to enable efficient, few-step T2V models; (2) We enhance CD with a generalized approach incorporating multiple teacher denoising steps to improve its effectiveness; (3) We propose to use a compact latent-space reward model for reward-based fine-tuning, which posts no requirement on the differentiability of original reward metrics and is more memory- and computation-efficient. All evaluations are conducted on large-scale T2V settings. Putting together these innovations, we present \texttt{DOLLAR} method with \textbf{D}isti\textbf{l}lation and \textbf{La}tent \textbf{R}eward \textbf{O}ptimization, to significantly advance the quality and efficiency of video generation and pave the way for real-time applications.


\section{Related Work}
\paragraph{Video Generation.}

Recent advancements have extended diffusion models from image synthesis to video generation, addressing the complexities of spatiotemporal data. Pioneering works like Video Diffusion Models \cite{ho2022video} adapted diffusion processes to handle temporal dynamics, enabling the creation of coherent and high-fidelity video clips. To enhance computational efficiency, Latent Diffusion Models (LDM) \cite{rombach2022high} perform diffusion modeling in compressed latent spaces, a strategy further refined for video by \cite{blattmann2023align}, \cite{harvey2022flexible}, Stable Video Diffusion \cite{blattmann2023stable} and VideoCrafter2 \cite{chen2024videocrafter2}. Text-to-video generation has progressed with models like Imagen Video \cite{ho2022video}, Make-A-Video \cite{singer2022make}, Phenaki \cite{villegas2022phenaki}, CogVideo \cite{hong2022cogvideo}, CogVideoX \cite{yang2024cogvideox}, Text2Video-Zero \cite{khachatryan2023text2video}, and ModelScopeT2V~\cite{wang2023modelscope}, which generate videos conditioned on textual descriptions, as known as the text-to-video (T2V) models. Hybrid approaches, such as Dual Diffusion Models \cite{xiao2023dual}, combine diffusion models with other generative frameworks to improve temporal coherence and resolution. There are also some recent advanced methods, like Lumiere~\cite{bar2024lumiere}, SF-V~\cite{zhang2024sf}, LaVie~\cite{wang2023lavie}, Pyramidal Flow Matching~\cite{jin2024pyramidal}. Diffusion transformer (DiT)~\cite{peebles2023scalable} further improves the scalability of the diffusion models by incorporating the transformer architecture, which allows to accommodate training videos with various resolutions and durations~\cite{polyak2024movie}. Despite these advancements, challenges like computational cost, temporal consistency, and suitable evaluation metrics remain, guiding future research in diffusion model-based video generation.

\paragraph{Efficiency of Diffusion Models.}
Diffusion models have achieved state-of-the-art results in generative tasks but are computationally intensive, requiring hundreds of sampling steps. DDIM \cite{song2021denoising} reduced the number of steps at inference time, but performance degrades significantly if it comes into the few-step regime. To address this, knowledge distillation for generative models is proposed to transfer knowledge from pre-trained teachers to students~\cite{luhman2021knowledge}.
Progressive Distillation \cite{salimans2022progressive} condensed multiple iterations into a single forward pass. With the distribution matching objective between the teacher and student models, score distillation is initially proposed for 3D generation with diffusion models~\cite{poole2022dreamfusion}. Variational score distillation (VSD) is later applied in 3D~\cite{wang2024prolificdreamer} and image generation~\cite{yin2024one, luo2024diff, xie2024distillation, salimans2024multistep}. Combining adversarial training with diffusion models is proposed for few-step image generation~\cite{xiao2021tackling, yin2024improved}.
Another branch of methods post alternative restrictions on the diffusion trajectories.
Consistency models \cite{song2023consistency} enabled one-step generation by training models to output consistent results across different noise levels. Latent consistency model (LCM)~\cite{luo2023latent} distills image diffusion models into consistency models. VideoLCM~\cite{wang2023videolcm} and AnimateLCM~\cite{wang2024animatelcm} apply consistency distillation from diffusion video models.  DPM-Solver \cite{lu2022dpmsolver} introduced a fast ODE solver, reducing diffusion sampling to around 10 steps.
Rectified flow~\cite{liu2022flow, lipman2022flow} is a special case of diffusion model, which enforces the straightness of the denoising trajectory during training to achieve high-quality few-step sampling.  Instaflow~\cite{liu2023instaflow} adopts this method to achieve one-step sampling for image generation. Limitations exist for present methods: VSD-based approach suffers from model collapse by generating less diverse samples after distillation; Consistency model methods tend to have lower fidelity and the samples are usually qualitatively worse than the teacher model.
\paragraph{Reward-based Fine-tuning.} To further improve image and video generation quality in aspects like aesthetic quality and text-image alignment, researchers recently proposed various reward-based fine-tuning methods for diffusion models~\cite{xu2024imagereward,chung2022diffusion, clark2023directly, prabhudesai2024video, li2024t2v, black2023training, clark2023directly, domingo2024adjoint}. The most common ones are direct reward gradients from a differentiable reward model. 
ReFL~\cite{xu2024imagereward} backpropagates the reward gradient through one-step predicted $x_0$ in DMs, similar as diffusion posterior sampling~\cite{chung2022diffusion}. 
DRaFT-$K$~\cite{clark2023directly} truncated the reward gradient backpropagation in diffusion process to latest $K$ steps. VADER~\cite{prabhudesai2024video} applies this on diffusion video models.
T2V-Turbo~\cite{li2024t2v} applies reward gradient for video diffusion models, with the gradients backpropagated through both the reward model and the decoder. The gradient is applied on distilled consistency models for one-step generation, to avoid multi-step backpropagation through DMs. Different from these, denoising diffusion policy optimization (DDPO)~\cite{black2023training} treats the denoising process as decision making process and applies stochastic policy gradient algorithms like REINFORCE and PPO to optimize it, without requiring the differentiable reward function. However, DDPO is found to be less sample efficient as reward-gradient method due to lack of the gradient information~\cite{clark2023directly}. To leverage the rich reward gradient information and bypass backpropagation through the large reward model and the decoder, we propose to use the latent reward model for gradient-based fine-tuning of diffusion models.
Adjoint Matching~\cite{domingo2024adjoint} casts the reward fine-tuning as a stochastic optimal control problem and proposes the memoryless flow matching method to ensure fine-tuned models converge to the tilted distribution.

\section{Methodology}
\label{sec:method}
\begin{figure*}[htbp]
    \centering
\includegraphics[width=\textwidth]{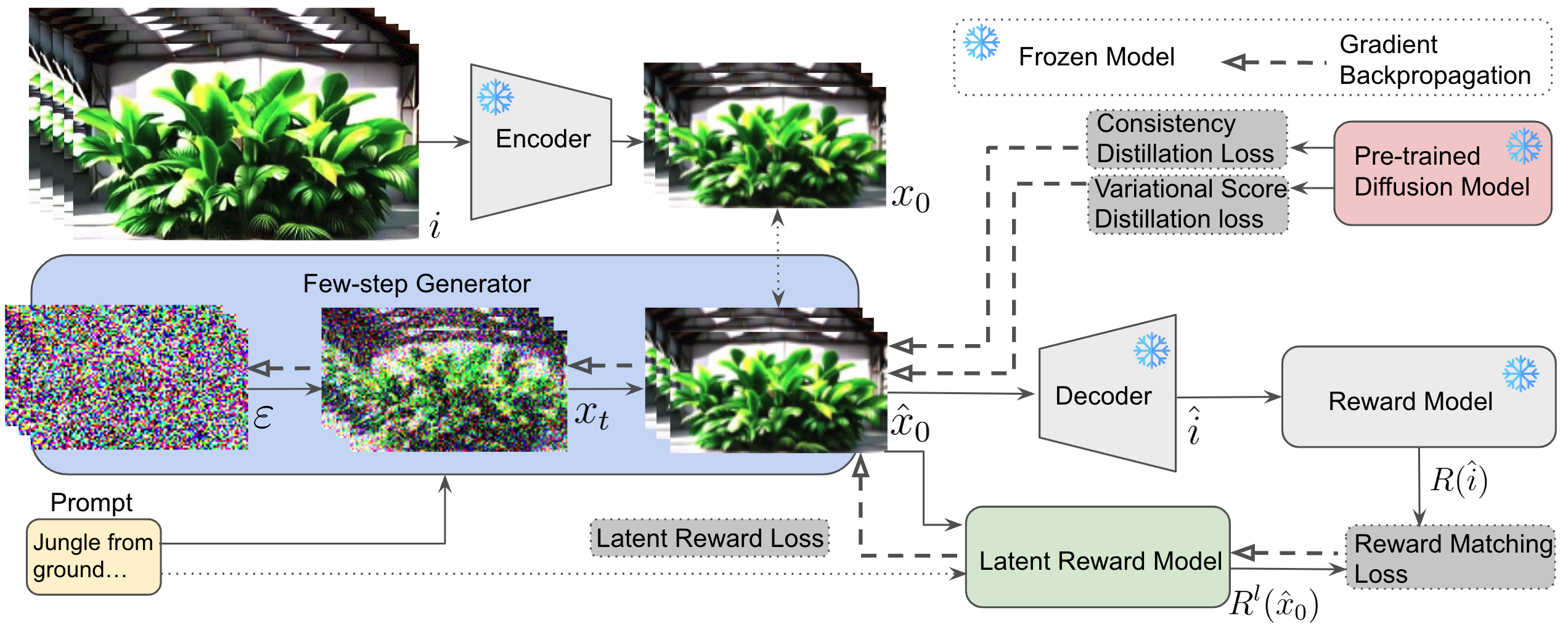}
    \caption{
    Method Overview: The few-step generator $G_\theta$ is trained to generate high-quality samples from random noise in latent space, guided by a combination of variational score distillation (VSD), consistency distillation (CD), and latent reward model (LRM) fine-tuning objectives. VSD loss enhances sample quality, albeit with a risk of mode collapse, while CD loss increases sample diversity without compromising generation quality. The LRM enables reward-based optimization to further improve sample quality, by bypassing the large, pixel-space reward model and the decoder, thereby reducing memory usage and removing the need for differentiable reward models.
    }
    \label{fig:diagram}
    \vspace{-.5cm}
\end{figure*}
The overview of our method is shown in Fig.~\ref{fig:diagram}.

\subsection{Diffusion Model}
\label{sec:diffusion}
Suppose the data distribution is $x_0\sim q(x_0)$, the diffusion model approximates this distribution by gradually denoising along a Markov chain. Forward diffusion follows $x_t\coloneqq F(x_0, t)=a_t x_0+b_t\varepsilon, \varepsilon\sim\mathcal{N}(0, \mathbf{I})$. For DDPM,
\begin{align}
    a_t=\sqrt{\bar{\alpha}_t}, b_t=\sqrt{1-\bar{\alpha}_t}
    \label{eq:ddpm_schedule}
\end{align}
with $\bar{\alpha}_t=\Pi_{i=1}^t \alpha_i$ following a pre-specified noise schedule $\alpha_t, t\in[T]$. For the general variance-preserving schedule~\cite{song2021score}, it satisfies $a_t^2+b_t^2=1$, therefore it can be equivalently written as $x_t=\cos(t) x_0 + \sin(t) \varepsilon, t\in[0, \frac{\pi}{2}]$ with a simple mapping of time sequences. Standard diffusion model optimization with velocity prediction $v_\theta$ follows the loss:
\begin{align}
    \mathcal{L}_\text{V}(\theta)&=\mathbb{E}_{x_0\sim q(x_0), \varepsilon\sim\mathcal
    {N}(0, \mathbf{I}), t}\big[w_t||v_\theta(x_t, t)-v_t||_2^2\big]\label{eq:diff_v_loss}\\
    v_t &= -\sin(t)x_0 + \cos(t)\varepsilon
    \label{eq:diff_v}
\end{align}
For rectified flow (RF)~\cite{liu2022flow} or flow matching~\cite{lipman2022flow}, $a_t=1-t, b_t=t, t\in[0,1]$, with a constant velocity target $v_t=\varepsilon-x_0, \forall t\in[0,1]$.

After training, the reverse diffusion process follows $x_{t-1}\coloneqq\text{Denoise}(x_t, t, \theta)$ iterative denoising. The training and inference details of our models refer to Sec.~\ref{app:training_inference}.

\subsection{Consistency Distillation}
\label{sec:cd}

Consistency model \cite{song2023consistency} enforces the consistency loss as the distillation method from a pre-trained teacher model $v_{\theta'}$, with a discrete sub-sampled time schedule $t_1=\epsilon< t_2<\dots < t_N=T$:
\begin{small}
\begin{align}
    \mathcal{L}_\text{CD}(\theta)&=\mathbb{E}_{x_0\sim q(x_0), t_n}[\lambda(t_n)d(f_\theta(x_{t_{n+m}}, t_{n+m}),f_{\theta^-}(\hat{x}_{t_n}, t_{n})]\label{eq:cd_loss}\\
    \hat{x}_{t_n}&=\text{Denoise}^m(x_{t_{n+m}}, t_{n+m}, t_n, \theta')
    \label{eq:xt_ode}
\end{align}
\end{small}
where $\lambda(t_n)$ is a time dependent coefficient usually set as constant in practice, $d(\cdot, \cdot)$ is a distance metric like MSE or Huber loss, and $\theta^-$ is exponential moving average of $\theta$. Instead of traditional one-step denoising with the teacher model, we apply a generalized CD with $\text{Denoise}^m(\cdot)$ indicating the $m$-step denoising function as defined by Eq.~\eqref{eq:denoise}, which iteratively predicts the sequence $(\hat{x}_{t_{n+m-1}}, \dots, \hat{x}_{t_{n}}|x_{t_{n+m}})$. This is practically found to improve generation quality. The student consistency function $f_\theta$ can be reparameterized from the neural network prediction, similar as in LCM~\cite{luo2023latent}:
\begin{align*}
    f_\theta(x_{t_n},t_n) &= c_{\text{skip}} x_{t_n} + c_{\text{out}}x_\theta(x_{t_n}, t_n)
\end{align*}

To enhance the distillation for conditional generation with conditional variable $c\in\mathcal{C}$ (\emph{e.g.}, text prompts), we applied the classifier-free guidance (CFG)~\cite{ho2022classifier} augmentation for the teacher denoising function, similar as VideoLCM~\cite{wang2023videolcm}, but for $v_\theta$-prediction in our case:
\begin{small}
\begin{align}
 v^w_\theta(x_{t_n}, t_n, c)=v_\theta(x_{t_n}, t_n, c) + w\big(v_\theta(x_{t_n}, t_n, c) - v_\theta(x_{t_n}, t_n, \varnothing)\big)
 \label{eq:cfg}
\end{align}
\end{small}
This is applied in replacement of $v_\theta$ in Eq.~\eqref{eq:x_v_trans} for conditional generation.

\subsection{Variational Score Distillation}
\label{sec:vsd}
Variational score distillation (VSD)~\cite{wang2024prolificdreamer} is proposed with the objective of distribution matching between the teacher and student models, by approximating the scores with properly trained diffusion models. Specifically, the loss of minimizing the Kullback-Leibler (KL) divergence between real (teacher) sample distribution $p_\text{real}$ and fake (student) sample distribution $p_\text{fake}$ has the form:
\begin{align}
    \mathcal{L}_\text{VSD}&\coloneqq D_\text{KL}(p_\text{fake}||p_\text{real})=\mathbb{E}_{x\sim p_\text{fake}}[\log \frac{p_\text{fake}(x)}{p_\text{real}(x)}]\\
    &=\mathbb{E}_{\varepsilon\sim \mathcal{N}(0, \mathbf{I}), x=G_\theta(\varepsilon)}[\log \frac{p_\text{fake}(x)}{p_\text{real}(x)}]
    \label{eq:vsd_loss}
\end{align}
and the derivative is,
\begin{small}
\begin{align}
    \nabla_\theta D_\text{KL}=\mathbb{E}_{\varepsilon\sim \mathcal{N}(0, \mathbf{I}), x=G_\theta(\varepsilon)}[-(s_\text{real}(x)-s_\text{fake}(x))\nabla_\theta G_\theta(\varepsilon)]
\end{align}
\end{small}
with score functions $s_\text{real}(x)=\nabla_x \log p_\text{real}(x)$ and $s_\text{fake}(x)=\nabla_x \log p_\text{fake}(x)$ for two distributions. $G_\theta(\cdot)$ is the generation process by the student network through iteratively denoising the noisy training samples. The real score $s_\text{real}$ is estimated with the pretrained teacher model.

\begin{figure}[htbp]
    \centering
\includegraphics[width=0.98\columnwidth]{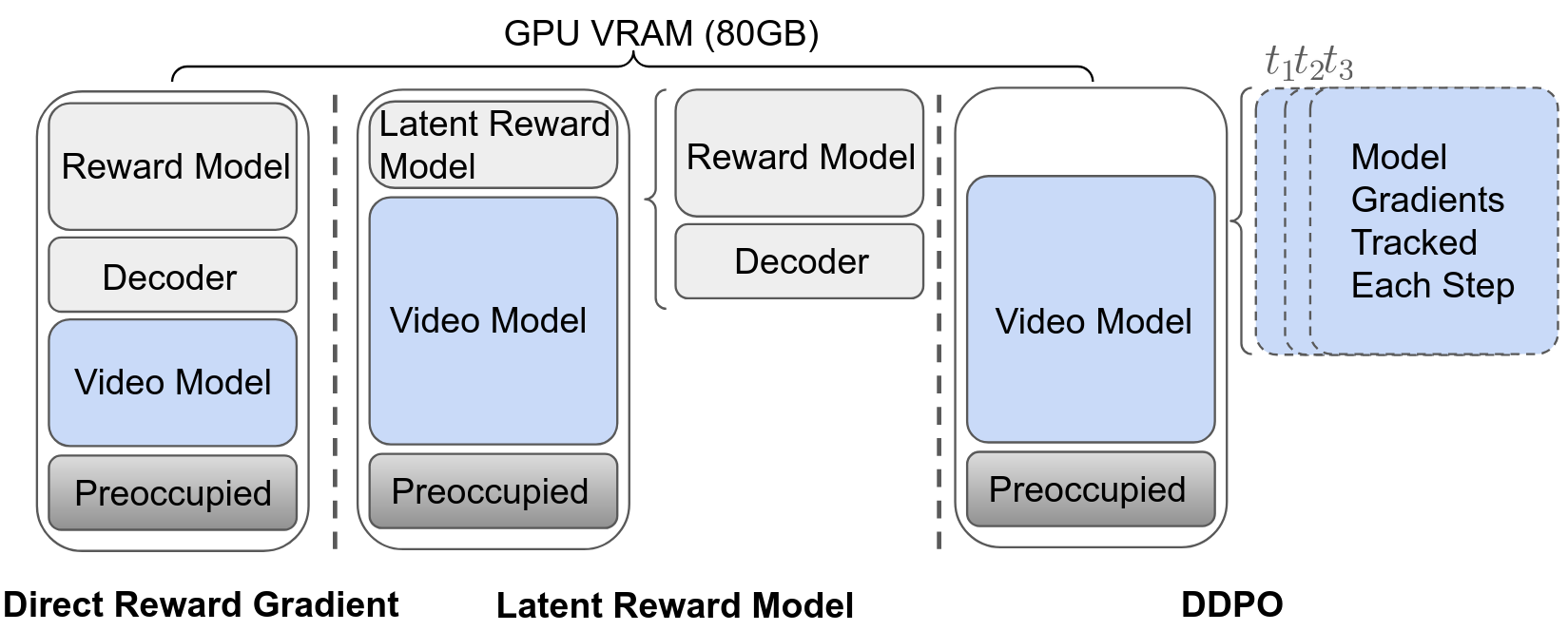}
    \caption{Comparison of different reward fine-tuning methods: 
    (1) Direct reward gradient methods are limited to small reward or video models or short input sequences, and they also require a differentiable reward model; (2) The latent reward model is compact and bypasses the decoder for gradient-based optimization, making it suitable when large reward models and decoders exceed available VRAM; (3) DDPO is similarly constrained by VRAM limits when handling large video models and tracking log-probabilities of samples over multiple steps.
    }
    \label{fig:compare_lrm}
    \vspace{-.5cm}
\end{figure}

For accurately estimating the fake score $s_\text{fake}$, the fake diffusion model  $v_{\theta_\text{fake}}$ is initialized from the teacher and dynamically adapts according to the student sample distribution. For the score estimation purpose, the fake score model is updated with the diffusion loss $\mathcal{L}_\text{CV}(\theta_\text{fake})$ following Eq.~\eqref{eq:insta_loss}, on student generated samples.

State-of-the-art distillation method distribution matching distillation (DMD)~\cite{yin2024one, yin2024improved} mainly applies the VSD loss for student distillation. We abandon the adversarial loss in \cite{yin2024improved} since the GAN training is usually unstable and the improvement can be marginal. We replace the adversarial training with the consistency distillation loss and reward model fine-tuning, which generates richer gradient signals as evidenced in our experiments.

\subsection{Latent Reward Fine-tuning}

Reward fine-tuning is an effective approach to align the sample distribution with the specified preference metric in the post-training phase, as evidenced Fig.~\ref{fig:compare_hps}.

Previous reward-based optimization methods either (1) requires direct gradients from the pixel-space reward models~\cite{xu2024imagereward, clark2023directly}, or (2) relies on the log-probability estimation of the samples for multiple diffusion steps like DDPO~\cite{black2023training}, as compared in Fig.~\ref{fig:compare_lrm}. One major drawback of (1) is that it only works for differentiable reward function, while not feasible for non-differentiable ones like JPEG compressibility~\cite{black2023training}, etc. Apart from that, the reward models usually work for raw RGB pixel space, which requires the reward gradient to backpropagate through not only the large reward models, but also the decoder, as the practical framework usually follows LDM~\cite{rombach2022high} for latent space modeling. Method (2) is found to be less efficient in reward optimization due to lack of rich reward gradient information~\cite{clark2023directly}, and occupies more memory due to gradient estimation over multiple diffusion steps. A concurrent work~\cite{li2024reward} explores the reward in latent space, but is limited to image generation. Video models have a pressing need for compact reward models, as a largely unexplored area, where leveraging image-based rewards for video generation remains non-trivial.

We propose to learn a dual latent reward model (LRM) for directly optimizing the video diffusion model in the latent space, which supports any type of reward metrics as detailed in Sec.~\ref{app_sec:lrm_reward_types}. Here we take image rewards as an example. Consider a provided image reward model $\mathcal{R}:\mathcal{I}\rightarrow \mathbb{R}$ with RGB image $i\in\mathcal{I}$ as its input, we approximate the LRM $\mathcal{R}^l_\phi:\mathcal{X}\bigcup \mathcal{X}'\rightarrow \mathbb{R}$ with loss:
\begin{align}
    \mathcal{L}_\text{LRM}(\phi)=\mathbb{E}_{x\in{\mathcal{X}\bigcup \mathcal{X}'}}\big[\big(\mathcal{R}(\text{Dec}(x))- \mathcal{R}^l_\phi(x)\big)^2\big]
    \label{eq:lrm_loss}
\end{align}
where $\mathcal{X}'=\{G_\theta(\varepsilon)\}$ is the set of generated samples from the generator, and $\text{Dec}(\cdot)$ is the decoder. We use both real images and generated images to improve the robustness of learned LRM on generated samples. To alleviate the computational burden in training the LRM, we apply Eq.~\eqref{eq:x_v_trans} for single-step prediction of generated samples, rather than iteratively denoising along the entire trajectory. This approach significantly reduces memory usage by avoiding gradient backpropagation through the iterative sampling process. Although our distilled student models operate with a maximum of 4 sampling steps, memory usage can still be intensive if samples are generated with a full denoising process. Tab.~\ref{tab:model_comparison_memory} compares the parameter counts and memory costs for pixel-space reward models and LRMs on video samples with a batch size of 1. HPSv2 and PickScore are two pixel-space reward models used in our experiments (as described in Sec.~\ref{sec:reward_tune}).

\begin{table}[htbp]
\centering
\caption{Comparison of parameters and GPU memory (VRAM) costs and for pixel-space HPSv2, PickScore reward models and LRMs. }
\resizebox{0.8\columnwidth}{!}{
\begin{tabular}{@{}l|c|c|c@{}}
\toprule
Model                     & \# Parameters   & Forward VRAM   & Backward VRAM   \\ \midrule
Image LRM       & 189,441            & 8.998 MB                 & 17.772 MB                     \\
Text-image LRM & 763,009            & 15.500 MB                & 26.277 MB                          \\
HPSv2/PickScore    & 632 million        & 5.926 GB                & $>$90 GB                             \\ \bottomrule
\end{tabular}
}
\label{tab:model_comparison_memory}
\vspace{-.4cm}
\end{table}

With the compact and differentiable LRM on the latent space, we apply direct reward gradient optimization to fine-tune the diffusion model:
\begin{align}
    \mathcal{L}_\text{FT}(\theta; \phi) = -\mathbb{E}_{\varepsilon\sim \mathcal{N}(0, \mathbf{I})}[\mathcal{R}^l_\phi(G_\theta(\varepsilon))]
    \label{eq:lrm_ft_loss}
\end{align}

In practice, we can either pre-train the LRM first and then fine-tune the diffusion model with a fixed LRM, or train the LRM and fine-tune the diffusion model iteratively. For simplicity, we adopt the second approach. If the original reward model is conditioned on additional text input, $\mathcal{R}(i, c)$, the LRM also operates conditionally as $\mathcal{R}^l(x, c), c \in \mathcal{C}$. \textit{The LRM method can accommodate any type of reward models, including image, video, text-image, and text-video rewards.} For image-only LRMs, we use a convolutional neural network, while for text-image LRMs, we apply a cross-attention module after the convolutional feature extractor to integrate text embeddings with image features. For video-based LRMs, the 2D convolution is replaced with a 3D convolutional neural network. Additional details are provided in Sec.~\ref{app_sec:lrm_reward_types}.

\section{Multi-Objective Distillation}
Distillation using VSD (as in DMD method) alone can result in severe mode collapse, while CD (as in LCM method) tends to produce lower-quality samples by averaging across sample distributions, with evidence shown in Fig.~\ref{fig:full_compare}. By integrating consistency distillation, variational score distillation, and latent reward fine-tuning objectives, our method enables few-step generation of high-quality, diverse samples after distillation, optimized by the following loss function:
\begin{align}
    \mathcal{L}(\theta) = \mathcal{L}_\text{VSD}(\theta) + \beta_\text{CD} \mathcal{L}_\text{CD}(\theta) + \beta_\text{FT} \mathcal{L}_\text{FT}(\theta;\phi)
\end{align}
During distillation, the fake score network is updated with $\mathcal{L}_\text{CV}(\theta_\text{fake})$, and the LRM $\mathcal{R}^l_\phi$ is updated using $\mathcal{L}_\text{LRM}(\phi)$. 
Pseudo-code of our method is provided in Sec.~\ref{app_sec:code}.

\section{Experiments}
\subsection{Implementation}
\label{sec:implement}
\paragraph{Student and teacher models.}
The video diffusion model in our experiments is based on the diffusion transformer (DiT) architecture~\cite{peebles2023scalable}. Although our methodology is architecture-agnostic and could be applied to models like U-Net~\cite{rombach2022high}, we select the transformer due to its scalability. The teacher and student T2V diffusion models are same as a modified variant of Open-Sora~\cite{opensora} and follow the LDM framework~\cite{rombach2022high, blattmann2023align}, utilizing DiT modeling in the latent space encoded with a pretrained 3D variational autoencoder (VAE)~\cite{kingma2013auto, opensora}. The 3D VAE encodes and decodes videos chunk-by-chunk to alleviate the computational burden, encoding chunks of 16 video frames into 5 latent embeddings. These embeddings are then patchified into sequences as inputs to the DiT. Leveraging the DiT architecture, the model can accommodate arbitrary video durations and resolutions; however, our experiments constrain the video generation setting to 128 frames at a resolution of $192 \times 320$, resulting in a patchified sequence of length 9600.


\paragraph{Model training and inference.}
The teacher model employs DDPM with 1000 sampling steps but uses DDIM for inference with 50 steps, while the student model is distilled to a 4-step sampling similar as previous work~\cite{li2024t2v2}, with CFG weights $7.5$ for CD. The distillation experiment for each student is on $8$ A100 GPUs.
More implementation details see Sec.~\ref{app:implementation_details}.

\paragraph{Reward Metrics.}
We utilize Human Preference Score v2 (HPSv2)~\cite{wu2023human} and PickScore~\cite{kirstain2023pick} as the text-image reward models. Both are fine-tuned CLIP-type models trained on extensive text-to-image datasets with human preferences. While our methods are compatible with directly optimizing the model using VBench reward metrics, we intentionally avoid doing so, as VBench scores serve as one of the final evaluation criteria. Optimizing directly for specific VBench metrics or simple image rewards such as JPEG compressibility can significantly improve reward scores, but it may also lead to overoptimization for specific metrics and degrade overall generation quality. Consequently, we adopt the more general preference-based reward models, HPSv2 and PickScore, by default for reward fine-tuning. Nonetheless, our method remains compatible with other reward models.

\paragraph{Evaluation.}
To faithfully reflect model performance, we apply both automatic evaluation benchmark VBench~\cite{huang2024vbench} and human evaluation for our results. VBench assesses 16 dimensions encompassing both video visual quality and semantic alignment aspects for T2V models, with higher scores indicating better performance in each metric. Following the standard VBench evaluation protocol, we use a set of 946 long prompts, generating five videos per prompt. Final scores for each dimension are averaged across all generated videos for that metric.

Additionally, we examine the impact of prompt length by comparing the performance of long descriptive prompts with short prompts in VBench evaluation. To further assess text-video alignment capabilities, we sample the distilled student models with various styles and motions.

\subsection{Comparison with SOTA Methods and Models}
\begin{figure}[htbp]
    \centering
\includegraphics[width=0.8\columnwidth]{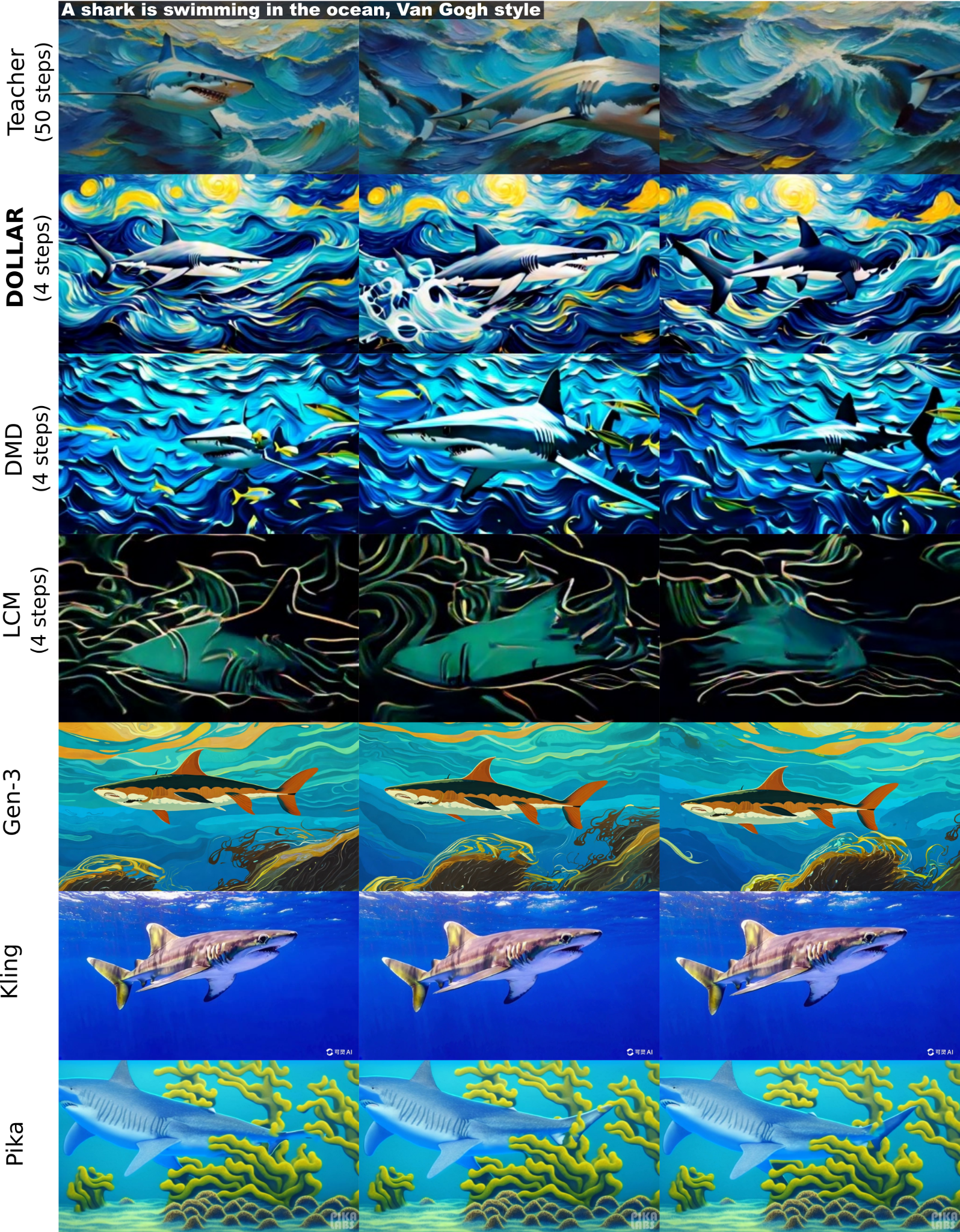}
    \caption{Video samples synthesized with our method and SOTA distillation methods and top video models (no distillation) on VBench.
    }
    \label{fig:full_compare}
\end{figure}

\paragraph{Comparison with SOTA Distillation Methods.}
In Fig.~\ref{fig:full_compare}, we compare the samples of our distilled models with SOTA distillation methods including (1). VSD based method for DMD~\cite{yin2024one, yin2024improved}; (2). CD based method typically for LCM~\cite{luo2023latent} and VideoLCM~\cite{wang2023videolcm}, while also applied by T2V-Turbo~\cite{li2024t2v2} and FastVideo~\cite{zhang2025fast}. These distillation methods apply 4-step inference. Samples by our teacher model with 50 DDIM steps are also displayed. LCM generates blurry samples with lower quality than others, due to the high volatility in training and sensitivity to hyperparameters, although it is possible to achieve better performance with comprehensive hyperparameter search. DMD has lower fidelity and sample diversity compared with ours. The LRM on distilled few-step models further enhances quality and alignment, surpassing the teacher model's performance limits. Due to the gradient backpropagation nature of reward gradients, applying them to teacher models with large sampling steps is challenging, whereas they are well-suited for distilled students.


\begin{table}[htbp]
\centering
\caption{Comparison of VBench scores for different models.}
\resizebox{\columnwidth}{!}{%
\begin{tabular}{@{}cccccc|ccc@{}}
\toprule
Model & Pika & Gen-2 & Gen-3 & Kling & T2V-Turbo & Teacher & Our (PickScore) & Our (HPSv2) \\ \midrule
Subject C.& \underline{96.76} & \underline{97.61} & 97.10 & \textbf{98.33} & 96.28 & 83.99 & 93.77 & 92.57 \\
Background C. & \textbf{98.95} & \underline{97.61} & 96.62 & \underline{97.60} & 97.02 & 93.78 & 96.80 & 96.14 \\
Temporal F. & \textbf{99.77} & \underline{99.56} & 98.61 & \underline{99.30} & 97.48 & 96.42 & 96.30 & 97.48 \\
Motion S. & \underline{99.51} & \textbf{99.58} & 99.23 & \underline{99.40} & 97.34 & 98.09 & 97.76 & 98.59 \\
Dynamic D.& 37.22 & 18.89 & 60.14 & 61.21 & 49.17 & \textbf{99.44} & \underline{75.83} & \underline{81.67} \\
Aesthetic Q. & 63.15 & \textbf{66.96} & \underline{63.34} & 46.94 & 63.04 & 61.21 & \underline{63.80} & 63.14 \\
Imaging Q.& 62.33 & \underline{67.42} & 66.82 & 65.62 & \textbf{72.49} & 63.87 & \underline{69.40} & 65.61 \\
Object C.& 87.45 & 90.92 & 87.81 & 87.24 & \textbf{93.96} & 85.79 & \underline{91.63} & \underline{93.84} \\
Multiple O. & 46.69 & 55.47 & 53.64 & \underline{68.05} & 54.65 & 52.59 & \underline{69.71} & \textbf{72.21} \\
Human A.& 88.00 & 89.20 & 96.40 & 93.40 & 95.20 & \textbf{99.60} & \underline{99.00} & \underline{99.00} \\
Color & 85.31 & \underline{89.49} & 80.90 & \textbf{89.90} & \textbf{89.90} & 77.00 & 77.95 & 74.78 \\
Spatial R.& 65.65 & 66.91 & 65.09 & \textbf{73.03} & 38.67 & 51.40 & \underline{68.56} & \underline{68.35} \\
Scene & 44.80 & 48.91 & \underline{54.57} & 50.86 & \textbf{55.58} & 49.99 & \underline{55.06} & 52.72 \\
T. Style & 24.44 & 24.12 & 24.71 & 24.17 & \underline{25.51} & \textbf{26.45} & 24.64 & \underline{25.23} \\
A. Style & 21.89 & 24.31 & \textbf{24.86} & 19.62 & 24.42 & \underline{24.83} & \underline{24.45} & 23.50 \\
Overall C. & 25.47 & 26.17 & 26.69 & 26.42 & \textbf{28.16} & \underline{27.89} & \underline{26.93} & 26.85 \\
\midrule
Quality & 82.68 & 82.47 & \textbf{84.11} & 83.39 & 82.57 & 81.89 & \underline{83.49} & \underline{83.83} \\
Semantic & 71.26 & 73.03 & 75.17 & \underline{75.68} & 72.57 & 73.71 & \textbf{77.90} & \underline{77.51} \\
\textbf{Total} & 80.40 & 80.58 & \underline{82.32} & 81.85 & 81.01 & 80.25 & \underline{82.37} & \textbf{82.57} \\
\bottomrule
\end{tabular}%
}
\label{tab:compare_vbench}
\end{table}

\begin{figure}[htbp]
    \centering
\includegraphics[width=\textwidth]{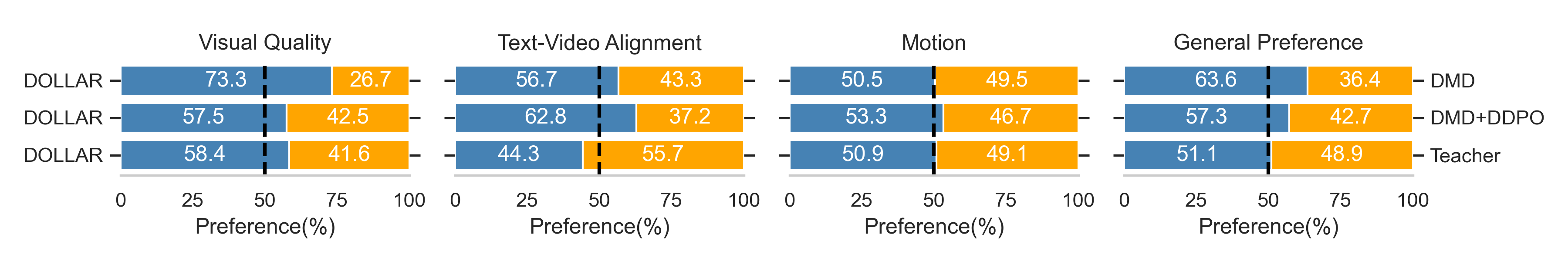}
    \caption{Human preference results for our DOLLAR method (with HPSv2) against baselines. The percentage indicates the preference rate.
    }
    \label{fig:preference}
\end{figure}
\paragraph{Comparison with SOTA Models on VBench.}
The VBench evaluation results are summarized in Tab.~\ref{tab:compare_vbench}, with the highest in bold and 2nd and 3rd underlined. Our distilled methods with VSD+CD+LRM achieve superior performance over the baselines including Pika~\cite{pikalabs2023}, Gen-2, Gen-3, Kling~\cite{kuaishou2024} and T2V-Turbo~\cite{li2024t2v} with full inference steps, as well as our teacher model. The visualization is displayed in Fig.~\ref{fig:full_compare}. The highest semantic scores of our models indicate a significant improvement over baselines for text-video alignment. The quality score, which reflects the visual quality, is heavily affected by the frame consistency metrics like subject consistency, background consistency, temporal flickering and motion smoothness, which are usually high if there is a lack of motions in the videos. Our models have significantly higher dynamics degree for motions as shown in the table and Fig.~\ref{fig:full_compare}. The total score is a weighted sum of all metrics showing the general preference of the videos, and our method achieves 82.37 and 82.57 surpassing all models in the table, as well as outperforming the teacher model. The students achieve higher scores in 9-10 metrics (out of 16) than the teacher. It indicates that the performance of our method is not upper bounded by the teacher model, which is beyond the VSD loss for student and teacher distribution matching. The additional CD loss enforces the self-consistency of model prediction on noisy real images. It provides the source of signals to improve the student model over teacher model on quality and semantic performances, which are further boosted by LRM fine-tuning.

\paragraph{Human Evaluation.}
We further conduct human evaluation to visually compare the generated videos for different models, over four independent metrics: visual quality, text-video alignment, motion, and general preference. The main evaluation results are shown in Fig.~\ref{fig:preference}. From the evaluated results, our method with HPSv2 reward is preferred more than the DDPO method (by $57.3\%$) and teacher model (by $51.1\%$), and performs similarly with the Gen-3 model (by $45.6\%$) in  terms of general preference. The visual quality of our distilled students is significantly higher than both teacher (by $58.4\%$) and Gen-3 (by $55.9\%$). Moreover, we find that, PickScore increases visual quality, but likely leads to worse motion performance. HPSv2 reward tuning not only increases the visual quality, but has better motion and text-video alignment.

\subsection{Comparison with Pixel-Space Reward}

\label{sec:reward_tune}
We compare our LRM with pixel-space reward fine-tuning method DDPO~\cite{black2023training}, which applies the REINFORCE algorithm to optimize the diffusion model by treating the diffusion process as a MDP. More details refer to Sec.~\ref{sec:ddpo}. Direct reward gradient methods like ReFL~\cite{xu2024imagereward} and DRaFT~\cite{clark2023directly} exceed single-GPU memory capacity in our case, thus are not included as baselines.

\begin{figure}[htbp]
    \centering
\includegraphics[width=\columnwidth]{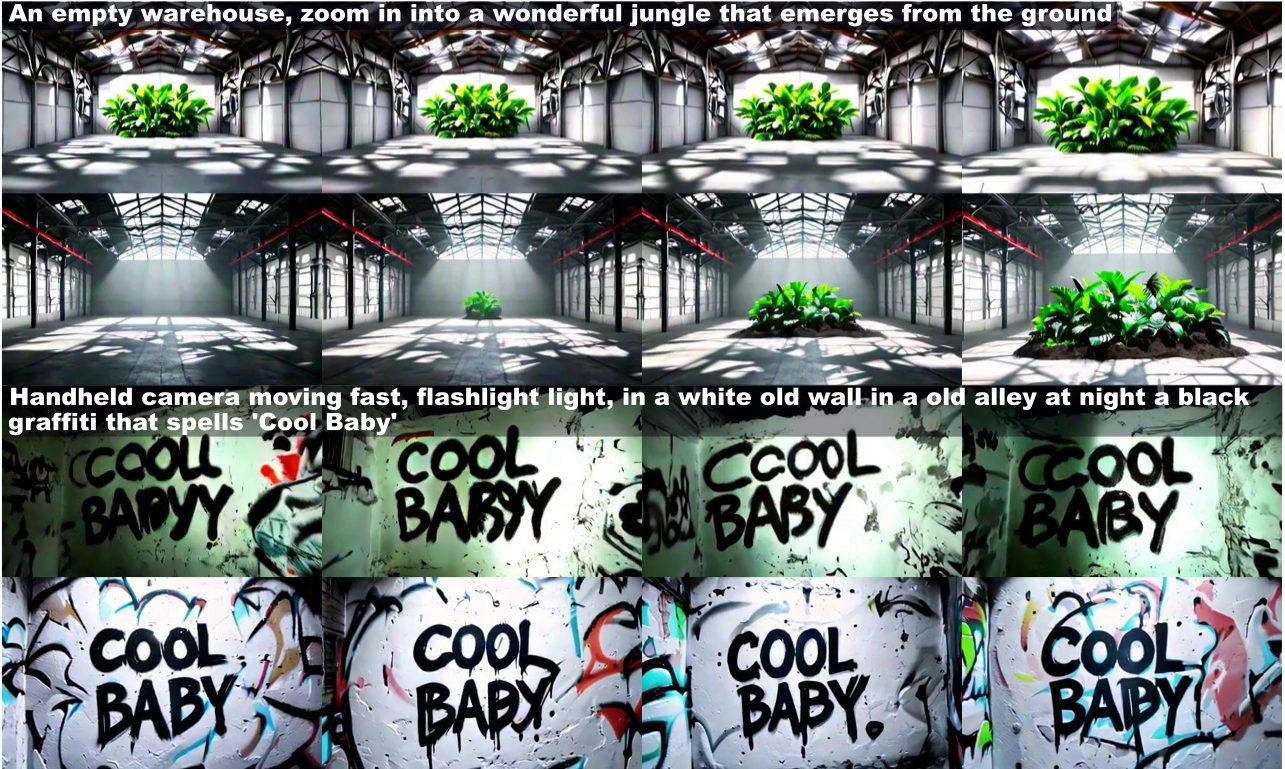}
    \caption{Compare the generated samples with (first line) and without (second line) reward fine-tuning for two samples. First sample: 4 frames are extracted from one sampled video per method along the time sequence. Second sample: one frame is extracted from one video, with 4 videos sampled by the same prompt.
    }
    \label{fig:compare_hps}
    \vspace{-.5cm}
\end{figure}

Tab.~\ref{tab:reward_method_compare} compares VBench scores and final reward values for LRM and DDPO. The last row ``Reward'' indicates the corresponding reward value after fine-tuning, for example, PickScore value is reported if the model is fine-tuned with PickScore reward model, and similar for HPSv2. The mean and standard deviation values are reported with 500 videos generated under VBench prompts. As visualized in Fig.~\ref{fig:compare_hps}, the reward fine-tuning helps to improve the text-image alignment for the first prompt by more explicitly exhibiting the ``emerging'' effect, and improves the accuracy of text display in frames for the second prompt. The lighting style is also improved through fine-tuning. More details of DDPO and LRM methods including the learning curves for reward optimization are provided in Sec.~\ref{app_sec:rm_finetune}.


\subsection{Ablation of Distillation Methods}
\label{sec:distill_compare}
Tab.~\ref{tab:diversity} shows the ablation of our distillation method by comparing it with VSD and VSD+CD. VSD has comparable performances with teacher, with additional CD loss it increases the sample diversity. Our VSD+CD+LRM method achieves high sample quality and diversity overall.
\begin{table}[htbp]
\centering
\caption{Comparison of teacher and students with different distillation methods, with 4-step sampling for student models.}
\resizebox{0.8\columnwidth}{!}{
\begin{tabular}{p{2.5cm}|c|ccc@{}}
\toprule
Model & Teacher & \multicolumn{3}{|c}{Student} \\ \midrule
Method & DDIM 50 steps & VSD & VSD+CD & VSD+CD+LRM \\ \midrule
Quality Score & 81.89 & 80.95 & 82.16 & \textbf{83.83} \\ \hline
Semantic Score & 73.71 & 76.61 & 74.58 & \textbf{77.51} \\ \hline
Total Score & 80.25 & 80.08 & 80.65 & \textbf{82.57} \\ \hline
Vendi (Pixel)$\uparrow$ & $1.46\pm0.14$ & $1.49\pm0.14$  & $1.59\pm0.17$ &  $\mathbf{1.60\pm0.14}$ \\ \hline
Vendi (Inception)$\uparrow$ & $\mathbf{2.34\pm0.16}$ & $1.91\pm0.14$  & $2.14\pm0.15$ & $1.98\pm0.14$\\ 
\bottomrule
\end{tabular}
}
\label{tab:diversity}
\end{table}

\paragraph{Diversity Measure.}
The diversity of model generation is not captured by the VBench. We conduct both qualitative and quantitative comparison of generation diversity for different distillation methods. We quantitatively measure the diversity of sampled videos with Vendi score~\cite{friedman2022vendi}, which is based on the similarity matrix for the sample set. The mean and standard deviations across prompted video samples are reported in Tab.~\ref{tab:diversity}.
We find that the Inception-based Vendi score aligns better with visual inspection than pixel-based alternatives. While VSD produces high-quality samples, it tends to lead to mode collapse. Our method addresses this diversity limitation by incorporating CD loss, and further enhances generation quality through LRM fine-tuning.





\paragraph{Inference Time.}
Tab.~\ref{tab:inference_time_percentage} presents the per-sample inference time consumption for the teacher model using 50-step DDIM inference, and for student models with 1, 2, and 4 inference steps. Here, ``diffusion time" refers solely to the diffusion sampling in latent space, while ``inference time" encompasses the complete generation process for one video, including text encoding, diffusion sampling, and decoding of latent outputs.
The inference experiments are conducted on a single A100 80GB GPU, with mean and standard deviation calculated over 100 samples. With parallel sampling across multiple GPUs, the amortized time per sample can be further minimized. The reported values indicate the percentage of the teacher model's inference time, excluding amortization effects. Distilled student models significantly accelerate diffusion sampling compared to the teacher, achieving speedups from $\times 15.6$ (4 steps) to $\times 278.6$ (1 step).


\begin{table}[htbp]
\centering
\caption{Time consumption of teacher and distilled student models (as percentage of teacher's total inference time) with different numbers of function evaluations (NFE)}
\begin{tabular}{@{}l|c|ccc@{}}
\toprule
Model & Teacher & \multicolumn{3}{c}{Student} \\ \midrule
Steps (NFE) & 50  & 4 & 2 & 1 \\ \hline
Diffusion Time (\%) & $91.94\pm0.32$ & $5.88\pm0.03$ & $2.16\pm0.01$ & $0.33\pm0.02$ \\ \hline
Inference Time (\%) & $100.00\pm0.66$ & $13.06\pm0.17$ & $9.30\pm0.11$ & $7.45\pm0.12$ \\
\bottomrule
\end{tabular}
\label{tab:inference_time_percentage}
\end{table}


\begin{table}[htbp]
\centering
\caption{Comparison of LRM with DDPO using VBench and training reward metrics. DOLLAR is our final method with VSD+CD+LRM.}
\resizebox{\columnwidth}{!}{
\begin{tabular}{@{}l|ccc|ccc@{}}
\toprule
Reward Model & \multicolumn{3}{c|}{PickScore} & \multicolumn{3}{c}{HPSv2} \\ \midrule
Method & VSD+DDPO & VSD+LRM & DOLLAR &  VSD+DDPO & VSD+LRM & DOLLAR   \\ \midrule
Quality & 82.99 & \textbf{84.01} & 83.49  & 82.97 & 83.53 & 83.83\\ \hline
Semantic & 77.26 & 72.51 & \textbf{77.90}  & 74.56 & 75.67 & 77.51\\ \hline
Total Score & 81.84  & 81.71 & 82.37  & 81.29 & 81.96 & \textbf{82.57}\\ \hline
Reward & 0.207 & 0.207 & \textbf{0.210 }& 0.271 & 0.276 & \textbf{0.277} \\
\bottomrule
\end{tabular}
}
\label{tab:reward_method_compare}
\end{table}

\subsection{Ablation Studies}
\label{sec:exp_ablation}
\paragraph{Distillation Timesteps.}
Our proposed method supports an arbitrary subset of timesteps for teacher sampling. By default, we use 4-step sampling for the student model to balance quality and efficiency, as discussed in Sec.~\ref{sec:implement}. Here, we investigate the impact of varying the number of sampling steps during distillation, specifically testing 1 step (timestep [999]), 2 steps (timesteps [499, 999]), and 4 steps (timesteps [249, 499, 749, 999]) with equal spacing. While our approach does not require equal spacing, this configuration is used for consistency in this experiment.
The evaluated VBench scores are reported in Tab.~\ref{tab:infer_steps}.
All three distilled student models with VSD loss demonstrate comparable or even superior performances relative to the teacher model with 50 inference steps. The slight differences can be attributed to checkpoint selection and evaluation variance. From visual inspection and human evaluation, we find that models with more inference steps tend to perform better, which may not be fully captured by the minor differences in VBench scores.

\begin{table}[htbp]
    \centering
    \begin{minipage}{0.5\columnwidth}
        \centering
\resizebox{0.8\columnwidth}{!}{       
\begin{tabular}{c|c|ccc}
\toprule
Model & Teacher & \multicolumn{3}{|c}{Student (VSD)} \\ \midrule
Inference Steps & $50$ & $1$ & $2$ & $4$ \\ \midrule
Quality & 81.89 & 81.61  & 82.71 & 80.95 \\ \hline
Semantic & 73.71 & 76.66  & 73.86 & 76.61 \\ \hline
Total & 80.25  & 80.62  & 80.94 & 80.08 \\
\bottomrule
\end{tabular}
}
        \caption{Comparison of the number of inference steps for distilled students with VSD using VBench (long prompt).}
        \label{tab:infer_steps}
    \end{minipage}%
    \hspace{-.1pt}
    \begin{minipage}{0.48\columnwidth}
        \centering
\resizebox{0.8\columnwidth}{!}{
\begin{tabular}{c|c|cc}
\toprule
Model & Teacher & \multicolumn{2}{|c}{Student (VSD+CD)} \\ \midrule
CD w/ Denoise$^m$ & - & $m=1$ & $m=5$ \\ \midrule
Quality & 81.89 & 80.75  & 82.16  \\ \hline
Semantic & 73.71 & 71.57  & 74.58 \\ \hline
Total & 80.25  & 78.92  & 80.65 \\
\bottomrule
\end{tabular}
}
        \caption{Effect of the number of teacher denoising steps in consistency distillation (CD) on VBench (long prompt).}
        \label{tab:cd_steps}
    \end{minipage}
\end{table}

\paragraph{Consistency Distillation Denoising Steps.}
In Sec.~\ref{sec:cd}, we introduced the consistency distillation method with a multi-step teacher denoising function: Denoise$^m(\cdot)$. We ablate the choice of $m$ in experiments and find that a larger value like $m=5$ improves distillation performance, as detailed in Tab.~\ref{tab:cd_steps}. The student models follow 4-step schedule, and CD loss is applied on a 50-step DDIM schedule with step size $20$ as previously discussed.

\section{\texttt{DOLLAR} Method Details}
\subsection{Pseudo-code}
\label{app_sec:code}
The pseudo-code of our \texttt{DOLLAR} method is displayed as Alg.~\ref{alg:code}
\begin{algorithm}
    [htbp]
    \caption{Training procedure of \texttt{DOLLAR}}
    \small
    \begin{algorithmic}
        [1] \STATE \textbf{Input:} Pretrained teacher model $v_{\theta'}$ by $\mathcal{L}_\text{CV}$ Eq.~\eqref{eq:insta_loss}, pretrained encoder and decoder, dataset $\mathcal{D}=\{(c, i)\}$
        \STATE \textbf{Output:} Distilled student few-step generator $G_\theta$
        \STATE \texttt{//Initialize student and fake score model from teacher}
        \STATE $\theta\leftarrow \theta'$, $\theta_\text{fake}\leftarrow \theta'$
        \WHILE{\text{train}}
        \STATE Sample batch $(c,i)\sim\mathcal{D}$, encode $x\leftarrow \text{Encoder}(i)$
        \STATE \texttt{//Update the generator with distillation}
        \STATE $\hat{x}\leftarrow G_\theta(c, \varepsilon), \varepsilon\sim\mathcal{N}(0, \mathbf{I})$
        \STATE Uniformly sample $t_n$, forward diffusion $x_{t_{n+m}}\leftarrow F(x, t_{n+m})$
        \STATE $\mathcal{L}_\text{G}=\mathcal{L}_\text{VSD}(\theta;\theta', \theta_\text{fake}, \hat{x}, c)+\eta_1\mathcal{L}_\text{CD}(\theta;\theta', x_{t_{n+m}}, c)$ \texttt{//VSD by Eq.~\eqref{eq:vsd_loss}, CD by Eq.~\eqref{eq:cd_loss}}
        \STATE $G_\theta\leftarrow \text{GradientDescent}(\theta, \mathcal{L}_\text{G})$
        \STATE \texttt{//Update fake score model}
        \STATE Uniformly sample $t$, forward diffusion $x_t\leftarrow F(\hat{x}, t)$
        \STATE $\theta_\text{fake}\leftarrow \text{GradientDescent}(\theta_\text{fake}, \mathcal{L}_\text{CV}(x_t))$ \texttt{//Eq.~\eqref{eq:insta_loss}}
        \STATE \texttt{//Train latent reward model}
        \STATE Merge batch $\tilde{x}=x\bigcup \hat{x}$
        \STATE $\mathcal{R}^l_\phi\leftarrow \text{GradientDescent}(\phi, \mathcal{L}_\text{LRM}(\phi; \tilde{x}, \mathcal{R}))$ \texttt{//Eq.~\eqref{eq:lrm_loss}}
        \STATE \texttt{//Update the generator with latent reward fine-tuning}
        \STATE $G_\theta\leftarrow \text{GradientDescent}(\theta, \mathcal{L}_\text{FT}(\theta; \hat{x}, \mathcal{R}^l))$ \texttt{//Eq.~\eqref{eq:lrm_ft_loss}}
        \ENDWHILE
    \end{algorithmic}
\label{alg:code}
\end{algorithm}

\subsection{Diffusion Model Training and Inference}
\label{app:training_inference}
\paragraph{Conjugate Prediction Objective.}
Instead of applying noise prediction in previous work~\cite{ho2020denoising, rombach2022high} and the standard velocity prediction objective as in Instaflow~\cite{liu2023instaflow}, we apply a \textit{conjugate} velocity prediction objective:
\begin{align}
    \mathcal{L}_\text{CV}(\theta)=\mathbb{E}_{x_0\sim q(x_0), \varepsilon\sim\mathcal
    {N}(0, \mathbf{I}), t}\big[||v_\theta(x_t, t)-(x_0 - \varepsilon)||_2^2\big]
    \label{eq:insta_loss}
\end{align}
\begin{figure}[htbp]
    \centering
\includegraphics[width=0.4\columnwidth]{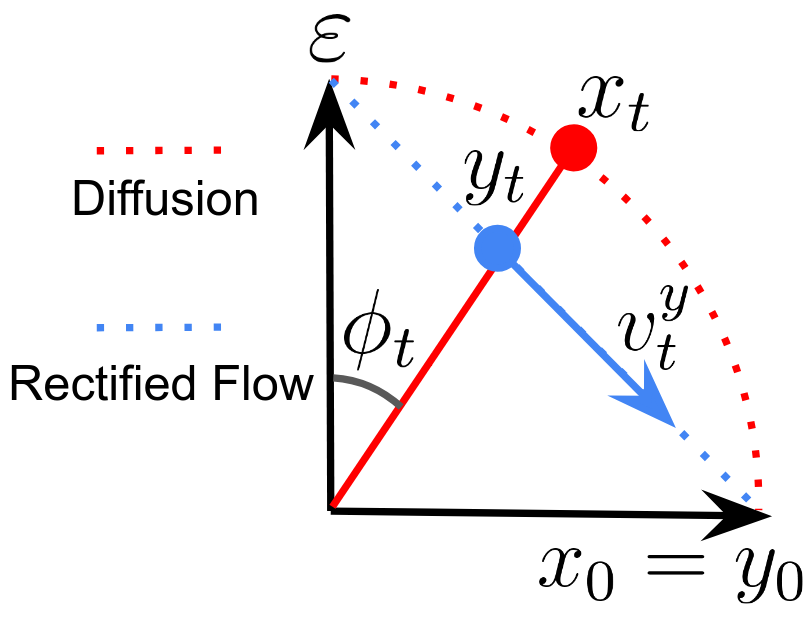}
    \caption{Demonstration of the conjugate velocity prediction: relationship of $v$-prediction for diffusion and rectified flow.
    }
    \label{fig:conjugate}
\end{figure}
with the sample $x_t$ being diffused along the diffusion trajectory according to the schedule defined as Eq.~\eqref{eq:ddpm_schedule}. The model is parameterized to predict velocity $v_t$ on RF trajectory at each timestep $t$, with a constant target $(x_0-\varepsilon)$ (we take a reverse here as opposed to standard RF for notation clarity), as visualized in Fig.~\ref{fig:conjugate}. The predicted velocity $v_\theta(x_t, t)=v_t^y$ is the velocity on RF as the conjugate point $y_t$ of sample $x_t$ along the diffusion trajectory. This is practically easier to learn compared to the time-varying velocity as in Eq.~\eqref{eq:diff_v}.

\paragraph{Why Conjugate Prediction Objective?} The rectified flow loss a commonly applied generative modeling objective in StableDiffusion3~\cite{esser2024scaling}, MovieGen~\cite{polyak2025moviegencastmedia}, Hunyuan Video~\cite{kong2024hunyuanvideo}, etc. 
While prior video generation models apply DDPM noise schedule, direct velocity prediction along the diffusion trajectory following Eq.~\eqref{eq:diff_v_loss} has a time-varying target as Eq.~\eqref{eq:diff_v}, which is practically harder to learn compared with the constant velocity objective in Eq.~\eqref{eq:insta_loss}. However, the standard rectified flow [29] target is not along the diffusion trajectory, which requires to model the trajectory with a different noise schedule other than the DDPM one. The conjugate objective generates variance-preserving noise samples along the diffusion trajectory while predicts a constant velocity, through the conjugate relationship between the diffusion trajectory and rectified flow, therefore it is easier to learn.

\paragraph{Inference.}
After training, the reverse diffusion process follows:
\begin{footnotesize}
\begin{align}
     &x_{t-1}\coloneqq\text{Denoise}(x_t, t, \theta) \nonumber\\
     &=(\sqrt{\bar{\alpha}_{t-1}}-\sqrt{1-\bar{\alpha}_{t-1}-\sigma_t^2}\frac{\sqrt{\bar{\alpha}_t}}{\sqrt{1-\bar{\alpha}_t}})\hat{x}_0 + \frac{\sqrt{1-\bar{\alpha}_{t-1}}}{\sqrt{1-\bar{\alpha}_t}}x_t+\sigma_t \varepsilon
     \label{eq:denoise}
\end{align}
\end{footnotesize}
with $\hat{x}_0=\frac{x_t+\sqrt{1-\bar{\alpha}_t}v_\theta(x_t, t)}{\sqrt{\bar{\alpha}_t}+\sqrt{1-\bar{\alpha}_t}}$ as the predicted original samples.
Proofs see Sec.~\ref{sec:math}.

\subsection{Implementation Details}
\label{app:implementation_details}
Both the teacher and student models are trained on internal image and video datasets with text captioning, comprising approximately $O(100\text{M})$ images and $O(1\text{M})$ videos. The teacher model employs standard DDPM settings with 1000 sampling steps: $t\in[1,\dots,1000]$. For inference, the teacher model utilizes DDIM sampling to generate high-quality samples in 50 steps, with $t_n\in[19, 39, \dots, 999]$. After distillation, the student model adopts a default 4-step sampling protocol, as in previous work~\cite{yin2024improved}, using timesteps $[249, 499, 749, 999]$. Additionally, we explore 1-step ($[999]$) and 2-step ($[499, 999]$) generation configurations for the student model in Sec.~\ref{sec:exp_ablation}. Consistency distillation (CD, discussed in Sec.~\ref{sec:cd}) follows a DDIM schedule with $N=50$ steps, as implemented in LCM~\cite{luo2023latent}. For teacher inference, we apply classifier-free guidance (CFG)~\cite{ho2022classifier} augmentation with a weight of $w=7.5$ in CD as specified in Eq.~\eqref{eq:cfg} and $w=3.5$ for the real score network in VSD, The fake score network and distilled student inference do not employ CFG. In the VSD loss, we adhere to the update ratio as 5 for the fake score update over generator update, as suggested in previous work~\cite{yin2024improved}, to ensure training stability. All experiments are conducted with a batch size of 1 per GPU due to the large model size and limited VRAM, utilizing 8 GPUs in parallel for each run. All student models are distilled up to $4\times 10^4$ iterations, with moderate model selection. Video samples are generated with 128 frames at a resolution of $192\times 320$. We set $\beta_\text{CD}=0.5$ and $\beta_\text{FT}=1.0$ to roughly match the magnitude of each loss without more fine-grained balance. This simple strategy is sufficient that the dominance of one loss over others does not appear throughout our experiments, which verifies the robust training of our framework at large scale.

To reduce VRAM occupancy on GPUs, we employ gradient checkpointing and fully sharded data parallel (FSDP)~\cite{zhao2023pytorch}, enabling sharding of model weights and gradients across GPUs in a data-parallel fashion. Additionally, we utilize mixed precision training with the Bfloat16 data type. For fine-tuning with LRMs, we apply gradient accumulation over 7 steps to stabilize training due to the small batch size (=1) used. 

\subsection{Student-Teacher Parameterization}
There are two different ways for student-teacher parameterization: \textbf{homogeneous} and \textbf{heterogeneous}. 

For homogeneous student-teacher parameterization, the networks of student and teacher both follow the same variable prediction, \emph{i.e.}, $v$-prediction in our setting, with a transformation:
\begin{align}
    x_\theta(x_t,t)&=\frac{x_t+\sqrt{1-\bar{\alpha}_t}v_\theta(x_t, t)}{\sqrt{\bar{\alpha}_t}+\sqrt{1-\bar{\alpha}_t}}
    \label{eq:x_v_trans}
\end{align}
which is proved in Sec.~\ref{sec:math}.
The student model $v_\theta$ will be initialized from teacher model $v_{\theta'}$ at the beginning of distillation. 

For heterogeneous student-teacher parameterization, the student network can directly predict $x_\theta$ without leveraging Eq.~\eqref{eq:x_v_trans}.
For the best usage of teacher model in student distillation, we adopt the homogeneous parameterization by default.

\subsection{Derivations}
\label{sec:math}

\subsubsection{Proof of Eq.~\eqref{eq:denoise}}
We start from the forward diffusion process of DDPM~\cite{ho2020denoising}.
The distribution of one-step diffusion process $q(x_t|x_{t-1})=\mathcal{N}(x_t; \sqrt{\alpha_t}x_{t-1}, (1-\alpha_t)\mathbf{I})$ can be equivalently written as:
\begin{align}x_t=\sqrt{\alpha_t}x_{t-1}+\sqrt{1-\alpha_t}\varepsilon, \quad \varepsilon\sim\mathcal{N}(0, \mathbf{I})
    \label{eq:x_t}
\end{align}
with $t\in[T]$.

By chain rule, we have
\begin{align}x_t=\sqrt{\bar{\alpha}_t}x_0+\sqrt{1-\bar{\alpha}_t}\varepsilon
    \label{eq:xt_x0}
\end{align}
with $\bar{\alpha}_t=\Pi_{i=1}^t \alpha_i$. Equivalently, we have $x_t\sim q(x_t|x_0)=\mathcal{N}(x_t;\sqrt{\bar{\alpha}_t}x_0, (1-\bar{\alpha}_t)\mathbf{I})$.
This equation is also used to predict:
\begin{align}
\hat{x}_0=\frac{1}{\sqrt{\bar{\alpha}_t}}x_t-\frac{\sqrt{1-\bar{\alpha}_t}}{\sqrt{\bar{\alpha}_t}}\epsilon_\theta
\label{eq:tweedie}
\end{align}
which is called the Tweedie's formula. $\epsilon_\theta$ is the approximated prediction of $\varepsilon$ with a parameterized model by $\theta$.

Proof of the denoising function Eq.~\eqref{eq:denoise} in reverse diffusion process is as follows:
\begin{align*}
    &x_{t-1}=\sqrt{\bar{\alpha}_{t-1}}\hat{x}_0 + \sqrt{1-\bar{\alpha}_{t-1}-\sigma_t^2}\epsilon_\theta+\sigma_t \varepsilon \\
    &=\sqrt{\bar{\alpha}_{t-1}}\hat{x}_0 + \sqrt{1-\bar{\alpha}_{t-1}-\sigma_t^2}(\frac{1}{\sqrt{1-\bar{\alpha}_t}}x_t \\
    &- \frac{\sqrt{\bar{\alpha}_t}}{\sqrt{1-\bar{\alpha}_t}}\hat{x}_0)+\sigma_t \varepsilon\\
    &=(\sqrt{\bar{\alpha}_{t-1}}-\sqrt{1-\bar{\alpha}_{t-1}-\sigma_t^2}\frac{\sqrt{\bar{\alpha}_t}}{\sqrt{1-\bar{\alpha}_t}})\hat{x}_0 \\
    &+ \frac{\sqrt{1-\bar{\alpha}_{t-1}}}{\sqrt{1-\bar{\alpha}_t}}x_t+\sigma_t \varepsilon
\end{align*}
with the first equation follows the posterior sampling in DDIM paper~\cite{song2021denoising}. The second is to plug in the Tweedie's formula. We have the variance term $\sigma_t^2=\frac{(1-\alpha_t)(1-\bar{\alpha}_{t-1})}{1-\bar{\alpha}_t}$.

\subsubsection{Proof of Eq.~\eqref{eq:x_v_trans}}
Following the Instaflow objective as Eq.~\eqref{eq:insta_loss},
the network directly predicts $v_\theta$, to approximate the target velocity $\tilde{v}^y$ along the rectified flow (RF) trajectory, as the difference of the clean sample and Gaussian noise:
\begin{align}
    v_\theta\approx \tilde{v}^y = x_0 - \varepsilon
    \label{eq:v_predict}
\end{align}
Since the RF sample $y_t$ is a scaled version of diffusion sample $x_t$ as:
\begin{align}
y_t&=\frac{x_t}{\sqrt{\bar{\alpha}_t}+\sqrt{1-\bar{\alpha}_t}}=\gamma_t x_0 + (1-\gamma_t)\varepsilon, \label{eq:y_x}\\\gamma_t&=\frac{\sqrt{\bar{\alpha}_t}}{\sqrt{\bar{\alpha}_t}+\sqrt{1-\bar{\alpha}_t}},
\label{eq:rf_formula}
\end{align}
which satisfies $y_0=x_0$.

Given the velocity prediction $v_\theta$, we can derive the prediction of original sample $x_\theta$ as following, by replacing $x_0$ with prediction $x_\theta$ in Eq.~\eqref{eq:v_predict} and \eqref{eq:y_x}:
\begin{align}
    \gamma_t x_\theta &= y_t - (1-\gamma_t)(x_\theta - v_\theta^y)\\
    x_\theta &= y_t + (1-\gamma_t)v_\theta = y_t + \frac{\sqrt{1-\bar{\alpha}_t}}{\sqrt{\bar{\alpha}_t}+\sqrt{1-\bar{\alpha}_t}}v_\theta
\end{align}
which concludes the proof.

\subsection{Inference Time Analysis}
Here, we provide a detailed analysis of the inference time experimental results presented in the main paper, as shown in Sec.~\ref{sec:distill_compare} Tab.~\ref{tab:inference_time_percentage}. Absolute time costs are not reported, as they are influenced by hardware-specific factors and inference configurations such as batch size and the number of GPUs used. Instead, relative time consumption is emphasized as a more reliable metric for cross-configuration comparisons.

Notably, the relationship between diffusion sampling time and the number of sampling steps is not strictly linear. For example, the first diffusion sampling step accounts for only $0.33\%$ of the total inference time, making it approximately 6.2 times faster than subsequent steps. This discrepancy is likely due to the faster inference process for initial Gaussian noise inputs or the relatively low hardware cache occupation during early inference stages.

Furthermore, the difference between the total inference time and the diffusion sampling time includes additional costs for text preprocessing and encoding, as well as decoding from the latent space back to the original pixel space. These processes collectively account for approximately $7\%$ of the total inference time.

\section{Reward Model Fine-Tuning}
\label{app_sec:rm_finetune}

\subsection{Evidence of Fine-tuning Effect}
\label{app_sec:evidence_finetune}
\begin{figure}[htbp]
    \centering
\includegraphics[width=\columnwidth]{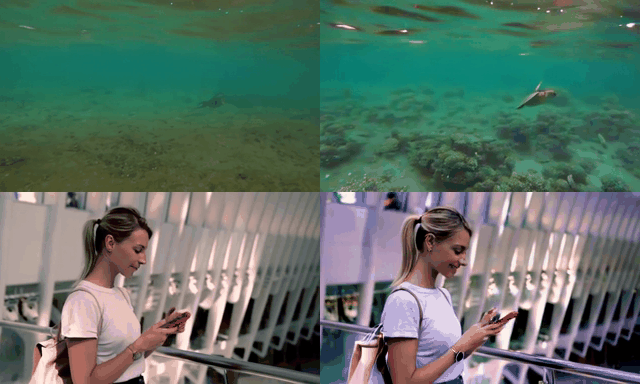}
    \caption{Visualization of samples in training dataset (left) and samples generated with reward tuning using HPSv2 reward (right).
    }
    \label{fig:compare_lrm_hps_x0}
\end{figure}

As shown in Fig.~\ref{fig:compare_lrm_hps_x0}, the samples generated after reward model tuning can have a substantial difference from the original training samples in dataset (left in the figure), for aspects of aesthetic quality, lighting condition, colors, etc.

\subsection{Direct Reward Gradient}
In this section, we discuss in details why the direct reward gradient methods like ReFL~\cite{xu2024imagereward} and DRaFT~\cite{clark2023directly}, cannot fit into the memory efficiently.

Take the HPSv2~\cite{wu2023human} model as an example. It applies fine-tuned version of ViT-H/14 variant of CLIP model, which contains 32 image transformer layers and 24 text transformer layers, each with 16 heads. This constitutes a total of 633 million parameters. Even with FP16 data type, the model weights will occupy 1.25 GB memory. Even for a batch size of 1, the input video tensor of size $(128, 3, 192, 320)$ occupies about 6 GB memory for forward inference only. Backpropagation through the model will drastically increases the memory cost due to gradients storage. Moreover, the memory occupancy roughly scales linearly with the batch size, making it hard to scale up. PickScore~\cite{kirstain2023pick} with CLIP-H model has the similar memory cost in practice. Comparison of parameter numbers and memory costs for reward models and LRMs is shown in Tab.~\ref{tab:model_comparison_memory}. If we take sub-sampling in videos to extract frames for reward optimization, the backward memory (VRAM) cost for different number of frames $H$ is shown in Tab.~\ref{tab:model_comparison_memory_batch}. It indicates that even with frame sub-sampling, the memory cost can still be too large to afford in video model training.

\begin{table}[htbp]
\centering
\caption{Backward memory (VRAM) costs for HPSv2, PickScore reward models with different numbers ($H$) of image ($192\times320$) frames. }
\resizebox{0.5\columnwidth}{!}{
\begin{tabular}{@{}l|cccc@{}}
\toprule
Model                     &  $H=12$   &  $H=24$   &  $H=64$ & $H=128$   \\ \midrule
HPSv2/PickScore    & 12.373 GB        & 20.577 GB  & 48.413 GB              & $>$90 GB                             \\ \bottomrule
\end{tabular}
}
\label{tab:model_comparison_memory_batch}
\end{table}

Given the diffusion modeling in latent space, direct reward gradient methods will also need to backpropagate the gradients from reward model through the large pretrained decoder, this further increases the burden on memory usage.

\subsection{Latent Reward Model For Different Reward Types}
\label{app_sec:lrm_reward_types}

\begin{table*}[htbp]
\centering
\caption{Summary of latent reward models for different pixel-space reward metrics.}
\resizebox{\textwidth}{!}{
\begin{tabular}{l|l|p{6cm}|p{4cm}}
\hline
\textbf{Reward Type} & \textbf{LRM Function} & \textbf{Architecture} & \textbf{Examples} \\ \hline
Image Reward & $\mathcal{R}^l_\phi(x):\mathcal{X} \rightarrow \mathbb{R}$ & 2D CNN backbone & LAION aesthetic~\cite{schuhmann2022laion}, JPEG compressibility~\cite{black2023training} \\ \hline
Text-Image Reward & $\mathcal{R}^l_\phi(x, c):\mathcal{X} \times \mathcal{C} \rightarrow \mathbb{R}$ & 2D CNN + text embedding, cross-attention & HPS~\cite{wu2023human_a, wu2023human}, ImageReward~\cite{xu2024imagereward}, PickScore~\cite{kirstain2023pick} \\ \hline
Video Reward & $\mathcal{R}^l_\phi(\boldsymbol{x}):\mathcal{X}^H \rightarrow \mathbb{R}$ & 2D CNN with average frame reward, or 3D CNN backbone & VBench quality scores~(subject consistency, motion smoothness, etc) \\ \hline
Text-Video Reward & $\mathcal{R}^l_\phi(\boldsymbol{x}, c):\mathcal{X}^H \times \mathcal{C} \rightarrow \mathbb{R}$ & 2D CNN with average frame reward, or 3D CNN backbone, + text embedding, cross-attention & ViCLIP~\cite{wang2023internvid}, VideoScore~\cite{he2024videoscore}, InternVideo2~\cite{wang2024internvideo2}, VBench semantic scores~(object class, human action, color, etc) \\ \hline
\end{tabular}
}
\label{tab:lrm_types}
\end{table*}

The proposed latent reward model method is compatible with any type of reward metrics as introduced previously, regardless of its differentiability and input formats. Here we consider several types of commonly used reward metrics: image reward, text-image reward, video reward and text-video reward. For each category, we provide examples and explain how LRM, with its diverse architectures, supports these metrics. A summary of this compatibility is provided in Tab.~\ref{tab:lrm_types}, with further details outlined below:

\begin{itemize}
    \item Image reward: $\mathcal{I}\rightarrow \mathbb{R}$.

    The LRM is $\mathcal{R}^l_\phi(x):\mathcal{X}\rightarrow \mathbb{R}, x=\text{Encode}(i),i\in\mathcal{I}$. It has the image backbone as a 2D convolutional neural network (CNN).

    Examples include LAION aesthetic quality~\cite{schuhmann2022laion}, JPEG compressibility~\cite{black2023training}.
    
    \item Text-image reward: $\mathcal{C}\times\mathcal{I}\rightarrow \mathbb{R}$. 

    The LRM is $\mathcal{R}^l_\phi(x, c):\mathcal{X}\times\mathcal{C}\rightarrow \mathbb{R}, x=\text{Encode}(i),i\in\mathcal{I}$. It has the image backbone as a 2D CNN and text embedding $e_c$ as inputs, with a cross-attention module for mixing image features $e_x$ and text features $e_c$: $\text{Softmax}(\mathbf{Q}(e_x) \cdot \mathbf{K}(e_c)^\top)\cdot \mathbf{V}(e_c)$.
    
    Examples include human preference score (HPS)~\cite{wu2023human_a, wu2023human}, ImageReward~\cite{xu2024imagereward}, PickScore~\cite{kirstain2023pick}.
    
    \item Video reward: $\mathcal{I}^H\rightarrow \mathbb{R}$ where $H$ is the number of frames in each video. 

    The LRM can be either (1). $\mathcal{R}^l_\phi(x):\mathcal{X}\rightarrow \mathbb{R}, x=\text{Encode}(i),i\in\mathcal{I}$ using a 2D CNN image backbone with average frame reward $\frac{1}{H}\sum_{k=1}^H\mathcal{R}^l_\phi(x_k)$ as video reward or (2). $\mathcal{R}^l_\phi(x_1, \dots, x_H):\mathcal{X}^H\rightarrow \mathbb{R}$ using a 3D CNN as video backbone.
    
    Examples include 7 quality scores in VBench (subject consistency, background consistency, motion smoothness, etc).
    
    \item Text-video reward: $\mathcal{C}\times\mathcal{I}^H\rightarrow \mathbb{R}$. 

    The LRM can be either (1). $\mathcal{R}^l_\phi(x, c):\mathcal{X}\times\mathcal{C}\rightarrow \mathbb{R}$ using a 2D CNN image backbone with average frame reward $\frac{1}{H}\sum_{k=1}^H\mathcal{R}^l_\phi(x_k, c)$ as video reward or (2). $\mathcal{R}^l_\phi(x_1, \dots, x_H, c):\mathcal{X}^H\times\mathcal{C}\rightarrow \mathbb{R}$ using a 3D CNN as video backbone, with additional text embedding $e_c$ as inputs, and cross-attention for mixing image features $e_x$ and text features $e_c$: $\text{Softmax}(\mathbf{Q}(e_x) \cdot \mathbf{K}(e_c)^\top)\cdot \mathbf{V}(e_c)$.

    Examples include  ViCLIP~\cite{wang2023internvid}, VideoScore~\cite{he2024videoscore}, InternVideo2~\cite{wang2024internvideo2} and 9 semantic score metrics in VBench (object class, human action, color, etc).
\end{itemize}

\paragraph{Architecture Details.}
The image only LRM $\mathcal{R}^l_\phi(x)$ has architecture detailed in Tab.~\ref{tab:conv_only_reward_model_architecture}.
The text-image LRM $\mathcal{R}^l_\phi(x,c)$ has architecture detailed in Tab.~\ref{tab:cross_attention_reward_model_architecture}. For video LRM and text-video LRM, we apply the same architectures with frame averaging in our experiments.

\paragraph{Discussions.}
The latent reward model can be utilized in two ways: it can either be pretrained or trained concurrently with the student model during fine-tuning, as demonstrated in our experiments. Furthermore, this approach can also be extended to fine-tune the teacher model. Alternatively, one could bypass the reward model in pixel space entirely and directly employ a latent reward model from the outset. However, we argue that such an approach is likely to be limited to specific fixed latent spaces and may lack generalizability across models. This is because pretrained encoder-decoder models can vary significantly and often do not share a unified latent space, particularly in existing image and video models.

\begin{table*}[htbp]
\centering
\caption{Architecture of the image latent reward model}
\resizebox{\textwidth}{!}{
\begin{tabular}{@{}lccccccc@{}}
\toprule
Layer                       & Input Shape           & Output Shape          & Kernel Size & Stride & Padding & Number of Parameters \\ \midrule
\textbf{Input}              & (batch, C, H, W)   &                       &            &        &         &                       \\
\textbf{Conv2d + GroupNorm + SiLU} & (batch, C, H, W) & (batch, 128, 6, 10)  & 4x4        & 4      & 1       & 24,704                \\
\textbf{Conv2d + GroupNorm + SiLU} & (batch, 128, 6, 10) & (batch, 128, 3, 5)   & 3x3        & 2      & 1       & 147,584               \\
\textbf{AdaptiveAvgPool2d}  & (batch, 128, 3, 5)    & (batch, 128, 1, 1)    & -          & -      & -       & 0                     \\
\textbf{Conv2d}             & (batch, 128, 1, 1)    & (batch, 128, 1, 1)    & 1x1        & 1      & 0       & 16,512                \\
\textbf{Flatten}            & (batch, 128, 1, 1)    & (batch, 128)          & -          & -      & -       & 0                     \\
\textbf{Linear}             & (batch, 128)          & (batch, 1)            & -          & -      & -       & 129                   \\ \midrule
\textbf{Total Parameters}   &                       &                       &            &        &         & 189,441               \\ \bottomrule
\end{tabular}
}
\label{tab:conv_only_reward_model_architecture}
\end{table*}

\begin{table*}[htbp]
\centering
\caption{Architecture of the text-image latent reward model}
\resizebox{\textwidth}{!}{
\begin{tabular}{@{}lccccccc@{}}
\toprule
Layer                                & Input Shape            & Output Shape           & Kernel Size / Projection & Stride & Padding & Number of Parameters \\ \midrule
\textbf{Input Image}                 & (batch, C, H, W)    &                        &                          &        &         &                       \\
\textbf{Conv2d + GroupNorm + SiLU}   & (batch, C, H, W)    & (batch, 128, 6, 10)    & 4x4                      & 4      & 1       & 24,704                \\
\textbf{Conv2d + GroupNorm + SiLU}   & (batch, 128, 6, 10)    & (batch, 128, 3, 5)     & 3x3                      & 2      & 1       & 147,584               \\
\textbf{AdaptiveAvgPool2d}           & (batch, 128, 3, 5)     & (batch, 128, 1, 1)     & -                        & -      & -       & 0                     \\
\textbf{Conv2d}                      & (batch, 128, 1, 1)     & (batch, 128, 1, 1)     & 1x1                      & 1      & 0       & 16,512                \\
\textbf{Flatten (Image Features)}    & (batch, 128, 1, 1)     & (batch, 128)           & -                        & -      & -       & 0                     \\ \midrule
\textbf{Input Text}                  & (batch, L, D)     &                        &                          &        &         &                       \\
\textbf{Text MLP}                    & (batch, L, D)     & (batch, 256, 128)      & -                        & -      & -       & 524,544               \\
\textbf{Average Pooling (Text Features)} & (batch, 256, 128) & (batch, 128)           & -                        & -      & -       & 0                     \\ \midrule
\textbf{Query Projection (Linear)}   & (batch, 128)           & (batch, 128)           & -                        & -      & -       & 16,512                \\
\textbf{Key Projection (Linear)}     & (batch, 128)           & (batch, 128)           & -                        & -      & -       & 16,512                \\
\textbf{Value Projection (Linear)}   & (batch, 128)           & (batch, 128)           & -                        & -      & -       & 16,512                \\
\textbf{Attention Mechanism (Softmax)} & (batch, 1, 1)       & (batch, 1, 1)          & -                        & -      & -       & 0                     \\
\textbf{Final Linear (Output Layer)} & (batch, 128)           & (batch, 1)             & -                        & -      & -       & 129                   \\ \midrule
\textbf{Total Parameters}            &                        &                        &                          &        &         & 763,009               \\ \bottomrule
\end{tabular}
}
\label{tab:cross_attention_reward_model_architecture}
\vspace{-.4cm}
\end{table*}

\subsection{Latent Reward Model Training}

\begin{figure}[htbp]
    \centering
\includegraphics[width=0.49\columnwidth]{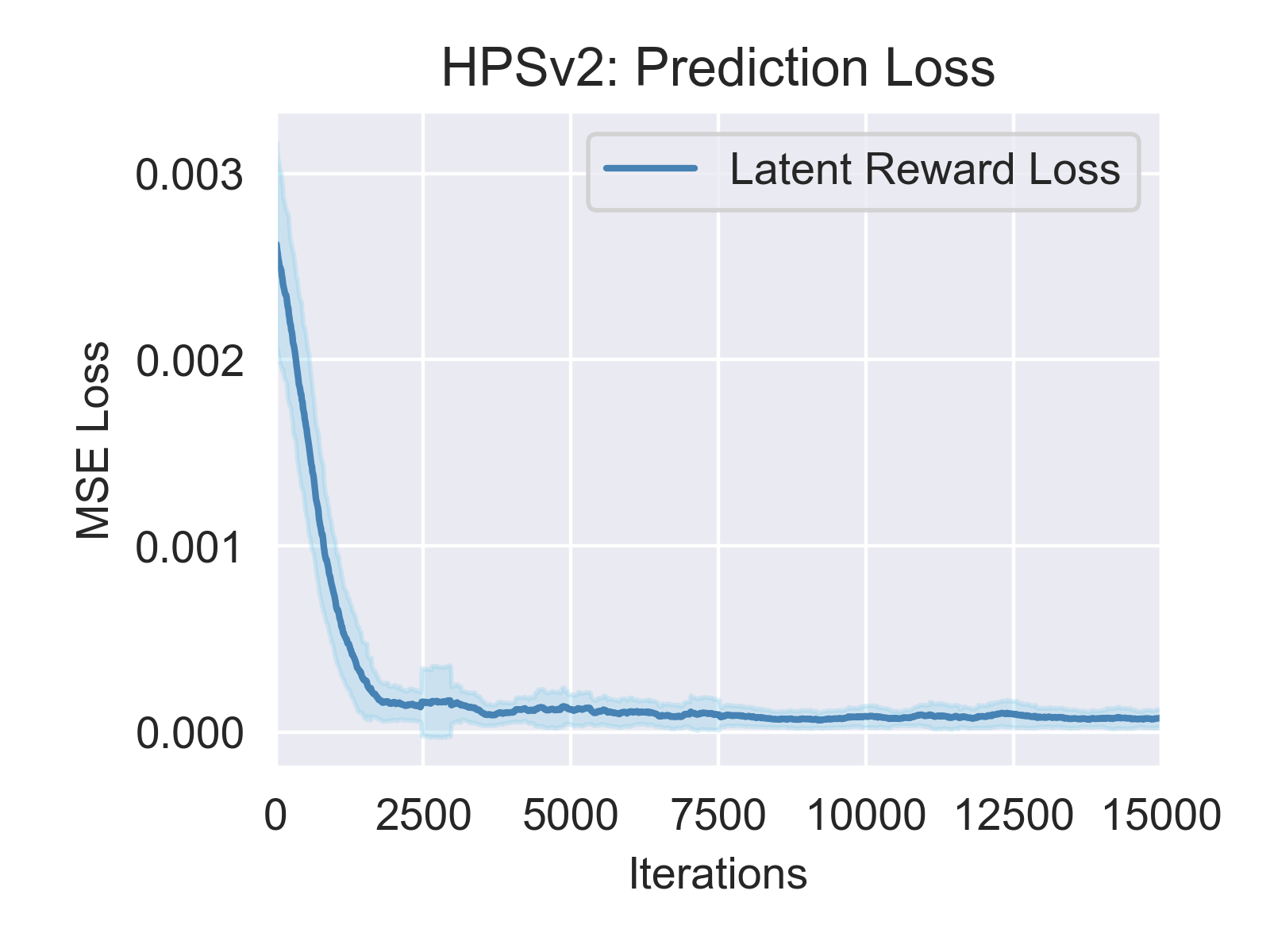}
\includegraphics[width=0.49\columnwidth]{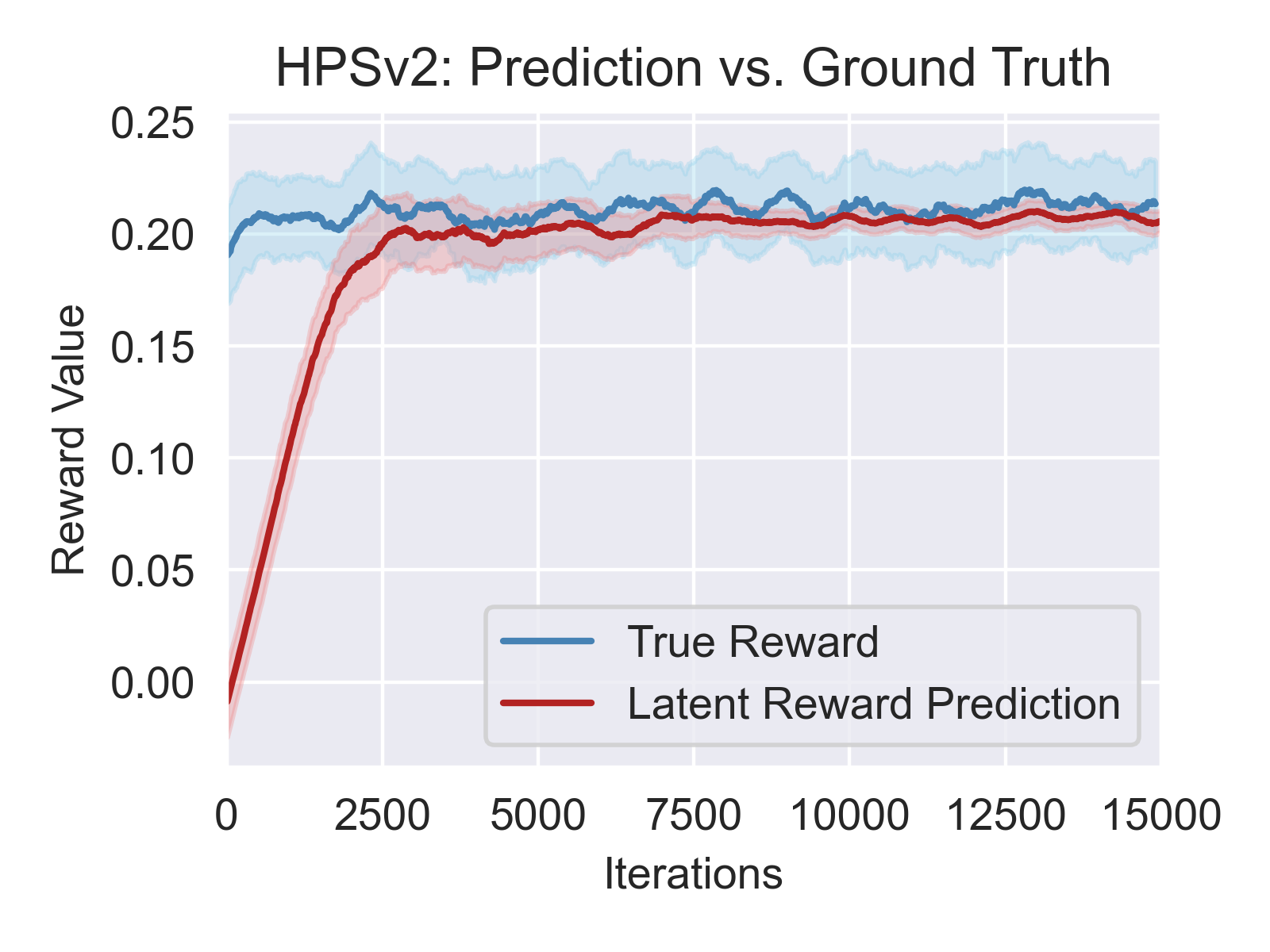}
    \caption{The learning process of LRM with HPSv2 reward.
    }
    \label{fig:lrm_hps_detail}
\end{figure}

\begin{figure}[htbp]
    \centering
\includegraphics[width=0.49\columnwidth]{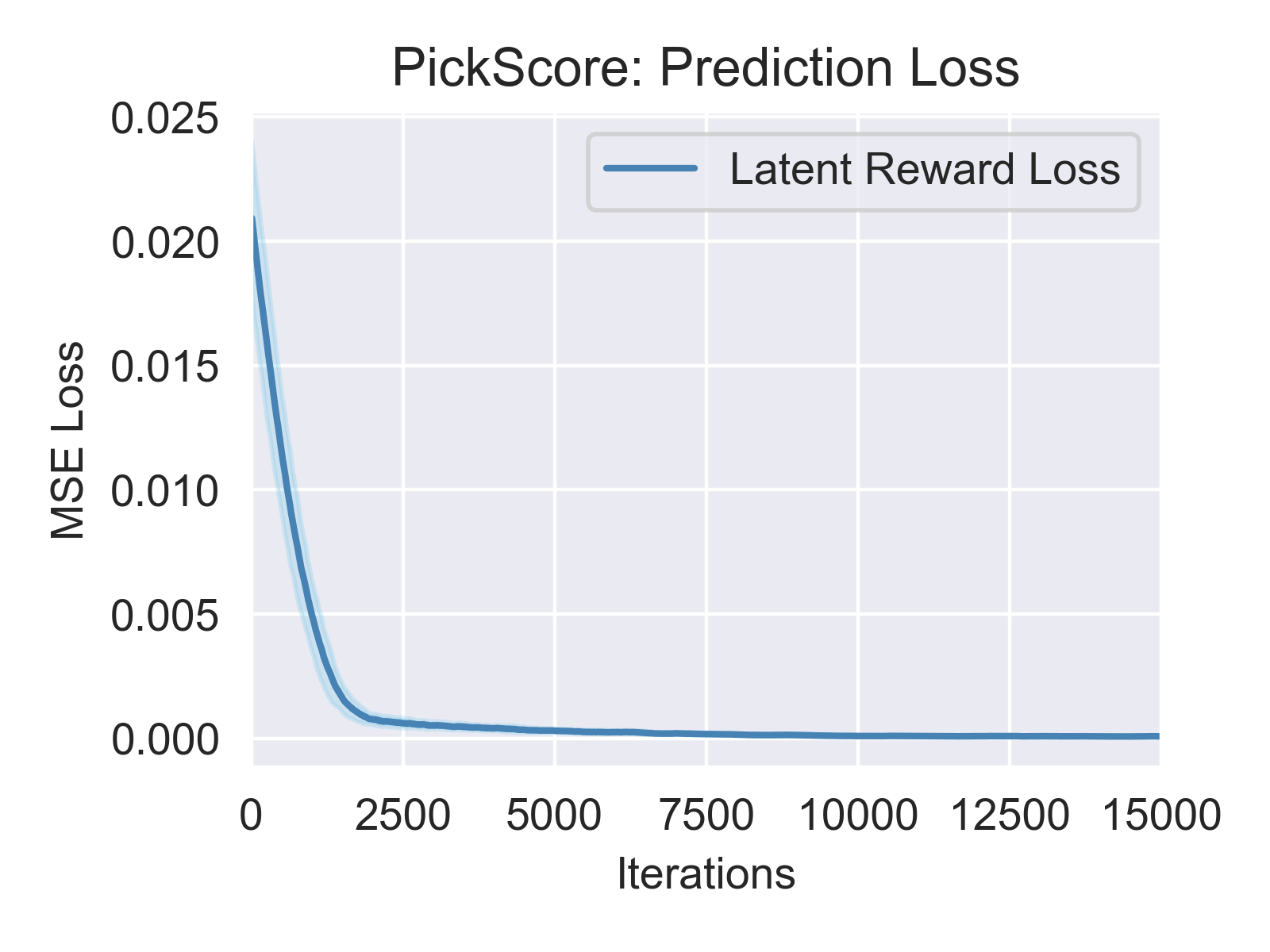}
\includegraphics[width=0.49\columnwidth]{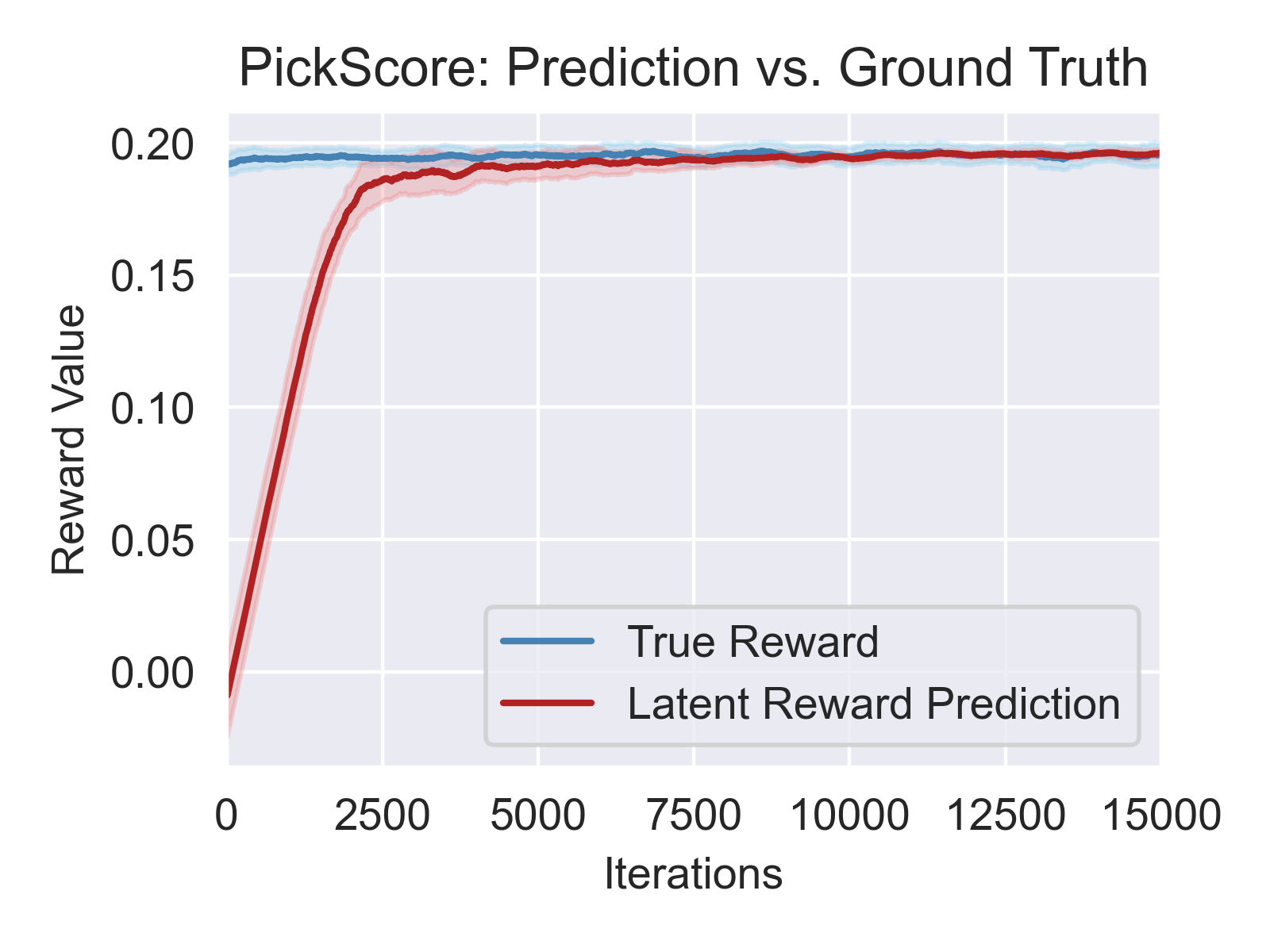}
    \caption{The learning process of LRM with PickScore reward.
    }
    \label{fig:lrm_pick_detail}
\end{figure}

Fig.~\ref{fig:lrm_hps_detail} and Fig.~\ref{fig:lrm_pick_detail} show the learning curves of latent reward models (LRMs) with two original pixel-space rewards HPSv2 and PickScore, respectively, during the distillation process. The loss for training is VSD+LRM. Left figure displays the MSE loss for LRM prediction against the ground-truth pixel-space reward value. Right figure displays the LRM predicted reward values $\mathcal{R}_\phi^l(x_0, c)$ and ground truth reward values $\mathcal{R}(x_0, c)$ on training samples from the dataset $x_0\sim\mathcal{X}$. This demonstrates that the LRM achieves rapid convergence within 2000–3000 training iterations, even when operating in a significantly lower-dimensional latent space.  The small approximation errors ensure the effectiveness of fine-tuning with learned LRM.

\subsection{Latent Reward Model Fine-tuning}
\begin{figure}[htbp]
    \centering
\includegraphics[width=0.99\columnwidth]{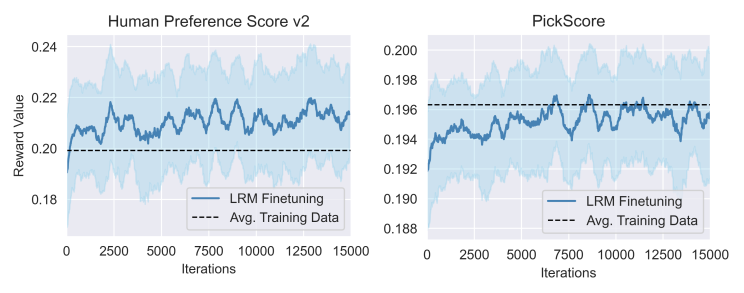}
    \caption{Latent reward model fine-tuning process under reward metrics HPSv2 and PickScore.
    }
    \label{fig:hpsv2}
\end{figure}

Fig.~\ref{fig:hpsv2} displays the predicted reward values $\mathcal{R}^l_\phi(\hat{x}_0, c)$ with LRM for generated samples ($\hat{x}_0\sim\mathcal{X}'$, by Eq.~\eqref{eq:x_v_trans}) during the distillation process with VSD+LRM loss, for two reward metrics HPSv2 and PickScore, respectively. The horizontal dashed lines are the average reward values of the samples in training dataset. For HPSv2, the reward values of generated samples surpass the training data quickly with the LRM fine-tuning. For PickScore, the reward values of generated samples also gradually increase to be close to the training data.

\subsection{Denoising Diffusion Policy Optimization}
\label{sec:ddpo}
Denoising Diffusion Policy Optimization (DDPO)~\cite{black2023training} serves as the baseline for comparison with our proposed LRM method.
DDPO applies the REINFORCE algorithm to optimize the diffusion model by treating the diffusion process as a MDP. It requires to estimate the log-probabilities for the sample at all diffusion steps, which are then summed over and weighted by the final reward as the optimization objective. Considering memory constraints, our method is suited for few-step sampling models or configurations with gradient truncation along the diffusion trajectory. In our experiments, memory limitations prevent log-probability estimation over more than 2 steps. Therefore, we employ a truncation step of 2 for the student model (\emph{i.e.}, log-probability estimation at timesteps [249, 499]). This truncation approach has been validated in previous work~\cite{clark2023directly, ren2024diffusion}. We apply $\text{DDPO}_\text{SF}$ for online policy gradient in our experiments.

By applying the REINFORCE algorithm on denoising process of diffusion models, the DDPO$_\text{SF}$ algorithm follows the score function policy gradient:
\begin{align}
   \nabla_\theta \mathcal{J}=\mathbb{E}[\sum_{t=1}^T \nabla_\theta \log p_\theta (x_{t-1}|x_t, c)R(x_0, c)]
\end{align}
This is the online version for gradient estimation, which requires to sample $x_{t-1}$ as well as calculating the probabilities $p_\theta (x_{t-1}|x_t, c)$ along the sampling process at the same time, such that the model parameters $\theta$ remain the same for sampling and probability estimation. The update will only take one step to preserve the online estimation property. Original paper~\cite{black2023training} also proposes another version for offline policy gradient estimation with importance sampling to allow multi-step updates. As log-probability $\log p_\theta (x_{t-1}|x_t, c)$ needs to be estimated during the sampling process, we cannot take sampling process as Eq.~\eqref{eq:denoise}, but estimating the posterior mean $\mu_\theta$ and standard deviation $\sigma$ instead:
\begin{align}
    \mu_\theta(x_{t-1};x_t)&= \frac{(1 - \alpha_t)\sqrt{\bar{\alpha}_t}}{1-\bar{\alpha}_t}x_\theta  + \frac{\sqrt{\alpha}_t(1-\bar{\alpha}_{t-1})}{1-\bar{\alpha}_t} x_t\nonumber\\
    \sigma_t &= \sqrt{(1 - \alpha_t)\frac{1-\bar{\alpha}_{t-1}}{1-\bar{\alpha}_t}}
    \label{eq:mean_std}
\end{align}
with $x_\theta$ following Eq.~\eqref{eq:x_v_trans}. $x_{t-1}$ will be sampled from $\mathcal{N}(\mu_\theta(x_{t-1};x_t), \sigma_t)$, with log-probability of the sample as:
\begin{align}
    \log p_\theta(x_{t-1}|\mu_\theta, \sigma, c)=-\frac{1}{2} \left( \frac{(x_{t-1} - \mu_\theta)^2}{\sigma^2} + \log(2\pi \sigma^2)\right)
    \label{eq:log_prob}
\end{align}
The practical procedure of DDPO$_\text{SF}$ is outlined in Alg.~\ref{alg:ddpo}. Due to VRAM memory constraints, we employ the REINFORCE policy gradient with truncation, allowing gradient tracking for a maximum of $N=2$ steps during training.
Specifically, for a student model with a sampling time sequence $[T, \dots, t_\text{min}]=[999, 749, 499, 249 ]$, the gradient update steps will only take the last two steps $t_n\in\{499, 249\}$, rather than all timesteps. This truncation is used to estimate the log-probabilities of samples at $t_{n-1}$. Here, $\text{Dec}(\cdot)$ represents the pretrained video decoder, while the reward model $R$ operates in the original pixel space. We use $.\text{detach}()$ to indicate a stop-gradient function.

\begin{algorithm}[htbp]
    \caption{DDPO practical procedure}
    \small
    \begin{algorithmic}
        [1] \STATE \textbf{Input:} Distilled student model $G_{\theta}$, dataset $\mathcal{D}=\{(c, i)\}$
        \STATE \textbf{Output:} Fine-tuned student few-step generator $G_\theta$
        \WHILE{\text{train}}
        \STATE \texttt{//Sample from random noise along entire diffusion trajectory}
        \STATE $x_T\leftarrow \varepsilon\sim\mathcal{N}(0, \mathbf{I})$
        \FOR{$t_n\in [T, \dots, t_\text{min}]$}
        \STATE Get posterior Gaussian $(\mu_\theta, \sigma)$ with $v_\theta(x_{t_n}\text{.detach()},t_n)$ \texttt{//Eq.~\eqref{eq:mean_std}}
        \STATE Sample $x_{t_{n-1}}\sim\mathcal{N}(\mu_\theta, \sigma \mathbf{I})$
        \STATE Estimate $\log p_\theta(x_{t_{n-1}}|x_{t_n}, c)$ \texttt{//Eq.~\eqref{eq:log_prob}}
        \ENDFOR
        \STATE Get reward $R=R(\text{Dec}(\hat{x}_0), c)\text{.detach()}$
        \STATE \texttt{//REINFORCE policy gradient with truncation}
        \STATE $\mathcal{L}_{\text{DDPO}_\text{SF}}=-\sum_n^N \log p_\theta(x_{t_{n-1}}\text{.detach()}|x_{t_n}, c) \cdot R$
        \STATE $G_\theta\leftarrow \text{GradientDescent}(\theta, \mathcal{L}_{\text{DDPO}_\text{SF}})$
        \ENDWHILE
    \end{algorithmic}
\label{alg:ddpo}
\end{algorithm}

\begin{figure}[htbp]
    \centering
\includegraphics[width=0.48\columnwidth]{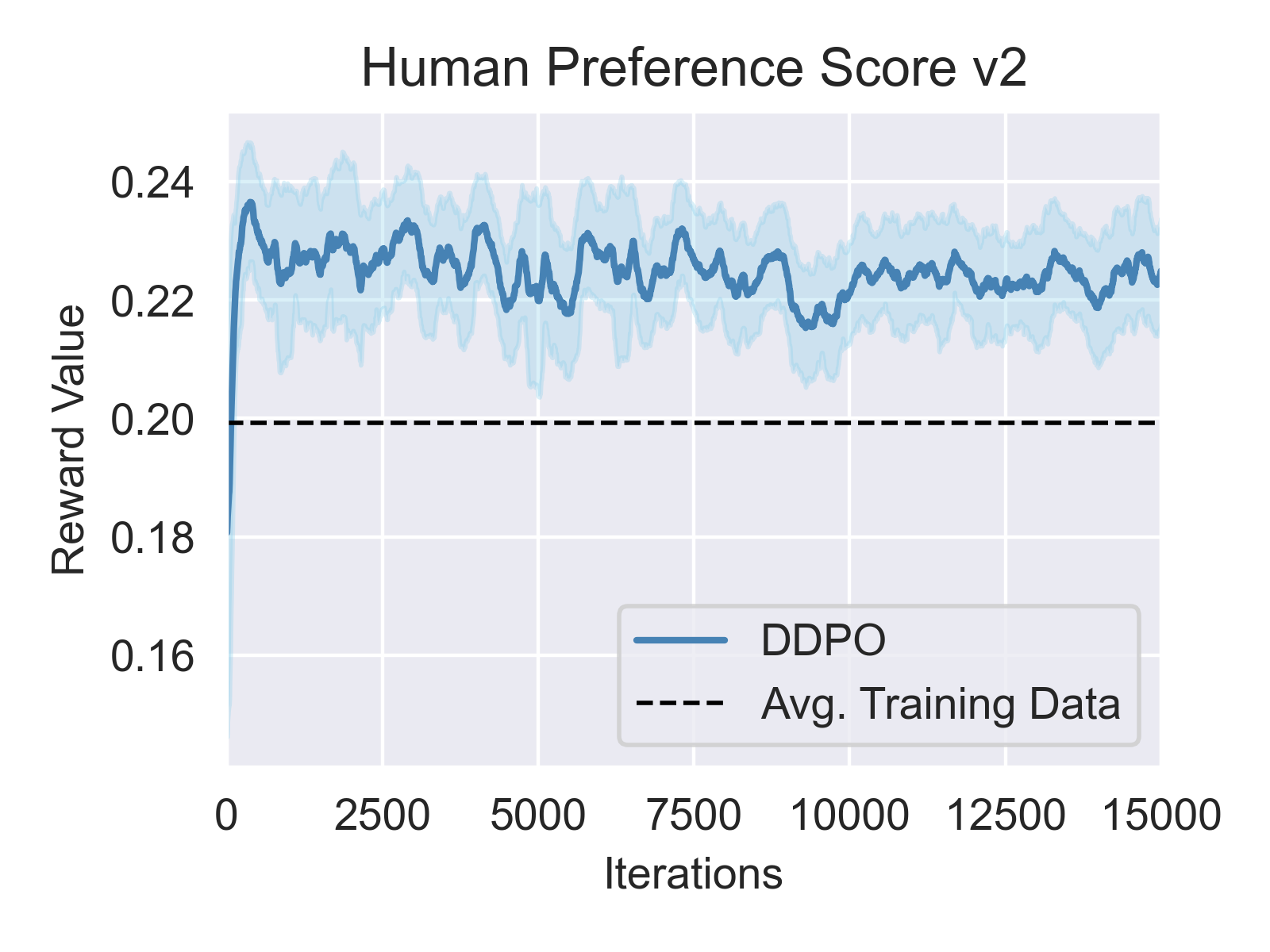}
\includegraphics[width=0.48\columnwidth]{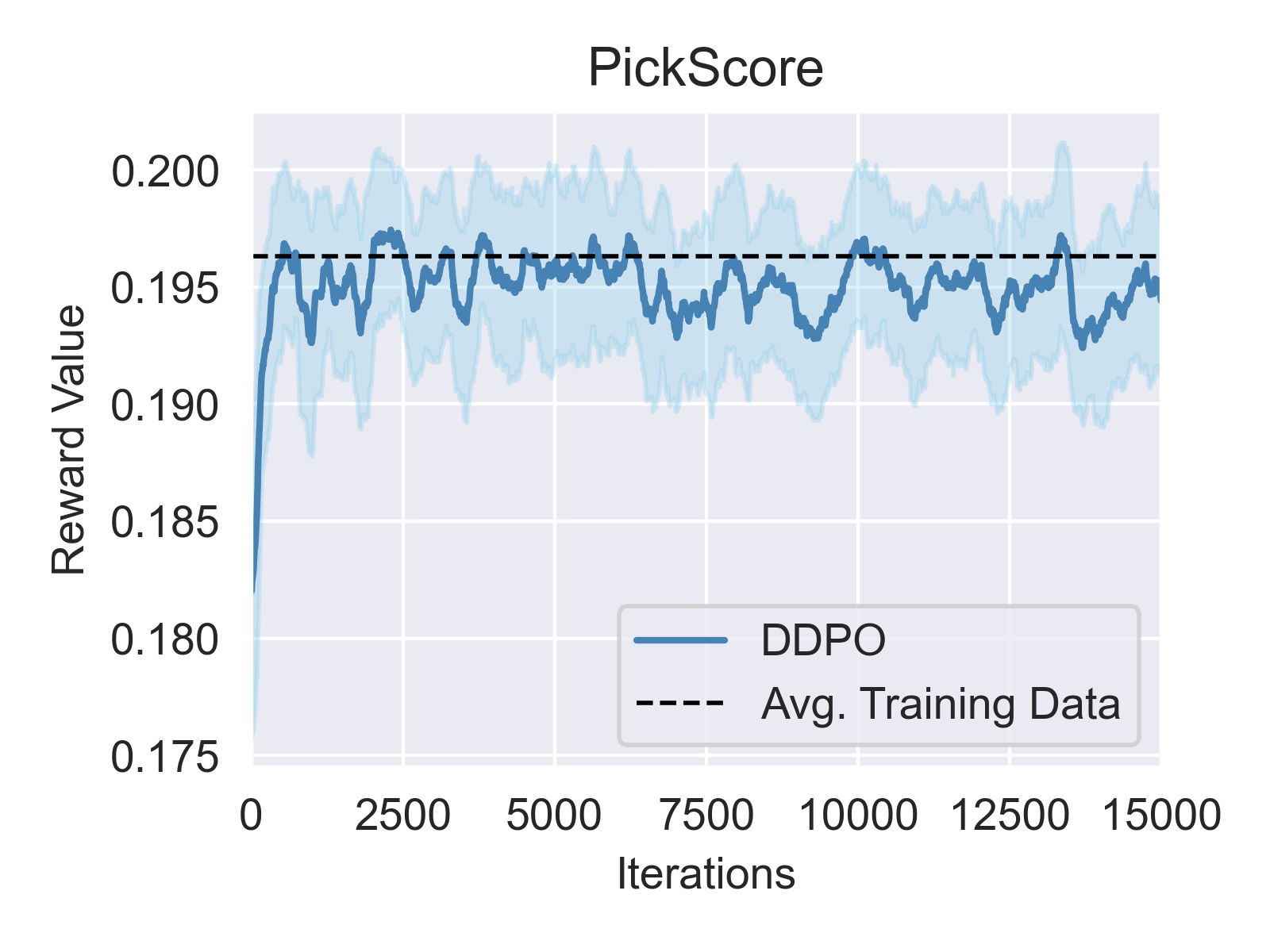}
    \caption{Reward model fine-tuning process with VSD+DDPO under reward models HPSv2 and PickScore.
    }
    \label{fig:ddpo_curves}
\end{figure}

\paragraph{Learning Curves.}
The training process of VSD+DDPO for two reward metrics are shown in Fig.~\ref{fig:ddpo_curves}. The learning curve shows the reward values $\mathcal{R}(x_0,c)$ for generated samples $\hat{x}_0$ through iterative denoising along the full diffusion trajectories, during the fine-tuning process.

The learning curves of DDPO are not directly comparable to those of the LRM methods shown in Fig.~\ref{fig:hpsv2}. This difference arises because DDPO samples across the entire diffusion trajectory to obtain the predicted $\hat{x}_0$ for reward evaluation, whereas LRM performs one-step prediction using $x_\theta=\frac{x_t+\sqrt{1-\bar{\alpha}_t}v^w_\theta(x_t, t)}{\sqrt{\bar{\alpha}_t}+\sqrt{1-\bar{\alpha}_t}}$, as defined in Eq.~\eqref{eq:x_v_trans}. Consequently, the LRM samples tend to be noisier and yield lower rewards during fine-tuning. A fair comparison involves evaluating the rewards of the final generated samples after the model fine-tuning, as presented in Tab.~\ref{tab:reward_method_compare} of main paper.

\chapter{Video World Model\label{ch:video_world_model}}
\begin{center}
\begin{quote}
This section is based on paper ``\textit{Learning World Models for Interactive Video Generation}''~\cite{chen2025learning} written in collaboration with Taiye Chen, Xun Hu and Chi Jin, previously published at NeurIPS 2025.
\end{quote}
\end{center}

\section{Introduction}

\begin{wrapfigure}{tr}{0.5\textwidth}
    \centering
    \includegraphics[width=1.0\linewidth]{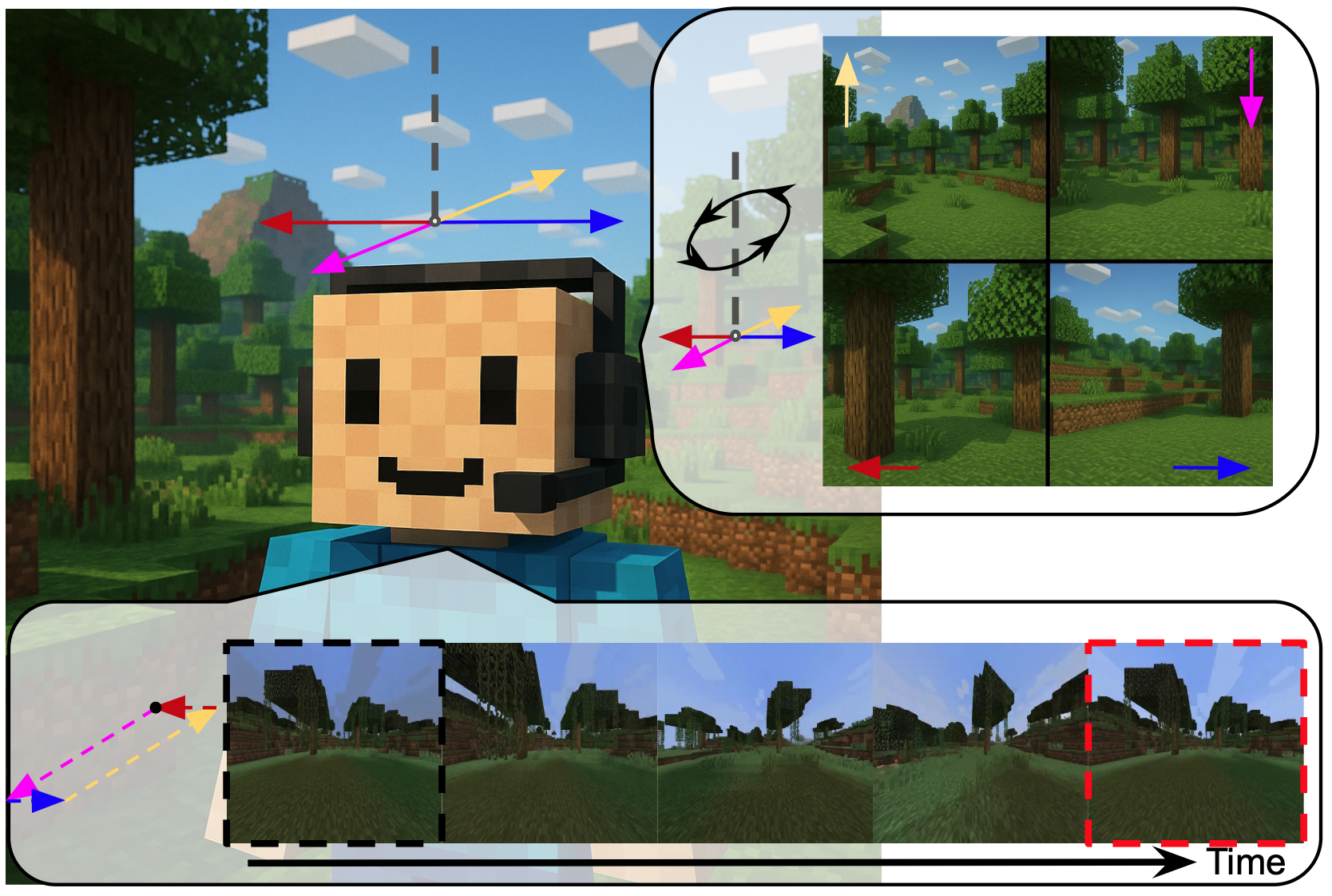}
    \captionsetup{width=0.5\textwidth}
    \caption{A world model possesses memory capabilities and enables faithful long-term future prediction by maintaining awareness of its environment and generating predictions based on the current state and actions. Example is in Minecraft game.}
    \label{fig:teaser}
\end{wrapfigure}

Foundational world models capable of simulating future outcomes based on different actions are crucial for effective planning and decision-making~\cite{watter2015embed, ha2018recurrent, hafner2020mastering}. To achieve this, these models must exhibit both interactivity, allowing for action conditioning, and spatiotemporal consistency over long horizons. While recent advancements in video generation, particularly diffusion models~\cite{sohl2015deep, song2019generative, ho2020denoising, song2021score}, have shown promise, extending them to generate long, interactive, and consistent videos remains a significant challenge~\cite{videoworldsimulators2024, bruce2024genie, valevski2024diffusion}.

Autoregressive approaches~\cite{weissenborn2019scaling, harvey2022flexible, li2024arlon, xie2024progressive}, which generate videos frame by frame or chunk by chunk conditioned on previous outputs, are a natural fit for modeling long temporal dependencies and incorporating interactivity. However, these methods face significant challenges stemming from two fundamental, often coupled, limitations: \textbf{compounding errors} and \textbf{insufficient memory mechanisms}. Compounding errors arise as small inaccuracies in early predictions accumulate over time, leading to significant divergence from plausible future states. Our analysis suggests this may be inherent to current autoregressive paradigms. Insufficient memory mechanisms hinder the models' ability to maintain consistent object identities, spatial layouts, and world states over extended durations, resulting in inconsistent world models. These two issues often exacerbate one another, making long-term consistent generation difficult.

Inspired by the success of large language models (LLMs)~\cite{achiam2023gpt}\cite{touvron2023llama} in handling long sequences, we investigate analogous techniques for video generation. Extending the context window, while potentially alleviating compounding errors to some degree, introduces substantial computational and memory overhead. More critically, we find that unlike LLMs, current video generation models exhibit weaker in-context learning capabilities, making longer context less effective in resolving fundamental consistency issues. Similarly, retrieval-augmented generation (RAG)~\cite{gao2023retrieval}\cite{zhao2024retrieval}, a powerful technique for incorporating external knowledge in LLMs, shows limited benefits in our experiments with video models. Neither static retrieval with heuristic sampling nor dynamic retrieval based on similarity search significantly improved world model consistency.

These findings suggest that implicitly learning world consistency solely from autoregressive prediction on pixel or latent representations is insufficient. We argue that explicit \textbf{global state conditioning} is necessary. Incorporating explicit representations like world maps, object states, or coordinate systems as conditioning information could provide the necessary grounding for generating consistent long-term interactive simulations.

Furthermore, evaluating the specific failure modes of long video generation demands appropriate metrics. Existing metrics often conflate the distinct issues of compounding errors and long-term consistency (memory faithfulness), providing a coupled assessment that obscures the underlying problems. To enable a clearer analysis, we advocate for and introduce a decoupled evaluation strategy by separately quantify the severity of compounding errors and the faithfulness of memory retrieval in long interactive video generation.

Our main contributions are: (1). We systematically decouple and analyze the challenges of compounding errors and insufficient memory in autoregressive video generation for interactive world modeling. (2). We propose video retrieval augmented generation (VRAG) with explicit global state conditioning, which significantly improves long-term spatiotemporal coherence and reduces compounding errors for interactive video generation. (3). We conduct a comprehensive comparison with various long-context methods adapted from LLM techniques, including position interpolation, neural memory augmentation, and historical frame retrieval, demonstrating their limited effectiveness due to the inherent weak in-context learning capabilities of video diffusion models.
This work sheds light on the fundamental obstacles in building consistent, interactive video world models and provides a benchmark and evaluation framework for future research in this direction.


\section{Related Works}

\paragraph{Video Diffusion Models}
Diffusion generative modeling has significantly advanced the fields of image and video generation~\cite{blattmann2023align, harvey2022flexible, 10377444, blattmann2023stable, chen2024videocrafter2, ho2022video, singer2022make, hong2022cogvideo, yang2024cogvideox, wang2023modelscope, ding2024dollar, opensora}. Latent video diffusion models~\cite{blattmann2023stable} operate on video tokens within a latent space derived from a variational auto-encoder (VAE)~\cite{kingma2013auto}, building upon prior work in latent image diffusion models~\cite{rombach2022high}.
The Diffusion Transformer (DiT)~\cite{peebles2023scalable} introduced the Transformer~\cite{vaswani2017attention} backbone as an alternative to the previously prevalent U-Net architecture~\cite{ho2022video, blattmann2023stable, chen2024videocrafter2} in diffusion models.

\paragraph{Long Video Generation}
Autoregressive video generation~\cite{weissenborn2019scaling, harvey2022flexible, li2024arlon, xie2024progressive, hong2024slowfast, wu2024ivideogpt, kim2024fifo, feng2024matrix, magi1, henschel2024streamingt2v} represents a natural approach for long video synthesis by conditioning on preceding frames, drawing inspiration from successes in large language models. This can be implemented using techniques such as masked conditional video diffusion~\cite{voleti2022mcvd, hong2024slowfast} or Diffusion Forcing~\cite{chen2024diffusion}. Diffusion Forcing introduces varying levels of random noise per frame to facilitate autoregressive generation conditioned on frames at inference time. Furthermore, the autoregressive framework naturally supports interactive world simulation by allowing action inputs at each step to influence future predictions.
Nevertheless, compounding errors remain a significant challenge in long video generation, particularly within the autoregressive paradigm, as will be discussed subsequently. 

\paragraph{Interactive Video World Models}
World models~\cite{watter2015embed, ha2018recurrent, hafner2020mastering} are simulation systems designed to predict future trajectories based on the current state and chosen actions. Diffusion-based world models~\cite{ding2024diffusion, alonso2024diffusion, valevski2024diffusion} facilitate the modeling of high-dimensional distributions, enabling high-fidelity prediction of diverse trajectories, even directly in pixel space.
The Sora model~\cite{videoworldsimulators2024} introduced the concept of leveraging video generation models as world simulators. Extending video generation models with interactive capabilities has led to promising applications in diverse domains, including game simulation like Genie~\cite{bruce2024genie}, GameNGen~\cite{valevski2024diffusion}, Oasis~\cite{oasis2024}, Gamegen-x~\cite{che2024gamegen}, The Matrix~\cite{feng2024matrix}, Mineworld~\cite{guo2025mineworld}, GameFactory~\cite{yu2025gamefactory} and so on~\cite{alonso2024diffusion}, autonomous driving~\cite{hu2023gaia}, robotic manipulation~\cite{wu2024ivideogpt, azzolini2025cosmos}, and navigation~\cite{bar2024navigation}. 
While existing work on interactive video world models has made significant engineering advances, there remains a notable gap in systematically analyzing and addressing the fundamental challenges underlying long-term consistency and compounding errors.

A lack of spatiotemporal consistency is a primary bottleneck for developing internal world models using current video generation techniques. One line of research addressing this involves predicting the underlying 3D world structure like Genie2~\cite{parkerholder2024genie2}, Aether~\cite{team2025aether}, Gen3C~\cite{ren2025gen3c} and others~\cite{liu2024reconx, gao2024cat3d, zhen2025tesseract}; however, these approaches often suffer from lower resolution compared to direct video generation due to the complexity of 3D representations, exhibit limited interaction capabilities, and typically operate only within localized regions. Consequently, our work focuses on enhancing the consistency of video-based world models~\cite{valevski2024diffusion, hong2024slowfast, xiao2025worldmem}.
SlowFast-VGen~\cite{hong2024slowfast} employs a dual-speed learning system to progressively trained LoRA modules for memory recall, utilizing semantic actions but offering limited interactivity. Concurrent work~\cite{xiao2025worldmem} explores interactive world simulation through the integration of supplementary memory blocks.








\section{Methodology}
\subsection{Preliminary: Latent Video Diffusion Model}
Video diffusion models have emerged as a powerful framework for video generation. We adopt a latent video diffusion model~\cite{blattmann2023stable} that operates in a compressed latent space rather than pixel space for computational efficiency. 
Specifically, given an input video sequence $\boldsymbol{x} \in \mathbb{R}^{L \times H \times W \times 3}$, we first encode it into a latent representation $\boldsymbol{z} = \mathcal{E}(\boldsymbol{x})$ using a pretrained variational autoencoder (VAE). The forward process gradually adds Gaussian noise to the latent according to a variance schedule $\{\beta_t\}_{t=1}^T$:
\begin{equation}
    q(\boldsymbol{z}_t|\boldsymbol{z}_{t-1}) = \mathcal{N}(\boldsymbol{z}_t; \sqrt{1-\beta_t}\boldsymbol{z}_{t-1}, \beta_t\mathbf{I})
\end{equation}
The model learns to reverse this process by predicting the noise $\boldsymbol{\epsilon}_\theta$ at each step:
\begin{equation}
    \mathcal{L} = \mathbb{E}_{t,\boldsymbol{\epsilon},\boldsymbol{z}}[\|\boldsymbol{\epsilon} - \boldsymbol{\epsilon}_\theta(\boldsymbol{z}_t, t)\|_2^2]
\end{equation}
where $\boldsymbol{z}_t = \sqrt{\bar{\alpha}_t} \boldsymbol{z}_0 + \sqrt{1-\bar{\alpha}_t} \boldsymbol{\epsilon}$ with $\boldsymbol{\epsilon} \sim \mathcal{N}(0,\mathbf{I})$.

At inference time, we can sample new videos by starting from random noise $\boldsymbol{z}_T \sim \mathcal{N}(0,\mathbf{I})$ and iteratively denoising:
\begin{equation}
    \boldsymbol{z}_{t-1} = \frac{1}{\sqrt{\alpha_t}}(\boldsymbol{z}_t - \frac{\beta_t}{\sqrt{1-\bar{\alpha}_t}}\boldsymbol{\epsilon}_\theta(\boldsymbol{z}_t,t)) + \sigma_t\boldsymbol{\epsilon}
\end{equation}
where $\alpha_t = 1-\beta_t$ and $\bar{\alpha}_t = \prod_{s=1}^t \alpha_s$.
The final latent sequence $\boldsymbol{z}_0$ is decoded back to pixel space using the decoder $\mathcal{D}$ to obtain the generated video.

\begin{figure}[htbp]
    \centering
    \includegraphics[width=1\linewidth]{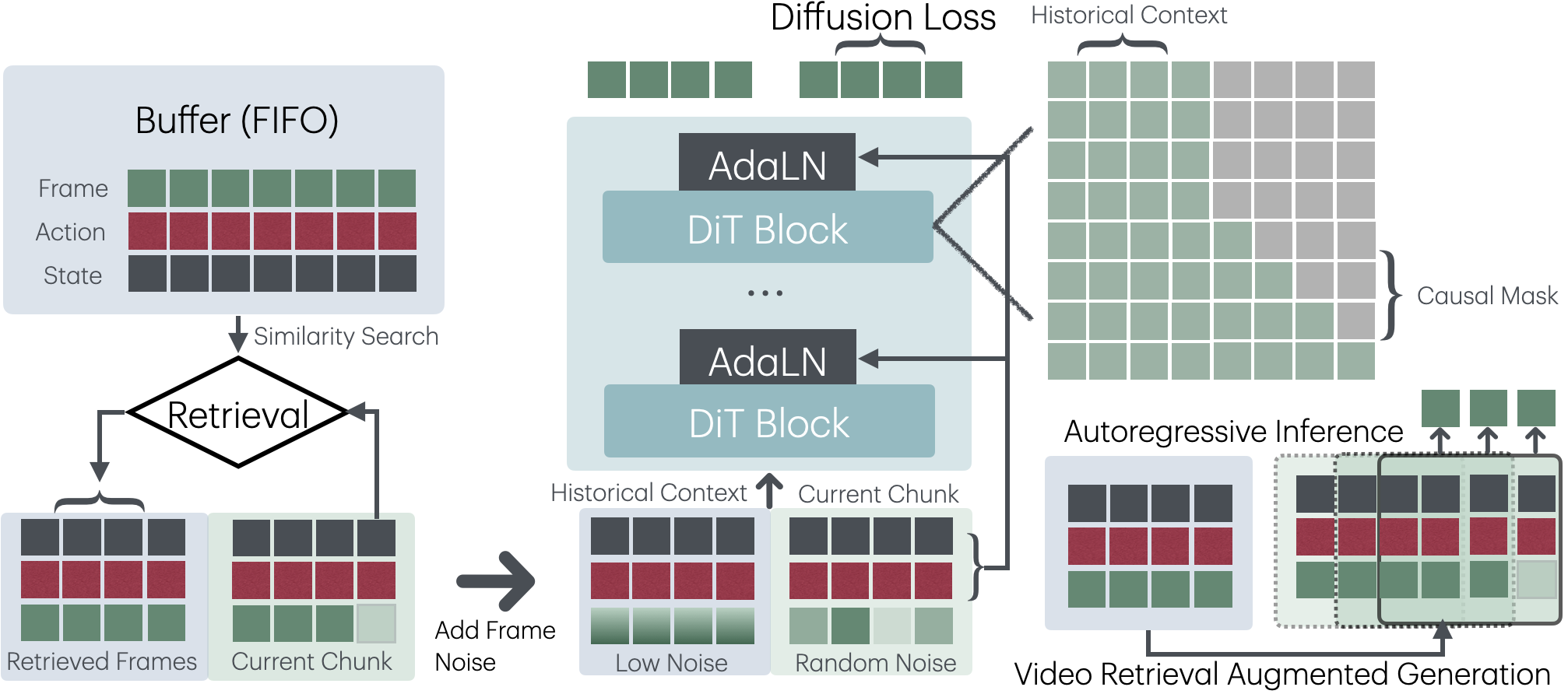}
    \caption{Overview of our VRAG framework for interactive video generation. The framework incorporates global state conditioning and memory retrieval mechanisms to ensure spatiotemporal consistency and mitigate error accumulation. During both training and inference, retrieved memory serves as context for joint self-attention in spatiotemporal DiT blocks. The model employs per-frame noise injection during training to facilitate autoregressive sampling at inference time.}
    \label{fig:diagram}
\end{figure}

\subsection{Interactive Long Video Generation}
\label{subsec:interactive}
To enable interactive long video generation conditioned on action sequences, we augment the base diffusion model with two 
techniques: (1) additional action condition with adaptive layer normalization (AdaLN), and (2) random frame noise for 
autoregressive modeling, as shown in diagram Fig.~\ref{fig:diagram}.

\paragraph{Action Conditioning}
To enable interactive video generation conditioned on action sequences, we augment the base diffusion model with adaptive layer normalization (AdaLN). Given an action sequence $\boldsymbol{a} \in \mathbb{R}^{L \times A}$ where $A$ is the action dimension, we first embed it into a latent space using a learnable embedding layer: 
$e_a = \text{Embed}(\boldsymbol{a}) \in \mathbb{R}^{L \times D_e}$ where $D_e$ is the embedding dimension. For each normalization layer in the diffusion model, we learn action-dependent scale and shift parameters through linear projections: $\gamma_a = e_a W_\gamma  + b_\gamma \in \mathbb{R}^{L \times D_h}, \beta_a = e_a W_\beta  + b_\beta \in \mathbb{R}^{L \times D_h}$, where $D_h$ matches the hidden dimension of the feature maps. We have $\text{AdaLN}(h) = \gamma_a \odot \text{LayerNorm}(h) + \beta_a$, where $h \in \mathbb{R}^{L \times D_h}$ represents the intermediate feature maps and $\odot$ denotes dot production. 

\paragraph{Autoregressive Video Generation}
To enable long video generation, we adopt an autoregressive approach where we generate frames sequentially. At each step, we condition on a fixed-length context window $L_c$ of previously generated frames. However, naive autoregressive generation with teacher forcing can suffer from large compounding errors where mistakes accumulate over time. We apply the Diffusion Forcing~\cite{chen2024diffusion} technique during training.

Specifically, during training, we randomly add noise to each frame in the entire input video sequence according to the diffusion schedule: $z^i_t = \sqrt{\bar{\alpha}_t} z^i_0 + \sqrt{1-\bar{\alpha}_t} \epsilon^i, \epsilon^i \sim \mathcal{N}(0,\mathbf{I})$, where $z^i_t$ represents the noised latent of the $i$-th frame. This forces the model to be robust to noise in the conditioning frames and prevents it from relying too heavily on the context. With above two techniques, the training objective for action-conditioned autoregressive video models become:
\begin{equation}
    \mathcal{L}_\text{DF} = \mathbb{E}_{[t],\boldsymbol{\epsilon},\mathbf{z},a}[\|\boldsymbol{\epsilon}_{[t]} - \boldsymbol{\epsilon}_\theta(\mathbf{z}_{[t]}, [t], \boldsymbol{a})\|_2^2], \quad \boldsymbol{\epsilon}_{[t]}=\{\epsilon^i_t\}_{i=1}^L, \mathbf{z}_{[t]}=\{z^i_t\}_{i=1}^L
\end{equation}
where $[t]$ is vector of $L$ timesteps with different $t\in[T]$ for each frame. The noise prediction model $\boldsymbol{\epsilon}_\theta$ conditioned on both the action sequence $\boldsymbol{a}$ and noised frames $\mathbf{z}_t$.

\paragraph{Architecture}
We apply diffusion transformer (DiT) for video generation modeling. We adopt spatiotemporal DiT block with separate spatial and temporal attention modules. Rotary Position Embedding (RoPE)~\cite{su2024roformer} is applied for both attention modules, and temporal attention is implemented with causal masking.

\subsection{Retrieval Augmented Video World Model with Global State}
\label{sec:rag and global state}
While the vanilla model in Sec.~\ref{subsec:interactive} provides a foundation for interactive video generation, it lacks robust mechanisms for maintaining long-term consistency and world model coherence. To address these limitations, we integrate memory retrieval and context enhancement with inspiration from LLMs, and incorporate video-specific approaches such as historical frame buffer and global state conditioning. These enhancements enable more consistent and coherent autoregressive video generation by providing the model with better access to historical context and spatial awareness.

\paragraph{Global State Conditioning}


To enhance spatial consistency in video generation, we incorporate global state information—specifically the character's current coordinates and pose—as an additional conditioning signal. 
The global state vector $s\in\mathbb{R}^S$ consists of two key components: $s_\text{pos}$ representing 3D position coordinates and $s_\text{ori}$ capturing orientation angles.
Given an action sequence ${\boldsymbol{a}} \in \mathbb{R}^{L \times A}$ and the global state sequence ${\boldsymbol{s}} \in \mathbb{R}^{L \times S}$, both are transformed by a learnable embedding layer, $e_c = \text{Embed}_c(\boldsymbol{a},\boldsymbol{s})$, to produce conditioning features. These features are then fed into AdaLN layers within the diffusion model. This mechanism allows the model to modulate its generation process, adapting to both the input actions and the character's spatial context, thereby improving overall coherence.

\paragraph{Video Retrieval Augmented Generation (VRAG)}
Beyond global state conditioning, we propose memory retrieval augmented generation to enhance the model's ability to leverage historical context while maintaining temporal coherence, namely video retrieval augmented generation (VRAG). 
For VRAG, we combine the concatenated historical and current frames with their corresponding action sequences $\tilde{\boldsymbol{a}} \in \mathbb{R}^{L \times A}$ and global state sequences $\tilde{\boldsymbol{s}}=[\boldsymbol{s}_\text{hist}, \boldsymbol{s}] \in \mathbb{R}^{L \times S}$ as conditional inputs to the model. The historical frames are retrieved from a fixed-length buffer $\mathcal{B}$, which stores previously generated frames. The per-frame retrieval process is based on a heuristic sampling strategy, where we select the most relevant historical frames based on similarity search to concatenate with the current context. 
The similarity score based on global state is defined as:
\begin{align}
    r(\hat{s}) &= f_{\text{sim}}(\hat{s} \odot w, s_{L-1} \odot w), \hat{s}\in \mathcal{B}
\end{align}
where $f_\text{sim}$ is a distance metric (e.g., Euclidean distance) between the history frame and the last frame to be predicted $s_{L-1}$, and $w\in\mathbb{R}^S$ is a weight vector that modulates the importance of different state components. The top $L_h$ most similar historical states and frames are selected and sorted to form the retrieved context.
Unlike RAG in LLMs which leverages strong in-context learning capabilities, \textbf{video diffusion models exhibit weak in-context learning abilities, making direct inference with historical frames as context ineffective}, as demonstrated later in our experiments. To address this limitation, we propose VRAG training with key modifications to the standard RAG approach, enabling effective memory-augmented video generation.

During training, we retrieve historical frames $\mathbf{z}_{\text{hist}} \in \mathbb{R}^{L_h \times D}$ and concatenate them with the current context window $\mathbf{z} \in \mathbb{R}^{L_c \times D}$ to form the extended context $\tilde{\mathbf{z}} = [\mathbf{z}_{\text{hist}}, \mathbf{z}]$. For effective VRAG, we make several key modifications: (1). To distinguish retrieved frames from normal context frames, we modify the RoPE embeddings by adding a temporal offset $\Delta t$ to the retrieved frames' position indices. (2). Additionally, we apply lower noise levels $\beta_{t'} < \beta_t$ to the retrieved frames $\mathbf{z}_{\text{hist}}$ to simulate partially denoised historical frames during inference. This enhances the robustness of the model with imperfect historical frames generated previously during the autoregressive process. The model is trained to denoise for the entire context $\tilde{\mathbf{z}}$ including both retrieved and current frames.
(3). To ensure the model focuses on denoising the current context while leveraging historical information, we mask the diffusion loss $\mathcal{L}_{\text{DF}}$ for retrieved frames. (4). Furthermore, for retrieved frames, we only condition on their global states $\boldsymbol{s}_{\text{hist}}\in \mathbb{R}^{L_h \times S}$, masking out action conditions $\boldsymbol{a}_{\text{hist}}\in \mathbb{R}^{L_h \times A}$ to avoid temporal discontinuity in action sequences. This selective conditioning approach helps maintain spatial consistency while preventing action-related artifacts from propagating through the generation process. Overall, the training objective of VRAG on diffusion models is defined as:
\begin{align}
    \mathcal{L}_\text{VRAG} &= \mathbb{E}_{[t],[t'],\boldsymbol{\epsilon},\tilde{\mathbf{z}},a,s}[\|\boldsymbol{\epsilon}_t - \boldsymbol{\epsilon}_\theta(\tilde{\mathbf{z}}_{\tilde{t}}, \tilde{t}, \tilde{\boldsymbol{a}}, \tilde{\boldsymbol{s}})\|_2^2 \odot \mathbf{m}],\\
     \tilde{\mathbf{z}}_{\tilde{t}} = [\mathbf{z}_{\text{hist},[t']}, &\mathbf{z}_{[t]}], \quad \tilde{\boldsymbol{a}} = [\varnothing_{L_h}, \boldsymbol{a}], \quad \tilde{\boldsymbol{s}} = [\boldsymbol{s}_{\text{hist}}, \boldsymbol{s}], \quad \mathbf{m} = [\mathbf{0}_{L_h}, \mathbf{1}_{L_c}],
\end{align}
where $\tilde{t}$ is a concatenation of $[t']$ and $[t]$, with $t'<t$ and $t',t \in [T]$.



\subsection{Long-context Extension Baselines}
\label{subsec:baseline}
To investigate whether established long-context extension techniques from LLMs can effectively enhance video generation models, we design three complementary approaches that leverage either explicit frame context or neural memory hidden states, based on vanilla models in Sec.~\ref{subsec:interactive}. These methods serve as baseline comparisons to our main approach, specifically targeting the model's ability to maintain spatial coherence and temporal consistency in long video generation. Through these baselines, we aim to verify the in-context learning capabilities of video diffusion models and assess their effectiveness in handling extended sequences.

\paragraph{Long-context Enhancement}
We extend the temporal context window using YaRN~\cite{peng2023yarn} modification for RoPE in temporal attention. RoPE encodes relative positions via complex-valued rotations, where the inner product between query $\mathbf{q}_m$ and key $\mathbf{k}_n$ depends on relative distance $(m-n)$. YaRN extends the context window by applying a frequency transformation to the rotary position embeddings. This transformation scales the rotation angles in a way that preserves the relative positioning information while allowing the model to handle longer video sequences, after small-scale fine-tuning on longer video clips.

\paragraph{Frame Retrieval from History Buffer}
We implement a fixed-length buffer $\mathcal{B}$ storing historical latent frames with a heuristic sampling strategy. The buffer is partitioned into $N_S=5$ exponentially decreasing segments $G_j$, where $L_j = L_1 \cdot \alpha^{j-1}$. From each segment $G_j$, we sample $k$ frames to form subset $F_j$. The retrieved memory $\mathbf{z}_{\text{mem}} = [F_1, \dots, F_{N_S}]$ is concatenated with current frame window $\mathbf{z}$ as additional context: $\tilde{\mathbf{z}}=[\mathbf{z}_{\text{mem}}, \mathbf{z}]$, which is then passed into the spatiotemporal DiT blocks. This design ensures higher sampling density for recent frames, emphasizing recent visual information while maintaining access to historical context for temporal consistency.

\paragraph{Neural Memory Augmented Attention} 
Instead of using explicit frames as context in above two methods, we explore a neural memory mechanism to store and retrieve hidden states. This approach is inspired by the success of Infini-attention~\cite{munkhdalai2024leavecontextbehindefficient} in LLMs, which utilizes a compressed memory representation to enhance attention mechanisms. The model processes video in overlapping segments to maintain temporal continuity. For each video segment $\mathbf{z}_s$, we compute query $\mathbf{q}_s$, key $\mathbf{k}_s$ and value $\mathbf{v}_s$ matrices. 
The model retrieves hidden state $\mathbf{A}_{\text{mem}}$ from compressive memory $\mathbf{M}_{s-1}$: $\mathbf{A}_{\text{mem}} = \frac{\sigma(\mathbf{q}_s)\mathbf{M}_{s-1}}{\sigma(\mathbf{q}_s)\mathbf{n}_{s-1}}$.
Memory $\mathbf{M}_{s-1}$ and normalization vector $\mathbf{n}_{s-1}$ are then updated.
The final attention output combines retrieved hidden state $\mathbf{A}_{\text{mem}}$ and standard attention using learnable gating to maintain visual consistency across the long video sequence.

\paragraph{Frame Pack}
As another baseline, we follow the Frame Pack~\citep{zhang2025packinginputframecontext} to compress historical frames as context. Three input compression kernels with different kernel sizes-(2, 4, 4), (4, 8, 8), and (8, 16, 16)-are employed to condense the historical frames into a fixed-length context. This approach essentially achieves frame compression through importance sampling with recency bias, which enables a larger field of view while maintaining lower computational costs. However, the prioritization of most recent frames can be suboptimal in many cases for long video generation especially when considering the memory issue. Our VRAG based on frame relevance provides theoretically better historical information retrieval. Moreover, our method is actually orthogonal to the frame compression technique in Frame Pack. We leave the combined methods as future work.


More details of the above methods can be found in the supplementary material.

\section{Experiments}
\label{gen_inst}
\begin{figure}[htbp]
    \centering
    \includegraphics[width=\linewidth]{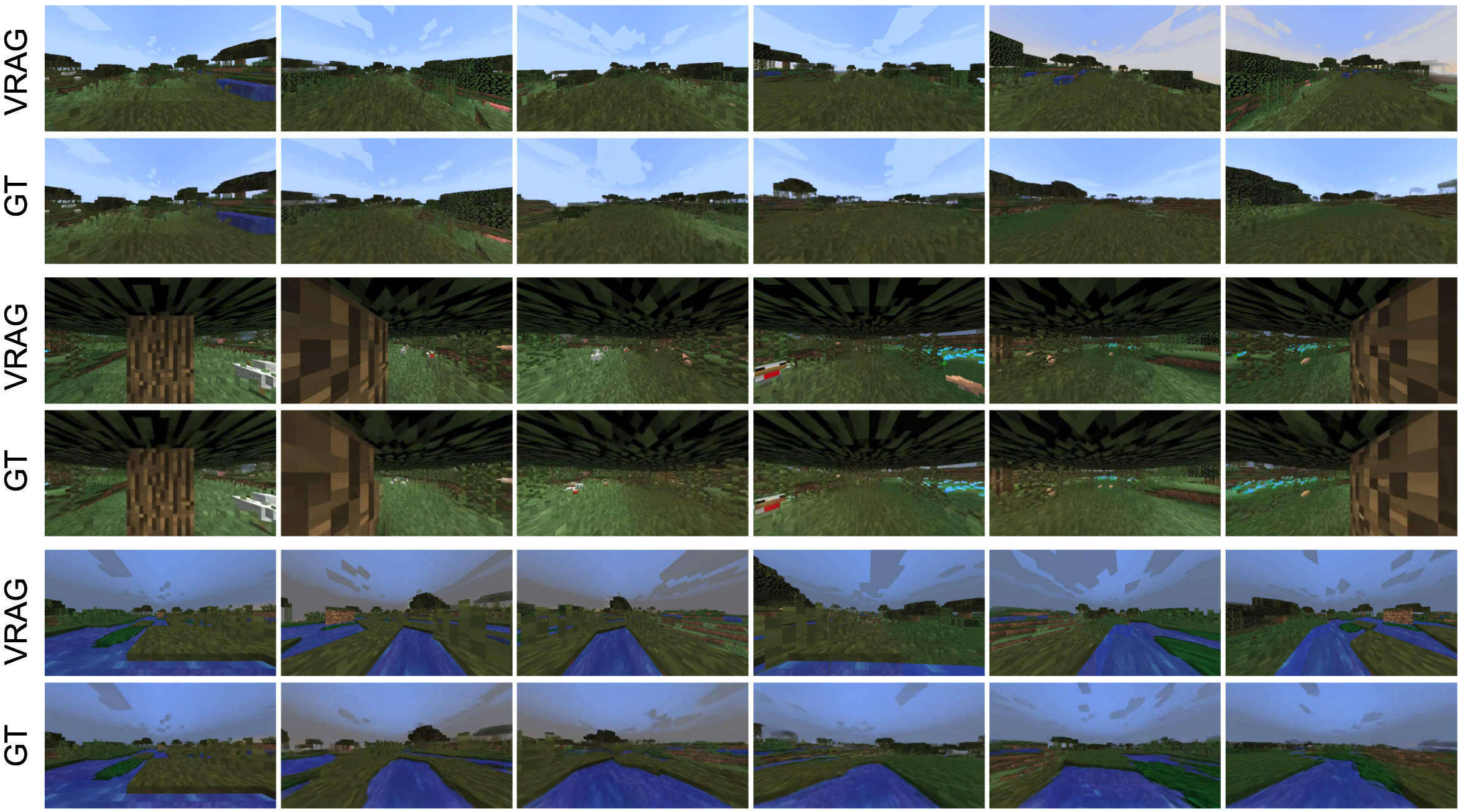}
    \caption{Visual comparison of VRAG with ground truth videos on world coherence evaluation. With 100 initial frames as history buffer, VRAG predicts 200 subsequent frames.}
    \label{fig:visual_compare}
\end{figure}

\subsection{Datasets and Evaluation Protocol}

For training, we collected 1000 long Minecraft gameplay videos (17 hours total) using MineRL~\cite{guss2019minerl}. All videos have a fixed resolution of 640×360 pixels. Each sequence spans 1200 frames, annotated with action vectors (forward/backward movement, jumping, camera rotation) and world coordinates (x, y, z positions and yaw angle).

For evaluation, we assembled two distinct test sets: (1) for compounding error evaluation, we use 20 long videos of 1200 frames with randomized actions and locations, and (2) for world coherence, we use 60 carefully curated 300-frame video sequences designed to systematically assess spatiotemporal consistency. These curated sequences feature controlled motion patterns including in-place rotation, direction reversal, and circular trajectory following. The first 100 frames of each sequence serve as initialization buffer for methods requiring buffer frames or are excluded from evaluation for others. Each model autoregressively generates next single frame with stride 1 until the desired length.


We evaluate the models against ground-truth test sets using several metrics: Structural Similarity Index (SSIM)~\cite{wang2004image} to measure spatial consistency, Peak Signal-to-Noise Ratio (PSNR) for pixel-level reconstruction quality, Learned Perceptual Image Patch Similarity (LPIPS)~\cite{zhang2018unreasonable} to assess perceptual similarity. For the compounding error evaluation, we find SSIM more accurately reflect the faithfulness of frames over long sequences.

\subsection{Training Details}
\label{sec:training_details}
\begin{wrapfigure}{tr}{0.7\textwidth}
    \centering
    \includegraphics[width=1.0\linewidth]{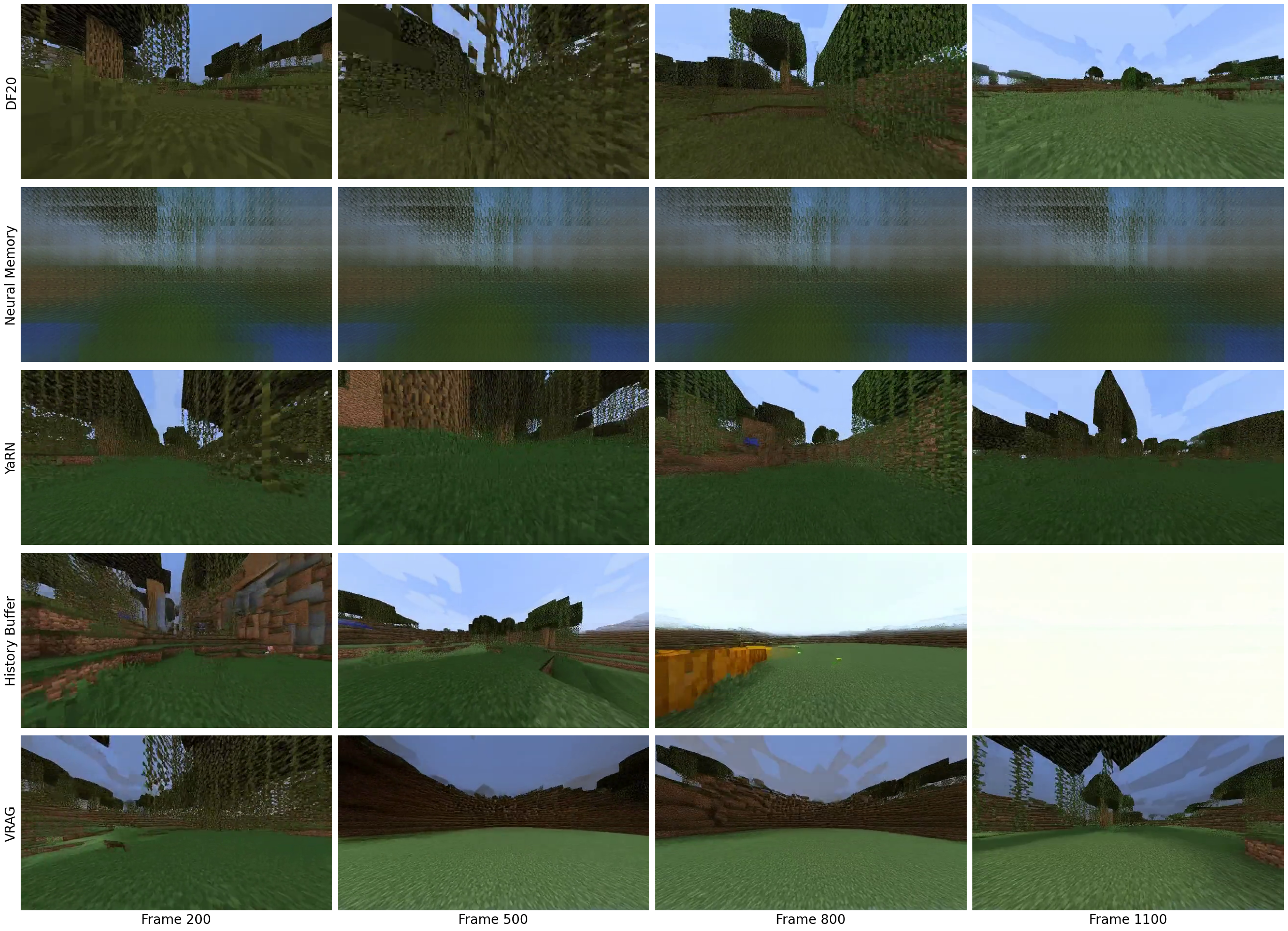}
    \captionsetup{width=0.7\textwidth}
    \caption{Visual comparison of different methods, evaluated for world coherence analysis.}
    \label{fig:visual_compare_coherence}
\end{wrapfigure}
A consistent window size of 20 frames is applied for both model training and evaluation for fair comparison. For vanilla Diffusion Forcing, we additionally train a variant with window sizes of 10 frame for context length evaluation. For our VRAG method, we combine 10 retrieved frames with 10 current frames for both training and inference. We represent the agent's state using a global state vector $s=[x,y,z, \text{yaw}]$ during training, which can be extended to incorporate a full 3D pose representation when needed.  To facilitate training convergence, these values are normalized relative to the initial state, thereby reducing the complexity of the diffusion process. The YaRN implementation extends the vanilla model (window size 20) by replacing position embeddings with YaRN and stretching factor $4$, followed by fine-tuning for $10^4$ steps on 80-frame sequences. During evaluation of Yarn, we use a 40-frame window.
The Infini-attention with neural memory employs a sliding window size 20 and stride 10, using the first 10 frames for memory state updates and the last 10 for local attention computation. The History Buffer method maintains a 124-frame buffer partitioned into 5 exponentially decreasing segments ($L_1 = 2, \alpha = 2$), sampling 2 frames per segment to form 10 historical frames that are concatenated with the 10 current frames. All models are trained for 3 epochs on the dataset, with a batch size of 32 across 8 A100 GPUs.

\subsection{World Coherence Results \label{sec:world coherence results}}

\begin{figure}[htbp]
    \centering
    \begin{minipage}{0.5\textwidth}
        \centering
        \includegraphics[width=\linewidth]{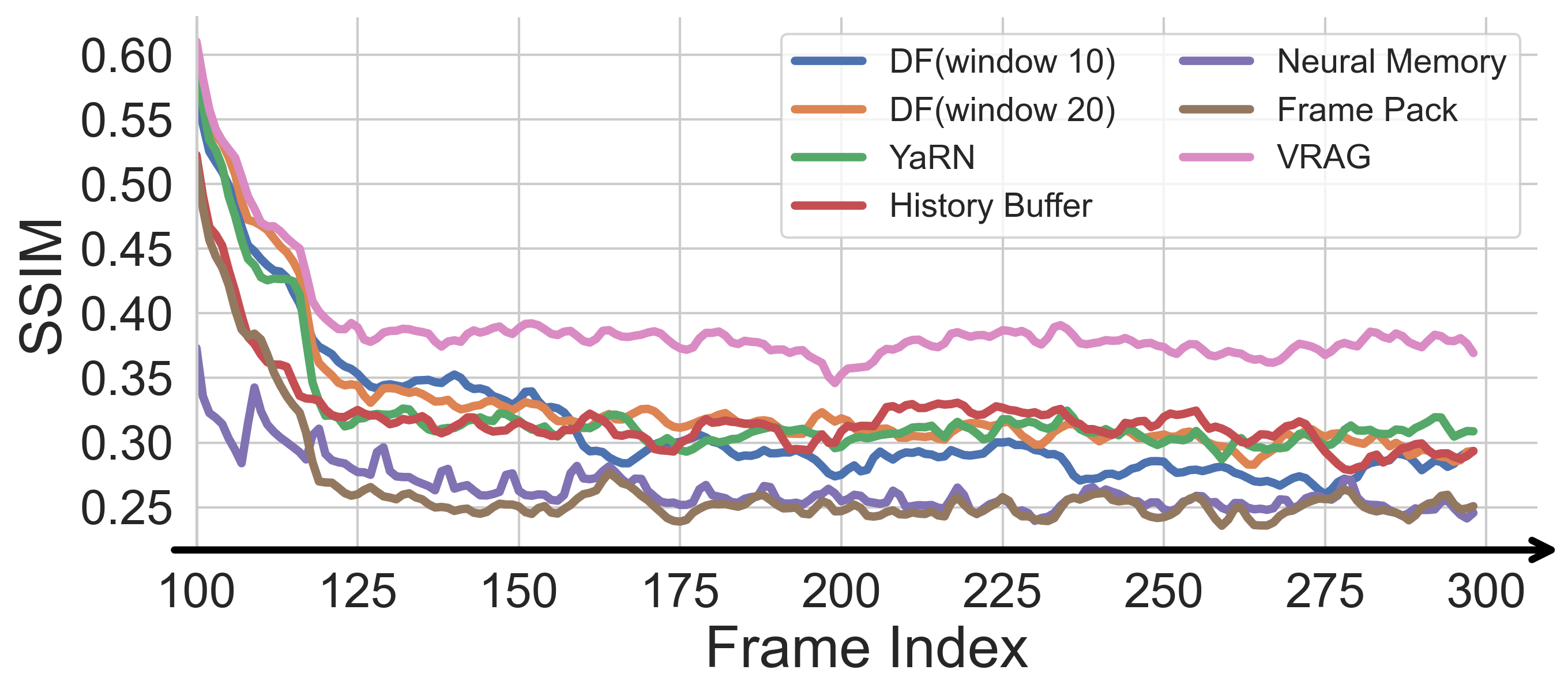}
        \caption{SSIM scores over time for different methods on world coherence evaluation.}
        \label{fig:world_coherence}
    \end{minipage}
    \hfill
    \begin{minipage}{0.49\textwidth}
        \centering
        \begin{adjustbox}{max width=\textwidth}
        \begin{tabular}{lccc}
            \hline
            Method & SSIM $\uparrow$ & PSNR $\uparrow$ & LPIPS $\downarrow$ \\
            \hline
            DF (window 10) & 0.455 & 16.161 & 0.509 \\
            DF (window 20) & 0.466 & 16.643 & 0.538 \\
            YaRN & 0.462 & 16.567 & 0.532 \\
            History Buffer & 0.459 & 16.922 & 0.543 \\
            Frame Pack & 0.421 & 16.372 & 0.574 \\
            \textbf{VRAG} & \textbf{0.506} & \textbf{17.097} & \textbf{0.506} \\
            \hline
        \end{tabular}
        \end{adjustbox}
        \captionof{table}{Quantitative comparison of world coherence across different methods, evaluated on videos with 300 frames.}
        \label{tab:world_coherence}
    \end{minipage}
\end{figure}

We investigate the spatiotemporal consistency of internal world models by evaluating the predicted videos given initial frames and action sequences. As visualized in Fig.~\ref{fig:visual_compare_coherence}, our VRAG provides an effective approach to enhance the model's ability to leverage historical context for improving world coherence. Fig.~\ref{fig:visual_compare} shows more visual comparison of VRAG with ground truth videos.
We evaluate the world coherence of different methods using multiple metrics.  Fig. \ref{fig:world_coherence} shows the SSIM scores over time, while Tab. \ref{tab:world_coherence} presents a comprehensive comparison across all metrics. Our VRAG method achieves the best performance across all metrics, demonstrating its superior ability to maintain world coherence in generated videos.
Our experimental results demonstrate that expanding the window size from 10 to 20 frames in the baseline DF model improves world coherence, indicating that longer context windows enhance consistency. However, further context extension using YaRN shows no improvement over the vanilla DF model. This suggests that YaRN's context extension capabilities, while effective in language models, do not transfer effectively to video generation for maintaining world coherence. Similarly, the History Buffer method fails to effectively utilize historical frames for spatiotemporal consistency without explicit in-context training.
These findings from both YaRN and History Buffer approaches reveal that video diffusion models at the current scale possess limited in-context learning capabilities, preventing them from effectively leveraging historical frames for maintaining long-term consistency. The Neural Memory method performs poorly due to its instability in model training.

\subsection{Compounding Error Results \label{sec:compounding error results}}
\begin{figure}[htbp]
    \centering
    \includegraphics[width=0.9\textwidth]{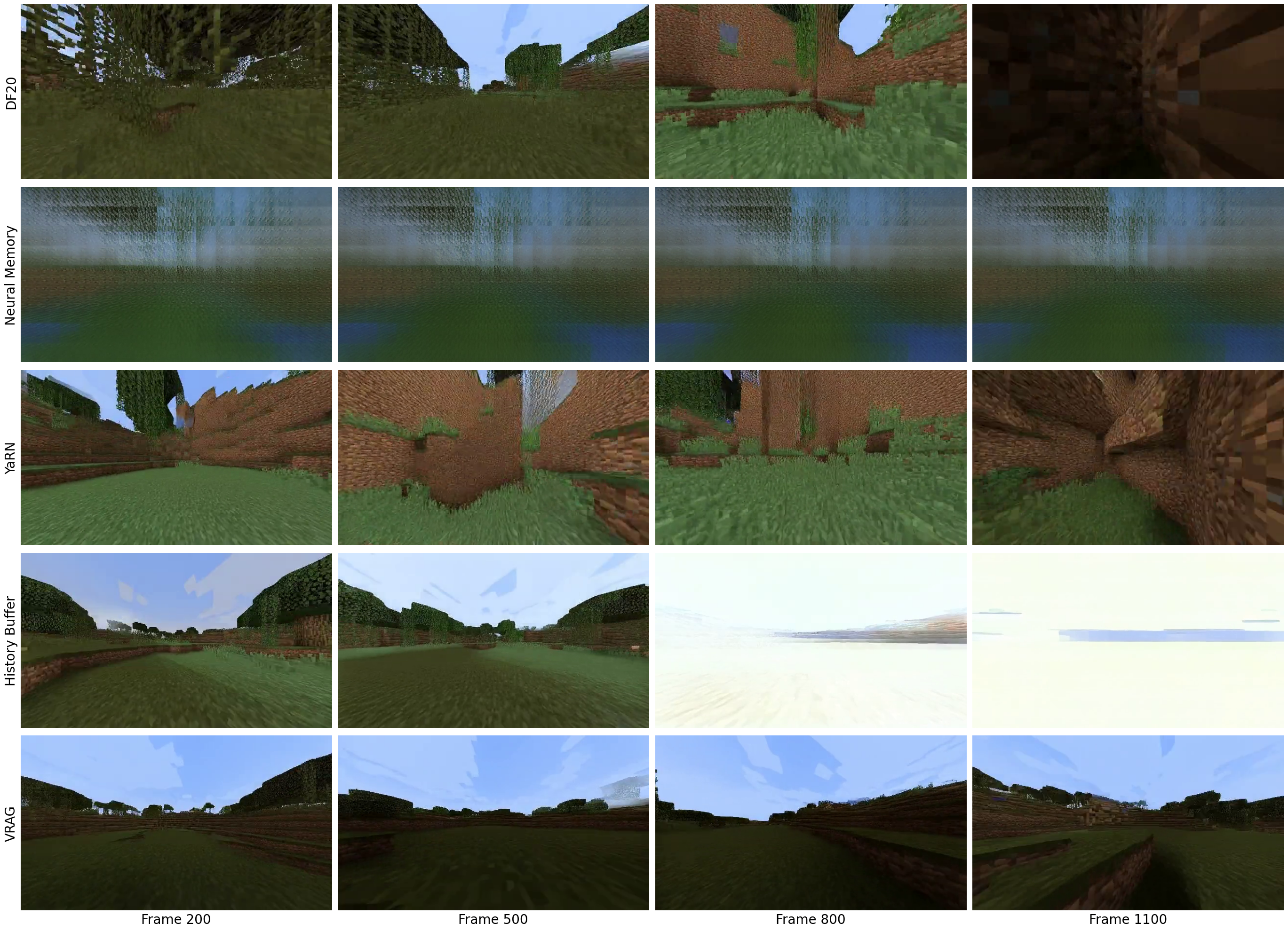}
    \caption{Visual comparison of long-term video prediction (1200 frames) across different methods, evaluated for compounding error analysis.}
    \label{fig:visual_compounding_error}
\end{figure}
\begin{figure}[htbp]
    \centering
    \begin{minipage}{0.48\textwidth}
        \centering
        \includegraphics[width=\linewidth]{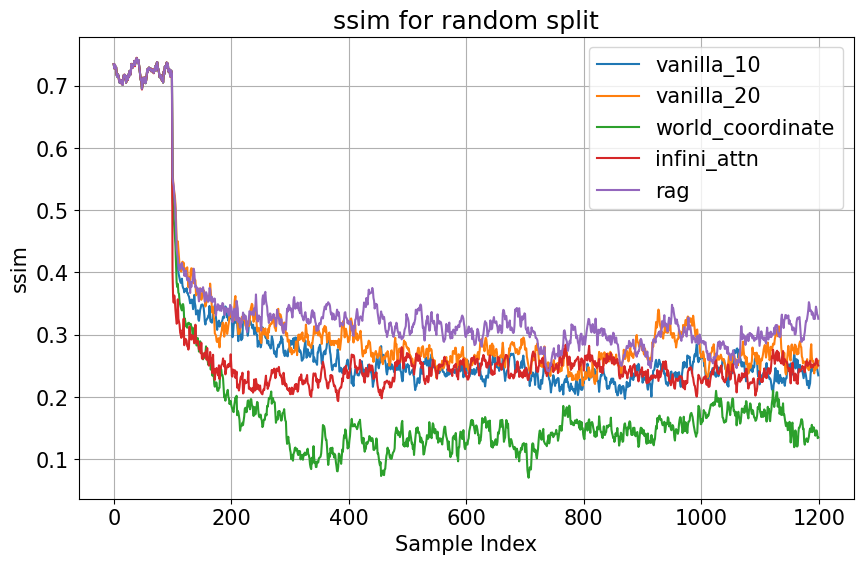}
        \caption{SSIM scores over time for compounding error evaluation}
        \label{fig:compounding_error_all}
    \end{minipage}
    \hfill
    \begin{minipage}{0.48\textwidth}
        \centering
        \begin{tabular}{lc}
            \hline
            Method & SSIM $\uparrow$ \\
            \hline
            DF (window 10) & 0.297 \\
            DF (window 20) & 0.321 \\
            YaRN & 0.316 \\
            History Buffer & 0.188 \\
            Neural Memory & 0.283 \\
            \textbf{VRAG} & \textbf{0.349} \\
            \hline
        \end{tabular}
        \captionof{table}{Average SSIM scores across all frames in compounding error evaluation}
        \label{tab:compounding_error}
    \end{minipage}
\end{figure}

We evaluate the compounding error in long video generation across different methods using the SSIM metric. As shown in Fig. \ref{fig:compounding_error_all} and Tab. \ref{tab:compounding_error}, our VRAG method achieves superior performance with an SSIM score of 0.349, demonstrating better structural similarity preservation compared to baseline methods. Increasing the window size in DF from 10 to 20 frames improves SSIM, indicating that longer context helps mitigate compounding errors. However, this improvement is still inferior to VRAG's performance, suggesting that our retrieval-augmented approach provides more effective long-term consistency.
As visualized in Fig.~\ref{fig:visual_compounding_error}, our VRAG method generates more coherent and consistent frames over long sequences, while other methods exhibit noticeable artifacts and inconsistencies. The History Buffer method performs poorly, with an SSIM score of 0.188, indicating that naive historical frame retrieval without effective in-context training fails to maintain long-term consistency. 
Given its limited performance in the world coherence experiments (Sec.~\ref{sec:world coherence results}), we exclude the Neural Memory method from visualization in this longer video prediction visualization.

\subsection{VBench Evaluation}


We evaluate the long video generation with five Video Quality metrics in VBench~\citep{10657096} for generated videos in Sec.~\ref{sec:compounding error results}. The evaluation results on VBench (higher is better) are shown in the Tab.~\ref{tab:vbench}. As demonstrated in the results, our method outperforms all other baselines across all metrics in both temporal quality and video frame quality. The Neural Memory baseline has Aesthetic Quality 0.343 and Imaging Quality score 0.3597 respectively, which are significantly lower than other baselines, therefore not listed here. Our VRAG shows better temporal consistency compared with all baseline methods, and the high video frame quality indicates the results are not over-smoothed.

\subsection{Extension: Real World Setting}


\begin{figure}[htbp]    \centering
        \includegraphics[width=\linewidth]{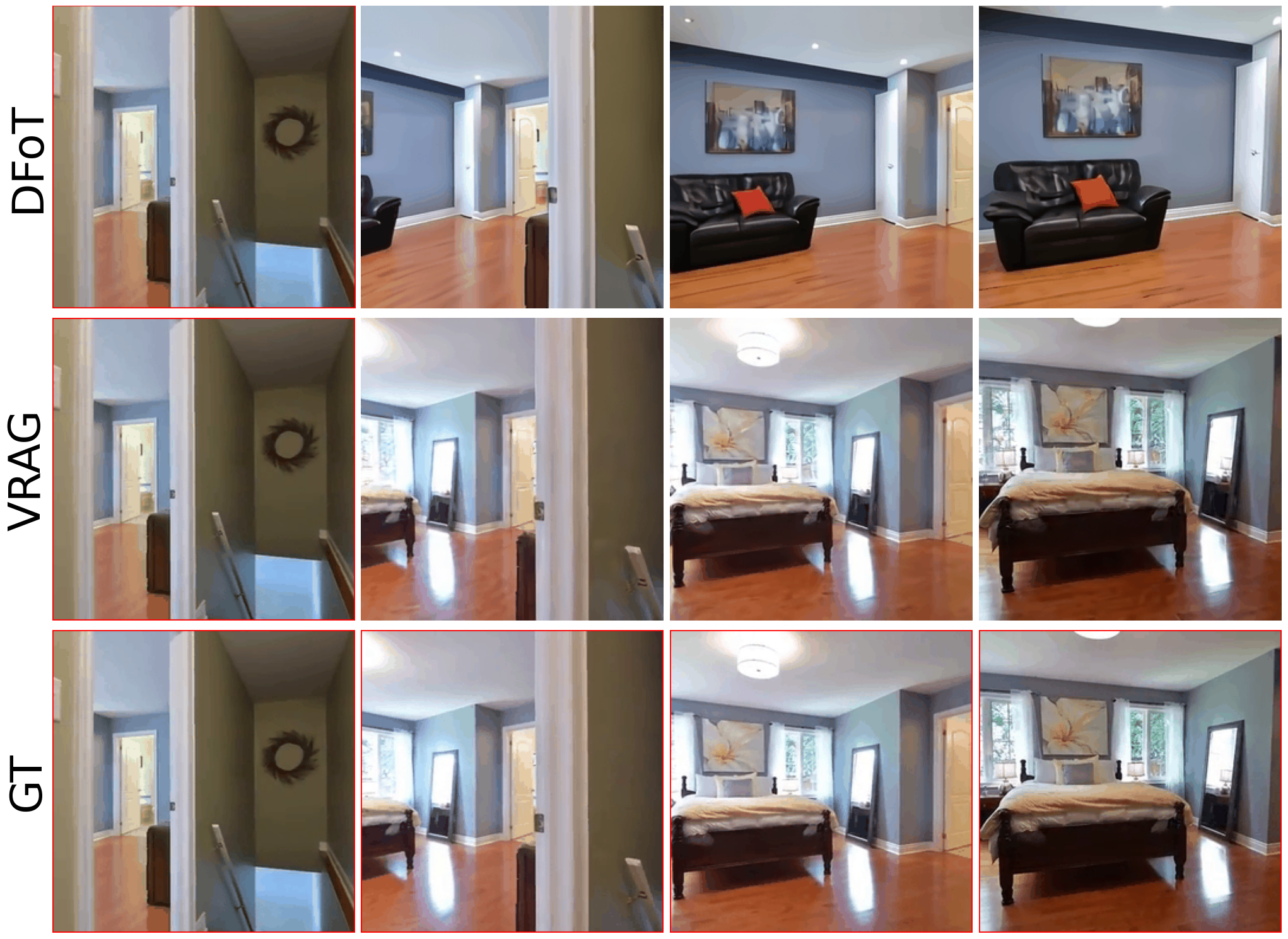}
      \captionsetup{width=0.6\textwidth}
        \caption{Visualized video frames on RealEstate10K dataset. Red blocks indicate the ground-truth frames.}
    \label{fig:realestate}
\end{figure}

\begin{table}[htbp]
    \centering
    \begin{minipage}[t]{0.72\textwidth}
        \centering
    \begin{adjustbox}{max width=\textwidth}
    \begin{tabular}{l|ccccc}
    \hline
    Method & \makecell{Background\\Consistency} & \makecell{Temporal \\Flickering} & \makecell{Motion \\Smoothness} & \makecell{Aesthetic \\Quality} & \makecell{Imaging \\Quality} \\
    \hline
    DF20 & 0.9668 & 0.9485 & 0.9582 & 0.5272 & 0.6058 \\
    YaRN & 0.9686 & 0.9401 & 0.9523 & 0.5252 & 0.6323 \\
    History Buffer & 0.9664 & 0.9475 & 0.9579 & 0.5167 & 0.6253 \\
    VRAG & \textbf{0.9686} & \textbf{0.9511} & \textbf{0.9603} & \textbf{0.5295} & \textbf{0.6444} \\
    \hline
    \end{tabular}
    \end{adjustbox}
    \captionof{table}{Evaluation results on five Video Quality metrics in VBench.}
    \label{tab:vbench}
    \end{minipage}
    \hfill
    \begin{minipage}[t]{0.27\textwidth}
        \centering
        \begin{adjustbox}{max width=\textwidth}
        \begin{tabular}{lcc}
            \hline
            Metric & DFoT &  VRAG  \\
            \hline
            SSIM $\uparrow$ & 0.4436 & \textbf{0.9116} \\
            PSNR $\uparrow$ & 13.03 & \textbf{32.21} \\
            LPIPS $\downarrow$ & 0.4469 & \textbf{0.1146} \\
            FVD $\downarrow$ & 337.5 & \textbf{221} \\
            \hline
        \end{tabular}
        \end{adjustbox}
        \captionof{table}{Quantitative comparison on RealEstate10K dataset.}
        \label{tab:realestate}
    \end{minipage}
\end{table}

We conduct additional experiments in real-world setting beyond Minecraft simulation to show generalization of our approach. Specifically, following the experimental setup of Diffusion Forcing Transformer (DFoT) \citep{song2025historyguidedvideodiffusion}, our VRAG model is initialized from pre-trained DFoT and finetuned on the RealEstate10K dataset \citep{realestate} with additionally retrieved historical context, as described in Sec.~\ref{sec:rag and global state}.

After fine-tuning for just 2 epochs (10\% of the original training steps), our method significantly outperforms the DFoT baseline in terms of memorization capability. The visualized frames are presented in Fig.~\ref{fig:realestate} and quantitative results are summarized in Tab.~\ref{tab:realestate}. The results effectively demonstrate the generalization of our approach beyond Minecraft for solving the memory issue in long video prediction.

\subsection{Ablation: Memory and Training of VRAG}

\begin{figure}[htbp]
    \centering
    \begin{minipage}{0.42\textwidth}
        \centering
        \includegraphics[width=\linewidth]{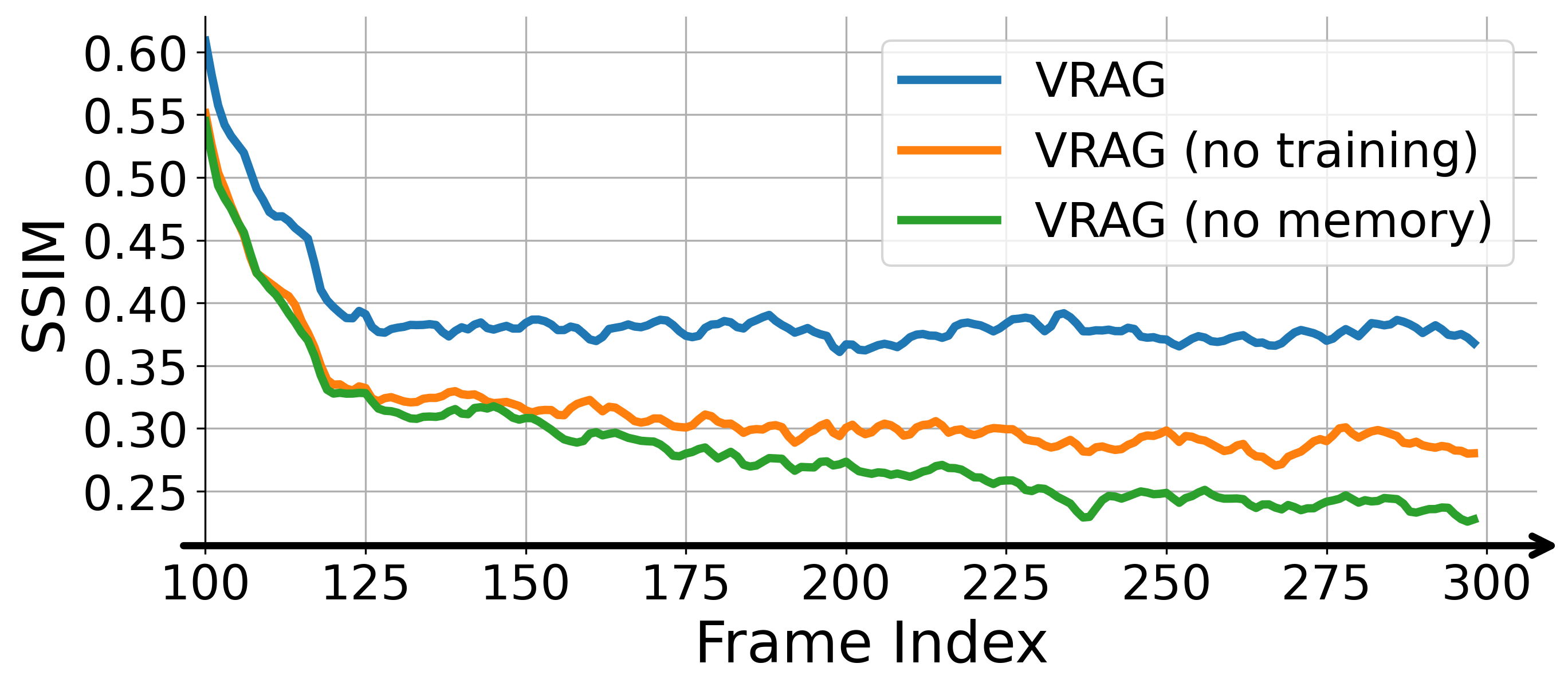}
        \caption{Comparison of SSIM scores over time for VRAG variants.}
        \label{fig:ablation_ssim}
    \end{minipage}
    \hfill
    \begin{minipage}{0.57\textwidth}
        \centering
        \begin{tabular}{@{}lccc@{}}
            \hline
            Method & SSIM $\uparrow$ & PSNR $\uparrow$ & LPIPS $\downarrow$ \\
            \hline
            \textbf{VRAG} & \textbf{0.506} & \textbf{17.097} & \textbf{0.506} \\
            VRAG (no training) & 0.455 & 16.670 & 0.528 \\
            VRAG (no memory) & 0.436 & 16.372 & 0.547 \\
            \hline
        \end{tabular}
        \captionof{table}{Ablation study of VRAG components. We compare the full model with variants that remove either the memory component (additional global state conditioning only) or training component (in-context learning only).}
        \label{tab:ablation_vrag}
    \end{minipage}
\end{figure}

We ablate the key designs for VRAG methods, including the memory and training components. The ablation results are shown in Fig. \ref{fig:ablation_ssim} and Tab. \ref{tab:ablation_vrag}. We compare the full VRAG model with two variants: (1) VRAG without the memory component, which only uses additional global state conditioning, and (2) VRAG without the training component, i.e., vanilla model with retrieval augmented generation for in-context learning at inference. The ablation study is conducted on the world coherence evaluation dataset.

The ablation results reveal several key insights about VRAG components. First, removing the memory component leads to the largest performance drop across all metrics, with SSIM decreasing by 13.8\% and LPIPS increasing by 8.1\%. This demonstrates that the memory mechanism is crucial for maintaining spatiotemporal consistency and quality. Second, removing the training component also causes significant degradation, with SSIM dropping by 10.1\% and LPIPS increasing by 4.3\%, highlighting the weak capabilities of in-context learning for current video models. The full VRAG model achieves the best performance across all metrics, showing that both components work synergistically to improve video generation quality.

\section{Baseline Method Details}
For the baseline methods in Sec.~\ref{subsec:baseline}, we implemented the following techniques to enhance the temporal context window of our video generation model.
\paragraph{Long-context Enhancement}

To extend the temporal context window of our video generation model, we apply the YaRN~\cite{peng2023yarn} modification for ROPE in temporal attention module for improved extrapolation. RoPE encodes relative position via complex-valued rotations, such that the inner product between the $m$-th query $\mathbf{q}_m$ and $n$-th key $\mathbf{k}_n$ depends only on the relative distance $(m - n)$:
\begin{align}
\langle \mathbf{q}_m, \mathbf{k}_n \rangle &=  \langle f_{\mathbf{W}_q}(\mathbf{z}_m, m), f_{\mathbf{W}_k}(\mathbf{z}_n, n) \rangle_{\mathbb{R}} \\
&= \mathrm{Re}\left( \langle (\mathbf{W}_q \mathbf{z}_m)e^{im\theta}, (\mathbf{W}_k \mathbf{z}_n)e^{in\theta} \rangle\right)\\
&= \mathrm{Re} \left(  (\mathbf{W}_q \mathbf{z}_m)(\mathbf{W}_k \mathbf{z}_n)^* \cdot e^{i  (m - n)} \theta\right)\\
&= g(\mathbf{z}_m, \mathbf{z}_n, m-n)
\end{align}
where $\text{Re}[\cdot]$ is real part of complex values and $(\cdot)^*$ represents conjugate of complex numbers, $\mathbf{z}_m, \mathbf{z}_n\in \mathbb{R}^D$ are input vectors, $\mathbf{W}_q$, $\mathbf{W}_k$ are learned projections, and $\theta \in \mathbb{R}^D$ encodes rotation frequencies per dimension: $\theta_d = b^{-2d/D}, \text{with } b = 10000$.

YaRN modifies modifies the rotated input vector $f_\mathbf{W}(\mathbf{z}_m, m, \theta_d)$ by applying a frequency transformation:
\begin{equation}
f'_{\mathbf{W}}(\mathbf{z}_m, m, \theta_d) = f_{\mathbf{W}}(\mathbf{z}_m, g(m), h(\theta_d))
\end{equation}
with $g(m) = m$ and frequency warping function:
\begin{equation}
h(\theta_d) = \left(1 - \gamma(r_d)\right) \cdot \frac{\theta_d}{s} + \gamma(r_d) \cdot \theta_d
\end{equation}
Here, $s$ is a stretching factor and $r_d = L_c / \lambda_d$ is the context-to-wavelength ratio with $\lambda_d = 2\pi / \theta_d = 2\pi \left(b^{\prime}\right) ^{2d/D}$ and $b^{\prime} = bs^{\frac{D}{D-2}}$. The ramp function $\gamma(\cdot)$ interpolates low-frequency dimensions to improve extrapolation while preserving high-frequency components.

\paragraph{Frame Retrieval from History Buffer}
We also experimented with a fixed-length buffer $\mathcal{B}$ that stores a history of previously generated latent frames, employing a heuristic sampling strategy for retrieval. Following \cite{jiang2024loopy}, this strategy involves partitioning $\mathcal{B}$ into $N_S=5$ segments $G_j$ for $j \in \{1, \dots, N_S\}$, ordered from oldest ($G_1$) to most recent ($G_{N_S}$). The total number of frames in the buffer is $N_B = \sum_{j=1}^{N_S} |G_j|$. The lengths of these segments, $L_j = |G_j|$, decrease exponentially (e.g., $L_j = L_1 \cdot \alpha^{j-1}$ for a base $\alpha < 1$ and $L_1$ being the length of the oldest segment $G_1$), ensuring that more recent segments are shorter. From each segment $G_j$, $k$ frames are randomly sampled to form a subset $F_j \subseteq G_j$ (where $|F_j|=k$). The retrieved memory $\mathbf{z}_{\text{mem}}$ is constructed as the concatenation of these sampled frames, $\mathbf{z}_{\text{mem}} = [F_1, F_2, \dots, F_{N_S}]$, totaling $N_S \cdot k$ frames. This design with recency bias implies that the sampling density $k/L_j$ is higher for more recent segments, thereby placing greater emphasis on recent information. This retrieved information $\mathbf{z}_{\text{mem}}$ is concatenated with current frame window $\mathbf{z}$ along temporal dimension as additional context: $\tilde{\mathbf{z}}=[\mathbf{z}_{\text{mem}}, \mathbf{z}]$, which is then passed as input to the spatiotemporal DiT blocks, enabling the model to jointly attend to both recent and historical frames.


\paragraph{Neural Memory Augmentation}

To extend video generation capabilities to longer sequences beyond a fixed attention window while retaining memory of past scenes, we adapt Infini-attention~\cite{munkhdalai2024leavecontextbehindefficient} as a neural memory mechanism for our video diffusion model. Infini-attention is a recurrent mechanism that augments standard dot-product attention (local context) with a compressed summary of past context (global context) stored in an evolving memory. 

The model processes the video in segments using a sliding window. To maintain the high degree of temporal continuity crucial for video generation, we employ overlapping segments. This is a modification from the original Infini-attention, which typically processes non-overlapping segments.
The input latent video segment $\mathbf{z}_s\in \mathbb{R}^{N\times D}$ ($s$ is segment index) is processed to derive query $\mathbf{q}_s$, key $\mathbf{k}_s$ and value $\mathbf{v}_s$ matrices using standard attention mechanisms. Key-value pairs from processed segments are incrementally summarized and stored in a compressive memory $\mathbf{M}$, which can be efficiently queried by subsequent segments using their query vectors. After each slide, the model first retrieves a hidden state $\mathbf{A}_\text{mem}$ by querying the compressive memory $\mathbf{M}_{s-1}$:
\begin{equation}
    \mathbf{A}_{\text{mem}} = \frac{\sigma(\mathbf{q}_s)\mathbf{M}_{s-1}}{\sigma(\mathbf{q}_s)\mathbf{n}_{s-1}}
\end{equation}
where $\sigma(\cdot)$ is an element-wise nonlinear activation function (e.g., ELU$(\cdot)+1$) and $\mathbf{n}_{s-1}$ is a normalization vector (accumulated up to segment $s-1$).

Next, the compressive memory $\mathbf{M}_s $ and normalization vector $\mathbf{n}_s$ are updated using the KV entries of the current segment $s$:
\begin{equation}
\begin{aligned}
    \mathbf{M}_s &= \mathbf{M}_{s-1} + \sigma(\mathbf{k}_s)^T \left(\mathbf{v}_s - \frac{\sigma(\mathbf{k}_s)\mathbf{M}_{s-1}}{\sigma(\mathbf{k}_s)\mathbf{n}_{s-1}}\right) \\
    \mathbf{n}_s &= \mathbf{n}_{s-1} + \sigma(\mathbf{k}_s)^T \mathbf{1}_N
\end{aligned}
\end{equation}
Here, $N$ is the length of the current segment $s$. $\sigma(\cdot)$ is applied element-wise, and $\mathbf{1}_N$ is an $N \times 1$ vector of ones.

The final attention output for segment $s$, denoted $\mathbf{A}_s$, combines the standard dot-product attention output $\mathbf{A}_{\text{local}}$ (local context from the current segment) with the retrieved memory state $\mathbf{A}_{\text{mem}}$ (global context from past segments) using a learnable gating scalar $\beta \in \mathbb{R}$:
\begin{equation}
    \mathbf{A}_s = \text{sigmoid}(\beta)\odot \mathbf{A}_{\text{mem}} + (1-\text{sigmoid}(\beta))\odot \mathbf{A}_{\text{local}}
\end{equation}
As in standard multi-head attention, a final linear projection is applied to $\mathbf{A}_s$ to produce the output of the Infini-attention layer.

\section{Additional Experiments}

\subsection{Analysis of Compounding Error Evaluation Metrics}

\begin{figure}[htbp]
    \centering
    \begin{subfigure}[b]{\textwidth}
        \includegraphics[width=\textwidth]{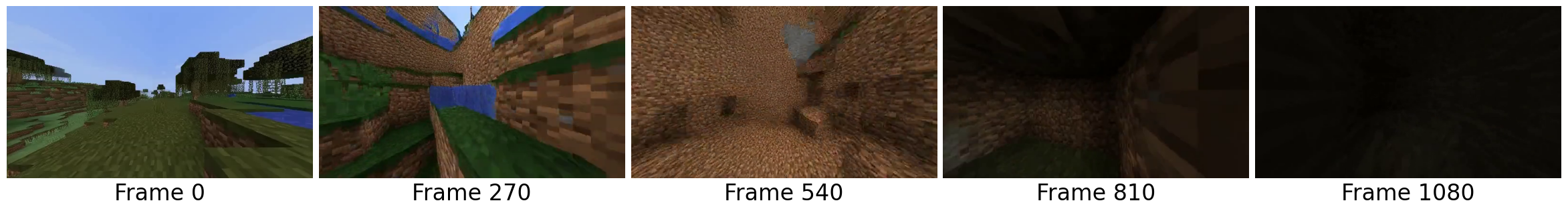}
    \end{subfigure}
    \vspace{0.5cm}     
    \begin{subfigure}[b]{\textwidth}
        \includegraphics[width=\textwidth]{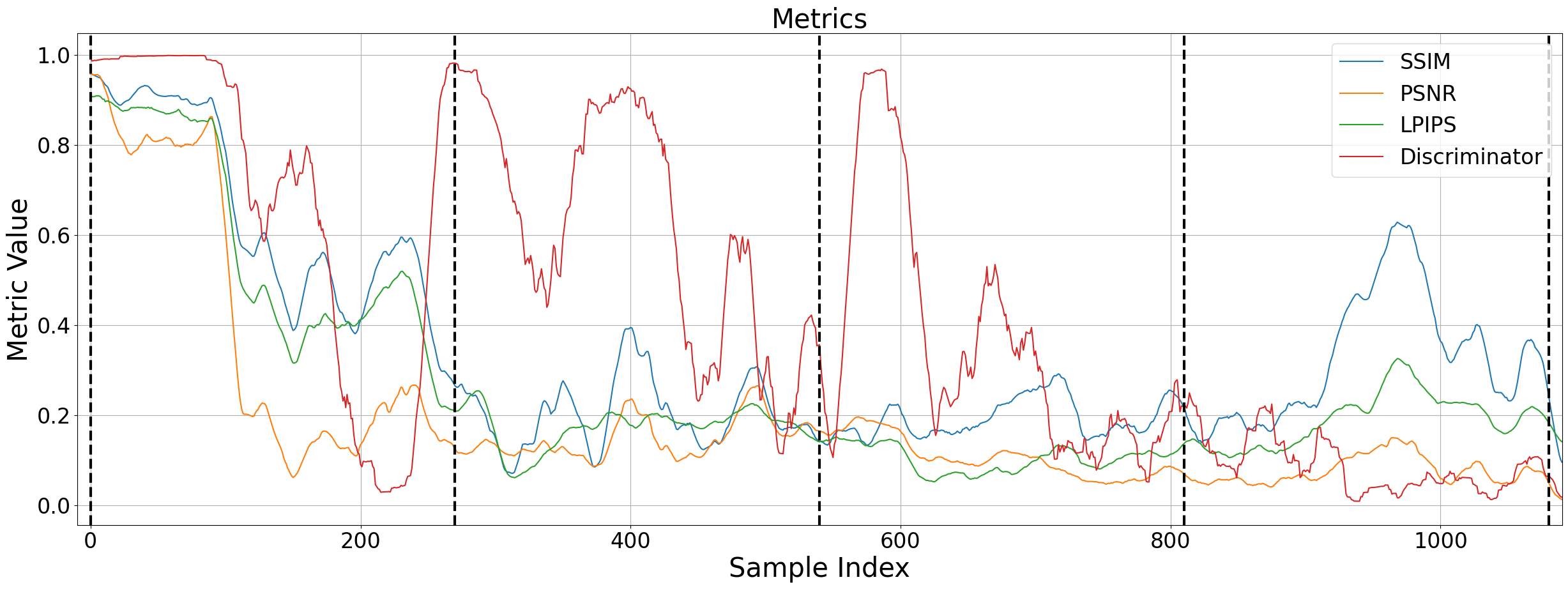}
    \end{subfigure}
    \caption{Comparison of SSIM, PSNR, LPIPS, and discriminator metrics. All metrics are normalized to the [0,1] range, where higher values indicate better performance for all scores. The discriminator score can accurately capture variations in generated image quality, while the other metrics are affected by distribution shift and fail to properly reflect compounding errors.}
    \label{fig:discriminator}
\end{figure}

Traditional metrics like SSIM, PSNR, and LPIPS measure pixel-level or feature-level differences between original and generated images. However, these metrics lose effectiveness when the generated video sample deviates significantly from the original video sample, especially for the compounding error evaluation, even if they falls in the same distribution and are visually reasonable. As shown in Figure~\ref{fig:discriminator}, we normalize all metrics to a 0-1 scale where higher values indicate better generation quality (with SSIM score flipped). While all metrics perform well on the initial frame (index 0), assigning high scores to ground truth, their values begin to deteriorate after frame 100.

To address this limitation, we developed a discriminator-based evaluation metric. We train a discriminator using 1000 videos from the vanilla DF model (window size 20), with each video containing 1000 frames. This yielded a dataset of $10^6$ ground truth frames and $10^6$ generated frames as fake ones. We implemented the discriminator as a binary classifier using a lightweight architecture with 4 ResNet blocks. Too large discriminator architecture will lead to less meaningful discriminative signals. Each block contains two convolutional layers with batch normalization and activation functions. This design provides discriminative outputs while maintains computational efficiency.

As shown in Fig.~\ref{fig:discriminator}, the decrease of discriminator value faithfully reflects the distortions in generated images, while other metrics decline for two reasons: image quality degradation and distributional shift from the original video. This shift prevents traditional metrics from accurately assessing generation performance in terms of the compounding error. For instance, while the 270th frame shows significantly better generation quality than the 1080th frame, SSIM, PSNR, and LPIPS assign similar scores to both. This indicates that the distribution shift has become the dominant factor in lowering the metric scores, making these metrics unreliable for evaluating compounding error in long-range video generation.

Unlike traditional metrics, the discriminator's evaluation remains robust to distribution shifts since it doesn't depend on the original image, but rather depending only on the distortion of the generated images. This makes the discriminator score a more reliable metric for evaluating compounding errors in this case. However, the discriminator approach has several limitations. First, training requires sampling from a pre-trained diffusion model, which incurs computational overhead. Second,  the training of the discriminator heavily depends on human judgment. We find that even a shallow ResNet architecture can effectively distinguish between ground truth and generated images. This suggests that an overly complex model might assign uniformly low scores to all generated content, making the discriminator metric less meaningful to look at. Finally, the discriminator shows limited generalization capability. When evaluating videos generated by new methods or datasets, the discriminator may be deceived into assigning inappropriately high scores. Therefore we do not report the discriminator score in the main paper, and advocate more investigation into faithful evaluation of compounding error in future work.

\subsection{Vanilla Long-context Extension vs. YaRN}

\begin{figure}[htbp]
    \centering
    \includegraphics[width=0.8\linewidth]{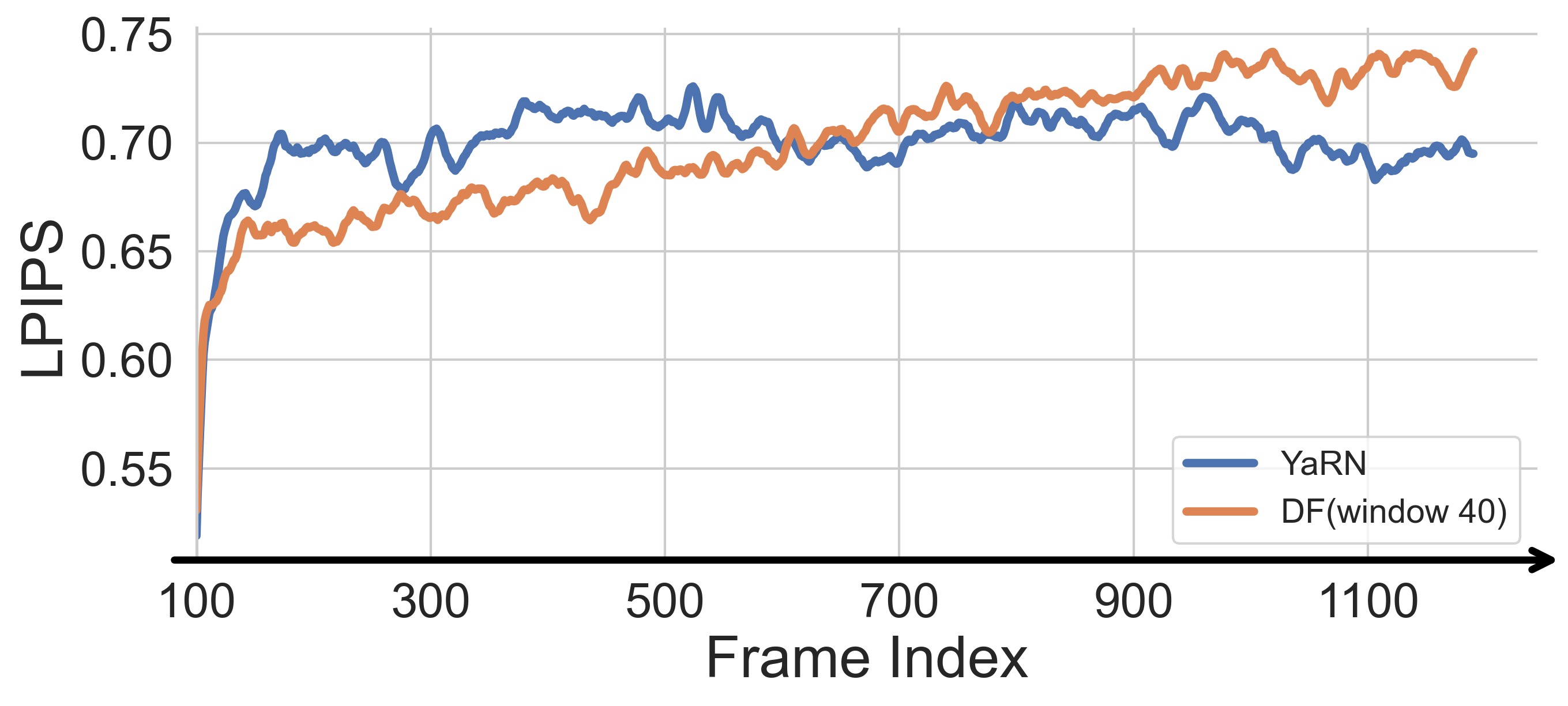}
    \caption{Comparison of vanilla long-context extension for DF model and YaRN with window length of 40 frames at inferences. Lower is better for LPIPS score.}
    \label{fig:lpips_for_yarn}
    \vspace{-2.5mm}
\end{figure}

\begin{figure}[htbp]
    \centering
    \includegraphics[width=0.8\linewidth]{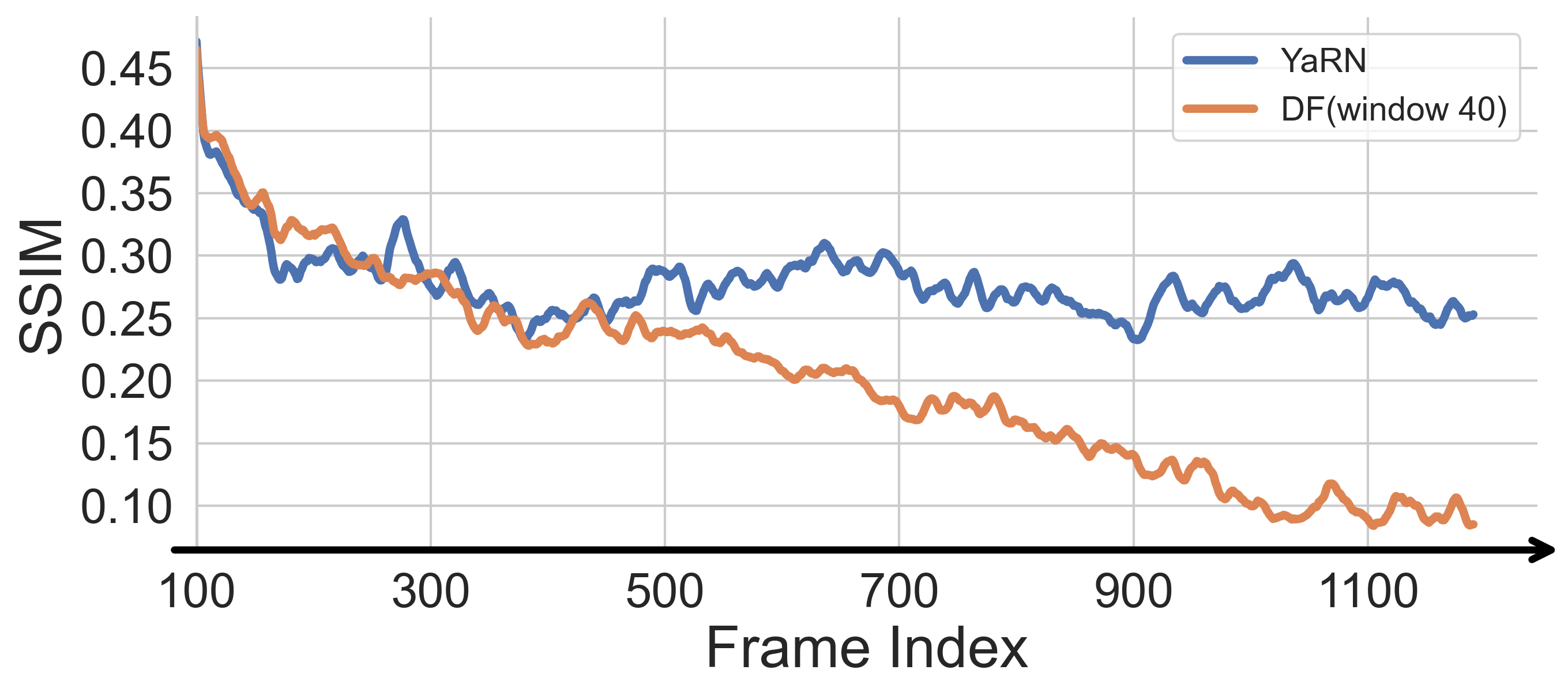}
    \caption{Comparison of vanilla long-context extension for DF model and YaRN with window length of 40 frames at inferences. Higher is better for SSIM score.}
    \label{fig:ssim_for_yarn}
    \vspace{-2.5mm}
\end{figure}

\begin{figure}[htbp]
    \centering
    \includegraphics[width=0.8\linewidth]{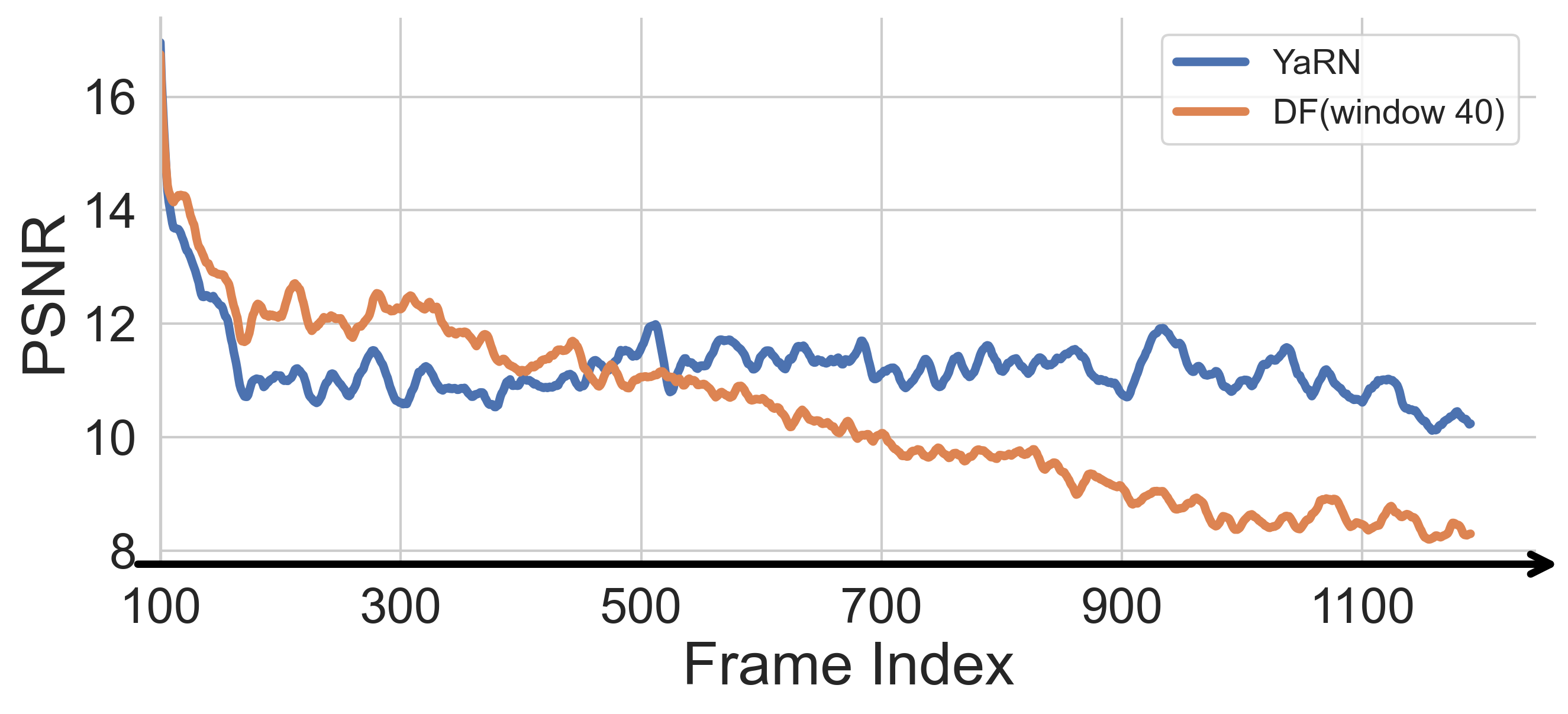}
    \caption{Comparison of vanilla long-context extension for DF model and YaRN with window length of 40 frames at inferences. Higher is better for PSNR score.}
    \label{fig:psnr_for_yarn}
    \vspace{-2.5mm}
\end{figure}

To ensure a fair comparison, we evaluate YaRN against a baseline that directly extrapolates the vanilla model's window size from 20 during training to 40 at inference, to match the inference window length as YaRN in our experiments as Sec.~\ref{gen_inst}. Evaluation of quantitative metrics LPIPS, SSIM and PSNR shown in Figures~\ref{fig:lpips_for_yarn},~\ref{fig:ssim_for_yarn}, and~\ref{fig:psnr_for_yarn} indicates that, YaRN maintains lower compounding error for long video generation (1100 frames). This demonstrates YaRN's effectiveness in extending the context window of diffusion video models to 40 frames after minimal fine-tuning. Vanilla extension of context length on DF models performs poorly due to out-of-distribution window size at inference.

\begin{figure}[htbp]
    \centering
    \includegraphics[width=0.8\linewidth]{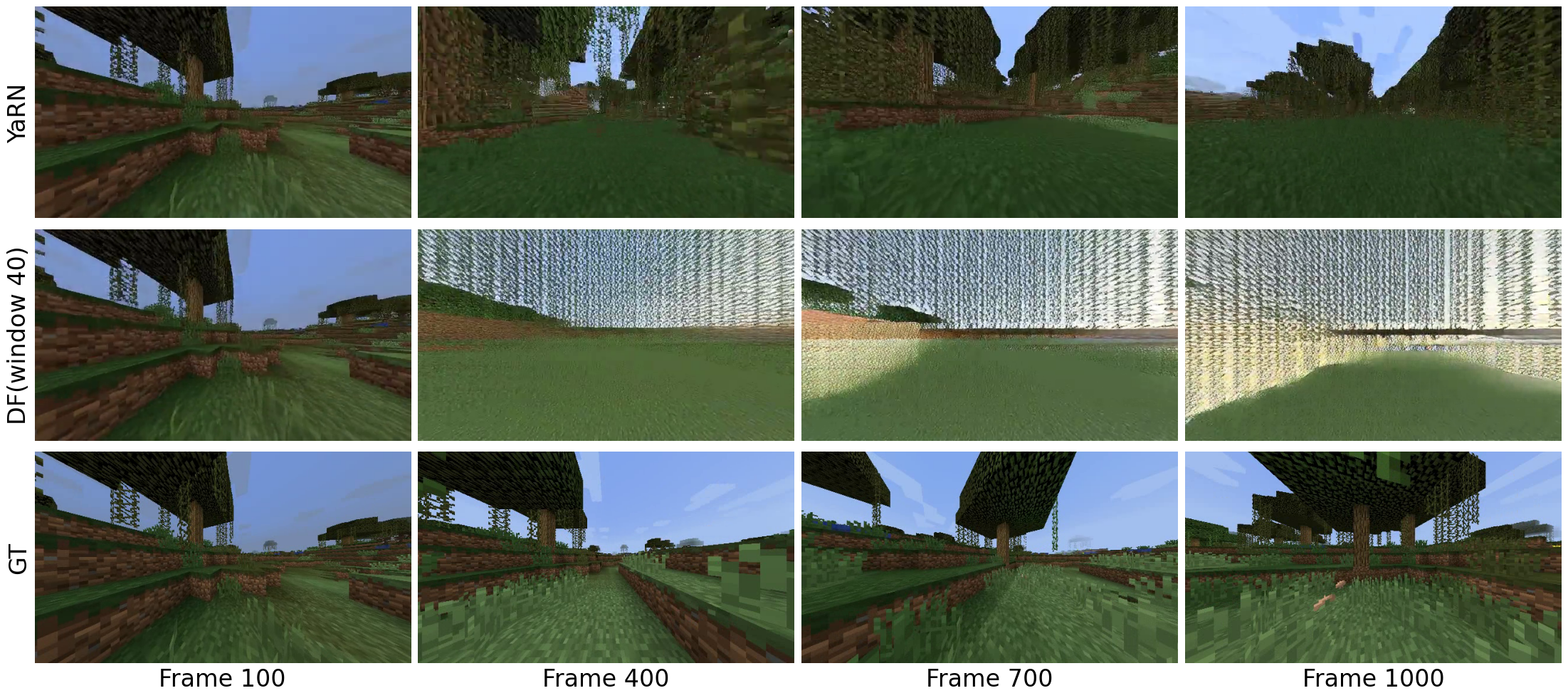}
    \caption{Visual comparison of vanilla long-context extension for DF model and YaRN. Both models are inferred with 40 frames window.}
    \label{fig:yarn vs vanilla with 40}
    \vspace{-5mm}
\end{figure}

While YaRN effectively extends the context window, its performance improvements are constrained by the inherent limitations of diffusion models in in-context learning. As demonstrated in Figure~\ref{fig:yarn vs vanilla with 40}, the model exhibits difficulties in effectively leveraging long-range dependencies, leading to suboptimal spatialtemporal consistency against the ground truth. In addition, YaRN also requires greater computational overheads during inference as it has a larger window size compared with other methods in our experiments in Sec.~\ref{gen_inst}, making it less suitable for real-time gameplay applications.

\subsection{More Discussions on Main Results}
\begin{figure}[htbp]
    \centering
    \includegraphics[width=0.8\linewidth]{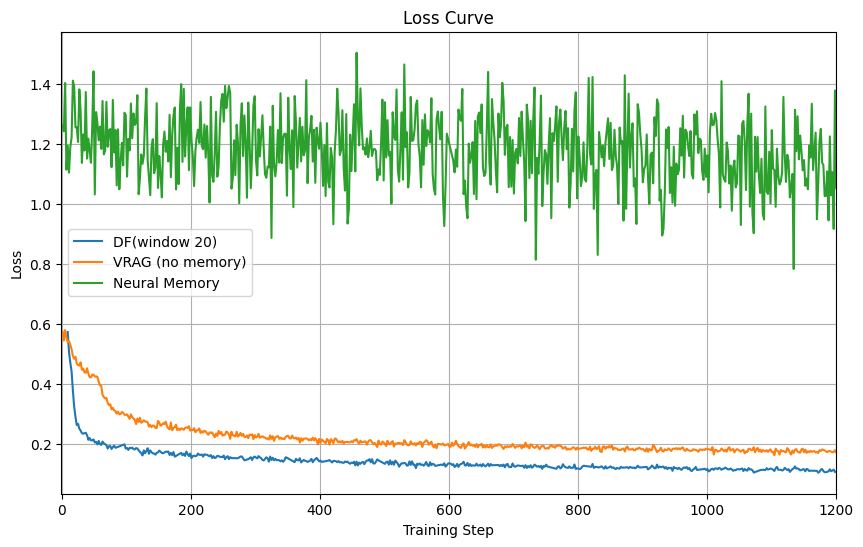}
    \caption{Training Loss Curves}
    \label{fig:loss_curve}
    \vspace{-8mm}
\end{figure}

For the main results in Sec.~\ref{sec:world coherence results} and Sec.~\ref{sec:compounding error results}, we provide more discussions here.
The Infini-attention model faces significant training challenges due to its global attention mechanism. As evidenced in Figure~\ref{fig:loss_curve}, the model struggles to converge during training. For VRAG without memory component, we incorporated global state conditioning (specifically [$x, y, z,$ yaw]) into the input. However, 
compared to the vanilla diffusion model, the training process becomes significantly more difficult. This may be due to the 
higher dimensionality and larger ranges of the spatial condition, whereas the action condition mostly consists of binary states ([0, 1]), making it harder for the model to learn and increasing perplexity.

\subsection{Predicted Global State}
In the paper, our main experiments are conducted with the access to the ground-truth global state as conditions during training and inference. However, the practical usage may require the global state to be also predicted based on historical states and actions. To ablate this effect, 
we trained a pose (global state) prediction model that takes the current frame and action as inputs and outputs the predicted pose change. The next post can be derived by adding the predicted pose change to the current pose. Its architecture consists of only a few convolutional layers and fully connected layers, with a very small inference time overhead. At evaluation, we apply this trained predictor to predict the global state at next step, and generate videos based on the predicted global state. Following the same setting as in Sec.~\ref{sec:world coherence results}, the experimental results (for 300 frames prediction) are summarized in Table~\ref{tab:predictor}.

\begin{table}[htbp]
    \centering
    \begin{tabular}{l|ccc}
        \hline
         Method & SSIM$\uparrow$ &PSNR$\uparrow$ & LPIPS$\downarrow$ \\
        \hline
         DF20 & 0.466& 16.643 &0.538 \\
         VRAG (predicted pose) & 0.500 & \textbf{17.116} & \textbf{0.506} \\
         VRAG & \textbf{0.506} & 17.097 & \textbf{0.506} \\
        \hline
    \end{tabular}
    \caption{Ablation study of replacing the ground-truth global state with predicted ones by a trained pose predictor.}
    \label{tab:predictor}
\end{table}

As shown in the table, the evaluation results are nearly identical with or without the pose prediction, since the pose prediction is a relatively simple task compared with the video generation. This proves the feasibility of using the predicted global state without significant video performance degradation.

\subsection{Memory and Time Overhead}
We also compare the memory usage and inference time of VRAG against several baselines: diffusion forcing with 10 and 20 context frames, and YaRN with 40 context frames.

\begin{table}[htbp]
    \centering
    \begin{tabular}{l|cccc}
        \hline
         Method & DF10 & DF20 & YaRN & VRAG \\
        \hline
        Context length (Frame) & 10 & 20 & 40 & 20 \\
        Memory usage (MB) & 4420 & 4448 & 4543 & 4452 \\
        Inference Time (min) & 9 & 12 & 23 & 12 \\
        \hline
    \end{tabular}
    \caption{Memory usage and inference-time of different methods}
    \label{tab:memory_inference_time}
\end{table}

As demonstrated in the Table~\ref{tab:memory_inference_time}, VRAG's GPU memory usage and inference time overhead are nearly identical to DF20. The inference time is derived for autoregressive generation over 600 frames. Meanwhile, the computation for the retrieval operations can be entirely performed on the CPU, with its memory footprint being only num\_frame $\times$ action\_dim $\times$ 4 Bytes = 9.4 KB in our experiments, which is almost negligible.

In summary, VRAG incurs almost no additional inference-time overhead compared to standard diffusion forcing. The memory and computational cost introduced by the retrieval mechanism are negligible, as it only involves similarity calculations between a set of vectors.


\chapter{World Model With Memory\label{ch:world_model_mem}}
\begin{center}
\begin{quote}
This section is based on paper ``\textit{Recurrent Autoregressive Diffusion: Global Memory Meets Local Attention}''~\cite{chen2025recurrent} written in collaboration with Taiye Chen, Anjian Li, Christina Zhang, Zeqi Xiao, Yisen Wang and Chi Jin.
\end{quote}
\end{center}
\section{Introduction}

World models have attracted considerable interest from the research community for their pivotal roles in data synthesis, model-based planning, simulation, and beyond. Leveraging the recent great progress in generative models, video diffusion models have become one of the most promising approaches for efficient and scalable world models. Opposed to representation world models \cite{bardes2024revisitingfeaturepredictionlearning, assran2025vjepa2selfsupervisedvideo} that learn latent representations of the environment, video diffusion models \cite{videoworldsimulators2024, parkerholder2024genie2, genie3} directly achieve world modeling in high-dimensional video pixel space.

Despite remarkable advancements, existing video world models face significant challenges in maintaining spatiotemporal consistency. For instance, models like Oasis \cite{oasis2024}—centered on Minecraft gameplay—and foundational models such as Cosmos \cite{nvidia2025cosmosworldfoundationmodel} both struggle with severe forgetting issues. This difficulty primarily stems from the inherently limited attention window of diffusion transformer (DiT) \cite{peebles2023scalable} architecture. Given the low information density of video data, tokenizing video sequences often results in context lengths that quickly exceed these attention limits. Consequently, frames outside the active attention window are effectively disregarded, resulting in visible temporal and spatial inconsistencies. While recent research has explored alternative approaches to mitigate this, such as leveraging 3D representations \cite{marble2025} to bolster spatial-temporal consistency, they usually lack the interactiveness and scalability of pixel-based video diffusion models.

A central challenge in applying DiT to long video generation is the ineffective compression of historical context through key-value (KV) caching in standard attention mechanisms. As video sequences extend, this approach results in scalability bottlenecks, since the memory and computation requirements grow proportionally to sequence length. Recent efforts have sought to address this limitation by incorporating recurrent neural networks (RNNs)—including Mamba~\cite{mamba, mamba2, wang2025lingen, po2025longcontextstatespacevideoworld} and Test-Time Training (TTT) approaches~\cite{sun2025learninglearntesttime, zhang2025testtimetrainingright}—within the DiT framework. However, these integrations often suffer from two major issues: (1) chunk-wise autogressive processing, common in TTT-style models, relies heavily on hidden state propagation, causing the model to lose direct access to the dense contextual information present in recent frames and resulting in pixel-level inconsistencies across chunk boundaries; and (2) the introduction of recurrency breaks the parallelizable nature of attention during training, thereby increasing computational inefficiency and exacerbating the gap between training and inference procedures.

In this work, we introduce a memory mechanism into DiT architecture for video world models, with several novel designs to overcome the above challenges. We advocate for an explicit frame-wise autoregressive rollout, which better preserves contextual integrity, and introduce a hidden-state prefetch mechanism to recover parallelism in attention computation. Building on these insights, we propose Recurrent Autoregressive Diffusion (RAD)—a unified framework that addresses both the limitations of historical compression and the efficiency constraints imposed by recurrency in long video diffusion.


In this paper, our main contributions are as follows:
\begin{itemize}
    \item We introduce \textbf{Recurrent Autoregressive Diffusion (RAD)}, a unified framework for long-term video generation with global memory and local attention, and systematically compare different RNN architectures—including LSTM~\cite{6795963}, Mamba2, and TTT—within autoregressive video generation framework. Our results reveal that LSTM, despite its simplicity, delivers robust performance and often surpasses more recent RNN variants on challenging long video benchmarks.
    \item We provide a comprehensive analysis of both \textbf{chunk-wise} and \textbf{frame-wise} autoregressive paradigms, demonstrating that explicit frame-wise generation with context overlap substantially enhances spatiotemporal consistency by leveraging immediate contextual information, and reduces reliance on persistent hidden state propagation—thereby mitigating pixel-level inconsistencies and improving the fidelity of long video synthesis.
    \item We design and implement a \textbf{hidden-state prefetch mechanism}, which overcomes the inherent trade-off between recurrency and training parallelism. This mechanism enables fully parallel attention computation during training while retaining the benefits of recurrent historical compression, significantly improving efficiency for large-scale, long-sequence modeling.
\end{itemize}

\section{Related work}

\paragraph{Video Diffusion Model} The remarkable success of diffusion models originated in image synthesis \cite{Rombach_2022_CVPR, ramesh2022hierarchicaltextconditionalimagegeneration} and was later extended to video generation \cite{NEURIPS2022_39235c56, opensora, opensora2, blattmann2023align, hong2022cogvideo, yang2024cogvideox}. Current state-of-the-art video diffusion models typically employ VAEs \cite{kingma2022autoencodingvariationalbayes} to map videos from the pixel level to a latent space, while the model architecture has evolved from the UNet \cite{ronneberger2015unetconvolutionalnetworksbiomedical, chen2024videocrafter2, blattmann2023stable} to diffusion transformer (DiT) \cite{peebles2023scalable}. 

\paragraph{Video World Model} World models \cite{watter2015embed, ha2018recurrent, hafner2020mastering} predict future states based on the current state and input actions. They hold broad application prospects in fields such as autonomous driving \cite{hu2023gaia, ren2025cosmosdrivedreamsscalablesyntheticdriving}, navigation \cite{bar2024navigation}, and robotic manipulation \cite{wu2024ivideogpt, azzolini2025cosmos, ge2025} and games \cite{valevski2024diffusion, oasis2024, che2024gamegen, guo2025mineworld, yu2025gamefactory}. The capability of video diffusion models to synthesize high-quality videos makes them promising candidates as video world models. Since the proposal of ``video generation models as world simulators'' by Sora \cite{videoworldsimulators2024}, a series of foundational world models have emerged—such as Genie2 \cite{parkerholder2024genie2}, Genie3 \cite{genie3}, and Cosmos \cite{nvidia2025cosmosworldfoundationmodel}—demonstrating remarkable video generation quality and interactivity. 
To achieve video world models, recent advancement improves video generation models in multiple aspects: autoregressive inference for extending video durations \cite{chen2024diffusion, huang2025selfforcingbridgingtraintest, cui2025self, Yin_2025_CVPR}, memory mechanism for improving spatiotemporal consistency \cite{chen2025learningworldmodelsinteractive, xiao2025worldmemlongtermconsistentworld, yu2025cam}, physics modeling \cite{kang2024far}, etc.


\paragraph{Diffusion Model with Memory}

While scalable DiT-based video diffusion models have shown strong performance on short clips, applying full attention to long videos incurs prohibitive computational and memory costs. To enable long‑video generation without global attention, additional memory mechanisms are required, which can be classified into \textit{context memory}, \textit{hidden‑state memory}, and \textit{weight memory}. Context memory approaches treat past frames as conditions for autoregressive (AR) prediction. With a limited context window, they either adopt recency‑biased frame selection—as in diffusion forcing \cite{chen2024diffusion, song2025historyguidedvideodiffusion} and self‑forcing \cite{huang2025selfforcingbridgingtraintest, cui2025self}—or employ similarity‑based retrieval, such as VRAG \cite{chen2025learningworldmodelsinteractive} and WorldMem \cite{xiao2025worldmemlongtermconsistentworld}. Hidden‑state memory methods compress historical information into recurrent states, e.g., by inserting Mamba layers before attention \cite{po2025longcontextstatespacevideoworld}. However, the recurrent structure disrupts the temporal parallelism of the original DiT, adding extra computational overhead. Weight memory techniques use inner‑loop losses to update specific weights as memory while sliding over small chunks, as seen in Test‑Time Training (TTT) blocks \cite{11095233, zhang2025testtimetrainingright}. Since adjacent attention windows do not overlap, the updated weights must fully encode all necessary history from prior chunks, imposing high demands on memory capacity for storage and retrieval. In contrast, our approach performs efficient memory compression only for history beyond the context window. It bridges context memory and hidden‑state memory through frame‑wise sliding during autoregressive generation, balancing efficiency and long‑range consistency.

\section{Preliminaries}

\subsection{Recurrent Neural Networks}\label{sec:preli_rnn}

\paragraph{RNN}
Assuming $x_t \in \mathbb{R}^{D}$ denotes the input of time $t$, 
The general form of a recurrent neural network (RNN) can be described as:
\begin{align*}
    h_t &= f(h_{t-1}, x_t, y_{t-1}; \theta) \\
    y_t &= g(h_t, x_t)
\end{align*}
where $h_t$ is the hidden state at time $t$, $x_t$ is the input, $y_t$ is the output, $f(\cdot)$ is a parameterized nonlinear function (typically involving affine transformation and activation), $\theta$ represents all trainable parameters, and $g(\cdot)$ maps the hidden state to the output.

\paragraph{LSTM}
\begin{align*}
    \begin{bmatrix}
        f_t \\
        i_t \\
        o_t \\
        g_t
    \end{bmatrix}
    &=
    \begin{bmatrix}
        \sigma\\
        \sigma\\
        \sigma\\
        \tanh
    \end{bmatrix}
    W
    \begin{bmatrix}
        y_{t-1} \\
        x_t
    \end{bmatrix}
    \\
    C_t &= f_t \odot C_{t-1} + i_t \odot g_t \\
    y_t &= o_t \odot \tanh(C_t)
\end{align*}

where $f_t, i_t, o_t, g_t$ denote forget gate, input gate, output gate, candidate cell state at time $t$, $\sigma, tanh$ denote sigmoid activation function and hyperbolic tangent activation function, $W$ denote weight and bias matrices for respective gates. The vectors $C_t$ and $y_t$ denote the cell state and output at time $t$, respectively, encapsulating all compressed memory information accumulated up to and including time step $t$.

\paragraph{Other RNNs} Apart from LSTM, Mamba and TTT are two popular recurrent architectures that update hidden states over time. Their key difference lies in how the new state is computed. LSTM updates both its hidden state and cell state using three sources of information—the current input $x_t$, the previous hidden output $y_{t-1}$, and the previous cell state $C_{t-1}$. In contrast, Mamba and TTT follow a simpler RNN-style update that depends only on the current input and the previous hidden state, i.e., $h_t = f(h_{t-1}, x_t)$. More details are provided in Appendix. 



\subsection{Video Diffusion model}

\paragraph{Latent Video Diffusion Model}
We adopt a latent video diffusion model~\cite{blattmann2023stable} that first encodes pixel space into a latent representation $\boldsymbol{z} = \mathcal{E}(\boldsymbol{x})$ using a pretrained variational autoencoder (VAE). The forward process gradually adds Gaussian noise to the latent according to a variance schedule $\{\beta_t\}_{t=1}^T$:
\begin{equation}
    q(\boldsymbol{z}_t|\boldsymbol{z}_{t-1}) = \mathcal{N}(\boldsymbol{z}_t; \sqrt{1-\beta_t}\boldsymbol{z}_{t-1}, \beta_t\mathbf{I})
\end{equation}

The model learns to reverse this process by predicting the noise $\boldsymbol{\epsilon}_\theta$ at each step:
\begin{equation}
    \mathcal{L} = \mathbb{E}_{t,\boldsymbol{\epsilon},\boldsymbol{z}}[\|\boldsymbol{\epsilon} - \boldsymbol{\epsilon}_\theta(\boldsymbol{z}_t, t)\|_2^2]
\end{equation}
where $\boldsymbol{z}_t = \sqrt{\bar{\alpha}_t} \boldsymbol{z}_0 + \sqrt{1-\bar{\alpha}_t} \boldsymbol{\epsilon}$ with $\boldsymbol{\epsilon} \sim \mathcal{N}(0,\mathbf{I})$.

At inference time, we can sample new videos by starting from random noise $\boldsymbol{z}_T \sim \mathcal{N}(0,\mathbf{I})$ and iteratively denoising:
\begin{equation}
    \boldsymbol{z}_{t-1} = \frac{1}{\sqrt{\alpha_t}}(\boldsymbol{z}_t - \frac{\beta_t}{\sqrt{1-\bar{\alpha}_t}}\boldsymbol{\epsilon}_\theta(\boldsymbol{z}_t,t)) + \sigma_t\boldsymbol{\epsilon}
\end{equation}
where $\alpha_t = 1-\beta_t$ and $\bar{\alpha}_t = \prod_{s=1}^t \alpha_s$.
The final latent sequence $\boldsymbol{z}_0$ is decoded back to pixel space using the decoder $\mathcal{D}$ to obtain the generated video.

\paragraph{Diffusion Forcing}
To enable long video generation, we apply the Diffusion Forcing~\cite{chen2024diffusion} technique. During training, we randomly add noise to each frame in the entire input video sequence according to the diffusion schedule: $z^i_t = \sqrt{\bar{\alpha}_t} z^i_0 + \sqrt{1-\bar{\alpha}_t} \epsilon^i, \epsilon^i \sim \mathcal{N}(0,\mathbf{I})$, where $z^i_t$ represents the noised latent of the $i$-th frame, and the training objective for action-conditioned autoregressive video models become:
\begin{align*}
    \mathcal{L}_\text{DF} &= \mathbb{E}_{[t],\boldsymbol{\epsilon},\mathbf{z},a}[\|\boldsymbol{\epsilon} - \boldsymbol{\epsilon}_\theta(\mathbf{z}_{[t]}, [t], \boldsymbol{a})\|_2^2] \\ \boldsymbol{\epsilon}&=\{\epsilon^i\}_{i=1}^L, \mathbf{z}_{[t]}=\{z^i_t\}_{i=1}^L
\end{align*}
where $[t]$ is vector of $L$ timesteps with different $t\in[T]$ for each frame and $\boldsymbol{a}$ is an action sequence $\boldsymbol{a} \in \mathbb{R}^{L \times A}$. The noise prediction model $\boldsymbol{\epsilon}_\theta$ conditioned on both the action sequence $\boldsymbol{a}$ and noised frames $\mathbf{z}_{[t]}$.

\section{Methodology}
\begin{figure*}
    \centering
    \includegraphics[width=0.8\linewidth]{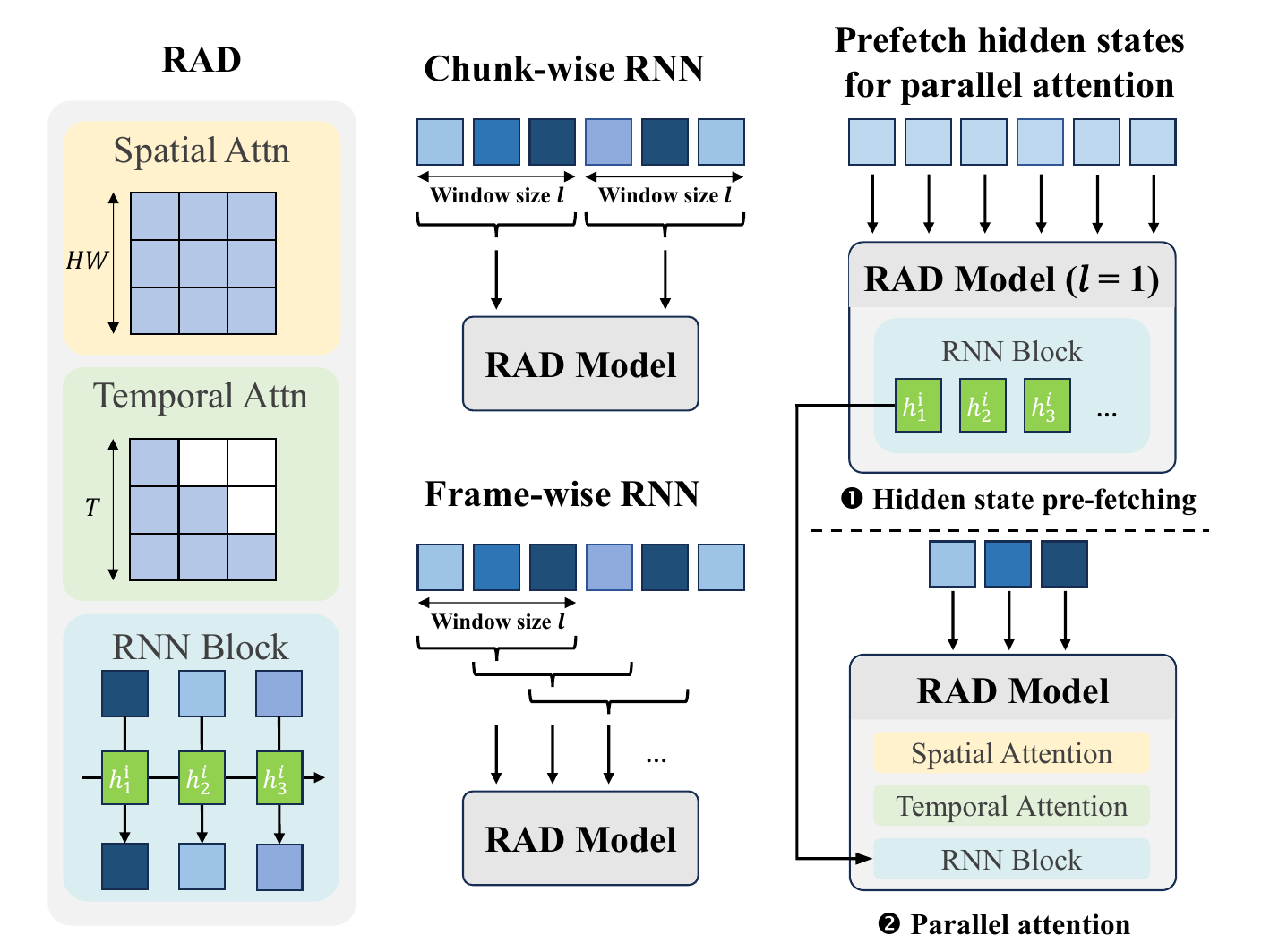}
    \caption{Training paradigm for Recurrent Autoregressive Diffusion with global memory and local attention: The model has three components in each DiT block, including spatial attention, temporal attention and RNN memory block. It supports both chunk-wise and frame-wise autoregressive generation with different attention mechanisms. For efficient training of frame-wise RNN, we (1) pre-fetch the hidden states from clean sample sequence to enable parallel attention computation across entire long sequences, and (2) conduct diffusion model forward in standard manner to get diffusion loss. This improves efficiency and fidelity for large-scale, long-sequence video modeling. $h^i_j$ is the $i$-th layer hidden state for frame index $j$.}
    \label{fig:train_paradigm}
\end{figure*}


To address the limitations of fixed-size context windows in video diffusion models, we propose the integration of \textbf{global memory with local attention} mechanisms. This approach enables the model to effectively capture long-term dependencies in video sequences while maintaining high fidelity in generated frames.

\subsection{Recurrent Autoregressive Diffusion}

\begin{figure}[htbp]
    \centering
    \includegraphics[width=0.96\linewidth]{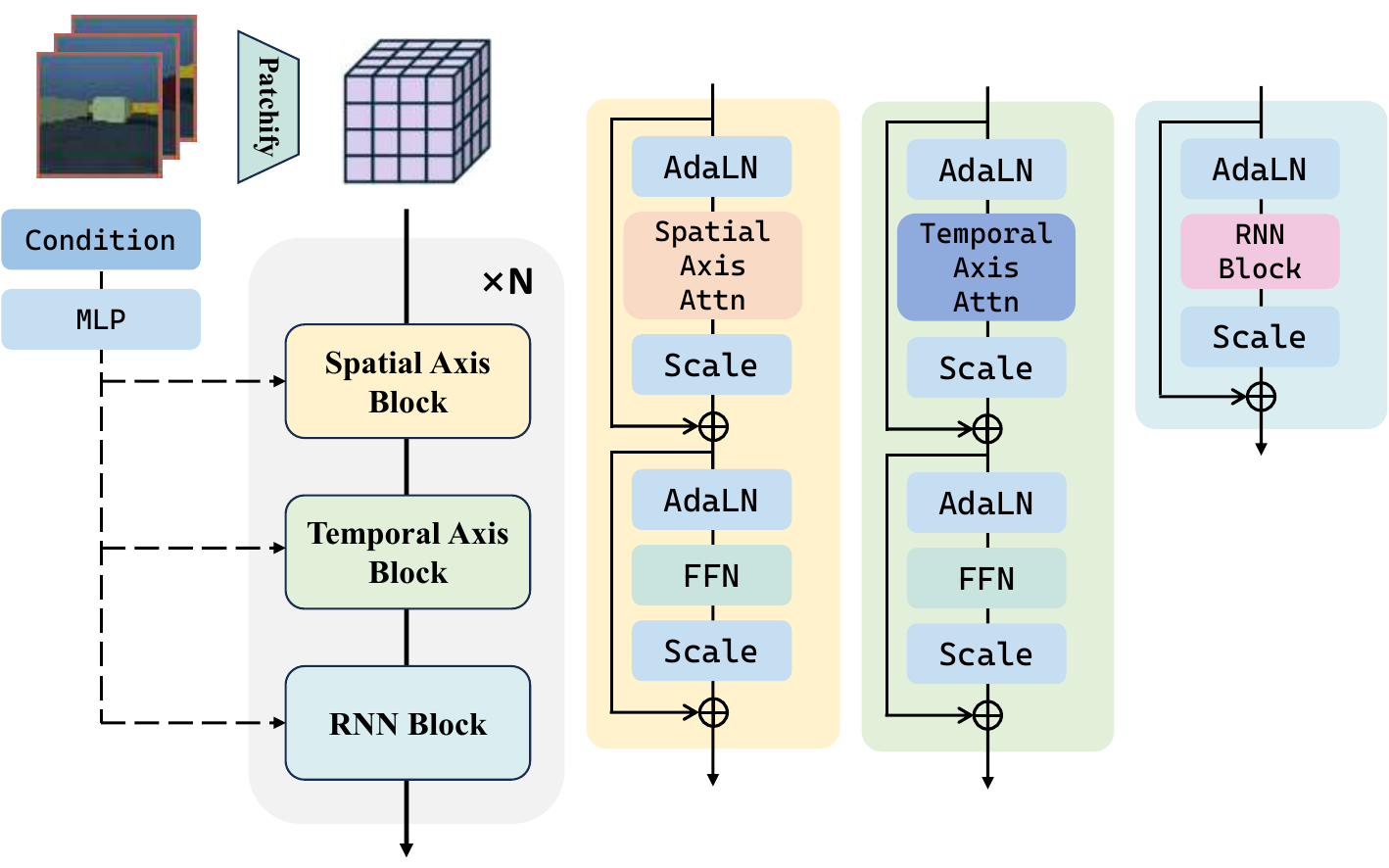}
    \caption{Recurrent Autoregressive Diffusion model architecture: The RAD model consists of $N$ DiT blocks, each of which contains three blocks: a Spatial Axis Block, a Temporal Axis Block, and an RNN Block. The timestep and action condition are processed through an MLP and then used to control each block via adaLN.}
    \label{fig:model_arch}
\end{figure}

We introduce the Recurrent Autoregressive Diffusion (RAD) model, which integrates a Recurrent Neural Network (RNN) block into the Diffusion Transformer (DiT) architecture~\cite{peebles2023scalable} to carry global memory information. The overall architecture of the RAD model is illustrated in \cref{fig:model_arch}.
Following the designs by previous work \cite{opensora, oasis2024}, we decompose the attention mechanism in our DiT into two distinct modules: Spatial Axis Attention and Temporal Axis Attention. Each layer of the DiT therefore comprises three primary components—the two attention modules as standard architecture and an additional RNN block. The Rotary Position Embedding (RoPE)~\cite{su2024roformer} is applied in both spatial and temporal dimensions, to enhance the capacity of both attention modules to capture positional dependencies. Conditioning information, specifically the timestep and action condition, is incorporated into the RAD model via adaptive Layer Normalization (adaLN). RAD also additionally applies the action conditioning on RNN blocks, apart from the attention modules. This design is verified to effectively enhance the action control and improve video fidelity, with experiments in \cref{sec:abla_cond}.
For the RNN block, we compare three alternatives including LSTM, TTT and Mamba's SSM block~\cite{11095233}, and find that LSTM performs the best in our RAD model, shown in experiment \cref{sec:experiment}. 

In the following sections we discuss two key design choices made to optimize the integration of RNNs within the DiT framework: (a). frame-wise autoregression for better context consistency (\cref{sec:chunkwise}) and (b). hidden state pre-fetching for parallel attention computation (\cref{sec:prefetch}).

\subsection{Chunk-wise and Frame-wise Autoregression}\label{sec:chunkwise}

\begin{figure}[ht]
    \centering
    \includegraphics[width=0.7\linewidth]{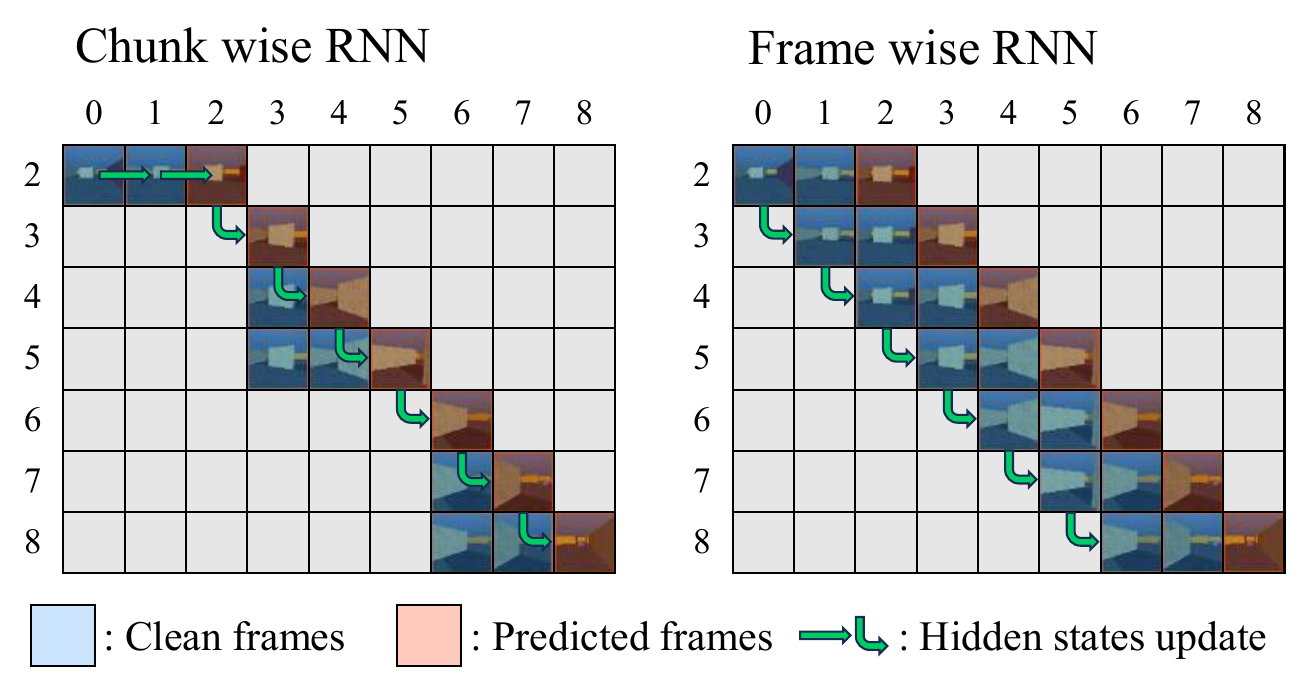}
    \caption{Comparison of effective temporal attention maps for chunk-wise and frame-wise RNNs, with chunk size 3 and 2 initial context frames. The horizontal axis of the graph represents the frame index, while the vertical axis represents the index of the currently predicted frame.}
    \label{fig:attn_map}
\end{figure}
In the RAD framework, the RNN block processes temporal information in an autoregressive manner, which can be implemented via two distinct modes: chunk-wise and frame-wise autoregression. Both approaches are seamlessly integrated with the temporal attention mechanisms of the DiT architecture, as illustrated in \cref{fig:train_paradigm} and \cref{fig:attn_map}.

For chunk-wise autoregression, the attention windows are non-overlapping. During training, the input video sequence is divided into chunks based on the model's window size. Within each chunk, local attention is computed, while global temporal dependencies are preserved by propagating the RNN hidden states across chunk boundaries. To ensure that the model operates autoregressively within each chunk, we apply a causal mask to the temporal dimension of the attention mechanism. Importantly, this approach contrasts with existing methods~\cite{11095233, zhang2025testtimetrainingright} that focus on fine-tuning diffusion models trained for full-sequence denoising. Instead, our method ensures both architectural and procedural consistency between training and inference: the model processes data during inference in exactly the same manner as during training, thereby supporting robust temporal generalization and faithful sequence modeling.

For the frame-wise RNN mode, building on Diffusion Forcing \cite{chen2024diffusion}, we employ a frame-by-frame autoregressive generation scheme. This method allows the model to fully leverage the attention mechanism for transmitting pixel-level information across frames. However, this comes at the cost of significant computational overhead when naively applying sliding window-based training. To circumvent this inefficiency, we introduce a Hidden State Pre-fetching strategy, detailed in a subsequent section. During inference, as depicted in \cref{fig:attn_map}, we slide the window one frame at a time, with only the first frame in each window responsible for updating the RNN's hidden state. This procedure directly mirrors the window size of 1 used in the hidden state pre-fetch step at training. Additionally, hidden state updates are performed only at the final step of the DDIM process, ensuring that all memory inputs consist of clean frames—thereby maintaining consistency with the training process and promoting more stable generation quality.


\subsection{Hidden State Pre-fetch for Parallel Attention}\label{sec:prefetch}

\begin{figure}
    \centering
    \includegraphics[width=0.8\linewidth]{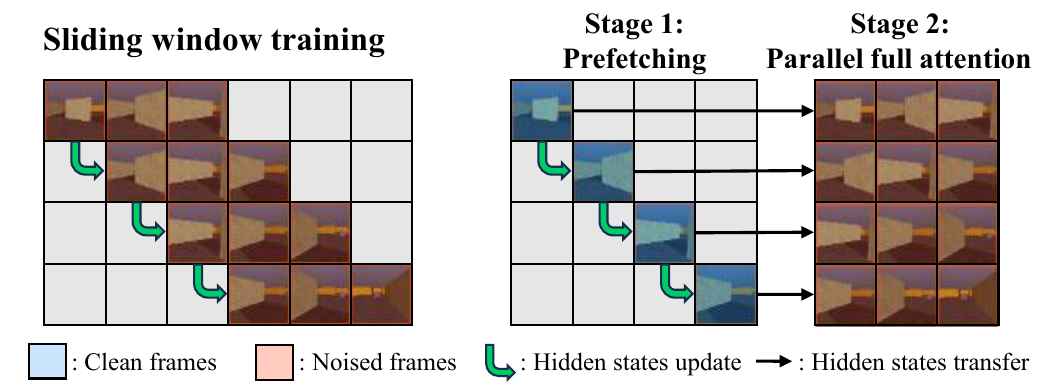}
    \caption{Comparison of standard forward (left) and hidden states prefetching process (right), for frame-wise sliding window.}
    \label{fig:prefetching}
\end{figure}

Our training strategy for frame-wise RNN is depicted in the right panel of \cref{fig:train_paradigm} and \cref{fig:prefetching}. A well-known limitation of RNNs is their inherent sequential dependence: each timestep’s output is tightly linked to the previous hidden state. This strict temporal recursion inhibits parallelization and considerably slows training, posing a particular challenge for long sequences and large-scale datasets. When RNNs are further integrated with attention mechanisms, the subsequent attention computation needs to follow the same recurrence along the input sequence during training, resulting in excessive computational bottlenecks. It becomes expensive to maintain a fully sequential temporal loop during training.

To mitigate these challenges, we employ a \textbf{hidden state pre-fetching} scheme that partially decouples the RNN from the attention modules. In this framework, the RNN maintains its temporal recurrence, but the attention operations can proceed in parallel, significantly improving training efficiency. Without this technique, a frame-wise sliding with window size $l$ over a sequence length $L$ requires $(L-l+1)$ times sequential attention computation, while parallel attention only conducts once at training in our case.  Another pivotal benefit of this approach is that only clean frames, rather than noised frames from the diffusion process \cite{po2025longcontextstatespacevideoworld}, are used as inputs to RNN memory during prefetching, a design choice that accelerates training convergence (see \cref{sec:clean_mem} for empirical analysis).

Concretely, for frame-wise autoregression, we first apply the RAD model with a window size of 1 across the clean frames of each training sequence. This step independently computes all RNN hidden states $\{h^d_t\}^{d\in [D]}_{t\in[T]}$ across frames in sequence and DiT layers, where each $h^d_t$ aggregates contextual information up to $t$-th frame and $d$-th DiT layer. Assuming the previous hidden state stacks sufficient context information, the pre-fetched hidden states are equivalent to the standard DiT with RNN by sliding-window. However, after pre-fetching all hidden states, the attention modules within each window—despite the sequential RNN update—can now be processed in parallel. By setting the prefetched hidden states for RNN layers at corresponding positions, we can compute the attention modules in RAD with a normal window size as usual to get the diffusion loss for the entire sequence. In our experiments, we do not calculate the diffusion losses for all sub-sequences (length $l$) in given sample (length $L$), but randomly sample partial of them to reduce computation cost.





\section{Experiments}
\label{sec:experiment}

\subsection{Datasets and Evaluation Protocol}\label{sec:dataset}


\paragraph{Maze Dataset} For our small-scale experimental setting, we employ the Memory Maze dataset~\cite{pasukonis2022memmaze}, which consists of approximately 30000 training videos, each depicting agent navigation within a $15 \times 15$ maze environment and comprising 1000 frames. For the Maze Dataset, all models were trained from scratch during the training process. For evaluation, we collect an additional set of 200 maze videos, with frame counts ranging from 100 to 300. In each sequence, the agent traverses from a designated start position to a target location and then returns along the same path. We define the initial 60\% of frames—corresponding to the outbound trajectory—as the contextual input, while the remaining 40\% serve as the prediction targets for model evaluation. This setup is specifically designed to rigorously test the model's capacity for long-term memory and context utilization.  The aerial visualization of the maze data can be found in the Appendix.

\paragraph{Minecraft Dataset} For large-scale experiments, we utilize the MineRL~\cite{guss2019minerl} to generate 20000 training sequences following the protocols in VRAG \cite{chen2025learningworldmodelsinteractive}. Each video contains 1200 frames. For evaluation, we collect 60 sequences incorporating distinct action patterns, such as rotation in place, which are intended to probe the model's memory capabilities under diverse behaviors. To ensure fair comparison, we first train a standard diffusion forcing model on the entire 20000-sample training set. Subsequently, this pretrained base model is fine-tuned with different RAD-RNN architectures and training paradigms, each for the same 5000 optimization steps. This procedure enables a controlled assessment of architectural and methodological differences under consistent pretraining conditions.

\paragraph{Evaluation Metrics} 
We employ three widely adopted metrics to quantitatively evaluate model performance: Structural Similarity Index (SSIM)~\cite{wang2004image}, which assesses spatial consistency in generated frames; Peak Signal-to-Noise Ratio (PSNR), which measures pixel-level reconstruction fidelity; and Learned Perceptual Image Patch Similarity (LPIPS)~\cite{zhang2018unreasonable}, which evaluates perceptual similarity. Notably, SSIM—when directly comparing generated outputs to ground truth—places a stronger emphasis on the preservation of memory and spatial-temporal consistency, whereas PSNR and LPIPS are more indicative of overall frame quality.

\subsection{Maze Results}\label{sec:maze_results}

\begin{table}[ht]
    \centering
    \begin{minipage}{0.48\textwidth}
        \centering        
        \caption{Experiment results on Maze Dataset: chunk-wise (``-c'') and frame-wise (``-f'') autoregressive modes}
        \resizebox{\textwidth}{!}{
        \begin{tabular}{c|c|ccc}
        \toprule
            Model & RNN Type & PSNR $\uparrow$ & SSIM $\uparrow$ & LPIPS $\downarrow$ \\
            \midrule
            DF & None & 14.73 & 0.32 & 0.51 \\
            \midrule
            \multirow{3}{*}{RAD} & Mamba2-c & 13.80 & 0.31 & 0.58 \\
            & TTT-c & 14.36 & 0.35 & 0.53 \\
            & LSTM-c & \textbf{15.64} & \textbf{0.43} & \textbf{0.47} \\
            \midrule
            \multirow{3}{*}{RAD} & Mamba2-f & 15.35 & \textbf{0.41} & 0.51 \\
            & TTT-f & \textbf{15.50} & \textbf{0.41} & 0.52 \\
            & LSTM-f & \textbf{15.50} & \textbf{0.41} & \textbf{0.45} \\
        \bottomrule
        \end{tabular}
        }
        \label{tab:rnn_maze}
    \end{minipage}
    \hfill
    \begin{minipage}{0.48\textwidth}
        \centering
        \caption{Experiment results on Minecraft dataset: chunk-wise (``-c'') and frame-wise (``-f'') autoregressive modes}
        \resizebox{\textwidth}{!}{
        \begin{tabular}{c|c|ccc}
        \toprule
            Model & RNN Type & PSNR $\uparrow$ & SSIM $\uparrow$ & LPIPS $\downarrow$ \\
            \midrule
            DF & None & 15.65 & 0.45 & 0.53 \\
            \midrule
            \multirow{3}{*}{RAD} & Mamba2-c & 12.72 & 0.33 & 0.63 \\
            & TTT-c & 14.04 & 0.38 & 0.56 \\
            & LSTM-c & \textbf{14.24} & \textbf{0.39} & \textbf{0.55} \\
            \midrule
            \multirow{3}{*}{RAD} & Mamba2-f & 16.70 & \textbf{0.46} & 0.47 \\
            & TTT-f & \textbf{16.72} & \textbf{0.46} & 0.47 \\
            & LSTM-f & 16.59 & \textbf{0.46} & \textbf{0.46} \\
        \bottomrule
        \end{tabular}
        }

        \label{tab:rnn_mc}
    \end{minipage}
\end{table}

\paragraph{RNN Methods}

We compare LSTM, Mamba2, and TTT within the RAD architecture as described in \cref{sec:chunkwise}, trained under identical settings for three epochs. \Cref{tab:rnn_maze} summarizes the results for both chunk-wise (``-c'') and frame-wise (``-f'') autoregressive modes.

In the chunk-wise setting, LSTM delivers the strongest performance across all metrics, improving notably over both the Diffusion Forcing (DF) baseline and the other recurrent variants. Mamba2-c and TTT-c, in contrast, perform worse than the baseline in PSNR and LPIPS, which focus more on image quality. This behavior reflects the burden placed on the recurrent module: without overlapped attention, all pixel-level continuity between chunks must be carried through hidden states or memory weights alone. LSTM benefits structurally from its separation of short-term memory (previous output $y_{t-1}$) and long-term memory (cell state $C_{t-1}$), which aligns well with this requirement. Mamba2 and TTT—designed for global compression rather than fine-grained pixel transport—struggle in comparison.

In the frame-wise setting, all recurrent variants perform similarly. With a sliding step of 1, consecutive frames share an attention window, allowing pixel-level information to propagate directly through attention rather than via hidden states. This eliminates the main failure mode of Mamba2-c and TTT-c, enabling Mamba2-f, TTT-f, and LSTM-f to reach comparable quality with only small metric differences. Correspondingly, LSTM’s structural advantage is diminished, as its explicit short-memory $y_{t-1}$ pathway becomes less necessary when attention already provides strong local continuity.

\paragraph{Frame-wise vs. Chunk-wise Autoregression}

The cross-paradigm comparison in \cref{tab:rnn_maze} highlights how the autoregressive design dictates the relative strength of each recurrent architecture.

In the \textbf{chunk-wise} mode, the absence of local cross-chunk attention forces hidden states to serve as the sole channel for transmitting pixel-level information. This creates excessive demands on its capacity for information storage and retrieval with the memory mechanism: it must encode both global memory and the local scene details needed to maintain visual consistency. Under this constraint, architectural differences become pronounced. Mamba2 and TTT underperform because they are not optimized to shuttle high-frequency information through memory alone, whereas LSTM’s disentanglement of local (via $y_{t-1}$) and global (via $C_{t-1}$) information is highly compatible with the chunk-wise sliding training and inference paradigm. As a result, LSTM-c achieves the strongest performance.

In the \textbf{frame-wise} mode, attention spans all consecutive frames, restoring direct pixel-level communication. Therefore, Hidden states focus mostly on global information, reducing the need for local detail retention. Once this burden is lifted, Mamba2 and TTT improve substantially and converge in performance with LSTM. The recurrent pathway in LSTM becomes partly redundant, which explains the disappearance of its earlier advantage.

Overall, the results indicate that the suitability of an RNN for autoregressive diffusion depends strongly on the temporal granularity of attention: when attention cannot bridge boundaries (chunk-wise), architectures with explicit local–global separation excel; when attention is continuous (frame-wise), architectural differences matter far less. We argue that hidden states could focus more on global information, while high-frequency temporal local information ought to be transmitted primarily through local attention.

\subsection{Minecraft Results}\label{sec:minecraft_results}

\begin{figure*}[htbp]
    \centering
    \includegraphics[width=0.9\linewidth]{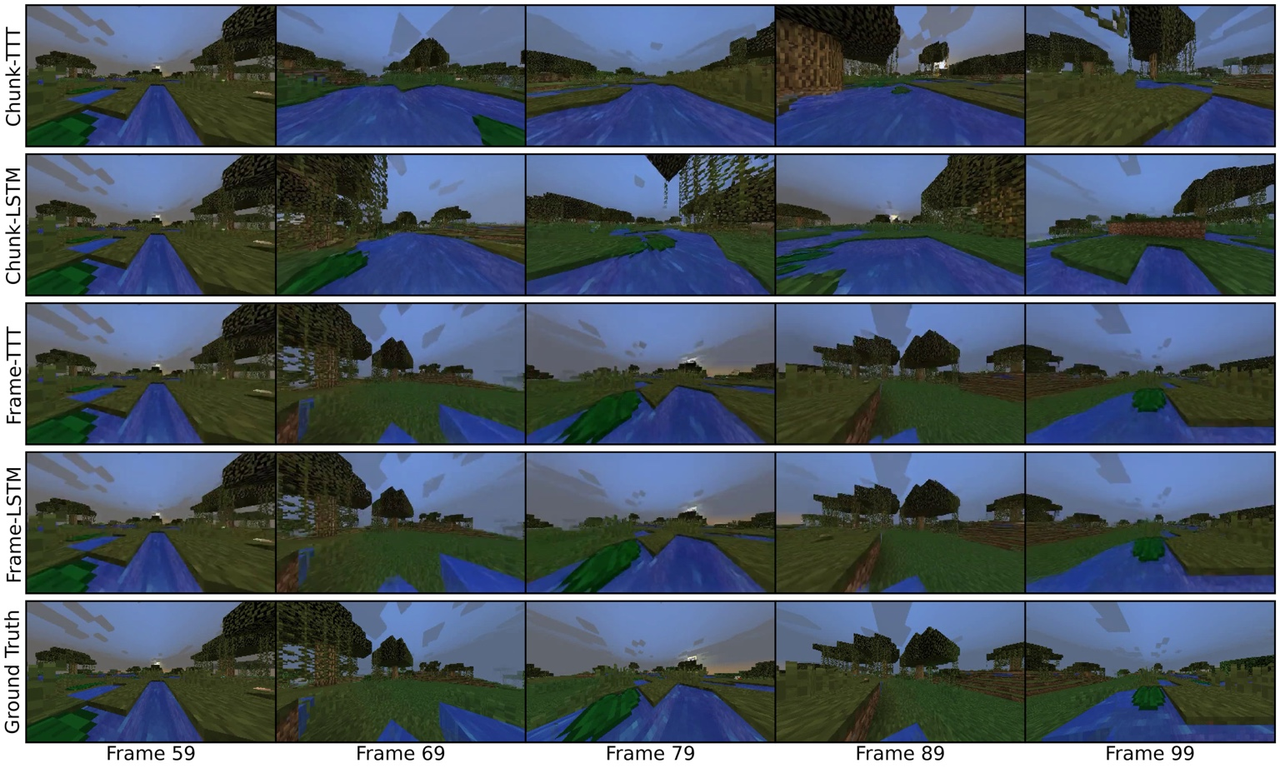}
    \caption{Visualization results on Minecraft dataset. Frame-wise RNN can effectively memorize and reconstruct scenes, maintaining consistency with the ground truth. In contrast, due to the lack of overlap attention, Chunk-wise RNN is unable to reconstruct historical scenes in datasets with higher information density}
    \label{fig:mc_vis}
\end{figure*}

\paragraph{Frame-wise vs. Chunk-wise Autoregression}
\Cref{tab:rnn_mc} presents the results of applying chunk-wise and frame-wise RAD with different RNN variants to the Minecraft dataset. The trends broadly match those observed on the Maze dataset, but with important differences driven by the substantially higher visual and structural complexity of Minecraft videos.

In the \textbf{chunk-wise} setting, LSTM again outperforms Mamba2 and TTT, consistent with its structural ability to balance short-term and long-term memory. However, unlike in the Maze experiments, all RNN variants fall short of the standard Diffusion Forcing baseline. Minecraft scenes contain dense textures, rich geometry, and rapid viewpoint changes, making it difficult for any recurrent hidden state to fully compress and transmit pixel-level information across chunk boundaries. As a result, the architectural limitations of relying solely on hidden states for inter-chunk communication become much more pronounced, leading to degraded reconstruction quality and weaker scene consistency.

In the \textbf{frame-wise} setting, the RAD models with three different RNN types exhibit similar performance, significantly surpassing the chunk-wise counterparts, because local visual information can propagate directly through attention, removing the need to encode high-frequency details in the hidden state. Under this more favorable regime, all three RNN types achieve significantly better performance than their chunk-wise counterparts.

Qualitative results in \cref{fig:mc_vis} highlight this contrast. Frame-wise TTT and frame-wise LSTM produce videos that maintain strong memory and stay aligned with ground truth across long horizons. Conversely, chunk-wise TTT and chunk-wise LSTM exhibit clear failures in memorizing scene layout and object configuration, reinforcing the difficulty of relying solely on hidden states to carry rich Minecraft-level detail across chunks.

Overall, these experiments further support the conclusion that the granularity of the autoregressive window plays a critical role. Chunk-wise autoregression imposes an unrealistic compression burden for visually complex environments, while frame-wise autoregression leverages attention to maintain local fidelity, allowing all RNN architectures to operate on more global signals and achieve substantially better results.





\paragraph{Performance Curve}

\begin{figure}[ht]
  \centering
  \begin{subfigure}{0.3\linewidth}
    \includegraphics[width=\linewidth]{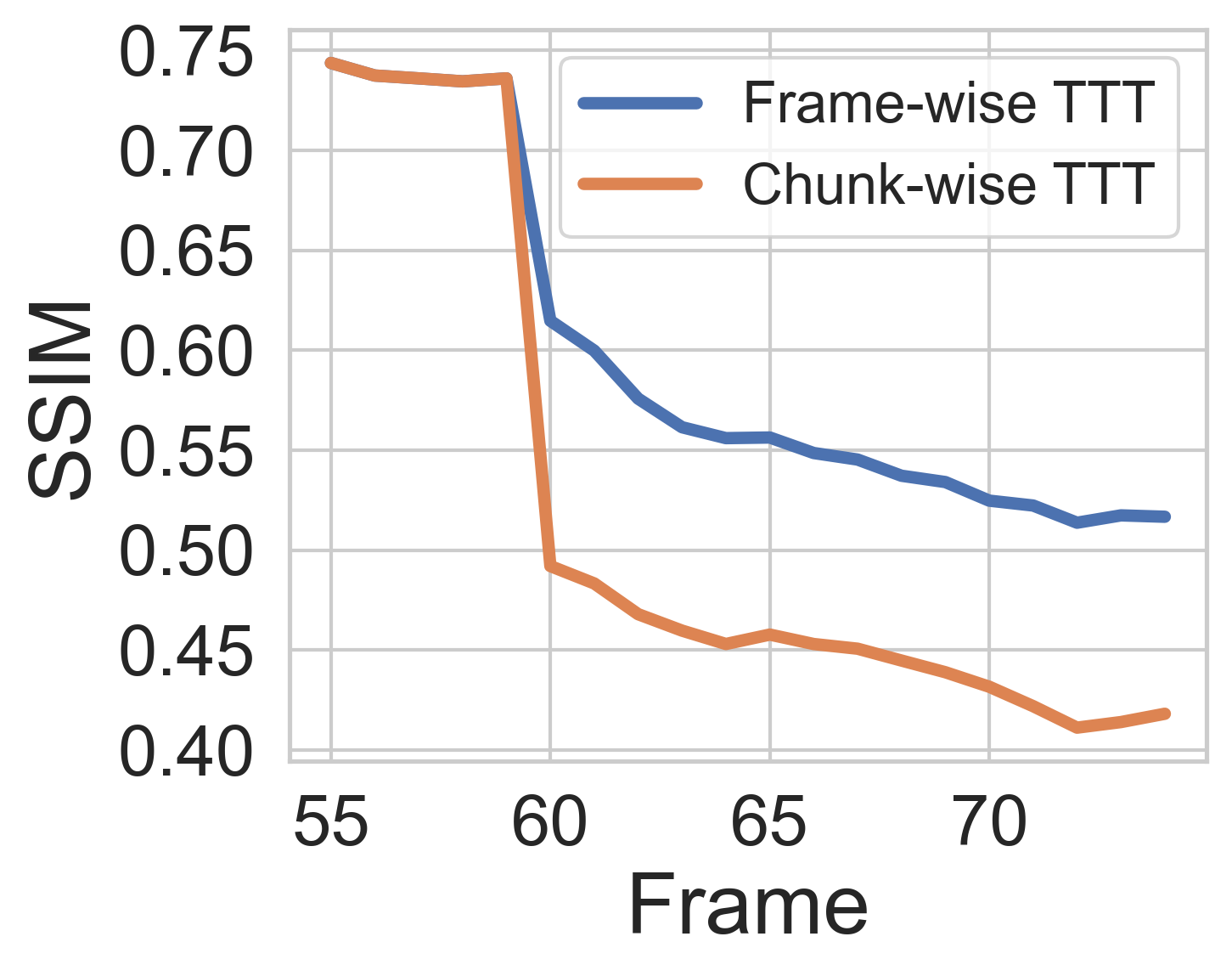}
    \caption{SSIM of chunk-wise and frame-wise TTT}
    \label{fig:mc_compare_ssim}
  \end{subfigure}
  \hspace{.6cm}
  \begin{subfigure}{0.3\linewidth}
    \includegraphics[width=\linewidth]{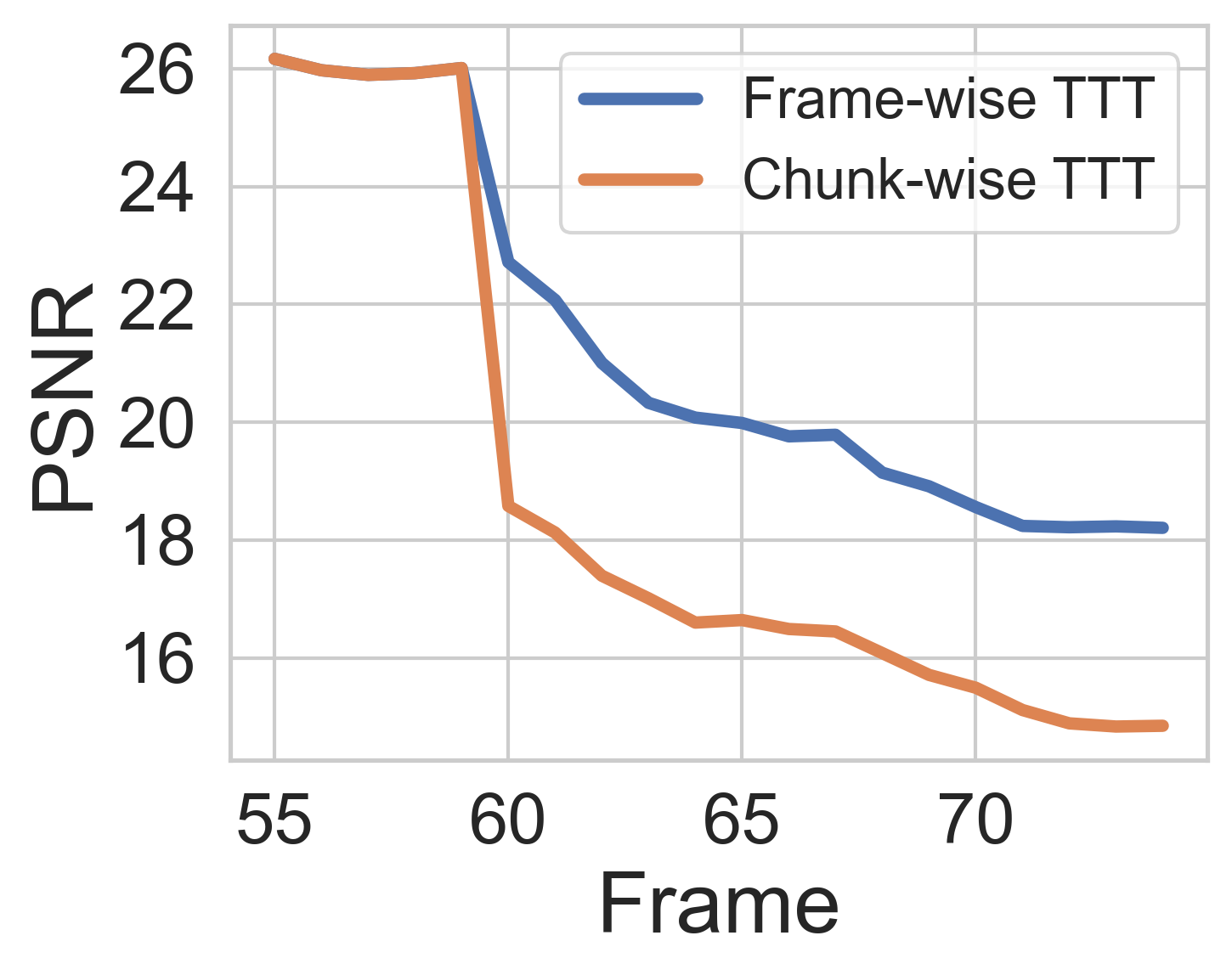}
    \caption{PSNR of chunk-wise and frame-wise TTT}
    \label{fig:mc_compare_psnr}
  \end{subfigure}
  \caption{Comparison of chunk-wise TTT and frame-wise TTT}
  \label{fig:mc_compare}
\end{figure}

To provide a more detailed comparison between chunk-wise and frame-wise RNN approaches, we present the SSIM and PSNR value curves as a function of frame index in \cref{fig:mc_compare}. Notably, both methods exhibit a decrease in performance at the 60-th frame, corresponding to the first frame predicted by the model. However, this drop is substantially more pronounced for the chunk-wise RNN, underscoring its limited capacity to effectively convey pixel-level information across chunk boundaries. 

\subsection{VBench Results}

We add VBench~\cite{huang2023vbench} evaluation results beyond just pixel similarity. As shown in \cref{tab:vbench}, the results on VBench demonstrate similar outcomes to those in \cref{sec:maze_results} and \cref{sec:minecraft_results}. The LSTM-based RNN achieves better performance, enabling the generation of higher-quality videos.

\begin{table}[h!]
    \centering
    \caption{VBench evaluation results for different autoregressive model variants.}
    \resizebox{\columnwidth}{!}{
        \begin{tabular}{l|ccccc|ccccc}
        \hline
 & \multicolumn{5}{c|}{Maze Dataset} & \multicolumn{5}{c}{Minecraft Dataset} \\
\hline
Metrics & \makecell{Background\\Consistency} & \makecell{Temporal \\Flickering} & \makecell{Motion \\Smoothness} & \makecell{Aesthetic \\Quality} & \makecell{Imaging \\Quality} & \makecell{Background\\Consistency} & \makecell{Temporal \\Flickering} & \makecell{Motion \\Smoothness} & \makecell{Aesthetic \\Quality} & \makecell{Imaging \\Quality} \\
 \hline
LSTM-c & \textbf{91.29} & \textbf{93.49} & 71.4 & \textbf{26.29} & 51.44 & 97.42 & 93.96 & 95.12 & 57.58 & \textbf{69.34}  \\
Mamba-c & 91.07 & 93.07 & 69.98 & 25.51 & 50.7 & 97.27 & 93.79 & 94.96 & 57.22 & 69.27 \\
TTT-c & 90.89 & 92.21 & 70.71 & 26 & \textbf{51.6} & \textbf{97.44} & 93.91 & 95.06 & 57.55 & 69.15 \\
LSTM-f & 91.16 & 93.01 & \textbf{71.48} & 26.11 & 50.94 & 97.38 & \textbf{94.21} & \textbf{95.3} & 55.71 & 66.72 \\ 
Mamba-f & 90.78 & 92.97 & 70.78 & 25.76 & 50.94 & 97.37 & 94.17 & 95.26 & \textbf{55.73} & 66.4 \\ 
TTT-f & 90.99 & 93.2 & 70.87 & 25.82 & 51.08 & 97.38 & 94.14 & 95.25 & 55.64 & 66.26 \\ 
\hline
        \end{tabular}
    }
    \label{tab:vbench}
\end{table}

\subsection{Computational Resource Analysis}

Directly comparing the computational efficiency of different RNN architectures is inherently challenging, as their implementations and underlying optimizations can vary substantially. For example, LSTM benefits from extensive low-level optimizations, such as those provided by cuDNN, whereas TTT currently lacks such specialized enhancement. Despite these differences, we report a summary of computational resource costs for reference. All reported results are obtained from experiments on the Minecraft dataset, conducted using 8 NVIDIA L40 GPUs and a batch size of 3.

As presented in \cref{tab:computational}, although the LSTM architecture features the highest parameter count, its GPU memory consumption and training time per step are comparable to those of Mamba2. These discrepancies in parameterization reflect intrinsic differences in model design rather than unfair experimental conditions, permitting a reasonable assessment.

\begin{table}[htbp]
  \centering
  \begin{minipage}[t]{0.53\textwidth}
    \centering
    \caption{Comparison of computational resource requirements among different RNN block types. ``Time'' denotes the duration of a single training step for an RAD model (same DiT) with the specified RNN type, while ``Params'' indicates the total number of parameters in the RNN block.}
    \resizebox{\columnwidth}{!}{
    \begin{tabular}{c|ccc}
    \toprule
        RNN Type & GPU Mem (GB) & Time (s/step) & Params (M) \\
        \midrule
        LSTM & 10.0 & 3.7 & 195 \\
        Mamba2 & 11.1 & 3.6 & 153\\
        TTT & 16.0 & 12.9 & 118 \\
    \bottomrule
    \end{tabular}
    }
    \label{tab:computational}
  \end{minipage}
  \hfill
  \begin{minipage}[t]{0.44\textwidth}
    \centering
        \caption{Ablation study on the effect of integrating conditional action information into the RNN input in the chunk-wise setting.}
    \resizebox{\columnwidth}{!}{
    \begin{tabular}{c|c|ccc}
    \toprule
        Model & +Action & PSNR $\uparrow$ & SSIM $\uparrow$ & LPIPS $\downarrow$ \\
        \midrule
        \multirow{2}{*}{RAD-Mamba2} & N  & 14.00  & 0.35 & 0.57 \\
         & Y & 13.80 & 0.31 & 0.58 \\
        \midrule
        \multirow{2}{*}{RAD-TTT}& N & 14.44 & 0.37 & 0.53 \\
         & Y & 14.36 & 0.35 & 0.53 \\
        \midrule
        \multirow{2}{*}{RAD-LSTM} & N  & 14.88 & 0.38 & 0.52 \\
        & Y & \textbf{15.64} & \textbf{0.43} &\textbf{0.47} \\
    \bottomrule
    \end{tabular}
    }
    \label{tab:rnn_chunk_no_comb}
  \end{minipage}
\end{table}

\section{Ablation study}
\subsection{Noise Level of Memory Frames}\label{sec:clean_mem}

\begin{figure}[htbp]
  \centering
  \begin{minipage}[b]{0.4\textwidth}
    \centering
    \includegraphics[width=\linewidth]{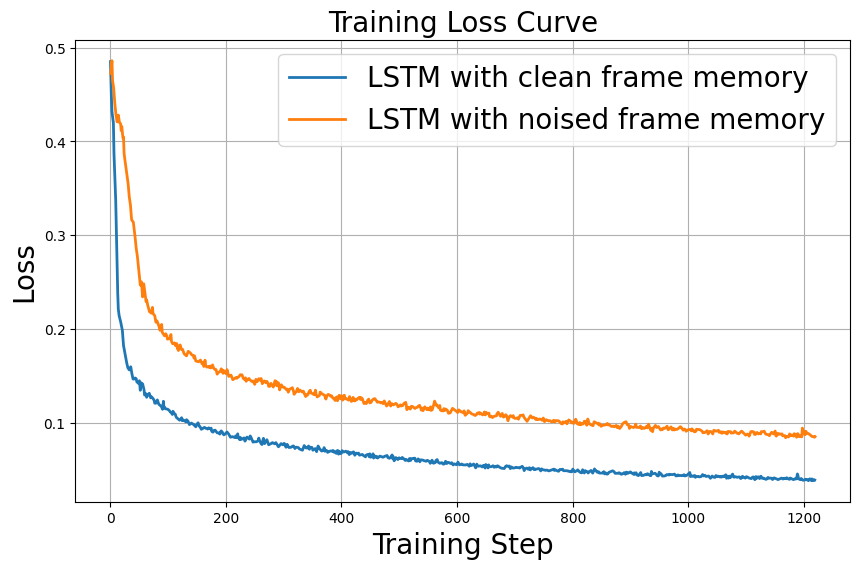}
    \caption{Training loss curves for frame-wise LSTM with clean frame memory and noised frame memory.}
    \label{fig:lnr}
  \end{minipage}
  \hspace{0.5cm}
  \begin{minipage}[b]{0.3\textwidth}
    \centering
    \resizebox{\textwidth}{!}{
    \begin{tabular}{c|cc}
      \toprule
      Metrics & noised & clean \\
      \midrule
      PSNR $\uparrow$ & 14.70 & \textbf{15.30} \\
      SSIM $\uparrow$ & 0.38 &  \textbf{0.41} \\
      LPIPS $\downarrow$ & 0.52 &  \textbf{0.50} \\
      \bottomrule
    \end{tabular}
    }
    \captionof{table}{Evaluation results for frame-wise LSTM with clean frame memory and noised frame memory on different metrics.}
    \label{tab:lnr}
  \end{minipage}
\end{figure}

We conducted an ablation study to evaluate the impact of noise levels applied to memory frames that are fed into hidden states of RNN during training. Our findings indicate that introducing higher noise levels makes model optimization more challenging, leading to increased training loss and degraded evaluation metrics, as illustrated in \cref{fig:lnr} and \cref{tab:lnr}. These results highlight the importance of our design choice: all clean frames used for memory are pre-processed in the pre-fetching stage, slid by a DiT model with a window size of 1 and decoupled from the denoising stage. 

This approach shares some similarities with the strategy of Po et al. \cite{po2025longcontextstatespacevideoworld}, where the initial $n$ frames of each training sequence are left clean and excluded from the diffusion loss calculation. However, our method is more comprehensive, where all frames passed into the RNN hidden states remain clean, rather than solely the initial portion. Moreover, our approach does not interfere with the computation of the diffusion loss, further enhancing training efficiency.

\subsection{Strided Chunk-wise Autoregression}\label{sec:abla_stride}
We further investigate the partially overlap attention window in chunk-wise RNN, by keeping the same model size and training data as Sec.~\ref{sec:maze_results}, while setting the stride of the sliding window to 10 frames (consistent during both training and inference). In other words, for frame-wise RNN, the stride is 1; for chunk-wise RNN, the stride equals the window size (20). The experimental results, as shown in Tab~\ref{tab:partial_overlap}, indicate that this method does not show advantage compared to both frame-wise RNN and chunk-wise RNN.

\begin{table}[h]
    \centering
    \caption{Strided Chunk-wise Autoregression for LSTM on Maze Dataset}
    \begin{resizebox}{0.8\columnwidth}{!}{
    \begin{tabular}{l|c|ccc}
    \toprule
    RNN Type & Stride & PSNR $\uparrow$ & SSIM $\uparrow$ & LPIPS $\downarrow$ \\
    \midrule
    LSTM-c & 20 & \textbf{15.64} & \textbf{0.43} & 0.47  \\
    LSTM-f & 1 & 15.50 & 0.41 & \textbf{0.45} \\
    LSTM-partial-overlap & 10 & 15.06 & 0.40 & 0.55 \\
    \bottomrule
    \end{tabular}}
    \end{resizebox}
    \label{tab:partial_overlap}
\end{table}

\subsection{Action Condition Design}\label{sec:abla_cond}
We performed an ablation study to investigate the influence of RNN block design, focusing on how conditional information is incorporated. In the standard approach, timestep and action condition information are introduced to the DiT block via AdaLN, which may limit the RNN block’s ability to utilize these signals. To address this, we experiment with passing the action condition through a linear projection and concatenate it directly with the output of the attention block over the hidden dimension, using this combination as the input to the RNN block.

As shown in \cref{tab:rnn_chunk_no_comb}, this modification leads to a notable improvement for LSTM, but produces only limited gains for TTT and Mamba architectures. Remarkably, LSTM continues to outperform both TTT and Mamba even without this integration, highlighting its inherent superiority for modeling sequential information.


\section{Implementationation Details}

\subsection{Mamba and TTT}
\label{sec:mamba_ttt}
\paragraph{Mamba}
\begin{align*}
    H_t = AH_{t-1} + BX_{t-1};\quad X_t = CH_t + DX_{t-1}
\end{align*}
where $H_t$ are latent states, and $A,B,C,D$ are linear projections of input $X_t$, i.e., $A_t$ := $\text{Linear}_{\theta_A}$ ($X_t$) and similarly for $B_t$, $C_t$, and $D_t$. This is the mathematical formulation of Mamba for autoregressive tasks. In a multi-layer setting, the output from the previous timestep does not serve as the input for the next timestep, but rather as the input for the next layer. Therefore, it can be expressed as:

\begin{align*}
    H_t = AH_{t-1} + BX_{t};\quad Y_t = CH_t + DX_t
\end{align*}

\paragraph{TTT}
\begin{align*}
    W_t &= W_{t-1}-\eta\nabla \mathcal{L}(W_{t-1};x_t) \\
    y_t &= f(\theta_Qx_t;W_t)
\end{align*}
where the self-supervised loss $\mathcal{L}$ is often defined as $\mathcal{L}(W; x_t) = ||f(\theta_Kx_t;W)-\theta_Vx_t||^2$, $\theta_Q,\theta_K,\theta_V$ are trainable parameters, and $W_t$ is a hidden state matrix at time $t$. TTT-linear learns per-instance weights $W_t \in \mathbb{R}^{d \times d}$, but with often small MLPs or projections. TTT updates global weights per step; typically, there is no recurrent hidden state transfer beyond $\mathcal{L}(W; x_t)$.

\paragraph{Comparison}
As presented in ~\cref{tab:comparison}, we provide a comparison of different RNN types, where $h$ is LSTM hidden size, $n$ is Mamba SSM state size.

\begin{table*}[htbp]
  \centering
  \caption{Comparison of computational complexity between LSTM, Mamba (SSM), and Test-Time Training (TTT-linear with $d\times d$ memory weights) for input tensor $(B,H,W,L,d)$.}
  \label{tab:comparison}
  \begin{resizebox}{\columnwidth}{!}{
  \begin{tabular}{l|c|c|c}
  \hline
  \textbf{Aspect} & \textbf{LSTM} & \textbf{Mamba (SSM)} & \textbf{TTT} \\
  \hline
  Computation (FLOPs) & $O(BHWL(dh + h^2))$ & $O(BHWL\,dn)$ & $O(BHWL(d^2 + C))$ \\
  Parameter count & $O(h(d+h))$ & $O(dn) + O(d^2)$ (if projected) & $O(d^2)$ \\
  Training Memory & $O(BHWLh)$ & $O(BHWL(d+n))$ & $O(BHWd)$ \\
  Inference Memory & $O(BHWh)$ & $O(BHW(d+n))$ & $O(BHWd)$ \\
  Sequence scaling & Linear in $L$ & Linear in $L$ & Linear in $L$ \\
  \hline
  \end{tabular}
  }
  \end{resizebox}
\end{table*}

\subsection{Efficient Parallelization of Attention in Frame-wise RNNs}

To enable efficient parallel computation of attention in frame-wise RNNs, we employ distinct parallelization strategies for two key stages: (1) pre-fetching hidden states and (2) diffusion forward computation.

In the first stage, we parallelize attention computation within each DiT layer along the temporal dimension of the sequence. We apply sliding window with size $l=1$ to pre-fetch hidden states. Taking the Maze setting as an example, we have a large number of chunks since each training sample containing $L=1001$ frames. For parallel computation we further split these chunks for each sample into small chunk batches, with a total of $b=128$ batches to satisfy the GPU memory constraint. Note that this is different from the global batch size $B=64$, which is the number of sample of length $L$ in each batch. The training data is partitioned into batches accordingly, and for each layer, we sequentially process the sample batch such that both spatial and temporal attention are computed in parallel across samples in the batch. After that, the RNN forward operation is conducted subsequently within each layer. Crucially, the spatial attention ($[BL, H, W, d]$) mechanism operates independently across frames, while the temporal attention ($[BHW\frac{L}{l}, l, d]$) employs a window of size $l=1$, thus preserving temporal causality as in causal attention. By the end of this process, we obtain all hidden states for each frame and layer.

For the second stage, we perform parallelization along the entire temporal dimension for diffusion forward computation to get the diffusion loss. We apply a normal window size $l'=20$ for attention computation. To mitigate the risk of GPU memory exhaustion, we randomly sample $N$ subsequences from the full sequence for evaluating the diffusion loss. The selected $N$ hidden states and their associated noisy frame inputs are concatenated along the batch dimension ($[BNl', H, W, d]$ for spatial and $[BHWN, l', d]$ for temporal), enabling the simultaneous computation of attention. This dual-stage parallelization framework is designed to maximize memory efficiency without exceeding memory constraints, with the only recurrence occurring during the RNN forward pass.

\singlespacing
\bibliographystyle{plainnat}


\bibliography{thesis}

@article{peters2008natural,
  title={Natural actor-critic},
  author={Peters, Jan and Schaal, Stefan},
  journal={Neurocomputing},
  volume={71},
  number={7-9},
  pages={1180--1190},
  year={2008},
  publisher={Elsevier}
}

@article{chen2025recurrent,
  title={Recurrent autoregressive diffusion: Global memory meets local attention},
  author={Chen, Taiye and Ding, Zihan and Li, Anjian and Zhang, Christina and Xiao, Zeqi and Wang, Yisen and Jin, Chi},
  journal={arXiv preprint arXiv:2511.12940},
  year={2025}
}

@inproceedings{haarnoja2018soft,
  title={Soft Actor-Critic: Off-Policy Maximum Entropy Deep Reinforcement Learning with a Stochastic Actor},
  author={Haarnoja, Tuomas and Zhou, Aurick and Abbeel, Pieter and Levine, Sergey},
  booktitle={International Conference on Machine Learning},
  pages={1861--1870},
  year={2018}
}

@inproceedings{konda2000actor,
  title={Actor-critic algorithms},
  author={Konda, Vijay R and Tsitsiklis, John N},
  booktitle={Advances in neural information processing systems},
  pages={1008--1014},
  year={2000}
}

@article{lowe2017multi,
  title={Multi-agent actor-critic for mixed cooperative-competitive environments},
  author={Lowe, Ryan and Wu, Yi I and Tamar, Aviv and Harb, Jean and Pieter Abbeel, OpenAI and Mordatch, Igor},
  journal={Advances in neural information processing systems},
  volume={30},
  year={2017}
}

@article{go,
author = {Silver, David and Schrittwieser, Julian and Antonoglou, Ioannis and Huang, Aja and Guez, Arthur and Hubert, Thomas and Baker, Lucas and Lai, Matthew and Bolton, Adrian and Chen, Yutian and Lillicrap, Timothy and Hui, Fan and Sifre, Laurent and Driessche, George and Graepel, Thore and Hassabis, Demis},
year = {2017},
month = {10},
pages = {354-359},
title = {Mastering the game of Go without human knowledge},
volume = {550},
journal = {Nature},
doi = {10.1038/nature24270}
}

@inproceedings{baker2019emergent,
  title={Emergent Tool Use From Multi-Agent Autocurricula},
  author={Baker, Bowen and Kanitscheider, Ingmar and Markov, Todor and Wu, Yi and Powell, Glenn and McGrew, Bob and Mordatch, Igor},
  booktitle={International Conference on Learning Representations},
  year={2019}
}

@misc{magi1,
      title={MAGI-1: Autoregressive Video Generation at Scale},
      author={Sand-AI},
      year={2025},
      url={https://static.magi.world/static/files/MAGI_1.pdf},
}

@article{liu2024reconx,
  title={Reconx: Reconstruct any scene from sparse views with video diffusion model},
  author={Liu, Fangfu and Sun, Wenqiang and Wang, Hanyang and Wang, Yikai and Sun, Haowen and Ye, Junliang and Zhang, Jun and Duan, Yueqi},
  journal={arXiv preprint arXiv:2408.16767},
  year={2024}
}

@misc{xiao2025worldmemlongtermconsistentworld,
      title={WORLDMEM: Long-term Consistent World Simulation with Memory}, 
      author={Zeqi Xiao and Yushi Lan and Yifan Zhou and Wenqi Ouyang and Shuai Yang and Yanhong Zeng and Xingang Pan},
      year={2025},
      eprint={2504.12369},
      archivePrefix={arXiv},
      primaryClass={cs.CV},
      url={https://arxiv.org/abs/2504.12369}, 
}

@article{kang2024far,
  title={How far is video generation from world model: A physical law perspective},
  author={Kang, Bingyi and Yue, Yang and Lu, Rui and Lin, Zhijie and Zhao, Yang and Wang, Kaixin and Huang, Gao and Feng, Jiashi},
  journal={arXiv preprint arXiv:2411.02385},
  year={2024}
}

@article{gao2024cat3d,
  title={Cat3d: Create anything in 3d with multi-view diffusion models},
  author={Gao, Ruiqi and Holynski, Aleksander and Henzler, Philipp and Brussee, Arthur and Martin-Brualla, Ricardo and Srinivasan, Pratul and Barron, Jonathan T and Poole, Ben},
  journal={arXiv preprint arXiv:2405.10314},
  year={2024}
}

@misc{ren2025cosmosdrivedreamsscalablesyntheticdriving,
      title={Cosmos-Drive-Dreams: Scalable Synthetic Driving Data Generation with World Foundation Models}, 
      author={Xuanchi Ren and Yifan Lu and Tianshi Cao and Ruiyuan Gao and Shengyu Huang and Amirmojtaba Sabour and Tianchang Shen and Tobias Pfaff and Jay Zhangjie Wu and Runjian Chen and Seung Wook Kim and Jun Gao and Laura Leal-Taixe and Mike Chen and Sanja Fidler and Huan Ling},
      year={2025},
      eprint={2506.09042},
      archivePrefix={arXiv},
      primaryClass={cs.CV},
      url={https://arxiv.org/abs/2506.09042}, 
}

@article{zhen2025tesseract,
  title={TesserAct: Learning 4D Embodied World Models},
  author={Zhen, Haoyu and Sun, Qiao and Zhang, Hongxin and Li, Junyan and Zhou, Siyuan and Du, Yilun and Gan, Chuang},
  journal={arXiv preprint arXiv:2504.20995},
  year={2025}
}

@article{henschel2024streamingt2v,
  title={Streamingt2v: Consistent, dynamic, and extendable long video generation from text},
  author={Henschel, Roberto and Khachatryan, Levon and Hayrapetyan, Daniil and Poghosyan, Hayk and Tadevosyan, Vahram and Wang, Zhangyang and Navasardyan, Shant and Shi, Humphrey},
  journal={arXiv preprint arXiv:2403.14773},
  year={2024}
}

@article{ren2025gen3c,
  title={Gen3c: 3d-informed world-consistent video generation with precise camera control},
  author={Ren, Xuanchi and Shen, Tianchang and Huang, Jiahui and Ling, Huan and Lu, Yifan and Nimier-David, Merlin and M{\"u}ller, Thomas and Keller, Alexander and Fidler, Sanja and Gao, Jun},
  journal={arXiv preprint arXiv:2503.03751},
  year={2025}
}

@article{team2025aether,
  title={Aether: Geometric-aware unified world modeling},
  author={Team, Aether and Zhu, Haoyi and Wang, Yifan and Zhou, Jianjun and Chang, Wenzheng and Zhou, Yang and Li, Zizun and Chen, Junyi and Shen, Chunhua and Pang, Jiangmiao and others},
  journal={arXiv preprint arXiv:2503.18945},
  year={2025}
}

@article{voleti2022mcvd,
  title={Mcvd-masked conditional video diffusion for prediction, generation, and interpolation},
  author={Voleti, Vikram and Jolicoeur-Martineau, Alexia and Pal, Chris},
  journal={Advances in neural information processing systems},
  volume={35},
  pages={23371--23385},
  year={2022}
}

@article{karras2022elucidating,
  title={Elucidating the design space of diffusion-based generative models},
  author={Karras, Tero and Aittala, Miika and Aila, Timo and Laine, Samuli},
  journal={Advances in Neural Information Processing Systems},
  volume={35},
  pages={26565--26577},
  year={2022}
}

@article{achiam2023gpt,
  title={Gpt-4 technical report},
  author={Achiam, Josh and Adler, Steven and Agarwal, Sandhini and Ahmad, Lama and Akkaya, Ilge and Aleman, Florencia Leoni and Almeida, Diogo and Altenschmidt, Janko and Altman, Sam and Anadkat, Shyamal and others},
  journal={arXiv preprint arXiv:2303.08774},
  year={2023}
}

@article{li2024reward,
  title={Reward guided latent consistency distillation},
  author={Li, Jiachen and Feng, Weixi and Chen, Wenhu and Wang, William Yang},
  journal={arXiv preprint arXiv:2403.11027},
  year={2024}
}

@article{friedman2022vendi,
  title={The vendi score: A diversity evaluation metric for machine learning},
  author={Friedman, Dan and Dieng, Adji Bousso},
  journal={arXiv preprint arXiv:2210.02410},
  year={2022}
}

@article{ho2022classifier,
  title={Classifier-free diffusion guidance},
  author={Ho, Jonathan and Salimans, Tim},
  journal={arXiv preprint arXiv:2207.12598},
  year={2022}
}

@article{yang2024cogvideox,
  title={Cogvideox: Text-to-video diffusion models with an expert transformer},
  author={Yang, Zhuoyi and Teng, Jiayan and Zheng, Wendi and Ding, Ming and Huang, Shiyu and Xu, Jiazheng and Yang, Yuanming and Hong, Wenyi and Zhang, Xiaohan and Feng, Guanyu and others},
  journal={arXiv preprint arXiv:2408.06072},
  year={2024}
}

@article{zhao2023pytorch,
  title={Pytorch fsdp: experiences on scaling fully sharded data parallel},
  author={Zhao, Yanli and Gu, Andrew and Varma, Rohan and Luo, Liang and Huang, Chien-Chin and Xu, Min and Wright, Less and Shojanazeri, Hamid and Ott, Myle and Shleifer, Sam and others},
  journal={arXiv preprint arXiv:2304.11277},
  year={2023}
}

@article{polyak2024movie,
  title={Movie gen: A cast of media foundation models},
  author={Polyak, Adam and Zohar, Amit and Brown, Andrew and Tjandra, Andros and Sinha, Animesh and Lee, Ann and Vyas, Apoorv and Shi, Bowen and Ma, Chih-Yao and Chuang, Ching-Yao and others},
  journal={arXiv preprint arXiv:2410.13720},
  year={2024}
}

@article{kirstain2023pick,
  title={Pick-a-pic: An open dataset of user preferences for text-to-image generation},
  author={Kirstain, Yuval and Polyak, Adam and Singer, Uriel and Matiana, Shahbuland and Penna, Joe and Levy, Omer},
  journal={Advances in Neural Information Processing Systems},
  volume={36},
  pages={36652--36663},
  year={2023}
}

@article{ren2024diffusion,
  title={Diffusion policy policy optimization},
  author={Ren, Allen Z and Lidard, Justin and Ankile, Lars L and Simeonov, Anthony and Agrawal, Pulkit and Majumdar, Anirudha and Burchfiel, Benjamin and Dai, Hongkai and Simchowitz, Max},
  journal={arXiv preprint arXiv:2409.00588},
  year={2024}
}

@article{luo2023latent,
  title={Latent consistency models: Synthesizing high-resolution images with few-step inference},
  author={Luo, Simian and Tan, Yiqin and Huang, Longbo and Li, Jian and Zhao, Hang},
  journal={arXiv preprint arXiv:2310.04378},
  year={2023}
}

@article{chung2022diffusion,
  title={Diffusion posterior sampling for general noisy inverse problems},
  author={Chung, Hyungjin and Kim, Jeongsol and Mccann, Michael T and Klasky, Marc L and Ye, Jong Chul},
  journal={arXiv preprint arXiv:2209.14687},
  year={2022}
}

@article{jin2024pyramidal,
  title={Pyramidal Flow Matching for Efficient Video Generative Modeling},
  author={Jin, Yang and Sun, Zhicheng and Li, Ningyuan and Xu, Kun and Jiang, Hao and Zhuang, Nan and Huang, Quzhe and Song, Yang and Mu, Yadong and Lin, Zhouchen},
  journal={arXiv preprint arXiv:2410.05954},
  year={2024}
}

@article{black2023training,
  title={Training diffusion models with reinforcement learning},
  author={Black, Kevin and Janner, Michael and Du, Yilun and Kostrikov, Ilya and Levine, Sergey},
  journal={arXiv preprint arXiv:2305.13301},
  year={2023}
}

@misc{pikalabs2023,
  author       = {{Pika Labs}},
  title        = {{Pika Labs}},
  howpublished = {\url{https://www.pika.art/}},
  note         = {Accessed: September 25, 2023}
}

@misc{kuaishou2024,
  author       = {Kuaishou},
  title        = {Kling},
  year         = {2024},
  howpublished = {\url{https://kling.kuaishou.com/en}},
  note         = {Accessed: [today's date]}
}

@article{poole2022dreamfusion,
  title={Dreamfusion: Text-to-3d using 2d diffusion},
  author={Poole, Ben and Jain, Ajay and Barron, Jonathan T and Mildenhall, Ben},
  journal={arXiv preprint arXiv:2209.14988},
  year={2022}
}

@inproceedings{wang2023score,
  title={Score jacobian chaining: Lifting pretrained 2d diffusion models for 3d generation},
  author={Wang, Haochen and Du, Xiaodan and Li, Jiahao and Yeh, Raymond A and Shakhnarovich, Greg},
  booktitle={Proceedings of the IEEE/CVF Conference on Computer Vision and Pattern Recognition},
  pages={12619--12629},
  year={2023}
}

@article{wang2024prolificdreamer,
  title={Prolificdreamer: High-fidelity and diverse text-to-3d generation with variational score distillation},
  author={Wang, Zhengyi and Lu, Cheng and Wang, Yikai and Bao, Fan and Li, Chongxuan and Su, Hang and Zhu, Jun},
  journal={Advances in Neural Information Processing Systems},
  volume={36},
  year={2024}
}

@article{prabhudesai2024video,
  title={Video Diffusion Alignment via Reward Gradients},
  author={Prabhudesai, Mihir and Mendonca, Russell and Qin, Zheyang and Fragkiadaki, Katerina and Pathak, Deepak},
  journal={arXiv preprint arXiv:2407.08737},
  year={2024}
}

@inproceedings{liu2023instaflow,
  title={Instaflow: One step is enough for high-quality diffusion-based text-to-image generation},
  author={Liu, Xingchao and Zhang, Xiwen and Ma, Jianzhu and Peng, Jian and others},
  booktitle={The Twelfth International Conference on Learning Representations},
  year={2023}
}

@article{bar2024lumiere,
  title={Lumiere: A space-time diffusion model for video generation},
  author={Bar-Tal, Omer and Chefer, Hila and Tov, Omer and Herrmann, Charles and Paiss, Roni and Zada, Shiran and Ephrat, Ariel and Hur, Junhwa and Liu, Guanghui and Raj, Amit and others},
  journal={arXiv preprint arXiv:2401.12945},
  year={2024}
}

@article{wang2023lavie,
  title={Lavie: High-quality video generation with cascaded latent diffusion models},
  author={Wang, Yaohui and Chen, Xinyuan and Ma, Xin and Zhou, Shangchen and Huang, Ziqi and Wang, Yi and Yang, Ceyuan and He, Yinan and Yu, Jiashuo and Yang, Peiqing and others},
  journal={arXiv preprint arXiv:2309.15103},
  year={2023}
}

@article{polino2018model,
  title={Model compression via distillation and quantization},
  author={Polino, Antonio and Pascanu, Razvan and Alistarh, Dan},
  journal={arXiv preprint arXiv:1802.05668},
  year={2018}
}

@article{kingma2013auto,
  title={Auto-encoding variational bayes},
  author={Kingma, Diederik P},
  journal={arXiv preprint arXiv:1312.6114},
  year={2013}
}

@article{song2019generative,
  title={Generative modeling by estimating gradients of the data distribution},
  author={Song, Yang and Ermon, Stefano},
  journal={Advances in neural information processing systems},
  volume={32},
  year={2019}
}

@article{zhang2024sf,
  title={SF-V: Single Forward Video Generation Model},
  author={Zhang, Zhixing and Li, Yanyu and Wu, Yushu and Xu, Yanwu and Kag, Anil and Skorokhodov, Ivan and Menapace, Willi and Siarohin, Aliaksandr and Cao, Junli and Metaxas, Dimitris and others},
  journal={arXiv preprint arXiv:2406.04324},
  year={2024}
}

@inproceedings{rombach2022high,
  title={High-resolution image synthesis with latent diffusion models},
  author={Rombach, Robin and Blattmann, Andreas and Lorenz, Dominik and Esser, Patrick and Ommer, Bj{\"o}rn},
  booktitle={Proceedings of the IEEE/CVF conference on computer vision and pattern recognition},
  pages={10684--10695},
  year={2022}
}

@inproceedings{blattmann2023align,
  title={Align your latents: High-resolution video synthesis with latent diffusion models},
  author={Blattmann, Andreas and Rombach, Robin and Ling, Huan and Dockhorn, Tim and Kim, Seung Wook and Fidler, Sanja and Kreis, Karsten},
  booktitle={Proceedings of the IEEE/CVF Conference on Computer Vision and Pattern Recognition},
  pages={22563--22575},
  year={2023}
}

@software{opensora,
  author = {Zangwei Zheng and Xiangyu Peng and Tianji Yang and Chenhui Shen and Shenggui Li and Hongxin Liu and Yukun Zhou and Tianyi Li and Yang You},
  title = {Open-Sora: Democratizing Efficient Video Production for All},
  month = {March},
  year = {2024},
  url = {https://github.com/hpcaitech/Open-Sora}
}

@inproceedings{ho2020denoising,
  title={Denoising diffusion probabilistic models},
  author={Ho, Jonathan and Jain, Ajay and Abbeel, Pieter},
  booktitle={Advances in Neural Information Processing Systems},
  volume={33},
  pages={6840--6851},
  year={2020}
}

@article{domingo2024adjoint,
  title={Adjoint Matching: Fine-tuning Flow and Diffusion Generative Models with Memoryless Stochastic Optimal Control},
  author={Domingo-Enrich, Carles and Drozdzal, Michal and Karrer, Brian and Chen, Ricky TQ},
  journal={arXiv preprint arXiv:2409.08861},
  year={2024}
}

@inproceedings{song2021score,
  title={Score-based generative modeling through stochastic differential equations},
  author={Song, Yang and Sohl-Dickstein, Jascha and Kingma, Diederik P. and Kumar, Abhishek and Ermon, Stefano and Poole, Ben},
  booktitle={International Conference on Learning Representations},
  year={2021}
}

@inproceedings{sohl2015deep,
  title={Deep unsupervised learning using nonequilibrium thermodynamics},
  author={Sohl-Dickstein, Jascha and Weiss, Eric A. and Maheswaranathan, Niru and Ganguli, Surya},
  booktitle={International Conference on Machine Learning},
  pages={2256--2265},
  year={2015}
}

@article{ho2022video,
  title={Video Diffusion Models},
  author={Ho, Jonathan and Salimans, Tim and Gritsenko, Alexey and Chan, William and Norouzi, Mohammad and Fleet, David J.},
  journal={arXiv preprint arXiv:2204.03458},
  year={2022}
}

@article{harvey2022flexible,
  title={Flexible Diffusion Modeling of Long Videos},
  author={Harvey, William and N{\o}rskov, S{\o}ren and K{\"o}lch, Niklas and Vogiatzis, George},
  journal={arXiv preprint arXiv:2205.11495},
  year={2022}
}

@article{singer2022make,
  title={Make-A-Video: Text-to-Video Generation without Text-Video Data},
  author={Singer, Uriel and Polyak, Adam and Nachmani, Eliya and Dahan, Guy and Shechtman, Eli and Hacohen, Haggai},
  journal={arXiv preprint arXiv:2209.14792},
  year={2022}
}

@article{villegas2022phenaki,
  title={Phenaki: Variable Length Video Generation from Open Domain Textual Descriptions},
  author={Villegas, Ruben and Yang, Jiahui and Tulyakov, Sergey and Kautz, Jan and Hong, Seungjun},
  journal={arXiv preprint arXiv:2210.02399},
  year={2022}
}

@article{xiao2023dual,
  title={Dual Diffusion Models for High-Fidelity Video Generation},
  author={Xiao, Tong and Liu, Peng and Yang, Yi},
  journal={arXiv preprint arXiv:2301.06513},
  year={2023}
}

@article{hong2022cogvideo,
  title={CogVideo: Large-scale Pretraining for Text-to-Video Generation with Transformers},
  author={Hong, Yu and Wei, Jing and Liu, Xing and Wang, Xiaodi and Bai, Yutong and Li, Haitao and Zhang, Ming and Xu, Hao},
  journal={arXiv preprint arXiv:2205.15868},
  year={2022}
}

@article{khachatryan2023text2video,
  title={Text2Video-Zero: Zero-Shot Text-to-Video Generation using Pretrained Text-to-Image Diffusion Models},
  author={Khachatryan, Levon and Davy, Adrien and Emond, Baptiste and Wang, Jun},
  journal={arXiv preprint arXiv:2302.01327},
  year={2023}
}

@article{wang2023modelscope,
  title={Modelscope text-to-video technical report},
  author={Wang, Jiuniu and Yuan, Hangjie and Chen, Dayou and Zhang, Yingya and Wang, Xiang and Zhang, Shiwei},
  journal={arXiv preprint arXiv:2308.06571},
  year={2023}
}

@inproceedings{chen2024videocrafter2,
  title={Videocrafter2: Overcoming data limitations for high-quality video diffusion models},
  author={Chen, Haoxin and Zhang, Yong and Cun, Xiaodong and Xia, Menghan and Wang, Xintao and Weng, Chao and Shan, Ying},
  booktitle={Proceedings of the IEEE/CVF Conference on Computer Vision and Pattern Recognition},
  pages={7310--7320},
  year={2024}
}

@article{wu2023human,
  title={Human preference score v2: A solid benchmark for evaluating human preferences of text-to-image synthesis},
  author={Wu, Xiaoshi and Hao, Yiming and Sun, Keqiang and Chen, Yixiong and Zhu, Feng and Zhao, Rui and Li, Hongsheng},
  journal={arXiv preprint arXiv:2306.09341},
  year={2023}
}

@article{blattmann2023stable,
  title={Stable video diffusion: Scaling latent video diffusion models to large datasets},
  author={Blattmann, Andreas and Dockhorn, Tim and Kulal, Sumith and Mendelevitch, Daniel and Kilian, Maciej and Lorenz, Dominik and Levi, Yam and English, Zion and Voleti, Vikram and Letts, Adam and others},
  journal={arXiv preprint arXiv:2311.15127},
  year={2023}
}

@article{liu2022flow,
  title={Flow straight and fast: Learning to generate and transfer data with rectified flow},
  author={Liu, Xingchao and Gong, Chengyue and Liu, Qiang},
  journal={arXiv preprint arXiv:2209.03003},
  year={2022}
}

@misc{kong2024hunyuanvideo,
      title={HunyuanVideo: A Systematic Framework For Large Video Generative Models}, 
      author={Weijie Kong and Qi Tian, et al},
      year={2024},
      archivePrefix={arXiv preprint arXiv:2412.03603},
      primaryClass={cs.CV},
      url={https://arxiv.org/abs/2412.03603}, 
}

@misc{polyak2025moviegencastmedia,
      title={Movie Gen: A Cast of Media Foundation Models}, 
      author={Adam Polyak and Amit Zohar, et al},
      year={2025},
      eprint={2410.13720},
      archivePrefix={arXiv},
      primaryClass={cs.CV},
      url={https://arxiv.org/abs/2410.13720}, 
}

@inproceedings{esser2024scaling,
  title={Scaling rectified flow transformers for high-resolution image synthesis},
  author={Esser, Patrick and Kulal, Sumith and Blattmann, Andreas and Entezari, Rahim and M{\"u}ller, Jonas and Saini, Harry and Levi, Yam and Lorenz, Dominik and Sauer, Axel and Boesel, Frederic and others},
  booktitle={Forty-first international conference on machine learning},
  year={2024}
}

@article{lipman2022flow,
  title={Flow matching for generative modeling},
  author={Lipman, Yaron and Chen, Ricky TQ and Ben-Hamu, Heli and Nickel, Maximilian and Le, Matt},
  journal={arXiv preprint arXiv:2210.02747},
  year={2022}
}

@article{wang2024animatelcm,
  title={Animatelcm: Accelerating the animation of personalized diffusion models and adapters with decoupled consistency learning},
  author={Wang, Fu-Yun and Huang, Zhaoyang and Shi, Xiaoyu and Bian, Weikang and Song, Guanglu and Liu, Yu and Li, Hongsheng},
  journal={arXiv preprint arXiv:2402.00769},
  year={2024}
}

@article{wang2023videolcm,
  title={Videolcm: Video latent consistency model},
  author={Wang, Xiang and Zhang, Shiwei and Zhang, Han and Liu, Yu and Zhang, Yingya and Gao, Changxin and Sang, Nong},
  journal={arXiv preprint arXiv:2312.09109},
  year={2023}
}

@inproceedings{song2021denoising,
  title={Denoising Diffusion Implicit Models},
  author={Song, Jiaming and Meng, Chenlin and Ermon, Stefano},
  booktitle={International Conference on Learning Representations},
  year={2021}
}

@article{salimans2022progressive,
  title={Progressive Distillation for Fast Sampling of Diffusion Models},
  author={Salimans, Tim and Ho, Jonathan},
  journal={arXiv preprint arXiv:2202.00512},
  year={2022}
}

@article{luhman2021knowledge,
  title={Knowledge Distillation for Generative Models},
  author={Luhman, Eric and Luhman, Tobias},
  journal={arXiv preprint arXiv:2106.05237},
  year={2021}
}

@article{clark2023directly,
  title={Directly fine-tuning diffusion models on differentiable rewards},
  author={Clark, Kevin and Vicol, Paul and Swersky, Kevin and Fleet, David J},
  journal={arXiv preprint arXiv:2309.17400},
  year={2023}
}

@article{schuhmann2022laion,
  title={Laion-5b: An open large-scale dataset for training next generation image-text models},
  author={Schuhmann, Christoph and Beaumont, Romain and Vencu, Richard and Gordon, Cade and Wightman, Ross and Cherti, Mehdi and Coombes, Theo and Katta, Aarush and Mullis, Clayton and Wortsman, Mitchell and others},
  journal={Advances in Neural Information Processing Systems},
  volume={35},
  pages={25278--25294},
  year={2022}
}

@article{xu2024imagereward,
  title={Imagereward: Learning and evaluating human preferences for text-to-image generation},
  author={Xu, Jiazheng and Liu, Xiao and Wu, Yuchen and Tong, Yuxuan and Li, Qinkai and Ding, Ming and Tang, Jie and Dong, Yuxiao},
  journal={Advances in Neural Information Processing Systems},
  volume={36},
  year={2024}
}

@article{wang2023internvid,
  title={Internvid: A large-scale video-text dataset for multimodal understanding and generation},
  author={Wang, Yi and He, Yinan and Li, Yizhuo and Li, Kunchang and Yu, Jiashuo and Ma, Xin and Li, Xinhao and Chen, Guo and Chen, Xinyuan and Wang, Yaohui and others},
  journal={arXiv preprint arXiv:2307.06942},
  year={2023}
}

@article{he2024videoscore,
  title={Videoscore: Building automatic metrics to simulate fine-grained human feedback for video generation},
  author={He, Xuan and Jiang, Dongfu and Zhang, Ge and Ku, Max and Soni, Achint and Siu, Sherman and Chen, Haonan and Chandra, Abhranil and Jiang, Ziyan and Arulraj, Aaran and others},
  journal={arXiv preprint arXiv:2406.15252},
  year={2024}
}

@inproceedings{huang2024vbench,
  title={Vbench: Comprehensive benchmark suite for video generative models},
  author={Huang, Ziqi and He, Yinan and Yu, Jiashuo and Zhang, Fan and Si, Chenyang and Jiang, Yuming and Zhang, Yuanhan and Wu, Tianxing and Jin, Qingyang and Chanpaisit, Nattapol and others},
  booktitle={Proceedings of the IEEE/CVF Conference on Computer Vision and Pattern Recognition},
  pages={21807--21818},
  year={2024}
}

@article{wang2024internvideo2,
  title={Internvideo2: Scaling video foundation models for multimodal video understanding},
  author={Wang, Yi and Li, Kunchang and Li, Xinhao and Yu, Jiashuo and He, Yinan and Chen, Guo and Pei, Baoqi and Zheng, Rongkun and Xu, Jilan and Wang, Zun and others},
  journal={arXiv preprint arXiv:2403.15377},
  year={2024}
}

@inproceedings{wu2023human_a,
  title={Human preference score: Better aligning text-to-image models with human preference},
  author={Wu, Xiaoshi and Sun, Keqiang and Zhu, Feng and Zhao, Rui and Li, Hongsheng},
  booktitle={Proceedings of the IEEE/CVF International Conference on Computer Vision},
  pages={2096--2105},
  year={2023}
}

@inproceedings{peebles2023scalable,
  title={Scalable diffusion models with transformers},
  author={Peebles, William and Xie, Saining},
  booktitle={Proceedings of the IEEE/CVF International Conference on Computer Vision},
  pages={4195--4205},
  year={2023}
}

@article{li2024t2v,
  title={T2V-Turbo: Breaking the Quality Bottleneck of Video Consistency Model with Mixed Reward Feedback},
  author={Li, Jiachen and Feng, Weixi and Fu, Tsu-Jui and Wang, Xinyi and Basu, Sugato and Chen, Wenhu and Wang, William Yang},
  journal={arXiv preprint arXiv:2405.18750},
  year={2024}
}

@article{li2024t2v2,
  title={T2v-turbo-v2: Enhancing video generation model post-training through data, reward, and conditional guidance design},
  author={Li, Jiachen and Long, Qian and Zheng, Jian and Gao, Xiaofeng and Piramuthu, Robinson and Chen, Wenhu and Wang, William Yang},
  journal={arXiv preprint arXiv:2410.05677},
  year={2024}
}

@article{yin2024improved,
  title={Improved Distribution Matching Distillation for Fast Image Synthesis},
  author={Yin, Tianwei and Gharbi, Micha{\"e}l and Park, Taesung and Zhang, Richard and Shechtman, Eli and Durand, Fredo and Freeman, William T},
  journal={arXiv preprint arXiv:2405.14867},
  year={2024}
}

@article{zhang2025fast,
  title={Fast Video Generation with Sliding Tile Attention},
  author={Zhang, Peiyuan and Chen, Yongqi and Su, Runlong and Ding, Hangliang and Stoica, Ion and Liu, Zhenghong and Zhang, Hao},
  journal={arXiv preprint arXiv:2502.04507},
  year={2025}
}

@inproceedings{yin2024one,
  title={One-step diffusion with distribution matching distillation},
  author={Yin, Tianwei and Gharbi, Micha{\"e}l and Zhang, Richard and Shechtman, Eli and Durand, Fredo and Freeman, William T and Park, Taesung},
  booktitle={Proceedings of the IEEE/CVF Conference on Computer Vision and Pattern Recognition},
  pages={6613--6623},
  year={2024}
}

@article{xiao2021tackling,
  title={Tackling the Generative Learning Trilemma with Denoising Diffusion GANs},
  author={Xiao, Tianyu and Bahri, Dara and Stanczuk, Pawel Lucjan and Ceylan, Duygu and McAuley, Julian and Vahdat, Arash and Kautz, Jan},
  journal={arXiv preprint arXiv:2112.07804},
  year={2021}
}

@article{goodfellow2014generative,
  title={Generative adversarial nets},
  author={Goodfellow, Ian and Pouget-Abadie, Jean and Mirza, Mehdi and Xu, Bing and Warde-Farley, David and Ozair, Sherjil and Courville, Aaron and Bengio, Yoshua},
  journal={Advances in neural information processing systems},
  volume={27},
  year={2014}
}

@article{kim2023consistency,
  title={Consistency trajectory models: Learning probability flow ode trajectory of diffusion},
  author={Kim, Dongjun and Lai, Chieh-Hsin and Liao, Wei-Hsiang and Murata, Naoki and Takida, Yuhta and Uesaka, Toshimitsu and He, Yutong and Mitsufuji, Yuki and Ermon, Stefano},
  journal={arXiv preprint arXiv:2310.02279},
  year={2023}
}

@article{xie2024distillation,
  title={EM Distillation for One-step Diffusion Models},
  author={Xie, Sirui and Xiao, Zhisheng and Kingma, Diederik P and Hou, Tingbo and Wu, Ying Nian and Murphy, Kevin Patrick and Salimans, Tim and Poole, Ben and Gao, Ruiqi},
  journal={arXiv preprint arXiv:2405.16852},
  year={2024}
}

@article{lu2024simplifying,
  title={Simplifying, Stabilizing and Scaling Continuous-Time Consistency Models},
  author={Lu, Cheng and Song, Yang},
  journal={arXiv preprint arXiv:2410.11081},
  year={2024}
}

@article{luo2024diff,
  title={Diff-instruct: A universal approach for transferring knowledge from pre-trained diffusion models},
  author={Luo, Weijian and Hu, Tianyang and Zhang, Shifeng and Sun, Jiacheng and Li, Zhenguo and Zhang, Zhihua},
  journal={Advances in Neural Information Processing Systems},
  volume={36},
  year={2024}
}

@article{salimans2024multistep,
  title={Multistep Distillation of Diffusion Models via Moment Matching},
  author={Salimans, Tim and Mensink, Thomas and Heek, Jonathan and Hoogeboom, Emiel},
  journal={arXiv preprint arXiv:2406.04103},
  year={2024}
}

@article{lu2022dpmsolver,
  title={{DPM-Solver}: A Fast ODE Solver for Diffusion Probabilistic Model Sampling in Around 10 Steps},
  author={Lu, Chao and Zhou, Yuhao and Bao, Fan and Chen, Jianfei and Li, Chongxuan and Zhu, Jun},
  journal={arXiv preprint arXiv:2206.00927},
  year={2022}
}

@article{mishra2023reorientdiff,
  title={ReorientDiff: Diffusion Model based Reorientation for Object Manipulation},
  author={Mishra, Utkarsh A and Chen, Yongxin},
  journal={arXiv preprint arXiv:2303.12700},
  year={2023}
}

@article{schulman2015high,
  title={High-dimensional continuous control using generalized advantage estimation},
  author={Schulman, John and Moritz, Philipp and Levine, Sergey and Jordan, Michael and Abbeel, Pieter},
  journal={arXiv preprint arXiv:1506.02438},
  year={2015}
}

@article{hansen2022modem,
  title={MoDem: Accelerating Visual Model-Based Reinforcement Learning with Demonstrations},
  author={Hansen, Nicklas and Lin, Yixin and Su, Hao and Wang, Xiaolong and Kumar, Vikash and Rajeswaran, Aravind},
  journal={arXiv preprint arXiv:2212.05698},
  year={2022}
}

@article{xu2023controllable,
  title={Controllable Video Generation by Learning the Underlying Dynamical System with Neural ODE},
  author={Xu, Yucheng and Li, Nanbo and Goel, Arushi and Guo, Zijian and Yao, Zonghai and Kasaei, Hamidreza and Kasaei, Mohammadreze and Li, Zhibin},
  journal={arXiv preprint arXiv:2303.05323},
  year={2023}
}

@article{jia2023chain,
  title={Chain-of-Thought Predictive Control},
  author={Jia, Zhiwei and Liu, Fangchen and Thumuluri, Vineet and Chen, Linghao and Huang, Zhiao and Su, Hao},
  journal={arXiv preprint arXiv:2304.00776},
  year={2023}
}

@article{du2023learning,
  title={Learning Universal Policies via Text-Guided Video Generation},
  author={Du, Yilun and Yang, Mengjiao and Dai, Bo and Dai, Hanjun and Nachum, Ofir and Tenenbaum, Josh and Schuurmans, Dale and Abbeel, Pieter},
  journal={arXiv preprint arXiv:2302.00111},
  year={2023}
}

@article{nakamoto2023cal,
  title={Cal-QL: Calibrated Offline RL Pre-Training for Efficient Online Fine-Tuning},
  author={Nakamoto, Mitsuhiko and Zhai, Yuexiang and Singh, Anikait and Mark, Max Sobol and Ma, Yi and Finn, Chelsea and Kumar, Aviral and Levine, Sergey},
  journal={arXiv preprint arXiv:2303.05479},
  year={2023}
}

@article{janner2022planning,
  title={Planning with diffusion for flexible behavior synthesis},
  author={Janner, Michael and Du, Yilun and Tenenbaum, Joshua B and Levine, Sergey},
  journal={arXiv preprint arXiv:2205.09991},
  year={2022}
}

@article{ajay2022conditional,
  title={Is Conditional Generative Modeling all you need for Decision-Making?},
  author={Ajay, Anurag and Du, Yilun and Gupta, Abhi and Tenenbaum, Joshua and Jaakkola, Tommi and Agrawal, Pulkit},
  journal={arXiv preprint arXiv:2211.15657},
  year={2022}
}

@article{chitnis2023iql,
  title={IQL-TD-MPC: Implicit Q-Learning for Hierarchical Model Predictive Control},
  author={Chitnis, Rohan and Xu, Yingchen and Hashemi, Bobak and Lehnert, Lucas and Dogan, Urun and Zhu, Zheqing and Delalleau, Olivier},
  journal={arXiv preprint arXiv:2306.00867},
  year={2023}
}

@inproceedings{chen2021decision,
title={Decision Transformer: Reinforcement Learning via Sequence Modeling},
author={Lili Chen and Kevin Lu and Aravind Rajeswaran and Kimin Lee and Aditya Grover and Michael Laskin and Pieter Abbeel and Aravind Srinivas and Igor Mordatch},
booktitle={Thirty-Fifth Conference on Neural Information Processing Systems},
year={2021},
url={https://openreview.net/forum?id=a7APmM4B9d}
}

@inproceedings{janner2021offline,
title={Offline Reinforcement Learning as One Big Sequence Modeling Problem},
author={Michael Janner and Qiyang Li and Sergey Levine},
booktitle={Thirty-Fifth Conference on Neural Information Processing Systems},
year={2021},
url={https://openreview.net/forum?id=wgeK563QgSw}
}

@article{nair2020awac,
  publtype={informal},
  author={Ashvin Nair and Murtaza Dalal and Abhishek Gupta and Sergey Levine},
  title={Accelerating Online Reinforcement Learning with Offline Datasets},
  year={2020},
  cdate={1577836800000},
  journal={CoRR},
  volume={abs/2006.09359},
  url={https://arxiv.org/abs/2006.09359}
}

@misc{kostrikov2021offline,
    title={Offline Reinforcement Learning with Implicit Q-Learning},
    author={Ilya Kostrikov and Ashvin Nair and Sergey Levine},
    year={2021},
    eprint={2110.06169},
    archivePrefix={arXiv},
    primaryClass={cs.LG}
}

@inproceedings{fujimoto2021minimalist,
title={A Minimalist Approach to Offline Reinforcement Learning},
author={Scott Fujimoto and Shixiang Gu},
booktitle={Thirty-Fifth Conference on Neural Information Processing Systems},
year={2021},
url={https://openreview.net/forum?id=Q32U7dzWXpc}
}

@article{vaswani2017attention,
  title={Attention is all you need},
  author={Vaswani, Ashish and Shazeer, Noam and Parmar, Niki and Uszkoreit, Jakob and Jones, Llion and Gomez, Aidan N and Kaiser, {\L}ukasz and Polosukhin, Illia},
  journal={Advances in neural information processing systems},
  volume={30},
  year={2017}
}

@article{jackson2024policy,
  title={Policy-Guided Diffusion},
  author={Jackson, Matthew Thomas and Matthews, Michael Tryfan and Lu, Cong and Ellis, Benjamin and Whiteson, Shimon and Foerster, Jakob},
  journal={arXiv preprint arXiv:2404.06356},
  year={2024}
}

@article{peng2019advantage,
  title={Advantage-weighted regression: Simple and scalable off-policy reinforcement learning},
  author={Peng, Xue Bin and Kumar, Aviral and Zhang, Grace and Levine, Sergey},
  journal={arXiv preprint arXiv:1910.00177},
  year={2019}
}

@article{wang2022diffusion,
  title={Diffusion policies as an expressive policy class for offline reinforcement learning},
  author={Wang, Zhendong and Hunt, Jonathan J and Zhou, Mingyuan},
  journal={arXiv preprint arXiv:2208.06193},
  year={2022}
}

@article{ding2023consistency,
  title={Consistency Models as a Rich and Efficient Policy Class for Reinforcement Learning},
  author={Ding, Zihan and Jin, Chi},
  journal={arXiv preprint arXiv:2309.16984},
  year={2023}
}

@article{song2020score,
  title={Score-based generative modeling through stochastic differential equations},
  author={Song, Yang and Sohl-Dickstein, Jascha and Kingma, Diederik P and Kumar, Abhishek and Ermon, Stefano and Poole, Ben},
  journal={arXiv preprint arXiv:2011.13456},
  year={2020}
}

@inproceedings{fujimoto2018addressing,
  title={Addressing function approximation error in actor-critic methods},
  author={Fujimoto, Scott and Hoof, Herke and Meger, David},
  booktitle={International conference on machine learning},
  pages={1587--1596},
  year={2018},
  organization={PMLR}
}

@article{kaiser2019model,
  title={Model-based reinforcement learning for atari},
  author={Kaiser, Lukasz and Babaeizadeh, Mohammad and Milos, Piotr and Osinski, Blazej and Campbell, Roy H and Czechowski, Konrad and Erhan, Dumitru and Finn, Chelsea and Kozakowski, Piotr and Levine, Sergey and others},
  journal={arXiv preprint arXiv:1903.00374},
  year={2019}
}

@article{fu2020d4rl,
  title={D4rl: Datasets for deep data-driven reinforcement learning},
  author={Fu, Justin and Kumar, Aviral and Nachum, Ofir and Tucker, George and Levine, Sergey},
  journal={arXiv preprint arXiv:2004.07219},
  year={2020}
}

@inproceedings{yamagata2023q,
  title={Q-learning decision transformer: Leveraging dynamic programming for conditional sequence modelling in offline rl},
  author={Yamagata, Taku and Khalil, Ahmed and Santos-Rodriguez, Raul},
  booktitle={International Conference on Machine Learning},
  pages={38989--39007},
  year={2023},
  organization={PMLR}
}

@article{kumar2020conservative,
  title={Conservative q-learning for offline reinforcement learning},
  author={Kumar, Aviral and Zhou, Aurick and Tucker, George and Levine, Sergey},
  journal={arXiv preprint arXiv:2006.04779},
  year={2020}
}

@article{emmons2021rvs,
  title={RvS: What is Essential for Offline RL via Supervised Learning?},
  author={Emmons, Scott and Eysenbach, Benjamin and Kostrikov, Ilya and Levine, Sergey},
  journal={arXiv preprint arXiv:2112.10751},
  year={2021}
}

@article{kumar2019stabilizing,
  title={Stabilizing off-policy q-learning via bootstrapping error reduction},
  author={Kumar, Aviral and Fu, Justin and Tucker, George and Levine, Sergey},
  journal={arXiv preprint arXiv:1906.00949},
  year={2019}
}

@inproceedings{sutton2000policy,
  title={Policy gradient methods for reinforcement learning with function approximation},
  author={Sutton, Richard S and McAllester, David A and Singh, Satinder P and Mansour, Yishay},
  booktitle={Advances in neural information processing systems},
  pages={1057--1063},
  year={2000}
}

@article{lillicrap2015continuous,
  title={Continuous control with deep reinforcement learning},
  author={Lillicrap, Timothy P and Hunt, Jonathan J and Pritzel, Alexander and Heess, Nicolas and Erez, Tom and Tassa, Yuval and Silver, David and Wierstra, Daan},
  journal={arXiv preprint arXiv:1509.02971},
  year={2015}
}

@article{schulman2017proximal,
  title={Proximal policy optimization algorithms},
  author={Schulman, John and Wolski, Filip and Dhariwal, Prafulla and Radford, Alec and Klimov, Oleg},
  journal={arXiv preprint arXiv:1707.06347},
  year={2017}
}

@inproceedings{schulman2015trust,
  title={Trust region policy optimization},
  author={Schulman, John and Levine, Sergey and Abbeel, Pieter and Jordan, Michael and Moritz, Philipp},
  booktitle={International conference on machine learning},
  pages={1889--1897},
  year={2015},
  organization={PMLR}
}

@article{meng2021offline,
  title={Offline Pre-trained Multi-Agent Decision Transformer: One Big Sequence Model Conquers All StarCraftII Tasks},
  author={Meng, Linghui and Wen, Muning and Yang, Yaodong and Le, Chenyang and Li, Xiyun and Zhang, Weinan and Wen, Ying and Zhang, Haifeng and Wang, Jun and Xu, Bo},
  journal={arXiv preprint arXiv:2112.02845},
  year={2021}
}

@inproceedings{zheng2022online,
  title={Online decision transformer},
  author={Zheng, Qinqing and Zhang, Amy and Grover, Aditya},
  booktitle={international conference on machine learning},
  pages={27042--27059},
  year={2022},
  organization={PMLR}
}

@inproceedings{fujimoto2019off,
  title={Off-policy deep reinforcement learning without exploration},
  author={Fujimoto, Scott and Meger, David and Precup, Doina},
  booktitle={International Conference on Machine Learning},
  pages={2052--2062},
  year={2019},
  organization={PMLR}
}

@article{wu2019behavior,
  title={Behavior regularized offline reinforcement learning},
  author={Wu, Yifan and Tucker, George and Nachum, Ofir},
  journal={arXiv preprint arXiv:1911.11361},
  year={2019}
}

@article{jaques2019way,
  title={Way off-policy batch deep reinforcement learning of implicit human preferences in dialog},
  author={Jaques, Natasha and Ghandeharioun, Asma and Shen, Judy Hanwen and Ferguson, Craig and Lapedriza, Agata and Jones, Noah and Gu, Shixiang and Picard, Rosalind},
  journal={arXiv preprint arXiv:1907.00456},
  year={2019}
}

@article{lee2022multi,
  title={Multi-Game Decision Transformers},
  author={Lee, Kuang-Huei and Nachum, Ofir and Yang, Mengjiao and Lee, Lisa and Freeman, Daniel and Xu, Winnie and Guadarrama, Sergio and Fischer, Ian and Jang, Eric and Michalewski, Henryk and others},
  journal={arXiv preprint arXiv:2205.15241},
  year={2022}
}

@article{hansen2023td,
  title={TD-MPC2: Scalable, Robust World Models for Continuous Control},
  author={Hansen, Nicklas and Su, Hao and Wang, Xiaolong},
  journal={arXiv preprint arXiv:2310.16828},
  year={2023}
}

@inproceedings{nichol2021improved,
  title={Improved denoising diffusion probabilistic models},
  author={Nichol, Alexander Quinn and Dhariwal, Prafulla},
  booktitle={International Conference on Machine Learning},
  pages={8162--8171},
  year={2021},
  organization={PMLR}
}

@article{janner2019trust,
  title={When to trust your model: Model-based policy optimization},
  author={Janner, Michael and Fu, Justin and Zhang, Marvin and Levine, Sergey},
  journal={Advances in neural information processing systems},
  volume={32},
  year={2019}
}

@article{yu2020mopo,
  title={Mopo: Model-based offline policy optimization},
  author={Yu, Tianhe and Thomas, Garrett and Yu, Lantao and Ermon, Stefano and Zou, James Y and Levine, Sergey and Finn, Chelsea and Ma, Tengyu},
  journal={Advances in Neural Information Processing Systems},
  volume={33},
  pages={14129--14142},
  year={2020}
}

@article{robine2023transformer,
  title={Transformer-based World Models Are Happy With 100k Interactions},
  author={Robine, Jan and H{\"o}ftmann, Marc and Uelwer, Tobias and Harmeling, Stefan},
  journal={arXiv preprint arXiv:2303.07109},
  year={2023}
}

@article{micheli2022transformers,
  title={Transformers are sample efficient world models},
  author={Micheli, Vincent and Alonso, Eloi and Fleuret, Fran{\c{c}}ois},
  journal={arXiv preprint arXiv:2209.00588},
  year={2022}
}

@article{schrittwieser2020mastering,
  title={Mastering atari, go, chess and shogi by planning with a learned model},
  author={Schrittwieser, Julian and Antonoglou, Ioannis and Hubert, Thomas and Simonyan, Karen and Sifre, Laurent and Schmitt, Simon and Guez, Arthur and Lockhart, Edward and Hassabis, Demis and Graepel, Thore and others},
  journal={Nature},
  volume={588},
  number={7839},
  pages={604--609},
  year={2020},
  publisher={Nature Publishing Group UK London}
}

@article{ye2021mastering,
  title={Mastering atari games with limited data},
  author={Ye, Weirui and Liu, Shaohuai and Kurutach, Thanard and Abbeel, Pieter and Gao, Yang},
  journal={Advances in Neural Information Processing Systems},
  volume={34},
  pages={25476--25488},
  year={2021}
}

@article{hafner2019dream,
  title={Dream to control: Learning behaviors by latent imagination},
  author={Hafner, Danijar and Lillicrap, Timothy and Ba, Jimmy and Norouzi, Mohammad},
  journal={arXiv preprint arXiv:1912.01603},
  year={2019}
}

@article{xiao2019learning,
  title={Learning to combat compounding-error in model-based reinforcement learning},
  author={Xiao, Chenjun and Wu, Yifan and Ma, Chen and Schuurmans, Dale and M{\"u}ller, Martin},
  journal={arXiv preprint arXiv:1912.11206},
  year={2019}
}

@article{sutton1991dyna,
  title={Dyna, an integrated architecture for learning, planning, and reacting},
  author={Sutton, Richard S},
  journal={ACM Sigart Bulletin},
  volume={2},
  number={4},
  pages={160--163},
  year={1991},
  publisher={ACM New York, NY, USA}
}

@article{lambert2022investigating,
  title={Investigating compounding prediction errors in learned dynamics models},
  author={Lambert, Nathan and Pister, Kristofer and Calandra, Roberto},
  journal={arXiv preprint arXiv:2203.09637},
  year={2022}
}

@article{hafner2020mastering,
  title={Mastering atari with discrete world models},
  author={Hafner, Danijar and Lillicrap, Timothy and Norouzi, Mohammad and Ba, Jimmy},
  journal={arXiv preprint arXiv:2010.02193},
  year={2020}
}

@article{ha2018world,
  title={World models},
  author={Ha, David and Schmidhuber, J{\"u}rgen},
  journal={arXiv preprint arXiv:1803.10122},
  year={2018}
}

@article{asadi2019combating,
  title={Combating the compounding-error problem with a multi-step model},
  author={Asadi, Kavosh and Misra, Dipendra and Kim, Seungchan and Littman, Michel L},
  journal={arXiv preprint arXiv:1905.13320},
  year={2019}
}

@article{kidambi2020morel,
  title={Morel: Model-based offline reinforcement learning},
  author={Kidambi, Rahul and Rajeswaran, Aravind and Netrapalli, Praneeth and Joachims, Thorsten},
  journal={Advances in neural information processing systems},
  volume={33},
  pages={21810--21823},
  year={2020}
}

@article{hansen2022temporal,
  title={Temporal difference learning for model predictive control},
  author={Hansen, Nicklas and Wang, Xiaolong and Su, Hao},
  journal={arXiv preprint arXiv:2203.04955},
  year={2022}
}

@article{hafner2023mastering,
  title={Mastering Diverse Domains through World Models},
  author={Hafner, Danijar and Pasukonis, Jurgis and Ba, Jimmy and Lillicrap, Timothy},
  journal={arXiv preprint arXiv:2301.04104},
  year={2023}
}

@article{chen2022transdreamer,
  title={Transdreamer: Reinforcement learning with transformer world models},
  author={Chen, Chang and Wu, Yi-Fu and Yoon, Jaesik and Ahn, Sungjin},
  journal={arXiv preprint arXiv:2202.09481},
  year={2022}
}

@article{kingma2021variational,
  title={Variational diffusion models},
  author={Kingma, Diederik and Salimans, Tim and Poole, Ben and Ho, Jonathan},
  journal={Advances in neural information processing systems},
  volume={34},
  pages={21696--21707},
  year={2021}
}

@inproceedings{zheng2023semi,
  title={Semi-supervised offline reinforcement learning with action-free trajectories},
  author={Zheng, Qinqing and Henaff, Mikael and Amos, Brandon and Grover, Aditya},
  booktitle={International Conference on Machine Learning},
  pages={42339--42362},
  year={2023},
  organization={PMLR}
}

@inproceedings{wu2018group,
  title={Group normalization},
  author={Wu, Yuxin and He, Kaiming},
  booktitle={Proceedings of the European conference on computer vision (ECCV)},
  pages={3--19},
  year={2018}
}

@article{mish2019self,
  title={A self regularized non-monotonic activation function [J]},
  author={Mish, Misra D},
  journal={arXiv preprint arXiv:1908.08681},
  year={2019}
}

@article{janner2020gamma,
  title={gamma-models: Generative temporal difference learning for infinite-horizon prediction},
  author={Janner, Michael and Mordatch, Igor and Levine, Sergey},
  journal={Advances in Neural Information Processing Systems},
  volume={33},
  pages={1724--1735},
  year={2020}
}

@article{zheng2023guided,
  title={Guided Flows for Generative Modeling and Decision Making},
  author={Zheng, Qinqing and Le, Matt and Shaul, Neta and Lipman, Yaron and Grover, Aditya and Chen, Ricky TQ},
  journal={arXiv preprint arXiv:2311.13443},
  year={2023}
}

@article{feinberg2018model,
  title={Model-based value estimation for efficient model-free reinforcement learning},
  author={Feinberg, Vladimir and Wan, Alvin and Stoica, Ion and Jordan, Michael I and Gonzalez, Joseph E and Levine, Sergey},
  journal={arXiv preprint arXiv:1803.00101},
  year={2018}
}

@inproceedings{silver2014deterministic,
  title={Deterministic policy gradient algorithms},
  author={Silver, David and Lever, Guy and Heess, Nicolas and Degris, Thomas and Wierstra, Daan and Riedmiller, Martin},
  booktitle={International conference on machine learning},
  pages={387--395},
  year={2014},
  organization={Pmlr}
}

@article{garg2023extreme,
  title={Extreme q-learning: Maxent RL without entropy},
  author={Garg, Divyansh and Hejna, Joey and Geist, Matthieu and Ermon, Stefano},
  journal={arXiv preprint arXiv:2301.02328},
  year={2023}
}

@article{deisenroth2013survey,
  title={A survey on policy search for robotics},
  author={Deisenroth, Marc Peter and Neumann, Gerhard and Peters, Jan and others},
  journal={Foundations and Trends{\textregistered} in Robotics},
  volume={2},
  number={1--2},
  pages={1--142},
  year={2013},
  publisher={Now Publishers, Inc.}
}

@article{dean2020sample,
  title={On the sample complexity of the linear quadratic regulator},
  author={Dean, Sarah and Mania, Horia and Matni, Nikolai and Recht, Benjamin and Tu, Stephen},
  journal={Foundations of Computational Mathematics},
  volume={20},
  number={4},
  pages={633--679},
  year={2020},
  publisher={Springer}
}

@inproceedings{hafner2019learning,
  title={Learning latent dynamics for planning from pixels},
  author={Hafner, Danijar and Lillicrap, Timothy and Fischer, Ian and Villegas, Ruben and Ha, David and Lee, Honglak and Davidson, James},
  booktitle={International conference on machine learning},
  pages={2555--2565},
  year={2019},
  organization={PMLR}
}

@inproceedings{thrun2014issues,
  title={Issues in using function approximation for reinforcement learning},
  author={Thrun, Sebastian and Schwartz, Anton},
  booktitle={Proceedings of the 1993 connectionist models summer school},
  pages={255--263},
  year={2014},
  organization={Psychology Press}
}

@article{williams2015model,
  title={Model predictive path integral control using covariance variable importance sampling},
  author={Williams, Grady and Aldrich, Andrew and Theodorou, Evangelos},
  journal={arXiv preprint arXiv:1509.01149},
  year={2015}
}

@inproceedings{nagabandi2018neural,
  title={Neural network dynamics for model-based deep reinforcement learning with model-free fine-tuning},
  author={Nagabandi, Anusha and Kahn, Gregory and Fearing, Ronald S and Levine, Sergey},
  booktitle={2018 IEEE international conference on robotics and automation (ICRA)},
  pages={7559--7566},
  year={2018},
  organization={IEEE}
}

@inproceedings{amos2021model,
  title={On the model-based stochastic value gradient for continuous reinforcement learning},
  author={Amos, Brandon and Stanton, Samuel and Yarats, Denis and Wilson, Andrew Gordon},
  booktitle={Learning for Dynamics and Control},
  pages={6--20},
  year={2021},
  organization={PMLR}
}

@incollection{sutton1990integrated,
  title={Integrated architectures for learning, planning, and reacting based on approximating dynamic programming},
  author={Sutton, Richard S},
  booktitle={Machine learning proceedings 1990},
  pages={216--224},
  year={1990},
  publisher={Elsevier}
}

@article{song2023consistency,
  title={Consistency models},
  author={Song, Yang and Dhariwal, Prafulla and Chen, Mark and Sutskever, Ilya},
  journal={arXiv preprint arXiv:2303.01469},
  year={2023}
}

@article{chi2023diffusion,
  title={Diffusion policy: Visuomotor policy learning via action diffusion},
  author={Chi, Cheng and Feng, Siyuan and Du, Yilun and Xu, Zhenjia and Cousineau, Eric and Burchfiel, Benjamin and Song, Shuran},
  journal={arXiv preprint arXiv:2303.04137},
  year={2023}
}

@article{hansen2023idql,
  title={Idql: Implicit q-learning as an actor-critic method with diffusion policies},
  author={Hansen-Estruch, Philippe and Kostrikov, Ilya and Janner, Michael and Kuba, Jakub Grudzien and Levine, Sergey},
  journal={arXiv preprint arXiv:2304.10573},
  year={2023}
}

@article{watkins1992q,
  title={Q-learning},
  author={Watkins, Christopher JCH and Dayan, Peter},
  journal={Machine learning},
  volume={8},
  pages={279--292},
  year={1992},
  publisher={Springer}
}

@article{ghasemipour2022so,
  title={Why so pessimistic? estimating uncertainties for offline rl through ensembles, and why their independence matters},
  author={Ghasemipour, Kamyar and Gu, Shixiang Shane and Nachum, Ofir},
  journal={Advances in Neural Information Processing Systems},
  volume={35},
  pages={18267--18281},
  year={2022}
}

@inproceedings{ronneberger2015u,
  title={U-net: Convolutional networks for biomedical image segmentation},
  author={Ronneberger, Olaf and Fischer, Philipp and Brox, Thomas},
  booktitle={Medical Image Computing and Computer-Assisted Intervention--MICCAI 2015: 18th International Conference, Munich, Germany, October 5-9, 2015, Proceedings, Part III 18},
  pages={234--241},
  year={2015},
  organization={Springer}
}

@article{nguyen2022conserweightive,
  title={ConserWeightive Behavioral Cloning for Reliable Offline Reinforcement Learning},
  author={Nguyen, Tung and Zheng, Qinqing and Grover, Aditya},
  journal={arXiv preprint arXiv:2210.05158},
  year={2022}
}

@misc{
alonso2024diffusion,
title={Diffusion World Models},
author={Eloi Alonso and Adam Jelley and Anssi Kanervisto and Tim Pearce},
year={2024},
url={https://openreview.net/forum?id=bAXmvOLtjA}
}

@article{yang2023learning,
  title={Learning interactive real-world simulators},
  author={Yang, Mengjiao and Du, Yilun and Ghasemipour, Kamyar and Tompson, Jonathan and Schuurmans, Dale and Abbeel, Pieter},
  journal={arXiv preprint arXiv:2310.06114},
  year={2023}
}

@article{lu2023synthetic,
  title={Synthetic experience replay},
  author={Lu, Cong and Ball, Philip J and Parker-Holder, Jack},
  journal={arXiv preprint arXiv:2303.06614},
  year={2023}
}

@article{rigter2023world,
  title={World Models via Policy-Guided Trajectory Diffusion},
  author={Rigter, Marc and Yamada, Jun and Posner, Ingmar},
  journal={arXiv preprint arXiv:2312.08533},
  year={2023}
}

@article{zhang2023learning,
  title={Learning unsupervised world models for autonomous driving via discrete diffusion},
  author={Zhang, Lunjun and Xiong, Yuwen and Yang, Ze and Casas, Sergio and Hu, Rui and Urtasun, Raquel},
  journal={arXiv preprint arXiv:2311.01017},
  year={2023}
}

@article{williams1992simple,
  title={Simple statistical gradient-following algorithms for connectionist reinforcement learning},
  author={Williams, Ronald J},
  journal={Machine learning},
  volume={8},
  pages={229--256},
  year={1992},
  publisher={Springer}
}

@article{huang2021towards,
  title={Towards general function approximation in zero-sum markov games},
  author={Huang, Baihe and Lee, Jason D and Wang, Zhaoran and Yang, Zhuoran},
  journal={arXiv preprint arXiv:2107.14702},
  year={2021}
}

@article{bai2022near,
  title={Near-Optimal Learning of Extensive-Form Games with Imperfect Information},
  author={Bai, Yu and Jin, Chi and Mei, Song and Yu, Tiancheng},
  journal={arXiv preprint arXiv:2202.01752},
  year={2022}
}

@article{kozuno2021model,
  title={Model-Free Learning for Two-Player Zero-Sum Partially Observable Markov Games with Perfect Recall},
  author={Kozuno, Tadashi and M{\'e}nard, Pierre and Munos, R{\'e}mi and Valko, Michal},
  journal={arXiv preprint arXiv:2106.06279},
  year={2021}
}

@article{jin2021power,
  title={The Power of Exploiter: Provable Multi-Agent RL in Large State Spaces},
  author={Jin, Chi and Liu, Qinghua and Yu, Tiancheng},
  journal={arXiv preprint arXiv:2106.03352},
  year={2021}
}

@article{bai2020near,
  title={Near-Optimal Reinforcement Learning with Self-Play},
  author={Bai, Yu and Jin, Chi and Yu, Tiancheng},
  journal={Advances in Neural Information Processing Systems},
  year={2020}
}

@inproceedings{bailey2018multiplicative,
  title={Multiplicative weights update in zero-sum games},
  author={Bailey, James P and Piliouras, Georgios},
  booktitle={Proceedings of the 2018 ACM Conference on Economics and Computation},
  pages={321--338},
  year={2018}
}

@article{brown1951iterative,
  title={Iterative solution of games by fictitious play},
  author={Brown, George W},
  journal={Activity analysis of production and allocation},
  volume={13},
  number={1},
  pages={374--376},
  year={1951},
  publisher={New York}
}

@article{hu2003nash,
  title={Nash Q-learning for general-sum stochastic games},
  author={Hu, Junling and Wellman, Michael P},
  journal={Journal of machine learning research},
  volume={4},
  number={Nov},
  pages={1039--1069},
  year={2003}
}

@inproceedings{badia2020agent57,
  title={Agent57: Outperforming the atari human benchmark},
  author={Badia, Adri{\`a} Puigdom{\`e}nech and Piot, Bilal and Kapturowski, Steven and Sprechmann, Pablo and Vitvitskyi, Alex and Guo, Zhaohan Daniel and Blundell, Charles},
  booktitle={International Conference on Machine Learning},
  pages={507--517},
  year={2020},
  organization={PMLR}
}

@article{jin2021v,
  title={V-Learning--A Simple, Efficient, Decentralized Algorithm for Multiagent RL},
  author={Jin, Chi and Liu, Qinghua and Wang, Yuanhao and Yu, Tiancheng},
  journal={arXiv preprint arXiv:2110.14555},
  year={2021}
}

@inproceedings{liu2021sharp,
  title={A sharp analysis of model-based reinforcement learning with self-play},
  author={Liu, Qinghua and Yu, Tiancheng and Bai, Yu and Jin, Chi},
  booktitle={International Conference on Machine Learning},
  pages={7001--7010},
  year={2021},
  organization={PMLR}
}

@inproceedings{farina2020stochastic,
  title={Stochastic regret minimization in extensive-form games},
  author={Farina, Gabriele and Kroer, Christian and Sandholm, Tuomas},
  booktitle={International Conference on Machine Learning},
  pages={3018--3028},
  year={2020},
  organization={PMLR}
}

@article{lanctot2009monte,
  title={Monte Carlo sampling for regret minimization in extensive games},
  author={Lanctot, Marc and Waugh, Kevin and Zinkevich, Martin and Bowling, Michael},
  journal={Advances in neural information processing systems},
  volume={22},
  year={2009}
}

@inproceedings{foerster2018counterfactual,
  title={Counterfactual multi-agent policy gradients},
  author={Foerster, Jakob and Farquhar, Gregory and Afouras, Triantafyllos and Nardelli, Nantas and Whiteson, Shimon},
  booktitle={Proceedings of the AAAI conference on artificial intelligence},
  volume={32},
  number={1},
  year={2018}
}

@article{mnih2013playing,
  title={Playing atari with deep reinforcement learning},
  author={Mnih, Volodymyr and Kavukcuoglu, Koray and Silver, David and Graves, Alex and Antonoglou, Ioannis and Wierstra, Daan and Riedmiller, Martin},
  journal={arXiv preprint arXiv:1312.5602},
  year={2013}
}

@article{sunehag2017value,
  title={Value-decomposition networks for cooperative multi-agent learning},
  author={Sunehag, Peter and Lever, Guy and Gruslys, Audrunas and Czarnecki, Wojciech Marian and Zambaldi, Vinicius and Jaderberg, Max and Lanctot, Marc and Sonnerat, Nicolas and Leibo, Joel Z and Tuyls, Karl and others},
  journal={arXiv preprint arXiv:1706.05296},
  year={2017}
}

@article{daskalakis2018last,
  title={Last-iterate convergence: Zero-sum games and constrained min-max optimization},
  author={Daskalakis, Constantinos and Panageas, Ioannis},
  journal={arXiv preprint arXiv:1807.04252},
  year={2018}
}

@article{mcaleer2021xdo,
  title={XDO: A double oracle algorithm for extensive-form games},
  author={McAleer, Stephen and Lanier, John and Baldi, Pierre and Fox, Roy},
  journal={arXiv preprint arXiv:2103.06426},
  year={2021}
}

@inproceedings{mcmahan2003planning,
  title={Planning in the presence of cost functions controlled by an adversary},
  author={McMahan, H Brendan and Gordon, Geoffrey J and Blum, Avrim},
  booktitle={Proceedings of the 20th International Conference on Machine Learning (ICML-03)},
  pages={536--543},
  year={2003}
}

@article{dinh2021online,
  title={Online Double Oracle},
  author={Dinh, Le Cong and Yang, Yaodong and Tian, Zheng and Nieves, Nicolas Perez and Slumbers, Oliver and Mguni, David Henry and Ammar, Haitham Bou and Wang, Jun},
  journal={arXiv preprint arXiv:2103.07780},
  year={2021}
}

@inproceedings{domahidi2013ecos,
  title={ECOS: An SOCP solver for embedded systems},
  author={Domahidi, Alexander and Chu, Eric and Boyd, Stephen},
  booktitle={2013 European Control Conference (ECC)},
  pages={3071--3076},
  year={2013},
  organization={IEEE}
}

@article{brown2019superhuman,
  title={Superhuman AI for multiplayer poker},
  author={Brown, Noam and Sandholm, Tuomas},
  journal={Science},
  volume={365},
  number={6456},
  pages={885--890},
  year={2019},
  publisher={American Association for the Advancement of Science}
}

@article{silver2016mastering,
  title={Mastering the game of Go with deep neural networks and tree search},
  author={Silver, David and Huang, Aja and Maddison, Chris J and Guez, Arthur and Sifre, Laurent and Van Den Driessche, George and Schrittwieser, Julian and Antonoglou, Ioannis and Panneershelvam, Veda and Lanctot, Marc and others},
  journal={nature},
  volume={529},
  number={7587},
  pages={484},
  year={2016},
  publisher={Nature Publishing Group}
}

@inproceedings{bai2020provable,
  title={Provable self-play algorithms for competitive reinforcement learning},
  author={Bai, Yu and Jin, Chi},
  booktitle={International Conference on Machine Learning},
  pages={551--560},
  year={2020},
  organization={PMLR}
}

@book{filar2012competitive,
  title={Competitive Markov decision processes},
  author={Filar, Jerzy and Vrieze, Koos},
  year={2012},
  publisher={Springer Science \& Business Media}
}

@incollection{littman1994markov,
  title={Markov games as a framework for multi-agent reinforcement learning},
  author={Littman, Michael L},
  booktitle={Machine learning proceedings 1994},
  pages={157--163},
  year={1994},
  publisher={Elsevier}
}

@article{shapley1953stochastic,
  title={Stochastic games},
  author={Shapley, Lloyd S},
  journal={Proceedings of the national academy of sciences},
  volume={39},
  number={10},
  pages={1095--1100},
  year={1953},
  publisher={National Acad Sciences}
}

@article{vinyals2019grandmaster,
  title={Grandmaster level in StarCraft II using multi-agent reinforcement learning},
  author={Vinyals, Oriol and Babuschkin, Igor and Czarnecki, Wojciech M and Mathieu, Micha{\"e}l and Dudzik, Andrew and Chung, Junyoung and Choi, David H and Powell, Richard and Ewalds, Timo and Georgiev, Petko and others},
  journal={Nature},
  volume={575},
  number={7782},
  pages={350--354},
  year={2019},
  publisher={Nature Publishing Group}
}

@inproceedings{xie2020learning,
  title={Learning zero-sum simultaneous-move markov games using function approximation and correlated equilibrium},
  author={Xie, Qiaomin and Chen, Yudong and Wang, Zhaoran and Yang, Zhuoran},
  booktitle={Conference on Learning Theory},
  pages={3674--3682},
  year={2020},
  organization={PMLR}
}

@inproceedings{heinrich2015fictitious,
  title={Fictitious self-play in extensive-form games},
  author={Heinrich, Johannes and Lanctot, Marc and Silver, David},
  booktitle={International conference on machine learning},
  pages={805--813},
  year={2015},
  organization={PMLR}
}

@misc{slimevolleygym,
  author = {David Ha},
  title = {Slime Volleyball Gym Environment},
  year = {2020},
  publisher = {GitHub},
  journal = {GitHub repository},
  howpublished = {\url{https://github.com/hardmaru/slimevolleygym}},
}

@misc{yu2021surprising,
      title={The Surprising Effectiveness of MAPPO in Cooperative, Multi-Agent Games}, 
      author={Chao Yu and Akash Velu and Eugene Vinitsky and Yu Wang and Alexandre Bayen and Yi Wu},
      year={2021},
      eprint={2103.01955},
      archivePrefix={arXiv},
      primaryClass={cs.LG}
}

@misc{heinrich2016deep,
      title={Deep Reinforcement Learning from Self-Play in Imperfect-Information Games}, 
      author={Johannes Heinrich and David Silver},
      year={2016},
      eprint={1603.01121},
      archivePrefix={arXiv},
      primaryClass={cs.LG}
}

@article{lanctot2019openspiel,
  title={OpenSpiel: A framework for reinforcement learning in games},
  author={Lanctot, Marc and Lockhart, Edward and Lespiau, Jean-Baptiste and Zambaldi, Vinicius and Upadhyay, Satyaki and P{\'e}rolat, Julien and Srinivasan, Sriram and Timbers, Finbarr and Tuyls, Karl and Omidshafiei, Shayegan and others},
  journal={arXiv preprint arXiv:1908.09453},
  year={2019}
}

@article{serrino2019finding,
  title={Finding friend and foe in multi-agent games},
  author={Serrino, Jack and Kleiman-Weiner, Max and Parkes, David C and Tenenbaum, Joshua B},
  journal={arXiv preprint arXiv:1906.02330},
  year={2019}
}

@article{zha2021douzero,
  title={DouZero: Mastering DouDizhu with Self-Play Deep Reinforcement Learning},
  author={Zha, Daochen and Xie, Jingru and Ma, Wenye and Zhang, Sheng and Lian, Xiangru and Hu, Xia and Liu, Ji},
  journal={arXiv preprint arXiv:2106.06135},
  year={2021}
}

@article{berner2019dota,
  title={Dota 2 with large scale deep reinforcement learning},
  author={Berner, Christopher and Brockman, Greg and Chan, Brooke and Cheung, Vicki and D{\k{e}}biak, Przemys{\l}aw and Dennison, Christy and Farhi, David and Fischer, Quirin and Hashme, Shariq and Hesse, Chris and others},
  journal={arXiv preprint arXiv:1912.06680},
  year={2019}
}

@article{arulkumaran2017deep,
  title={Deep reinforcement learning: A brief survey},
  author={Arulkumaran, Kai and Deisenroth, Marc Peter and Brundage, Miles and Bharath, Anil Anthony},
  journal={IEEE Signal Processing Magazine},
  volume={34},
  number={6},
  pages={26--38},
  year={2017},
  publisher={IEEE}
}

@book{dong2020deep,
  title={Deep Reinforcement Learning},
  author={Dong, Hao and Dong, Hao and Ding, Zihan and Zhang, Shanghang and Chang},
  year={2020},
  publisher={Springer}
}

@book{sutton2018reinforcement,
  title={Reinforcement learning: An introduction},
  author={Sutton, Richard S and Barto, Andrew G},
  year={2018},
  publisher={MIT press}
}

@article{vinyals2017starcraft,
  title={StarCraft II: A New Challenge for Reinforcement Learning},
  author={Vinyals, Oriol and Ewalds, Timo and Bartunov, Sergey and Georgiev, Petko and Vezhnevets, Alexander Sasha and Yeo, Michelle and Makhzani, Alireza and K{\"u}ttler, Heinrich and Agapiou, John and Schrittwieser, Julian and others},
  year={2017}
}

@misc{samvelyan2019starcraft,
      title={The StarCraft Multi-Agent Challenge}, 
      author={Mikayel Samvelyan and Tabish Rashid and Christian Schroeder de Witt and Gregory Farquhar and Nantas Nardelli and Tim G. J. Rudner and Chia-Man Hung and Philip H. S. Torr and Jakob Foerster and Shimon Whiteson},
      year={2019},
      eprint={1902.04043},
      archivePrefix={arXiv},
      primaryClass={cs.LG}
}

@article{nash1950equilibrium,
  title={Equilibrium points in n-person games},
  author={Nash, John F and others},
  journal={Proceedings of the national academy of sciences},
  volume={36},
  number={1},
  pages={48--49},
  year={1950},
  publisher={USA}
}

@article{Bellemare_2013,
   title={The Arcade Learning Environment: An Evaluation Platform for General Agents},
   volume={47},
   ISSN={1076-9757},
   url={http://dx.doi.org/10.1613/jair.3912},
   DOI={10.1613/jair.3912},
   journal={Journal of Artificial Intelligence Research},
   publisher={AI Access Foundation},
   author={Bellemare, M. G. and Naddaf, Y. and Veness, J. and Bowling, M.},
   year={2013},
   month={Jun},
   pages={253–279}
}

@misc{pettingzoo,
      title={PettingZoo: Gym for Multi-Agent Reinforcement Learning}, 
      author={Justin K. Terry and Benjamin Black and Mario Jayakumar and Ananth Hari and Ryan Sullivan and Luis Santos and Clemens Dieffendahl and Niall L. Williams and Yashas Lokesh and Caroline Horsch and Praveen Ravi},
      year={2021},
      eprint={2009.14471},
      archivePrefix={arXiv},
      primaryClass={cs.LG}
}

@misc{wang2019benchmarking,
      title={Benchmarking Model-Based Reinforcement Learning}, 
      author={Tingwu Wang and Xuchan Bao and Ignasi Clavera and Jerrick Hoang and Yeming Wen and Eric Langlois and Shunshi Zhang and Guodong Zhang and Pieter Abbeel and Jimmy Ba},
      year={2019},
      eprint={1907.02057},
      archivePrefix={arXiv},
      primaryClass={cs.LG}
}

@inproceedings{khan2022darefightingice,
  title={DareFightingICE Competition: A Fighting Game Sound Design and AI Competition},
  author={Khan, Ibrahim and Van Nguyen, Thai and Dai, Xincheng and Thawonmas, Ruck},
  booktitle={2022 IEEE Conference on Games (CoG)},
  pages={478--485},
  year={2022},
  organization={IEEE}
}

@article{yao2022learning,
  title={Learning distributed and fair policies for network load balancing as markov potential game},
  author={Yao, Zhiyuan and Ding, Zihan},
  journal={Advances in Neural Information Processing Systems},
  volume={35},
  pages={28815--28828},
  year={2022}
}

@article{li2024fightladder,
  title={FightLadder: A benchmark for competitive multi-agent reinforcement learning},
  author={Li, Wenzhe and Ding, Zihan and Karten, Seth and Jin, Chi},
  journal={arXiv preprint arXiv:2406.02081},
  year={2024}
}

@article{andrychowicz2020matters,
  title={What matters in on-policy reinforcement learning? a large-scale empirical study},
  author={Andrychowicz, Marcin and Raichuk, Anton and Sta{\'n}czyk, Piotr and Orsini, Manu and Girgin, Sertan and Marinier, Raphael and Hussenot, L{\'e}onard and Geist, Matthieu and Pietquin, Olivier and Michalski, Marcin and others},
  journal={arXiv preprint arXiv:2006.05990},
  year={2020}
}

@article{ouyang2022training,
  title={Training language models to follow instructions with human feedback},
  author={Ouyang, Long and Wu, Jeffrey and Jiang, Xu and Almeida, Diogo and Wainwright, Carroll and Mishkin, Pamela and Zhang, Chong and Agarwal, Sandhini and Slama, Katarina and Ray, Alex and others},
  journal={Advances in neural information processing systems},
  volume={35},
  pages={27730--27744},
  year={2022}
}

@article{andrychowicz2020learning,
  title={Learning dexterous in-hand manipulation},
  author={Andrychowicz, OpenAI: Marcin and Baker, Bowen and Chociej, Maciek and Jozefowicz, Rafal and McGrew, Bob and Pachocki, Jakub and Petron, Arthur and Plappert, Matthias and Powell, Glenn and Ray, Alex and others},
  journal={The International Journal of Robotics Research},
  volume={39},
  number={1},
  pages={3--20},
  year={2020},
  publisher={SAGE Publications Sage UK: London, England}
}

@article{brohan2022rt,
  title={Rt-1: Robotics transformer for real-world control at scale},
  author={Brohan, Anthony and Brown, Noah and Carbajal, Justice and Chebotar, Yevgen and Dabis, Joseph and Finn, Chelsea and Gopalakrishnan, Keerthana and Hausman, Karol and Herzog, Alex and Hsu, Jasmine and others},
  journal={arXiv preprint arXiv:2212.06817},
  year={2022}
}

@article{moravvcik2017deepstack,
  title={Deepstack: Expert-level artificial intelligence in heads-up no-limit poker},
  author={Morav{\v{c}}{\'\i}k, Matej and Schmid, Martin and Burch, Neil and Lis{\`y}, Viliam and Morrill, Dustin and Bard, Nolan and Davis, Trevor and Waugh, Kevin and Johanson, Michael and Bowling, Michael},
  journal={Science},
  volume={356},
  number={6337},
  pages={508--513},
  year={2017},
  publisher={American Association for the Advancement of Science}
}

@article{li2020suphx,
  title={Suphx: Mastering mahjong with deep reinforcement learning},
  author={Li, Junjie and Koyamada, Sotetsu and Ye, Qiwei and Liu, Guoqing and Wang, Chao and Yang, Ruihan and Zhao, Li and Qin, Tao and Liu, Tie-Yan and Hon, Hsiao-Wuen},
  journal={arXiv preprint arXiv:2003.13590},
  year={2020}
}

@book{dresher2016advances,
  title={Advances in Game Theory.(AM-52), Volume 52},
  author={Dresher, Melvin and Shapley, Lloyd S and Tucker, Albert William},
  volume={52},
  year={2016},
  publisher={Princeton University Press}
}

@article{nichol2018retro,
  title={Gotta Learn Fast: A New Benchmark for Generalization in RL},
  author={Nichol, Alex and Pfau, Vicki and Hesse, Christopher and Klimov, Oleg and Schulman, John},
  journal={arXiv preprint arXiv:1804.03720},
  year={2018}
}

@article{murphy2013hacking,
  title={Hacking public memory: Understanding the multiple arcade machine emulator},
  author={Murphy, David},
  journal={Games and Culture},
  volume={8},
  number={1},
  pages={43--53},
  year={2013},
  publisher={Sage Publications Sage CA: Los Angeles, CA}
}

@article{stable-baselines3,
  author  = {Antonin Raffin and Ashley Hill and Adam Gleave and Anssi Kanervisto and Maximilian Ernestus and Noah Dormann},
  title   = {Stable-Baselines3: Reliable Reinforcement Learning Implementations},
  journal = {Journal of Machine Learning Research},
  year    = {2021},
  volume  = {22},
  number  = {268},
  pages   = {1-8},
  url     = {http://jmlr.org/papers/v22/20-1364.html}
}

@article{daskalakis2020independent,
  title={Independent policy gradient methods for competitive reinforcement learning},
  author={Daskalakis, Constantinos and Foster, Dylan J and Golowich, Noah},
  journal={Advances in neural information processing systems},
  volume={33},
  pages={5527--5540},
  year={2020}
}

@article{lanctot2017unified,
  title={A unified game-theoretic approach to multiagent reinforcement learning},
  author={Lanctot, Marc and Zambaldi, Vinicius and Gruslys, Audrunas and Lazaridou, Angeliki and Tuyls, Karl and P{\'e}rolat, Julien and Silver, David and Graepel, Thore},
  journal={Advances in neural information processing systems},
  volume={30},
  year={2017}
}

@article{schaul2015prioritized,
  title={Prioritized experience replay},
  author={Schaul, Tom and Quan, John and Antonoglou, Ioannis and Silver, David},
  journal={arXiv preprint arXiv:1511.05952},
  year={2015}
}

@article{tesauro1995temporal,
  title={Temporal difference learning and TD-Gammon},
  author={Tesauro, Gerald and others},
  journal={Communications of the ACM},
  volume={38},
  number={3},
  pages={58--68},
  year={1995}
}

@article{go2023phase,
  title={A Phase-Change Memristive Reinforcement Learning for Rapidly Outperforming Champion Street-Fighter Players},
  author={Go, Shao-Xiang and Jiang, Yu and Loke, Desmond K},
  journal={Advanced Intelligent Systems},
  volume={5},
  number={11},
  pages={2300335},
  year={2023},
  publisher={Wiley Online Library}
}

@article{palmas2022diambra,
  title={DIAMBRA Arena: a New Reinforcement Learning Platform for Research and Experimentation},
  author={Palmas, Alessandro},
  journal={arXiv preprint arXiv:2210.10595},
  year={2022}
}

@article{brown2018superhuman,
  title={Superhuman AI for heads-up no-limit poker: Libratus beats top professionals},
  author={Brown, Noam and Sandholm, Tuomas},
  journal={Science},
  volume={359},
  number={6374},
  pages={418--424},
  year={2018},
  publisher={American Association for the Advancement of Science}
}

@article{resnick2018pommerman,
  title={Pommerman: A multi-agent playground},
  author={Resnick, Cinjon and Eldridge, Wes and Ha, David and Britz, Denny and Foerster, Jakob and Togelius, Julian and Cho, Kyunghyun and Bruna, Joan},
  journal={arXiv preprint arXiv:1809.07124},
  year={2018}
}

@article{pan2022mate,
  title={Mate: Benchmarking multi-agent reinforcement learning in distributed target coverage control},
  author={Pan, Xuehai and Liu, Mickel and Zhong, Fangwei and Yang, Yaodong and Zhu, Song-Chun and Wang, Yizhou},
  journal={Advances in Neural Information Processing Systems},
  volume={35},
  pages={27862--27879},
  year={2022}
}

@inproceedings{song2020arena,
  title={Arena: A general evaluation platform and building toolkit for multi-agent intelligence},
  author={Song, Yuhang and Wojcicki, Andrzej and Lukasiewicz, Thomas and Wang, Jianyi and Aryan, Abi and Xu, Zhenghua and Xu, Mai and Ding, Zihan and Wu, Lianlong},
  booktitle={Proceedings of the AAAI conference on artificial intelligence},
  volume={34},
  pages={7253--7260},
  year={2020}
}

@article{bard2020hanabi,
  title={The hanabi challenge: A new frontier for ai research},
  author={Bard, Nolan and Foerster, Jakob N and Chandar, Sarath and Burch, Neil and Lanctot, Marc and Song, H Francis and Parisotto, Emilio and Dumoulin, Vincent and Moitra, Subhodeep and Hughes, Edward and others},
  journal={Artificial Intelligence},
  volume={280},
  pages={103216},
  year={2020},
  publisher={Elsevier}
}

@inproceedings{kurach2020google,
  title={Google research football: A novel reinforcement learning environment},
  author={Kurach, Karol and Raichuk, Anton and Sta{\'n}czyk, Piotr and Zaj{\k{a}}c, Micha{\l} and Bachem, Olivier and Espeholt, Lasse and Riquelme, Carlos and Vincent, Damien and Michalski, Marcin and Bousquet, Olivier and others},
  booktitle={Proceedings of the AAAI conference on artificial intelligence},
  volume={34},
  pages={4501--4510},
  year={2020}
}

@article{peng2021facmac,
  title={Facmac: Factored multi-agent centralised policy gradients},
  author={Peng, Bei and Rashid, Tabish and Schroeder de Witt, Christian and Kamienny, Pierre-Alexandre and Torr, Philip and B{\"o}hmer, Wendelin and Whiteson, Shimon},
  journal={Advances in Neural Information Processing Systems},
  volume={34},
  pages={12208--12221},
  year={2021}
}

@article{mohanty2020flatland,
  title={Flatland-rl: Multi-agent reinforcement learning on trains},
  author={Mohanty, Sharada and Nygren, Erik and Laurent, Florian and Schneider, Manuel and Scheller, Christian and Bhattacharya, Nilabha and Watson, Jeremy and Egli, Adrian and Eichenberger, Christian and Baumberger, Christian and others},
  journal={arXiv preprint arXiv:2012.05893},
  year={2020}
}

@inproceedings{mordatch2018emergence,
  title={Emergence of grounded compositional language in multi-agent populations},
  author={Mordatch, Igor and Abbeel, Pieter},
  booktitle={Proceedings of the AAAI conference on artificial intelligence},
  volume={32},
  year={2018}
}

@inproceedings{zheng2018magent,
  title={Magent: A many-agent reinforcement learning platform for artificial collective intelligence},
  author={Zheng, Lianmin and Yang, Jiacheng and Cai, Han and Zhou, Ming and Zhang, Weinan and Wang, Jun and Yu, Yong},
  booktitle={Proceedings of the AAAI conference on artificial intelligence},
  volume={32},
  year={2018}
}

@article{suarez2021neural,
  title={The neural mmo platform for massively multiagent research},
  author={Suarez, Joseph and Du, Yilun and Zhu, Clare and Mordatch, Igor and Isola, Phillip},
  journal={arXiv preprint arXiv:2110.07594},
  year={2021}
}

@article{zhou2020smarts,
  title={Smarts: Scalable multi-agent reinforcement learning training school for autonomous driving},
  author={Zhou, Ming and Luo, Jun and Villella, Julian and Yang, Yaodong and Rusu, David and Miao, Jiayu and Zhang, Weinan and Alban, Montgomery and Fadakar, Iman and Chen, Zheng and others},
  journal={arXiv preprint arXiv:2010.09776},
  year={2020}
}

@article{sukhbaatar2016learning,
  title={Learning multiagent communication with backpropagation},
  author={Sukhbaatar, Sainbayar and Fergus, Rob and others},
  journal={Advances in neural information processing systems},
  volume={29},
  year={2016}
}

@inproceedings{zhang2019cityflow,
  title={Cityflow: A multi-agent reinforcement learning environment for large scale city traffic scenario},
  author={Zhang, Huichu and Feng, Siyuan and Liu, Chang and Ding, Yaoyao and Zhu, Yichen and Zhou, Zihan and Zhang, Weinan and Yu, Yong and Jin, Haiming and Li, Zhenhui},
  booktitle={The world wide web conference},
  pages={3620--3624},
  year={2019}
}

@article{hu2023marllib,
  title={MARLlib: A Scalable and Efficient Multi-agent Reinforcement Learning Library},
  author={Hu, Siyi and Zhong, Yifan and Gao, Minquan and Wang, Weixun and Dong, Hao and Liang, Xiaodan and Li, Zhihui and Chang, Xiaojun and Yang, Yaodong},
  journal={Journal of Machine Learning Research},
  volume={24},
  number={315},
  pages={1--23},
  year={2023}
}

@article{beattie2020deepmind,
  title={Deepmind lab2d},
  author={Beattie, Charles and K{\"o}ppe, Thomas and Du{\'e}{\~n}ez-Guzm{\'a}n, Edgar A and Leibo, Joel Z},
  journal={arXiv preprint arXiv:2011.07027},
  year={2020}
}

@article{brockman2016openai,
  title={Openai gym},
  author={Brockman, Greg and Cheung, Vicki and Pettersson, Ludwig and Schneider, Jonas and Schulman, John and Tang, Jie and Zaremba, Wojciech},
  journal={arXiv preprint arXiv:1606.01540},
  year={2016}
}

@article{papoudakis2020benchmarking,
  title={Benchmarking multi-agent deep reinforcement learning algorithms in cooperative tasks},
  author={Papoudakis, Georgios and Christianos, Filippos and Sch{\"a}fer, Lukas and Albrecht, Stefano V},
  journal={arXiv preprint arXiv:2006.07869},
  year={2020}
}

@inproceedings{sarkar2022pantheonrl,
  title={Pantheonrl: A marl library for dynamic training interactions},
  author={Sarkar, Bidipta and Talati, Aditi and Shih, Andy and Sadigh, Dorsa},
  booktitle={Proceedings of the AAAI Conference on Artificial Intelligence},
  volume={36},
  pages={13221--13223},
  year={2022}
}

@article{hu2021rethinking,
  title={Rethinking the implementation tricks and monotonicity constraint in cooperative multi-agent reinforcement learning},
  author={Hu, Jian and Jiang, Siyang and Harding, Seth Austin and Wu, Haibin and Liao, Shih-wei},
  journal={arXiv preprint arXiv:2102.03479},
  year={2021}
}

@article{rashid2020monotonic,
  title={Monotonic value function factorisation for deep multi-agent reinforcement learning},
  author={Rashid, Tabish and Samvelyan, Mikayel and De Witt, Christian Schroeder and Farquhar, Gregory and Foerster, Jakob and Whiteson, Shimon},
  journal={The Journal of Machine Learning Research},
  volume={21},
  number={1},
  pages={7234--7284},
  year={2020},
  publisher={JMLRORG}
}

@article{yu2022surprising,
  title={The surprising effectiveness of ppo in cooperative multi-agent games},
  author={Yu, Chao and Velu, Akash and Vinitsky, Eugene and Gao, Jiaxuan and Wang, Yu and Bayen, Alexandre and Wu, Yi},
  journal={Advances in Neural Information Processing Systems},
  volume={35},
  pages={24611--24624},
  year={2022}
}

@article{zhou2023malib,
  title={MALib: A parallel framework for population-based multi-agent reinforcement learning},
  author={Zhou, Ming and Wan, Ziyu and Wang, Hanjing and Wen, Muning and Wu, Runzhe and Wen, Ying and Yang, Yaodong and Yu, Yong and Wang, Jun and Zhang, Weinan},
  journal={Journal of Machine Learning Research},
  volume={24},
  number={150},
  pages={1--12},
  year={2023}
}

@article{ding2022deep,
  title={A Deep Reinforcement Learning Approach for Finding Non-Exploitable Strategies in Two-Player Atari Games},
  author={Ding, Zihan and Su, Dijia and Liu, Qinghua and Jin, Chi},
  journal={arXiv preprint arXiv:2207.08894},
  year={2022}
}

@article{silver2018general,
  title={A general reinforcement learning algorithm that masters chess, shogi, and Go through self-play},
  author={Silver, David and Hubert, Thomas and Schrittwieser, Julian and Antonoglou, Ioannis and Lai, Matthew and Guez, Arthur and Lanctot, Marc and Sifre, Laurent and Kumaran, Dharshan and Graepel, Thore and others},
  journal={Science},
  volume={362},
  number={6419},
  pages={1140--1144},
  year={2018},
  publisher={American Association for the Advancement of Science}
}

@inproceedings{tan1993multi,
  title={Multi-agent reinforcement learning: Independent vs. cooperative agents},
  author={Tan, Ming},
  booktitle={Proceedings of the tenth international conference on machine learning},
  pages={330--337},
  year={1993}
}

@article{elo1978rating,
  title={The rating of chessplayers: Past and present},
  author={Elo, Arpad E and Sloan, Sam},
  journal={(No Title)},
  year={1978}
}

@article{desmouceaux20186lb,
  title={6lb: Scalable and application-aware load balancing with segment routing},
  author={Desmouceaux, Yoann and Pfister, Pierre and Tollet, J{\'e}r{\^o}me and Townsley, Mark and Clausen, Thomas},
  journal={IEEE/ACM Transactions on Networking},
  volume={26},
  number={2},
  pages={819--834},
  year={2018},
  publisher={IEEE}
}

@article{sivakumar2019mvfst,
  title={Mvfst-rl: An asynchronous rl framework for congestion control with delayed actions},
  author={Sivakumar, Viswanath and Rockt{\"a}schel, Tim and Miller, Alexander H and K{\"u}ttler, Heinrich and Nardelli, Nantas and Rabbat, Mike and Pineau, Joelle and Riedel, Sebastian},
  journal={arXiv preprint arXiv:1910.04054},
  year={2019}
}

@INPROCEEDINGS{10377444,
  author={Esser, Patrick and Chiu, Johnathan and Atighehchian, Parmida and Granskog, Jonathan and Germanidis, Anastasis},
  booktitle={2023 IEEE/CVF International Conference on Computer Vision (ICCV)}, 
  title={Structure and Content-Guided Video Synthesis with Diffusion Models}, 
  year={2023},
  volume={},
  number={},
  pages={7312-7322},
  doi={10.1109/ICCV51070.2023.00675}}

@article{jiang2024loopy,
  title={Loopy: Taming audio-driven portrait avatar with long-term motion dependency},
  author={Jiang, Jianwen and Liang, Chao and Yang, Jiaqi and Lin, Gaojie and Zhong, Tianyun and Zheng, Yanbo},
  journal={arXiv preprint arXiv:2409.02634},
  year={2024}
}

@article{reuss2023goal,
  title={Goal-Conditioned Imitation Learning using Score-based Diffusion Policies},
  author={Reuss, Moritz and Li, Maximilian and Jia, Xiaogang and Lioutikov, Rudolf},
  journal={arXiv preprint arXiv:2304.02532},
  year={2023}
}

@article{pearce2023imitating,
  title={Imitating human behaviour with diffusion models},
  author={Pearce, Tim and Rashid, Tabish and Kanervisto, Anssi and Bignell, Dave and Sun, Mingfei and Georgescu, Raluca and Macua, Sergio Valcarcel and Tan, Shan Zheng and Momennejad, Ida and Hofmann, Katja and others},
  journal={arXiv preprint arXiv:2301.10677},
  year={2023}
}

@article{zhu2023guiding,
  title={Guiding online reinforcement learning with action-free offline pretraining},
  author={Zhu, Deyao and Wang, Yuhui and Schmidhuber, J{\"u}rgen and Elhoseiny, Mohamed},
  journal={arXiv preprint arXiv:2301.12876},
  year={2023}
}

@article{marden2009cooperative,
  title={Cooperative control and potential games},
  author={Marden, Jason R and Arslan, G{\"u}rdal and Shamma, Jeff S},
  journal={IEEE Transactions on Systems, Man, and Cybernetics, Part B (Cybernetics)},
  volume={39},
  number={6},
  pages={1393--1407},
  year={2009},
  publisher={IEEE}
}

@misc{ali2019reinforcement,
  title={Reinforcement Learning Algorithm for Load Balancing in Self-Organizing Networks},
  author={Ali, Mohammed Shabbir and Coucheney, Pierre and Coupechoux, Marceau},
  year={2019},
  publisher={Wiley}
}

@article{macua2018learning,
  title={Learning parametric closed-loop policies for markov potential games},
  author={Macua, Sergio Valcarcel and Zazo, Javier and Zazo, Santiago},
  journal={arXiv preprint arXiv:1802.00899},
  year={2018}
}

@inproceedings{mguni2021learning,
  title={Learning in nonzero-sum stochastic games with potentials},
  author={Mguni, David H and Wu, Yutong and Du, Yali and Yang, Yaodong and Wang, Ziyi and Li, Minne and Wen, Ying and Jennings, Joel and Wang, Jun},
  booktitle={International Conference on Machine Learning},
  pages={7688--7699},
  year={2021},
  organization={PMLR}
}

@article{de2020independent,
  title={Is independent learning all you need in the starcraft multi-agent challenge?},
  author={de Witt, Christian Schroeder and Gupta, Tarun and Makoviichuk, Denys and Makoviychuk, Viktor and Torr, Philip HS and Sun, Mingfei and Whiteson, Shimon},
  journal={arXiv preprint arXiv:2011.09533},
  year={2020}
}

@article{li2020multi,
  title={Multi-agent trust region policy optimization},
  author={Li, Hepeng and He, Haibo},
  journal={arXiv preprint arXiv:2010.07916},
  year={2020}
}

@book{fudenberg1998theory,
  title={The theory of learning in games},
  author={Fudenberg, Drew and Drew, Fudenberg and Levine, David K and Levine, David K},
  volume={2},
  year={1998},
  publisher={MIT press}
}

@article{yang2020overview,
  title={An overview of multi-agent reinforcement learning from game theoretical perspective},
  author={Yang, Yaodong and Wang, Jun},
  journal={arXiv preprint arXiv:2011.00583},
  year={2020}
}

@inproceedings{zimmer2021learning,
  title={Learning fair policies in decentralized cooperative multi-agent reinforcement learning},
  author={Zimmer, Matthieu and Glanois, Claire and Siddique, Umer and Weng, Paul},
  booktitle={International Conference on Machine Learning},
  pages={12967--12978},
  year={2021},
  organization={PMLR}
}

@article{jiang2019learning,
  title={Learning fairness in multi-agent systems},
  author={Jiang, Jiechuan and Lu, Zongqing},
  journal={Advances in Neural Information Processing Systems},
  volume={32},
  year={2019}
}

@article{leonardos2021global,
  title={Global convergence of multi-agent policy gradient in markov potential games},
  author={Leonardos, Stefanos and Overman, Will and Panageas, Ioannis and Piliouras, Georgios},
  journal={arXiv preprint arXiv:2106.01969},
  year={2021}
}

@inproceedings{fox2022independent,
  title={Independent natural policy gradient always converges in Markov potential games},
  author={Fox, Roy and Mcaleer, Stephen M and Overman, Will and Panageas, Ioannis},
  booktitle={International Conference on Artificial Intelligence and Statistics},
  pages={4414--4425},
  year={2022},
  organization={PMLR}
}

@article{candogan2011flows,
  title={Flows and decompositions of games: Harmonic and potential games},
  author={Candogan, Ozan and Menache, Ishai and Ozdaglar, Asuman and Parrilo, Pablo A},
  journal={Mathematics of Operations Research},
  volume={36},
  number={3},
  pages={474--503},
  year={2011},
  publisher={INFORMS}
}

@article{sandholm2001potential,
  title={Potential games with continuous player sets},
  author={Sandholm, William H},
  journal={Journal of Economic theory},
  volume={97},
  number={1},
  pages={81--108},
  year={2001},
  publisher={Elsevier}
}

@article{monderer1996potential,
  title={Potential games},
  author={Monderer, Dov and Shapley, Lloyd S},
  journal={Games and economic behavior},
  volume={14},
  number={1},
  pages={124--143},
  year={1996},
  publisher={Elsevier}
}

@article{yao2022reinforced,
  title={Reinforced Cooperative Load Balancing in Data Center},
  author={Yao, Zhiyuan and Ding, Zihan and Clausen, Thomas},
  journal={arXiv preprint arXiv:2201.11727},
  year={2022}
}

@article{concury2020,
  title  = {Concury: A Fast and Light-weight Software Cloud Load Balancer},
  author = {Shi, Shouqian and Yu, Ye and Xie, Minghao and Li, Xin and Li, Xiaozhou and Zhang, Ying and Qian, Chen},
  year   = {2020},
  pages  = {14}
}

@article{prism2020,
  title     = {Lb scalability: Achieving the right balance between being stateful and stateless},
  author    = {Cohen, Reuven and Kadosh, Matty and Lo, Alan and Sayah, Qasem},
  journal   = {IEEE/ACM Transactions on Networking},
  year      = {2021},
  publisher = {IEEE}
}

@inproceedings{facebook-dc-traffic,
  place        = {New York, NY, USA},
  series       = {SIGCOMM ’15},
  title        = {Inside the Social Network’s (Datacenter) Network},
  isbn         = {978-1-4503-3542-3},
  url          = {http://doi.acm.org/10.1145/2785956.2787472},
  doi          = {10.1145/2785956.2787472},
  abstractnote = {Large cloud service providers have invested in increasingly larger datacenters to house the computing infrastructure required to support their services. Accordingly, researchers and industry practitioners alike have focused a great deal of effort designing network fabrics to efficiently interconnect and manage the traffic within these datacenters in performant yet efficient fashions. Unfortunately, datacenter operators are generally reticent to share the actual requirements of their applications, making it challenging to evaluate the practicality of any particular design. Moreover, the limited large-scale workload information available in the literature has, for better or worse, heretofore largely been provided by a single datacenter operator whose use cases may not be widespread. In this work, we report upon the network traffic observed in some of Facebook’s datacenters. While Facebook operates a number of traditional datacenter services like Hadoop, its core Web service and supporting cache infrastructure exhibit a number of behaviors that contrast with those reported in the literature. We report on the contrasting locality, stability, and predictability of network traffic in Facebook’s datacenters, and comment on their implications for network architecture, traffic engineering, and switch design.},
  note         = {event-place: London, United Kingdom},
  booktitle    = {Proceedings of the 2015 ACM Conference on Special Interest Group on Data Communication},
  publisher    = {ACM},
  author       = {Roy, Arjun and Zeng, Hongyi and Bagga, Jasmeet and Porter, George and Snoeren, Alex C.},
  year         = {2015},
  pages        = {123–137},
  collection   = {SIGCOMM ’15}
}

\end{document}